# یادگیری ماشین

## و علم داده

### مبانی، مفاهیم، الگوریتم‌ها و ابزارها

**تالیف و گردآوری: میلاد وزان**

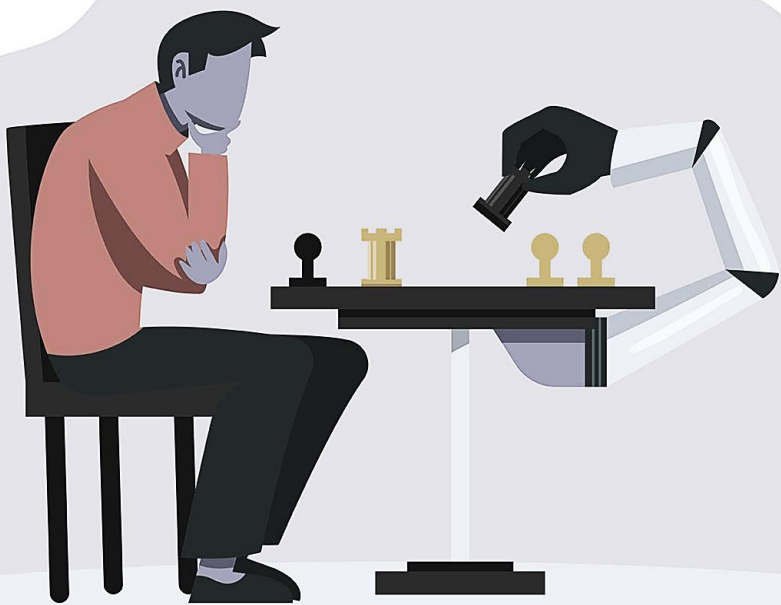

به نام خدا

# یادگیری ماشین و علم داده

## مبــانی، مفاهیم، الگوریتم‌ها و ابزارهــا

**تالیف و گردآوری: میلاد وزان**

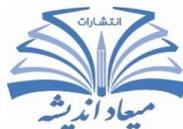

Miaadpub.ir

# پیش‌گفتار

امروزه اگر برجسته‌ترین رهبران تجارت جهانی را گردهم بیاورید و از آنان بخواهید که بزرگ‌ترین تفاوت بین کسب‌وکار در قرن بیستم در مقابل قرن بیست‌ویکم را بیان کنند، با احتمال خیلی زیاد، آن‌ها یک کلمه را خواهند گفت: **داده**.

از آغاز قرن، مقدار داده‌ها با ظهور رسانه‌های اجتماعی، تلفن‌های هوشمند، اینترنت اشیا و دیگر پیشرفت‌های فن‌آوری با سرعت شگفت‌انگیزی افزایش‌یافته است. تخمین زده می‌شود که بیش از ۹۰ درصد از کل داده‌های ایجاد شده توسط انسان‌ها در پنج سال گذشته تولید شده‌اند. این انفجار اطلاعات به عنوان "**کلان داده**" شناخته می‌شود و به طور کامل جهان اطراف ما را متحول خواهد کرد. رشد نمایی در تولید داده‌ها، سازمان‌ها را بیش از هر اندازه‌ای به این فکر فروبرده است که چگونه می‌توانند از داده‌ها برای تحقق منافع تجاری خود استفاده کنند. در همین حال، افراد به طور فزاینده‌ای به دنبال توسعه مهارت‌های داده خود برای برجسته کردن رزومه، پیشرفت شغلی و کسب امنیت شغلی هستند.

امروزه، داده‌ها به‌عنوان ابزار و سوختی برای کسب‌وکارها در جهت بدست آوردن بینش‌های مهم و بهبود عملکرد آن‌ها هستند. علم داده‌ها تقریبا تمام صنایع جهان را تحت سلطه خود درآورده است. امروزه هیچ صنعتی در دنیا وجود ندارد که از داده‌ها استفاده نکند. اما چه کسی این بینش را بدست خواهد آورد؟ چه کسی تمام داده‌های خام را پردازش می‌کند؟ همه چیز توسط یک تحلیل‌گر داده یا یک دانشمند داده انجام می‌شود. این دو محبوب‌ترین نقش شغلی در این زمینه هستند. چراکه شرکت‌ها در سراسر جهان به دنبال بیشترین استفاده از داده‌ها هستند. برای افرادی که به دنبال پتانسیل شغلی بلند مدت هستند، مشاغل علم داده از مدت‌هاست که گزینه‌ی مطلوب و اطمینان‌بخشی هستند. این جریان با ادغام هوش مصنوعی و یادگیری ماشین در زندگی روزمره ما و اقتصاد، احتمالا ادامه‌دار خواهد بود.

داده‌ها فقط در مورد گذشته به ما نمی‌گویند. اگر آن را با دقت و با روش‌های دقیق مدل‌سازی کنیم، می‌توانیم الگوها و همبستگی‌هایی را برای پیش‌بینی بازارهای سهام، تولید توالی پروتئین، کشف ساختارهای بیولوژیکی همانند ویروس‌ها و موارد دیگر پیدا کنیم. با این حال، مدل‌سازی حجم زیادی از داده‌ها به صورت دستی کار خسته‌ کننده‌ای است. برای رسیدن به این هدف، به الگوریتم‌های یادگیری ماشین روی آورده‌ایم که می‌توانند به ما در استخراج اطلاعات از داده‌ها کمک کنند. یادگیری ماشین خودکارسازی وظایفی که معمولا به هوش انسانی نیاز دارند را انجام می‌دهند. به نوبه خود، یادگیری ماشین را می‌توان به عنوان استفاده و توسعه سیستم‌های رایانه‌ای تعریف کرد که قادر به یادگیری و تطبیق بدون برنامه‌ریزی صریح هستند. این سیستم‌ها از الگوریتم‌ها و مدل‌های آماری برای تحلیل و استنتاج از الگوهای موجود در داده‌ها استفاده

میکنند. الگوریتمهای یادگیری ماشین میتوانند مسائل پیچیدهای را که انجام دستی آنها غیرعملی و یا حتی غیرممکن است را حل کنند و توزیعها، الگوها و همبستگیها را بیاموزند تا دانش درون دادهها را آشکار کنند. الگوریتمها این کار را باکاوش یک مجموعه داده و ایجاد یک مدل تقریبی بر روی توزیع دادهها انجام میدهند، به طوریکه وقتی دادههای جدید و دیده نشده را تغذیه میکنیم، نتایج خوبی به همراه خواهد داشت.

به طور کلی، الگوریتمهای یادگیری ماشین میتوانند با سه رویکرد متفاوت توانایی یادگیری بدست آورند:

- **یادگیری بانظارت:** یادگیری بانظارت در یادگیری ماشین روشی است که مدل را برای پیشبینی نتیجه بر اساس دادههای برچسبگذاری شده ایجاد میکند. وجود دادههای برچسبدار به این معنا است که برای هر نمونهای از مجموعه داده، یک پاسخ یا راهحل داده شده است. به عنوان یک مثال ساده، اگر بخواهیم مدل یادگیری ماشین ما پیشبینی در مورد میوههای سیب یا موز داشته باشد، برچسب مقادیر "سیب" یا "موز" را به همراه مجموعهای از ویژگیها همانند وزن، طول، عرض و هر اندازهگیری مرتبط از میوههای موجود را میگیرد. بیایید به یک مثال مرتبطتر با کسبوکار نگاه کنیم؛ ریزش مشتری. برای درک بهتر ریزش مشتری، ابتدا باید تجزیه و تحلیل کنید که چه شاخصهایی ممکن است منجر به خروج مشتری شود. مجموعه داده شما برای این نوع مدل شامل متغیرهای شاخصی همانند، روزهای سپریشده پس از آخرین خرید، میانگین مقدار خرید و همچنین متغیر پیشبینیکننده برچسبگذاری شده، یعنی اینکه آیا فرد هنوز مشتری است یا خیر. از آنجایی که ما دادههای گذشته در مورد وضعیت مشتری را داریم، ایجاد مدلی با این نوع مجموعه داده میتواند کاندیدای عالی برای یادگیری بانظارت باشد.

- **یادگیری غیرنظارتی:** یادگیری غیرنظارتی در یادگیری ماشین زمانی است که نمونهها را بدون هیچ راهنمایی به الگوریتم ارائه میکنیم و ایجاد برچسب را به الگوریتم واگذار میکنیم. به عبارت دیگر، یادگیری غیرنظارتی، به یافتن الگوهای پنهان از دادههای بدونبرچسب و ایجاد گروهها و خوشهها میپردازد. به عنوان مثال، زمانی که ما باید بفهمیم که چگونه گروههای داخل مجموعه داده مشتری را میتوان بر اساس ویژگیها و رفتارهایشان به بخشهای مشابه دستهبندی کرد. یادگیری غیرنظارتی اغلب برای تجزیه و تحلیل اکتشافی و تشخیص ناهنجاری استفاده میشود. چراکه به یافتن نحوه ارتباط بخشهای داده و اینکه چه روندهایی ممکن است وجود داشته باشد، کمک میکند. آنها میتوانند برای پیشپردازش دادههای شما قبل از استفاده از الگوریتم یادگیری بانظارت استفاده شوند.

- **یادگیری تقویتی:** یادگیری تقویتی تکنیکی است که بازخورد آموزشی را با استفاده از مکانیزم پاداش ارائه می‌کند. فرآیند یادگیری به عنوان یک عامل رخ می‌دهد که با یک محیط تعامل دارد و روش‌های مختلف را برای رسیدن به یک نتیجه امتحان می‌کند. عامل هنگامی که به وضعیت مطلوب یا نامطلوب برسد پاداش یا تنبیه دریافت می‌کند. از طریق این بازخورد یادگیری، عامل یاد می‌گیرد که کدام حالت‌ها منجر به نتایج خوب و کدام منجر به شکست می‌شود و باید از آن‌ها اجتناب کرد. به عنوان مثال، زمانی که برای عملکرد موفقیت‌آمیز در یک محیط رقابتی، مانند یک بازی ویدیویی یا بازار سهام، به نرم‌افزار نیاز داریم، می‌توانیم از یادگیری تقویتی استفاده کنیم. در این حالت، نرم‌افزار شروع به فعالیت در محیط می‌کند و مستقیما از خطاهای خود درس می‌گیرد تا زمانی که مجموعه‌ای از قوانینی را پیدا کند که موفقیت آن را تضمین می‌کند. یادگیری تقویتی به داده‌های برچسب‌دار، همانند یادگیری بانظارت نیاز ندارد. علاوه بر این، حتی از یک مجموعه داده بدون‌برچسب، همانند یادگیری غیرنظارتی استفاده نمی‌کند. یادگیری تقویتی به جای تلاش برای کشف رابطه در یک مجموعه داده، به طور مداوم در بین نتایج تجربیات گذشته خود و همچنین ایجاد تجربیات جدید بهینه می‌شود. به عبارت دیگر، مجموعه داده‌ها و نتایج جدیدی را با هر تلاش ایجاد می‌کند.

یادگیری بانظارت در علم داده بسیار مهم است، چراکه به ما اجازه می‌دهد تا چیزی را که نسل بشر آرزوی آن را داشته است، انجام دهیم: **پیش‌بینی.** پیش‌بینی در تجارت و برای سودمندی کاربرد خیلی زیادی دارد و ما را قادر می‌سازد تا بهترین اقدام را انجام دهیم، چراکه از طریق پیش‌بینی، نتیجه احتمالی یک موقعیت را می‌دانیم.

یادگیری بانظارت ممکن است برای برخی همانند جادوگری به نظر برسد. با این حال، یادگیری بانظارت به هیچ وجه جادو نمی‌کند. بلکه، یادگیری بانظارت بر اساس دستاوردهای انسان در ریاضیات و آمار و با استفاده از تجربیات و مشاهدات انسانی و تبدیل آن‌ها به پیش‌بینی‌های دقیق به گونه‌ای که ذهن هیچ انسانی قادر به انجام آن نیست، کمک می‌کند.

با این حال، یادگیری بانظارت تنها در شرایط مطلوب خاصی می‌تواند پیش‌بینی کند. از این‌رو، برای این کار، بسیار مهم است که نمونه‌هایی از گذشته داشته باشیم که بتوانیم از آن‌ها قوانین و نکاتی استخراج کنیم که می‌تواند از جمع‌بندی آن‌ها یک پیش‌بینی بسیار محتمل ایجاد شود.

### چرا یادگیری ماشین؟

حجم داده‌های در دسترس ما به طور مداوم در حال افزایش است. ماشین‌ها از این داده‌ها برای یادگیری و بهبود نتایج و ارائه آن‌ها به ما استفاده می‌کنند. این نتایج می‌تواند در ارائه بینش‌های ارزشمند و همچنین اتخاذ تصمیمات تجاری آگاهانه بسیار مفید باشد. یادگیری ماشین به طور

مداوم در حال رشد است و در پی آن، کاربردهای یادگیری ماشین نیز رو به رشد هستند. ما بیش از آنچه می‌دانیم از یادگیری ماشین در زندگی روزمره خود استفاده می‌کنیم. یادگیری ماشین خود را وارد زندگی روزمره ما کرده است، حتی بدون اینکه ما متوجه شویم. الگوریتم‌های یادگیری ماشین به دنیای اطراف ما قدرت داده‌اند. ناگفته نماند که *آینده از قبل در اینجاست* و یادگیری ماشین نقش مهمی در نحوه‌ی پندارهای معاصر ما دارد.

امروزه یادگیری ماشین تمام توجهی که به آن نیاز دارد را دارد. یادگیری ماشین می‌تواند بسیاری از وظایف را به‌طور خودکار انجام دهد، به خصوص آن‌هایی که تنها انسان‌ها می‌توانند با هوش ذاتی خود انجام دهند. تکثیر این هوش در ماشین‌ها تنها با کمک یادگیری ماشین قابل دستیابی است.

با کمک یادگیری ماشین، کسب‌وکارها می‌توانند کارهای روتین را خودکار کنند. همچنین به خودکارسازی و ایجاد مدل‌هایی برای تجزیه و تحلیل داده‌ها کمک می‌کند. صنایع مختلف برای بهینه‌سازی عملکرد خود و اتخاذ تصمیمات هوشمندانه به مقادیر زیادی از داده‌ها وابسته هستند. یادگیری ماشین به ایجاد مدل‌هایی کمک می‌کند تا بتوانند مقادیر زیادی از داده‌های پیچیده را پردازش و تجزیه و تحلیل کنند و نتایج دقیقی را ارائه کنند. این مدل‌ها، دقیق و مقیاس‌پذیر هستند و با تابع زمانی کمتری کار می‌کنند. با ساخت چنین مدل‌هایِ دقیق یادگیری ماشین، کسب‌وکارها می‌توانند از فرصت‌های سودآور استفاده کنند و از ریسک‌های ناشناخته اجتناب کنند.

تشخیص تصویر، تولید متن، طبقه‌بندی متن، تشخیص بیماری‌ها و بسیاری از موارد دیگر در دنیای واقعی کاربرد دارند. از این‌رو، این امر زمینه را برای درخشش کارشناسان یادگیری ماشین به عنوان یک متخصص حرفه‌ای افزایش می‌دهد. علاوه بر این، با توجه به سرعت سریعی که در جهش‌های فناوری انجام شده است، بسیاری از شرکت‌ها از فناوری عقب مانده‌اند. تحول دیجیتال صنعت بزرگی است و حقیقت موضوع این است که به اندازه کافی متخصص یادگیری ماشین برای پاسخگویی به نیازهای صنعت جدید وجود ندارد.

اگر می‌خواهید حرفه‌ی خود را به سطح دیگری ببرید، یادگیری ماشین می‌تواند این کار را برای شما انجام دهد. *اگر به دنبال این هستید که خود را درگیر کاری کنید که شما را بخشی از چیزی کند که هم جهانی و هم مرتبط با معاصر باشد، یادگیری ماشین می‌تواند این کار را برای شما انجام دهد.*

## ارتباط یادگیری ماشین و علم داده

یادگیری ماشین، تنها در صورتی می‌تواند بینش‌های ارزشمندی ارائه دهد که داده‌های باکیفیت را دریافت کند. از این‌رو، بدون استفاده از داده‌های تمیز، سازگار و با کیفیت، بینش معنی‌دار کمی (در صورت وجود) می‌توان تولید کرد. در عین حال، دانشمند داده به یادگیری ماشین نیاز

دارد، زیرا درک و پیش‌بینی دقیق نتایج حاصل از حجم عظیمی از داده‌های پیچیده که سازمان‌ها در اختیار دارند، عملاً غیرممکن است.

دانشمند داده همچنین باید حس تجاری عالی داشته باشد تا بفهمد چه چیزی باعث موفقیت کسب‌وکار می‌شود، کجا می‌توان آن را بهبود بخشید و گزینه‌های احتمالی برای دستیابی به آن نتیجه چیست. یادگیری ماشین را می‌توان آمار کاربردی برای انجام این امر در نظر گرفت. به عبارت دیگر، یادگیری ماشین، ادغام علوم رایانه و رشته‌های مختلف ریاضی است که در آن از مفاهیم علوم رایانه برای ساختن مدل‌های ریاضی قوی استفاده می‌شود که می‌تواند مجموعه‌ای از مسائل مشابه و مرتبط را حل کند.

دانشمندان داده نگران به اطمینان از این‌که آیا مدل یادگیری ماشین به اهداف پروژه دست می‌یابد، هستند. اینجاست که مجموعه مهارت‌های تجاری شاید مهم‌ترین مهارت برای داشتن باشد. برای موفقیت در توسعه مدل یادگیری ماشین، دانشمند داده باید درک معقولی از مشکل در دست و اهداف پروژه داشته باشد. بدون این مورد، احتمال موفقیت کمی برای هر برنامه علم داده و مدل یادگیری ماشین وجود دارد. در این راستا، تقریباً ۸۰ درصد از زمان یک دانشمند داده صرف کاوش، تمیزسازی و آماده‌سازی داده‌ها می‌شود. انجام درست این کار بخشی اساسی از فرآیند است. پس از تکمیل، دانشمند داده می‌تواند توسعه مدل یادگیری ماشین را شروع کند. آن‌ها می‌توانند مدل‌های مختلف را آزمایش و مقایسه کنند و سپس امیدوارکننده‌ترین نامزد را برای عرضه در محیطِ تولید، بهینه‌سازی کنند.

یکی از موثرترین راه‌ها برای مشاهده‌ی پذیرش این مدل‌ها از طریق مصورسازی داده‌ها است. ارائه‌ یک گزارش با داده‌ها، رهبران کسب وکار را قادر می‌سازد تا تصمیمات آگاهانه‌تری بگیرند که می‌تواند به نفع سازمان باشد. جدای از تهیه داده‌ها، این *مصورسازی شاید مهم‌ترین گام در کمک به اطمینان از موفقیت پروژه باشد.*

بنابراین، با وجود اینکه مدل‌های یادگیری ماشین مهم هستند، موفقیت آن‌ها به شدت وابسته به توانایی تیم داده‌ها برای درک و فراهم کردن داده‌های ساختار یافته با اطلاعات عالی است که به مدل اجازه می‌دهد تا پیش‌بینی‌های دقیق انجام دهد.

علم داده و یادگیری ماشین به یکدیگر وابسته هستند و برای موفقیت یک سازمانِ داده‌محور بسیار مهم هستند. با این حال، همه چیز به کیفیت داده‌های مورد استفاده بستگی دارد.

### درباره‌ی کتاب

کتاب حاضر شامل دو بخش **مقدمات** و **یادگیری ماشین** است که در مجموع شامل **۹** فصل می‌شود. عمده‌ی مطالب اصلی و مهم کتاب در بخش دوم می‌باشد. از این‌رو، اگر با مفاهیم برنامه‌نویسی و پیش‌پردازش داده‌ها آشنایی دارید، <span style="color:red">**بخش اول کتاب مناسب شما نیست**</span> و می‌توانید از این بخش عبور کنید و مستقیماً وارد بخش یادگیری ماشین شوید.

**مخاطبین**

این کتاب می‌تواند به عنوان یک درس اختیاری برای دانشجویان سال آخر در مقطع **کارشناسی** و یک کتاب درسی برای دانشجویان مقطع **کارشناسی ارشد** در رشته‌های **مهندسی** و **علوم کامپیوتر** باگرایش هوش مصنوعی در نظر گرفته شود. همچنین این کتاب می‌تواند مرجع مناسبی برای تمامیِ علاقه‌مندان به یادگیری ماشین و علم‌داده، از محققین در رشته‌های مختلف گرفته تا پزشکان باشد.

از خوانندگان محترم تقاضا می‌شود انتقادات، پیشنهادات و یا در صورت مشاهده‌ی هرگونه اشکال در کتاب، اینجانب را آگاه سازند:

vazanmilad@gmail.com

میلاد وزان

بیست‌ویکم دی‌ماه هزار و چهارصد ـ خراسان رضوی،کاشمر

"هر چیزی را باید تا حد امکان ساده کرد، اما نه ساده‌تر"

آلبرت اینشتین

# فهرست مطالب

# بخش اول: مقدمات

## فصل۱: علم داده

## فصل ۲: مقدمه‌ای بر پایتون

## فصل ۳: داده

**بخش دوم: یادگیری ماشین**

## فصل ۴: مقدمه‌ای بر یادگیری ماشین

## فصل۵: انتخاب و ارزیابی مدل

## فصل ۶: یادگیری بانظارت

# فصل ۷: یادگیری عمیق

## فصل۸: یادگیری غیرنظارتی

## فصل۹: مباحث منتخب

# بخش اول

## مقدمات

**شامل فصل‌های:**

فصل اول: علم داده

فصل دوم: مقدمه‌ای بر پایتون

فصل سوم: داده

# علم داده

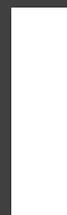

# علم داده چیست؟

در قرن گذشته نفت به عنوان طلای سیاه در نظر گرفته شد و با انقلاب صنعتی و ظهور صنعت خودرو، نفت منبع محرک اصلی تمدن بشر شد. با این حال، با گذشت زمان ارزش آن به دلیل بدست آوردن منابع تجدیدپذیر جایگزین انرژی کاهش یافت. در قرن بیست‌ویکم، نیروی محرک جدید که همان داده‌ها هستند بر صنایع تاثیر گذاشته است. این داده‌ها هر جنبه‌ای از هر چیزی که ما می‌دانیم و انجام می‌دهیم و نشان می‌دهند و اطراف ما را احاطه کرده‌اند. حتی صنایع خودرو از داده‌ها برای تأمین استقلال و بهبود ایمنی وسایل نقلیه خود استفاده می‌کنند.

امروزه علم داده را می‌توان نیروی برقی دانست که به صنایع قدرت می‌دهد و انقلاب عظیمی را تقریبا در تمام صنایع به ارمغان آورده است. صنایع به داده‌ها نیاز دارند تا عملکرد خود را بهبود ببخشند، رشد کسب‌وکار خود را افزایش دهند و محصولات بهتری برای مشتریان خود فراهم کنند. با این همه، هرچند که اکثر سازمان‌ها به داده‌ها برای پیشبرد تصمیمات تجاری خود تاکید دارند، اما داده‌ها به تنهایی هدف نیستند. اگر نتوان بینش ارزشمندی که منجر به اقدامات آگاهانه‌تر می‌شود را بدست آورد، حقایق و ارقام بی‌معنی هستند.

علم داده به عنوان یک مسیر شغلی حرفه‌ای و امیدوارکننده در حال تکامل است. چه‌بسا شنیده باشید هاروارد شغل جذاب قرن بیست‌ویکم را دانشمند داده معرفی کرده است. علم داده آینده هوش مصنوعی و حرفه‌ی آینده است. صنایع در حال تبدیل شدن به داده‌محوری هستند و به دانشمندان داده نیاز دارند تا از آن‌ها در تصمیم‌گیری‌های هوشمندانه و ایجاد محصولات بهتر کمک بگیرند. همچنین در دنیای امروز، نیاز به داشتن سواد داده به یک ضرورت تبدیل شده‌است. ما باید یاد بگیریم که چگونه داده‌های خام می‌توانند به محصولاتی معنی‌دار تبدیل شوند. ما باید تکنیک‌هایی را بیاموزیم و الزامات تجزیه و تحلیلِ بینشِ از داده‌ها را درک کنیم. داده‌ها پتانسیل بکری را در اختیار دارند که باید به منظور توسعه محصولات مفید، محقق شوند. با ظهور فن‌آوری‌های یادگیری ماشین و یادگیری عمیق، پیش‌بینی و دسته‌بندیِ هوشمندانه داده‌ها امکان‌پذیر شده است.

کلان داده و علم داده کلید آینده هستند. از همین‌رو بسیار مهم است که بدانیم علم داده چیست و چگونه می‌تواند بر کسب‌وکار و محیط اطراف ما تاثیرگذار باشد. باید پویا باشیم، با تکنولوژی کار کنیم و به جلو حرکت کنیم و قبل از اینکه خیلی دیر شود، علم داده را بیاموزیم.

## تعریف | علم داده

علم داده زمینه‌ی تحقیقاتی است که با ترکیبی از ابزارهای مختلف، الگوریتم‌ها، اصول یادگیری ماشین، متن‌کاوی، ریاضیات و آماربه کشف الگوهای پنهان از داده‌های خام می‌پردازد.

شاغلین در حوزه‌ی علم داده، با اعمال الگوریتم‌های یادگیری ماشین، ریاضیات و آمار به انواع مختلفی از داده‌ها، سعی می‌کنند سیستم هوش مصنوعی بسازند تا به انجام وظایفی بپردازد که معمولا نیاز به هوش انسانی دارد و یا مسائل پیچیده را به مسائل کوچک‌تر تقسیم کند تا دیدگاه و دانشی را از آن‌ها بدست آورند. به نوبه خود، این سیستم‌ها دیدگاهی تولید می‌کنند که نقش بسیار مهمی در پیشبرد اهداف تحلیل‌گران و کاربران تجاری به‌همراه دارد. به‌طور خلاصه می‌توان گفت، *هدف علم داده توضیح فرآیندها از طریق داده‌های موجود است.* انتظار می‌رود این توضیح به اندازه کافی دقیق باشد تا بتواند پیش‌بینی انجام دهد. هدف نهایی از این تفسیرها و توضیحات این است که تصمیماتی آگاهانه بر اساس دانش استخراج شده از این داده‌ها انجام دهیم.

## کلان داده چیست؟

داده‌ها، پایه و اساس علم داده هستند؛ داده‌ها همان مولفه‌های اصلی هستند که تمام تجزیه و تحلیل‌ها بر پایه‌ی آن‌ها استوار است. در زمینه علم داده، می‌توان این داده‌ها را به دو گروه تقسیم‌بندی کرد: **داده‌های سنتی و کلان داده.**

داده‌های سنتی، به داده‌هایی اشاره دارد که در پایگاه‌های داده‌ای که تحلیل‌گران می‌توانند در یک رایانه آن‌ها را مدیریت کنند، تهیه و ذخیره‌سازی شده است. این داده‌ها در قالب جدولی هستند که حاوی مقادیر عددی یا متنی است. البته باید گفت که واژه‌ی "سنتی" چیزی است که بیشتر استفاده می‌کنیم تا بتوانیم بهتر تمایز بین کلان داده و انواع دیگر داده را نشان دهیم. در طرف مقابل، کلان داده، داده‌هایی بزرگ‌تر از داده‌ها سنتی هستند و معمولا در یک شبکه گسترده از رایانه‌ها توزیع می‌شوند.

**تعریف**    **کلان داده**

کلان داده به مجموعه داده‌های ساختاریافته پیچیده و بدون ساختار با حجم بالا اشاره دارد که به سرعت تولید و از انواع مختلفی از منابع به‌دست آمده و سبب افزایش بینش و تصمیم‌گیری می‌شوند.

کلان داده به گروه بزرگی از داده‌های ناهمگن اشاره دارد که از منابع گوناگونی بدست می‌آید و شامل انواع مختلفی از داده‌ها به صورت زیر می‌شود:

- **داده‌های بدون ساختار:** شبکه‌های اجتماعی، ایمیل‌ها، وبلاگ‌ها، توییت‌ها، تصاویر دیجیتال، داده‌های تلفن همراه، صفحات وب و غیره.
- **نیمه‌ساخت یافته:** فایل‌های XML، فایل‌های متنی و غیره.
- **ساختار یافته:** پایگاه داده‌ها و سایر فرمت‌های ساختاری.

کلان داده اساسا یک کاربرد ویژه از علم داده است که در آن مجموعه داده‌ها بسیار بزرگ است و نیاز به غلبه بر چالش‌های منطقی برای مقابله آن‌ها دارد. علم داده یک رویکرد علمی است که ایده‌های الگوریتمی، محاسباتی و ابزارهای رایانه‌ای را برای پردازش این داده‌های بزرگ اعمال می‌کند.

به طور کلی نگرانی اصلی کلان داده، ذخیره، استخراج، پردازش و تجزیه‌وتحلیل در این مجموعه‌ی عظیم از داده‌ها است. پردازش و تحلیل این مجموعه داده‌های بزرگ اغلب به دلیل محدودیت‌های محاسباتی امکان‌پذیر نیست. از همین‌رو روش‌ها و ابزارهای ویژه‌ای به عنوان مثال: الگوریتم، نرم‌افزار، برنامه‌نویسی موازی و غیره را مورد نیاز دارد.

## تفاوت علم داده با کلان داده

در زیر تفاوت‌های بین علم داده و کلان داده فهرست شده‌اند:

- سازمان‌ها به داده‌های بزرگ نیاز دارند تا کارایی خود را بهبود بخشند، رشد کسب‌وکار خود را افزایش دهند و محصولات بهتری برای مشتریان خود فراهم کنند. در حالی که علم داده روش‌ها و سازوکارهای درک و استفاده از پتانسیل داده‌های بزرگ را به موقع فراهم می‌کند.
- در حال حاضر، برای سازمان‌ها هیچ محدودیتی برای مقدار داده‌های ارزشمندی که می‌تواند جمع‌آوری شود وجود ندارد. اما برای استفاده از همه‌ی این داده‌ها برای استخراج اطلاعات معنی‌دار برای تصمیمات سازمانی، علم داده مورد نیاز است.
- علم داده‌ها به وضوح از رویکردهای نظری و عملی برای کاوش اطلاعات از داده‌های بزرگ استفاده می‌کند که نقش مهمی در استفاده از پتانسیل داده‌های بزرگ ایفا می‌کند. کلان داده‌ها را می‌توان به عنوان استخری از داده‌ها در نظر گرفت که اعتبار ندارد، مگر اینکه با استدلال قیاسی و استقرایی تجزیه و تحلیل شود.
- تجزیه و تحلیل داده‌های بزرگ به داده کاوی مرتبط می‌شود. اما علم داده‌ها از الگوریتم‌های یادگیری ماشین برای طراحی و توسعه مدل‌های آماری برای تولید دانش از حجم عظیم کلان داده استفاده می‌کند.

از تفاوت‌های بالا بین کلان بزرگ و علم داده، ممکن است مشخص شود که علم داده‌ها در مفهوم کلان داده گنجانده شده‌است. علم داده نقش مهمی در بسیاری از حوزه‌های کاربردی بازی می‌کند. علم داده بر روی داده‌های بزرگ کار می‌کند تا از طریق تجزیه و تحلیل پیش‌گویانه نتایج مفیدی بدست آورد که در آن نتایج برای اتخاذ تصمیمات هوشمندانه مورد استفاده قرار می‌گیرند.

تفاوت اساسی بین کلان داده و علم داده را می‌توان با ذکر یک مثال بهتر درک کرد. علم داده همانند یک کتاب است که در آن شما می‌توانید یک راه حل برای مشکلات خود پیدا کنید. از سوی دیگر، کلان داده را می‌توان به عنوان یک کتابخانه بزرگ در نظر گرفت که در آن تمام پاسخ‌های سوالات در آنجا هستند، اما یافتن پاسخ به سوالات شما دشوار است.

# چرا علم داده را یاد بگیریم؟

ما در جالب‌ترین زمان تاریخ بشر زندگی می‌کنیم. دوره‌ای که در آن داده‌ها به یک کالا تبدیل شده‌اند که ارزشمندتر از نفت و طلا هستند. میزان داده‌های تولید شده در سطح جهانی بی‌سابقه است و انتظار می‌رود که با افزایش بیشتر جمعیت جهان و دسترسی بیشتر به اینترنت همچنان ادامه پیدا کند.

این داده‌های تولید شده منابع بسیار با ارزشی هستند و علم در رمزگشایی آن‌هاست. همچنین، تغییرات انقلاب گسترده‌ای در الگوی رفتاری مشتریان در خریدهای برخط، سرمایه‌گذاری در بازار سهام و... بوجود آمده است که هر کدام از این فعالیت‌ها نیاز به تجزیه و تحلیل عمیق از داده‌ها را طلب می‌کند. همین‌جا است که علم داده پا به عرصه می‌گذارد و مورد تقاضای شرکت‌ها، سازمان‌ها و... قرار می‌گیرد.

# تجزیه و تحلیل داده

تجزیه و تحلیل داده‌ها فرآیند جمع‌آوری، مدل‌سازی و تحلیل داده‌ها برای استخراج بینش‌هایی است که در تصمیم‌گیری‌ها، کمک‌کننده هستند. به شخصی که عهده‌دار این تجزیه و تحلیل‌ها می‌باشد، **تحلیل‌گر داده** گویند. یک تحلیل‌گر داده، داده‌ها را از طریق چندین روش مانند تمیزسازی داده‌ها، تبدیل داده‌ها و مدل‌سازی داده‌ها استخراج می‌کند. روش‌های و تکنیک‌های متعددی برای تجزیه و تحلیل، بسته به صنعت و هدف وجود دارد. تجزیه و تحلیل داده‌ها به صنایع این امکان را می‌دهد تا پرس‌وجوهای سریع را پردازش کنند و نتایج عملی را که در مدت زمان کوتاهی مورد نیاز است، تولید کند.

> تحلیل‌گران داده، فارغ از اینکه در کدام صنعت فعالیت می کنند، می‌توانند زمان خود را صرف توسعه سیستم‌هایی برای جمع‌آوری داده‌ها و جمع‌آوری یافته‌های خود در گزارش‌هایی کنند که می‌تواند به پیشرفت و بهبود شرکت آن‌ها کمک کند. تحلیل‌گران داده می توانند در هر بخشی از فرایند تجزیه و تحلیل دخلیل باشند. در نقش یک تحلیل‌گر داده، شما می‌توانید درگیر همه چیز شوید، از راه اندازی یک سیستم تجزیه و تحلیل تا ارائه بینش بر اساس داده‌هایی که گردآوری می‌کنید. حتی ممکن است از شما خواسته شود که دیگران را در سیستم جمع‌آوری داده‌های خود آموزش دهید.

با این که علم داده و تجزیه و تحلیل داده ممکن است از زمینه‌ی مشترک آمار حاصل شود، اما نقش‌ها و ریشه‌های آن‌ها متفاوت است. با این حال اکثر مردم فکر می‌کنند که علم داده و تجزیه و تحلیل داده مشابه یک‌دیگر هستند. به منظور درک تفاوت‌های آن‌ها، باید آن‌ها را مورد ارزیابی قرار دهیم. در زیر برخی از این تفاوت‌ها فهرست شده‌اند:

- اولین تفاوت کلیدی میان دانشمند داده و تحلیل‌گر داده این است که، در حالی که تحلیل‌گر داده‌ها با حل مسائل سر و کار دارد، یک دانشمند داده، مشکلات را شناسایی کرده و سپس آن‌ها را حل می‌کند. تحلیل‌گران داده‌ها توسط شرکت‌ها استخدام می‌شوند تا مشکلات تجاری خود را حل کنند. نقش یک تحلیل‌گر داده، یافتن روند بهتر فروش و یا استفاده از آمار خلاصه‌ای برای توصیف تراکنش‌های مشتری است. از سوی دیگر، یک دانشمند داده‌ها نه تنها مشکلات و مسائل را حل می‌کند، بلکه مشکلات را در وهله اول نیز شناسایی می‌کند.

- تحلیل‌گران داده‌ها به مهارت‌های ارتباطی و فراست تجاری نیازی ندارند. تحلیل‌گر داده‌ها محدود به مرزهای تجزیه و تحلیل داده‌ها است. آن‌ها برای ارتباط دادن نتایج با تیم مورد نیاز نیستند و به آن‌ها در تصمیم‌گیری‌های مبتنی‌بر داده کمک می‌کنند. با این حال، یک دانشمند داده‌ها باید مهارت داستان‌سرایی و مهارت‌های مدیریتی را به منظور ترجمه یافته‌های خود به استراتژی‌های تجاری داشته باشد. بنابراین، یک دانشمند داده نقش مهمی در روند تصمیم‌گیری شرکت دارد.

- یکی دیگر از تفاوت‌های بین یک دانشمند داده و تحلیل‌گر داده تفاوت در داده‌گردانی داده‌ها است. تحلیل‌گر داده‌ها از پرسش‌وجوهای SQL برای بازیابی و مدیریت داده‌های ساختار یافته استفاده می‌کند. درمقابل، دانشمندان داده‌ها از NoSQL برای داده‌های بدون‌ساختار استفاده می‌کند. بنابراین، دانشمندان داده مسئول مدیریت هر دو نوع داده‌های بدون ساختار و ساختاری هستند.

- تحلیل‌گر داده‌ها با توسعه مدل‌سازی پیشگویانه یا ابزار آماری برای پیش‌بینی داده‌ها سروکار ندارد. با این حال، دانشمندان داده نیاز به دانش یادگیری ماشین برای ساخت مدل‌های پیش‌بینی قدرتمند دارند. این مدل‌های پیش‌بینی مدل‌های رگرسیون و دسته‌بندی هستند.

- دانشمندان داده نیاز به تنظیم مدل‌های داده در جهت ایجاد بهتر محصولات داده‌ها دارند. همچنین نیازمند بهینه‌سازی عملکرد مدل‌های یادگیری ماشین است. این مورد توسط تحلیل‌گران داده‌ها مورد نیاز نیست. بنابراین، نقش یک دانشمند داده نه تنها شامل ساخت مدل‌ها، بلکه تنظیم و حفظ آن‌ها نیز می‌شود.

تجزیه و تحلیل داده و علم داده همپوشانی قابل توجهی با یکدیگر دارند و در عین حال کاملا از هم متمایز هستند. تحلیل‌گران داده بر روی اینجا و اکنون تمرکز می‌کنند، در حالی‌که دانشمندان داده آنچه را که ممکن است اتفاق بیافتد را پیش‌بینی می‌کنند.

تحلیل‌گران داده اغلب دانشمندان سطح پایین داده‌ها هستند که بیشتر وقت خود را صرف تجزیه و تحلیل داده‌ها و ارائه پیشنهادها می‌کنند. با این حال، آن‌ها معمولا نیازی به ایجاد برنامه‌های فنی و الگوریتم‌های یادگیری ماشین ندارند. به این دلیل که تحلیل‌گران داده، برخلاف دانشمندان داده، ارتباط چندانی با تجزیه و تحلیل‌های پیش‌گویانه ندارند. آن‌ها با داده‌های موجود کار می‌کنند و خلاصه‌ای از انواع جزئیات عملکرد شرکت را ارائه می‌دهند.

# مسئولیت‌های اصلی یك تحلیل‌گر داده

پاسخ به پرسش "**تحلیل‌گر داده چه کاری انجام می‌دهد؟**" بسته به نوع سازمان و میزان اتخاذ تصمیمات مبتنی‌بر داده توسط یک کسب و کار متفاوت خواهد بود. هرچند، مسئولیت‌های یک تحلیل‌گر داده به‌طور معمول شامل موارد زیر می‌شود:

- **توسعه و پیاده‌سازی پایگاه‌داده، سیستم‌های جمع‌آوری داده و راهبردهای دیگری که کارایی و کیفیت آماری را بهینه می‌کنند.**

- **استخراج داده‌ها از منابع اولیه و ثانویه، سپس سازماندهی مجدد داده‌ها در قالبی که به راحتی توسط انسان یا ماشین قابل خواندن باشد.**

- **استفاده از ابزارهای آماری برای تفسیر مجموعه‌های داده، توجه ویژه به جریان‌ها[1] و الگوهایی که می‌توانند برای تحلیل‌های تشخیصی و پیش‌گویانه ارزشمند باشند.**

- **همکاری با برنامه‌نویسان، مهندسین و رهبران سازمانی برای شناسایی فرصت‌ها برای بهبود فرآیند، پیشنهاد اصلاح سیستم و تدوین سیاست‌هایی برای مدیریت داده‌ها.**

- **ایجاد مستندسازی مناسب که به ذینفعان اجازه می‌دهد مراحل فرآیند تحلیل داده‌ها را درک کنند و در صورت لزوم تحلیل را تکرار کنند.**

# انواع تجزیه و تحلیل داده

تجزیه و تحلیل داده‌ها یک ابزار ضروری برای بهینه‌سازیِ عملکردِ کلیِ هر شرکت و سازمانی است. با اجرای برخی از تجزیه و تحلیل داده‌ها در مدل‌های تجاری خود، شرکت‌ها احتمالا می‌توانند تصمیمات بهتری بگیرند، فرآیندهای خود را بهینه کنند و هزینه‌های خود را کاهش دهند. چهار نوع تحلیل داده‌ها وجود دارد که در تمامی صنایع کاربرد دارند. اما بهترین نوع مدل تجزیه و تحلیل داده برای هر شرکت و سازمانی چیست؟

درحالی‌که ما این بخش‌ها را به دسته‌های مجزا تقسیم می‌کنیم، همه آن‌ها با هم ارتباط دارند و به یکدیگر متصل می‌شوند. روش‌های تجزیه و تحلیل به چهار دسته‌ی عمده به ترتیب پیچیدگی: تحلیل توصیفی، تشخیصی، پیش‌گویانه و تجویزی، تقسیم‌بندی می‌شوند. همانطور که شروع به حرکت از ساده‌ترین نوع به پیچیده‌تر می‌کنید، میزان دشواری و منابع مورد نیاز افزایش می‌یابد، در عین حال، سطح بینش نیز افزایش می‌یابد.

---

[1] trends

## تجزیه و تحلیل توصیفی[1]- چه اتفاقی افتاده است

تجزیه و تحلیل توصیفی، ساده‌ترین، رایج‌ترین و اولین گام مهم برای انجام هر فرآیند تحلیلی آماری است و هدف آن پاسخ به سوال است که *چه اتفاقی افتاده است؟* به عبارت دیگر توصیفی از آن‌چه در گذشته اتفاق افتاده است را بررسی می‌کند: درآمد ماهانه، فروش شش ماه اخیر، بازدید سالانه وب‌سایت و غیره. این کار با خلاصه‌سازی داده‌های گذشته، دست‌ورزی و تفسیر داده‌های خام از منابع مختلف برای تبدیل آن به بینش‌های ارزشمند انجام می‌شود. این تحلیل به ما نحوه‌ی توزیع داده‌ها را ارائه می‌دهد، به شناسایی دورریزها کمک می‌کند و ما را قادر می‌سازد تا رابطه‌ی بین متغیرها را شناسایی کنیم، در نتیجه داده‌ها را برای انجام تجزیه و تحلیل آماری بیشتر آماده کنیم.

*تجزیه و تحلیل توصیفی چگونه می‌تواند در دنیای واقعی کمک کند؟* به‌عنوان مثال در یک بخش مراقبت‌های بهداشتی، بیان می‌کند که تعداد زیادی از افراد در مدت‌زمان کوتاهی در اتاق اورژانس پذیرفته می‌شوند. تجزیه و تحلیل توصیفی به شما می‌گوید که این اتفاق می‌افتد و *داده‌های درلحظه[2]* را با تمام آمار مربوطه (تاریخ وقوع، حجم، جزئیات بیمار و غیره) فراهم می‌کند.

انجام تجزیه و تحلیل توصیفی ضروری است. چرا که به ما این امکان را می‌دهد تا داده‌های خود را به روشی معنی‌دار ارائه کنیم. اگرچه لازم به ذکر است که این تجزیه و تحلیل به خودی خود به شما اجازه نخواهد داد تا نتایج آینده را پیش‌بینی کنید و یا پاسخ به سوالاتی مانند این که چرا یک اتفاق افتاد را پیش‌بینی کنید، اما داده‌های شما سازمان‌دهی شده و با تبدیل مجموعه داده‌های بزرگ به مقدار کمی از اطلاعات که درک آن آسان‌تر است آماده انجام تجزیه و تحلیل بیشتر می‌شود. این کار را معمولا با جمع‌بندی و برجسته‌سازی ویژگی‌های اصلی مورد علاقه ما و همچنین استفاده از نمودارها و سایر نمایش‌های کاربرپسند انجام می دهد.

نکته منفی این است که تجزیه و تحلیل توصیفی یک پدیده را برجسته می‌کند بدون اینکه به‌طور دقیق دلیل وقوع آن را توضیح دهد. به همین دلیل، چنین رویکردی باید همیشه با سایر تجزیه و تحلیل‌ها ترکیب شود تا منافع واقعی کسب‌وکار شما را بدست آورد.

## تجزیه و تحلیل تشخیصی[3]- چرا این اتفاق افتاد

تجزیه و تحلیل تشخیصی که اغلب به عنوان تجزیه و تحلیل علت ریشه‌ای[4] شناخته می‌شود، یک نوع تجزیه و تحلیل پیشرفته است که یک گام فراتر می‌رود تا داده‌ها و یا محتوا را برای پاسخ به پرسش "چرا این اتفاق افتاد داد؟" بررسی کند. تجزیه و تحلیل تشخیصی با روش‌هایی مانند

---

داده‌کاوی، استخراج داده‌ها و هم‌بستگی مشخص می‌شود و نگاهی عمیق‌تر به داده‌ها دارد تا علل رویدادها و رفتارها را درک کنید و به شما این امکان را می‌دهد تا اطلاعات خود را سریع‌تر درک کنید و به پرسش‌های مهم نیروی کار پاسخ دهید. در مثال مراقبت‌های بهداشتی که پیش‌تر گفته شد، تجزیه و تحلیل تشخیصی داده‌ها را بررسی کرده و هم‌بستگی‌ها را ایجاد می‌کند. برای مثال، ممکن است به شما کمک کند تا مشخص کنید که تمام علایم بیمار: "تب بالا، سرفه خشک و خستگی" اشاره به یک عامل عفونی دارد. حال شما یک توضیح در مورد افزایش ناگهانی حجم در اورژانس دارید.

کلید موفقیت این رویکرد، دسترسی گسترده به داده‌ها است. مانند تجزیه و تحلیل توصیفی، تجزیه و تحلیل تشخیصی به داده‌های گذشته "داخلی" نیاز دارد، اما برخلاف مورد قبلی، تجزیه و تحلیل تشخیصی اغلب شامل اطلاعات خارجی از طیف گسترده‌ای از منابع برای تعیین جزئیات آنچه اتفاق داده‌است، می‌شود. به عنوان مثال ممکن است متوجه شوید که درآمد وب‌سایت شما در سه ماهه‌ی اخیر کاهش یافته است. این امر می‌تواند با کاهش هزینه‌های تبلیغات و همچنین تغییر در الگوریتم گوگل ارتباط داشته باشد. برای یافتن سرنخ‌هایی از این روند، لازم است که داده‌ها را از منابع مختلف مانند ثبت‌نام‌های جدید و موارد دیگر شناسایی کنیم تا اینکه یک الگوی مشکوک پیدا شود.

## تجزیه و تحلیل پیش‌گویانه[1]- چه اتفاقی خواهد افتاد

تجزیه و تحلیل پیش‌گویانه با تشخیص تمایلات در تجزیه و تحلیل‌های تشخیصی و توصیفی، نتایج احتمالی را تعیین می‌کند. تجزیه و تحلیل پیش‌گویانه داده‌های گذشته را می‌گیرد و آن را به یک مدل یادگیری ماشین تغذیه می‌کند که الگوهای کلیدی را در نظر می‌گیرد. سپس، مدل به داده‌های فعلی اعمال می‌شود تا پیش‌بینی کند که چه اتفاقی می‌افتد. این امر این امکان را برای یک سازمان فراهم می‌سازد تا یک اقدام پیش‌گیرانه‌ای انجام دهد. به عنوان مثال، مانند تماس با یک مشتری که به بعید است تمدید قرارداد کند. در بیمارستانی که مثال زدیم، تجزیه و تحلیل پیش‌گویانه ممکن است افزایش بیماران بستری‌شده چند هفته آینده را در بخش اورژانس پیش‌بینی کند. براساس الگوهای موجود در داده‌ها، این بیماری به سرعت در حال گسترش است.

## تجزیه و تحلیل تجویزی[2]- چه اقدامی باید انجام شود

تجزیه و تحلیل تجویزی فرآیندی است که داده‌ها را تجزیه و تحلیل می‌کند و توصیه‌های فوری در مورد چگونگی بهینه‌سازی شیوه‌های کسب‌وکار متناسب با چندین نتیجه پیش‌بینی شده را ارائه می‌دهد. در واقع، تجزیه و تحلیل تجویزی "آنچه که *ما می‌دانیم می‌گیرد (داده‌ها)*"، داده‌ها

---

را برای پیش‌بینی اینکه چه اتفاقی می‌افتد درک می‌کند و بهترین مراحل روبه‌جلو را براساس شبیه‌سازی آگاهانه پیشنهاد می‌کند و مشخص می‌کند که پیامدهای احتمالی هر یک از آن‌ها چیست. هدف تجزیه و تحلیل تجویزی، که قطعا پیشرفته‌ترین گروه در فهرست ما است، پیشنهاد یک روش عملی برای جلوگیری از مشکلات آینده یا بدست آوردن حداکثر سود از یک روند امیدوار کننده است.

تجزیه و تحلیل تجویزی هم به تحلیل توصیفی و هم با تحلیل پیش‌گویانه ارتباط دارد. در حالی که تجزیه و تحلیل توصیفی به دنبال ارائه‌ی بینشی در مورد آنچه اتفاق افتاده است و تجزیه و تحلیل پیش‌گویانه به مدل سازی و پیش‌بینی آنچه ممکن است رخ دهد کمک می‌کند، تجزیه و تحلیل تجویزی با توجه به پارامترهای شناخته شده، به دنبال تعیین بهترین راه حل یا نتیجه از میان گزینه‌های مختلف است.

تجزیه و تحلیل تجویزی همچنین می‌تواند گزینه‌های تصمیم‌گیری برای چگونگی استفاده از یک فرصت آینده یا کاهش ریسک آینده را پیشنهاد کند و پیامدهای هر گزینه‌ی تصمیم‌گیری را نشان دهد. در عمل تجزیه و تحلیل تجویزی می‌تواند به طور مستمر و خودکار داده‌های جدید را پردازش کند تا صحت پیش‌بینی را بهبود بخشد و گزینه‌های تصمیم‌گیری بهتری را ارائه دهد. به مثال بیمارستان برگردیم: اکنون که می‌دانید بیماری در حال گسترش است، ابزار تجزیه و تحلیل تجویزی ممکن است پیشنهاد کند که شما تعداد کارمندان را برای درمان موثر بیماران افزایش دهید.

تجزیه و تحلیل تجویزی، پیشرفت طبیعی حاصل از روش‌های تجزیه و تحلیل توصیفی و پیش‌گویانه است و برای حذف حدس‌ورزی[1] از تجزیه و تحلیل داده‌ها، یک گام فراتر می‌رود. همچنین سبب صرفه‌جویی زمان دانشمندان داده و بازاریابان که در تلاش برای درک این که داده‌های آن‌ها چه معنایی دارند و چه نقاطی را می‌توان بهم متصل کرد تا تجربه کاربری کاملا شخصی و مطلوبی را به مخاطبان خود ارائه دهد، می‌شود. سازمان‌های آینده‌نگر از تجزیه و تحلیل‌های مختلف برای تصمیم‌گیری‌های هوشمندانه استفاده می‌کنند که به کسب‌وکار کمک می کند و یا در مورد بیمارستانی که مثال زده شد، جان انسان‌ها را نجات می‌دهد.

---

[1] guesswork

## چرخه دوام[1] علم داده

چرخه حیات علم داده‌ها شامل پنج مرحله است. دانشمندان دادهٔ اثربخش آن‌هایی هستند که می‌توانند هر کدام از این فازها را اجرا کنند. این پنج مرحله به شرح زیر هستند:

■ **مرحله اول: جمع‌آوری داده‌ها**

همان‌طور که از نام آن پیداست، اینجا جایی است که جمع‌آوری داده‌ها انجام می‌شود. کاربران هر روز میلیون‌ها داده تولید می‌کنند. هر پیوند کلیک شده، جستجوی انجام شده، عکس بارگذاری شده و پیام ارسال شده به انبار داده اضافه می‌شود. به این ترتیب، روند جمع‌آوری داده‌ها چیزی بی‌اهمیت نیست. باید مشخص شود که کدام داده‌ها مربوط به پروژه هستند. همچنین، کارِ شناسایی مکان جمع‌آوری داده‌ها نیز وجود دارد. انواع مختلفی از منابع داده‌ها از طریق سایت‌های خبری، نظرسنجی‌ها و غیره در یک سایت و رسانه‌های اجتماعی به صورت برخط در دسترس هستند.

■ **مرحله دوم: تمیزسازی داده‌ها**

مهم است که بدانیم داده‌های جمع‌آوری‌شده در فاز اول بدون‌ساختار هستند. یک دانشمند داده باید داده‌های خام را تمیزسازی و آن‌ها را طبقه‌بندی کند. این به معنای جست‌وجوی هرگونه ناسازگاری (داده‌های تکراری، داده‌های ناهنجار و غیره) در داده‌ها در جهت جلوگیری از هرگونه خطا در مراحل بعدی است. به دلیل وظایفی که در دست انجام است، مرحله دوم معمولاً بخش زمان‌بر یک پروژه علم داده است.

■ **مرحله سوم: کاوش داده‌ها**

تجزیه و تحلیل بعد از تمیزسازی مجموعه داده‌ها شروع می‌شود. دانشمندان داده، داده‌ها را به دقت بررسی می‌کنند تا ایده گسترده‌تری از الگوها و روندهای کلیدی مجموعه داده‌ها بدست آورند. مصورسازی و تجزیه و تحلیل آماری بر این مرحله حاکم است. کاوش داده‌ها نکاتی را که به تجزیه و تحلیل بیشتر نیاز دارند برجسته می‌کند. ابزارهای مصورسازی همچنین به دانشمندان داده اجازه می‌دهد تا موارد دورریز را یادداشت کنند و این موارد را بیشتر کاوش کنند.

■ **مرحله چهارم: مدل‌سازی داده‌ها**

مدل‌سازی در قلب روش پژوهش علم داده است. آن درک رابطه بین عناصر داده و نگاشت آن‌ها را تشکیل می‌دهد. از طریق مدل سازی داده‌ها ، یک دانشمند داده می‌بیند که مهم‌ترین عناصر چگونه با یکدیگر تعامل دارند و در کنار هم قرار می‌گیرند. روش‌های مختلفی برای

---

[1] Life Cycle

ساخت مدل وجود دارد. این مدل‌سازی می‌تواند از طریق روش‌های یادگیری ماشین یا مدل‌سازی آماری باشد. *تنها پس از مدل‌سازی، یک دانشمند داده شروع به استخراج بینش از آن می‌کند.*

- **مرحله پنجم: تفسیر داده‌ها**

  پس از آنکه از داده‌ها بینش معناداری را استخراج کردید، نوبت به آخرین مرحله از چرخه دوام علم داده یعنی تفسیر داده‌ها می‌رسد. اگر می‌خواهید اکتشافات ارزشمند شما به مرحله اجرا درآید، باید بتوانید این بینش‌ها را به شکلی جذاب و قابل فهم ارائه دهید تا ذینفعان پروژه بتوانند آن را به آسانی درک کنند.

## دانشمند داده

دانشمندان داده، نسل جدیدی از متخصصین تحلیل‌گر داده‌ها هستند که مهارت‌های فنی برای حل مسائل پیچیده را دارند و همچنین کنجکاو در کشف مسائلی هستند که باید حل شوند. کار یک دانشمند داده، ارائه‌ی دقیق‌ترین پیش‌بینی‌ها است. این کار نیاز به استفاده از فن‌آوری‌های پیشرفته تجزیه و تحلیل، از جمله یادگیری ماشین و مدل‌سازی پیش‌گویانه دارد. نقش دانشمند داده، مجموعه‌ای از چندین نقشِ فنیِ سنتی است، از جمله ریاضیدان، متخصص آمار و متخصص کامپیوتر. دانشمندان داده یک دهه پیش، زیاد توجه کسی را به خود جلب نمی‌کردند، اما محبوبیت ناگهانی آن‌ها نشان‌دهنده این است که چگونه شرکت‌ها در حال حاضر به کلان داده‌ها فکر می‌کنند. دیگر نمی‌توان انبوهی از داده‌های بدون‌ساختار را نادیده گرفت و فراموش کرد. این یک معدن طلای مجازی است که به افزایش درآمد کمک می‌کند.

بسیاری از دانشمندان داده کار خود را به عنوان متخصص آمار یا تحلیل‌گر داده آغاز کردند. اما همان‌طور که کلان داده (و فن‌آوری‌های ذخیره‌سازی و پردازش داده‌های بزرگ) شروع به رشد و تکامل کردند، این نقش‌ها نیز تکامل یافته و می‌یابند. داده‌ها دیگر فقط پس‌اندیشه‌ای برای پردازش فن‌آوری اطلاعات نیستند. این اطلاعات کلیدی است که نیاز به تجزیه و تحلیل، کنجکاوی خلاقانه و یک مهارت برای ترجمه ایده‌های برتر فن‌آوری به روش‌های جدید برای سود بردن است.

در تجارت، دانشمندان داده به طور معمول در تیم‌ها کار می‌کنند تا کلان داده‌ها را برای اطلاعاتی که می‌تواند در پیش‌بینی رفتار مشتری و شناسایی فرصت‌های سرمایه‌گذاری جدید مورد استفاده قرار گیرد، بکار گیرند. در بسیاری از سازمان‌ها، دانشمندان داده همچنین وظیفه تعیین بهترین روش‌های جمع‌آوری داده‌ها، استفاده از ابزارهای تجزیه و تحلیل و تفسیر داده‌ها را دارند.

دانشمندان داده کل چرخه دوام داده‌ها، از جمع‌آوری و سازماندهی تا تجزیه و تحلیل و تفسیر را مدیریت می‌کنند. بینش آن‌ها معمولا آینده‌نگر است. این به این معنی است که آن‌ها داده‌های گذشته مربوط را ارزیابی می‌کنند و بینش‌هایی را استخراج می‌کنند که می‌تواند به عنوان پایه‌ای برای ایجاد تغییرات بالقوه در رفتار یا روند مصرف‌کننده استفاده شود. این به سازمان‌ها امکان می‌دهد تا با راهبردهای بلند مدت ظاهر شوند.

## مهندس داده

یک دانشمند دادهٔ خوب، تنها به خوبی داده‌هایی است که به آن دسترسی دارد. اکثر شرکت‌ها داده‌های خود را در قالب‌های مختلف در پایگاه‌های داده و پرونده‌های متنی ذخیره می‌کنند. اینجاست که مهندسان داده وارد می شوند؛ آن‌ها خط لوله‌ای که این داده‌ها را به قالب‌هایی تبدیل می‌کنند که دانشمندان داده می‌توانند از آن‌ها استفاده کنند. وظیفهٔ اصلی آن‌ها، تهیه داده‌ها برای استفاده‌های تحلیلی یا عملیاتی است. آن‌ها داده‌ها را یکپارچه کرده، تحلیل کرده و از آن برای استفاده در برنامه‌های تجزیه و تحلیل استفاده می‌کنند. هدف آن‌ها این است که داده‌ها را به‌راحتی در دسترس قرار دهند و اکوسیستم داده‌های بزرگ سازمان خود را بهینه کنند. مهندسان داده به همان اندازهٔ دانشمندان داده مهم هستند، اما معمولا کم‌تر دیده می‌شوند.

درحالی‌که علم داده و دانشمندان داده به طور خاص به کاوش داده‌ها، یافتن بینش در آن‌ها و ساخت الگوریتم‌های یادگیری ماشین مشغول هستند، مهندس داده به کارکرد این الگوریتم‌ها در زیرساخت تولید و ایجاد خط لوله‌ی داده توجه دارد. بنابراین، مهندس داده یک نقش مهندسی در یک تیم علم داده یا هر پروژه مرتبط با داده است، جایی که نیاز به ایجاد و مدیریت زیرساخت‌های فناوری یک بستر داده دارد.

## نقش مهندس داده

مهندسان داده بر جمع‌آوری و آماده‌سازی داده‌ها برای استفاده توسط دانشمندان و تحلیل‌گران داده متمرکز هستند. مهندسان داده سه نقش اصلی را به‌شرح زیر برعهده می‌گیرند:

■ **کارشناس عمومی:** مهندسان داده با تمرکز عمومی، معمولا در تیم‌های کوچک کار می‌کنند. آن‌ها ممکن است مهارت بیشتری نسبت به اکثر مهندسین داده‌ها داشته باشند، اما دانش کم‌تری از معماری سیستم‌ها دارند. یک دانشمند داده که می‌خواهد مهندس داده شود، با نقش کارشناس عمومی به‌خوبی هم‌خوانی دارد. بدون مهندس داده، تحلیل‌گران و دانشمندان داده چیزی برای تجزیه و تحلیل ندارند. از این رو، مهندس داده یکی از اعضای مهم تیم علمی داده است.

- **خط‌لوله ــ محور:** در شرکت‌های متوسط، این مهندسان داده معمولا در کنار دانشمندان داده کار می‌کنند تا از داده‌های جمع‌آوری‌شده به‌طور مفید استفاده کنند. آگاهی از علم رایانه و سیستم‌های توزیع شده برای مهندسین خط‌لوله ــ محور برای انجام چنین تجزیه‌هایی ضروری است.

- **پایگاه‌داده ــ محور:** این مهندسین داده وظیفه پیاده‌سازی، نگهداری و جمع‌آوری پایگاه داده‌های تجزیه و تحلیل را دارند. این نقش معمولا در شرکت‌های بزرگ‌تر وجود دارد که داده‌ها در چندین پایگاه داده توزیع می‌شوند. مدیریت جریان داده‌ها یک شغل تمام‌وقت است و مهندسین داده در این نقش به‌طور کامل بر روی پایگاه‌داده تجزیه و تحلیل تمرکز می‌کنند. به عنوان مهندس دادی پایگاه‌داده- محور، باید در پایگاه داده‌های متعدد کار کنید و جدول‌ها را در انبار داده توسعه دهید.

## حوزه‌ها و مهارت‌های اساسی مطالعه در علم داده

علم داده اصطلاح گسترده‌ای است که برای تسلط در آن باید در حوزه‌های مختلف مهارت داشت. در زیر چند مورد از حوزه‌ها و جنبه‌های اساسی که برای تسلط در علم داده مورد نیاز است، فهرست شده‌اند.

### یادگیری ماشین

برای یک دانشمند داده، یادگیری ماشین یک مهارت اصلی است. ایده اصلی یادگیری ماشین این است که به ماشین‌ها اجازه دهد به‌طور مستقل با استفاده از انبوه داده‌هایی که به‌عنوان ورودی به ماشین تغذیه می‌شوند، بیاموزد. با پیشرفت فن‌آوری، ماشین‌ها آموزش می‌بینند تا همانند یک انسان در قابلیت تصمیم‌گیری رفتار کنند.

### یادگیری عمیق

یادگیری عمیق اغلب در علم داده‌ها مورد استفاده قرار می‌گیرد. چراکه در مقایسه با روش‌های یادگیری ماشین سنتی، بسیار بهتر عمل می‌کند. در مقایسه یادگیری ماشین با یادگیری عمیق می‌توان این‌گونه بیان کرد، در حالی‌که یادگیری عمیق به‌طور خودکار ویژگی‌ها را از ساختار داده‌ها استخراج می‌کند، این عمل توسط یادگیری ماشین باید به‌صورت دستی انجام گیرد و اگر در تصمیم‌گیری حل مسئله پیش‌بینی‌های نادرستی را انجام دهد، آنگاه متخصص یا برنامه‌نویس باید صراحتا به حل این مشکل بپردازد.

**ریاضیات**

برای ارتقا مهارت‌های خود در یادگیری ماشین، یک دانشمند داده باید دانش عمیقی از ریاضیات داشته باشد. دو موضوع مهم در ریاضیات از نظر کاربرد در علم داده جبرخطی و حسابان است. در حالی که جبرخطی تماما در مورد مطالعه بردارها و توابع خطی است، حسابان به مطالعه ریاضیاتی تغییرات پیوسته می‌پردازد. بسیاری از مفاهیم جبرخطی مانند تنسورها و بردارها در بسیاری از زمینه‌های یادگیری ماشین استفاده می‌شوند. به‌طور مشابه، حسابان در حوزه‌های مختلف یادگیری ماشین مانند تکنیک‌های بهینه‌سازی، مورد نیاز است.

**آمار و احتمالات**

جهان یک جهان احتمالی است، بنابراین ما با داده‌هایی کار می‌کنیم که احتمالاتی هستند؛ بدین معنی که با توجه به مجموعه‌ای خاص از پیش‌شرط‌ها، داده‌ها تنها بخشی از زمان را به شما نشان خواهند داد. برای استفاده درست از علم داده‌ها، فرد باید با احتمالات و آمار آشنا باشد. *آمار و احتمالات پیش‌نیازترین زمینه در علم داده است و داشتن دانش خوب در این زمینه الزامی است.*

**پردازش زبان طبیعی**

در حوزه علم داده، پردازش زبان طبیعی، یک جزء بسیار مهم و باکاربردهای وسیع در بخش‌های مختلف صنایع و شرکت‌ها می‌باشد. برای انسان درک زبان آسان است، با این حال، ماشین‌ها به اندازه کافی قادر به تشخیص آن نیستند. پردازش زبان طبیعی شاخه‌ای از هوش مصنوعی است که بر پرکردنِ شکاف بین ارتباطات انسان و ماشین تمرکز دارد تا ماشین را قادر به تفسیر و درک کند.

**مصورسازی داده‌ها**

مصورسازی داده‌ها یکی از مهم‌ترین شاخه‌های علم داده است. به بیان ساده، مصورسازی شامل نمایش داده‌ها در قالب نمودارها و گراف‌ها می‌باشد.

**زبان برنامه‌نویسی**

یک دانشمند داده به غیر از مهارت‌های اساسی رایانه همانند، تسلط در Microsoft Excel، باید مهارت برنامه‌نویسی داشته باشد تا بتواند برای کار با داده‌ها (پردازش، مصورسازی و غیره) و استفاده از مهارت‌های یادگیری ماشین و یادگیری عمیق در پیاده‌سازی پروژه‌ها از آن بهره ببرد.

**الگوریتم**

از آنجایی که همه سیستم‌های یادگیری ماشین براساس الگوریتم‌ها ساخته شده‌اند، پیش‌نیاز بسیار مهم این است که یک دانشمند داده، درک اساسی از الگوریتم‌ها و نحوه طراحی آن‌ها را داشته باشد.

## کاربرد علم داده

اکنون که از اهمیت علم داده، پیش‌نیازها و مهارت‌های مورد نیاز برای آن آگاهی یافتید، مهم است که بدانید علم داده چگونه می‌تواند در دنیای واقعی به‌کار گرفته شود و خواهیم دید که چگونه علم داده، امروزه جهان را متحول کرده است. از همین‌رو، در ادامه لیستی از برنامه‌های کاربردی علم داده‌ها را فهرست کرده‌ایم تا با کاربردهای آن بیشتر آشنا شوید.

### حمل و نقل

مهم‌ترین پیشرفت یا تحولی که علم داده در حمل و نقل داشته است، معرفی اتومبیل‌های خودران است. علم داده با تجزیه و تحلیل گسترده الگوهای مصرف سوخت، نظارت فعال بر وسیله نقلیه و رفتار راننده، جای پای محکمی در صنعت حمل و نقل ایجاد کرده است و با ارائه محیط‌های رانندگی ایمن‌تر برای رانندگان، بهینه‌سازی عملکرد خودرو، افزودن خودمختاری به اتومبیل‌ها و موارد دیگر، سبب تحول در حمل و نقل گردیده است. با استفاده از یادگیری تقویتی و خودمختاری، تولیدکنندگان خودرو می‌توانند خودروهای هوشمند و مسیرهای منطقی بهتری را بسازند.

### تشخیص ریسک و کلاه‌برداری

علم داده برای نخستین بار در امور مالی و بانکی مورد استفاده قرار گرفت. بسیاری از موسسات مالی در پایان هر سال دچار بدهی‌ها و ضررهایی بودند. بنابراین شیوه‌های علم داده به عنوان راه حل در نظر گرفته شد. آن‌ها برای تجزیه و تحلیل احتمال ریسک، یاد گرفتند که داده‌ها را بر اساس مشخصات مشتری، هزینه‌های گذشته و سایر متغیرهای ضروری جدا کنند. از این‌رو، می‌توانند بازاریابی هدفمند را بر اساس میزان درآمد هر سال مشتری انجام دهند.

### ژنتیک و ژنومیکس

علم داده به متخصصان علوم زیستی کمک می‌کند تا واکنش ژن‌ها به داروهای مختلف را تجزیه و تحلیل کنند. هدف آن درک و مطالعه تأثیر DNA بر سلامت فرد است که سعی می‌کند ارتباطات بیولوژیکی بین بیماری‌ها، ژن‌ها و پاسخ دارویی را بیابد.

### توسعه دارو

اکتشاف و کشف یک داروی جدید مستلزم سال‌ها تحقیق و آزمایش است تا به مرحله تولید برسد و در نهایت مجوز دریافت آن به فروشگاه های پزشکی و بیمارستان‌ها برای بیماران ارائه شود. می‌توان از الگوریتم‌های یادگیری ماشین و علم داده برای ساده‌سازی فرآیند و کاهش زمان لازم برای غربالگری اولیه ترکیبات دارویی مورد استفاده برای تولید دارو استفاده کرد. الگوریتم‌ها و علم داده همچنین می‌توانند نحوه واکنش بدن به ترکیبات خاص دارویی را با استفاده از

شبیه‌سازی‌ها و مدل‌های مختلف آماری و ریاضی پیش‌بینی کنند. این در مقایسه با آزمایش‌های سنتی آزمایشگاهی، بسیار سریع‌تر است. مدل‌ها همچنین می‌توانند نتایج آینده را با دقت بیشتری پیش‌بینی کنند.

## خلاصه فصل

- صنایع در حال تبدیل شدن به داده‌محوری هستند و به دانشمندان داده نیاز دارند تا از آن‌ها در تصمیم‌گیری‌های هوشمندانه و ایجاد محصولات بهتر کمک بگیرند.
- هدف علم داده توضیح فرآیندها از طریق داده‌های موجود است.
- کلان داده به گروه بزرگی از داده‌های ناهمگن اشاره دارد که از منابع گوناگونی بدست می‌آید.
- کلان داده اساسا یک کاربرد ویژه از علم داده است.
- روش‌های تجزیه و تحلیل به چهار دسته‌ی عمده به ترتیب پیچیدگی: تحلیل توصیفی، تشخیصی، پیش‌گویانه و تجویزی، تقسیم‌بندی می‌شوند.
- کار یک دانشمند داده، ارائه دقیق‌ترین پیش‌بینی‌ها است.
- یک دانشمند داده‌ی خوب، تنها به خوبی داده‌هایی است که به آن دسترسی دارد.
- مهندسان داده بر جمع‌آوری و آماده‌سازی داده‌ها برای استفاده توسط دانشمندان و تحلیل‌گران داده متمرکز هستند.

## مراجع برای مطالعه بیشتر

# ۲ مقدمه‌ای بر پایتون

اهداف:

- دلیل انتخاب پایتون
- مفاهیم برنامه‌نویسی در پایتون
- آشنایی با کتابخانه NumPy

# پایتون چیست؟

به عنوان دانشمند داده جدید، مسیر شما با زبان برنامه‌نویسی که باید یاد بگیرید شروع می‌شود. در میان تمام زبان‌هایی که می‌توانید انتخاب کنید، پایتون، محبوب‌ترین زبان برای یک دانشمند داده است. پایتون یک زبان برنامه‌نویسی شیگرا و سطح بالا است که برای طیف گسترده‌ای از مسائل با مقیاس و پیچیدگی‌های مختلف استفاده می‌شود. برخلاف بسیاری از زبان‌های مشابه، تسلط و یادگیری آن آسان است و برای مبتدیان بسیار ایده‌آل می‌باشد. اما این آسان بودن دلیلی بر کم اهمیت بودن آن نیست و حتی برای کاربران پیشرفته نیز به اندازه کافی قدرتمند می‌باشد. علاوه براین، پایتون، پرکاربردترین ابزار علم داده است و در بیشتر لیست اعلان‌های شغلی علم داده از آن به عنوان یک الزام نام برده می‌شود.

## دلیل انتخاب پایتون در علم داده؟

زبان پایتون یکی از با ارزش‌ترین و جالب‌ترین زبان‌ها برای تجزیه و تحلیل داده‌ها است و محبوبیت آن در دنیای تجزیه و تحلیل داده‌ها و علم داده، روز به روز در حال افزایش است. از آنجاکه پایتون یکی از انعطاف‌پذیرترین زبان‌های برنامه‌نویسی است، از همین‌رو مورد علاقه علم داده است. همچنین، افرادی که می‌خواهند وارد دنیای علم داده شوند، پایتون را به بسیاری از زبان‌های برنامه‌نویسی دیگر ترجیح می‌دهند. چراکه مجبور نیستند زمان زیادی را برای یادگیری آن صرف کنند. همچنین، بسته‌هایی در پایتون وجود دارد که به‌طور خاص برای کارهای مشخصی طراحی شده‌اند، از جمله، pandas، NumPy و SciPy. به‌طور خلاصه می‌توان گفت، پایتون به دلیل ویژگی‌هایی که در زیر آن‌ها را فهرست کرده‌ایم، علم داده را به تسخیر خود در آورده است:

- **سادگی:** پایتون یکی از آسان‌ترین زبان‌ها برای شروع است. همچنین، این سادگی شما را از امکاناتی که به آن‌ها نیاز دارید محدود نمی‌کند.
- **کتابخانه‌ها و چارچوب‌ها:** به دلیل محبوبیت، پایتون صدها کتابخانه و چارچوب مختلف را دارا است که به روند توسعه شما کمک بسیار زیادی می‌کند و در زمان بسیار صرفه‌جویی می‌کنند. به عنوان یک دانشمند داده، متوجه می‌شوید که بسیاری از این کتابخانه‌ها به علم داده و یادگیری ماشین متمرکز هستند.
- **جامعه‌ی عظیم:** یکی از دلایل مشهور بودن پایتون وجود یک جامعه‌ی عظیم متشکل از مهندسان و دانشمندان داده است. ممکن است فکر کنید که این نباید یکی از دلایل اصلی انتخاب شما برای پایتون باشد، اما حقیقت برعکس است. اگر از نظرات و پشتیبانی‌های متخصصان دیگر استفاده نکنید، مسیر یادگیری شما دشوار خواهد بود.

- **قابل توجه برای یادگیری عمیق:** پایتون دارای بسته‌های زیادی مانند: keras، Tensorflow و PyTorch است که به دانشمندان داده کمک می‌کند تا الگوریتم‌های یادگیری عمیق را ایجاد کند.
- **مصورسازی بهتر داده‌ها:** مصورسازی برای دانشمندان داده کلیدی و مهم است، چرا که به آن‌ها کمک می‌کند تا داده‌ها را بهتر درک کنند. پایتون با وجود کتابخانه‌هایی مانند: ggplot، Matplotlib، NetworkX و غیره می‌تواند به شما در مصورسازی‌های خیره‌کننده کمک کند.

# نصب پایتون

## بارگیری

در این بخش مراحل نصب پایتون را در سیستم عامل Windows معرفی می‌کنیم. از آنجایی که هیچ محیط پایتون داخلی در سیستم عامل ویندوز وجود ندارد، باید به‌طور مستقل نصب شود. بسته نصب را می‌توان از وب سایت رسمی پایتون (www.Python.org) بارگیری کرد. پس از باز کردن وب سایت رسمی، نوار ناوبری را که دارای دکمه "بارگیری"[۱] است جستجو کنید. وب سایت، پیوندی را به‌طور پیش فرض توصیه می‌کند، چرا که می تواند سیستم عامل شما را شناسایی کرده و آخرین نسخه Python ۳.x را توصیه کند. پس از ورود به صفحه بارگیری نسخه مربوطه، مقدمه‌ای اساسی در مورد محیطی که می‌خواهید بارگیری کنید وجود دارد. چندین نسخه مختلف عمدتا برای سیستم عامل‌های مختلف طراحی شده‌اند. بسته به ۳۲ یا ۶۴ بیتی بودن سیستم، می‌توانید فایل‌های مختلفی را برای بارگیری انتخاب کنید. در صفحه جدیدی که باز می شود، می‌توانیم نسخه‌های دیگری را نیز پیدا کنیم، از جمله آخرین نسخه آزمایشی و نسخه مورد نیاز. اگر می‌خواهید نسخه ۶۴ بیتی ۳/۹/۶ را نصب کنید، روی پیوند ارائه شده در صفحه کنونی کلیک کنید.

## نصب

پس از بارگیری پایتون، نوبت به نصب آن می‌رسد. نصب بسته ویندوز بسیار آسان است. درست مانند نصب سایر برنامه‌های ویندوز، ما فقط باید گزینه مناسب را انتخاب کرده و روی دکمه "بعدی" کلیک کنیم تا نصب کامل شود. هنگامی که گزینه‌ها در هنگام نصب ظاهر می‌شوند، برای رفتن به مرحله بعدی عجله نکنید. چرا که برای راحتی در آینده، باید یک دکمه را انتخاب کنید.

---

[۱] download

پس از علامت‌گذاری دکمه "Add Python ۳٫۹٫٦ to PATH" به متغیر محیط[1]، می‌توان در آینده دستورات پایتون را مستقیما و به‌راحتی در خط فرمان Windows اجرا کرد. پس از انتخاب "Add Python ۳٫۹٫٦ to PATH"، نصب دلخواه را انتخاب کنید. البته امکان انتخاب مکان نصب نیز وجود دارد، که به‌طور پیش فرض در دایرکتوری کاربری در درایو C نصب شده است. با این حال، بهتر است بدانید دایرکتوری کاربر چیست تا بتوانید در مواقع ضروری فایل‌های Python.exe نصب شده را پیدا کنید. دستورالعمل‌ها را ادامه دهید تا پایتون با موفقیت در سیستم نصب شود.

## شروع کار با پایتون

راه‌اندازی پایتون به دو صورت امکان‌پذیر است:

۱) **با استفاده از IDLE خود پایتون.** اگر می‌خواهید پایتون را اجرا کنید، می‌توانید روی دکمه "شروع" در دسکتاب ویندوز کلیک و در کادر "جستجو" عبارت "IDLE" را تایپ کنید تا به‌طور سریع وارد "read-evaluate-print-loop" شوید. پس از اجرای برنامه، با تصویری همانند زیر روبه‌رو می‌شوید:

IDLE یک IDE (محیط توسعه یک‌پارچه[2]) ساده خود پایتون است که یک ویرایشگر رابط گرافیکی را در اختیار کاربران قرار می‌دهد. عملکرد آن ساده به نظر می‌رسد و برای مبتدیان یادگیری زبان پایتون مناسب است. توسط IDLE یک محیط REPL ارائه می‌شود، یعنی ورودی کاربر () را می‌خواند، ارزیابی و محاسبه می‌کند ()، سپس نتیجه را چاپ می‌کند () و یک پیغام "حلقه" (منتظر ورودی بعدی) ظاهر می‌شود.

---

[1] environment variable

[2] Integrated Development Environment

۲)  **با استفاده از Windows Prompt.** راه دیگر برای راه‌اندازی پایتون، اجرای برنامه‌های پایتون از طریق خط فرمان ویندوز است. برای این کار کلیدهای **"Win+R"** را فشار دهید تا کادر اعلان باز شود و سپس در کادر باز شده، **"cmd"** را وارد کنید. اگر در هنگام نصب پایتون "Add Python ۳.x to PATH" را علامت زده باشید، پایتون نصب‌شده به متغیر محیط ویندوز اضافه شده است. حال با وارد کردن کلمه **"python"** پس از ظاهر شدن "‹" پایتون با موفقیت اجرا می‌شود و با تصویری همانند زیر روبه‌رو می‌شوید:

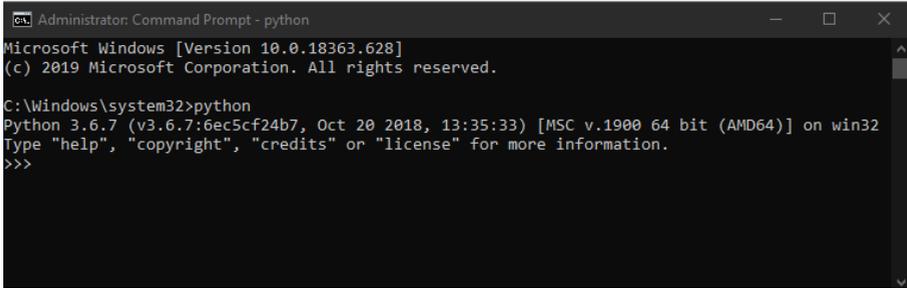

اعلان "«««" بیانگر این است که نصب با پایتون موفقیت‌آمیز بوده و پایتون شروع به‌کار کرده است.

## کتابخانه‌ها و مدیریت آن‌ها در پایتون

کتابخانه‌ها در پایتون تکه کدهایی (شامل مجموعه‌ای از توابع و روش‌ها) با قابلیت استفاده مجدد هستند که به کاربر این امکان را می‌دهند تا بدون اینکه کاربر به نوشتن آن‌ها بپردازید، برای آن‌ها فراهم آورند. گاهی حتی ممکن ساعت‌ها وقت خود را صرف نوشتن یک اسکریپت برای یک کار خاص کنید که به زمان اجرای $O(n^3)$ نیاز دارد. حال آنکه، بسیار محتمل است که کتابخانه‌ای برای آن در پایتون وجود داشته باشد که حاوی همین عملکرد باشد و در $O(n)$ اجرا شود.

به‌عنوان یک مبتدی همیشه این سوال مطرح می‌شود که از چه کتابخانه‌ای برای شروع علم داده استفاده کنم تا پیاده‌سازی آن راحت‌تر و آسان‌تر باشد؟ هزاران کتابخانه در پایتون وجود دارد. با این حال، در ادامه تنها لیستی از کتابخانه‌های مهم پایتون در حوزه علم داده را فهرست کرده و کاربرد هر یک بیان خواهیم کرد.

## NumPy

NumPy یک کتابخانه اساسی برای محاسبات ریاضی و علمی است. این کتابخانه، آرایه‌ها و ماتریس‌های بزرگ و چند بعدی، به‌همراه مجموعه‌ای بزرگ از توابع ریاضی سطح بالا برای کار برروی این آرایه‌ها را پشتیبانی می‌کند. بدون شک، NumPy کتابخانه‌ای است که اگر از علاقه‌مندان به علم داده هستید، باید آن را بیاموزید.

**نصب:** `>>> pip install numpy`

## Keras

Keras یک کتابخانه منبع باز پایتون است که به‌طور گسترده برای آموزش مدل‌های یادگیری عمیق استفاده می‌شود. این کتابخانه یک رابط، برای چارچوب TensorFlow فراهم می‌کند و آزمایش‌های سریع با شبکه‌های عصبی عمیق را ممکن می‌سازد. علاوه براین، استفاده از این کتابخانه بسیار ساده است.

**نصب:** `>>> pip install keras`

## TensorFlow

TensorFlow یکی از پرکاربردترین کتابخانه‌های پایتون برای پردازش و مدل‌سازی داده‌ها است و در کنار آن یک کتابخانه مهم یادگیری ماشین در پایتون است. TensorFlow بر اساس گراف‌های جریان داده که دارای گره‌ها و یال‌ها هستند کار می‌کند. از آنجایی که مکانیزم اجرا به‌صورت گراف است، اجرای کد TensorFlow در حین استفاده از GPU بسیار ساده‌تر است.

`>>> pip install tensorflow`

## PyTorch

PyTorch یک چارچوب یادگیری ماشین منبع باز و یادگیری عمیق است که توسط محققین هوش مصنوعی فیس‌بوک توسعه یافته است. در سراسر جهان بسیاری از دانشمندان داده به‌طور گسترده از PyTorch برای پردازش زبان طبیعی و مسائل بینایی رایانه استفاده می‌کنند. دانشمندان داده می‌توانند گراف‌های محاسباتی را به‌صورت پویا از طریق PyTorch طراحی کنند.

**نصب:** `>>> pip install torch torchvision torchaudio`

## Scrapy

Scrapy یکی از محبوب‌ترین کتابخانه‌های پایتون برای استخراج داده‌ها از وب‌سایت‌ها است. این کتابخانه به دریافت داده‌ها از وب‌سایت‌ها به شیوه‌ای کارآمد کمک می‌کند. Scrapy کمک می‌کند تا داده‌های ساختاریافته‌ای را از وب دریافت کرد که بعدا می‌توان از آن‌ها برای مدل یادگیری ماشین خود استفاده کرد.

**نصب:** `>>> pip install Scrapy`

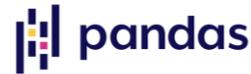

## BeautifulSoup

BeautifulSoup یکی از بهترین و محبوب‌ترین کتابخانه‌های خزنده وب است که می‌تواند برای استخراج داده‌ها از فایل‌های HTML و XML استفاده شود. این ابزار کمک می‌کند اسنادی را که از وب استخراج شده، تمیز کرده و تجزیه کنید. این امر، سبب صرفه‌جویی در ساعت‌ها یا روزها کار برنامه‌نویسان و تحلیل‌گران می‌شود.

**نصب:**     `>>> pip install beautifulsoup4`

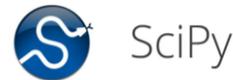

## Pandas

Pandas یکی از کتابخانه‌های مهم علم داده است که برای ایجاد ساختار داده استفاده می‌شود. Pandas انعطاف‌پذیری قدرتمندی در ایجاد ساختار داده‌ها برای علم داده ایجاد می‌کند. چرا که می‌تواند ساختار داده‌های چند بعدی، جدولی، ناهمگن و غیره را ایجاد کند. علاوه‌براین، از این کتابخانه برای دست‌ورزی و تجزیه و تحلیل داده‌ها استفاده می‌شود.

`>>> pip install pandas`     **نصب:**

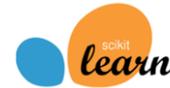

## SciPy

SciPy یکی دیگر از کتابخانه‌های پایتون است که برای حل مسائل علمی و ریاضی استفاده می‌شود و برروی افزونه NumPy ساخته شده است. محاسبات عددی جنبه مهمی از علم داده است و SciPy می‌تواند دانشمندان داده را در حل مسائل پیچیده ریاضی راهنمایی کند. می‌توان گفت SciPy نسخه پیشرفته NumPy است که دارای ویژگی های اضافی مانند نسخه کامل جبرخطی است. SciPy سریع و قدرت محاسبه بالایی دارد.

**نصب:**     `>>> pip install scipy`

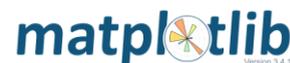

## Scikit-Learn

دانشمندان داده از Scikit-Learn برای مدل‌سازی آماری داده‌ها شامل طبقه‌بندی، کاهش ابعاد، خوشه‌بندی و رگرسیون استفاده می‌کنند. Scikit-Learn بر اساس کتابخانه‌های NumPy و Matplotlib ساخته شده است. کاهش ابعاد داده‌ها یکی از مفیدترین قابلیت‌های Scikit-Learn است. چراکه داده‌های حاصل‌شده پیچیدگی کمتری خواهند داشت.

`>>> pip install scikit-learn`     **نصب:**

## Matplotlib

ترسیم نمودار یکی از مراحله‌های اساسی در طول تجزیه و تحلیل و مدیریت داده‌ها است. Matplotlib یکی از رایج‌ترین کتابخانه‌ها در جامعه پایتون برای ترسیم و مصورسازی داده‌های ایستا، متحرک و تعاملی است.

**نصب:**     `>>> pip install matplotlib`

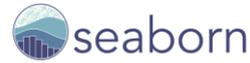

**Seaborn**

Seaborn یک کتابخانه پایتون است که بر اساس Matplotlib ساخته شده است و به‌طور گسترده‌ای برای مصورسازی داده‌ها استفاده می‌شود. دانشمندان داده با استفاده از این کتابخانه می‌توانند نقشه‌های حرارتی را ایجاد کنند. تعداد گزینه‌های ارائه شده توسط Seaborn برای مصورسازی داده‌ها بسیار زیاد است.

**نصب:**
```
>>> pip install seaborn
```

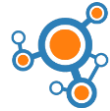

**NetworkX**

NetworkX یک کتابخانه پایتون برای ایجاد، دستورزی، مطالعه ساختار و تجزیه و تحلیل شبکه‌های پیچیده بزرگ است. NetworkX ، علاوه بر مصورسازی‌های دو بعدی و سه بعدی بسیار خوب، بسیاری از معیارها و الگوریتم‌های استاندارد گراف در اختیار کاربر قرار می‌دهد، مانند کوتاه‌ترین مسیر، مرکزیت، رتبه‌بندی صفحه و غیره.

**نصب:**
```
>>> pip install networkx
```

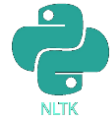

**NLTK**

مجموعه ابزار زبان طبیعی، که به اختصار NLTK نامیده می‌شود، یک کتابخانه مهم پایتون است که توسط دانشمندان داده برای کارهای مختلف مربوط به پردازش زبان طبیعی همانند، برچسب‌گذاری متن، نشانه‌گذاری، ریشه‌یابی و غیره استفاده می‌شود.

**نصب:**
```
>>> pip install nltk
```

## نصب، حذف و بروزرسانی کتابخانه‌ها

برای مدیریت کتابخانه‌های پایتون باید از Pip استفاده کنید. Pip یک ابزار ضروری است که به شما امکان می‌دهد بسته‌های مورد نیاز خود را بارگیری، بروزرسانی و حذف کنید. علاوه‌براین با استفاده از آن می‌توان وابستگی‌های مناسب و سازگاری بین نسخه‌ها را بررسی کنید.

نصب یک کتابخانه با استفاده از Pip در خط فرمان ویندوز صورت می‌گیرد. برای مثال فرض کنید می‌خواهیم کتابخانه NumPy را نصب کنیم. مراحل زیر نحوه نصب این کتابخانه را نشان می‌دهد:

- ابتدا کلیدهای "Win+R" را فشار دهید تا کادر اعلان باز شود و سپس در کادر باز شده، "cmd" را وارد کنید. سپس دستور زیر را در خط فرمان وارد کنید:

```
> pip install numpy
```

```
Administrator: Command Prompt                                    —    □    ✕
Microsoft Windows [Version 10.0.18363.628]
(c) 2019 Microsoft Corporation. All rights reserved.

C:\Windows\system32>pip install numpy
```

▪ برای اطمینان از نصب کتابخانه، از خط فرمان پایتون را اجراکرده و دستور زیر را بنویسید:

```
>>> import numpy
```

▪ اگر کتابخانه به‌درستی نصب شده باشد پیغامی مشاهده نمی‌شود. در صورتی کتابخانه در رایانه شما نصب نشده باشد با اجرای دستور فوق، این پیغام را مشاهده خواهید کرد:

```
Traceback (most recent call last):
  File "<stdin>", line 1, in <module>
ImportError: No module named numpy
```

برای حذف یک کتابخانه (برای مثال کتابخانه numpy) از دستور زیر استفاده می‌شود:

```
> pip uninstall numpy
```

گاهی اوقات، شما در موقعیتی قرار می‌گیرید که مجبورید یک کتابخانه را ارتقا دهید. چراکه نصب یک کتابخانه دیگر نیاز به نسخه جدیدتری از کتابخانه نصب شده در رایانه شما دارد و حتی شاید قصد داشته باشید از مزایای نسخه بروزشده که دارای ویژگی‌های اضافی است، بهره‌مند شوید. از همین‌رو، برای بروزرسانی یک کتابخانه (برای مثال کتابخانه numpy) در خط فرمان دستور زیر را اجرا کنید:

```
> pip install --upgrade numpy
```

برای نمایش نسخه نصب‌شده یک کتابخانه می‌توان به این صورت عمل کرد:

```
>>> import numpy
>>> numpy.__version__
'1.19.2'
```

# جوپیتر نوت‌بوک

جوپیتر نوت‌بوک یک ابزار فوق‌العاده قدرتمند برای توسعه و ارائه پروژه‌های علم داده به‌صورت تعاملی است که می‌تواند علاوه بر اجرای کد، شامل متن، تصویر، صدا و یا ویدیو باشد. یک نوت‌بوک، کد و خروجی آن را با مصورسازی، متن روایی، معادلات ریاضی و سایر رسانه‌ها در قالب یک سند واحد ترکیب می‌کند. به عبارت دیگر، یک نوت‌بوک، یک سند واحد است که در آن می‌توانید کد را اجرا کنید، خروجی را نمایش دهید و همچنین توضیحات، فرمول‌ها، نمودارها را اضافه، تاکار خود را شفاف‌تر، قابل فهم، تکرارپذیر و قابل اشتراک‌گذاری کنید. سایر بسترهای مشابه، به‌عنوان مثال Spyder پنجره‌های متعددی را در اختیار کاربران قرار می‌دهند که سبب پیچیدگی می‌شوند. نوت بوک‌های جوپیتر با دادن تنها یک پنجره به کاربر که در آن قطعه کدهای اجراشده و خروجی‌های آن‌ها بصورت داخلی نمایش داده می‌شوند، به ایجاد یک رابط کاربری قوی و کارآمد تبدیل شده است. این امر، به کاربران این امکان را می‌دهد تا کد را به‌طور موثر توسعه دهند و همچنین بتوانند به‌عنوان مرجع به کارهای قبلی نگاه کرده و حتی تغییراتی را در آن‌ها ایجاد کنند. در ادامه این بخش، نحوه نصب و استفاده از جوپیتر نوت‌بوک را برای پروژه‌های علم داده آموزش خواهیم داد.

## نصب جوپیتر

برای نصب جوپیتر نوت‌بوک، لازم است پایتون را از قبل نصب کرده باشید. حتی اگر قصد داشته باشید از جوپیتر برای سایر زبان‌های برنامه‌نویسی استفاده کنید، پایتون ستون اصلی جوپیتر است. برای نصب جوپیتر کافی است در خط فرمان ویندوز دستور زیر را بنویسید:

```
> pip install jupyter
```

## اجرای جوپیتر و ایجاد یک نوت‌بوک جدید

برای اجرای جوپیتر خط فرمان را باز کرده و دستور زیر را در آن بنویسید:

```
> jupyter notebook
```

پس از اجرای دستور فوق، مرورگر وب پیش‌فرض شما با جوپیتر راه‌اندازی می‌شود. هنگام راه‌اندازی جوپیتر نوت‌بوک به دایرکتوری خط فرمان توجه فرمایید، چرا که این دایرکتوری به فهرست اصلی تبدیل می‌شود که بلافاصله در جوپیتر نوت‌بوک ظاهر می‌شود و تنها به پرونده‌ها

و زیردایرکتوری‌هایی موجود در آن دسترسی خواهید داشت. با اجرای دستور jupyter notebook با صفحه‌ای همانند زیر روبه‌رو می‌شوید:

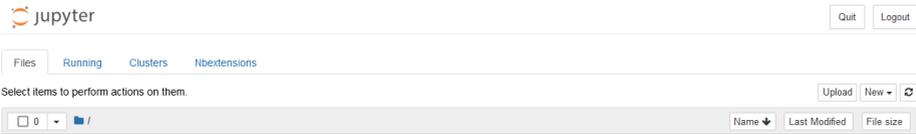

با این حال، این صفحه هنوز یک نوت‌بوک نیست و تنها میزکار جوپیتر است که برای مدیریت نوت‌بوک‌های جوپیتر شما طراحی شده است و آن را به‌عنوان راه‌اندازی برای پیگردی[1]، ویرایش و ایجاد نوت‌بوک‌های خود در نظر بگیرید. نوت‌بوک‌ها و میزکار جوپیتر مبتنی‌بر مرورگر است و جوپیتر یک سرور محلی پایتون راه‌اندازی می‌کند تا این برنامه‌ها را به مرورگر وب شما ارتباط دهد.

برای ایجاد یک نوت‌بوک جدید به به دایرکتوری که قصد دارید اولین نوت‌بوک خود را در ایجاد کنید بروید و بر روی دکمه کشویی "New" که در قسمت بالای میزکار در سمت راست است کلیک کرده و گزینه "Python 3"را انتخاب کنید:

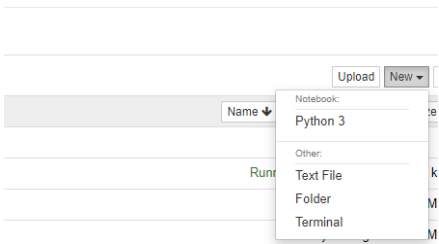

پس از آن، اولین نوت‌بوک شما در یک برگه جدید[2] همانند تصویر زیر باز می‌شود:

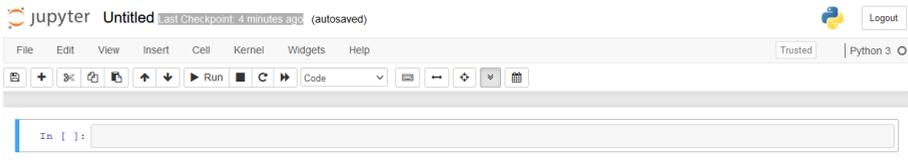

اگر به میزکار جوپیتر بازگردید، فایل جدید Untitled.ipynb را مشاهده خواهید کرد و باید متن سبز رنگی را مشاهده کنید که به شما می‌گوید نوت‌بوک شما در حال اجرا است.

---

[1] exploring

[2] new tab

# کار در جوپیتر نوت‌بوک

یک نوت‌بوک از سلول‌ها تشکیل شده است؛ جعبه‌هایی که حاوی کد یا متن قابل خواندن برای انسان هستند. هر یک این سلول‌ها دارای یک نوع[1] می‌توان آن را از گزینه‌های کشویی منو انتخاب کنید:

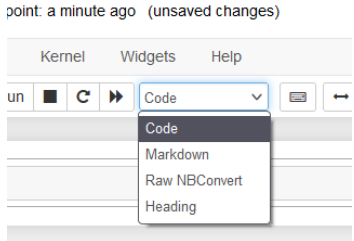

گزینه پیش‌فرض سلول‌ها "code" است. برای ایجاد سلول با قابلیت خواندن برای انسان (متن) باید از نوع سلول "Markdown" استفاده شود و براساس قراردادهای قالب‌بندی Markdown نوشته شود.

بیایید نحوه اجرای یک سلول را با یک مثال کلاسیک آزمایش کنیم. (!'print('Hello World را در یک سلول تایپ کنید و برروی دکمه ▶ Run در نوار ابزار بالا کلیک کنید یا دکمه‌های Ctrl+Enter را فشار دهید. نتیجه آن به این صورت خواهد بود:

Markdown یک زبان نشانه‌گذاری سبک برای قالب‌بندی متن ساده است. نحو[2] آن دارای تطابق با برچسب‌های[3] HTML است. اصول اساسی آن را با یک مثال سریع پوشش می‌دهیم، با قرار دادن متنی همانند تصویر مشاهده شده در صفحه بعدی در یک سلول که پیشتر نوع آن Markdown را انتخاب کرده‌اید، نتیجه آن را پس از اجرا در تصویر بعدی مشاهده می‌کنید. البته باید ذکر شود که پس از اجرای آن، این خروجی نمایش داده شده، همان سلول را تبدیل به متن می‌کند.

---

[1] type

[2] syntax

[3] tags

```
# This is a level 1 heading
## This is a level 2 heading
This is some plain text that forms a paragraph. Add emphasis via **bold** and __bold__, or *italic* and _italic_.

Paragraphs must be separated by an empty line.

* Sometimes we want to include lists.
* Which can be bulleted using asterisks.

1. Lists can also be numbered.
2. If we want an ordered list.

[It is possible to include hyperlinks](https://www.example.com)
```

# This is a level 1 heading

## This is a level 2 heading

This is some plain text that forms a paragraph. Add emphasis via **bold** and **bold**, or *italic* and *italic*.

Paragraphs must be separated by an empty line.

- Sometimes we want to include lists.
- Which can be bulleted using asterisks.

1. Lists can also be numbered.
2. If we want an ordered list.

It is possible to include hyperlinks

---

برای ایجاد یک سلول جدید در جوپیتر برروی دکمه  کلیک کنید و یا از طریق میانبر صفحه کلید، کلیدهای esc+b را فشار دهید. با این کار یک سلول جدید (به‌طور پیش‌فرض از نوع کد) در زیر سلولی که در حال حاضر انتخاب شده است، ایجاد می‌شود.

## برنامه‌نویسی در پایتون

## نحو در زبان پایتون

پایتون در ابتدا به‌عنوان یک زبان آموزشی توسعه داده شد، اما سهولت استفاده و نحو تمیز باعث شده است تا توسط مبتدیان و متخصصان مورد استقبال قرار گیرد. نحو تمیز پایتون باعث شده است که برخی آن را "شبه‌کد اجرایی"[1] بنامند. اغلب خواندن و فهمیدن یک خط از پایتون بسیار راحت‌تر از خواندن یک خط مشابه به‌عنوان مثال در زبان C است.

نحو در زبان برنامه‌نویسی به ساختار یک زبان اشاره دارد. به‌عبارت دیگر، نحو مجموعه‌ای از قوانینی است که نحوه نوشتن برنامه‌نویسی در یک زبان را مشخص می‌کند.

---

[1] executable pseudocode

## کلمات کلیدی

کلمات کلیدی برخی از کلمات رزرو شده و محفوظ در پایتون هستند که معانی خاصی دارند. از کلمات کلیدی برای تعریف دستور نحو و ساختار زبان استفاده میشود. کلمه کلیدی را نمیتوان به عنوان شناسه، تابع و نام متغیر استفاده کرد. کلمات کلیدی در پایتون به حروف کوچک و بزرگ حساس هستند، بنابراین باید آنها را همانطور که هستند، نوشت. همه کلمات کلیدی در پایتون به جز True، False و None با حروف کوچک نوشته میشوند. کلمات کلیدی ممکن است در نسخههای مختلف پایتون تغییر کند. برخی از موارد اضافی ممکن است اضافه یا برخی حذف شوند. همیشه میتوانید لیست کلمات کلیدی نسخه فعلی خود را با تایپ کردن دستورات زیر دریافت کنید:

```
In [1]:   import keyword
          keyword.kwlist
Out [1]:  ['False','None','True','and', 'as','assert','async','await',
          'break','class','continue','def','del','elif','else','except',
          'finally','for','from','global','if','import', 'in','is', 'lambda',
          'nonlocal', 'not','or', 'pass', 'raise', 'return', 'try', 'while',
          'with', 'yield']
```

## شناسه

شناسه نامی است که ما برای شناسایی یک متغیر، تابع، کلاس، ماژول یا شی میگذاریم. این امر به تمایز یک موجودیت از دیگری کمک میکند.

## قوانین نوشتن شناسه

برخی قوانین برای نوشتن شناسهها وجود دارد. اول از همه باید بدانیم که پایتون به حروف کوچک و بزرگ حساس است. این بدان معناست که Name و name دو شناسه متفاوت در پایتون هستند. در زیر چند قانون برای نوشتن شناسه در پایتون آورده فهرست شده است:

۱. شناسه ها میتوانند ترکیبی از حروف کوچک (a تا z) یا بزرگ (A تا Z) یا ارقام (۰ تا ۹) یا یک زیرینخط[1] ( _ ) باشند. نامهایی مانند myPython، my_Python و var_1 همگی معتبر هستند.

۲. شناسه نمیتواند با رقم شروع شود.

۳. از نمادهای خاصی همانند !، @، #، $، % و غیره بهعنوان شناسه نمیتوان استفاده کرد.

۴. شناسه میتواند هر طولی داشته باشد.

---

[1] underscore

**متغیر**

متغیر مکان نام‌گذاری شده‌ای است که برای ذخیره داده‌ها در حافظه استفاده می‌شود. این بدان معناست که هنگام ایجاد یک متغیر، مقداری از فضا در حافظه اشغال می‌کنید. به هر متغیر یک نام اختصاص داده می‌شود تا بتوان آن را از دیگر متغیرها شناسایی و به آن دسترسی پیدا کرد.

**تخصیص مقدار به متغیرها**

ایجاد متغیرها در پایتون ساده است، تنها باید نام متغیر را در سمت چپ = و مقدار متغیر را در سمت راست آن بنویسید:

```
In  [1]: num = 5
In  [2]: str = "Python"
```

لازم نیست نوع متغیر را به‌صراحت ذکر کنید، پایتون نوع را بر اساس مقداری که ما اختصاص می‌دهیم آن را استنباط می‌کند.

**تخصیص چندگانه**

پایتون به ما این امکان را می‌دهد که یک مقدار را به چند متغیر در یک دستور اختصاص دهیم که به آن تخصیص چندگانه گفته می‌شود. ما می‌توانیم تخصیص چندگانه را به دو صورت اعمال کنیم، یا با اختصاص یک مقدار واحد به چند متغیر یا اختصاص چند مقدار به چند متغیر. به مثال‌های زیر توجه کنید:

```
In  [3]:  a=b=c=20
          print("a:",a)
          print("b:",b)
          print("c:",c)
Out [3]:  a: 20
          b: 20
          c: 20
In  [4]:  a,b,c=1,2.54,"python"
          print("a:",a)
          print("b:",b)
          print("c:",c)
Out [4]:  a: 1
          b: 2.54
          c: python
```

# انواع داده

در برنامه‌نویسی، نوع داده یک مفهوم مهم است. متغیرها می‌توانند داده‌های با نوع‌های مختلف را ذخیره کنند و نوع‌های مختلف می‌توانند کارهای متفاوتی انجام دهند. از آن‌جایی که همه چیز در برنامه‌نویسی پایتون یک شی است، انواع داده‌ها در واقع کلاس هستند و متغیرها نمونه‌های (شی) این کلاس‌ها هستند. در این بخش تنها به بررسی نوع داده **عددی** و نوع داده **رشته** در پایتون پرداخته می‌شود و انواع دیگر در بخش داده‌ساختارها تشریح می‌شوند.

## عددی

در پایتون، نوع داده‌های عددی به داده‌هایی اشاره دارد که دارای ارزش عددی هستند. مقدار عددی می‌تواند عدد صحیح، عدد ممیز شناور و یا حتی اعداد مختلط باشد. این مقادیر به عنوان کلاس int، float و complex در پایتون تعریف می‌شوند. برای تعیین نوع داده از تابع ()type استفاده می‌شود.

```
In [1]:    a = 6
           b = 7.0
           c = 4 + 3j
           print("Type of a: ", type(a))
           print("Type of b: ", type(b))
           print("Type of c: ", type(c))
Out [1]:   Type of a:  <class 'int'>
           Type of b:  <class 'float'>
           Type of c:  <class 'complex'>
```

## رشته‌ها

در پایتون، رشته‌ها آرایه‌هایی از بایت‌ها هستند که کاراکترهای یونیکد را نشان می‌دهند. یک رشته مجموعه‌ای از یک یا چند کاراکتر است که در داخل " و یا داخل یک "" قرار می‌گیرد. در پایتون هیچ نوع داده کاراکتری وجود ندارد، یک کاراکتر یک رشته با طول یک است.

```
In [2]:    Str_1 = 'Python Data Types'
           Str_1
Out [2]:   'Python Data Types'
In [3]:    Str_2 = "data science"
           Str_2
Out [3]:   "data science"
In [3]:    type(Str_2)
Out [3]:   str
In [3]:    type("c")
Out [3]:   str
```

# عملگرها

از عملگرهای پایتون به‌طور کلی برای انجام عملیات برروی مقادیر و متغیرها استفاده می‌شود. آن‌ها نمادهای استانداردی هستند که به‌منظور انجام عملیات منطقی، حسابی و مقایسه‌ای مورد استفاده قرار می‌گیرند.

## عملگرهای حسابی

عملگرهای حسابی برای انجام عملیات ریاضی همانند جمع، تفریق، ضرب و تقسیم استفاده می‌شوند.

| عملگر | شرح عملکرد | نحو |
|---|---|---|
| + | جمع | x + y |
| − | تفریق | x − y |
| * | ضرب | x * y |
| / | تقسیم (شناور) | x / y |
| // | تقسیم (کف) | x // y |
| % | باقیمانده | x % y |
| ** | توان | x ** y |

In [1]:
```python
x=10
y=3
add = x+y
sub = x - y
mul = x * y
div1 = x / y
div2 = x // y
mod = x % y
p = x ** y
print("Addition:",add)
print("Subtraction:",sub)
print("Multiplication:",mul)
print("Division(float):",div1)
print("Division(floor):",div2)
print("Modulo:",mod)
print("Power:",p)
```

Out [1]:
```
Addition: 13
Subtraction: 7
Multiplication: 30
Division(float): 3.3333333333333335
Division(floor): 3
Modulo: 1
Power: 1000
```

## عملگرهای منطقی

عملگرهای منطقی (and، or و not) برروی عبارات منطقی اعمال می‌شوند و نتیجه آن‌ها True
و یا False است. از این عملگرها برای شرط‌های پیچیده استفاده می‌شود.

| عملگر | شرح عملکرد | نحو |
|---|---|---|
| **and** | در صورتی True که هر دو عملوند True باشد | x and y |
| **or** | در صورتی True که یکی از عملوندها True باشد | x or y |
| **not** | در صورتی True که عملوند False باشد | not x |

| In [1]: | `X = True`<br>`Y = False`<br>`print(X and Y)`<br>`print(X or Y)`<br>`print(not X)` |
|---|---|
| Out [1]: | False<br>True<br>False |

## عملگرهای مقایسه‌ای

این عملگرها مقادیر دو طرف خود را مقایسه کرده و رابطه بین آن‌ها را تعیین می‌کنند. عملگرهای
مقایسه‌ای را عملگرهای رابطه‌ای نیز می‌نامند.

| عملگر | شرح عملکرد | نحو |
|---|---|---|
| > | بزرگتر از: زمانی True است که عملوند چپ بزرگتر از عملوند راست باشد | x > y |
| < | کوچکتر از: زمانی True است که عملوند چپ کوچکتر از عملوند راست باشد | x < y |
| == | برابر است با: True اگر هر دو عملوند برابر باشند. | x == y |
| != | نابرابر: True اگر هر دو عملوند برابر نباشند. | x != y |
| >= | بزرگتر مساوی: اگر عملوند سمت چپ بزرگتر یا مساوی عملوند راست باشد | x >= y |
| <= | کوچکتر مساوی: اگر عملوند سمت چپ کوچکتر یا مساوی عملوند راست باشد | x <= y |

| In [1]: | `a = 11`<br>`b = 26`<br>`print(a > b)`<br>`print(a < b)`<br>`print(a == b)`<br>`print(a != b)`<br>`print(a >= b)`<br>`print(a <= b)` |
|---|---|
| Out [1]: | False<br>True<br>False |

```
True
False
True
```

## داده ساختارها

سازماندهی، مدیریت و ذخیره داده‌ها بسیار مهم است، چراکه دسترسی آسان‌تر و تغییرات کارآمد را امکان‌پذیر می‌سازد. داده ساختارها به شما این امکان را می‌دهد که داده‌های خود را به‌گونه‌ای سازماندهی کنید تا بتوانید مجموعه‌ای از داده‌ها را ذخیره کرده، آن‌ها را بهم مرتبط ساخته و براساس آن عملیاتی را انجام دهید. داده ساختارها بلوک‌های اساسی برای حل مسائل کارآمد در دنیای واقعی هستند. آن‌ها ابزارهای اثبات‌شده و بهینه‌سازی شده‌ای هستند که به شما چارچوبی آسان برای سازماندهی برنامه‌ها را می‌دهند.

پایتون به‌طور ضمنی از داده ساختارهایی پشتیبانی می‌کند که به شما امکان ذخیره و دسترسی به داده‌ها را می‌دهد. در زبان برنامه‌نویسی پایتون در مجموع چهار ساختار داده داخلی وجود دارد. این داده ساختارها شامل **لیست**، **تاپل**، **دیکشنری** و **مجموعه** می‌شود. داده ساختارهای پایتون ساده اما بسیار قدرتمند هستند. تسلط بر استفاده از آن‌ها بخش مهمی از تبدیل شدن به یک برنامه‌نویس ماهر پایتون است. علاوه‌بر این، هنگام حل مسائل برنامه‌نویسی در دنیای واقعی، کارفرمایان و استخدام‌کنندگان به‌دنبال زمان اجرا و بهره‌وری منابع هستند. دانستن اینکه کدام داده ساختار به بهترین نحو با راه حل فعلی مطابقت دارد، عملکرد برنامه را افزایش داده و زمان لازم برای ساخت آن را کاهش می‌دهد. به‌همین دلیل، اکثر شرکت‌های برتر نیاز به درک قوی از داده ساختارها دارند و آن را در مصاحبه‌های کدنویسی خود به شدت آزمایش می‌کنند.

<div dir="rtl">

**داده ساختارها به موارد زیر کمک می‌کند:**

- مجموعه داده‌ها را مدیریت و استفاده کنید.
- جستجوی سریع داده‌های خاص از پایگاه داده.
- ایجاد ارتباطات سلسله مراتبی یا رابطه‌ای بین نقاط داده
- ساده‌سازی و سرعت بخشیدن به پردازش داده‌ها

</div>

## لیست

یک لیست به‌عنوان مجموعه‌ای مرتب از عناصر (اقلام) تعریف می‌شود. به عبارت دیگر، یک لیست دنباله‌ای از عناصر را در خود نگه می‌دارد. ترتیب عناصر یک ویژگی ذاتی است که در طول عمر لیست ثابت می‌ماند. از آنجایی که همه چیز در پایتون یک شی محسوب می‌شود، ایجاد یک لیست در اصل ایجاد یک شی از یک نوع خاص است. هنگام ایجاد یک لیست، همه عناصر موجود در لیست باید در داخل [] قرار گرفته و با کاما از یکدیگر جدا شوند تا به پایتون

اطلاع داده شود که لیستی ایجاد شده است. یک لیست را می‌توان به‌صورت زیر در پایتون ایجاد کرد:

```
List_A = [item 1, item 2, item 3, ...., item n]
```

```
In  [1]: my_list = [1, 2, 3, 4]
In  [2]: my_list
Out [2]: [1, 2, 3, 4]
```

اگر هیچ عنصری را در داخل براکت قرار ندهید، یک لیست خالی به‌عنوان خروجی دریافت خواهید کرد:

```
In  [1]: my_list1 = []
In  [2]: my_list1
Out [2]: []
```

هر لیستی می‌تواند تعدادی عنصر با انواع مختلفی از داده‌ها باشد:

```
In  [3]: my_list = [1, 'example', 5.45]
In  [4]: my_list
Out [4]: [1, 'example', 5.45]
```

علاوه‌بر این، یک لیست می‌تواند یک لیست دیگر به عنوان یک عنصر داشته باشد. چنین لیستی به‌عنوان لیست تو در تو شناخته می‌شود:

```
In  [5]: my_list1 = [1.56, 'python']
In  [6]: my_list2 = ['example', 1]
In  [7]: my_list = [5, my_list1 ,'data scientist', my_list2]
In  [8]: my_list
Out [6]: [5, [1.56, 'python'], 'data scientist', ['example', 1]]
```

**افزودن عنصر**

افزودن عنصر به یک لیست در پایتون با استفاده از توابع ()append، ()extend و ()insert قابل دستیابی است:

- **تابع ()append** تمام عناصر منتقل‌شده به لیست را به‌عنوان یک عنصر واحد اضافه می‌کند.
- **تابع ()extend** عناصر را یکی یکی به لیست اضافه می‌کند.
- **تابع ()insert** یک عنصر را در شاخص (اندیس) مشخص به فهرست اضافه می‌کند.

```
In  [7]: my_list = [7, 2, 1]
In  [8]: my_list
```

```
Out [8]: [7, 2, 1]
In  [9]: my_list.append([44, 15,'python'])
In  [10]: my_list
Out [10]: [7, 2, 1, [44, 15, 'python']]
In  [11]: my_list.extend(['example',2])
In  [12]: my_list
Out [12]: [7, 2, 1, [44, 15, 'python'], 'example', 2]
In  [13]: my_list.insert(1, 'insert_example1')
In  [14]: my_list.insert(6, 'insert_example2')
In  [15]: my_list
Out [15]: [7, 'insert_e1', 2, 1, [44, 15, 'python'], 'example',
'insert_e2', 2]
```

**حذف عنصر**

حذف عنصر از یک لیست در پایتون با استفاده از توابع ()clear، ()pop، ()remove و ()del قابل دستیابی است:

- با استفاده از تابع ()clear تمام عناصر یک لیست حذف می‌شوند.
- تابع ()pop یک عنصر را براساس شاخص حذف می‌کند و مقدار آن را نیز در خروجی نمایش می‌دهد.
- با استفاده از تابع ()remove می‌توان یک عنصر را براساس مقدار آن حذف کرد.
- با استفاده از تابع ()del می‌توان عناصر یک آرایه را براساس شاخص حذف کرد. اولین شاخص ۰ و آخرین شاخص ۱ ــ است.

```
In  [16]: my_list = [0, 1, 2, 3, 4, 5, 6, 7, 8, 9]
In  [17]: my_list
Out [17]: [0, 1, 2, 3, 4, 5, 6, 7, 8, 9]
In  [18]: my_list.clear()
Out [18]: []
In  [19]: my_list = [8, 2, 3, 7, 9, 1]
In  [20]: my_list.pop(0)
Out [20]: 8
In  [21]: my_list.pop(4)
Out [21]: 1
In  [22]: my_list = [12, 1, 5, 2, 4]
In  [23]: my_list.pop(-3)
```

Out [23]: 5

In  [24]: my_list = ['Ali', 'Mohammad', 'Milad', 1, 5.69]

In  [25]: my_list.remove('Milad')

In  [25]: my_list

Out [25]: ['Ali', 'Mohammad', 1, 5.69]

In  [26]: my_list.remove(1)

In  [27]: my_list

Out [27]: ['Ali', 'Mohammad', 5.69]

In  [28]: my_list = [4, 7, 5, 1, 4]

In  [29]: del my_list[0]

In  [30]: my_list

Out [30]: [7, 5, 1, 4]

In  [31]: del my_list[-1]

In  [32]: my_list

Out [32]: [7, 5, 1]

## توابع دیگر

چندین تابع دیگر وجود دارد که هنگام کار با لیست‌ها می‌توان از آن‌ها استفاده کرد:

▪ تابع ()**len** طول یک لیست را باز می‌گرداند.

▪ تابع ()**index** شاخص یا همان اندیس یک عنصر را بر می‌گرداند (توجه: اگر یک عنصر چندین مرتبه در لیست ظاهر شده باشد، اولین شاخصی که مطابقت داده شود باز می‌گردد).

▪ با استفاده از تابع ()**sort** به‌صورت صعودی مرتب می‌شوند.

In  [33]: my_list1 = [4, 7, 5, 1, 4, 12]

In  [34]: len(my_list)

Out [34]: 6

In  [35]: my_list.index(5)

Out [35]: 2

In  [36]: my_list.sort()

In  [37]: my_list

Out [37]: [1, 4, 4, 5, 7, 12]

**تاپل**

یک تاپل یا چندتایی دنباله‌ای ثابت و تغییرناپذیر از شی‌ها در پایتون است. مهم‌ترین تفاوت آن‌ها با لیست‌ها همین تغییرناپذیری است. در حالی‌که لیست‌ها تغییرپذیر هستند، تاپل‌ها این ویژگی را ندارند.

ساده‌ترین راه برای ایجاد یک تاپل به‌صورت زیر است:

```
tuple_A = item 1, item 2, item 3,…, item n
```

استفاده از پرانتز در ایجاد تاپل اختیاری است، اما توصیه می‌شود که بین شروع و پایان تاپل تمایز قائل شوید:

```
tuple_A = (item 1, item 2, item 3,…, item n)
```

```
In  [1]: my_tuple = (1, 2, 3)
In  [2]: my_tuple
Out [2]: (1, 2, 3)
```

با فراخوانی tuple می‌توانید هر دنباله‌ای را به چند تاپل تبدیل کنید:

```
In   [3]: tuple([1, 3, 8])
Out [3]: (1, 3, 8)
In   [4]: tup_A = tuple('Python')
In   [5]: tup_A
Out [5]: ('P', 'y', 't', 'h', 'o', 'n')
```

**توابع**

چندین تابع وجود دارد که هنگام کار با تاپل‌ها می‌توان از آن‌ها استفاده کرد:

- **تابع ()len طول یک تاپل را باز می‌گرداند.**
- **تابع ()index شاخص یا همان اندیس یک عنصر را بر می‌گرداند.**
- **تابع ()max بزرگ‌ترین مقدار یک تاپل را بر می‌گرداند.**
- **تابع ()min کوچک‌ترین مقدار یک تاپل را بر می‌گرداند.**

```
In   [6]: tup_A = tuple('Python')
In   [7]: tup_A.index('y')
```

```
Out [7]: 1
In   [8]: my_tuple = (8, 1, 4, 5, 3)
In   [9]: max (my_tuple)
Out [9]: 8
In   [9]: min (my_tuple)
Out [9]: 1
```

هنگام نوشتن یک تاپل تنها با یک عنصر واحد، باید از کاما بعد از عنصر استفاده شود. یک تاپل با یک عنصر واحد می‌تواند به صورت زیر ایجاد شود:

tuple_A = (item 1,)

برای ایجاد یک تاپل خالی، کاربر باید یک جفت پرانتز خالی به‌صورت زیر ایجاد کند:

tuple_A = ()

```
In   [10]: Empty_tuple= ()
In   [11]: Empty_tuple
Out [11]: ()
```

**چرا تاپل بر لیست ترجیح داده می‌شود؟**
- تاپل‌ها سریع‌تر از لیست‌ها هستند. تاپل در یک بلوک واحد از حافظه ذخیره می‌شود. تاپل‌ها تغییرناپذیر هستند، بنابراین برای ذخیره عناصر جدید به فضای اضافی احتیاج ندارد.
- هنگامی که کاربر نمی‌خواهد داده‌ها تغییر کنند، تاپل ترجیح داده می‌شود. گاهی اوقات، کاربر می‌خواهد یک شی ایجاد کند که در طول عمر خود دست نخورده باقی بماند. تاپل‌ها تغییرناپذیر هستند، بنابراین می‌توان از آن‌ها برای جلوگیری از افزودن تصادفی، اصلاح یا حذف داده‌ها استفاده کرد.

### دیکشنری

دیکشنری در پایتون یک مجموعه نامرتب از مقادیر است که برخلاف سایر داده ساختارها که تنها یک مقدار را به عنوان یک عنصر نگه می‌دارد، برای ذخیره جفت‌های کلید ـ مقدار استفاده می‌شود. کلید ـ مقدار در دیکشنری برای بهینه‌سازی بیشتر ارائه گردیده است. در دیکشنری کلیدها باید یکتا باشند.

```
my_dictionary= {key 1 : value 1,  key 2 : value 2}
```

```
In  [1]: my_dict = {'First': 'Python', 'Second': 'Julia'}
In  [2]: my_dict
Out [2]: {'First': 'Python', 'Second': 'Julia'}
```

برای تغییر مقادیر دیکشنری، باید این کار را از طریق کلیدها انجام دهید. بنابراین، ابتدا به کلید
دسترسی پیدا کرده و سپس مقدار آن را تغییر دهید:

```
In  [3]: my_dict['Second'] = 'golang'
In  [4]: my_dict
Out [4]: {'First': 'Python', 'Second': 'golang'}
```

برای افزودن مقادیر، کافی است یک جفت کلید ـ مقدار دیگر را مطابق دستور زیر اضافه کنید:

```
In  [3]: my_dict['Third'] = 'Rust'
In  [4]: my_dict
Out [4]: {'First': 'Python', 'Second': 'golang', 'Third': 'Rust'}
```

برای حذف یک مقدار از تابع ()pop استفاده می‌شود (این تابع مقدار حذف شده را نیز بر
می‌گرداند):

```
In  [5]: my_dict.pop('Third')
Out [5]: Rust
In  [6]: my_dict
Out [6]: {'First': 'Python', 'Second': 'golang'}
```

برای پاک کردن کل دیکشنری، از تابع ()clear استفاده می‌شود:

```
In  [7]: my_dict.clear()
In  [8]: my_dict
Out [8]: {}
```

برای دسترسی به یک عنصر تنها کافی است که به کلید آن دسترسی پیدا کنید:

```
In  [9]: my_dict = {'First': 'Python', 'Second': 'Julia'}
In  [10]: my_dict['First']
```

Out [10]: Python

برای نمایش کلیدها از تابع ()keys و نمایش مقادیر از تابع ()values استفاده می‌شود. نمایش جفت کلید ــ مقدار با استفاده از تابع ()items امکان‌پذیر است:

```
In   [11]: my_dict = {'First': 'Python', 'Second': 'Julia'}
In   [12]: my_dict.keys()
Out [12]: dict_keys(['First', 'Second'])
In   [13]: my_dict.values()
Out [13]: dict_values(['Python', 'Julia'])
In   [14]: my_dict.items()
Out [14]: dict_items([('First', 'Python'), ('Second', 'Julia')])
```

### مجموعه

مجموعه به عنوان کلکسیونی از عناصر یکتا تعریف می‌شود که از ترتیب خاصی پیروی نمی‌کنند. مجموعه‌ها زمانی مورد استفاده قرار می‌گیرند که وجود یک شی در مجموعه‌ای از اشیا مهم‌تر از تعداد دفعات ظاهر شدن یا ترتیب اشیا باشد. در مجموعه‌ها، اگر داده‌ها بیش از یک بار تکرار شوند، فقط یک بار در مجموعه وارد می‌شوند. برخلاف تاپل‌ها، مجموعه‌ها قابل تغییر هستند؛ یعنی می‌توان آنها را اصلاح، اضافه، جایگزین و یا حذف کرد. یک مجموعه نمونه را می‌توان به صورت زیر نشان داد:

```
set_a = {"item 1", "item 2", "item 3",....., "item n"}
```

```
In   [1]: my_set = {2, 2, 3, 1, 4, 5, 5, 5}
In   [2]: my_set
Out [2]: {1, 2, 3, 4, 5}
```

برای افزودن یک عنصر می‌توان از تابع ()add استفاده کرد:

```
In   [3]: my_set = {8, 1, 5}
In   [4]: my_set.add(6)
In   [5]: my_set
Out [5]: {1, 5, 6, 8}
```

مجموعه‌ها از عملیات بر روی مجموعه‌های ریاضی همانند اجتماع، اشتراک و غیره پشتیبانی می‌کند. مثال زیر اجتماع دو مجموعه را نشان می‌دهد:

```
In  [3]: a = {1, 2, 3, 4, 5}
In  [4]: b = {6, 4, 5, 1, 3, 8, 7}
In  [5]: a.union(b)
Out [5]: {1, 2, 3, 4, 5, 6, 7, 8}
```

# ساختارهای کنترلی و حلقه‌ها

در زندگی روزمره ما هرروز تصمیماتی می‌گیریم و بر اساس تصمیمات اتخاذ شده، اقدامات بعدی را انجام می‌دهیم. بنابراین همه فعالیت‌های روزمره ما به تصمیماتی که می‌گیریم بستگی دارد. وضعیت مشابهی در زبان برنامه نویسی نیز ایجاد می‌شودکه در آن ما باید برخی تصمیمات را بگیریم و بر اساس آن برنامه اجرا می‌شود. در زبان برنامه‌نویسی این کار توسط ساختارهای کنترلی صورت می‌گیرد. به زبان ساده، کنترل جریان در برنامه‌نویسی، ترتیب انجام عملیات خاص است. بیایید باید یک مثال ساده شروع کنیم، فرض کنید می‌خواهیم اسکریپتی داشته باشیم که تحت برخی از شرایط متفاوت اجرا شود. به عنوان مثال، اگر دمای هوا را ۳ درجه سانتی‌گراد اندازه‌گیری کردیم، "هوا سرد است" را چاپ کند، اما اگر ۲۱ درجه سانتی‌گراد باشد، "هوا گرم است" را چاپ کند. در این حالت، برخی از شرایط بررسی می‌شوند و با توجه به یک شرط، یک کار اجرا می‌شود (یک عبارت خاص چاپ می‌شود).

## دستورات شرطی

در زبان های برنامه‌نویسی، بیشتر اوقات در پروژه‌ها باید جریان اجرای برنامه خود را کنترل کنیم. به‌عبارت دیگر، می‌خواهیم برخی از دستورات را تنها در صورت برآورده شدن شرایط داده شده اجرا کنیم. دستورات شرطی که به عنوان بیانیه‌های تصمیم‌گیری نیز شناخته می‌شوند، برای انجام این کار ایجاد شده‌اند و بسته به درست یا نادرست بودن یک شرط معین عمل می‌کنند. در پایتون می‌توانیم با استفاده از دستورات زیر به تصمیم‌گیری برسیم:

- **دستور if**
- **دستور if-else**
- **دستور elif**
- **دستور if-else تو در تو**

**دستور if**

در دستورات کنترلی، دستور if ساده‌ترین شکل آن است. یک شرط می‌گیرد و به دو صورت درست یا نادرست ارزیابی می‌شود:

```
if condition:
    statement 1
    statement 2
    statement n
```

**مثال:**

```
In  [1]:   num = 3
           if (num < 7):
             print("Num is smaller than 7")
Out [1]:   Num is smaller than 7
In  [2]:   a = 3
           b = 2
           if (a > b):
             print("a is greater than b")
Out [2]:   a is greater than b
```

**دستور** if-else

با استفاده از دستور if-else اگر شرط معینی درست باشد، دستورات موجود در داخل بلوک if اجرا می‌شود و اگر شرط نادرست باشد، بلوک else اجرا می‌شود:

```
if condition:
    statement 1
else:
    statement 2
```

**مثال:**

```
In  [1]:   a = 1
           b = 2
           if (a > b):
             print("a is greater than b")
           else:
             print("b is greater than a")
Out [1]:   b is greater than a
In  [2]:   passing_Score = 70
           my_Score = 59
           if(my_Score >= passing_Score):
             print("Congratulations! You passed the exam")
             print("You are passed in the exam")
           else:
```

```
            print("Sorry! You failed the exam")
Out [2]:    Sorry! You failed the exam
```

**دستور elif**

با کمک دستور elif می‌توانیم تصمیم پیچیده‌ای بگیریم. دستور elif چندین شرط را یک به یک به یک بررسی می‌کند و اگر شرط برآورده شد، آن بلوک از کد را اجرا می‌کند:

```
if condition-1:
    statement 1
elif condition-2:
    stetement 2
elif condition-3:
    stetement 3
    ...
else:
    statement
```

**مثال:**

```
In [1]:    num = -1
           if (num > 0):
            print("Number is positive")
           elif (num < 0):
            print("Number is negative")
           else:
            print("Number is Zero")
Out [1]:   Number is negative
```

**دستور if-else تو در تو**

دستور if-else تو در تو به این معنی است که دستور if یا if-else در داخل بلوک if یا if-else دیگر وجود دارد. این به‌نوبه خود به ما کمک می‌کند تا شرایط متعدد را در یک برنامه معین بررسی کنیم.

```
if conditon_outer:
    if condition_inner:
        statement of inner if
    else:
        statement of inner else:
    statement ot outer if
else:
    Outer else
statement outside if block
```

**مثال:**

```
In [1]:   num = 0
          if (num != 0):
            if (num > 0):
              print("Number is positive")
            else:
              print("Number is negative")
          else:
            print("Number is Zero")
Out [1]:  Number is Zero
```

### حلقه‌ها

به‌عنوان برنامه‌نویس، یکی از اهداف کلی ما نوشتن کد کارآمد است. هر کاری که انجام می‌دهیم باید حول محور ارائه یک تجربه کاربری خوب، کاهش منابع پردازنده و ایجاد برنامه‌هایی با کم‌ترین میزان کد ممکن باشد. یکی از راه‌هایی که می‌توانیم به آن برسیم استفاده از حلقه‌ها است که دو نوع از آن‌ها در پایتون وجود دارد. حلقه‌ها به ما اجازه می‌دهند تا هر زمان که بخواهیم قسمتی از کد را تکرار کنیم؛ مادامی که شرطی را که تعریف کرده‌ایم برآورده شود. حلقه‌ها به ما کمک می‌کنند تا تکرار کد خود را کاهش دهیم، زیرا این امکان را می‌دهد یک عملیات را چندین مرتبه اجرا کنند.

### حلقه for

با استفاده از حلقه for، می‌توان هر دنباله یا متغیر تکرارپذیر را پیمایش کرد. دنباله می‌تواند رشته، لیست، دیکشنری، مجموعه یا تاپل باشد. نحوه استفاده از حلقه for برای تکرار و پیمایش به‌صورت زیر است:

```
for iterator_var in sequence:
    statements(s)
```

**مثال:**

```
In [1]:   n = 4
          for i in range(n):
            print(i)
Out [1]:  0
          1
          2
          3
In [2]:   Str = 'Persian'
          for i in Str:
            print(i)
Out [2]:  P
```

```
        e
        r
        s
        i
        a
        n
In  [3]:  l = ["machine", "learning", "and"]
        for i in l:
            print(i)
Out [3]:  machine
        learning
        and
```

**حلقه while**

در پایتون، از حلقه while برای اجرای یک بلوک از دستورات استفاده می‌شود تا زمانی که
یک شرط مشخص برآورده شود و وقتی شرط نادرست شد، خط بلافاصله پس از حلقه برنامه
اجرا می‌شود. نحوه استفاده از حلقه while به‌صورت زیر است:

```
while expression:
    statement(s)
```

**مثال:**

```
In  [1]:  num = 10
        sum = 0
        i = 1
        while i <= num:
            sum = sum + i
            i = i + 1
        print("Sum of first 10 number is:", sum)
Out [1]:  Sum of first 10 number is: 55
In  [2]:  count = 0
        while (count < 3):
            count = count + 1
            print("Hello")
Out [2]:  Hello
        Hello
        Hello
```

# توابع

از توابع در برنامه‌نویسی برای مجموعه‌ای از دستورالعمل‌هایی که می‌خواهید به‌طور مکرر استفاده
کنید یا به‌دلیل پیچیدگی آن‌ها بهتر است در یک برنامه فرعی دیگر قرار گیرند و در مواقع ضروری

فراخوانی شوند، استفاده می‌شود. توابع به دو دلیل بخش مهمی از هر زبان برنامه‌نویسی هستند. اول، آن‌ها به شما امکان می‌دهند از کدی که نوشته‌اید مجددا استفاده کنید. به عنوان مثال، اگر با پایگاه داده کار می‌کنید، همیشه باید با پایگاه داده ارتباط برقرار کنید و آن را از جدولی که می‌خواهید به آن دسترسی داشته باشید مطلع کنید. با نوشتن یک تابع می‌توانید این کار را با نوشتن یک خط کد در هر برنامه‌ای که نیاز به دسترسی به پایگاه داده دارد انجام دهید. مزیت دیگر برای استفاده از یک تابع برای انجام چنین کاری این است که اگر احتیاج داشتید نوع پایگاه داده مورد استفاده خود را تغییر دهید، یا اگر در منطقی که برای اولین بار تابع را نوشتید نقصی تشخیص دادید، می‌توانید به سادگی یک نسخه واحد از تابع را ویرایش کرده و سایر برنامه‌ها می‌توانند نسخه تصحیح شده را استفاده کنند تا فورا بروز شود.

دلیل دوم استفاده از توابع این است که به شما اجازه می‌دهد تا به‌طور منطقی کارهای فرعی مختلف را که هنگام کار برروی یک برنامه هستید و همیشه نیاز به نوشتن آن‌ها دارید را جدا کنید. در مثال پایگاه داده، شما عموما باید به پایگاه داده متصل شوید و سپس یا پایگاه داده را جستجو کنید یا برخی تغییرات را ایجاد کنید. با نوشتن یک تابع برای اتصال، دومین مورد برای پرس‌وجو و سومی برای بروزرسانی، می‌توانید قسمت اصلی برنامه خود را بسیار مختصر بنویسید. اشکال زدایی چنین برنامه‌ای بسیار ساده‌تر می شود. زیرا، هنگامی که مجموعه‌ای از توابع توسعه و آزمایش شده است، تشخیص این‌که آیا مشکل ایجاد شده از یکی از توابع است یا کدی که آن را فراخوانی می‌کند دشوار نیست.

توابع به تقسیم برنامه‌ها به قطعات کوچکتر کمک می‌کند. همان‌طور که برنامه ما بزرگتر و بزرگتر می‌شود، توابع آن را سازمان یافته‌تر و قابل مدیریت‌تر می‌کنند. علاوه‌براین، از تکرار جلوگیری می‌کند. توابع می‌توانند هم داخلی باشند و هم توسط کاربر تعریف شوند.

## تعریف تابع

چهار مرحله برای تعریف یک تابع در پایتون به‌شرح زیر است:

۱. از کلمه کلیدی def برای اعلام تابع استفاده کنید و پس از آن یک نام برای تابع خود انتخاب کنید.

۲. پارامترها را به تابع اضافه کنید. آن‌ها را در داخل پرانتز قرار دهید و خط خود را با قرار دادن دو نقطه (:) به پایان برسانید.

۳. عبارت‌هایی که تابع باید اجرا کند را اضافه کنید.

۴. اگر تابع باید چیزی در خروجی نمایش دهد، تابع خود را با دستور return خاتمه دهید. بدون دستور return، تابع شما یک شی None را بر می‌گرداند.

نحو آن در پایتون به‌صورت زیر است:

```
def function_name(parameters):
    statement(s)
    return expression
```

**مثال:**

```
In  [1]:  def greet(name):
             print("Hello, " + name + ". Good morning!")
In  [2]:  greet("ali")
Out [2]:  Hello, ali. Good morning!
In  [3]:  def absolute_value(num):
             if num >= 0:
                 return num
             else:
                 return -num
In  [3]:  absolute_value(5)
Out [3]:   5
In  [4]:  absolute_value(-8)
Out [4]:   8
In  [5]:  def evenOdd(x):
             if (x % 2 == 0):
                 print("even")
             else:
                 print("odd")
In  [6]:  evenOdd(5)
Out [6]:  odd
In  [7]:  evenOdd(8)
Out [7]:  even
```

## کار با کتابخانه NumPy

NumPy یک کتابخانه پایتون است که برای کار با آرایه‌ها استفاده می‌شود. دلیل اهمیت آن برای علم داده با پایتون این است که بیشتر کتابخانه‌های موجود در یادگیری ماشین و یادگیری عمیق به NumPy به‌عنوان یکی از بلوک‌های اصلی خود تکیه می‌کنند، چراکه برای آن‌ها سرعت و منابع بسیار مهم است. ممکن است این سال برای شما پیش آید که چرا وقتی لیست‌های پایتون وجود دارند از آرایه‌های NumPy استفاده می‌کنیم؟ در پایتون لیست، هدف آرایه‌ها را ارائه می‌دهد. با این حال، پردازش آن‌ها کند است و کند بودن آن در نحوه ذخیره‌سازی یک شی در حافظه پنهان است. یک شی پایتون در واقع یک اشاره‌گر به یک مکان حافظه است که تمام جزئیات مربوط به شی را مانند بایت و مقدار آن را ذخیره می‌کند. اگرچه این اطلاعات اضافی چیزی است که پایتون را به یک زبان پویا تبدیل می‌کند، اما هزینه‌ای را نیز در پی دارد. برای غلبه

بر این مشکل، از آرایه‌های NumPy استفاده می‌کنیم که فقط شامل عناصر همگن هستند، یعنی عناصری که دارای نوع داده یکسانی هستند. این سبب می‌شود که ذخیره و دست‌ورزی آرایه کارآمدتر باشد. هدف NumPy ارائه یک شی آرایه‌ای است که حداکثر ۵۰ برابر سریع‌تر از لیست‌های سنتی پایتون باشد. آرایه‌های NumPy برخلاف لیست‌ها در یک مکان پیوسته در حافظه ذخیره می‌شوند، بنابراین فرآیندها می‌توانند به‌طور موثر به آن‌ها دسترسی داشته و آن‌ها را دست‌ورزی کنند. این رفتار در علوم رایانه ارجاع محلی[1] نامیده می‌شود. این دلیل اصلی سریع‌تر بودن NumPy از لیست‌ها است. همچنین، با آرایه‌های NumPy می‌توان عملیات عنصری را انجام داد، چیزی که با استفاده از لیست‌های پایتون امکان‌پذیر نیست! به‌همین دلیل است که آرایه‌های NumPy هنگام انجام عملیات ریاضی روی حجم زیادی از داده‌ها بر لیست‌های پایتون ترجیح داده می‌شوند.

NumPy مجموعه داده‌های بزرگ را به‌طور موثر و کارآمد مدیریت می‌کند. به‌عنوان یک دانشمند داده یا به عنوان یک متخصص علم داده، باید در مورد NumPy و نحوه عملکرد آن در پایتون درک کاملی داشته باشیم.

## وارد کردن NumPy

هر زمان که می‌خواهید از یک بسته یا کتابخانه در کد خود استفاده کنید، ابتدا باید آن را در دسترس قرار دهید. برای شروع استفاده از NumPy و همه توابع موجود در NumPy، باید آن را وارد کنید. این کار را می‌توان به‌راحتی با دستور import انجام داد:

```
In [1]:   import numpy as np
```

np مخفف NumPy است که توسط جامعه علم داده استفاده می‌شود. ما NumPy را به np کوتاه می‌کنیم تا در وقت خود صرفه‌جویی کرده و همچنین استاندارد بودن کد را حفظ کنیم تا هرکسی که با کد ما کار می‌کند به‌راحتی آن را بفهمد و اجرا کند.

## ایجاد یک آرایه NumPy

برای ایجاد یک آرایه پایه در NumPy، از متد ()np.array استفاده می‌شود. تنها چیزی که باید در آن جای داده شود، مقادیر آرایه به‌عنوان یک لیست است:

```
In [1]:   np.array([1,2,3,4])
Out [1]:  array([1, 2, 3, 4])
```

---

[1] locality of reference

این آرایه شامل مقادیر صحیح است. می‌توانید نوع داده‌ها را در آرگومان dtype مشخص کنید:

In [2]: np.array([1,2,3,4],dtype=np.float32)

Out [2]: array([1., 2., 3., 4.], dtype=float32)

با استفاده از براکت‌های مربعی ([]) می‌توانیم به عناصر آرایه دسترسی پیدا کنیم. هنگام دسترسی به عناصر، به یاد داشته باشید که نمایه‌سازی در NumPy از ۰ شروع می‌شود. این بدان معنی است که اگر می‌خواهید به اولین عنصر در آرایه خود دسترسی داشته باشید، با استفاده از ۰ به آن دسترسی خواهید داشت:

In [3]: a= np.array([5 , 1, 3, 7])
        a[0]
Out [3]:    5

آرایه‌های NumPy همچنین می‌توانند چند بعدی باشند:

In [4]: a = np.array([[1 , 5, 2], [6, 8, 1], [10, 3, 11]])
        a
Out [4]: array([[ 1,  5,  2],
               [ 6,  8,  1],
               [10,  3, 11]])
In [5]: a[0]

Out [5]: array([1, 5, 2])

In [6]: a[2]

Out [6]: array([10,  3, 11])

In [7]: a[0][0]

Out [7]:   1

## آرایه‌ای از صفرها

NumPy به شما امکان می‌دهد با استفاده از متد ()np.zeros یک آرایه از صفرها ایجاد کنید. تنها کاری که باید انجام دهید این است که شکل[1] آرایه مورد نظر را در آن قرار دهید:

In [1]: np.zeros(7)

Out [1]: array([0., 0., 0., 0., 0., 0., 0.])

---

[1] shape

آرایه قبلی یک آرایه ۱ بعدی است. برای ایجاد یک آرایه دو بعدی به‌صورت زیر عمل کنید:

```
In  [2]:   np.zeros((2,6))

Out [2]:   array([[0., 0., 0., 0., 0., 0.],
                  [0., 0., 0., 0., 0., 0.]])
```

## آرایه‌ای از یک‌ها

همچنین می‌توانید با استفاده از متد ()np.ones یک آرایه از یک‌ها ایجاد کنید:

```
In  [1]:   np.ones(6)

Out [1]:   array([1., 1., 1., 1., 1., 1.])
```

## افزودن، حذف و مرتب‌سازی عناصر

می‌توانید با استفاده از متد ()np.append عناصر را به آرایه خود اضافه کنید:

```
In  [1]:   a = np.array([5, 1, 2, 3, 9, 4, 7])
           a
Out [1]:   array([5, 1, 2, 3, 9, 4, 7])

In  [2]:   np.append(a, [12,2,1])

Out [2]:   array([ 5,  1,  2,  3,  9,  4,  7, 12,  2,  1])
```

برای حذف یک عنصر در یک مکان خاص از متد ()np.delete استفاده می‌شود:

```
In  [3]:   a = np.array([5, 1, 2, 3, 9, 4, 7])
           a
Out [3]:   array([5, 1, 2, 3, 9, 4, 7])

In  [4]:   np.delete(a, 0)

Out [4]:   array([1, 2, 3, 9, 4, 7])
```

برای هر برنامه‌نویس، پیچیدگی زمانی هر الگوریتم بسیار مهم است. مرتب‌سازی یک عملیات مهم و بسیار اساسی است که ممکن است به عنوان یک دانشمند داده روزانه از آن استفاده کنید. بنابراین، مهم است که از یک الگوریتم مرتب‌سازی خوب با حداقل پیچیدگی زمانی استفاده کنید. کتابخانه NumPy دارای طیف وسیعی از توابع مرتب‌سازی است که می‌توانید از آن‌ها برای مرتب‌سازی عناصر آرایه خود استفاده کنید:

```
In  [5]:   a = np.array([5,4,2,5,3,6,8,7,9,1,8])
           np.sort(a, kind='quicksort')
Out [5]:   array([1, 2, 3, 4, 5, 5, 6, 7, 8, 8, 9])
```

```
In   [6]:    a = np.array([[8,5,7,4,1,6],
                          [9,2,3,7,5,1]])
             np.sort(a, kind='mergresort')
Out [6]:     array([[1, 4, 5, 6, 7, 8],
                    [1, 2, 3, 5, 7, 9]])
```

### شناسایی شکل و اندازه یك آرایه

با استفاده از ndim می‌توان تعداد محورها یا ابعاد آرایه را بدست آورد:

```
In   [1]:    a = np.array([[8,5,7,4,1,6],
                          [9,2,3,7,5,1]])
             a.ndim
Out [1]:     2
```

size تعداد کل عناصر آرایه را به شما می‌گوید:

```
In   [2]:    a = np.array([[8,5,7,4,1,6],
                          [9,2,3,7,5,1]])
             a.size
Out [2]:     12
```

برای یافتن شکل آرایه از shape استفاده می‌شود:

```
In   [3]:    a = np.array([[8,5,7,4,1,6],
                          [9,2,3,7,5,1]])
             a.shape
Out [3]:     (2, 6)
```

## خلاصه فصل

- زبان پایتون یکی از با ارزش‌ترین و جالب‌ترین زبان‌ها برای تجزیه و تحلیل داده‌ها است.
- پایتون یکی از آسان‌ترین زبان‌ها برای شروع است. همچنین، این سادگی شما را از امکاناتی که به آن‌ها نیاز دارید محدود نمی‌کند.
- جوپیتر نوت‌بوک یک ابزار فوق‌العاده قدرتمند برای توسعه و ارائه پروژه‌های علم داده به‌صورت تعاملی است که می‌تواند علاوه بر اجرای کد، شامل متن، تصویر، صدا و یا ویدیو باشد.
- NumPy یک کتابخانه پایتون است که برای کار با آرایه‌ها استفاده می‌شود.
- هدف NumPy ارائه یک شی آرایه‌ای است که حداکثر ۵۰ برابر سریع‌تر از لیست‌های سنتی پایتون باشد.

## مراجع برای مطالعه بیشتر

# ۳

## داده

**اهداف:**

- آشنایی با انواع داده‌ها
- آماده‌سازی و تمیزسازی داده‌ها
- آشنایی با تکنیک‌های وب‌تراش
- نحوه‌ی وارد کردن داده‌ها با فرمت‌های متفاوت
- مصورسازی داده

## داده

علم داده تماما درباره آزمایش با داده‌های خام یا ساختاریافته است. داده‌ها محرکی هستند که می‌توانند یک کسب و کار را به مسیر درستی سوق دهند یا حداقل بینش‌های کاربردی ارائه دهند که می‌تواند به‌راحتی، راه‌اندازی محصولات جدید را سازماندهی کرده و یا تجربیات مختلف را امتحان کند. همه این موارد یک مولفه محرک مشترک دارند و آن *داده‌ها* هستند. ما در حال ورود به عصر دیجیتال هستیم که در آن داده‌های زیادی تولید می‌کنیم. وقتی این داده‌ها اهمیت زیادی در زندگی ما دارند، ذخیره و پردازش صحیح این داده‌ها بدون خطا اهمیت پیدا می‌کند. هنگام برخورد با مجموعه داده‌ها، نوع داده‌ها نقش مهمی را ایفا می‌کند تا تعیین کند که کدام استراتژی پیش‌پردازش برای یک مجموعه خاص برای بدست آوردن نتایج مناسب کار می‌کند یا چه نوع تحلیل آماری باید برای بهترین نتایج بکار رود. درک انواع مختلف داده‌ها به شما امکان می‌دهد نوع داده‌ای را که بیشتر با نیازها و اهداف شما مطابقت دارد انتخاب کنید. این که آیا شما یک تاجر، بازاریاب، دانشمند داده یا حرفه‌ای دیگری دارید که با انواع داده‌ها کار می‌کند، باید با لیست کلیدی از انواع داده‌ها آشنا باشید.

## نوع داده کیفی[1]

داده‌های کیفی یا داده‌های رسته‌ای[2]، شی مورد بررسی را با استفاده از مجموعه‌ای محدود از کلاس‌های گسسته توصیف می‌کند. این بدان معناست که این نوع داده‌ها را نمی‌توان به‌راحتی با استفاده از اعداد شمارش یا اندازه‌گیری کرد و بنابراین به رسته‌ها تقسیم می‌شود. جنسیت یک فرد (مرد ، زن یا دیگران) مثال خوبی از این نوع داده است. داده‌های کیفی می‌توانند به سوالاتی همانند "چگونه این اتفاق افتاده است؟" یا "چرا این اتفاق افتاده است؟" پاسخ دهند. جنسیت یک فرد، رنگ‌ها و قومیت‌ها مثال‌هایی از این نوع داده‌ها هستند. به عنوان مثال، تصور کنید دانش آموزی در یکی از جلسات کلاس یک پاراگراف از کتاب را می‌خواند. معلمی که به کتاب خواندن گوش می‌دهد درباره نحوه خواندن آن پاراگراف توسط کودک بازخورد می‌دهد. اگر معلم بر اساس شیوایی (زبان‌آوری)، لحن و تلفظ بدون دادن نمره به کودک بازخورد دهد، این به‌عنوان نمونه‌ای از داده‌های کیفی در نظر گرفته می‌شود. مثال دیگر می‌تواند مربوط به یک برند تلفن هوشمند باشد که اطلاعاتی در مورد رتبه فعلی، رنگ گوشی، رسته گوشی و غیره را ارائه می‌دهد.

---

[1] Qualitative Data Type

[2] Categorical Data

همه این اطلاعات را می‌توان به‌عنوان داده‌های کیفی طبقه‌بندی کرد. دو نوع کلی از داده‌های کیفی وجود دارد: داده‌های **اسمی**[1] و **ترتیبی**[2].

## اسمی

داده‌های اسمی به عنوان داده‌هایی تعریف می‌شوند که برای نام‌گذاری یا برچسب‌گذاری متغیرها، بدون هیچ مقدار کمی استفاده می‌شوند. معمولا هیچ ترتیب ذاتی برای داده‌های اسمی وجود ندارد. به‌عنوان مثال، رنگ یک تلفن هوشمند را می‌توان به‌عنوان نوع داده اسمی در نظر گرفت. زیرا نمی‌توانیم یک رنگ را با رنگ‌های دیگر مقایسه کنیم. به عبارت دیگر، نمی‌توان بیان کرد که "قرمز" بزرگتر از "آبی" است. به‌عنوان مثالی دیگر، رنگ چشم یک متغیر اسمی است که دارای چند دسته (آبی، سبز، قهوه‌ای) است و راهی برای مرتب‌سازی این دسته‌ها از بالاترین به‌کمترین وجود ندارد.

## ترتیبی

داده‌های ترتیبی یک نوع داده دسته‌بندی شده و دارای نظم طبیعی است. متغیرهای داده‌های ترتیبی به‌صورت مرتب فهرست شده‌اند. متغیرهای ترتیبی معمولا شماره‌گذاری می‌شوند تا ترتیب لیست را نشان دهند. با این حال، اعداد از نظر ریاضی اندازه‌گیری یا تعیین نمی‌شوند، بلکه فقط به‌عنوان برچسب نظرات تعیین می‌شوند. به عنوان مثال، اگر اندازه یک مارک لباس را در نظر بگیریم، می‌توانیم آن‌ها را به راحتی براساس برچسب نام آن‌ها به ترتیب کوچک، متوسط و بزرگ طبقه‌بندی کنیم.

> این دسته‌بندی‌ها به ما کمک می‌کنند تصمیم بگیریم که کدام استراتژی کدگذاری را می‌توان روی کدام نوع داده اعمال کرد. کدگذاری داده‌ها برای داده‌های کیفی مهم است، چراکه مدل‌های یادگیری ماشین نمی‌توانند این مقادیر را مستقیما به‌کار برند و لازم است به انواع عددی تبدیل شوند، زیرا مدل‌ها ماهیت ریاضی دارند. برای نوع داده‌های اسمی که مقایسه‌ای بین دسته‌ها وجود ندارد می‌توان از کدگذاری one-hot استفاده کرد و برای نوع داده‌های ترتیبی، می‌توان از کدگذاری label استفاده کرد که به شکلی از یک عدد صحیح است.

---

[1] Nominal

[2] Ordinal

# نوع داده کمی[1]

داده‌های کمی، داده‌هایی با قابل اندازه‌گیری هستند. به عبارت دیگر، می‌توان آن را محاسبه یا اندازه‌گیری کرد و مقدار عددی برای آن بدست آورد. قیمت یک تلفن هوشمند، تخفیف ارائه شده، فرکانس پردازنده تلفن هوشمند یا رم آن گوشی، همه این موارد در دسته انواع داده‌های کمی قرار می‌گیرند. نکته اصلی این است که تعداد نامحدودی از مقادیر که یک ویژگی می‌تواند داشته باشد وجود دارد. به عنوان مثال، قیمت یک تلفن هوشمند می‌تواند از مقدار $x$ تا هر مقداری متفاوت باشد. داده‌های گسسته و پیوسته دو نوع کلیدی از داده‌های کمی هستند.

### گسسته[2]

داده‌های گسسته قابلیت شمارش دارند و تنها شامل اعداد صحیح می‌شوند. تعداد بلندگوهای تلفن همراه، تعداد دوربین‌ها، هسته‌های پردازنده، تعداد سیم‌کارت‌های پشتیبانی شده همه این‌ها نمونه‌هایی از نوع داده گسسته است.

### پیوسته[3]

داده‌های پیوسته داده‌هایی هستند که می‌توان آن‌ها را به‌طور معناداری به سطوح دقیق‌تری تقسیم کرد. می‌توان آن را در مقیاس[4] یا پیوستار[5] اندازه‌گیری کرد و تقریبا می‌تواند هر مقدار عددی داشته باشد. به عنوان مثال، می‌توانید قد خود را با مقیاس‌های بسیار دقیق متر، سانتی‌متر، میلی‌متر و غیره اندازه بگیرید. شما می‌توانید داده‌های پیوسته را در اندازه‌گیری‌های مختلف عرض، دما، زمان و غیره ثبت کنید. اینجاست که تفاوت اصلی با انواع گسسته داده‌ها نمایان می‌شود. متغیرهای پیوسته می‌توانند هر مقداری را بین دو عدد بگیرند. به عنوان مثال، بین ۶۰ تا ۸۰ سانتی‌متر، میلیون‌ها عدد دیگر وجود دارد. یک قاعده خوب برای مشخص کردن پیوسته یا گسسته بودن داده‌ها این است که اگر بتوان نقطه اندازه‌گیری را به نصف کاهش داد و همچنان عدد بدست آمده منطقی بود، داده‌ها پیوسته هستند.

---

[1] Quantitative Data Type

[2] Discrete

[3] Continuous

[4] scale

[5] continuum

# داده‌های سری زمانی[1]

خواه بخواهیم روند بازارهای مالی را پیش‌بینی کنیم یا مصرف برق، زمان عامل مهمی است که اکنون باید در مدل‌های ما مورد توجه قرار گیرد. به عنوان مثال، جالب خواهد بود که پیش‌بینی کنیم در چه ساعتی از روز اوج مصرف برق وجود دارد. برای این انجام این کار کافی است تا از داده‌های سری زمانی استفاده کنیم. داده‌های سری زمانی دنباله‌ای از اعداد است که در فواصل منظم طی یک دوره زمانی جمع‌آوری می شوند. در یک سری زمانی، زمان اغلب متغیر مستقل است و هدف معمولا ایجاد پیش‌بینی برای آینده است.

تجزیه و تحلیل سری‌های زمانی یک روش خاص برای تجزیه و تحلیل دنباله‌ای از نقاط داده‌های جمع‌آوری‌شده در یک بازه زمانی است. در تجزیه و تحلیل سری‌های زمانی، تحلیل‌گران نقاط داده را در فواصل ثابت در یک بازه زمانی مشخص ثبت می کنند نه این‌که فقط نقاط داده را به‌صورت متناوب یا تصادفی ثبت کنند. با این حال، این نوع تجزیه و تحلیل صرفا عمل جمع‌آوری داده‌ها در طول زمان نیست. آنچه داده‌های سری زمانی را از سایر داده‌ها متمایز می‌کند این است که در تجزیه و تحلیل می‌تواند نشان دهد که چگونه متغیرها در طول زمان تغییر می‌کنند. به عبارت دیگر، زمان یک متغیر مهم است، چراکه نشان می‌دهد که چگونه داده‌ها در طول مسیر و همچنین نتایج نهایی تنظیم می‌شوند. این یک منبع اطلاعات اضافی و مجموعه ای از وابستگی‌ها بین داده‌ها را فراهم می‌کند.

تجزیه و تحلیل سری‌های زمانی به‌طور معمول به تعداد زیادی از نقاط داده برای اطمینان از ثبات و قابلیت اطمینان نیاز دارد. یک مجموعه داده بزرگ تضمین می‌کند که شما یک اندازه کافی از نمونه‌های شاخص داشته باشید و تجزیه و تحلیل داده‌های نویزدار را کاهش دهید. همچنین تضمین می‌کند که هرگونه الگوی روند یا الگوی کشف شده دور از ذهن نیست.

تجزیه و تحلیل سری‌های زمانی به سازمان‌ها کمک می‌کند تا علل زمینه‌ای روندها یا الگوهای نظامند[2] را در طول زمان درک کنند. با استفاده از مصورسازی داده‌ها، کاربران تجاری می‌توانند روندهای فصلی را ببینند و علت وقوع این روندها را عمیق‌تر بررسی کنند. با مدل‌های تجزیه و تحلیل مدرن، این مصورسازی‌ها می‌توانند بسیار فراتر از نمودارهای خطی باشند. وقتی سازمان‌ها داده‌ها را در فواصل زمانی ثابت تجزیه و تحلیل می‌کنند، می‌توانند از پیش‌بینی سری‌های زمانی نیز برای پیش‌بینی احتمال وقایع آینده استفاده کنند. پیش‌بینی سری‌های زمانی بخشی از تجزیه و تحلیل‌های پیشگویانه است. این می‌تواند به احتمال زیاد تغییرات در داده‌ها،

---

[1] Time Series Data

[2] systemic

مانند فصلی یا رفتار چرخه‌ای را نشان دهد، که درک بهتری از متغیرهای داده را فراهم می‌کند و به پیش‌بینی بهتر کمک می‌کند.

تجزیه و تحلیل سری‌های زمانی برای داده‌های غیر ثابت استفاده می‌شود؛ مواردی که به‌طور دائم در طول زمان در حال نوسان هستند یا تحت تاثیر زمان قرار می‌گیرند. صنایعی مانند امور مالی، خرده فروشی و اقتصاد اغلب از تجزیه و تحلیل سری‌های زمانی استفاده می‌کنند، چراکه ارز و فروش همیشه در حال تغییر است. تجزیه و تحلیل بازار سهام نمونه‌ای عالی از تجزیه و تحلیل سری‌های زمانی در عمل است. به همین ترتیب، تجزیه و تحلیل سری‌های زمانی برای پیش‌بینی تغییرات آب و هوا ایده آل است و به هواشناسان کمک می‌کند تا همه چیز را از گزارش آب و هوای فردا تا تغییرات آب و هوایی سال‌های آینده را پیش‌بینی کنند. نمونه‌هایی از تجزیه و تحلیل سری‌های زمانی در عمل عبارتند از:

- داده‌های آب و هوا
- اندازه‌گیری بارندگی
- اندازه‌گیری ضربان قلب[1] (EKG)
- نظارت بر مغز[2]
- فروش سه ماهه
- قیمت سهام
- معاملات خودکار سهام
- نرخ بهره

## مجموعه داده‌ها و ویژگی‌ها

مجموعه داده‌ها را اغلب می‌توان به عنوان مجموعه‌ای از اشیاء داده با ویژگی‌های یکسان در نظر گرفت. نام‌های دیگر برای یک شی داده عبارتند از: رکورد، نقطه، بردار، الگو، رویداد، مورد، نمونه، مثال، مشاهده یا موجودیت. به نوبه خود، اشیاء داده با تعدادی ویژگی توصیف می‌شوند که مشخصات اصلی یک شی را نشان می‌دهند، مانند زمان وقوع یک رویداد. به عنوان مثال، رنگ چشم در افراد مختلف متفاوت است، همچنین دمای یک جسم در طول زمان متفاوت است. باید توجه داشت که رنگ یک ویژگی نمادین با تعداد کمی از مقادیر ممکن است (قهوه‌ای، سیاه، آبی، سبز و غیره)، در حالی که دما یک ویژگی عددی با تعدادی نامحدود از مقادیر است. نام‌های دیگر یک ویژگی، متغیر، مشخصه، ویژگی یا بعد است.

---

[1] Heart rate monitoring

[2] Brain monitoring

اغلب یک مجموعه داده یک فایل است که در آن هر یک از شی‌ها در یک سطر هستند و هر ستون مربوط به یکی از ویژگی‌های این شی‌ها است. به عنوان مثال، جدول ۳-۱ یک مجموعه داده را نشان می‌دهد که شامل اطلاعات دانشجویان است. هر سطر به یک دانشجو اشاره دارد و هر ستون یک ویژگی است که برخی از جنبه‌های دانشجویان را توصیف می‌کند، همانند شماره دانشجویی، سال ورود، معدل و رشته تحصیلی.

جدول ۳-۱ نمونه‌ای از یک مجموعه داده شامل اطلاعات دانشجویان

| شماره دانشجویی | سال ورود | معدل | رشته تحصیلی |
|---|---|---|---|
| ۹۷۶۰۰۱ | ۱۳۹۷ | ۱۸/۴۵ | مهندسی کامپیوتر |
| ۹۷۴۱۲۰ | ۱۳۹۷ | ۱۹/۰۳ | علوم کامپیوتر |
| ۹۹۰۲۴۵ | ۱۳۹۹ | ۱۸/۹۵ | مهندسی برق |

## ویژگی‌ها کلی مجموعه داده‌ها

سه ویژگی کلی که هنگام استفاده از بسیاری از مجموعه داده‌ها کاربرد دارد و تأثیر قابل توجهی در استفاده از تکنیک‌های یادگیری ماشین دارد، عبارتند از: ابعاد[1]، پراکندگی[2] و وضوح[3].

■ **ابعاد:** ابعاد یک مجموعه داده، تعداد ویژگی‌هایی است که اشیاء در مجموعه داده دارند. داده‌های با بعد کم از نظر کیفی متفاوت از داده‌های با بعد متوسط یا زیاد هستند. در واقع، گاهی اوقات مسائل مربوط به تجزیه و تحلیل داده‌های با ابعاد بالا را مشقت بُعدچندی می‌نامند. به همین دلیل، یک انگیزه مهم در پیش‌پردازش داده‌ها کاهش ابعاد است.

■ **پراکندگی:** در یک مجموعه داده، پراکندگی تعداد رکوردهایی را در یک جدول نشان می‌دهد که مقداری ندارند. به عبارت دیگر در برخی از مجموعه داده‌ها اکثر ویژگی‌های یک شی دارای مقادیر ۰ هستند. از نظر عملی، پراکندگی یک مزیت است، چراکه معمولا تنها مقادیر غیرصفر باید ذخیره و دست‌ورزی شوند. این امر موجب صرفه‌جویی قابل توجهی در زمان محاسبه و ذخیره‌سازی می‌شود.

■ **وضوح (دقت نمایش):** اغلب بدست آوردن داده‌ها در سطوح مختلف وضوح ممکن است و همچنین ویژگی‌های داده‌ها در وضوح مختلف متفاوت است. به عنوان مثال، سطح زمین در وضوح چند متر بسیار ناهموار به نظر می‌رسد، اما در وضوح ده‌ها کیلومتر نسبتا هموار است. الگوهای موجود در داده‌ها نیز بستگی به سطح وضوح دارد. اگر سطح وضوح بسیار

---

[1] dimensionality

[2] sparsity

[3] resolution

خوب باشد، ممکن است یک الگو قابل مشاهده نباشد یا در بین نویزها حذف شود. اگر وضوح بسیار کلان باشد، ممکن است الگو ناپدید شود. به عنوان مثال، تغییرات فشار جوی در مقیاس ساعت، حرکت طوفان‌ها و دیگر سیستم‌های آب و هوایی را منعکس می‌کند. در مقیاس چند ماه، چنین پدیده‌هایی قابل تشخیص نیستند.

## نمونه‌هایی از داده‌های با ابعاد بالا

مثال‌های زیر مجموعه داده‌های با ابعاد بالا در زمینه‌های مختلف را نشان می‌دهد.

**مثال ۱:** داده‌های بهداشت و درمان

داده‌های با ابعاد بالا در مجموعه داده‌های بهداشت و درمان رایج است که تعداد ویژگی‌های یک فرد مشخص می‌تواند بسیار بزرگ باشد، به عنوان مثال، فشار خون، ضربان قلب در حالت استراحت، وضعیت سیستم ایمنی بدن، سابقه جراحی، قد، وزن، شرایط موجود و غیره. در این مجموعه داده‌ها، معمول است که تعداد ویژگی‌ها بیشتر از تعداد مشاهدات باشد.

**مثال ۲:** داده‌های مالی

داده‌های با ابعاد بالا همچنین در مجموعه داده‌های مالی رایج است که تعداد ویژگی‌های یک سهام مشخص می‌تواند بسیار زیاد باشد، به عنوان مثال، حجم معاملات، نسبت PE، نرخ سود سهام و غیره. در این نوع از داده‌ها، متداول است که تعداد ویژگی‌ها بسیار بیشتر از تعداد سهام فردی باشد.

**مثال ۳:** ژنومیک

داده‌های با ابعاد بالا اغلب در زمینه ژنومیک رخ می‌دهد که در آن تعداد ویژگی‌های ژن برای یک فرد مشخص می‌تواند عظیم باشد.

## نحوه مدیریت داده‌های با ابعاد بالا

امکان تجزیه و تحلیل تک تک ابعاد در سطح کوچک در داده‌های با ابعاد بالا وجود ندارد. ممکن است روزها یا ماه‌ها طول بکشد تا تجزیه و تحلیل معناداری انجام دهیم که به زمان و هزینه زیادی نیاز دارد. آموزش داده‌هایی با ابعاد بالا مشکلاتی را برای ما به دنبال داشت:

- با افزایش ابعاد، فضای مورد نیاز برای ذخیره اطلاعات افزایش می‌یابد.
- با افزایش ابعاد، امکان بیش‌برازش مدل نیز افزایش می‌یابد.
- ابعاد بیش‌تر، پیچیدگی زمانی بیشتری در آموزش مدل دارد.
- ما نمی‌توانیم یک داده با ابعاد بالا را تجسم کنیم. با کاهش ابعاد، داده‌ها را برای تجسم بهتر به دو بعدی یا سه بعدی کاهش می‌دهیم.

از همین‌رو، برای مقابله با این مشکلات نیاز است تا داده‌هایی با ابعاد بالا مدیریت شوند. دو روش معمول برای مقابله با داده‌های با ابعاد بالا وجود دارد:

## ۱. انتخاب ویژگی‌های کم‌تر

واضح‌ترین راه برای جلوگیری از برخورد با داده‌هایی با ابعاد بالا این است که به سادگی ویژگی‌های کم‌تری را از مجموعه داده انتخاب کنید. روش‌های مختلفی برای تصمیم‌گیری در مورد حذف ویژگی‌ها از مجموعه داده وجود دارد، از جمله:

- **حذف ویژگی‌هایی با بسیاری از مقادیر مفقود شده:** اگر یک ستون داده شده در مجموعه داده دارای مقادیر زیادی از داده‌های مفقود شده باشد، ممکن است بتوانید بدون از دست دادن اطلاعات زیاد آن را به طور کامل حذف کنید.

- **حذف ویژگی‌هایی با واریانس کم:** اگر یک ستون داده شده در مجموعه داده دارای مقادیری باشد که تغییرات بسیار کمی دارد، ممکن است بتوانید آن را حذف کنید، زیرا بعید است که اطلاعات مفیدی در مورد یک متغیر پاسخ[۱] در مقایسه با سایر ویژگی‌ها ارائه دهد.

- **حذف ویژگی‌هایی با همبستگی کم با متغیر پاسخ:** اگر ویژگی خاصی با متغیر پاسخ مورد علاقه ارتباط چندانی ندارد، احتمالاً می‌توانید آن را از مجموعه داده حذف کنید، زیرا بعید است که در یک مدل، ویژگی مفیدی باشد.

## ۲. استخراج ویژگی‌ها

یکی دیگر از تکنیک‌های متداول برای مدیریت داده‌هایی با ابعاد بالا استخراج ویژگی است. هدف استخراج ویژگی کاهش تعداد ویژگی‌های یک مجموعه داده با ایجاد ویژگی‌های جدید از ویژگی‌های موجود (و سپس حذف ویژگی‌های اصلی) است. این مجموعه جدید کاهش یافته از ویژگی‌ها، باید بتواند بیشتر اطلاعات موجود در مجموعه اصلی ویژگی‌ها را خلاصه کند. به این ترتیب، یک نسخه خلاصه از ویژگی‌های اصلی را می‌توان از ترکیب مجموعه اصلی ایجاد کرد.

> تفاوت بین انتخاب ویژگی و استخراج ویژگی این است که انتخاب ویژگی اهمیت ویژگی‌های موجود در مجموعه داده را رتبه‌بندی کرده و ویژگی‌های کمتر مهم را کنار بگذارد (هیچ ویژگی جدید ایجاد نمی‌شود)، در حالی‌که استخراج ویژگی‌ها با کاهش ویژگی‌ها به ایجاد ویژگی‌های جدیدی از ویگی‌های موجود منجر می‌شود.

---

[۱] Response Variable

## گردآوری و ساخت داده[1]

هر برنامه‌ای اهداف و الزامات خاص خود را دارد که باید برآورده شود. استراتژی‌های متفاوتی باید مورد استفاده قرار گیرد تا خروجی قابل اطمینانی در پایان بدست آید. چنین نیازی منجر به ایده‌ی ایجاد مجموعه داده‌های جدید می‌شود که بعدا می‌تواند برای اهداف متعددی مورد استفاده قرار گیرد. ایجاد یک مجموعه داده بزرگ در صورتی که به صورت دستی انجام شود یک کار خسته کننده است. اما روش‌هایی مانند وب تراش[2] و خزنده وب[3] می تواند جمع آوری داده‌ها را خودکار کرده و ایجاد مجموعه داده برای تجزیه و تحلیل را آسان‌تر کند.

## وب تراش

در دنیای رقابتی امروز همه به دنبال راه‌هایی برای نوآوری و استفاده از فناوری‌های جدید هستند. وب تراش راه حلی را برای کسانی که می‌خواهند به صورت خودکار به داده‌های ساختاریافته وب دسترسی پیدا کنند، ارائه می‌دهد. اگر وب سایت عمومی که می‌خواهید از آن اطلاعات دریافت کنید API نداشته باشد یا فقط دسترسی محدودی به داده‌ها داشته باشد، وب تراش مفید است.

به طور کلی، استخراج داده‌های وب توسط افراد و مشاغل مورد استفاده قرار می‌گیرد که می‌خواهند از حجم وسیعی از داده‌های وب در دسترس عموم، برای تصمیم‌گیری هوشمندانه استفاده کنند. این روش را می‌توان با استفاده از روش‌های سنتی رونوشت‌ـ درج[4] به صورت دستی انجام داد، اما در بیشتر موارد ابزارهای خودکار ترجیح داده می‌شوند، زیرا هزینه کم‌تری دارند و با سرعت بیشتری کار می‌کنند. کل فرآیند وب تراش را می‌توان به مراحل مختلف تقسیم کرد و به‌طور خلاصه به شرح زیر توضیح داد:

- **فاز اول ـ واکشی داده‌ها[5]:** در این مرحله وب سایت‌هایی باید انتخاب شوند که داده‌ها از آن قابل دسترسی باشند. سپس می‌توان با استفاده از پروتکل HTTP، یک پروتکل اینترنتی که برای ارسال و دریافت درخواست از یک سرور وب استفاده می‌شود، واکشی را انجام داد.

- **فاز دوم ـ استخراج اطلاعات[6]:** پس از واکشی اسناد HTML مورد نظر، مرحله بعد استخراج اطلاعات مورد نیاز ما از وب سایت است. این را می‌توان با استفاده از چندین تکنیک همانند تجزیه HTML، DOM، XPath و تطبیق الگوی متن انجام داد.

---

[1] Data collection

[2] web scraping

[3] web crawling

[4] copy-pasting

[5] Fetch data

[6] Extracting Information

▪ **فاز سوم_ تبدیل داده‌ها**[1]: پس از استخراج اطلاعات مورد نیاز از مکان‌های مورد نظر (URL)، داده‌ها در قالب بدون ساختار خواهند بود. سپس می‌توان آن را به صورت ساختار یافته مانند CSV، صفحه گسترده[2] یا pdf، برای ارائه یا ذخیره‌سازی تبدیل کرد.

## تکنیک‌های وب تراش

برای پیاده سازی وب تراش، چندین رویکرد وجود دارد که از بین آن‌ها می‌توان بر اساس نیاز برنامه‌نویس بهترین روش را انتخاب کرد. با توجه به موارد فوق، می‌توان مشاهده کرد که این تکنیک‌ها در مرحله استخراج مورد استفاده قرار می‌گیرند و عمدتا به دو دسته وب تراش دستی و وب تراش خودکارتقسیم می‌شوند. این بخش در مورد برخی از تکنیک‌های اصلی در هر دسته بحث می‌کند و زمینه‌ای برای انتخاب تکنیک مناسب در میان آن‌ها را ارائه می‌دهد.

### وب تراش دستی

جای تعجب نیست که وب تراش را می‌توان به صورت دستی انجام داد. رونوشت_ درج سنتی یک رویکرد دستی است که در آن داده هایی که باید از یک وب سایت استخراج شوند به صورت دستی به عنوان یک گروه رونوشت شده و در یک سند درج می‌شوند. سپس داده‌های مورد نیاز از گروه جمع‌آوری شده و به شکل ساختاریافته مرتب می‌شوند. گاهی اوقات، برای اطلاعات کم این روش می‌تواند بهترین تکنیک باشد. اما در روند ایجاد مجموعه داده‌های بزرگ، این تکنیک می‌تواند خسته کننده و مستعد خطا باشد، چراکه شامل کارهای دستی زیادی است. این روش مزایا و معایب زیادی دارد:

▪ **مزایا**
↳ مسلما ساده‌ترین روش وب تراش است، چراکه نیازی به یادگیری مهارت‌های جدید برای انجام این کار نیست.
↳ به افراد اجازه می‌دهد تا هر نقطه داده را بررسی کند و از خطاها جلوگیری کرده یا از داده‌های نامربوط در حین استخراج صرف نظر کند.
↳ با توجه به سرعت کند وب تراش دستی، بعید است که دسترسی به وب سایتی که از آن داده‌ها را استخراج می‌کنید، مسدود شود.

▪ **معایب**
↳ مسلما کندترین روش وب تراش است. حتی در حداکثر سرعت، یک ربات وب تراش به طور قابل توجهی سریع‌تر از یک انسان در استخراج داده‌ها عمل می‌کند.

---

[1] Data Transformation

[2] spreadsheet

← با دقت انسانی نیز خطای انسانی حاصل می‌شود. بسته به اهمیت صحت داده‌ها، خطای انسانی می‌تواند هزینه زیادی برای شما داشته باشد.

**وب تراش خودکار**

در طرف دیگر وب تراش دستی، وب تراش خودکار وجود دارد که به دلیل سهولت استفاده و صرفه‌جویی در زمان و هزینه، رواج فزاینده‌ای یافته‌اند. این رویکرد توسط تکنیک‌های متفاوتی انجام می‌شود که آن‌ها را در ادامه به شرح آن‌ها می‌پردازیم. این روش نیز مزایا و معایب زیادی دارد:

■ **مزایا**

← برای وب تراش، فوق‌العاده سریع عمل کرده و صدها رکورد را در چند ثانیه استخراج می‌کند.

← استفاده از آن‌ها آسان است. اکثر وب تراش‌های مدرن UIهای فوق‌العاده ساده‌ای را به اجرا در آورده‌اند که به هرکسی اجازه می‌دهد بدون نیاز به مهارت‌های کدنویسی، داده‌ها را از وب خارج کند.

■ **معایب**

← ممکن است به آموزش جزئی در مورد نحوه استفاده از خود ابزار نیاز داشته باشند. برخی از ابزارها با پیاده سازی UIها ساده‌ای با این مشکل مقابله می‌کنند.

← برخی از وب سایت‌ها به‌طور فعال سعی می‌کنند وب تراش‌های صفحات وب خود را مسدود کنند.

← عدم کنترل انسانی در حین استخراج داده‌ها. توصیه می‌شود قبل از استفاده از مجموعه داده‌های استخراج‌شده از وب تراش‌های خودکار، داده‌ها را بررسی کنید.

**تجزیه HTTP**

به طور کلی، تجزیه[1] فرایند تحلیل رشته‌ای از نمادها در زبان طبیعی، زبان رایانه یا ساختار داده‌ها مطابق با قوانین دستور زبان است. نتیجه تجزیه سند معمولا درختی با مجموعه‌ای از گره‌هایی است که نمایانگر ساختار آن است. در تجزیه HTML، پس از واکشی سند HTML، یک درخت از گره‌ها در حین تجزیه ایجاد می‌شود که از آن می‌توان اطلاعاتی همانند عناوین[2] صفحه، سرصفحه[3]، پاراگراف‌های صفحه را با تشخیص گره‌های HTML استخراج کرد. برخی از

---

[1] parsing

[2] heading

[3] title

زبان‌های پرس‌وجو داده‌های نیمه‌ساختاریافته، همانند XQuery و HTQL، می‌توانند برای تجزیه صفحات HTML و بازیابی و تغییر محتوای صفحه استفاده شوند.

## تجزیه DOM

DOM یک زیرساختار[1] رایج برای مدیریت ساختار سند است که با ایجاد یک رابط کاربری در دسترسی به ساختار و محتویات این اسناد بر روی اسناد XML کار می‌کند. همانند تجزیه کننده‌های HTML، هنگامی که یک سند XML با تجزیه‌کننده DOM واکشی و اعمال می‌شود، یک ساختار درختی شامل همه عناصر سند شکل می‌گیرد. با کمک DOM می‌توان محتویات و ساختار سند را بررسی و برای استخراج از آن استفاده کرد.

## تجزیه XPath

XPath مخفف زبان مسیر XML[2] است. از این فناوری می‌توان در اسناد XML برای دسترسی به عناصر مختلف در ساختار و محتوای آن‌ها استفاده کرد. XPath همچنین می‌تواند برای دسترسی در اسناد HTML استفاده شود، زیرا ساختار مشابهی با XML دارند. XPath به ما امکان می‌دهد به جای بررسی کل درخت، عباراتی را بنویسیم که بتوانند مستقیما به عناصر HTML دسترسی داشته باشند. به طور کلی، پس از تجزیه DOM، XPath می‌تواند به عنوان وب تراش در جهت استخراج داده‌ها استفاده شود. XPath یک زبان نیست، بلکه در قالب عباراتی ارائه می‌شود که به نحو خاص خود نیاز دارد.

## انتخاب کننده‌های CSS

از دیگر تکنیک‌های رایج وب تراش‌ها برای استخراج داده‌ها از اسناد HTML، استفاده از انتخاب کننده‌های CSS است. CSS[3] زبانی است برای یکپارچه‌سازی اسناد HTML استفاده می‌شود و عمدتا ارائه اسناد ساختاریافته همانند HTML و XML را توصیف می‌کند. بر اساس ویژگی‌های مختلف CSS، انتخاب کننده‌های مختلفی همانند نوع، ویژگی، شناسه و غیره برای نشان دادن ساختار و محتوای وب سایت استفاده می‌شود. از این موارد می‌توان برای مطابقت و استخراج عناصر HTML استفاده کرد.

---

**تطبیق الگوی متن**

تطبیق الگوی متن یک تکنیک تطبیق[1] است که در آن از عبارات منظم[2] برای مطابقت با برچسب‌های HTML و استخراج داده‌ها از اسناد HTML استفاده می‌شود. عبارات منظم، به‌طورکلی، توالی از حروف هستند که منجر به الگوی جستجو می‌شوند. از آنجایی که HTML تقریبا از رشته‌های زیادی تشکیل شده است، عبارات منظم می‌توانند در اینجا با مطابقت رشته‌های مختلف وارد عمل شوند. اما عبارات منظم ممکن است اولین گزینه در تجزیه HTML نباشند، چراکه شانس مواجهه با اشتباهاتی مثل برچسب‌های گم‌شده وجود دارد.

**کتابخانه‌های وب تراش**

مجموعه‌ی بسیار زیادی از کتابخانه‌های پایتون برای انجام وب تراش موجود است. اما کدام برای یک پروژه خاص باید انتخاب شود؟ کدام یک از این کتابخانه‌ها بیشترین انعطاف‌پذیری را دارند؟ هدف این بخش پاسخگویی به این سوالات از طریق مرور چندین کتابخانه مشهور پایتون برای وب تراش است که هر علاقه‌مندی باید از آن مطلع باشد.

**Requests**

Requests اساسی‌ترین کتابخانه پایتون برای وب تراش است. از طریق آن می‌توان درخواست‌های HTML را به یک سرور وب سایت ارسال کرد تا داده‌ها را از صفحه وب بازیابی کند. این کتابخانه در مرحله واکشی فرآیند خراشیدن وب مورد استفاده قرار می‌گیرد. این کتابخانه پایتون با ارائه انواع مختلف درخواست‌های HTTP مانند GET، POST انعطاف‌پذیری بالایی به کاربران می‌دهد. از آنجا که این یک کتابخانه پایه است که تنها می‌تواند برای واکشی صفحات وب استفاده شود، نمی‌توان به صورت جداگانه برای گردآوری داده‌ها از آن استفاده کرد. از همین رو برای بدست آوردن خروجی قابل اطمینان باید با کتابخانه‌های دیگر ترکیب شود.

چگونه می‌توانیم از این کتابخانه استفاده کنیم؟ بدست آوردن HTML خام یک صفحه وب کار ساده‌ای است، سپس باید آن را تجزیه کرده و داده‌های مورد نیاز خود را استخراج کنید. بیایید مثالی را ببینیم که در آن صفحه ویکی پدیا در مورد "یادگیری عمیق" را تراش[3] می‌دهیم.

کتابخانه Requests اغلب در کتابخانه‌های داخلی پایتون گنجانده شده است، با این حال اگر به دلایلی نمی‌توانید آن را وارد کنید، کافی است تا دستور زیر را در خط فرمان اجرا کنید:

```
> pip install requests
```

---

پس از نصب کتابخانه، باید آن را به پروژه خود وارد کنیم. سپس، باید یک درخواست GET به
URL مورد نظر ارائه دهیم:

```
In  [1]:   import requests
           r =requests.get('https://fa.wikipedia.org/wiki/یادگیری_عمیق')
           print(r.content)
Out [1]:   b'<!DOCTYPE    html>\n<html    class="client-nojs"
           lang="fa"  dir="rtl">\n<head>\n<meta  charset="UTF-
           ^"/>\n<title>.......
```

## LXML

همان‌طور که در قسمت قبل بیان شد Requests محدودیتی داشت که نمی‌توان از آن به عنوان
تجزیه‌کننده استفاده کرد. LXML، یک ابزار سریع برای تجزیه HTML و XML در پایتون است
که از آن می‌توان برای تجزیه و استخراج داده‌ها از صفحات وب استفاده کرد. این کتابخانه
سریع‌تر از اکثر تجزیه‌کنندگان است و از API پایتون استفاده می‌کند و کاربرد را آسان‌تر می‌کند.
اما با اسناد HTML ضعیف طراحی شده به خوبی کار نمی‌کند. از همین رو باعث می‌شود LXML
در مقایسه با سایر کتابخانه‌ها انعطاف‌پذیری کم‌تری داشته باشد.

برای نصب این کتابخانه کافی است که در خط فرمان دستور زیر را اجرا کنید:

```
> pip install lxml
```

در این مثال سعی می‌کنیم همه پیوندها را در یک صفحه وب نشان دهیم (با این حال فقط
تعدادی از آن‌ها را در خروجی به نمایش می‌گذاریم). ما مجددا از کتابخانه Requests برای
دریافت کد خام HTML صفحه وب و سپس تجزیه آن با استفاده از LXML استفاده می‌کنیم.

```
In  [1]:   import requests
           import lxml.html
           r =requests.get('https://fa.wikipedia.org/wiki/یادگیری_عمیق')
           content = r.content
           doc = lxml.html.fromstring(content)
           for element in doc.xpath('//a/@href'):
             print(element)
Out [1]:
           https://da.wikipedia.org/wiki/Deep_learning
           https://de.wikipedia.org/wiki/Deep_Learning
           https://en.wikipedia.org/wiki/Deep_learning
           https://es.wikipedia.org/wiki/Aprendizaje_profundo
           https://et.wikipedia.org/wiki/S%C3%BCgav%C3%B5pe
           https://eu.wikipedia.org/wiki/Ikaskuntza_sakon
```

https://fi.wikipedia.org/wiki/Syv%C3%A4oppiminen
https://fr.wikipedia.org/wiki/Apprentissage_profond
https://it.wikipedia.org/wiki/Apprendimento_profondo
https://ms.wikipedia.org/wiki/Pembelajaran_dalam
.....

در قطعه کد بالا، از عبارت XPath برای انتخاب تمام پیوندهایی که پیدا کردیم و چاپ آن‌ها استفاده کرده‌ایم.

## BeautifulSoup

BeautifulSoup یک کتابخانه پایتون است که برای استخراج اطلاعات از فایل‌های XML و HTML استفاده می‌شود. این کتابخانه به دلایلی زیبا نامیده می‌شود، زیرا به شما کمک می‌کند داده‌های استخراج شده را با سهولت تجزیه کنید، در آن حرکت کرده و فقط داده‌های مورد علاقه خود را انتخاب کنید. BeautifulSoup به دلیل سهولت استفاده محبوبیت خود را بدست آورده است، اما در مقایسه با LXML کندتر است. یکی از مزایای مهم این کتابخانه این است که برای هر نوع وب سایتی مناسب است (توانایی تشخیص کدگذاری[1] صفحه و در نتیجه دریافت اطلاعات دقیق‌تر از متن HTML را داراست) و می‌تواند همراه با Requests برای انجام موفقیت آمیز مراحل واکشی و استخراج استفاده شود. برای سرعت بخشیدن به روند، می‌توان آن را با تجزیه کننده LXML ترکیب کرد.

برای نصب این کتابخانه کافی است که در خط فرمان دستور زیر را اجرا کنید:

```
> pip install bs4
```

در این مثال سعی می‌کنیم همه پاراگراف‌های مثال قبل را تجزیه کرده و محتوای آن را چاپ کنیم (تنها چند مورد از آن‌ها در اینجا نمایش داده شده است):

```
In  [1]:   import requests
           from bs4 import BeautifulSoup
           r =requests.get('https://fa.wikipedia.org/wiki/یادگیری_عمیق')
           content = r.content
           soup = BeautifulSoup(content, features="html.parser")
           for element in soup.findAll('p'):
             print(element.text)
Out [1]:
           برای مثال، در پردازش تصویر، لایه‌های پست‌تر می‌توانند لبه‌ها را تشخیص دهند، در حالی‌که
           لایه‌های عالی‌تر ممکن است ویژگی‌های پرمعناتر برای انسان، همچون حروف یا چهره‌ها، را
           تشخیص دهند.
```

---

[1] encoding

تا قبل از پیدایش یادگیری عمیق، روش‌های یادگیری ماشین سنتی، بیش‌از حد به بازنمایی‌هایی (انتخاب ویژگی‌ها)که از داده‌ها بدست می‌آورند، وابسته بودند. این روش‌ها، نیاز به یک متخصص در دامنه موضوع داشت تا استخراج ویژگی‌ها را به‌صورت دستی انجام دهد. حال آنکه، این استخراج ویژگی‌ها به صورت دستی فرآیندی چالش‌انگیز و زمان‌بر است. پیدایش یادگیری عمیق توانست به‌سرعت جایگزین این روش‌های سنتی شود. چراکه می‌توانست استخراج ویژگی‌ها را به‌صورت خودکار متناسب با هر مساله بدست آورد

در قطعه کد قبلی به BeautifulSoup گفتیم که از تجزیه کننده "html.parser" برای محتوای استخراج شده استفاده کرده و همه برچسب‌های <P> را برای ما انتخاب کند.

## Selenium

کتابخانه‌هایی که تاکنون ذکر شده‌اند دارای محدودیتی هستند که نمی‌توانند با وب سایت‌های طراحی شده با جاوااکسریپت کار کنند. این امر سبب می‌شودکار کردن با صفحات وب پویا در وب تراش مشکل باشد. کار با صفحات وب پویا یکی از بزرگترین چالش‌ها در زمینه وب تراش است اما Selenium یکی از کتابخانه‌های پایتون است که می‌تواند بر این مشکل غلبه کند. Selenium یک ابزار منبع باز مبتنی بر وب و یک درایور وب است، به این معنی که می‌توانید از آن برای باز کردن یک صفحه وب، کلیک بر روی یک دکمه و دریافت نتایج استفاده کنید.

علیرغم قدرت آن، Selenium یک ابزار مبتدی است. همچنین به کد اجازه می‌دهد تا رفتار انسان را تقلید کند. با این حال، یکی از محدودیت‌های اصلی این کتابخانه بارگیری و اجرای جاوااسکریپت برای هر صفحه است که سبب می‌شود روندها کندتر شوند و برای پروژه‌های بزرگ مناسب نباشند.

برای نصب این کتابخانه کافی است که در خط فرمان دستور زیر را اجراکنید:

```
> pip install selenium
```

توجه داشته باشید که با استفاده از Selenium برای دریافت محتویات صفحه وب نیازی به‌کمک کتابخانه‌های دیگر نداریم، زیرا سلنیوم می‌تواند همه کارها را به تنهایی انجام دهد! پس از اینکه به درایور وب گفتیم که قصد داریم از chromedriver استفاده کنیم و کدام URL را باید خط تراش دهد، باید مشخص کنیم که در داده‌های استخراج شده در جستجوی چه چیزی هستیم. در این مثال سعی کنیم دوباره همه پیوندها را دریافت کنیم:

```
In     from selenium import webdriver
[1]:   from webdriver_manager.chrome import ChromeDriverManager
       import requests
       from selenium import webdriver
       chrome_options = webdriver.ChromeOptions()
       chrome_options.add_argument('--headless')
       chrome_options.add_argument('--no-sandbox')
```

```
    chrome_options.add_argument('--disable-dev-shm-usage')
    wd = webdriver.Chrome('chromedriver',chrome_options=chrome_options)
    driver=webdriver.Chrome('chromedriver',chrome_options=chrome_option
    s)
    driver.get('https://fa.wikipedia.org/wiki/یادگیری_عمیق')
    links = driver.find_elements_by_tag_name('a')
    for element in links:
      print(element.get_attribute('href'))
```

Out
[1]:
https://da.wikipedia.org/wiki/Deep_learning
https://de.wikipedia.org/wiki/Deep_Learning
https://en.wikipedia.org/wiki/Deep_learning
https://es.wikipedia.org/wiki/Aprendizaje_profundo
https://et.wikipedia.org/wiki/S%C%3BCgav%C%3Bope
https://eu.wikipedia.org/wiki/Ikaskuntza_sakon
https://fi.wikipedia.org/wiki/Syv%C%3A£oppiminen

# ذخیره‌سازی و ارائه داده‌ها

بسته به ماهیت داده‌ها می‌توان آن‌ها را در قالب‌های مختلف ذخیره کرد. برخی از قالب‌ها، داده‌ها را به‌گونه‌ای ذخیره می‌کنند که به راحتی توسط ماشین‌ها کنترل می‌شوند، در حالی که برخی دیگر، داده‌ها را به‌گونه‌ای ذخیره می‌کنند که برای انسان قابل خواندن است. اسناد مایکروسافت ورد نمونه‌ای از دومی است. در مقابل، CSV، JSON و XML نمونه‌هایی از موارد اول هستند. در ادامه این بخش، ابتدا مختصری در مورد هر یک از این قالب‌های ذخیره‌سازی داده‌ها بحث کرده و سپس نحوه خواندن چنین فایل‌هایی که توسط ماشین‌ها کنترل می‌شوند را بررسی خواهیم کرد.

## CSV (مقادیر جداشده با ویرگول[1])

اولین فایل قابل خواند توسط ماشین‌ها که با آن آشنا خواهیم شد فایل CSV است. در یک فایل CSV، ستون‌های داده‌ها با استفاده از ویرگول از یکدیگر جدا می‌شوند. این راهی برای تبادل داده‌های ساختاریافته در بین برنامه‌هایی است که لزوما نمی‌توانند مستقیما با یکدیگر صحبت کنند.

### نحوه وارد کردن داده‌های CSV در پایتون

روش‌های مختلفی برای خواندن یک فایل CSV وجود دارد که از ماژول CSV یا کتابخانه pandas استفاده می‌کند:

---

- **ماژول csv:** ماژول CSV یکی از ماژول‌های پایتون است که کلاس‌هایی را برای خواندن و نوشتن داده‌های جداول در قالب فایل CSV فراهم می‌کند.

- **کتابخانه pandas:** کتابخانه pandas یکی از کتابخانه‌های منبع ـ باز پایتون است که ابزارها و تکنیک‌های تجزیه و تحلیل داده‌ها ارائه می‌دهد.

**با استفاده از ()csv.reader**

با استفاده از چند خط کد زیر می‌توان فایل داده‌ای در قابل CSV را توسط پایتون باز کرده و داده‌ها را به شکلی تبدیل کنیم که پایتون بتواند آن‌ها را درک کند. در قطعه کد زیر از تابع ()csv.reader برای خواندن فایل data.csv استفاده می‌شود تا داده‌ها را به لیست‌ها نگاشت کند:

```
In [1]:   import csv

          # opening the CSV file
          with open('data.csv', mode ='r')as file:

           # reading the CSV file
           csvFile = csv.reader(file)

           # displaying the contents of the CSV file
           for lines in csvFile:
               print(lines)
Out [1]:
          ['id', 'name', 'age', 'Field of Study']
          ['9396321', 'sara', '26', 'electrical engineering']
          ['9496328', 'mahsa', '24', 'Computer Engineering']
          ['9896325', 'maryam', '23', 'computer science']
          ['9496352', 'sepideh', '25', 'electrical engineering']
          ['9896389', 'sima', '26', 'computer science']
          ['9896325', 'mina', '21', 'electrical engineering']
```

در ابتدا، فایل CSV با استفاده از متد ()open در حالت 'r' باز می‌شود (حالت خواندن را هنگام باز کردن یک فایل مشخص می‌کند) که شی فایل را بر می‌گرداند. سپس، خروجی تابع ()csv.reader(file) در متغیر csvFile ذخیره می‌شود. متغیر csvFile در حال حاضر یک خواننده CSV با فایل باز شده را در اختیار دارد. این خواننده CSV به ما امکان می‌دهد تا داده‌های فایل خود را با استفاده از دستورات ساده پایتون به راحتی مشاهده کنیم.

**توجه:** کلمه کلیدی "with" همراه با متد ()open استفاده می‌شود، زیرا مدیریت استثنا را ساده کرده و به طور خودکار فایل CSV را می‌بندد.

**با استفاده از ()csv.DictReader**

این روش مشابه روش قبلی است، ابتدا فایل CSV با استفاده از متد ()open باز می‌شود سپس با استفاده از کلاس DictReader از ماژول csv داده‌های موجود در فایل CSV به یک دیکشنری نگاشت می‌شود:

```python
import csv

# opening the CSV file
with open('data.csv', mode ='r')as file:

 # reading the CSV file
 csvFile = csv.DictReader(file)

 # displaying the contents of the CSV file
 for lines in csvFile:
     print(lines)
```

In [1]:

Out [1]:

```
OrderedDict([('id', '9396321'), ('name', 'sara'), ('age', '26'),
('Field of Study', 'electrical engineering')])
OrderedDict([('id', '9496328'), ('name', 'mahsa'), ('age',
'24'), ('Field of Study', 'Computer Engineering')])
OrderedDict([('id', '9896325'), ('name', 'maryam'), ('age',
'23'), ('Field of Study', 'computer science')])
OrderedDict([('id', '9496352'), ('name', 'sepideh'), ('age',
'25'), ('Field of Study', 'electrical engineering')])
OrderedDict([('id', '9896389'), ('name', 'sima'), ('age',
'26'), ('Field of Study', 'computer science')])
OrderedDict([('id', '9896325'), ('name', 'mina'), ('age', '21'),
('Field of Study', 'electrical engineering')])
```

**با استفاده از ()pandas.read_csv**

خواندن یک فایل CSV با استفاده از توابع کتابخانه pandas بسیار آسان و ساده است. برای خواندن داده‌های فایل‌های CSV از متد ()read_csv کتابخانه pandas استفاده می‌شود.

In [1]:

```python
import pandas

# reading the CSV file
csvFile = pandas.read_csv('data.csv')

# displaying the contents of the CSV file
print(csvFile)
```

```
Out [1]:
            id    name   age       Field of Study
    ۰   ۹۳۹۶۳۲۱   sara   ۲۶   electrical engineering
    ۱   ۹۴۹۶۳۲۸   mahsa   ۲۴   Computer Engineering
    ۲   ۹۸۹۶۳۲۵   maryam   ۲۳       computer science
    ۳   ۹۴۹۶۳۰۲   sepideh   ۲۵   electrical engineering
    ٤   ۹۸۹۶۳۸۹   sima   ۲۶       computer science
    ۵   ۹۸۹۶۳۲۵   mina   ۲۱   electrical engineering
```

در برنامه فوق، متد ()read_csv کتابخانه pandas فایل data.csv را خوانده و داده‌های آن را به یک لیست دو بعدی نگاشت می‌کند.

# XML (زبان نشانه‌گذاری گسترش‌پذیر[1])

XML طراحی شده‌است که هم قابل خواندن برای انسان و هم برای ماشین باشد، بنابراین می‌توان از آن برای ذخیره و انتقال داده‌ها استفاده کرد. در دنیای واقعی، سیستم‌های کامپیوتری و پایگاه‌های داده حاوی داده‌هایی با قالب‌های ناسازگار هستند، از آنجاکه داده‌های XML در قالب متن ساده ذخیره می‌شوند، راهی مستقل از نرم‌افزار و سخت‌افزار برای ذخیره داده‌ها را فراهم می‌کند. این امر ایجاد داده‌هایی را که می‌تواند توسط برنامه‌های مختلف به اشتراک گذاشته شود بسیار ساده‌تر می‌کند. در اینجا نمونه‌ای از صفحه XML را مشاهده می‌کنید:

```xml
<?xml version="۱٫۰" encoding="UTF-۸"?>
<breakfast_menu>
<food>
   <name>Belgian Waffles</name>
   <price>$۵٫۹۵</price>
   <description>
  Two of our famous Belgian Waffles with plenty of real maple syrup
   </description>
   <calories>۶۵۰</calories>
</food>
<food>
   <name>Strawberry Belgian Waffles</name>
   <price>$۷٫۹۵</price>
   <description>
   Light Belgian waffles covered with strawberries and whipped cream
   </description>
   <calories>۹۰۰</calories>
</food>
</breakfast_menu>
```

---

[1] eXtensible Markup Language

## نحوه وارد کردن داده‌های XML در پایتون

ماژول Elementree ابزارهای زیادی را برای دست‌ورزی فایل‌های XML در اختیار ما قرار می‌دهد. این ماژول در کتابخانه استاندارد پایتون موجود است، بنابراین نیازی به نصب هیچ ماژول خارجی برای استفاده از آن نیست. ماژول ElementTree روش‌هایی را برای نمایش کل سند XML به عنوان یک درخت ارائه می‌دهد. برای خواندن یک فایل XML، ابتدا کلاس ElementTree موجود در کتابخانه XML را وارد می‌کنیم. سپس، نام فایل XML را به متد ElementTree.parse() منتقل می‌کنیم تا تجزیه شروع شود. بعد از آن برچسب والد فایل XML را با استفاده از getroot() دریافت می‌کنیم. سپس برچسب والد فایل XML را نمایش می‌دهیم. برای بدست آوردن ویژگی‌های زیر- برچسب والد از root[٠].attrib استفاده می‌شود. با فرض اینکه یک فایل XML به‌صورت زیر داشته باشیم:

```xml
<model>
  <child name="Acer" qty="١٢">Acer is a laptop</child>
  <unique>Add model number here</unique>
  <child name="Onida" qty="١٠">Onida is an oven</child>
  <child name="Acer" qty="٧">Exclusive</child>
  <unique>Add price here</unique>
  <data>Add content here
    <family>Add company name here</family>
    <size>Add number of employees here</size>
  </data>
</model>
```

قطعه کد زیر نحوه خواندن آن را با استفاده از ماژول Elementree در پایتون نشان می‌دهد:

```python
In [1]:  import xml.etree.ElementTree as ET

         # Pass the path of the xml document
         tree = ET.parse('data-text.xml')

         # get the parent tag
         root = tree.getroot()

         # print the root (parent) tag along with its memory
         location
         print(root)

         # print the attributes of the first tag
         print(root[0].attrib)
```

```
# print the text contained within first subtag of the 5th
tag from the parent
print(root[5][0].text)
```
Out [1]:
```
<Element 'model' at 0x0000028C44F2F548>
{'name': 'Acer', 'qty': '12'}
Add company name here
```

# JSON (نشانه‌گذاری شیء جاوااسکریپت[1])

JSON یک قالب تبادل داده سبک و رایج است که نه تنها خواندن و نوشتن آن برای انسان‌ها آسان است، بلکه تجزیه و تولید آنها برای ماشین‌ها نیز آسان است. همچنین یکی از رایج‌ترین قالب‌های داده‌ای است که وب‌سایت‌ها هنگام انتقال داده‌ها به جاوااسکریپت روی صفحه از آن استفاده می‌کنند.

JSON بر روی دو ساختار ساخته شده است:

- **مجموعه‌ای از جفت‌های نام و مقدار.** در زبان‌های مختلف، این به عنوان یک شی، رکورد، دیکشنری، جدول درهم‌ساز، لیست کلید خورده یا آرایه انجمنی درک می‌شود.
- **فهرستی مرتب از مقادیر.** در اکثر زبان‌ها، این به عنوان یک آرایه، بردار، لیست یا دنباله قابل درک است.

هنگام تبادل اطلاعات بین مرورگر و سرویس‌دهنده[2]، داده‌ها فقط به صورت متن ارسال می‌شوند. JSON متنی است و ما می‌توانیم هر شی JavaScript را به JSON تبدیل کنیم و JSON را به سرویس‌دهنده ارسال کنیم. همچنین می‌توانیم هر JSON دریافت شده از سرویس‌دهنده را به اشیاء جاوااسکریپت تبدیل کنیم. به این ترتیب می‌توانیم با داده‌ها به عنوان اشیاء جاوااسکریپت بدون تجزیه و ترجمه پیچیده کار کنیم. اجازه دهید به نمونه‌هایی از نحوه ارسال و دریافت داده با استفاده از JSON نگاهی کنیم.

۱. **ارسال داده‌ها:** اگر داده‌ها در یک شی جاوااسکریپت ذخیره شوند، می‌توانیم آن را به JSON تبدیل کرده و به سرویس‌دهنده ارسال کنیم. در زیر یک مثال آورده شده است:

```
<!DOCTYPE html>
  <html>
  <body>
  <p id="demo"></p>
  <script>
    var obj = {"name":"Milad", "age":29, "state": "Tehran"};
```

---

[1] JavaScript Object Notation

[2] server

```
    var obj_JSON = JSON.stringify(obj);
    window.location = "json_Demo.php?x=" + obj_JSON;
  </script>
  </body>
  </html>
```

۲. **دریافت داده‌ها:** اگر داده‌های دریافت شده با فرمت JSON باشند، می‌توانیم آن‌ها را به یک شی جاوااسکریپت تبدیل کنیم. به عنوان مثال:

```
<!DOCTYPE html>
  <html>
  <body>
  <p id="demo"></p>
  <script>
    var obj_JSON = "{"name":"Milad", "age":29, "state": "Tehran"}";
    var obj = JSON.parse(obj_JSON);
    document.getElementById("demo").innerHTML=obj.name;
  </script>
  </body>
  </html>
```

## نحوه وارد کردن داده‌های JSON در پایتون

بارگذاری یک شی JSON در پایتون بسیار آسان است. پایتون دارای یک بسته داخلی به نام JSON است که می‌تواند برای کار با داده‌های JSON استفاده شود. این ماژول JSON متدهای زیادی را در اختیار ما قرار می‌دهد که در بین این متدها، متد ()loads به ما در خواندن فایل JSON کمک می‌کند. با فرض اینکه یک فایل JSON به‌صورت زیر داشته باشیم:

```
[
        {
                "Name": "Debian",
                "Version": "9",
                "Install": "apt",
                "Owner": "SPI",
                "Kernel": "4.9"
        },
        {
                "Name": "Ubuntu",
                "Version": "17.10",
                "Install": "apt",
                "Owner": "Canonical",
                "Kernel": "4.13"
        },
        {
                "Name": "Fedora",
                "Version": "26",
                "Install": "dnf",
                "Owner": "Red Hat",
                "Kernel": "4.13"
        },
```

```json
        {
                "Name": "CentOS",
                "Version": "7",
                "Install": "yum",
                "Owner": "Red Hat",
                "Kernel": "3.10"
        },
        {
                "Name": "OpenSUSE",
                "Version": "42.3",
                "Install": "zypper",
                "Owner": "Novell",
                "Kernel": "4.4"
        },
        {
                "Name": "Arch Linux",
                "Version": "Rolling Release",
                "Install": "pacman",
                "Owner": "SPI",
                "Kernel": "4.13"
        },
        {
                "Name": "Gentoo",
                "Version": "Rolling Release",
                "Install": "emerge",
                "Owner": "Gentoo Foundation",
                "Kernel": "4.12"
        }
]
```

قطعه کد زیر نحوه خواندن آن را با استفاده از ماژول JSON در پایتون نشان می‌دهد:

```python
In  [1]:   import json
           # Opening JSON file
           json_data = open('data.json').read()
           # returns JSON object as
           # a dictionary
           data = json.loads(json_data)
           # Iterating through the json
           # list
           for item in data:
              print (item)
Out [1]:
           {'Name': 'Debian', 'Version': '9', 'Install': 'apt', 'Owner':
           'SPI', 'Kernel': '4.9'}
           {'Name': 'Ubuntu', 'Version': '17.10', 'Install': 'apt',
           'Owner': 'Canonical', 'Kernel': '4.13'}
```

{'Name': 'Fedora', 'Version': '26', 'Install': 'dnf', 'Owner': 'Red Hat', 'Kernel': '4.13'}
{'Name': 'CentOS', 'Version': '7', 'Install': 'yum', 'Owner': 'Red Hat', 'Kernel': '3.10'}
{'Name': 'OpenSUSE', 'Version': '42.3', 'Install': 'zypper', 'Owner': 'Novell', 'Kernel': '4.4'}
{'Name': 'Arch Linux', 'Version': 'Rolling Release', 'Install': 'pacman', 'Owner': 'SPI', 'Kernel': '4.13'}
{'Name': 'Gentoo', 'Version': 'Rolling Release', 'Install': 'emerge', 'Owner': 'Gentoo Foundation', 'Kernel': '4.12'}

## پیش‌پردازش و آماده‌سازی داده‌ها

دور از تصور است که انتظار داشته باشیم داده‌ها کامل باشند. ممکن است مشکلاتی به دلیل خطای انسانی یا نقص در فرآیند جمع‌آوری داده‌ها وجود داشته باشد. ممکن است برخی مقادیر وجود نداشته باشد و در موارد دیگر، ممکن است اشیا جعلی یا تکراری وجود داشته باشد. به عنوان مثال، ممکن است دو پرونده متفاوت برای فردی وجود داشته باشد که اخیرا در دو آدرس مختلف زندگی کرده است. حتی اگر همه داده‌ها موجود باشند و خوب به نظر برسند، ممکن است ناسازگاری‌هایی وجود داشته باشد، به عنوان مثال، قد یک فرد ۲ متر است، اما وزن آن تنها ۲ کیلوگرم است.

علاوه بر این‌ها، بسیار نادر است که مجموعه داده‌ها به شکل مورد نیاز توسط الگوریتم‌های علم داده در دسترس باشند. اکثر الگوریتم‌های علم داده نیاز به ساختار داده‌ها در قالب جداول با رکورد در سطرها و ویژگی‌ها در ستون ها دارند. اگر داده‌ها در قالب دیگری باشند، باید داده‌ها را به گونه‌ای نگاشت کرد تا داده‌ها به ساختار مورد نیاز تبدیل شوند. از همین‌رو، داده‌ها نیاز به تمیزسازی و تبدیل دارند.

## تمیزسازی داده‌ها

تمیزسازی داده‌ها فرآیند آماده‌سازی داده‌ها برای تجزیه و تحلیل با حذف یا تغییر داده‌های نادرست، ناقص، بی‌ربط، تکراری یا قالب نامناسب است. این داده‌ها معمولا در تجزیه و تحلیل داده‌ها ضروری یا مفید نیستند، زیرا ممکن است روند کار را مختل کرده یا نتایج نادرستی را ارائه دهند. بسته به نحوه ذخیره اطلاعات و پاسخ‌های مورد نظر، روش‌های مختلفی برای تمیزسازی داده‌ها وجود دارد. تمیزسازی داده‌ها صرفا در مورد پاک کردن اطلاعات برای ایجاد فضا برای داده‌های جدید نیست، بلکه یافتن راهی برای به حداکثر رساندن دقت مجموعه داده‌ها است. تمیزسازی داده‌ها به عنوان یک عنصر اساسی در مبانی علم داده در نظر گرفته می‌شود، زیرا نقش مهمی در فرآیند تجزیه و تحلیل و کشف پاسخ‌های قابل اعتماد دارد. داده‌های نادرست

یا ناسازگار منجر به نتایج نادرست می‌شود. بنابراین، نحوه تمیزسازی و درک داده‌ها تاثیر زیادی بر کیفیت نتایج دارد.

به عنوان مثال، دولت ممکن است بخواهد آمار سرشماری جمعیت را تجزیه و تحلیل کند تا تصمیم بگیرد کدام مناطق به هزینه و سرمایه‌گذاری بیشتری در زیرساخت‌ها و خدمات نیاز دارند. در این مورد، دسترسی به داده‌های معتبر برای جلوگیری از تصمیمات مالی اشتباه مهم خواهد بود. یا در دنیای تجارت، داده‌های نادرست می‌تواند پرهزینه باشند. بسیاری از شرکت‌ها از مجموعه داده‌های اطلاعات مشتری استفاده می‌کنند که اطلاعاتی مانند اطلاعات تماس و آدرس را ثبت می‌کند. به عنوان مثال، اگر آدرس‌ها ناسازگار باشند، شرکت هزینه ارسال مجدد نامه یا حتی از دست دادن مشتریان را متحمل می‌شود.

> یک الگوریتم ساده می‌تواند بر یک الگوریتم پیچیده غلبه کند، تنها به این دلیل که داده‌های کافی و یا کیفیت بالا به آن داده شده است.

## مقادیر مفقود شده[1]

گاهی اوقات ممکن است داده‌ها در قالب مناسب باشند، اما برخی از مقادیر آن وجود نداشته باشند. جدولی را در نظر بگیرید که حاوی اطلاعات مشتری است که در آن برخی از شماره تلفن‌های خانه وجود ندارد. دلیل آن می‌تواند این باشد که برخی از مردم تلفن خانگی نداشته و در عوض از تلفن‌های همراه خود به عنوان تلفن اصلی استفاده می‌کنند. در سایر مواقع ممکن است داده‌ها به دلیل مشکلاتی در روند جمع‌آوری داده‌ها مفقود شوند. علاوه بر این، جامعیت ممکن است در زمان جمع‌آوری مهم تلقی نشده باشد. به عنوان مثال، هنگامی که ما شروع به جمع‌آوری اطلاعات مشتری کرده‌ایم، این اطلاعات محدود به یک شهر یا منطقه خاص بوده است، بنابراین جمع‌آوری کد منطقه برای یک شماره تلفن ضروری نبوده است. حال، ممکن است هنگامی که تصمیم به توسعه فراتر از آن شهر یا منطقه بگیریم دچار مشکل شویم. بنابراین، وقتی با داده‌های مفقود شده مواجه می‌شویم چه کنیم؟ هیچ پاسخ خوب و واحدی وجود ندارد. ما باید بر اساس شرایط، راهبرد مناسبی پیدا کنیم. داشتن مقادیر مفقود شده در داده‌های شما لزوما یک عقب‌نشینی نیست. با این حال، این فرصتی است برای انجام مهندسی ویژگی مناسب برای هدایت مدل به‌منظور تفسیر داده‌های مفقود شده به روش صحیح. چندین روش مختلف برای مقابله با این مشکل وجود دارد، اما هر روش مزایا و معایبی دارد. اولین قدم برای مدیریت مقادیر مفقود شده، درک دلیل عدم وجود مقادیر است. ردیابی منبع داده‌ها می‌تواند منجر به شناسایی مشکلات سیستمیک در طول ثبت داده‌ها یا خطاها در تبدیل داده شود. دانستن منبع یک مقدار از دست رفته، اغلب راهنمایی می‌کند که از کدام روش استفاده کنید. مقدار مفقود

---

[1] Missing Values

شده را می‌توان با طیف وسیعی از داده‌های مصنوعی جایگزین کرد تا بتوان مسئله را با تأثیرات ناچیزی در مراحل بعدی فرآیند علم داده مدیریت کرد. چندین راهبرد مختلف برای مدیریت داده‌های مفقود شده در ادامه فهرست شده است:

- **حذف اشیا یا ویژگی‌های داده:** یک راهبرد ساده و موثر حذف اشیاء با مقادیر گمشده است. با این حال، حتی یک شی داده حاوی برخی اطلاعات است و اگر مقدار زیادی از اشیا دارای مقادیر مفقود شده باشند، تجزیه و تحلیل قابل اعتماد می‌تواند دشوار یا غیرممکن باشد. با این وجود، اگر در یک مجموعه داده تنها چند شی آن دارای مقادیر مفقود شده باشد، حذف آن‌ها ممکن است مفید باشد. یک راهبرد مرتبط این است که ویژگی‌هایی را که دارای مقادیر مفقود شده هستند، حذف کنیم. با این حال، این عمل باید با احتیاط انجام شود، زیرا ویژگی‌های حذف شده ممکن است از ویژگی‌های مهم تجزیه و تحلیل باشند.

- **برآورد مقادیر مفقود شده:** گاهی اوقات داده‌های مفقود شده را می‌توان به‌طور قابل اعتماد تخمین زد. به عنوان مثال، یک سری زمانی را در نظر بگیرید که به‌طور منطقی تغییر می‌کند، اما دارای چند نقطه مفقود شده پراکنده است. در چنین مواردی می‌توان مقادیر مفقوده را با استفاده از مقادیر باقی‌مانده برآورد (درون‌یابی[1]) کرد. به عنوان مثال دیگر، یک مجموعه داده را در نظر بگیرید که دارای نقاط داده مشابه بسیاری است. در این وضعیت، اغلب از مقادیر ویژگی نقاط نزدیک به نقطه با مقدار مفقود شده برای برآورد مقدار از دست رفته استفاده می‌شود. اگر صفت پیوسته باشد، از میانگین مقدار ویژگی نزدیکترین همسایگان استفاده می‌شود. اگر صفت از نوع گسسته باشد، می‌توان متداول‌ترین مقدار ویژگی را در نظر گرفت.

- **نادیده گرفتن مقادیر گم شده در هنگام تجزیه و تحلیل:** رویکرد دیگر در برخورد با داده‌های مفقود شده، نادیده گرفتن این مقادیر است. به عنوان مثال، فرض کنید اشیا در حال خوشه‌بندی هستند و شباهت بین جفت شی داده باید محاسبه شود. اگر یک یا هر دو شی، دارای مقادیر مفقود شده برای برخی از ویژگی‌ها باشند، می‌توان شباهت را فقط با استفاده از ویژگی‌هایی که مقادیر مفقود شده ندارند محاسبه کرد.

---

[1] interpolated

### دادههای دورافتاده[1]

در یادگیری ماشین کیفیت دادهها به اندازه کیفیت مدل پیشگویانه یا طبقهبندی اهمیت دارد. با این حال، برخی اوقات در مجموعه داده، دادههایی ثبت شدهاند که بهطور چشمگیری با سایر موارد متفاوت هستند، آنها خود را در یک یا چند ویژگی متمایز میکنند. این دادهها که به عنوان دادههای دورافتاده شناخته میشوند، میتوانند (و احتمالاً خواهد داشت) باعث ایجاد ناهنجاری در نتایج بدست آمده از طریق الگوریتمها و سیستمهای تحلیلی شوند. در مدلهای بانظارت، دادههای دورافتاده میتوانند فرآیند آموزش را فریب دهند که این امر میتواند منجر به طولانی شدن زمان آموزش و یا منجر به ایجاد مدلهایی با دقت کم شود.

تفسیرپذیری یک مدل با مقادیر دورافتاده و شناخت دادههای دورافتاده در تجزیه و تحلیل دادهها دارای دو جنبه بسیار مهم است. اول اینکه، ممکن است با وجود دادههای دورافتاده، کل نتیجه یک تجزیه و تحلیل سوگیری منفی داشته باشد. دوم اینکه، ممکن است رفتار دادههای دورافتاده دقیقا همان چیزی است که به دنبال آن هستیم. در واقع، برخی اوقات دادههای دورافتاده میتوانند شاخصهای مفیدی باشند. به عنوان مثال، در برخی از کاربردهای تجزیه و تحلیل دادهها مانند تشخیص تقلب در کارت اعتباری، تجزیه و تحلیل دادههای دورافتاده اهمیت پیدا میکند، چراکه در اینجا استثنا و نه قاعده ممکن است برای تحلیلگر جالب باشد.

> یک داده دورافتاده را نمیتوان **نویز** یا **خطا** در نظر گرفت. با این حال، آنها مشکوک هستند که با روش مشابه بقیه دادهها (شیها) تولید نشدهاند.

### علل ایجاد دادههای دورافتاده

موارد زیر برخی از دلایل متداول وجود نقاط دورافتاده در یک مجموعه داده مشخص میباشد:

- **خطای اندازهگیری (خطای ابزارها):** زمانی ایجاد میشود که ابزار اندازهگیری مورد استفاده معیوب باشد.
- **خطای ورود دادهها (خطاهای انسانی):** خطاهای انسانی همانند خطاهایی که در حین جمعآوری، ضبط یا ورود دادهها ایجاد میشود، میتوانند باعث ایجاد فاصله زیاد در دادهها شوند.
- **خطای تجربی:** این خطاها در حین استخراج دادهها یا هنگام انجام آزمایش ایجاد میشود.
- **خطای پردازش دادهها:** هنگام دستورزی یا استخراج مجموعه دادهها ایجاد میشود.
- **خطای نمونهگیری:** این خطا زمانی اتفاق میافتد که فرد دادهها را از منابع اشتباه یا منابع مختلف استخراج یا مخلوط کند.

---

[1] Outliers

- **عمدی:** اینها موارد بیرونی ساختگی هستند که برای آزمایش روش‌های تشخیص ساخته شده‌اند.

- **طبیعی:** در فرآیند تولید، جمع‌آوری، پردازش و تجزیه و تحلیل داده‌ها، نقاط دورافتاده می‌توانند از منابع مختلف آمده و در ابعاد مختلف پنهان شوند. آنهایی که محصول خطا نیستند، دورافتاده طبیعی نامیده می‌نامند.

## تاثیر داده‌های دورافتاده در تجزیه و تحلیل

داده‌های دورافتاده تاثیر زیادی بر نتیجه تجزیه و تحلیل داده‌ها دارند. برخی از رایج‌ترین اثرات به شرح زیر است:

- **ممکن است تاثیر معنی‌داری بر میانگین و انحراف معیار داشته باشند.**
- **اگر پراکندگی نقاط دورافتاده به‌صورت تصادفی توزیع نشوند، می‌توانند هنجار[1] بودن را کاهش دهند.**
- **آنها می‌توانند باعث بایاس (سوگیری) یا تحت تاثیر قرار دادن برآوردها شوند.**
- **آنها می‌توانند بر فرض اساسی رگرسیون و سایر مدل‌های آماری تاثیر بگذارند.**

## انواع داده‌های دورافتاده

در علم داده و آمار، به‌طور کلی داده‌های دورافتاده را به سه دسته اصلی تقسیم‌بندی می‌کنند:

1. **داده‌های دورافتاده سراسری[2] (ناهنجاری نقطه):** اینها ساده‌ترین اشکال داده‌های دورافتاده هستند. اگر در یک مجموعه داده مشخص، یک نقطه داده به شدت از بقیه نقاط داده منحرف شود، به عنوان یک خروجی سراسری شناخته می‌شود. به عنوان مثال، در سیستم تشخیص نفوذ، اگر تعداد زیادی بسته در مدت زمان بسیار کوتاهی پخش شود، این ممکن است به عنوان یک خروجی سراسری در نظر گرفته شود و می‌توان گفت که آن سیستم خاص به طور بالقوه هک شده است.

2. **داده‌های دورافتاده محتوا محور[3] (مشروط):** اگر در یک مجموعه داده مشخص، یک شی داده فقط بر اساس یک محتوا (زمینه) یا شرایط خاص از سایر نقاط داده منحرف شود. یک نقطه داده ممکن است به دلیل یک وضعیت خاص بسیار فاصله داشته باشد و در شرایط دیگر رفتار عادی از خود نشان دهد. بنابراین، یک محتوا باید به عنوان بخشی از بیان مشکل مشخص شود تا نقاط دورافتاده محتوایی شناسایی شوند. ویژگی‌های نقاط داده بر اساس ویژگی‌های محتوایی و رفتاری تصمیم‌گیری می‌شود. به عنوان مثال، دمای

---

[1] normality

[2] Global outliers

[3] Contextual outliers

۴۰ درجه سانتی‌گراد ممکن است در محتوای "فصل زمستان" به عنوان یک دمای دورافتاده عمل کند، اما در محتوای "فصل تابستان" مانند یک نقطه داده معمولی رفتار می‌کند.

۳. **داده‌های دورافتاده جمعی**[1]: اگر در یک مجموعه داده مشخص، برخی از نقاط داده، از بقیه مجموعه داده به‌طور قابل توجهی منحرف شوند، ممکن است به عنوان نقاط دور افتاده جمعی نامیده شوند. باید توجه داشت این مقادیر نقاط داده، به‌صورت فردی به لحاظ محتوایی یا سراسری غیرعادی نیستند. برای تشخیص این نقاط دورافتاده، ممکن است به داده‌های پیشین در مورد رابطه بین آن اشیاء داده که رفتارهای دورافتاده را نشان می‌دهد، نیاز داشته باشیم.

**تشخیص نقاط دورافتاده**

روش‌های گوناگونی برای یافتن نقاط دورافتاده وجود دارد. همه این روش‌ها از رویکردی برای یافتن مقادیری استفاده می‌کنند که در مقایسه با سایر مجموعه داده غیر معمول است. در این‌جا ما تنها چند مورد از این تکنیک‌ها به شرح زیر فهرست کرده‌ایم:

▪ **مرتب‌سازی.** مرتب‌سازی ساده‌ترین تکنیک برای تجزیه و تحلیل داده‌های دورافتاده است. مجموعه داده خود را در هر نوع ابزار دست‌ورزی داده، مانند صفحه‌گسترده (یا جدول) بارگذاری کنید و مقادیر را بر اساس اندازه آن‌ها مرتب کنید. سپس، محدوده مقادیر نقاط مختلف داده را بررسی کنید. اگر هر نقطه داده به‌طور قابل توجهی بالاتر یا پایین‌تر از نقاط دیگر در مجموعه داده باشد، ممکن است به‌عنوان یک مورد دورافتاده تلقی شود. *روش مرتب‌سازی داده‌ها بر روی مجموعه داده کوچک بسیار مؤثر است.*

▪ **با استفاده از نمودارها.** یکی دیگر از تکنیک‌های تجزیه و تحلیل داده‌های دورافتاده، نمودار است. تمام نقاط داده را روی یک نمودار ترسیم کنید و ببینید کدام نقاط از بقیه فاصله دارند. با استفاده از رویکرد رسم نمودار در مقایسه با رویکرد مرتب‌سازی، می‌توانیم نقاط داده بیشتری را تجسم کنیم که مشاهده دورافتاده‌ها را آسان می‌کند. می‌توانیم نقاط دورافتاده را با استفاده از نمودار **جعبه‌ای**[2]، **بافت‌نگار**[3] و نمودار **نقطه‌ای**[4] تشخیص دهیم.

▪ **با استفاده از نمره z.** نمره z (نمره استاندارد) اندازه‌گیری یک رابطه نقطه‌ای با میانگین تمام نقاط در مجموعه داده‌ها است. زمانی که امتیازدهی به انجام رسید، مقادیر یک عدد مثبت

---

یا منفی دریافت می‌کنند. با محاسبه نمره $z$ [1] برای هر نقطه داده، به راحتی می‌توان دید که کدام نقاط داده‌ها به‌طور متوسط از میانگین فاصله می‌گیرند. این روش فرض می‌کند که متغیر دارای توزیع گاوسی است.

### داده‌های تکراری[2]

مشاهدات تکراری اغلب در طول جمع‌آوری داده‌ها اتفاق می‌افتد. هنگامی که مجموعه داده‌ها را از چندین جا ترکیب می‌کنید، داده‌ها را از طریق وب تراش جمع‌آوری می‌کنید و یا داده‌ها را از مشتریان یا چندین شعبه مختلف دریافت می‌کنید، فرصت‌هایی برای ایجاد داده‌های تکراری وجود دارد. حذف این داده‌های تکراری یکی از بزرگ‌ترین زمینه‌هایی است که باید در این فرآیند مورد توجه قرار گیرد.

### داده‌های بی‌ربط[3]

داده‌های بی‌ربط یا غیرضروری آن‌هایی هستند که در واقع مورد نیاز نیستند و در چارچوب مساله‌ای که سعی داریم آن را حل کنیم، مناسب نیستند. به عنوان مثال، اگر ما داده‌های مربوط به سلامت عمومی مردم را تجزیه و تحلیل می‌کنیم، شماره تلفن داده‌ای بی‌ربط است. به عنوان مثالی دیگر، اگر می‌خواهید داده‌های مربوط به نسل تازه را تجزیه و تحلیل کنید، اما مجموعه داده شما شامل نسل‌های قدیمی‌تر است، باید آن مشاهدات بی‌ربط را حذف کنید. این امر می‌تواند تجزیه و تحلیل را کارآمدتر کرده و سردرگمی را از هدف اصلی شما به حداقل برساند.

تنها در صورتی که مطمئن شده‌اید یک تکه داده (ویژگی) بی‌اهمیت است، می‌توانید آن را حذف کنید. در غیر این صورت، ماتریس همبستگی بین متغیرهای ویژگی را کشف کنید و حتی اگر متوجه هیچ ارتباطی نشده‌اید، باید از فردی که متخصص آن حوزه است سوال کنید. شاید ویژگی‌ای که از منظر شما بی‌ربط به نظر می‌رسد، از منظر حوزه‌ای مانند دیدگاه بالینی بسیار مرتبط باشد.

## تبدیل داده‌ها[4]

داده‌ها باید به گونه‌ای تبدیل شوند تا برای یک سیستم خوانا و سازگار باشند. در ادامه برخی از فرآیندهای مهمی که برای تبدیل داده‌ها مورد استفاده قرار می‌گیرد را شرح خواهیم داد.

---

[1] $z = \frac{x - \mu}{\sigma}$

که در این معادله، $x$ نمره‌ی خام، $\mu$ میانگین جمعیت و $\sigma$ انحراف معیار جمعیت است.

[2] Duplicate Data

[3] Irrelevant data

[4] Data transformation

**تجمیع**[1]

تجمیع داده‌ها روشی است که در آن داده‌های خام جمع‌آوری شده و به صورت خلاصه برای تجزیه و تحلیل استفاده می‌شود. به عنوان مثال، داده‌های خام را می‌توان در یک دوره زمانی معین جمع‌آوری کرد تا آماری همانند میانگین، حداقل، حداکثر و مجموع را ارائه دهد. پس از تجمیع داده‌ها و نوشتن آن‌ها به عنوان یک گزارش، می‌توانید داده‌های تجمیعی را تجزیه و تحلیل کنید تا در مورد منابع خاص، بینش کسب کنید. به عبارت دیگر، تجمیع داده‌ها می‌تواند تحلیل‌گران را قادر سازد تا در بازه زمانی معقول به حجم زیادی از داده‌ها دسترسی داشته و آن‌ها را بررسی کنند. یک ردیف از داده‌های تجمیعی می‌تواند صدها، هزاران یا حتی بیشتر از داده‌های ریز را نشان دهد. مجموعه‌ای از داده‌ها را در نظر بگیرید که شامل تراکنش‌هایی است که میزان فروش روزانه محصولات در مکان‌های مختلف یک فروشگاه برای روزهای در طی یک سال را ثبت می‌کند. یک راه برای تجمیع تراکنش‌های این مجموعه داده، تعویض تمام تراکنش یک فروشگاه با یک تراکنش واحد است. این امر باعث کاهش صدها یا هزاران تراکنشی می‌شود که روزانه در یک فروشگاه خاص رخ می‌دهد و تعداد اشیا داده به تعداد فروشگاه‌ها کاهش می‌یابد.

نمونه‌هایی از داده‌های تجمیعی شامل موارد زیر است:

- میزان مشارکت رای‌دهندگان بر اساس استان یا شهرستان. سوابق رای دهندگان جداگانه ارائه نمی‌شود، تنها مجموع آرا برای یک نامزد در یک منطقه خاص است.
- میانگین سن مشتری بر اساس محصول. هر مشتری جداگانه شناسایی نمی‌شود، اما برای هر محصول، میانگین سن مشتری ذخیره می‌شود.
- تعداد مشتریان بر اساس کشور. به جای بررسی هر مشتری، شماری از مشتریان هر کشور ارائه می‌شود.

**گسسته‌سازی**[2]

ما اغلب با داده‌هایی روبرو هستیم که از فرآیندهای پیوسته مانند دما، نور محیط و قیمت سهام یک شرکت جمع‌آوری می‌شوند. اما گاهی اوقات نیاز داریم که این مقادیر پیوسته را به قسمت‌های قابل کنترل‌تری تبدیل کنیم (چراکه برخی از الگوریتم‌های یادگیری ماشین، به ویژه الگوریتم‌های دسته‌بندی مستلزم این هستند که داده‌ها به صورت ویژگی‌های دسته‌ای باشند). نگاشت داده‌ها از مقادیر پیوسته به مقادیر گسسته، گسسته‌سازی نامیده می شود. به عبارت دقیق‌تر، گسسته‌سازی داده‌ها روشی برای تبدیل مقادیر ویژگی داده‌های پیوسته به مجموعه‌ای محدود از فواصل با

---

[1] Aggregation

[2] Discretization

حداقل از دست دادن داده است. می‌توانیم این مفهوم را باکمک یک مثال درک کنیم. فرض کنید ما یک ویژگی همانند سن با مقادیر داده شده به صورت زیر داشته باشیم:

۱،۵،۹،۴،۷،۱۱،۱۴،۱۷،۱۳،۱۸،۱۹،۳۱،۳۳،۳۶،۴۲،۴۴،۴۶،۷۰،۷۴،۷۸،۷۷ | سن

جدول زیر این داده‌ها را پس از گسسته‌سازی نشان می‌دهد:

| ویژگی | سن | سن | سن | سن |
|---|---|---|---|---|
| | ۱،۵،۴،۹،۷ | ۱۱،۱۴،۱۷،۱۳،۱۸،۱۹ | ۳۱،۳۳،۳۶،۴۲،۴۴،۴۶ | ۷۰،۷۴،۷۷،۷۸ |
| بعد ازگسسته‌سازی | بچه | جوان | بزرگسال | سالمند |

## هموارسازی[1]

هموارسازی داده‌ها با استفاده از الگوریتم‌های تخصصی برای حذف نویز از مجموعه داده انجام می‌شود. این فرآیند اجازه می‌دهد تا الگوهای مهم داده‌ها برجسته شوند. هموارسازی داده‌ها می‌تواند در پیش‌بینی روندها کمک کند. هرچند هموارسازی داده‌ها می‌تواند به افشای الگوها در داده‌های پیچیده کمک کند، با این حال، هموارسازی داده‌ها لزوما تفسیری از موضوع یا الگوهایی که به تشخیص آن کمک می‌کند را ارائه نمی‌دهد. گاهی اوقات هموارسازی داده‌ها ممکن است نقاط داده قابل استفاده را حذف کند. اگر مجموعه داده‌ها فصلی باشند و به‌طور کامل منعکس‌کننده واقعیت تولید شده توسط نقاط داده نباشند، ممکن است منجر به پیش‌بینی‌های نادرست شود.

## مقیاس‌بندی[2]

یادگیری ماشین همانند تهیه یک آب‌میوه مخلوط است. اگر می‌خواهیم بهترین آب‌میوه را بدست آوریم، باید همه میوه‌ها را نه بر اساس اندازه آن‌ها بلکه بر اساس نسبت مناسب آن‌ها مخلوط کنیم. به طور مشابه، در بسیاری از الگوریتم‌های یادگیری ماشین، برای آوردن همه ویژگی‌ها در یک وضعیت، باید مقیاس‌بندی را انجام دهیم تا یک عدد قابل توجه مدل را فقط به دلیل اندازه بزرگ آن‌ها تحت تاثیر قرار ندهد. **مقیاس‌بندی ویژگی‌ها**[3] در یادگیری ماشین یکی از مهم‌ترین مراحل در حین پیش‌پردازش داده‌ها قبل از ایجاد مدل یادگیری ماشین است. مقیاس‌بندی می‌تواند بین یک مدل یادگیری ماشین ضعیف و یک مدل بهتر تفاوت ایجاد کند.

متداول‌ترین تکنیک‌های مقیاس‌بندی ویژگی‌ها متعارف‌سازی و هنجارسازی هستند. هنجارسازی زمانی استفاده می‌شود که بخواهیم مقادیر خود را بین دو عدد، معمولا بین [۰،۱]

---

[1] Smoothing

[2] Scaling

[3] Feature scaling

یا [۱، ۱-] محدود کنیم. در حالی که متعارف‌سازی، داده‌ها را به میانگین صفر و واریانس ۱ تبدیل می‌کند.

**دلیل مقیاس‌بندی داده‌ها؟**

الگوریتم‌های یادگیری ماشین تنها اعداد را می‌بینند. از این‌رو، اگر تفاوت وسیعی در محدوده‌ی اعداد وجود داشته باشد این فرض اساسی را ایجاد می‌کنند که اعداد در محدوده بالاتر، برتری‌هایی دارند. بنابراین این تعداد قابل‌توجه شروع به ایفای نقش تعیین‌کننده‌تری در حین آموزش مدل می‌کند. علاوه‌بر این، الگوریتم‌های یادگیری ماشین روی اعداد کار می‌کنند و نمی‌دانند که این عدد نشان‌دهنده چه چیزی است. وزن ۱۰ گرم و قیمت ۱۰ دلار کاملا دو چیز متفاوت را نشان می‌دهد، چیزی که برای انسان ها بدیهی و روشن است، اما برای یک مدل به‌عنوان یک ویژگی، هر دو را یکسان تلقی می‌کند. فرض کنید دو ویژگی وزن و قیمت داشته باشیم که مقادیر وزن نسبت به قیمت اعداد بزرگ‌تری داشته باشند. از این‌رو، این فرض برای الگوریتم بوجود می‌آید که از آنجا که وزن از قیمت بزرگ‌تر است، بنابراین وزن مهم‌تر از قیمت است. به همین دلیل، این تعداد قابل توجه‌تر هنگام آموزش مدل، نقش تعیین‌کننده‌تری را ایفا می‌کنند. بنابراین، مقیاس‌بندی ویژگی‌ها برای آوردن همه ویژگی‌ها در یک وضعیت بدون هیچ‌گونه اهمیت اولیه نیاز است.

از دیگر دلایلی که مقیاس‌پذیری ویژگی‌ها اعمال می‌شود این است که تعداد کمی از الگوریتم‌ها همانند گرادیان کاهشی شبکه عصبی با مقیاس‌بندی ویژگی بسیار سریع‌تر از بدون آن همگرا می‌شود (شکل ۳ ـ ۱).

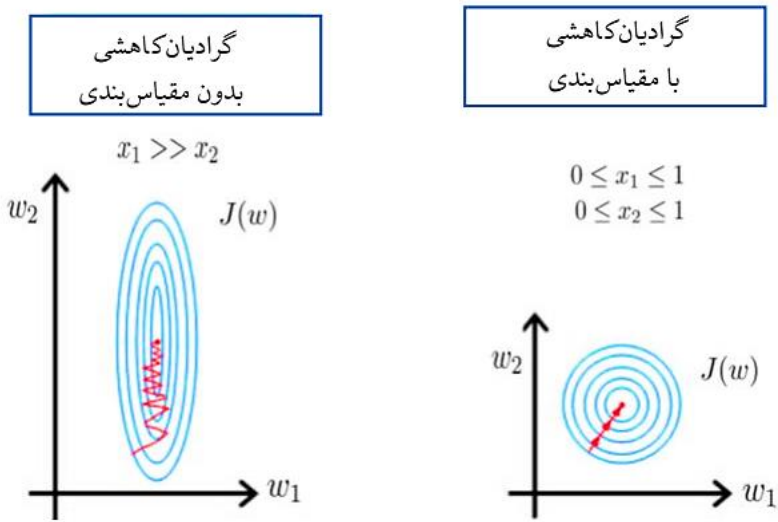

**شکل ۳ ـ ۱** گرادیان کاهشی با/بدون مقیاس‌بندی

## متعارف‌سازی[1]

متعارف‌سازی (استانداردسازی) یک تکنیک مهم است که بیشتر به عنوان یک مرحله پیش پردازش قبل از بسیاری از مدل‌های یادگیری ماشین انجام می‌شود تا محدوده ویژگی‌های مجموعه داده‌های ورودی را متعارف (استاندارد) کند. متعارف‌سازی زمانی ظاهر می‌شود که ویژگی‌های مجموعه داده‌های ورودی تفاوت‌های زیادی بین محدوده خود داشته باشند. به عبارت ساده‌تر، هنگامی که داده‌ها در واحدهای اندازه‌گیری مختلف اندازه‌گیری می‌شوند (به عنوان مثال، کیلوگرم، متر، کیلومتر و غیره). این تفاوت‌ها در محدوده ویژگی‌های اولیه باعث ایجاد مشکل در بسیاری از مدل‌های یادگیری ماشین می‌شود. به عنوان مثال، برای مدل‌هایی که مبتنی‌بر محاسبه فاصله هستند، اگر یکی از ویژگی‌ها دارای دامنه وسیعی از مقادیر باشد، فاصله توسط ویژگی خاصی تنظیم می‌شود. فرض کنید ما یک مجموعه داده دو بعدی با دو ویژگی ارتفاع برحسب متر و وزن برحسب کیلوگرم داریم که به ترتیب بین [۱ تا ۲] متر و [۳۰ تا ۹۰] کیلوگرم متغیر است. صرف نظر از این که بر اساس این مجموعه داده چه مدل مبتنی‌بر فاصله‌ای را انجام می‌دهید، ویژگی وزن بر ویژگی ارتفاع غالب می‌شود و سهم بیشتری در محاسبه فاصله خواهد داشت؛ تنها به این دلیل که مقادیر بیشتری در مقایسه با ارتفاع دارد. بنابراین، برای جلوگیری از این مشکل و راه حل آن، تبدیل ویژگی‌ها به مقیاس‌های قابل مقایسه با استفاده از متعارف‌سازی است.

## چگونه داده‌ها را متعارف کنیم؟

نمره Z که همچنین نمره متعارف[2] نامیده می‌شود یکی از رایج‌ترین روش‌ها برای متعارف‌سازی داده‌ها است که می‌توان با کسر میانگین و تقسیم آن بر انحراف معیار برای هر مقدار از هر ویژگی این کار را انجام داد. معادله ریاضی آن به‌صورت زیر می‌باشد:

$$z = \frac{x - \mu}{\sigma}$$

که در این معادله، $x$ نمره‌ی خام، $\mu$ میانگین جمعیت و $\sigma$ انحراف معیار جمعیت است.

پس از اتمام متعارف‌سازی، همه‌ی ویژگی‌ها دارای میانگین **صفر**، انحراف معیار **یک** و در نتیجه، مقیاس یکسان خواهند بود.

## چه زمانی داده‌ها را متعارف کنیم؟

---

همان‌طور که پیش‌تر به آن اشاره گردید، برای مدل‌های مبتنی‌بر فاصله، متعارف‌سازی انجام
می‌شود تا از ویژگی‌های با محدوده وسیع‌تر در جهت غلبه‌بر معیار فاصله جلوگیری شود. با این
حال، دلیل متعارف‌سازی داده‌ها برای همه‌ی مدل‌های یادگیری ماشین یکسان نیست و از یک
مدل به مدل دیگر متفاوت است. برخی از توسعه دهندگان یادگیری ماشین تمایل دارند داده‌های
خود را کورکورانه قبل از مدل یادگیری ماشین متعارف کنند، بدون اینکه تلاش کنند بفهمند
چرا باید از این روش استفاده شود. بنابراین، قبل از استفاده از هر یک مدل‌ها و روش‌های
یادگیری ماشین بهتر است بدانیم چه زمانی و چرا باید از متعارف‌سازی استفاده کنیم:

۱. **قبل از PCA:** در تحلیل مولفه اصلی[۱] (PCA)، ویژگی‌های با واریانس زیاد/محدوده
وسیع، وزن بیشتری نسبت به آن‌هایی که دارای واریانس کم هستند، بدست می‌آورند و
در نتیجه، آن‌ها به طور غیرمعقولی بر اولین اجزای اصلی تسلط می‌یابند. متعارف‌سازی
می‌تواند با ارائه وزن یکسان به همه ویژگی‌ها از این امر جلوگیری کند.

۲. **قبل از خوشه‌بندی:** مدل‌های خوشه‌بندی، الگوریتم‌های مبتنی‌بر فاصله هستند که
به‌منظور اندازه‌گیری شباهت بین مشاهدات، از معیار فاصله استفاده می‌کنند. بنابراین،
ویژگی‌های با محدوده بالا تاثیر بیشتری بر روی خوشه‌بندی خواهند داشت. از این‌رو،
متعارف‌سازی قبل از ایجاد یک مدل خوشه‌بندی مورد نیاز است.

۳. **قبل از KNN:** k ـ نزدیک‌ترین همسایگان[۲] یک الگوریتم دسته‌بندی مبتنی‌بر فاصله
است که مشاهدات جدید را بر اساس معیارهای تشابه (به عنوان مثال، معیارهای
فاصله) با مشاهدات برچسب‌گذاری شده از مجموعه آموزشی دسته‌بندی می‌کند.
متعارف‌سازی باعث می‌شود که همه متغیرها به یک اندازه در اندازه‌گیری تشابه سهیم
باشند.

۴. **قبل از SVM:** ماشین بردار پشتیبان[۳] سعی می‌کند فاصله بین صفحه جدا کننده و
بردارهای پشتیبان را حداکثر کند. اگر یک ویژگی دارای مقادیر بسیار بزرگ باشد،
هنگام محاسبه فاصله بر سایر ویژگی‌ها تسلط می‌یابد. بنابراین متعارف‌سازی به همه
ویژگی‌ها تاثیر یکسانی در معیار فاصله می‌دهد.

---

[۱] Principal Component Analysis

[۲] k-nearest neighbors

[۳] Support Vector Machine

## هنجارسازی[۱]

هنجارسازی بخشی از تکنیک‌های پیش‌پردازش و تمیزسازی داده‌ها و به عبارت کلی‌تر، نوعی مقیاس‌بندی ویژگی است. هدف اصلی این تکنیک این است که داده‌ها در همه‌ی رکوردها و زمینه‌ها یکدست شوند (بدون این‌که در محدوده مقادیر تفاوت ایجاد شود). این به ایجاد ارتباط بین داده‌های ورودی کمک می‌کند که به نوبه خود به تمیزسازی و بهبود کیفیت داده‌ها کمک می‌کند. این نوع از مقیاس‌بندی زمانی مورد استفاده قرار می‌گیرد که داده‌ها دارای گستره متنوعی (ویژگی‌ها دارای دامنه‌های متفاوت) باشند و الگوریتم‌هایی که بر روی آن‌ها آموزش داده می‌شود در مورد توزیع داده‌ها پیش‌فرض ایجاد نمی‌کند (همانند شبکه‌های عصبی).

هنجارسازی وزن/اهمیت مساوی به هر متغیر می‌دهد به‌طوری که هیچ متغیر واحدی عملکرد مدل را در یک جهت منحرف نمی‌کند؛ فقط به این دلیل که آن‌ها تعداد بیشتری هستند. رایج‌ترین و پرکاربردترین تکنیک هنجارسازی **مقیاس‌بندی مجدد**[۲] است که همچنین به عنوان هنجارسازی حداقل ـ حداکثر[۳] شناخته می‌شود و به صورت زیر محاسبه می‌شود:

$$\acute{x} = \frac{x - min(x)}{max(x) - min(x)}$$

## متعارف‌سازی یا هنجارسازی؟

هنجارسازی زمانی مفید است که بدانید توزیع داده‌های شما از توزیع گوسی (منحنی زنگوله‌ای) پیروی نمی‌کند. این می‌تواند در الگوریتم‌هایی مفید باشد که هیچ توزیعی از داده‌ها را فرض نمی‌کنند، همانند K ـ نزدیک‌ترین همسایه یا شبکه‌های عصبی. از طرف دیگر، متعارف‌سازی می‌تواند در مواردی مفید باشد که داده‌ها از توزیع گوسی پیروی می‌کنند (متعارف‌سازی فرض می‌کند که داده‌های شما دارای توزیع گوسی هستند). با این حال، لازم نیست که این امر لزوما درست باشد، اما اگر توزیع ویژگی شماگوسی باشد، این تکنیک مؤثرتر است. همچنین برخلاف هنجارسازی، متعارف‌سازی محدوده محدودیتی ندارد. بنابراین، حتی اگر در داده‌های خود مقدار زیادی داده‌ی دورافتاده داشته باشید، تحت تاثیر متعارف‌سازی قرار نخواهند گرفت. با این حال، انتخاب استفاده از هنجارسازی یا متعارف‌سازی بستگی به مساله شما و الگوریتم یادگیری ماشین شما دارد. هیچ قانون سختی وجود ندارد که به شما بگوید چه زمانی داده‌های خود را

---

[۱] normalization

[۲] Rescaling

[۳] min-max normalization

متعارف یا هنجار کنید. شما همیشه می‌توانید مدل خود را با داده‌های خام، هنجار و متعارف
شده برازش[1] کنید و عملکرد را برای بهترین نتایج مقایسه کنید.

> در حالی که هنجارسازی مقادیر اصلی را در محدوده خاص قرار می‌دهد، متعارف‌سازی
> آن‌ها را در یک توزیع که میانگین آن صفر و انحراف معیار آن یک است، قرار می‌دهد.

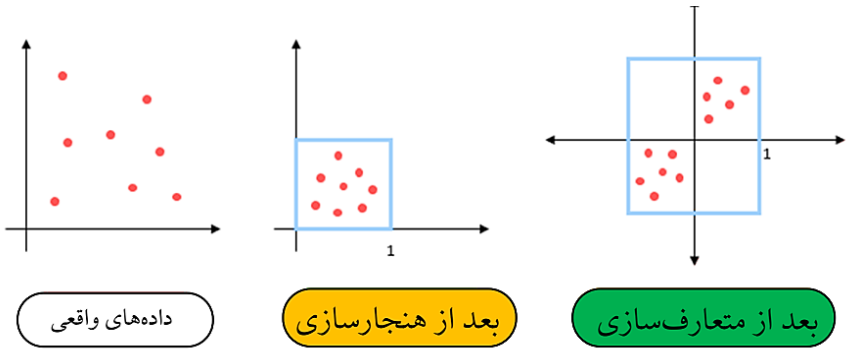

داده‌های واقعی | بعد از هنجارسازی | بعد از متعارف‌سازی

## مصورسازی داده‌ها

مصورسازی داده‌ها، نمایش داده‌ها یا اطلاعات در یک نمودار، گراف یا سایر قالب‌های بصری
است که ارتباط داده‌ها را به تصاویر منتقل می‌کند. مصورسازی داده‌ها یکی از مراحل مهم در
علم داده و یادگیری ماشین است، زیرا اجازه می‌دهد روندها و الگوها به راحتی دیده شوند. با
ظهور کلان داده، ما باید بتوانیم دسته‌های بزرگی از داده‌ها را تفسیر کنیم. ما به مصورسازی
داده‌ها نیاز داریم، چراکه خلاصه تصویری از داده‌ها، شناسایی الگوها و روندها را در مقایسه با
نگاه کردن به هزاران ردیف در یک صفحه گسترده آسان‌تر می‌کند. این شیوه کار مغز انسان است.
از آنجا که هدف از تجزیه و تحلیل داده‌ها کسب بینش است، داده‌ها وقتی مصور می‌شوند،
می‌توان اطلاعات بیشتری را از آن‌ها بدست آورد. حتی اگر یک تحلیل‌گر داده بتواند بدون
مصورسازی بینش‌هایی را از داده‌ها بیرون بکشد، انتقال مفهوم بدون مصورسازی دشوارتر
خواهد بود.

هنگامی که یک دانشمند داده در حال نوشتن الگوریتم‌های تجزیه و تحلیل پیشگویانه پیشرفته
است، مصورسازی خروجی‌ها برای نظارت بر نتایج و اطمینان از عملکرد مدل‌ها بسیار مهم
است. به این دلیل که مصورسازی الگوریتم‌های پیچیده به‌طور کلی آسان‌تر از خروجی‌های عددی
است. مصورسازی داده‌ها در پایتون شاید یکی از کاربردی‌ترین ویژگی‌هایی باشد که برای علم

---

[1] fitting

داده با پایتون در عصر امروز استفاده میشود. در ادامه این بخش پس از معرفی انواع نمودارهای مصورسازی دادهها و دلایل انتخاب هر یک از آنها، هر یک از این نمودارها را با کتابخانههای قدرتمند پایتون که در این زمینه وجود دارند، پیادهسازی میکنیم.

## اهمیت و مزایای مصورسازی دادهها

صرف نظر از اینکه چه حرفهای را انتخاب کردهاید، مصورسازی دادهها میتواند با نمایش دادهها به کارآمدترین شکل ممکن کمک کند. مصورسازی دادهها، دادههای خام را میگیرد، مدل میکند و دادهها را نمایش میدهد تا بتوان به نتیجهای رسید. این عمل میتواند به شرکتها کمک کند تا تشخیص دهند که کدام مناطق نیاز به بهبود دارند، چه عواملی بر رضایت و نارضایتی مشتریان تأثیر میگذارد و با محصولات خاص (کجا باید بروند و به چه کسانی باید فروخته شوند) چه کنند. دادههای مصورشده به ذینفعان، صاحبان مشاغل و تصمیمگیرندگان، پیشبینی بهتری از میزان فروش و رشد آینده میدهد.

مصورسازی دادهها بر تصمیمگیری سازمانها و شرکتها با نمایش بصری تعاملی دادهها تاثیر مثبت میگذارد. اکنون مشاغل میتوانند الگوها را سریعتر تشخیص دهند، چراکه میتوانند دادهها را به صورت گرافیکی یا تصویری تفسیر کنند. در اینجا به چند روش خاص اشاره میکنیم که مصورسازی دادهها میتواند به نفع یک سازمان باشد:

■ **همبستگی در روابط:** بدون مصورسازی دادهها، شناسایی همبستگی بین رابطه متغیرهای مستقل چالشبرانگیز است. با درک متغیرهای مستقل، میتوان تصمیمات تجاری بهتری گرفت.

■ **گرایشات (روندها) در گذر زمان:** این یکی از با ارزشترین برنامههای کاربردی مصورسازی دادهها است. پیشبینی بدون داشتن اطلاعات لازم از گذشته و حال غیرممکن است. گرایشات به مرور زمان به ما میگوید که کجا بودیم و بهطور بالقوه میتوانیم به کجا برسیم.

■ **بررسی بازار:** مصورسازی دادهها اطلاعات را از بازارهای مختلف میگیرد تا به شما بینشی دهد که توجه شما باید بر روی کدام مخاطبان متمرکز شود و از کدامیک باید دور بمانید. با نمایش این دادهها در نمودارهای مختلف، تصویر واضحتری از فرصتهای موجود در بازار بدست میآوریم.

■ **واکنش به بازار:** توانایی به دست آوردن سریع و آسان اطلاعات با دادههایی که به وضوح در میزکار عملکردی نمایش داده میشوند، به کسب و کارها اجازه میدهد تا به سرعت به یافتهها عمل کرده و به آنها پاسخ دهند و به جلوگیری از اشتباه کمک میکند.

سایر مزایای مصورسازی دادهها شامل موارد زیر است:

- توانایی جذب سریع اطلاعات، بهبود بینش و تصمیم‌گیری سریع‌تر
- درک بیشتر از گام‌های بعدی که باید برای بهبود سازمان برداشته شود
- توزیع آسان اطلاعات که فرصت به اشتراک گذاشتن بینش با همه افراد درگیر را افزایش می‌دهد.

## هدف از مصورسازی داده‌ها چیست؟

هدف مصورسازی داده‌ها کاملا واضح است؛ دادن مفهوم به داده‌ها و استفاده از اطلاعات برای منافع سازمان. علاوه بر این، داده‌ها پیچیده هستند و هر زمان که مصورسازی می‌شوند، ارزش بیشتری پیدا می‌کنند. بدون مصورسازی، یافتن سریع ارتباط از داده‌ها و شناسایی الگوها برای بدست آوردن بینش دشوار است. دانشمندان داده می‌توانند الگوها یا خطاها را بدون مصورسازی پیدا کنند. با این حال، انتقال یافته‌ها از داده‌ها و شناسایی اطلاعات حیاتی از آن‌ها بسیار مهم است. مصورسازی داده‌ها تاثیر پیام‌رسانی را برای مخاطبان شما تقویت می‌کند و نتایج تجزیه و تحلیل داده‌ها را به متقاعدکننده‌ترین حالت ارائه می‌دهد. مصورسازی داده‌ها به شما امکان می‌دهد حجم وسیعی از داده‌ها را در یک نگاه و به شیوه‌ای بهتر درک کنید. این به درک بهتر داده‌ها برای اندازه‌گیری تاثیر آن بر تجارت کمک می‌کند و بینش را به صورت بصری به مخاطبان داخلی و خارجی منتقل می‌کند.

## از چه نوع نمودار مصورسازی استفاده کنیم؟

قبل از شروع به بررسی انواع نمودارها، باید ۵ سوال مهم در مورد داده‌هایی که در اختیار دارید از خود بپرسید. این سوالات به شما کمک می‌کند تا داده‌های خود را بهتر درک کنید و از این رو، نوع نمودار مناسب را برای نشان دادن آن‌ها انتخاب کنید.

**۱.   موضوع (گزارشی) که داده‌های شما سعی در ارائه آن دارد چیست؟**

اولین چیزی که باید در مورد داده‌های خود بدانید این است که سعی دارد چه موضوع یا گزارشی را ارائه دهد؟ چرا این داده‌ها جمع‌آوری شد و چگونه؟ آیا داده‌های شما برای یافتن گرایشات جمع‌آوری شده‌اند؟ برای مقایسه گزینه‌های مختلف؟ آیا  توزیع را نشان می‌دهد؟ یا برای مشاهده رابطه بین مجموعه‌های مختلف مقادیر استفاده می‌شود؟ درک موضوع مبدا داده‌های شما و دانستن آنچه در تلاش است ارائه کند، انتخاب نوع نمودار را برای شما آسان‌تر می‌کند.

**۲.   نتایج خود را به چه کسی ارائه خواهید کرد؟**

بعد از اینکه موضوع پشت داده‌های خود را فهمیدید، در مرحله بعد، باید بدانید که نتایج خود را برای چه کسانی ارائه می‌دهید. اگر در حال تجزیه و تحلیل روندهای بازار سهام هستید و یافته‌های خود را به برخی از بازرگانان ارائه می‌دهید، ممکن است از نوع نمودار متفاوتی نسبت به زمانی که یافته خود را برای افرادی که شروع به کار در بازار سهام کرده‌اند، استفاده کنید. هدف کلی استفاده از مصورسازی داده‌ها این است که ارتباط داده‌ها کارآمدتر باشد. به همین دلیل، شما باید مخاطبان خود را بشناسید تا بتوانید بهترین نوع نمودار را برای نمایش داده‌های خود به آن‌ها انتخاب کنید.

## ۳. حجم داده‌های شما چقدر است؟

اندازه داده‌های شما به‌طور قابل توجهی بر نوع نموداری که استفاده می‌کنید تاثیر می‌گذارد. برخی از نمودارها نباید برای مجموعه داده‌های عظیم استفاده شوند، در حالی که برخی دیگر برای داده‌های بزرگ مناسب هستند. به عنوان مثال، نمودارهای دایره‌ای با مجموعه داده‌های کوچک بهتر کار می‌کنند. با این حال، اگر از مجموعه داده بزرگی استفاده می‌کنید، استفاده از نمودار نقطه‌ای (پراکندگی) منطقی‌تر خواهد بود. شما باید یک نوع نمودار را انتخاب کنید که متناسب با اندازه داده‌های شما باشد و بدون شلوغی، آن‌ها را به وضوح نشان دهد.

## ۴. نوع داده شما چیست؟

انواع مختلفی از داده‌ها وجود دارد، اسمی، ترتیبی، پیوسته یا گسسته. می‌توانید از نوع داده برای حذف برخی از انواع نمودار استفاده کنید. به عنوان مثال، اگر داده‌های پیوسته دارید، نمودار میله‌ای ممکن است بهترین انتخاب نباشد. ممکن است لازم باشد به جای آن از یک نمودار خطی استفاده کنید. به طور مشابه، اگر داده‌های گسسته دارید، استفاده از نمودار میله‌ای یا نمودار دایره‌ای ممکن است ایده خوبی باشد.

## ۵. عناصر مختلف داده‌های شما چگونه با یکدیگر ارتباط دارند؟

در نهایت، باید از خود بپرسید که عناصر مختلف داده‌های شما چگونه با هم ارتباط دارند. آیا ترتیب داده‌های شما بر اساس برخی عوامل همانند زمان، اندازه، نوع است؟ یا همبستگی بین متغیرهای مختلف؟ آیا داده‌های شما یک سری زمانی است؟ یا بیشتر توزیع است؟ رابطه‌ای بین مقادیر موجود در مجموعه داده شما بر تصمیم‌گیری برای انتخاب بهترین نمودار کمک می‌کند.

# انواع نمودارهای مصورسازی داده‌ها

اکنون که فهمیده‌ایم مصورسازی داده‌ها چیست و چه کاربردی دارد، بیایید انواع مختلف نمودارهایی که مصورسازی داده‌ها را انجام می‌دهد را بررسی کنیم.

## نمودار خطی

یک نمودار خطی برای نشان دادن تغییر داده در بازه زمانی پیوسته یا یک محدوده زمانی استفاده می‌شود. به عبارت دیگر، نمودار خطی به‌طور موثر زمانی مورد استفاده قرار می‌گیرد که می‌خواهیم روند را در طول زمان درک کنیم.

### چه زمانی از نمودار خطی استفاده کنیم؟

- اگر یک مجموعه داده پیوسته دارید که در طول زمان تغییر می‌کند.
- اگر مجموعه داده شما برای نمودار میله‌ای خیلی بزرگ است.
- زمانی که می‌خواهید روندها را برای دسته‌های مختلف در یک دوره زمانی یکسان نشان دهید و در نتیجه مقایسه را نشان دهید.
- اگر می‌خواهید به جای مقادیر دقیق، روندها را تجسم کنید.

### چه زمانی از نمودار خطی استفاده نکنیم؟

- نمودار خطی با مجموعه داده‌های بزرگ‌تر بهتر کار می‌کند، بنابراین اگر یک مجموعه داده کوچک داشته باشید، از نمودار میله‌ای به جای آن استفاده کنید.

### نمودار خطی با استفاده از Matplotlib

```python
# Importing packages
import matplotlib.pyplot as plt
# Define x and y values
x = [7, 14, 21, 28, 35, 42, 49]
y = [8, 13, 21, 30, 31, 44, 50]
# Plot a simple line chart without any feature
plt.plot(x, y)
plt.show()
```

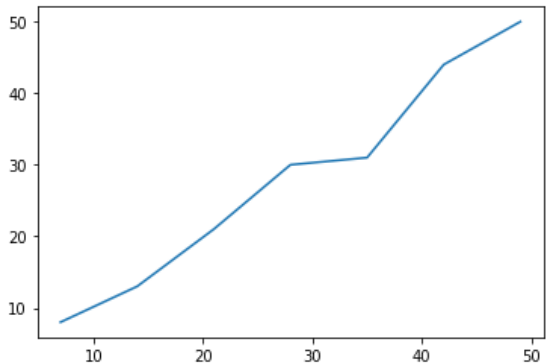

```
# Importing packages
import numpy as np
# Define x  value
x = np.random.randint(low=1, high=10, size=25)
plt.plot(x, linewidth=3)
plt.show()
```

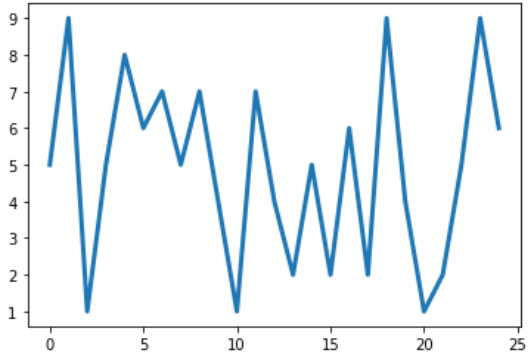

```
# Define x and y values
x = [7, 14, 21, 28, 35, 42, 49]
y = [8, 13, 21, 30, 31, 44, 50]
# Plot points on the line chart
plt.plot(x, y, 'o--', linewidth=2)
plt.show()
```

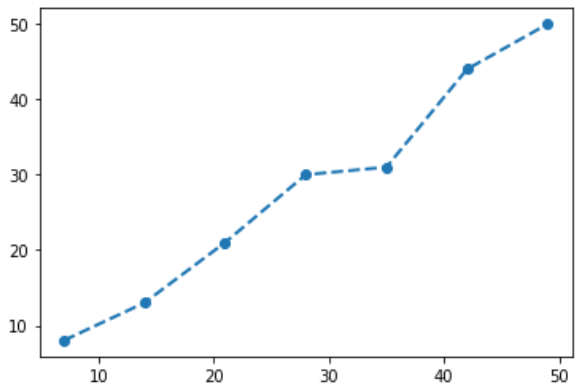

```
# Define x and y values
x = np.array([7, 11, 24, 28, 35, 34, 41])
y = np.array([8, 20, 13, 30, 31, 48, 50])

# Drawn a simple scatter plot for the data given
plt.scatter(x, y, marker='*', color='k')
```

```
# Generating the parameters of the best fit line
m, c = np.polyfit(x, y, 1)
# Plotting the straight line by using the generated parameters
plt.plot(x, m*x+c)
plt.show()
```

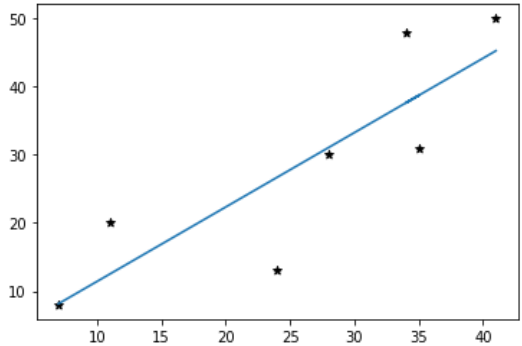

```
import pandas as pd
# Let's create a Dataframe using lists
Name = ['Sara', 'Mahsa', 'Zahra', 'Maryam', 'Ayda']
Score = ['19.02', '19.74', '18.34', '17.26', '19.87']
# Now, create a pandas dataframe using above lists
df_ = pd.DataFrame(
    {'Name' : Name, 'Score' : Score})
# Plotting the data from the dataframe created using matplotlib
plt.figure(figsize=(9, 5))
plt.plot(df_['Name'], df_['Score'], '-b', linewidth=2)
plt.xticks(rotation=60)
plt.xlabel('Name')
plt.ylabel('Score')
plt.show()
```

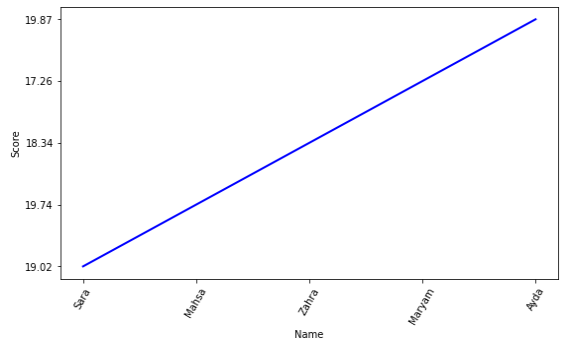

```python
#importing the required libraries
import matplotlib.pyplot as plt
import seaborn as sns
#Creating the dataset
df = sns.load_dataset("iris")
df=df.groupby('sepal_length')['sepal_width'].sum().to_frame().reset_index()
#Creating the line chart
plt.plot(df['sepal_length'], df['sepal_width'])
#Adding the aesthetics
plt.title('Chart title')
plt.xlabel('X axis title')
plt.ylabel('Y axis title')
#Show the plot
plt.show()
```

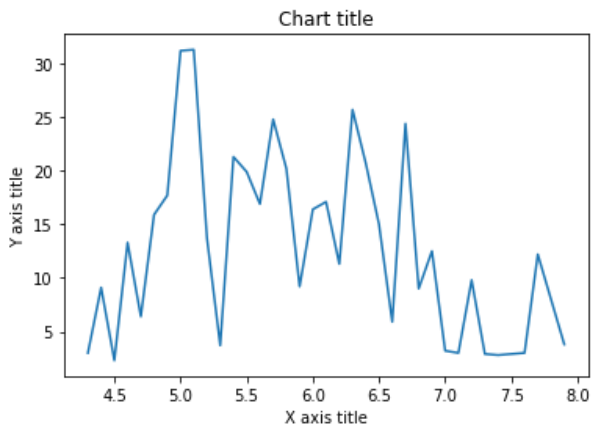

```python
import numpy as np
from mpl_toolkits import mplot3d

# Setting 3 axes for the graph
plt.axes(projection='3d')

# Define the z, y, x data
z = np.linspace(0, 1, 100)
x = 4.5 * z
y = 0.8 * x + 2

# Plotting the line
plt.plot(x, y, z, 'r', linewidth=2)
plt.title('Plot a line in 3D')
plt.show()
```

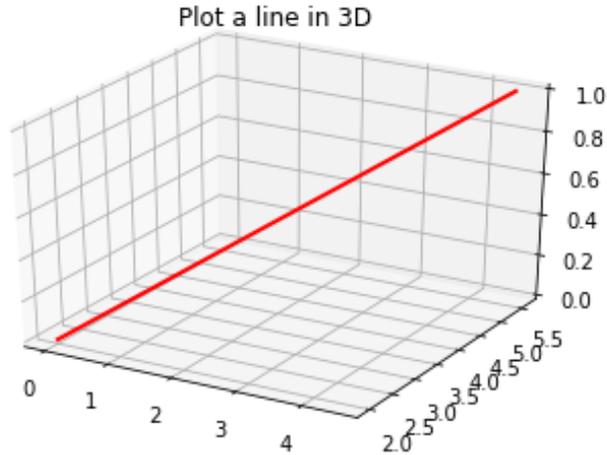

**نمودار خطی با استفاده از Seaborn**

```
#importing the required libraries
import seaborn as sns

# Define the  x and y data
x = [1, 2, 3, 4, 5]
y = [1, 5, 4, 7, 4]

sns.lineplot(x, y)
plt.show()
```

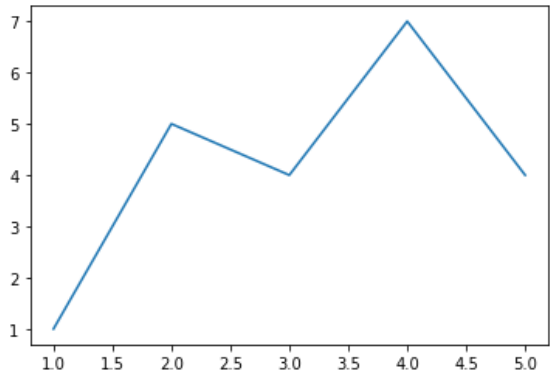

```
# Define the  x and y data
x = ['day 1', 'day 2', 'day 3']
y = [1, 5, 4]

sns.lineplot(x, y)
plt.show()
```

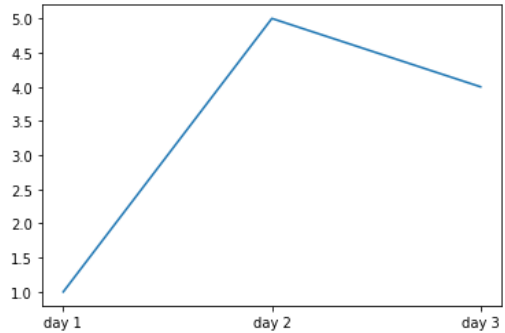

## نمودار میله‌ای

نمودار میله‌ای راهی برای نمایش مقادیر داده ارائه شده به صورت میله‌های افقی است و برای نمایش روند داده‌ها و مقایسه داده‌ها در زیر گروه‌های مختلف در کنار هم استفاده می‌شود.

**چه زمانی از نمودار میله‌ای استفاده کنیم؟**

- زمانی‌که نیاز به مقایسه چند دسته مختلف دارید.
- زمانی‌که باید نشان دهید که داده‌های بزرگ در طول زمان چگونه تغییر می‌کنند.
- اگر می‌خواهید مقادیر مثبت و منفی را در مجموعه داده نشان دهید.

**چه زمانی از نمودار میله‌ای استفاده نکنیم؟**

- اگر دسته‌های زیادی دارید. نمودار شما نباید بیش از ۱۰ میله داشته باشد.

**نمودار میله‌ای با استفاده از Matplotlib**

```python
import numpy as np
import matplotlib.pyplot as plt
# Dataset generation
objects = ('Python', 'C++', 'Julia', 'Go', 'Rust', 'c')
y_pos = np.arange(len(objects))
performance = [10,8,6,4,2,1]
#  Bar plot
plt.barh(y_pos, performance, align='center', alpha=0.5)
plt.yticks(y_pos, objects)
plt.xlabel('Usage')
plt.title('Programming language usage')
plt.show()
```

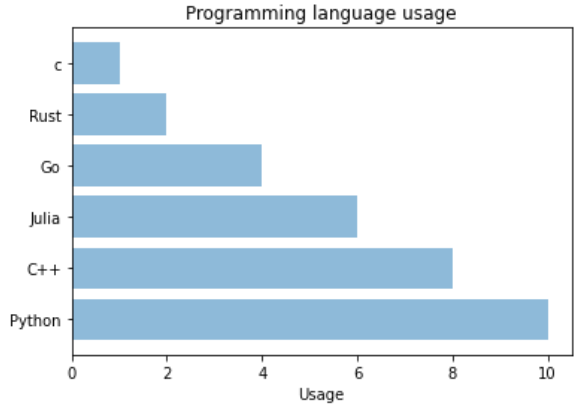

**نمودار میله‌ای با استفاده از Seaborn**

```
import matplotlib.pyplot as plt
import seaborn as sns

x = ['A', 'B', 'C']
y = [1, 5, 3]

sns.barplot(y, x)
plt.show()
```

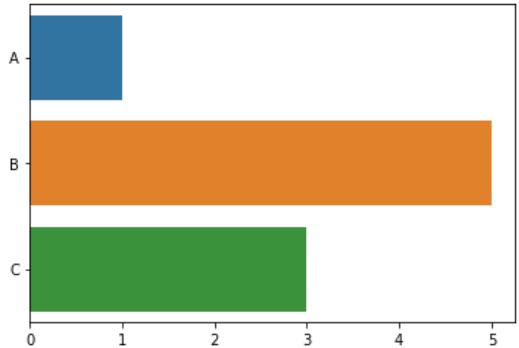

## نمودار ستونی

نمودار ستونی نوعی نمودار میله‌ای است که از میله‌های عمودی برای نشان دادن مقایسه بین دسته‌ها استفاده می‌کند. نمودارهای ستونی برای نشان دادن وضعیت در یک نقطه از زمان (مثلا تعداد محصولات فروخته شده در یک وب سایت) بهترین کاربرد را دارند. هدف اصلی آن‌ها جلب توجه به اعداد به جای روند است (روندها برای نمودار خطی مناسب‌تر هستند).

**چه زمانی از نمودار ستونی استفاده کنیم؟**

- زمانیکه نیاز دارید مقایسهایکنار هم، از مقادیر مختلف نشان دهید.
- زمانیکه میخواهید بر تفاوت بین ارزش ها تاکید کنید.
- زمانیکه میخواهید کل ارقام را به جای روندها برجسته کنید.

**چه زمانی از نمودار ستونی استفاده نکنیم؟**

- فقط برای مجموعه دادههای کوچک و متوسط مناسب است.
- تعداد ستونها نباید خیلی زیاد باشد.

**نمودار ستونی با استفاده از Matplotlib**

```python
import numpy as np
import matplotlib.pyplot as plt
# Dataset generation
data_dict = {'CSE':33, 'ECE':28, 'EEE':30}
courses = list(data_dict.keys())
values = list(data_dict.values())
fig = plt.figure(figsize = (10, 5))
# Bar plot
plt.bar(courses, values, color ='green',
        width = 0.5)
plt.xlabel("Courses offered")
plt.ylabel("No. of students enrolled")
plt.title("Students enrolled in different courses")
plt.show()
```

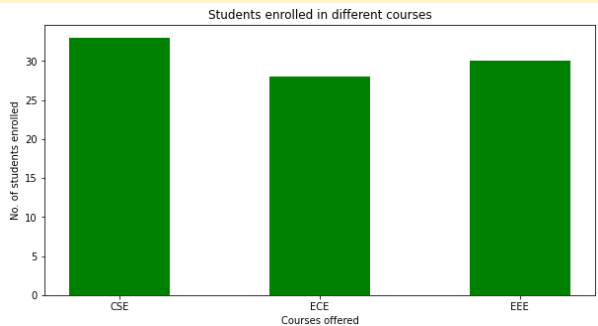

```python
import pandas as pd
plotdata = pd.DataFrame({

    "2018":[57,67,77,83],

    "2019":[68,73,80,79],

    "2020":[73,78,80,85]},
```

```
    index=["Django", "Gafur", "Tommy", "Ronnie"])
plotdata.plot(kind="bar",figsize=(15, 8))
plt.title("FIFA ratings")
plt.xlabel("Footballer")
plt.ylabel("Ratings")
```

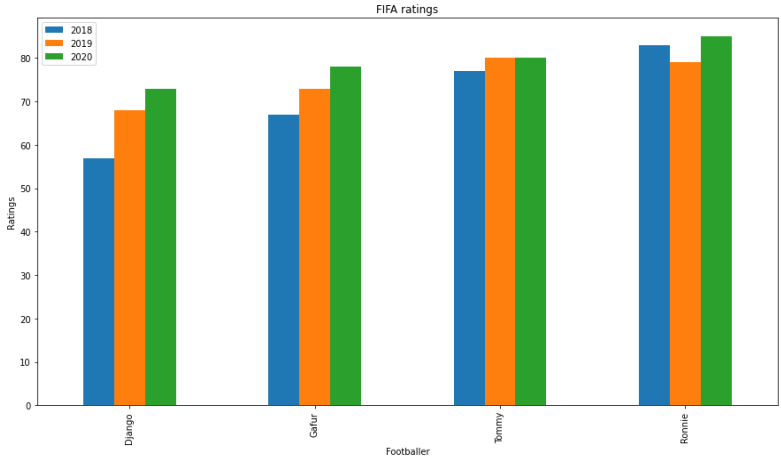

```
#Creating the dataset
df = sns.load_dataset('titanic')
df_pivot = pd.pivot_table(df,
values="fare",index="who",columns="class", aggfunc=np.mean)
#Creating a grouped bar chart
ax = df_pivot.plot(kind="bar",alpha=0.5)
#Adding the aesthetics
plt.title('Chart title')
plt.xlabel('X axis title')
plt.ylabel('Y axis title')
# Show the plot
plt.show()
```

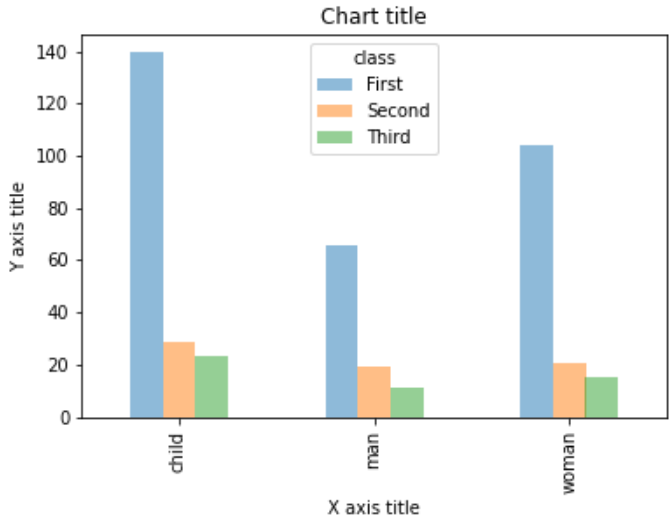

<div dir="rtl">

**نمودار ستونی با استفاده از Seaborn**

</div>

```python
#Reading the dataset
titanic_dataset = sns.load_dataset('titanic')
#Creating column chart
sns.barplot(x = 'who',y = 'fare',data = titanic_dataset,palette = "B
lues")
#Adding the aesthetics
plt.title('Chart title')
plt.xlabel('X axis title')
plt.ylabel('Y axis title')
# Show the plot
plt.show()
```

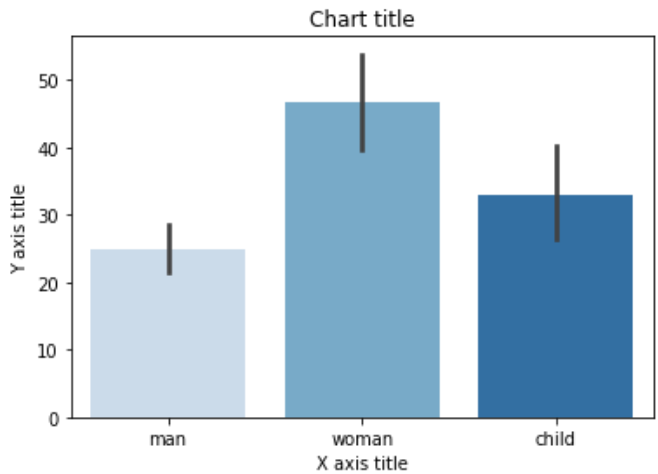

```
#Reading the dataset
titanic_dataset = sns.load_dataset('titanic')
#Creating the bar plot grouped across classes
sns.barplot(x = 'who',y = 'fare',hue = 'class',data = titanic_datase
t, palette = "Blues")
#Adding the aesthetics
plt.title('Chart title')
plt.xlabel('X axis title')
plt.ylabel('Y axis title')
# Show the plot
plt.show()
```

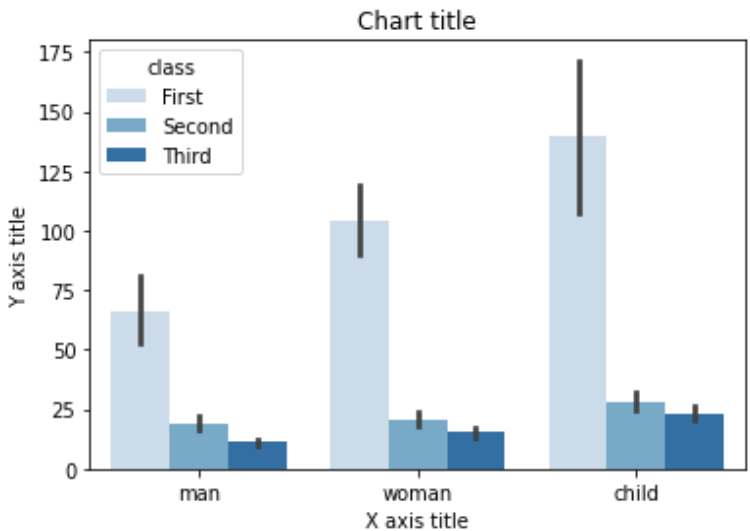

## نمودار دایره‌ای

نمودار دایره‌ای به‌طور گسترده در زمینه‌های مختلف برای نشان دادن نسبت طبقه‌بندی‌های مختلف و مقایسه طبقه‌بندی‌های مختلف توسط کمان استفاده می‌شود. نمودار دایره‌ای یک انتخاب عالی برای تجسم درصد است، چراکه هر عنصر را به عنوان بخشی از یک کل نشان می‌دهد.

**چه زمانی از نمودار دایره‌ای استفاده کنیم؟**

- وقتی نسبت‌ها و درصدهای نسبی یک مجموعه داده را نشان می‌دهید.
- بهترین استفاده با مجموعه داده‌های کوچک است.
- هنگام مقایسه تاثیر یک عامل بر دسته‌های مختلف.
- اگر تا ۶ دسته دارید.

● وقتی داده های شما اسمی هستند و ترتیبی نیستند.

**چه زمانی از نمودار دایره‌ای استفاده نکنیم؟**

● اگر مجموعه داده بزرگی دارید.

● اگر می‌خواهید یک مقایسه دقیق یا مطلق بین مقادیر انجام دهید.

**نمودار دایره‌ای با استفاده از Matplotlib**

```
#Creating the dataset
cars = ['AUDI', 'BMW', 'NISSAN',
    'TESLA', 'HYUNDAI', 'HONDA']
data = [20, 15, 15, 14, 16, 20]
#Creating the pie chart
plt.pie(data, labels = cars)
#Adding the aesthetics
plt.title('Chart title')
#Show the plot
plt.show()
```

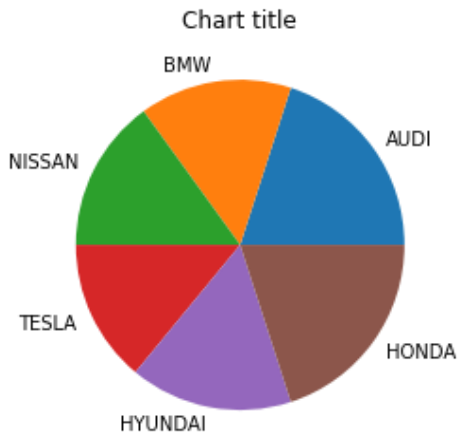

```
#Creating the dataset
cars = ['AUDI', 'BMW', 'NISSAN',
    'TESLA', 'HYUNDAI', 'HONDA']
data = [20, 15, 15, 14, 16, 20]
myexplode = [0.2, 0, 0, 0,0,0.6]
#Creating the pie chart
plt.pie(data, labels = cars,explode = myexplode)
#Adding the aesthetics
plt.title('Chart title')
#Show the plot
plt.show()
```

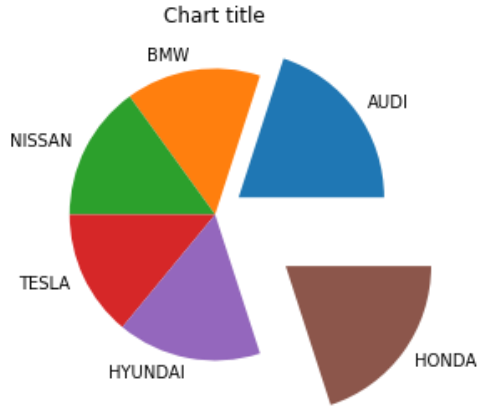

**نمودار دایره‌ای با استفاده از Seaborn**

```
import seaborn as sns

#define data
data = [15, 25, 25, 30, 5]
labels = ['Group 1', 'Group 2', 'Group 3', 'Group 4', 'Group 5']

#create pie chart
plt.pie(data, labels = labels)
plt.show()
```

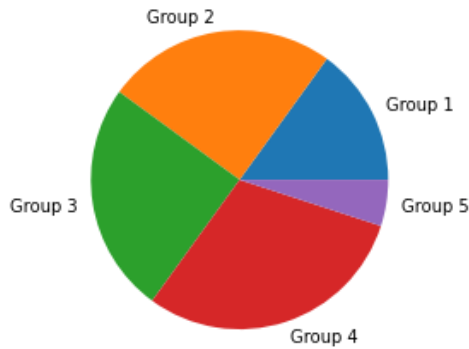

## نمودار نقطه‌ای

نمودار نقطه‌ای، نموداری است که رابطه بین دو متغیر را به‌صورت نقطه روی یک سیستم مختصات مستطیلی نشان می‌دهد. موقعیت نقطه با مقدار متغیر تعیین می‌شود. با مشاهده توزیع نقاط داده می‌توان همبستگی بین متغیرها را استنباط کرد. ایجاد نمودار نقطه‌ای به داده‌های زیادی نیاز دارد، در غیر این صورت همبستگی آشکار نمی‌شود.

**چه زمانی از نمودار نقطه‌ای استفاده کنیم؟**

- برای نشان دادن همبستگی و خوشه‌بندی در مجموعه داده‌های بزرگ.
- اگر مجموعه داده شما حاوی نقاطی است که دارای یک جفت مقدار هستند.
- زمانی که نیاز به مشاهده و نشان دادن روابط بین دو متغیر عددی دارید.
- اگر ترتیب نقاط در مجموعه داده ضروری نیست.

**چه زمانی از نمودار نقطه‌ای استفاده نکنیم؟**

- اگر مجموعه داده کوچکی دارید.
- اگر مقادیر موجود در مجموعه داده شما همبستگی ندارند.

**نمودار نقطه‌ای با استفاده از Matplotlib**

```python
import matplotlib.pyplot as plt
import numpy as np

#define data
x = np.array([5,7,8,7,2,17,2,9,4,11,12,9,6])
y = np.array([99,86,87,88,111,86,103,87,94,78,77,85,86])

#Show the plot
plt.scatter(x, y)
plt.show()
```

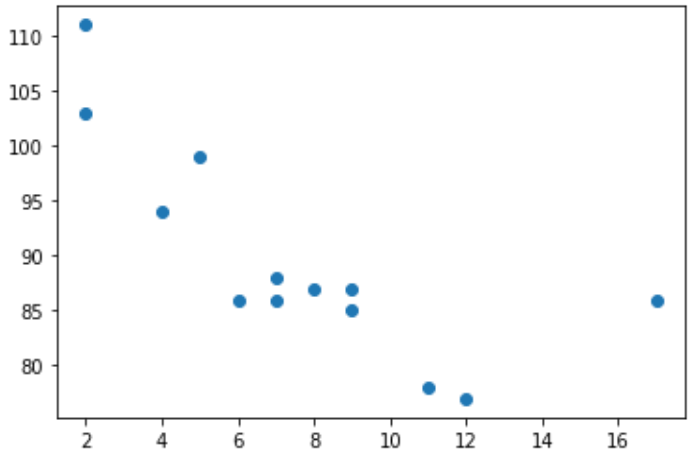

```
#Creating the dataset
df = sns.load_dataset("tips")
#Creating the scatter plot
plt.scatter(df['total_bill'],df['tip'],alpha=0.5 )
#Adding the aesthetics
plt.title('Chart title')
plt.xlabel('X axis title')
plt.ylabel('Y axis title')
#Show the plot
plt.show()
```

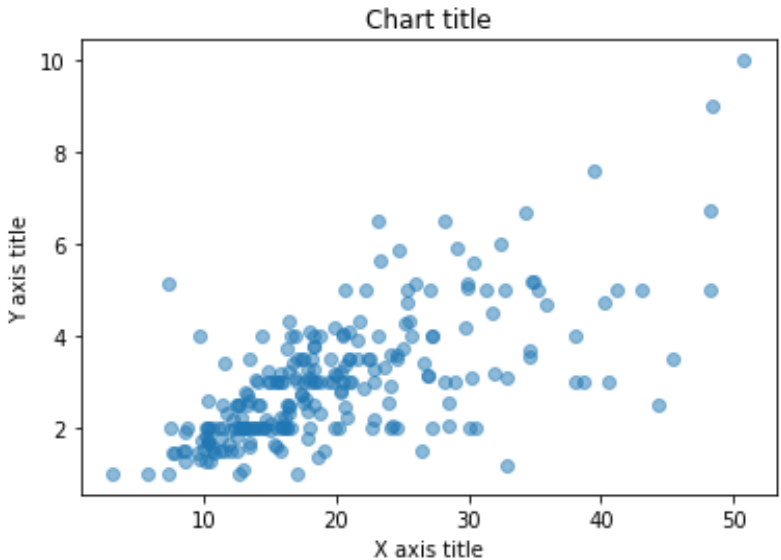

<div dir="rtl">

**نمودار نقطه‌ای با استفاده از Seaborn**

</div>

```
#Creating the dataset
bill_dataframe = sns.load_dataset("tips")
#Creating scatter plot
sns.scatterplot(data=bill_dataframe, x="total_bill", y="tip")
#Adding the aesthetics
plt.title('Chart title')
plt.xlabel('X axis title')
plt.ylabel('Y axis title')
# Show the plot
plt.show()
```

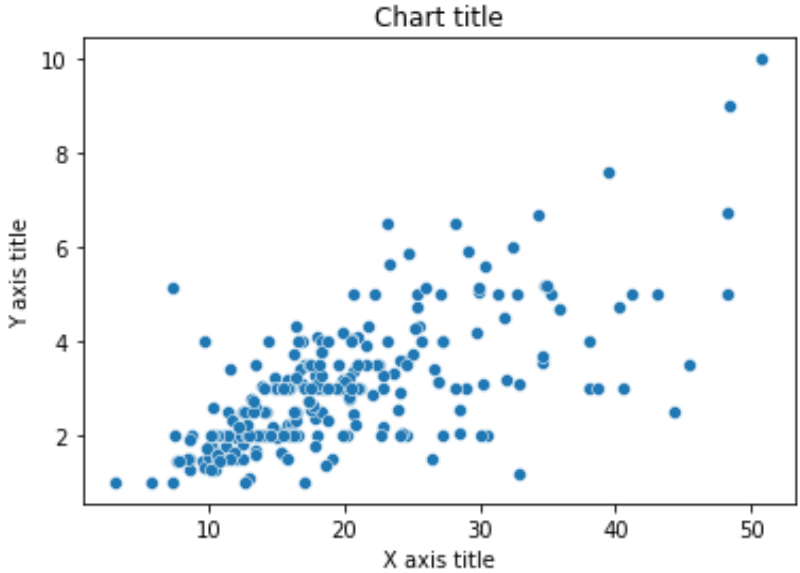

## نمودار مساحت[1]

نمودارهای مساحت برای ردیابی تغییرات در طول زمان برای یک یا چند گروه استفاده می‌شود. زمانی که بخواهیم تغییرات را در طول زمان برای بیش از یک گروه ثبت کنیم، نمودارهای مساحت نسبت به نمودارهای خطی ترجیح داده می‌شوند.

### چه زمانی از نمودار مساحت استفاده کنیم؟

- زمانی که می‌خواهید نه تنها کل مقادیر را دنبال کنید بلکه می‌خواهید از تفکیک آن به وسیله گروه‌ها نیز آگاه شوید.
- اگر می‌خواهید حجم داده‌های خود را به تصویر بکشید و نه تنها نسبت به زمان.

### چه زمانی از نمودار مساحت استفاده نکنیم؟

- نمی‌توان آن را با داده‌های گسسته استفاده کرد.

### نمودار مساحت با استفاده از Matplotlib

```
#Reading the dataset
x=range(1,6)
y=[ [1,4,6,8,9], [2,2,7,10,12], [2,8,5,10,6] ]
#Creating the area chart
ax = plt.gca()
ax.stackplot(x, y, labels=['A','B','C'],alpha=0.5)
```

---

[1] Area Chart

```
#Adding the aesthetics
plt.legend(loc='upper left')
plt.title('Chart title')
plt.xlabel('X axis title')
plt.ylabel('Y axis title')
#Show the plot
plt.show()
```

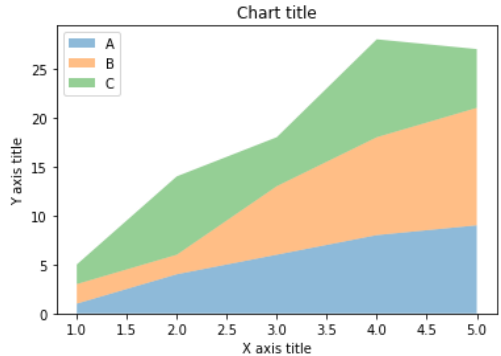

**نمودار مساحت با استفاده از Seaborn**

```
# Data
years_of_experience =[1,2,3]
salary=[ [6,8,10], [4,5,9], [3,5,7] ]
# Plot
plt.stackplot(years_of_experience,salary, labels=['Company A','C
ompany B','Company C'])
plt.legend(loc='upper left')
#Adding the aesthetics
plt.title('Chart title')
plt.xlabel('X axis title')
plt.ylabel('Y axis title')
# Show the plot
plt.show()
```

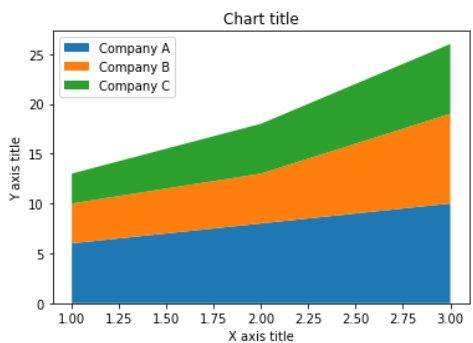

## نمودار حبابی[1]

نمودار حبابی یک نمودار چند متغیره و نوعی نمودار پراکندگی است که برای نمایش روابط بین سه متغیر استفاده می‌شود. مقادیر متغیرها برای هر نقطه با موقعیت افقی، موقعیت عمودی و اندازه نقطه نشان داده می‌شود.

### چه زمانی از نمودار حبابی استفاده کنیم؟

- اگر می‌خواهید مقادیر مستقل را با هم مقایسه کنید.
- اگر می‌خواهید توزیع یا رابطه را نشان دهید.
- وقتی می‌خواهید روابط بین سه متغیر را به تصویر بکشید و نشان دهید.

### چه زمانی از نمودار حبابی استفاده نکنیم؟

- اگر مجموعه داده کوچکی دارید.

### نمودار حبابی با استفاده از Matplotlib

```python
import matplotlib.pyplot as plt
import numpy as np

# create data
x = np.random.rand(40)
y = np.random.rand(40)
z = np.random.rand(40)
colors = np.random.rand(40)
# use the scatter function
plt.scatter(x, y, s=z*1000,c=colors)
plt.show()
```

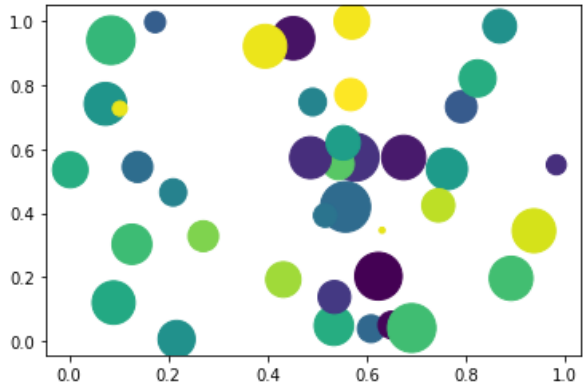

---

[1] Bubble Chart

**نمودار حبابی با استفاده از Seaborn**

```python
import matplotlib.pyplot as plt
import numpy as np
import seaborn as sns
import pandas as pd

# data
x=["IEEE", "Elsevier", "Others", "IEEE", "Elsevier", "Others"]
y=[7, 6, 2, 5, 4, 3]
z=["conference", "journal", "conference", "journal", "conference", "journal"]

# create pandas dataframe
data_list = pd.DataFrame(
    {'x_axis': x,
     'y_axis': y,
     'category': z
    })
# change size of data points
minsize = min(data_list['y_axis'])
maxsize = max(data_list['y_axis'])
# scatter plot
sns.catplot(x="x_axis", y="y_axis", kind="swarm", hue="category",sizes=(minsize*100, maxsize*100), data=data_list)
plt.grid()
```

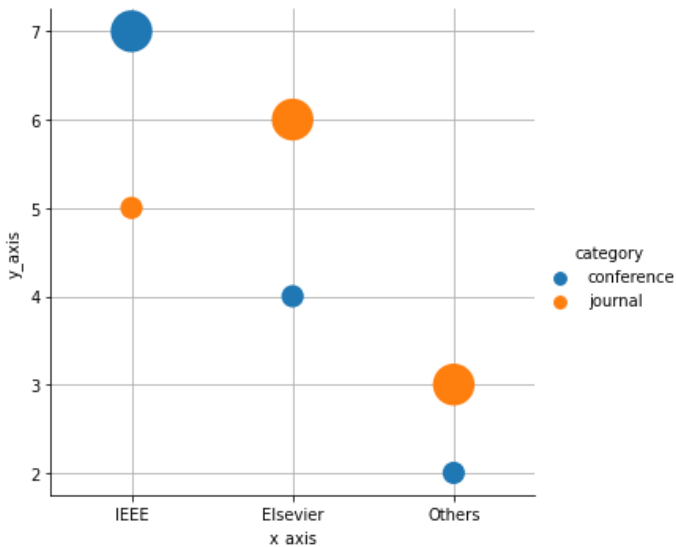

# خلاصه فصل

- داده‌ها محرکی هستند که می‌توانند یک کسب‌وکار را به مسیر درستی سوق دهند.
- مجموعه داده‌ها را اغلب می‌توان به عنوان مجموعه‌ای از اشیاء داده با ویژگی‌های یکسان در نظر گرفت.
- نام‌های دیگر برای یک شیء داده عبارتند از: رکورد، نقطه، بردار، الگو، رویداد، مورد، نمونه، مشاهده یا موجودیت.
- تمیزسازی داده‌ها فرآیند آماده‌سازی داده‌ها برای تجزیه و تحلیل با حذف یا تغییر داده‌های نادرست، ناقص، بی‌ربط، تکراری یا قالب نامناسب است.
- تمیزسازی داده‌ها به عنوان یک عنصر اساسی در مبانی علم داده در نظر گرفته می‌شود.
- یک الگوریتم ساده می‌تواند بر یک الگوریتم پیچیده غلبه کند، تنها به این دلیل که داده‌های کافی و باکیفیت بالا به آن داده شده است.
- کیفیت داده‌ها به اندازه کیفیت مدل پیش‌گویانه یا طبقه‌بندی اهمیت دارد.
- یک داده دورافتاده را نمی‌توان نویز یا خطا در نظر گرفت. با این حال، آن‌ها مشکوک هستند که با روش مشابه بقیه داده‌ها (شیء‌ها) تولید نشده‌اند.
- داده‌های دورافتاده تاثیر زیادی بر نتیجه تجزیه و تحلیل داده‌ها دارند.
- تنها در صورتی که مطمئن شده‌اید یک تکه داده (ویژگی) بی‌اهمیت است، می‌توانید آن را حذف کنید. شاید ویژگی‌ای که از منظر شما بی‌ربط به نظر می‌رسد، از منظر حوزه‌ای مانند دیدگاه بالینی بسیار مرتبط باشد.
- تجمیع داده‌ها روشی است که در آن داده‌های خام جمع‌آوری شده و به صورت خلاصه برای تجزیه و تحلیل استفاده می‌شود.
- نگاشت داده‌ها از مقادیر پیوسته به مقادیر گسسته، گسسته‌سازی نامیده می‌شود.
- هموارسازی داده‌ها با استفاده از الگوریتم‌های تخصصی برای حذف نویز از مجموعه داده انجام می‌شود.
- گاهی اوقات هموارسازی داده‌ها ممکن است نقاط داده قابل استفاده را حذف کند.
- متداول‌ترین تکنیک‌های مقیاس‌بندی ویژگی‌ها متعارف‌سازی و هنجارسازی هستند.
- گرادیان کاهشی شبکه عصبی با مقیاس‌بندی ویژگی بسیار سریع‌تر از بدون آن همگرا می‌شود.
- متعارف‌سازی زمانی ظاهر می‌شود که ویژگی‌های مجموعه داده‌ای ورودی تفاوت‌های زیادی بین محدوده خود داشته باشند.
- هنجارسازی زمانی مفید است که بدانید توزیع داده‌های شما از توزیع گوسی (منحنی زنگوله‌ای) پیروی نمی‌کند.

# مراجع برای مطالعه بیشتر

# بخش دوم

# یادگیری ماشین

**شامل فصل‌های:**

# ۴

## مقدمه‌ای بر یادگیری ماشین

**اهداف:**

- یادگیری ماشین چیست؟
- آشنایی با انواع رویکردهای یادگیری
- ارتباط یادگیری ماشین با سایر رشته‌ها
- کاربرد یادگیری ماشین
- آشنایی با ابزارهای یادگیری ماشین
- آشنایی با تفسیرپذیری و توضیح‌پذیری

# یادگیری ماشین چیست؟

در پی بارش نم‌نم باران، در خیابان خیس قدم می‌زنیم. با احساس نسیم ملایم و دیدن درخشش غروب آفتاب، شرط می‌بندیم که فردا باید هوا خوب باشد. ما همچنان انتظار داریم که نمرات تحصیلی خوبی در این ترم بعد از تلاش سخت و مطالعه‌های زیاد داشته باشیم. با نگاهی دقیق‌تر به این مثال‌ها، متوجه می‌شویم که بسیاری از پیش‌بینی‌های ما مبتنی‌بر تجربه است. به عنوان مثال، چرا فردا پس از مشاهده نسیم ملایم و درخشش غروب آفتاب، انتظار هوای خوبی را داریم؟ ما به این دلیل انتظار این هوای خوب را داریم، چراکه این تجربه برای ما حاصل شده است که آب و هوای روز بعد اغلب هنگامی خوب است که چنین صحنه‌ای را در حال حاضر تجربه می‌کنیم. به‌طور مشابه، تجربه یادگیری ما، به ما می‌گوید که کار سخت منجر به نمرات خوب تحصیلی می‌شود. ما به پیش‌بینی‌های خود اطمینان داریم، چراکه از تجربه آموخته‌ایم و تصمیمات مبتنی‌بر تجربه گرفته‌ایم. در حالیکه انسان‌ها از تجربه می‌آموزند، آیا رایانه‌ها نیز می‌توانند چنین کاری را انجام دهند؟ پاسخ "بله" است و یادگیری ماشین چیزی است که چنین کاری را انجام می‌دهد. یادگیری ماشین روشی است که با یادگیری تجربی از طریق روش‌های محاسباتی عملکرد سیستم را بهبود می‌بخشد (Zhou, 2021).

آرتور ساموئل، از پیشگامان در زمینه بازی‌های رایانه‌ای و هوش مصنوعی، اصطلاح "یادگیری ماشین" را در سال ۱۹۵۹ در IBM به کار برد. وی یادگیری ماشین را به‌عنوان "زمینه‌ای که به رایانه‌ها امکان یادگیری بدون برنامه‌ریزی صریح می‌دهد" تعریف کرد. یادگیری ماشین شاخه‌ای از هوش مصنوعی است که با هدف ایجاد رایانه‌هایی که توانایی یادگیری دارند، بوجود آمد. ایده اصلی اختراع یادگیری ماشین استدلال مبتنی‌بر نمونه است که فرآیند استدلال در مساله مورد نظر با مراجعه به نمونه‌های مشابه قبلی ممکن می‌شود. مثال‌های قبلی که برای ایجاد ظرفیت استفاده می‌شود، نمونه‌های (داده‌ها) آموزشی نامیده می‌شوند و فرآیند انجام این کار یادگیری نامیده می‌شود. اما قبل از آن که بیشتر با یادگیری ماشین آشنا شویم، بهتر است دریابیم که اساسا یادگیری چیست؟ یادگیری را می‌توان "بهتر شدن عملکرد در یک وظیفه خاص با استفاده تجربه و تمرین" تعریف کرد. رفتار هوشمندانه انسان از طریق یادگیری در تجربه‌ها به‌وقوع می‌پیوندد، یادگیری همان پدیدآورنده‌ی انعطاف‌پذیری در زندگی فردی است. یادگیری در انسان چیزی فراتر و حتی پیشرفته‌تر از پیشرفته‌ترین الگوریتم‌های یادگیری ماشین است. با این حال، یادگیری در رایانه‌ها چگونه اتفاق می‌افتد؟ در سیستم‌های رایانه‌ای، تجربه در قالب داده‌ها وجود دارند و وظیفه اصلی یادگیری ماشین توسعه الگوریتم‌های یادگیری است که از داده‌ها مدل می‌سازد. با تغذیه داده‌های تجربی به الگوریتم یادگیری ماشین، ما مدلی را بدست می‌آوریم که می‌تواند پیش‌بینی‌هایی را در مشاهدات جدید انجام دهد.

## تعریف یادگیری

"یک برنامه رایانه‌ای با در نظر گرفتن تجربه $E$ در مورد وظیفه $T$ برحسب معیار کارایی $P$ توانایی یادگیری خواهد داشت، اگر کارایی‌اش پس از تجربه $E$ برای وظیفه $T$ بهبود یابد."

مثال‌ها:

- ### یادگیری تشخیص دست‌خط
  - وظیفه $T$: تشخیص و طبقه‌بندی کلمات دست‌نویس در تصاویر
  - کارایی $P$: درصد کلماتی که به‌درستی طبقه‌بندی شده‌اند
  - تجربه $E$: مجموعه‌ای از کلمات دست‌نویس با طبقه‌بندی‌های داده شده

- ### یادگیری ربات راننده
  - وظیفه $T$: رانندگی در بزرگراه‌ها با استفاده از حسگرهای بینایی
  - کارایی $P$: میانگین مسافت طی شده قبل از خطا
  - تجربه $E$: یک توالی از تصاویر و فرمان هنگام هدایت توسط یک راننده انسانی

- ### یادگیری شطرنج
  - وظیفه $T$: بازی شطرنج
  - کارایی $P$: درصد بازی‌های برنده مقابل حریفان
  - تجربه $E$: انجام بازی‌های تمرینی برابر خودش

فرآیند یادگیری ماشین با استفاده از داده‌های خام در جهت استخراج اطلاعات مفید برای کمک به تصمیم‌گیری بهتر شروع می‌شود. از همین‌رو، یادگیری ماشین را می‌توان این‌گونه تعریف کرد: "یادگیری ماشین بر طراحی مدل‌هایی متمرکز است که در یک حوزه خاص، الگوریتم کامپیوتری برمبنای داده‌های آموزشی داده شده به مدل یادگیری را به‌صورت خودکار از طریق تجربه و آزمایش از داده‌ها بدست می‌آورد و عملکردش را بهبود می‌دهد تا در مواجه با داده‌های جدید در همان حوزه بتواند رفتاری مشابه انسان از خود نشان دهد". هرچند که ما یک تعریف کلی از یادگیری ماشین را ارائه کرده‌ایم، نویسندگان مختلف، تعریف‌های متفاوتی را برای یادگیری ماشین ارائه می‌کنند. در ادامه سه تعریف دیگر از یادگیری ماشین را فهرست کرده‌ایم:

- یادگیری ماشین عبارت است از، برنامه‌نویسی رایانه‌ها برای بهینه‌سازی معیار عملکرد با استفاده از داده‌های نمونه یا تجربه گذشته. ما یک مدل تعریف‌شده برای برخی پارامترها داریم و یادگیری، اجرای یک برنامه رایانه‌ای برای بهینه‌سازی پارامترهای مدل با استفاده از داده‌های آموزشی یا تجربه گذشته است. این مدل ممکن است پیش‌گویانه باشد تا

پیش‌بینی را در آینده انجام دهد، یا برای کسب دانش از داده‌ها باشد، یا هر دو. *(Alpaydin, 2004)*

▪ *زمینه تحقیقاتی که به‌عنوان یادگیری ماشین شناخته می‌شود، مربوط به چگونگی ساخت برنامه‌های رایانه‌ای است که به‌طور خودکار با تجربه بهبود می‌یابند.* *(Mitchell, 1997)*

▪ *یادگیری ماشین به این معناست که رایانه‌ها اعمال خود را تطبیق دهند یا تغییر داده یا تطبیق دهند (خواه این اقدامات پیش‌گویانه باشد، یا کنترل یک ربات را انجام دهند) تا این اقدامات دقیق‌تر شوند، که در آن دقت با میزان عملکردهای انتخاب شده صحیح، منعکس می‌شود.* *(Marsland, 2015)*

طبق این تعاریف از یادگیری ماشین دو پرسش اساسی بوجود می‌آید: نخست اینکه، یک رایانه چگونه می‌داند که در وظیفه خاص مورد انجامش، در حال بهبود عملکرداش است یا خیر. دوم اینکه، این برنامه چگونه درمی‌یابد که باید در انجام این وظیفه بهبود یابد. پاسخ به این پرسش‌ها دسته‌بندی از چند رویکرد متفاوت در یادگیری ماشین را بوجود می‌آورد که در این فصل به تشریح آن‌ها با عناوین یادگیری بانظارت، غیرنظارتی و تقویتی خواهیم پرداخت.

اغلب، روش‌های یادگیری ماشین به دو فاز تقسیم می‌شوند *(Hertzmann and Fleet, 2012)*:

۱. **آموزش**: یک مدل با استفاده از مجموعه داده‌های آموزشی، یادگیری بدست می‌آورد.

۲. **کاربرد**: مدل آموزش دیده شده، برای تصمیم‌گیری و پیش‌بینی برروی برخی از داده‌های آزمایشی (داده‌هایی که مدل در فرآیند آموزش آن‌ها را مشاهده نکرده است) استفاده می‌شود.

# یادگیری نظارتی

در رویکرد نظارتی، مجموعه‌ای از نمونه‌های آموزشی با پاسخ‌های صحیح (اهداف) به الگوریتم تغذیه می‌شود (هم داده‌های خام ورودی و هم نتایج آن‌ها را داریم) و الگوریتم سعی می‌کند براساس این داده‌ها و پاسخ‌های صحیح تابعی را بیاموزد (در طول زمان یاد می‌گیرد و باگذشت زمان دقیق‌تر می‌شود) تا بتواند مقادیر هدف را برای نمونه‌های جدید به درستی پیش‌بینی کند. به عبارت دیگر، هدف آن تطبیق سیستم به‌گونه‌ای است که برای ورودی‌های جدید، سیستم بتواند خروجی درستی را براساس آنچه تاکنون از داده‌های آموزشی فراگرفته است، پیش‌بینی کند. این نوع رویکرد از یادگیری، یادگیری از نمونه‌ها نیز نامیده می‌شود. در یادگیری نظارتی اگر داده‌های مسئله جهت یادگیری به‌صورت گسسته باشند، این مساله دسته‌بندی، و اگر مقادیر داده‌ها به‌صورت پیوسته باشند به آن رگرسیون گویند.

تعریف یادگیری نظارتی

یادگیری نظارتی جایی است که شما متغیرهای ورودی $(X)$ و یک متغیر خروجی $(Y)$ را دارید و از یک الگوریتم برای یادگیری تابع نگاشت از ورودی به خروجی استفاده می‌کنید:

$$Y = f(X)$$

هدف این است که تابع نگاشت به خوبی تخمین زده شود تا وقتی داده‌های ورودی جدید $(X)$ در اختیار الگوریتم قرار گرفت، بتواند متغیرهای خروجی $(Y)$ را برای آن داده‌ها پیش‌بینی کند.

از آنجایی که فرآیند یادگیری الگوریتم از مجموعه داده‌های آموزشی را می‌توان به‌عنوان یک معلم ناظر بر فرآیند یادگیری در نظر گرفت، به آن یادگیری نظارتی گویند. ما پاسخ‌های صحیح را می‌دانیم، الگوریتم به‌طور مکرر پیش‌بینی‌هایی را در مورد داده‌های آموزشی انجام می‌دهد و توسط معلم تصحیح می‌شود. یادگیری زمانی متوقف می‌شود که الگوریتم به سطح قابل قبولی از عملکرد برسد.

## مزایای یادگیری نظارتی

- یادگیری نظارتی برای استخراج نتیجه از تجربیات گذشته یا دانش قبلی مفید است.
- می‌تواند انواع مختلفی از مسائل محاسباتی عملی را حل کند.
- نتیجه در مقابل روش یادگیری غیرنظارتی از دقت بیشتری برخوردار است.
- قبل از ارائه داده‌ها برای آموزش دقیقا می‌دانید چند کلاس وجود دارد.

## معایب یادگیری نظارتی

- آموزش نیاز به زمان زیادی برای محاسبه دارد.
- در صورتی که مجموعه داده آزمایشی با مجموعه داده آموزشی متفاوت باشد، الگوریتم یادگیری نظارتی هنگام پیش‌بینی نتایج با مشکلاتی روبرو می‌شود.
- برچسب‌گذاری داده‌ها زمان و هزینه زیادی را در پی دارد و گاهی جمع‌آوری یک مجموعه داده برچسب‌دار کافی امکان‌پذیر نیست.
- یادگیری نظارتی محدود است به‌طوری که نمی‌توان برخی از کارهای پیچیده در یادگیری ماشین را با استفاده از آن انجام داد.
- یادگیری نظارتی نمی‌تواند اطلاعات ناشناخته‌ای از داده‌های آموزشی همانند یادگیری غیرنظارتی بدهد.
- در صورتی که مساله طبقه‌بندی باشد، اگر ورودی را که جزء هیچ یک از کلاس‌ها در مجموعه داده‌های آموزشی نیست را ارائه دهیم، ممکن است خروجی برچسب کلاس اشتباه باشد. برای مثال، فرض کنید شما یک طبقه‌بندی کننده تصویر با داده‌های گربه و

سگ آموزش داده‌اید. سپس، اگر تصویر زرافه را بدهید، خروجی ممکن است گربه یا سگ باشد، که درست نیست.

# یادگیری غیرنظارتی

در رویکرد غیرنظارتی، پاسخ‌های درست به الگوریتم ارائه نمی‌شود (داده‌ها برچسب‌دار نیستند)، اما در عوض الگوریتم سعی می‌کند شباهت‌های بین ورودی‌ها را مشخص کند تا ورودی‌هایی که دارای ویژگی مشترک هستند در کنار هم گروه‌بندی شوند. به عبارت دیگر، سیستم خروجی مناسب را ندارد، اما داده‌ها را کاوش کرده و می‌تواند از مجموعه داده‌ها استنباط‌هایی را برای توصیف ساختارهای پنهان از داده‌های بدون برچسب بدست آورد. خوشه‌بندی، قوانین انجمنی و کاهش ابعاد نمونه‌هایی از یادگیری غیرنظارتی هستند.

## تعریف یادگیری غیرنظارتی

یادگیری غیرنظارتی جایی است که شما فقط داده‌های ورودی ($X$) را دارید و هیچ متغیر خروجی مربوطه ندارید. هدف از یادگیری غیرنظارتی، کسب اطلاعات بیشتر از داده‌ها است.

به این مورد یادگیری غیرنظارتی گفته می‌شود، چراکه برخلاف یادگیری نظارتی پاسخ‌های صحیح و معلم وجود ندارد. کشف و ارائه ساختار جالب در داده‌ها به‌عهده خود الگوریتم‌ها است.

با الگوریتم‌های غیرنظارتی، نمی‌دانید چه چیزی می‌خواهید از مدل بدست آورید. شما احتمالاً مشکوک هستید که باید نوعی رابطه یا همبستگی بین داده‌های شما وجود داشته باشد، اما داده‌ها بسیار پیچیده هستند تا بتوان حدس زد. بنابراین در این موارد شما داده‌های خود را به داده‌های نرمال تبدیل می‌کنید تا قابل مقایسه شوند و سپس اجازه می‌دهید مدل کار کند و تلاش کند تا برخی از این روابط را پیدا کند. یکی از ویژگی‌های ویژه این مدل‌ها این است که در حالی‌که مدل می‌تواند روش‌های مختلفی را برای گروه‌بندی یا سفارش داده‌های شما پیشنهاد دهد، این به شما بستگی دارد که تحقیقات بیشتری برروی این مدل‌ها انجام دهید تا چیز مفیدی رونمایی کنید.

## مزایای یادگیری غیرنظارتی

- برچسب‌گذاری داده‌ها مستلزم کار و هزینه زیادی است. یادگیری غیرنظارتی با یادگیری از داده‌های بدون‌برچسب این مشکل را حل می‌کند.
- در یافتن الگوهایی از داده‌هایی که یافتن آن‌ها با استفاده از روش‌های معمول امکان‌پذیر نیست، بسیار مفید است.

- کاهش ابعاد داده‌ها با استفاده از این نوع یادگیری به‌راحتی انجام می‌شود.

**معایب یادگیری غیرنظارتی**

- نتیجه ممکن است در مقابل روش یادگیری نظارتی از دقت کمتری برخوردار باشد. چرا که ما هیچ‌گونه برچسبی برای داده‌ها در اختیار نداریم و مدل باید با دانش بدست آمده از داده‌های خام یاد بگیرد.
- هرچه تعداد ویژگی‌ها بیشتر شود، پیچیدگی بیشتر می‌شود.
- فرآیندی زمان‌بر است. چراکه مرحله یادگیری الگوریتم ممکن است زمان زیادی را صرف تجزیه و تحلیل و محاسبه همه احتمالات کند.

# یادگیری تقویتی

در یادگیری تقویتی، یک عامل سعی می‌کند یک مساله را با آزمایش و خطا از طریق تعامل با محیطی که پویایی آن برای عامل ناشناخته است، حل کند. عامل می‌تواند ضمن دریافت بازخورد فوری از محیط، وضعیت محیط را با اقدامات خود تغییر دهد. هدف عامل این است که با یافتن یک زنجیره بهینه از اقدامات، مساله را حل کند. اگرچه یادگیری تقویتی یکی از حوزه‌های یادگیری ماشین است، با این حال از جهات مختلفی با روش‌های یادگیری ماشین استاندارد (نظارتی و غیرنظارتی) تفاوت اساسی دارد. اول اینکه، یادگیری تقویتی وابسته به فراگیری داده‌ها نیست. در عوض، در یادگیری تقویتی عامل از تجربه خود که در طول تعامل با محیط ایجاد شده است یاد می‌گیرد و به ناظر وابسته نیست. دوم اینکه، یادگیری تقویتی به جای تجزیه و تحلیل داده‌ها، بر یافتن یک سیاست بهینه متمرکز است.

**تفاوت یادگیری نظارتی، غیرنظارتی و تقویتی**

یادگیری نظارتی زمانی اتفاق می‌افتد که مجموعه‌ای از نمونه‌های آموزشی با پاسخ‌های صحیح (اهداف) به الگوریتم تغذیه می‌شود و الگوریتم به عنوان راهنما از این پاسخ‌ها در حل مساله کمک می‌گیرد. در مقابل آن، یادگیری غیرنظارتی نیازی به داده‌های برچسب‌دار ندارد و این خود مدل است که به تنهایی و بدون یک ناظر خارجی با یافتن شباهت‌های بین ورودی‌ها و کشف الگوهای پنهان به حل مساله می‌پردازد. در مقابل این دو رویکرد، یادگیری تقویتی نیازی به مجموعه داده نداشته و ماشین یا عامل با محیط خود در تعامل است تا با آزمون و خطا و دریافت پاداش از محیط بهترین اقدام را در جهت حل مساله مورد نظر بدست آورد. یادگیری تقویتی دقیقا نظارتی نیست، چراکه به‌طور کامل به مجموعه داده‌های آموزشی (برچسب‌دار) تکیه نمی‌کند. در واقع یادگیری تقویتی متکی بر توانایی نظارت بر واکنش اقدامات انجام‌شده و

اندازگیری آن با پاداش است. همچنین، غیرنظارتی نیز نیست، چراکه ما از قبل می‌دانیم که چه زمانی "یادگیرنده" را مدل می‌کنیم، که پاداش مورد انتظار است. به‌طور خلاصه، در یادگیری نظارتی، هدف تولید فرمولی براساس مقادیر ورودی و خروجی است. در یادگیری بدون‌نظارت، ارتباطی بین مقادیر ورودی و گروه‌بندی آن‌ها پیدا می‌شود. در یادگیری تقویتی، یک عامل از طریق تعامل با محیط یاد می‌گیرد. بر این اساس می‌توان تفاوت این سه رویکرد از یادگیری ماشین را در جدول ۱ ـ ۱ مشاهده کرد.

جدول ۱ ـ ۱ مقایسه یادگیری نظارتی، غیرنظارتی و تقویتی

| شاخص | یادگیری بانظارت | یادگیری بدون‌نظارت | یادگیری تقویتی |
|---|---|---|---|
| تعریف | از طریق مجموعه داده دارای برچسب یاد می‌گیرد. | بدون راهنما از طریق داده‌های بدون‌برچسب آموزش داده می‌شود. | در تعامل با محیط کار می‌کند. |
| نوع داده‌ها | داده‌های برچسب‌دار | داده‌های بدون برچسب | بدون تعریف داده |
| نوع مساله | دسته‌بندی و رگرسیون | قوانین انجمنی و خوشه‌بندی | مبتنی بر پاداش |
| ناظر | ناظر اضافی | بدون ناظر | بدون ناظر |
| هدف | نگاشت داده‌های ورودی به خروجی‌های مشخص | کشف الگو | آموختن یک سری اقدامات |

## یادگیری انتقالی

یادگیری انتقالی بر استخراج داده‌ها از یک دامنه مشابه متمرکز است تا توانایی یادگیری را افزایش یا تعداد نمونه‌های برچسب‌دار مورد نیاز در یک دامنه هدف را کاهش دهد. در یادگیری انتقالی، یک مدل از دانش بدست آمده از کار قبلی برای بهبود تعمیم در مورد دیگری بهره می‌برد. هدف از یادگیری انتقالی بهبود فرآیند یادگیری وظایف جدید با استفاده از تجربه از حل مسائل قبلی است که تا حدودی مشابه هستند.

### تعریف یادگیری انتقالی

استفاده از مدلی پیش‌آموزش داده شده در راستای انتقال دانش از این مدل برای وظیفه‌ای مشابه، در جهت بهبود عملکرد این وظیفه جدید.

برای درک تعریف رسمی یادگیری انتقالی، لازم است در ابتدا دامنه و وظیفه تعریف شوند. دامنه مجموعه داده‌ای است که برای آموزش استفاده و دامنه به صورت $D = \{\chi, P(X)\}$ نشان داده می‌شود که شامل دو مولفه: $\chi$ فضای ویژگی و $P(X)$ یک توزیع احتمال که در این تعریف $X = \{x_i, ..., x_n\} \in \chi$ است. وظیفه را می‌توان با فضای برچسب $y$ و یک تابع مدل هدف $f(x)$ و به صورت $T = \{y, f(x)\}$ نشان داد. $f(x)$ را همچنین

می‌توان به عنوان یک تابع احتمال شرطی $P(y|x)$ نوشت. حال می‌توان یادگیری انتقالی را به‌طور رسمی به صورت زیر تعریف کرد:

با توجه به دامنه منبع $D_s$ و وظیفه یادگیری منبع $T_s$، دامنه هدف $D_t$ و وظیفه یادگیری هدف $T_t$، جایی که حجم $D_s$ها بیشتر از حجم $D_t$ها باشد، یادگیری انتقالی روشی است برای بهبود عملکرد مدل هدف $f_T(.)$ برای وظیفه یادگیری هدف $T_t$ با کسب دانش ضمنی از $D_s$ و $T_s$، جایی که $D_s \neq D_t$ و $T_s \neq T_t$ است.

## یادگیری چندوظیفه‌ای

یادگیری چندوظیفه‌ای یک الگوی آموزشی است که در راستای به حداکثر رساندن کارآیی مدل، چندین وظیفه مرتبط به‌طور همزمان یاد گرفته می‌شوند و همزمان چندین تابع ضرر بهینه می‌شوند. در این فرآیند، مدل از همه داده‌های موجود در وظایف مختلف برای یادگیری بازنمایی کلی داده‌ها که در زمینه‌های مختلف مفید هستند، استفاده کرده و آن‌ها را بین وظایف مختلف به اشتراک می‌گذارد. این اشتراک‌گذاری‌ها کارآیی مدل را افزایش داده (هر وظیفه می‌تواند از وظیفه دیگر بهره‌مند شود) و به‌طور بالقوه می‌تواند سرعت یادگیری سریع‌تری را به‌همراه داشته باشد. انگیزه اصلی یادگیری چندوظیفه‌ای ایجاد یک مدل "عمومی" است که می‌تواند چندین کار را به‌جای ایجاد چندین مدل "تخصصی" که فقط برای یک کار خاص آموزش دیده‌اند، در یک مدل به‌طور همزمان حل کند. از منظر زیست‌شناختی، یادگیری چندوظیفه‌ای الهام گرفته از روشی است که ما انسان‌ها یاد می‌گیریم. برای یادگیری وظایف جدید، معمولا ما دانشی راکه از یادگیری کارهای مرتبط بدست آورده‌ایم، بکار می‌گیریم.

### تفاوت یادگیری انتقالی با یادگیری چندوظیفه‌ای

یادگیری چندوظیفه‌ای متفاوت از یادگیری انتقالی است، و تفاوت آن‌ها در نحوه انتقال دانش است. وظایف به‌طور متوالی در یادگیری انتقالی یاد گرفته می‌شوند و از یکی به یکی دیگر منتقل می‌شود. در حالیکه، یادگیری چندوظیفه‌ای با به‌اشتراک گذاشتن اطلاعات بین همه وظایف به‌دنبال عملکرد خوب در تمام وظایف در نظر گرفته شده توسط یک مدل واحد به‌صورت همزمان به‌طور موازی است.

## یادگیری یک‌نمونه‌ای[1]

به‌طور معمول، طبقه‌بندی فرآیندی شامل تغذیه تعداد زیادی نمونه از هر کلاس به مدل است. یادگیری یک نمونه‌ای برخلاف مدل‌های سنتی یادگیری ماشین که از هزاران نمونه آموزشی برای

---

[1] One-shot Learning

یادگیری استفاده می‌کنند، نوعی یادگیری است که تنها از یا چند نمونه آموزشی برای یادگیری استفاده می‌کند. نمونه‌ای از کاربردهای این نوع یادگیری تشخیص چهره است. جایی که افراد باید با توجه به حالات مختلف چهره، شرایط نور، لوازم جانبی و مدل مو با توجه به یک یا چند عکس الگو، به‌درستی طبقه‌بندی شوند.

## یادگیری بدون‌نمونه[1]

امروزه بسیاری از روش‌های یادگیری ماشین بر طبقه‌بندی مواردی متمرکز هستند که کلاس‌های آن‌ها قبلا در آموزش دیده شده است. با این حال، در بسیاری از مسائل نیاز به طبقه‌بندی مواردی است که کلاس‌های آن‌ها قبلا دیده نشده است. یادگیری بدون‌نمونه یک روش یادگیری نظارتی ولی بدون داده‌های آموزشی از آن کلاس است. یادگیری بدون‌نمونه قادر است یک مساله را با وجود عدم دریافت هیچ نمونه آموزشی از آن مساله حل کند. برای مثال، تصور کنید که یک دسته از اجسام را در عکس‌ها تشخیص می‌دهید بدون اینکه قبلا عکسی از آن نوع شی را دیده باشید.

## یادگیری استقرایی[2] (یادگیری مفهومی[3])

یادگیری استقرایی که همچنین به‌عنوان یادگیری مفهومی شناخته می‌شود ، شامل ایجاد یک قانون تعمیم‌یافته برای تمام داده‌های تغذیه‌شده به الگوریتم است. در این رویکرد، داده‌ها را به عنوان ورودی و نتایج را به عنوان خروجی داریم و باید رابطه بین ورودی و خروجی را پیدا کنیم. این می‌تواند بسته به داده‌ها بسیار پیچیده باشد. با این حال روش موثری است که در یادگیری ماشین استفاده در زمینه‌های مختلف مانند فناوری تشخیص چهره، تشخیص بیماری درمان و غیره استفاده می‌شود. *این نوع یادگیری یک رویکرد پایین به بالا است.*

در این نوع از یادگیری، مدل ساخته‌شده به گونه‌ای است که اگر بتواند تقریب خوبی از تابع هدف را به مجموعه بزرگ از مجموعه آموزشی بسط دهد، می‌تواند این تابع هدف را در مورد نمونه‌های دیده‌نشده تخمین بزند. در این یادگیری تنها اطلاعات موجود، مجموعه داده‌های آموزشی است، بنابراین در بهترین حالت یک الگوریتم یادگیری می‌تواند فرضیه‌ای را ارایه دهد که تابع هدف را در نمونه‌های آموزشی تخمین می‌زند.

این یادگیری بسیار مهم است، زیرا به ما رابطه‌ای را از داده‌ها می‌دهد که برای ارجاع‌های بعدی کاربرد دارند. از این رویکرد زمانی استفاده می‌شود که تخصص انسانی زمانی که خروجی‌ها

---

[1] Zero-shot learning

[2] Inductive Learning

[3] Concept Learning

تغییر می‌کنند، کاربردی ندارد. به طور خلاصه، می‌توان گفت که در یادگیری استقرایی، ما نتایج حاصل از حقایق را تعمیم می‌دهیم. برای مثال:

**آ.** سیب یک میوه است.

**ب.** طعم سیب شیرین است.

**نتیجه:** همه‌ی میوه‌ها طعم شیرین دارند.

*این حوزه از یادگیری ماشین هنوز تحت مطالعه و تحقیق است. زیرا پیشنهادات زیادی برای بهبود در مورد کارایی و سرعت الگوریتم آن وجود دارد.*

یادگیری استقرایی همان چیزی است که ما معمولا به عنوان یادگیری بانظارت سنتی می‌شناسیم. ما یک مدل یادگیری ماشینی را بر اساس مجموعه داده‌های آموزشی برچسب‌گذاری شده‌ای که قبلاً داریم، می‌سازیم و آموزش می‌دهیم. سپس از این مدل آموزش‌دیده برای پیش‌بینی برچسب‌های مجموعه داده آزمایشی استفاده می‌کنیم که قبلاً هرگز با آن‌ها مواجه نشده‌ایم.

# یادگیری قیاسی[1]

درست مانند استدلال استقرایی، یادگیری قیاسی یا استدلال قیاسی شکل دیگری از استدلال است. در واقع، استدلال یک مفهوم هوش مصنوعی است و هر دو یادگیری استقرایی و قیاسی بخشی از آن هستند. استدلال قیاسی فرآیند استنباط دانش جدید از اطلاعات موجود قبلی است که به طور منطقی به هم مرتبط هستند و شکلی از منطق معتبر است، به این معنی که در صورت صحت ادعاها، نتیجه‌گیری باید صادق باشد.

برخلاف یادگیری استقرایی که مبتنی‌بر تعمیم حقایق خاص است، یادگیری قیاسی از حقایق و اطلاعات موجود استفاده می‌کند تا نتیجه‌گیری معتبری بدست آورد و از رویکرد *بالا به پایین استفاده می‌کند (کاملا مخالف یادگیری استقرایی است)*. نکته مهمی که باید به آن توجه کرد این است که در یادگیری قیاسی، نتایج قطعی است؛ یعنی بله یا خیر. در حالی که یادگیری استقرایی مبتنی‌بر احتمال است، یعنی می‌تواند از قوی تا *ضعیف* متغیر باشد.

اعتبار[2] استنتاج، صدقِ بیانِ استدلالِ قیاسی را تضمین می‌کند. استدلال قیاسی معمولا با اصول اولیه شروع می‌شود و به یک نتیجه خاص می‌رسد. برای مثال:

**آ.** همه گوشتخواران گوشت می‌خورند.

---

[1] Deductive Learning

[2] validity

**ب.** شیر یک گوشتخوار است.

**نتیجه:** شیر گوشت می‌خورد.

روش کلاسیک یادگیری ماشین از پارادایم علمی استقرا و قیاس پیروی می‌کند. در مرحله استقرایی مدل را از داده‌های خام (مجموعه آموزشی) یاد می‌گیریم و در مرحله قیاس مدل برای پیش‌بینی رفتار داده‌های جدید اعمال می‌شود.

# یادگیری تمثیلی[1]

در یادگیری تمثیلی، هم داده‌های آموزشی و هم داده‌های آزمایشی از قبل تحلیل می‌شوند. دانش بدست آمده از این مجموعه داده‌ها، دانشی است که مفید است. این مدل سعی می‌کند پس از یادگیری از مجموعه داده‌های آموزشی، برچسب‌ها را برای مجموعه داده‌های آزمایشی پیش‌بینی کند.

## تفاوت یادگیری استقرایی با یادگیری تمثیلی

یادگیری استقرایی چیزی نیست جز اصل پشت الگوریتم‌های یادگیری ماشین بانظارت که در آن یک مدل سعی می‌کند با بررسی الگوهای پنهان در داده‌های آموزشی، رابطه‌ای بین متغیرهای ویژگی و متغیر هدف ایجاد کند. اگرچه مدل در معرض دامنه محدودی از داده‌های آموزشی قرار دارد، یادگیری مدل مطابق با ماهیت عمومی داده‌ها خواهد بود به طوری که می‌تواند ارزش هر نقطه داده را برای یک مجموعه داده بدون‌برچسب (مجموعه داده آزمایشی) پیش‌بینی کند. تفاوت اصلی بین یادگیری استقرایی و یادگیری تمثیلی این است که در طول یادگیری تمثیلی، شما قبلا با مجموعه داده‌های آموزشی و آزمایشی هنگام آموزش مدل مواجه شده‌اید. با این حال، یادگیری استقرایی تنها با داده‌های آموزشی در هنگام آموزش مدل مواجه می‌شود و مدل آموخته شده را بر روی مجموعه داده‌ای که قبلا هرگز ندیده است، اعمال می‌کند.

یادگیری تمثیلی یک مدل پیش‌گویانه نمی‌سازد. اگر یک نقطه داده جدید به مجموعه داده آزمایشی اضافه شود، باید الگوریتم را از ابتدا دوباره اجرا کنیم، مدل را آموزش دهیم و سپس از آن برای پیش‌بینی برچسب‌ها استفاده کنیم. از سوی دیگر، یادگیری استقرایی یک مدل پیش‌گویانه می‌سازد. هنگامی که با نقاط داده جدید روبرو می‌شوید، نیازی به اجرای مجدد الگوریتم از ابتدا نیست. به عبارت ساده‌تر، یادگیری استقرایی سعی می‌کند یک مدل عمومی ایجاد کند که در آن هر نقطه داده جدید بر اساس مجموعه مشاهده شده از نقاط داده آموزشی پیش‌بینی شوند. در *اینجا می‌توانید هر نقطه‌ای را در فضای نقاط، فراتر از نقاط بدون برچسب پیش‌بینی کنید.* در مقابل، یادگیری تمثیلی مدلی را ایجاد می‌کند که با نقاط داده آموزشی و آزمایشی که قبلا مشاهده

---

[1] Transductive Learning

کرده است، متناسب باشد. این رویکرد با استفاده از دانش نقاط برچسب‌گذاری شده و اطلاعات اضافی، برچسب‌های داده‌های بدون‌برچسب را پیش‌بینی می‌کند. در مواردی که نقاط داده جدید توسط یک جریان ورودی[1] معرفی شوند، یادگیری تمثیلی می‌تواند پرهزینه شود. هر بار که یک نقطه داده جدید می‌رسد، باید همه چیز را دوباره اجرا کنید. از سوی دیگر، یادگیری استقرایی در ابتدا یک مدل پیش‌گویانه ایجاد می‌کند و نقاط داده جدید را می‌توان در مدت زمان بسیار کوتاهی با محاسبات کمتر برچسب‌گذاری کرد.

تصور کنید که یک مجموعه داده آموزشی دارید، اما تنها یک زیرمجموعه از آن دارای برچسب می‌باشد. به عنوان مثال، شما سعی دارید دسته‌بندی انجام دهید که آیا در یک تصویر یک گل وجود دارد یا نه. شما ۱۰۰۰۰۰ تصویر دارید، اما تنها ۱۰۰۰ عکس دارید که قطعا حاوی یک گل می‌باشد و ۱۰۰۰ عکس دیگر که می‌دانید حاوی یک گل نیست و در مورد ۹۸۰۰۰ نمونه دیگر شما هیچ ایده‌ای ندارید. به عبارت دیگر، شاید آن‌ها گل داشته باشند، شاید هم نه. یادگیری استقرایی با بررسی ۲۰۰۰ نمونه برچسب‌گذاری شده و ساختن یک دسته‌بند بر روی این ۲۰۰۰ نمونه کار می‌کند. در مقابل، یادگیری تمثیلی می‌گوید: "صبر کنید، شاید سایر ۹۸۰۰۰ دیگر برچسب *نداشته باشند، اما آن‌ها چیزی در مورد فضای مساله به من می‌گویند که شاید بتوانم از آن‌ها برای کمک به بهبود دقت خود استفاده کنم.*"

به طور خلاصه :

- یادگیری استقرایی مدل را با نقاط داده برچسب‌دار آموزش می‌دهد و سعی می‌کند برچسب نقاط داده بدون‌برچسب را پیش‌بینی کند. در مقابل، در یادگیری تمثیلی با مجموعه داده آموزشی و آزمایشی، آموزش داده می‌شود و سعی می‌کند برچسب نقاط داده بدون‌برچسب را پیش‌بینی کند.

- در یادگیری استقرایی، اگر یک نقطه داده بدون‌برچسب جدید معرفی شود، می توانیم از مدل آموزش دیده قبلی برای پیش‌بینی استفاده کنیم. با این حال، در یادگیری تمثیلی، ممکن است نیاز به آموزش مجدد کل مدل داشته باشیم.

- در یادگیری تمثیلی، مجموعه آزمون باید از قبل در دسترس باشد، به گونه‌ای که مدل سازی تمثیلی مساله از اطلاعات موجود از داده‌های آزمون بدون برچسب، برای دقت بهتر استفاده کند.

- یادگیری تمثیلی از نظر محاسباتی گران‌تر از یادگیری استقرایی است.

---

[1] input stream

# یادگیری فعال[1]

یادگیری فعال شاخه‌ای از یادگیری ماشین است که در آن یک الگوریتم یادگیری می‌تواند با کاربر ارتباط برقرار کند تا داده‌ها را با خروجی‌های مورد نظر علامت‌گذاری کند. یادگیرنده نمونه‌های آموزشی خود را بر اساس برخی استراتژی‌ها برای بهبود عملکرد خود می‌سازد. الگوریتم در یادگیری فعال به‌طور مستمر (فعال) زیرمجموعه نمونه‌هایی را که در مرحله بعدی دسته‌بندی می‌شوند، از مجموعه داده‌های بدون برچسب انتخاب می‌کند. باور اصلی پشت ایده الگوریتم یادگیرنده فعال این است که به یک الگوریتم یادگیری ماشین اجازه انتخاب داده‌هایی که باید از آنها بیاموزد را می‌دهد تا از نظر تئوری به درجه بالاتری از دقت با استفاده از تعداد محدودی از برچسب‌های آموزشی دست یابد. در نتیجه، یادگیرندگان موفق اجازه دارند در مرحله آموزش به صورت تعاملی سوال بپرسند. این پرس‌وجوها معمولا به شکل نمونه داده‌های بدون‌برچسب با درخواستی از حاشیه‌نویسی[2] انسانی برای علامت‌گذاری است. در نتیجه، یادگیری فعال جزئی از مدل *انسان در حلقه*[3] می‌شود، جایی که یکی از مهم‌ترین نمونه‌های عملکرد است.

هدف از یادگیری فعال افزایش عملکرد الگوریتم یادگیری ماشین و در عین حال ثابت نگه داشتن تعداد نمونه‌های آموزشی است. از این رویکرد اغلب زمانی استفاده می‌شود که تولید نمونه‌های آموزشی پرهزینه است و یا به زمان زیادی نیاز دارد. دو راهبرد اساسی در یادگیری فعال عبارتند از: **نمونه‌گیری عدم قطعیت**[4] و **نمونه‌گیری فضای نمونه**[5].

## نمونه‌گیری عدم قطعیت

در نمونه‌گیری عدم قطعیت، یادگیرنده نمونه‌های آموزشی خود را به صورت مکرر انتخاب می‌کند، به گونه‌ای که نمونه‌های انتخاب شده، نمونه‌هایی هستند که *یادگیرنده کمترین اطمینان را در مورد آنها دارد*. به طور معمول، یادگیرنده دارای مدلی است که بر روی نمونه‌های آموزشی فعلی خود آموزش داده شده است که به احتمال تعلق موارد به یک کلاس خاص را پیش‌بینی می‌کند. سپس، یادگیرنده از یک تکنیک کمینه‌سازی برای یافتن نمونه‌ای استفاده می‌کند که این احتمال را به حداقل می‌رساند (و از این رو مدل کمترین اطمینان را در مورد نحوه دسته‌بندی دارد). این نقطه توسط شخص ثالثی دسته‌بندی شده و به مجموعه آموزشی اضافه می‌شود. این فرآیند تا زمانی تکرار می‌شود که یادگیرنده از عملکرد دسته‌بند راضی باشد. پیاده‌سازی و استدلال پشت

---

[1] Active learning

[2] annotator

[3] human-in-the-loop model

[4] uncertainty sampling

[5] version space sampling

نمونه‌گیری عدم قطعیت بسیار ساده است که سبب شد این استراتژی در یادگیری فعال محبوب شود. نقطه ضعف این استراتژی این است که هرگاه یادگیرنده، از کلاسِ بخشی از فضای ورودی مطمئن شود (در حالی که ممکن است آن قسمت را اشتباه دسته‌بندی کند)، یادگیرنده نمونه‌های آموزشی جدیدی را از آن قسمت انتخاب نمی‌کند. این می‌تواند باعث شود که مدل نهایی در آن بخش از فضای ورودی شکست بخورد.

## نمونه‌گیری فضای نمونه

فضای نمونه یک مجموعه آموزشی، فضایی است که شامل تمام مدل‌هایی است که به‌درستی با نمونه‌های آموزشی مطابقت دارند. به عبارت دیگر، فضای نمونه ساختاری است که به حفظ تمام فرضیه‌هایی که قادر به دسته‌بندی کامل مشاهدات فعلی ما هستند، کمک می‌کند. ایده این است که سرعت یادگیری را با انتخاب نمونه‌ها به‌گونه‌ای افزایش دهیم که با هر برچسب‌گذاری، فضای نمونه به سرعت به حداقل برسد.

از تعریف فضای نمونه، نتیجه‌گیری می‌شود که افزودن یک نمونه آموزشی به مجموعه آموزشی تنها می‌تواند اندازه‌یِ فضای نمونه را کاهش دهد. این اساسِ نمونه‌گیری فضای نمونه است. در نمونه‌گیری از فضای نمونه، نمونه بعدی به‌گونه‌ای انتخاب می‌شود که کاهش اندازه فضای نمونه حداکثر باشد. این کار با انتخاب نمونه‌هایی انجام می‌شود که مدل‌های موجود در فضای نمونه با آن مخالفت می‌کنند.

# یادگیری برخط[1]

در ابتدایی‌ترین شکل، یادگیری برخط یک تکنیک یادگیری ماشین است که نمونه‌هایی از داده‌های بلادرنگ یک مشاهده را در یک زمان فرا می‌گیرد. به عبارت دیگر، الگوریتم‌های یادگیری برخط با داده‌هایی که در دسترس هستند کار می‌کنند. الگوریتم‌های اکیداً[2] برخط با رسیدن هر نمونه داده جدید به تدریج بهبود می‌یابند، سپس آن داده‌ها را دور می‌اندازند و دیگر از آن استفاده نمی‌کنند. این یک الزام نیست، اما معمولاً مطلوب است که یک الگوریتم برخط نمونه‌های قدیمی‌تر را در طول زمان فراموش کند تا بتواند با جمعیت‌های غیر_ایستا[3] سازگار شود.

یادگیری برخط را می‌توان برای مسائلی اعمال کرد که در آن‌ها نمونه‌ها در طول زمان معرفی می‌شوند و توزیع احتمال نمونه‌ها در طول زمان پیش‌بینی می‌شود. گرادیان کاهشی تصادفی با پس‌انتشار (همان‌طور که در شبکه‌های عصبی استفاده می‌شود) نمونه‌ای از این نوع یادگیری است.

---

[1] Online learning

[2] Strictly

[3] non-stationary

# یادگیری برون‌خط

الگوریتم‌های یادگیری برون‌خط با داده‌ها به صورت انبوه، از یک مجموعه داده کار می‌کنند. به عبارت دیگر، باید با استفاده از داده‌هایی که از قبل در اختیار دارند یاد بگیرد. الگوریتم‌های یادگیری اکیدا برون‌خط باید دوباره از ابتدا اجرا شوند تا از داده‌های تغییر یافته یاد بگیرند. ماشین‌های بردار پشتیبان و جنگل‌های تصادفی الگوریتم‌های اکیدا برون‌خط هستند (اگرچه محققان انواع برخط آن‌ها را ساخته‌اند).

## تفاوت یادگیری برون‌خط با برخط

تفاوت بین این دو نوع یادگیری را می‌توان با یک مثال ساده توضیح داد (*Dulhare et al., 2020*). فرض کنید دانش‌آموزی می‌خواهد جبرخطی را یاد بگیرد. در نوع اول یادگیری، این دانش‌آموز می‌تواند کتاب‌های متعدد جبرخطی را بخواند و بیاموزد. او پس از یادگیری از این کتاب‌ها، دیگر چیز جدیدی یاد نمی‌گیرد و بعد از آن فقط از دانش خود استفاده می‌کند. این یک نوع یادگیری برون‌خط است. در این روش یادگیری، تمامی داده‌ها در طول آموزش در دسترس است و پس از مرحله آموزش، دیگر یادگیری را نخواهیم داشت. در مقابل، در یادگیری دوم یا یادگیری برخط به این صورت است که دانش‌آموز ابتدا کتاب‌های خود را می‌خواند و می‌خواند. می‌آموزد و سپس ضمن استفاده از دانش خود، هرگاه کتاب جدیدی در زمینه جبرخطی پیدا کرد، مطالعه می‌کند و با خواندن آن، میزان یادگیری خود را بهبود می‌بخشد.

بنابراین، یادگیری برون‌خط مانند این است که شما مجموعه‌ای از کتاب‌ها دارید و باید آن‌ها را یاد بگیرید. همه منابع شما این کتاب‌ها هستند و در واقع همه داده‌ها را دارید، اما فرض کنید در مسیر زندگی با اطلاعات جدیدی که هر روز به شما داده می‌شود باید چیزهای جدیدی یاد بگیرید و به دانش قبلی خود بیافزایید. این دومین مورد از یادگیری، برخط است، یعنی زمانی که همه داده‌ها در حال حاضر در دسترس نیستند.

یادگیری برخط دو مزیت عمده دارد:

۱. **این روش می‌تواند داده‌های با حجم بسیار بالا را آموزش دهد.** مثلا داده‌هایی که به دلیل حجم زیاد در حافظه نیستند.

۲. **تغییراتی که ممکن است در ماهیت داده‌ها رخ دهد از این طریق پوشش داده می‌شود.** فرض کنید گوگل یک الگوریتم برای سیستم ایمیل خود ایجاد کرده است که به‌طور هوشمند ایمیل‌های هرزنامه را با الگوریتم‌های یادگیری ماشین شناسایی می‌کند. همانطور که ممکن است انتظار داشته باشید، محتوای ایمیل‌های هرزنامه به‌طور دائم در حال تغییر است و افرادی که ایمیل‌های هرزنامه ارسال می‌کنند هر روز خود را در برابر این الگوریتم‌های گوگل بهینه می‌کنند. بنابراین، الگوریتم تشخیص

ایمیل هرزنامه گوگل می‌تواند آموزش برخط را برای شناسایی ایمیل‌های هرزنامه‌ای که در طول زمان تغییر کرده‌اند انجام دهد. در واقع یادگیری الگوریتم با اصلاح محتوا و شکل ایمیل‌های هرزنامه بروز و تقویت می‌شود.

> از بین این دو نوع الگوریتم، الگوریتم‌های برخط متداول‌تر هستند. زیرا می‌توانید به راحتی یک الگوریتم برون‌خط را از یک الگوریتم اکیدا برخط به اضافه یک مجموعه داده ذخیره‌شده بسازید، اما عکس آن برای یک الگوریتم کاملا برون‌خط صادق نیست. با این حال، این لزوما باعث برتری آن‌ها نمی‌شود (اغلب از نظر بازده نمونه، هزینه CPU یا دقت هنگام استفاده از یک الگوریتم برخط مصالحه‌هایی انجام می‌شود). رویکردهایی مانند ریز-دسته‌ای در آموزش شبکه‌های عصبی را می‌توان به عنوان تلاش‌هایی برای یافتن حد وسط بین الگوریتم‌های برخط و برون‌خط در نظر گرفت.

## یادگیری دسته‌ای[1]

در الگوریتم‌های یادگیری دسته‌ای، داده‌های آموزشی از ابتدا به‌طور کامل موجود و در دسترس عامل یادگیری است و پس از مرحله آموزش، نمی‌توان داده‌های آموزشی جدید را به سیستم اضافه کرد. در این الگوریتم‌ها، اگر داده‌های آموزشی بیش از حد بزرگ باشد، دوره آموزشی طولانی و زمان‌بر خواهد بود و در برخی موارد ممکن است فضای کافی برای ذخیره کل داده‌های آموزشی وجود نداشته باشد.

## یادگیری افزایشی[2]

در الگوریتم‌های یادگیری افزایشی، داده‌های آموزشی ممکن است از ابتدا معلوم یا کامل نباشند و یا ممکن است در طول زمان اضافه شوند. به عبارت دیگر، این امکان وجود دارد که این الگوریتم‌ها پس از مرحله آموزش، داده‌های آموزشی جدیدی را وارد کنند. هدف این الگوریتم‌ها حفظ نتایج مراحل قبلی آموزش و بهبود عملکرد عامل یادگیری تنها با یادگیری الگوهای جدید است. در واقع، این الگوریتم‌ها بدون نیاز به آموزشِ مجددِ الگوریتم‌های قدیمی که ممکن است دیگر در دسترس نباشند، خود را با ورود الگوریتم‌های جدید تطبیق داده و بروز می‌کنند.

یادگیری افزایشی یکی از موضوعات مهم در یادگیری ماشین است. تعاریف و تفاسیر متنوعی از یادگیری افزایشی را می‌توان در متون از جمله یادگیری برخط، دسته‌بندی‌مجدد نادرست[3] نمونه‌های قبلی یا توسعه و هرس معماری خوشه‌ای یافت. هدف کلی سیستم مبتنی‌بر یادگیری افزایشی، بروزرسانی مفروضات و اطلاعات قبلی خود در هنگام معرفی نمونه‌های جدید، بدون

---

[1] Batch Learning

[2] Incremental Learning

[3] incorrect reclassification

استفاده مجدد از نمونه‌های قبلی است. با این حال، باید به این نکته توجه داشت که، چنین سیستمی هنگام ارائه نمونه‌های جدید، نتایج آموزش مراحل قبلی خود را بر روی نمونه‌های قدیمی فراموش نمی‌کند، بلکه دانش گذشته خود را نسبت به نمونه‌های جدید بهبود می‌بخشد. به عبارت دیگر، یادگیری افزایشی، زمانی یادگیری افزایشی خوانده می‌شود که دارای ویژگی‌های زیر باشد (*Dulhare et al., 2020*):

- **توانایی بدست آوردن دانش اضافی هنگام معرفی داده‌های جدید.**
- **توانایی نگهداری اطلاعات آموخته‌شده از مراحل قبلی یادگیری.**
- **امکان یادگیری کلاس جدید در صورت ارائه نمونه جدید.**

بر اساس مطالعات انجام شده بر روی الگوریتم‌ها با قابلیت یادگیری افزایشی، می‌توان این الگوریتم‌ها را بر اساس معیارهای مختلف دسته‌بندی کرد. به عنوان مثال، این الگوریتم‌ها را می‌توان بر اساس توانایی آن‌ها در نگهداری داده‌های آموزشی به سه دسته تقسیم کرد:

- **داده‌های کامل:** این مجموعه از الگوریتم‌ها قادر است تمام داده‌های آموزشی را بدون از دست دادن داده‌های قدیمی ذخیره کند. از مزایای این الگوریتم‌ها می‌توان به نوسازی و بروزرسانی کارآمد و دستیابی به دقت دقیق اشاره کرد. با این حال، به دلیل وجود تمام داده‌های آموزشی، این الگوریتم‌ها به فضای ذخیره‌سازی زیادی نیاز دارند.

- **داده‌های جزئی:** این الگوریتم‌ها فقط داده‌های خاصی را نگه می‌دارند. بنابراین آن‌ها به درجه‌ای از سازش بین دقت و استفاده از حافظه می‌رسند.

- **بدون‌داده:** این دسته الگوریتم‌ها فقط اطلاعات آماری مربوط به داده‌ها را ذخیره می‌کنند و تمام داده‌ها را دور می‌اندازند. بنابراین دقت این الگوریتم‌ها بسته به نوع داده‌های ذخیره شده کم‌تر از دو دسته بالا است. با این حال، میزان استفاده از حافظه چنین الگوریتم‌هایی نیز کم است.

# یادگیری خودنظارتی[1]

یادگیری خودنظارتی را می‌توان نسخه پیشرفته‌تری از یادگیری بدون‌نظارت نامید که به داده‌های نظارتی همراه با آن نیاز دارد. فقط در این مورد، برچسب‌گذاری داده‌ها توسط انسان انجام نمی‌شود و این خود مدل است که به برچسب‌گذاری را از داده‌ها بدست می‌آورد. از آنجایی که نیازی به بازخورد انسانی در زمینه برچسب‌گذاری داده ها ندارد، یادگیری خودنظارتی را می‌توان شکل مستقلی از یادگیری بانظارت در نظر گرفت. یادگیری خودنظارتی، برچسب‌گذاری را با کمک ابردادهای تعبیه شده به عنوان داده‌های نظارتی انجام می‌دهد.

---

[1] Self-Supervised Learning

# یادگیری چندنمونه‌ای[1]

یادگیری چندنمونه‌ای به عنوان نوعی از یادگیری بانظارت برای مسائل مربوط به دانش ناقص در مورد برچسب‌های نمونه‌های آموزشی پیشنهاد شده است. در یادگیری بانظارت، هر نمونه آموزشی با یک برچسب گسسته یا یک مقدار حقیقی اختصاص داده می‌شود. در مقابل، در یادگیری چندنمونه‌ای برچسب‌ها به کیف‌هایی از نمونه‌ها اختصاص داده می‌شوند (مجموعه‌ای مرتب از داده‌های آموزشی کیف نامیده می‌شود و کل کیف برچسب‌گذاری می‌شود). در حالت دودویی، اگر حداقل یک نمونه در آن کیف مثبت باشد، کیف دارای برچسب مثبت است و اگر همه نمونه‌های موجود در آن منفی باشند، کیف دارای برچسب منفی می‌شود. به عبارت دیگر، اگر یک نمونه با نتیجه مطابقت داشته باشد، کل کیف آن مثبت است و اگر مطابقت نداشته باشد، کل کیف برابر با منفی است. هدف چندنمونه‌ای دسته‌بندی کیف‌ها یا نمونه‌های دیده نشده بر اساس کیف‌های برچسب‌دار به عنوان داده‌های آموزشی است.

## یادگیری ماشین چگونه کار می‌کند؟

یادگیری ماشین به عنوان یک فرآیند خودکار که الگوها را از داده‌ها استخراج می‌کند، تعریف می‌شود. برای ساخت مدل‌های مورد استفاده در برنامه‌های تحلیل داده‌های پیش‌گویانه، از یادگیری ماشین بانظارت استفاده می‌کنیم. رویکرد یادگیری ماشین با نظارت به‌طور خودکار مدلی را از رابطه بین مجموعه‌ای از ویژگی‌های توصیفی و یک ویژگی هدف، بر اساس مجموعه‌ای از مثال‌ها یا نمونه‌های گذشته می‌آموزد. سپس، می‌توانیم از این مدل برای پیش‌بینی نمونه‌های جدید استفاده کنیم. این دو مرحله جداگانه در شکل ۴ ـ ۱ نشان داده شده است.

جدول ۴ ـ ۱ مجموعه‌ای از نمونه‌های گذشته یا مجموعه داده‌های وام مسکنی را که یک بانک در گذشته اعطا کرده است، فهرست کرده است. این مجموعه داده شامل ویژگی‌های توصیفی است که وام مسکن را توصیف می‌کند و یک ویژگی هدف که نشان می‌دهد که آیا متقاضی وام مسکن در نهایت نتوانسته وام را پرداخت کند یا آن را به‌طور کامل بازپرداخت کرده است. ویژگی‌های توصیفی دارای سه سابقه (اطلاعات) در مورد وام مسکن است: شغل (که می‌تواند اداری یا صنعتی باشد)، سن متقاضی و نسبت بین حقوق متقاضی و مبلغ وام گرفته‌شده (نسبت وام به حقوق). ویژگی هدف (نتیجه) به صورت پیش‌فرض یا بازپرداخت تنظیم شده است. در اصطلاح یادگیری ماشین، هر ردیف در مجموعه داده به عنوان یک **نمونه آموزشی** و مجموعه داده کلی به عنوان **مجموعه داده آموزشی** نامیده می‌شود.

---

[1] Multiple Instance Learning

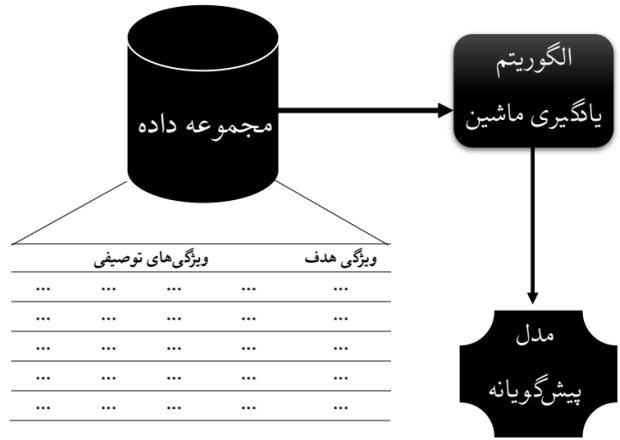

(آ) یادگیری یک مدل از مجموعه‌ای از نمونه‌های‌گذشته

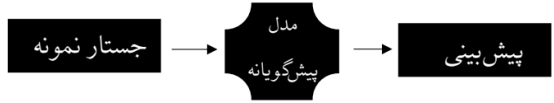

(ب) استفاده از مدل برای پیش‌بینی

**شکل ۴ ــ ۱** دو مرحله در یادگیری ماشین بانظارت: (آ) یادگیری و (ب) پیش‌بینی.

**جدول ۴ ــ ۱** مجموعه داده امتیازدهی وام

| نتیجه | نسبت وام به حقوق | سن | شغل | شماره |
|---|---|---|---|---|
| بازپرداخت | ۲/۹٦ | ۳٤ | صنعتی | ۱ |
| پیش‌فرض | ٤/٦٤ | ٤۱ | اداری | ۲ |
| پیش‌فرض | ۳/۲۲ | ۳٦ | اداری | ۳ |
| پیش‌فرض | ۳/۱۱ | ٤۱ | اداری | ٤ |
| پیش‌فرض | ۳/۸۰ | ٤۸ | صنعتی | ٥ |
| بازپرداخت | ۲/٥۲ | ٦۱ | صنعتی | ٦ |
| بازپرداخت | ۱/٥۰ | ۳۷ | اداری | ۷ |
| بازپرداخت | ۱/۹۳ | ٤۰ | اداری | ۸ |
| پیش‌فرض | ٥/۲٥ | ۳۳ | صنعتی | ۹ |
| پیش‌فرض | ٤/۱٥ | ۳۲ | صنعتی | ۱۰ |

نمونه‌ای از یک مدل پیش‌بینی بسیار ساده برای این مجموعه داده به‌صورت زیر می‌باشد:

**if** نسبت وام به حقوق $> 3$ **then**

پیش‌فرض=نتیجه

**else**

بازپرداخت=نتیجه

**end if**

می‌توان گفت که این مدل با این مجموعه داده سازگار است، چراکه هیچ نمونه‌ای در مجموعه داده وجود ندارد که مدل پیش‌بینی درستی برای آن انجام ندهد. هنگامی که درخواست‌های جدید وام مسکن انجام می‌شود، می‌توانیم از این مدل برای پیش‌بینی اینکه آیا متقاضی وام مسکن را بازپرداخت می‌کند یا آن را بازپرداخت نخواهد کرد، پیش‌بینی کنیم و بر اساس این پیش‌بینی تصمیمات وام‌دهی را اتخاذ کنیم.

الگوریتم‌های یادگیری ماشین فرآیند یادگیری مدلی را خودکار می‌کنند که رابطه بین ویژگی‌های توصیفی و ویژگی هدف در یک مجموعه داده را نشان می‌دهد. برای مجموعه داده‌های ساده همانند آنچه در جدول ٤ ــ ١ ارائه شده است، ممکن است بتوانیم به صورت دستی یک مدل پیش‌بینی ایجاد کنیم و در مثالی در این مقیاس، یادگیری ماشین کاربردی برای ما ندارد. حال فرض کنید همین مجموعه داده شامل ویژگی‌های توصیفی بیشتری باشد، همانند مقدار وامی که شخص دریافت می‌کند، حقوق دارنده وام مسکن، نوع ملکی که وام مسکن به آن مربوط می‌شود و تعداد دفعاتی که شخص وام گرفته است. حال، مدل پیش‌بینی ساده‌ای که فقط از ویژگی نیست وام به حقوق استفاده می‌کرد، دیگر با چنین مجموعه داده‌ای سازگار نیست. از این‌رو، پیدا کردن چنین قاعده‌ای در این مجموعه داده بسیار سخت است. به عبارت دیگر، یادگیری دستی این مدل با بررسی داده‌ها تقریبا غیرممکن است. اینجاست که یادگیری ماشین وارد عمل می‌شود، چرا که این کار برای الگوریتم یادگیری ماشین، بسیار ساده است. از این‌رو، وقتی می‌خواهیم مدل‌های پیش‌گویانه را از مجموعه داده‌های بزرگ با ویژگی‌های متعدد بسازیم، راه حا استفاده از یادگیری ماشین است.

الگوریتم‌های یادگیری ماشین با جستجوی مجموعه‌ای از مدل‌های پیش‌بینی ممکن در تلاش برای ساخت مدلی هستند که به بهترین شکل رابطه بین ویژگی‌های توصیفی و ویژگی هدف را در یک مجموعه داده نشان می‌دهد. یک معیار واضح برای هدایت این جستجو، جستجوی مدل‌هایی است که با داده‌ها سازگار باشد. با این حال، حداقل دو دلیل وجود دارد که چرا جستجوی ساده برای مدل‌های سازگار برای یادگیری مدل‌های پیش‌بینی مفید کافی نیست. اول اینکه، وقتی با مجموعه داده‌های بزرگ سروکار داریم، احتمالا نویز در داده‌ها نیز وجود دارد و مدل‌های پیش‌گویانه که با داده‌های پر نویز سازگار هستند، پیش‌بینی‌های نادرستی می‌کنند. دوم اینکه، در اکثر مواقع در پروژه‌های یادگیری ماشین، مجموعه آموزشی تنها یک نمونه کوچکی از مجموعه نمونه‌های ممکن در دامنه را نشان می‌دهد. در نتیجه، یادگیری ماشین یک **مساله بدطرح**[1] است، یعنی *مساله‌ای که تنها با استفاده از داده‌های موجود نمی‌توان راه حل منحصر به فردی برای آن تعیین کرد*. می‌توانیم با استفاده از مثالی نشان دهیم که چگونه یادگیری ماشین یک مساله بدطرح است که در آن تیم تجزیه و تحلیل در یک سوپرمارکت زنجیره‌ای می‌خواهد

---

[1] ill-posed problem

مشتریان خود را در گروه‌های جمعیتی مجرد، متاهل یا خانواده، تنها بر اساس عادات خریدشان طبقه‌بندی کند. مجموعه داده ارائه شده در جدول ٤ـ۲ شامل ویژگی‌های توصیفی است که عادات خرید پنج مشتری را توصیف می‌کند. ویژگی‌های توصیفی در این جدول نشان می‌دهند که آیا یک مشتری غذای کودک، نوشیدنی گازدار یا محصولات گیاهی ارگانیک خریداری می‌کند یا نه. هر ویژگی یکی از دو مقدار را در بر می‌گیرد، *بله* یا *خیر*. در کنار این ویژگی‌های توصیفی، یک ویژگی هدف وجود دارد که گروه جمعیتی را برای هر مشتری (مجرد، زوج یا خانواده) توصیف می‌کند. مجموعه داده ارائه شده در جدول ٤ـ۲ به عنوان یک مجموعه داده برچسب‌دار نامیده می‌شود، چراکه شامل مقادیری برای ویژگی هدف است.

**جدول ٤ـ۲** مجموعه داده خرده‌فروشی ساده

| هدف | محصولات گیاهی ارگانیک | نوشیدنی گازدار | غذای کودک | شماره |
|---|---|---|---|---|
| زوج | خیر | خیر | خیر | ۱ |
| خانواده | بله | بله | بله | ۲ |
| خانواده | خیر | بله | بله | ۳ |
| زوج | بله | خیر | خیر | ٤ |
| مجرد | بله | خیر | خیر | ٥ |

تصور کنید که ما سعی می‌کنیم یک مدل پیش‌بینی برای این سناریوی خرده‌فروشی با جستجوی مدلی مطابق با مجموعه داده یاد بگیریم. اولین کاری که باید انجام دهیم این است که بفهمیم چند مدل ممکن مختلف واقعا برای این سناریو وجود دارد. *این مرحله مجموعه مدل‌های پیش‌بینی راکه الگوریتم یادگیری ماشین جستجو می‌کند، تعریف می‌کند.* از منظر جستجو برای یک مدل سازگار، مهمترین ویژگی یک مدل پیش‌بینی این است که یک نگاشت *از هر ترکیب ممکنی از مقادیر ویژگی توصیفی را به یک پیش‌بینی برای ویژگی هدف تعریف می‌کند.* برای سناریوی خرده فروشی، تنها سه ویژگی توصیفی دودویی وجود دارد، از این‌رو ۸ $= ۲^۳$ ترکیب ممکن از مقادیر ویژگی توصیفی وجود دارد. با این حال، برای هر یک از این ۸ ترکیب ممکن از مقادیر ویژگی توصیفی، ۳ مقدار ویژگی هدف ممکن وجود دارد.، بنابراین، این بدان معناست که ٦٥٦۱ $= ۳^۸$ مدل پیش‌بینی ممکن وجود دارد که می‌توان از آن‌ها استفاده کرد. جدول (آ) ٤ـ۳ رابطه‌ای بین ترکیبی از مقادیر ویژگی توصیفی و مدل‌های پیش‌بینی را برای سناریوی خرده‌فروشی نشان می‌دهد. ترکیب ویژگی‌های توصیفی در سمت راست جدول فهرست شده است و مجموعه مدل‌های بالقوه برای این دامنه به صورت $M_۱$ تا $M_{٦٥٦۱}$ در سمت چپ جدول نشان داده شده است. با استفاده از مجموعه داده آموزشی جدول ٤ـ۲ یک الگوریتم یادگیری ماشین مجموعه کامل ٦٥٦۱ مدل پیش‌بینی ممکن برای این سناریو را تنها به مدل‌هایی کاهش می‌دهد که با نمونه‌های آموزشی سازگار هستند. جدول (ب) ٤ـ۳ این را نشان می‌دهد. ستون‌های خالی در جدول مدل‌هایی را نشان می‌دهد که با داده‌های آموزشی سازگار نیستند.

جدول (ب) ۳-۴ همچنین نشان می‌دهد که مجموعه داده آموزشی حاوی نمونه‌ای برای هر ترکیب ممکن از مقادیر ویژگی توصیفی نیست. به عبارت دیگر، هنوز تعداد زیادی مدل پیش‌بینی بالقوه وجود دارد که پس از حذف مدل‌های ناسازگار با مجموعه داده آموزشی سازگار می‌مانند. یعنی، سه ترکیب مقدار ویژگی توصیفی باقی‌مانده است که مقدار ویژگی هدف درست برای آن‌ها مشخص نیست. بنابراین، ۲۷ = $3^3$ مدل بالقوه وجود دارد که با داده‌های آموزشی سازگار می‌مانند. سه مورد از این موارد $M_2$، $M_4$ و $M_0$ در جدول نشان داده شده است. از آنجایی که نمی‌توان یک مدل منسجم واحد را بر اساس مجموعه داده‌های آموزشی نمونه یافت، می‌گوییم یادگیری ماشین اساسا یک مشکل بدطرح است.

**جدول ۳-۴** مدل‌های پیش‌بینی بالقوه (آ) قبل و (ب) پس از در دسترس بودن داده‌های آموزشی

(آ) قبل از اینکه داده‌های آموزشی در دسترس قرار گیرد

| $M_{6561}$ | ... | $M_5$ | $M_4$ | $M_3$ | $M_2$ | $M_1$ | هدف | محصولات گیاهی .. | نوشیدنی گازدار | غذای کودک | شماره |
|---|---|---|---|---|---|---|---|---|---|---|---|
| زوج | | زوج | زوج | مجرد | زوج | زوج | ؟ | خیر | خیر | خیر | ۱ |
| مجرد | | زوج | زوج | مجرد | زوج | مجرد | ؟ | بله | خیر | خیر | ۲ |
| خانواده | | مجرد | زوج | مجرد | خانواده | خانواده | ؟ | خیر | بله | خیر | ۳ |
| زوج | | مجرد | زوج | مجرد | مجرد | مجرد | ؟ | بله | بله | خیر | ٤ |
| خانواده | | خانواده | خانواده | خانواده | زوج | زوج | ؟ | خیر | خیر | بله | ٥ |
| زوج | | خانواده | خانواده | خانواده | خانواده | زوج | ؟ | بله | خیر | بله | ٦ |
| مجرد | | خانواده | خانواده | خانواده | خانواده | مجرد | ؟ | خیر | بله | بله | ۷ |
| خانواده | | زوج | خانواده | خانواده | مجرد | مجرد | ؟ | بله | بله | بله | ۸ |

(ب) پس از اینکه داده‌های آموزشی در دسترس هستند

| ... | $M_5$ | $M_4$ | $M_2$ | هدف | محصولات گیاهی .. | نوشیدنی گازدار | غذای کودک | شماره |
|---|---|---|---|---|---|---|---|---|
| | زوج | زوج | زوج | زوج | خیر | خیر | خیر | ۱ |
| | زوج | زوج | زوج | زوج | بله | خیر | خیر | ۲ |
| | مجرد | زوج | خانواده | ؟ | خیر | بله | خیر | ۳ |
| | مجرد | زوج | مجرد | مجرد | بله | بله | خیر | ٤ |
| | خانواده | خانواده | زوج | ؟ | خیر | خیر | بله | ٥ |
| | خانواده | خانواده | خانواده | خانواده | بله | خیر | بله | ٦ |
| | خانواده | خانواده | خانواده | خانواده | خیر | بله | بله | ۷ |
| | زوج | خانواده | مجرد | ؟ | بله | بله | بله | ۸ |

ممکن است فکر کنیم که داشتن چندین مدل که با داده‌ها سازگار هستند چیز خوبی است. با این حال، مشکل این است که اگرچه این مدل‌ها در مورد اینکه کدام پیش‌بینی باید برای نمونه‌های مجموعه داده آموزشی انجام شود، توافق دارند، اما در مورد اینکه کدام پیش‌بینی‌ها باید برای نمونه‌هایی که در مجموعه داده آموزشی نیستند و باید بازگردانده شوند، موافق نیستند (درست عمل نمی‌کند). به عنوان مثال، اگر یک مشتری جدید شروع به خرید به سوپرمارکت کند و غذای کودک، نوشیدنی گازدار و محصولات گیاهی ارگانیک بخرد، مجموعه مدل‌های ثابت ما با توجه

به پیش‌بینی‌ای که باید برای این مشتری برگردانده شود، با یکدیگر تناقض دارند. به عنوان مثال، $M_1$ هدف را مجرد، $M_2$ هدف را خانواده و $M_3$ هدف را زوج برمی‌گرداند.

معیار سازگاری با داده‌های آموزشی هیچ راهنمایی را با توجه به این که کدام یک از مدل‌های سازگار در برخورد با جستارهایی[1] (پرس‌وجو) که خارج از مجموعه داده آموزشی هستند، فراهم نمی‌کند. در نتیجه آن، ما نمی‌توانیم از مجموعه مدل‌های سازگار برای پیش‌بینی این جستارها استفاده کنیم. در واقع، جستجوی مدل‌های پیش‌گویانه که با مجموعه داده‌ها مطابقت دارند، معادل به با خاطرسپاری مجموعه داده است. براین اساس، هیچ یادگیری از این طریق صورت نمی‌گیرد. چراکه مجموعه مدل‌های سازگار به ما چیزی در مورد رابطه اساسی بین ویژگی‌های توصیفی و هدف، فراتر از آنچه که یک نگاه ساده به مجموعه داده آموزشی ارائه می‌دهد، به ما ارائه نمی‌دهد.

اگر قرار است یک مدل پیش‌گویانه مفید باشد، باید بتواند برای جستارهایی که در داده‌ها وجود ندارند، پیش‌بینی خوبی انجام دهد. یک مدل پیش‌گویانه که پیش‌بینی‌های درست را برای این پرسش‌ها انجام می‌دهد، رابطه اساسی بین ویژگی‌های توصیفی و هدف را نشان می‌دهد و گفته می‌شود که به خوبی **تعمیم**[2] می‌یابد. عملا، *هدف یادگیری ماشین یافتن مدل پیش‌گویانه‌ای است که بهترین تعمیم را دارد.* برای یافتن این بهترین مدل، یک الگوریتم یادگیری ماشین باید از معیارهایی برای انتخاب از بین مدل‌های نامزدی که در طول جستجوی خود در نظر می‌گیرد استفاده کند. با توجه به اینکه سازگاری با مجموعه داده‌ها معیار مناسبی برای انتخاب بهترین مدل پیش‌گویانه نیست، از کدام معیار باید استفاده کنیم؟ پاسخ‌های بالقوه زیادی برای این سوال وجود دارد و به همین دلیل است که الگوریتم‌های مختلف یادگیری ماشین وجود دارند. هر الگوریتم یادگیری ماشین از معیارهای انتخاب مدل متفاوتی برای هدایت جستجوی خود برای بهترین مدل پیش‌گویانه استفاده می‌کند. بنابراین، وقتی انتخاب می‌کنیم از یک الگوریتم یادگیری ماشین به جای الگوریتم دیگری استفاده کنیم، عملا، انتخاب می‌کنیم که از یک معیار انتخاب مدل به جای دیگری استفاده کنیم.

همه معیارهای مختلف انتخاب مدل شامل مجموعه‌ای از مفروضات در مورد ویژگی‌های مدل است که ما می‌خواهیم الگوریتم استنتاج کند. مجموعه‌ی مفروضاتی که معیارهای انتخاب مدل یک الگوریتم یادگیری ماشین را تعریف می‌کند، به عنوان **باياس استقرایی**[3] الگوریتم

---

[1] queries

[2] generalize

[3] inductive bias

یادگیری ماشین شناخته می‌شود. دو نوع بایاس استقرایی وجود دارد که الگوریتم یادگیری ماشین می‌تواند از آن‌ها استفاده کند: **بایاس محدودیت**[1] و یک **بایاس ترجیحی**[2].

تعریف　　بایاس محدودیت

محـدود کردن فضای فرضیه به طوری که عناصر موجود در فضای فرضیه محـدود می‌شوند و تعـداد جستجوها بدون تاثیر بر جستجو کاهش می‌یابد.

بایاس محدودیت، قدرت بازنمایی یک الگوریتم یا مجموعه فرضیه‌هایی است که الگوریتم ما در نظر خواهد گرفت.

تعریف　　بایاس ترجیحی

تغییر روش جستجو و جستجوی ناقص تمام فضاهای فرضیه.

بایاس ترجیحی به سادگی همان چیزی است که یک الگوریتم یادگیری با نظارت ترجیح می‌دهد. به عنوان مثال، یک الگوریتم درخت تصمیم ممکن است درختان کوتاه‌تر و کم‌تر پیچیده را ترجیح دهد. به عبارت دیگر، این بایاس، باور الگوریتم ما در مورد آنچه که یک فرضیه خوب را می‌سازد است.

یک بایاس محدودیت، مجموعه مدل‌هایی را که الگوریتم در طول فرآیند یادگیری در نظر می‌گیرد، محدود می‌کند. یک بایاس ترجیحی، الگوریتم یادگیری را هدایت می‌کند تا مدل‌های خاصی را بر مدل‌های دیگر ترجیح دهد. نکته مهم این است که *استفاده از یک بایاس استقرایی یک پیش‌نیاز ضروری برای رخ‌دادن یادگیری است*. بدون بایاس استقرایی، یک الگوریتم یادگیری ماشین نمی‌تواند چیزی فراتر از آنچه در داده است را بیاموزد. *به‌طور خلاصه، یادگیری ماشین با جستجو در میان مجموعه‌ای از مدل‌های بالقوه برای یافتن مدل پیش‌گویانه‌ای که به بهترین وجه، فراتر از مجموعه داده تعمیم می‌یابد، کار می‌کند.* الگوریتم‌های یادگیری ماشین از دو منبع اطلاعاتی برای هدایت این جستجو استفاده می‌کنند، مجموعه داده‌ی آموزشی و بایاسِ استقراییِ فرض شده توسط الگوریتم.

## بایاس استقرایی و ضرورت آن در یادگیری

همه روش‌های یادگیری دارای بایاس استقرایی هستند. *بایاس استقرایی فرآیند یادگیری اصول کلی بر اساس نمونه‌های خاص است*. به عبارت دیگر، این همان کاری است که هر الگوریتم

---

یادگیری ماشین انجام می‌دهد؛ زمانی که یک پیش‌بینی برای هر نمونه آزمایشی دیده‌نشده بر اساس تعداد محدودی از نمونه‌های آموزشی تولید می‌کند. در واقع، بایاس استقرایی به محدودیت‌هایی اشاره دارد که توسط مفروضات ایجاد شده در روش یادگیری اعمال می‌شود. بایاس استقرایی به این معنی است که راه‌حل‌های بالقوه‌ای وجود دارند که ما نمی‌توانیم کشف کنیم و در نتیجه در فضای نمونه که بررسی می‌کنیم موجود نیستند. ممکن است این یک محدودیت بسیار بد به نظر برسد. اما، در واقع *بایاس استقرایی برای یادگیری ضروری است.* برای داشتن یک یادگیرنده بدون‌بایاس، فضای نمونه باید شامل هر فرضیه ممکنی باشد که احتمالاً می‌تواند بیان شود. این یک محدودیت شدید ایجاد می‌کند: راه حلی که یادگیرنده تولید می‌کند هرگز نمی‌تواند کلی‌تر از مجموعه کامل داده‌های آموزشی باشد. به عبارت دیگر، می‌تواند داده‌هایی را که قبلاً دیده بود طبقه‌بندی کند، اما قادر به تعمیم به‌منظور طبقه‌بندی داده‌های جدید و نادیده نخواهد بود.

> **بدون بایاس استقرایی، یادگیرنده نمی‌تواند بهتر از حدس زدن تصادفی، از مثال‌های مشاهده‌شده به نمونه‌های جدید تعمیم دهد.**

در قرن چهاردهم، ویلیام اکام "**تیغ اوکام**"[1] را پیشنهاد کرد که به‌طور ساده بیان می‌کند بهتر است ساده‌ترین فرضیه را برای توضیح هر پدیده‌ای انتخاب کنید. می‌توانیم این را نوعی بایاس استقرایی در نظر بگیریم که بیان می‌کند، بهترین فرضیه برای برازش مجموعه‌ای از داده‌های آموزشی، ساده‌ترین فرضیه است. این به این معنی است که اگر دو الگوریتم تقریبا عملکرد مشابهی برای معیارهای ارزیابی در یک پروژه خاص داشته باشند، ما باید "ساده‌تر" را ترجیح دهیم.

اما "ساده‌تر" در این زمینه به چه معناست؟ به‌طور کلی می‌توان اینگونه برداشت کرد که ساده‌تر به معنای الگوریتمی است که کم‌ترین پیچیدگی را برای استنتاج دارد (چراکه به عنوان مثال از متغیرهای کم‌تری استفاده می‌کند و به مهندسی ویژگی کم‌تری نیاز دارد) و تفسیر آن آسان‌تر است. *با این حال، باید توجه داشته باشید که این نوع معاوضه‌ها معمولاً زمانی معنا پیدا می‌کنند که دقت مدل ساده‌تر، حداقل در محدوده مشابه مدل پیچیده‌تر باشد. در سناریویی که در آن نیم‌درصد دقت ممکن است به میلیون‌ها دلار درآمد اضافی یا صرفه‌جویی در هزینه تبدیل شود، ممکن است مدلی را انتخاب کنید که تفسیر آن سخت‌تر است یا به زمان بیشتری برای توسعه نیاز دارد.*

بهتر است همیشه در هر پروژه یادگیری ماشین در ابتدا تمرکز را بر روی مساله‌ی تجاری که قصد دارید به آن پاسخ دهید و با فرمول‌بندی معیارهای کلیدی موفقیت برای تجزیه و تحلیل شروع کنید. با فرض این‌که تمام معیارهای کلیدی دیگر (تقریباً) برابر هستند، در آن زمان از تیغ

---

[1] Occam's Razor

*اوکام* استفاده کنید و مدلی را انتخاب کنید که ساده‌ترین برای تفسیر، توضیح، توسعه و نگهداری باشد. به عبارت دیگر، *ساده‌ترین مدل را ترجیح دهید که به اندازه کافی دقیق باشد، اما اطمینان حاصل کنید که فضای مساله را به‌خوبی می‌شناسید تا بدانید "به اندازه کافی دقیق"* در عمل به چه معناست. زیرا همان‌طور که اینشتین (شاید بزرگترین شاگرد اوکام)، زمانی گفت: "هر چیزی را باید تا حد امکان ساده کرد، اما نه ساده‌تر".

# کاربردهای یادگیری ماشین

یادگیری ماشین واژه‌ای باب‌شده در فناوری امروز است و روز به روز در حال رشد است. در صنعت، یادگیری ماشین راه را برای دست‌آوردها و ابزارهای فناوری هموار کرده است که چند سال پیش غیرممکن بود. علاوه بر این‌ها، ما در زندگی روزمره خود از یادگیری ماشین استفاده می‌کنیم حتی بدون اینکه چنین چیزی را بدانیم. در ادامه برخی از پرطرفدارترین برنامه‌های کاربردی یادگیری ماشین در دنیای واقعی را فهرست کرده‌ایم.

## تشخیص تصویر

تشخیص تصویر یکی از رایج‌ترین،مهم‌ترین و قابل توجه‌ترین کاربردهای یادگیری ماشین است. از آن برای شناسایی اشیاء، اشخاص، مکان‌ها و غیره استفاده می‌شود. این تکنیک برای تجزیه و تحلیل بیشتر مانند تشخیص الگو و یا تشخیص چهره به‌کار گرفته می‌شود.

## تشخیص گفتار

تشخیص گفتار فرآیند تبدیل دستورالعمل‌های صوتی به متن است که هم‌چنین به عنوان "گفتار به متن" یا "تشخیص گفتار رایانه‌ای" شناخته می‌شود. در حال حاضر، الگوریتم‌های یادگیری ماشین به‌طور گسترده‌ای توسط برنامه‌های مختلف تشخیص گفتار استفاده می‌شود.

## توصیه‌گر محصول

یکی از شناخته‌شده‌ترین کاربردهای یادگیری ماشین توصیه محصول است. توصیه محصول یکی از ویژگی‌های بارز تقریبا هر وب‌سایت تجارت الکترونیک امروزه است که یک برنامه پیشرفته از تکنیک‌های یادگیری ماشین است. با استفاده از یادگیری ماشین و هوش مصنوعی، وب‌سایت‌ها رفتار شما را بر اساس خرید قبلی، الگوی جستجوی شما، سابقه سبد خرید شما پیگیری می‌کنند و توصیه‌های محصول را ارائه می‌دهند.

## ترجمه زبان

یادگیری ماشین نقش مهمی در ترجمه یک زبان به زبان دیگر دارد. امروزه اگر از مکان جدیدی بازدید می‌کنیم و از زبان آن آگاهی نداریم، به هیچ وجه مشکلی ایجاد نمی‌کند، چراکه یادگیری

ماشین با تبدیل متن به زبان‌های شناخته شده به ما کمک می‌کند. GNMT گوگل[1] این ویژگی را ارائه می‌دهد که یک یادگیری ماشین عصبی است.

## خدمات مالی

بانک‌ها و سایر شرکت‌های مالی از فناوری یادگیری ماشین برای دو هدف اصلی استفاده می‌کنند: شناسایی بینش‌های ارزشمند در مورد داده‌ها و کاهش ریسک. این بینش‌ها می‌توانند فرصت‌های سرمایه‌گذاری را شناسایی کنند یا به سرمایه‌گذاران در یافتن زمان مناسب برای معامله کمک کنند. علاوه بر این، می‌تواند مشتریان پرخطر را شناسایی کند یا از تجزیه و تحلیل‌های سایبری برای شناسایی هشدارهای کلاهبرداری استفاده شود.

## نفت و گاز

یافتن منابع جدید انرژی، تجزیه و تحلیل مواد معدنی در زمین، پیش‌بینی خرابی سنسور پالایشگاه، تسهیل توزیع نفت برای افزایش بهره‌وری و هزینه و کاربردهای متعدد دیگری در یادگیری ماشین وجود دارد که همچنان در حال گسترش است.

## حمل و نقل

تجزیه و تحلیل داده‌ها برای شناسایی روندها و الگوها نقش مهمی در صنعت حمل و نقل ایفا می‌کند و به ساده‌سازی مسیرها و پیش‌بینی مسائل بالقوه افزایش سودآوری کمک می‌کند. برای شرکت‌های حمل و نقل کالا، حمل و نقل عمومی و سایر سازمان‌های حمل و نقل، تجزیه و تحلیل داده‌ها و مدل‌سازی یادگیری ماشین ابزار مهمی هستند.

## مراقبت‌های بهداشتی

رقابت امروز استفاده از یادگیری ماشین برای تجزیه و تحلیل در زمینه پزشکی است. بسیاری از استارت آپ‌های مختلف به دنبال مزایای یادگیری ماشین با داده‌های انبوه هستند تا حرفه‌ای‌ترین مراقبت‌های پزشکی را با هدف مشترک اتخاذ معقول‌ترین تصمیم‌ها ارائه دهند. امروزه، مصرف‌کنندگان بی‌شماری، حتی با تلفن‌های هوشمند خود می‌توانند طیف وسیعی از اطلاعات سلامتی را به‌طور منظم اندازه‌گیری کنند. سیستم‌های یادگیری ماشین می‌توانند مدلی از وضعیت سلامت فرد ارائه دهند و از توصیه‌هایی که سیستم به‌روزرسانی می‌کند برای بهبود سلامت فرد استفاده کنند.

---

[1] Google Neural Machine Translation

# ارتباط با سایر زمینه‌ها

یادگیری ماشینی یک زمینه چند رشته‌ای است و در واقع ارتباط تنگاتنگی بین این رشته و سایر علوم وجود دارد، به عنوان مثال، به برخی از این زمینه‌های مهم مرتبط با یادگیری ماشین در ادامه این بخش پرداخته شده است.

## هوش مصنوعی

هوش مصنوعی اساسا سیستمی است که هوشمند به نظر می‌رسد. با این حال، این تعریف خیلی دقیق و خوبی نیست. اما هوش مصنوعی دقیقا به چه معناست؟ بر اساس فرهنگ لغت کالینز، هوش مصنوعی *"شبیه‌سازی انسان‌ها و رفتارهای ذهنی آن‌ها توسط یک برنامه رایانه‌ای است"*. به عبارت ساده‌تر، سیستمی که می‌تواند رفتار انسان را تقلید کند. این رفتارها شامل حل مساله، یادگیری و برنامه‌ریزی است که برای مثال از طریق تجزیه و تحلیل داده‌ها و شناسایی الگوهای درون آن به منظور تکرار آن رفتارها بدست می‌آید.

به عبارت دیگر، کد، فناوری یا الگوریتمی که بتواند مقوله فهم شناختی را تقلید کند که در خود یا در دستاوردهای آن پدیدار می‌شود، هوش مصنوعی است. از این رو، هوش مصنوعی سازنده‌ی (پایه‌ریز) یادگیری ماشین است. در واقع، یادگیری ماشین زیرمجموعه اصلی هوش مصنوعی است و می‌تواند ماشین‌ها را قادر سازد تا با استفاده از روش‌های آماری، تجربیات خود را با کیفیت‌تر و دقیق‌تر کنند. این امر به رایانه‌ها و ماشین‌ها امکان می‌دهد تا دستورات را بر اساس داده‌ها و یادگیری خود اجرا کنند. این برنامه‌ها یا الگوریتم‌ها به گونه‌ای طراحی شده‌اند که بتوانند در طول زمان اطلاعات بیشتری کسب کنند و با داده‌های جدید بهتر شوند و تطبیق پیدا کنند.

## داده‌کاوی[1]

یادگیری ماشین و داده‌کاوی از تکنیک‌های یکسانی استفاده می‌کنند و و هم‌پوشانی دارند. با این حال، در حالی که یادگیری ماشین برمبنای ویژگی‌ها به یادگیری از داده‌های آموزشی متمرکز است، داده‌کاوی بر یافتن ویژگی‌های داده‌های ناشناخته متمرکز است (این مرحله استخراج دانش در پایگاه داده است). داده‌کاوی از بسیاری از روش‌های یادگیری ماشین اما با اهداف متفاوت استفاده می‌کند. از سوی دیگر، یادگیری ماشین همچنین از روش‌های داده‌کاوی به‌عنوان یادگیری بدون نظارت یا به‌عنوان یک مرحله پیش‌پردازش برای بهبود دقت یادگیرنده استفاده می‌کند.

---

[1] data mining

## یادگیری ماشین و داده‌کاوی چه وجه مشترکی دارند؟

هم داده‌کاوی و هم یادگیری ماشین در علم داده مورد استفاده قرار می‌گیرند که منطقی است، چراکه هر دو از داده‌ها استفاده می‌کنند. آن‌ها مکمل یکدیگر هستند، هر دوی آن‌ها اشتراکات زیادی دارند، اما به اهداف متفاوتی می‌رسند. هم داده‌کاوی و هم یادگیری ماشین در بازاریابی، شناسایی کارت اعتباری تقلبی، تجارت الکترونیک و خرده‌فروشی محبوب هستند. دانشمندان داده و مهندسان داده از هر دو برای کمک به مشاغل استفاده می‌کنند. به عنوان مثال، هم یادگیری ماشین و هم داده‌کاوی، مدیریت موجودی کارآمد، کنترل کیفیت و کارایی عملیاتی را بدون دخالت انسان ممکن می‌سازند. وقتی صحبت از یادگیری ماشین و داده‌کاوی به میان می‌آید، هم‌پوشانی‌های زیادی وجود دارد و افراد (به اشتباه) به جای یکدیگر از آن استفاده می‌کنند. اما درک تفاوت بین آن‌ها مهم است، زیرا بسته به اهداف و منابع خود از فرآیندها و معماری‌های مختلف استفاده خواهید کرد. وقتی از یادگیری ماشین و داده‌کاوی بدرستی استفاده می‌کنید، در مسیر درستی قرار دارید تا داده‌های خام را به بینش‌های ارزشمندی تبدیل کنید که بر نتیجه شما تاثیر می‌گذارد. این بینش‌ها می‌توانند عملیاتی، استراتژیک یا آماری باشند. به عنوان مثال، در یک انبار، ما از داده‌کاوی و تشخیص الگو برای حل مشکلات مسیریابی چیننده[1] استفاده می‌کنیم. در این سناریو، داده‌کاوی از تکنیک‌های یادگیری ماشین برای تخمین دقیق طول کوتاه‌ترین مسیر ممکن برای افزایش کارایی استفاده می‌کند.

داده‌کاوی قلب هوش مصنوعی، یادگیری ماشین، یادگیری عمیق و آمار است. در حالی که در ۳۰ سال گذشته به شهرت رسید، سابقه‌ای بیش از ۲۰۰ سال دارد. دانشمندان داده از تکنیک‌های داده‌کاوی برای یافتن الگوهای پنهان اما مفید در پایگاه‌های داده بزرگ استفاده می‌کنند که نمی‌توانیم از طریق تکنیک‌های پرس‌وجو و گزارش به این سوالات بپردازیم. از آنجایی که داده‌ها به سرعت و به صورت تصاعدی رشد می‌کنند، باید از این روش‌ها برای تجزیه و تحلیل و پیش‌بینی مفید، استفاده کنیم. تکنیک‌های یادگیری ماشین به پردازش سریع داده‌ها کمک می‌کنند و به‌طور خودکار نتایج را بسیار سریع‌تر بدست می‌آورند. تکنیک‌های داده‌کاوی الگوها و روندها را در مجموعه داده‌های گذشته برای پیش‌بینی نتایج آینده برجسته می‌کند. این نتایج به شکل نمودار، گراف و موارد دیگر است.

## تفاوت یادگیری ماشین و داده‌کاوی در چیست؟

یادگیری ماشین بخشی از هوش مصنوعی است که به سیستم‌ها توانایی یادگیری و بهبود خودکار بر اساس تجربه را می‌دهد. در این سناریو، می‌توانیم الگوریتم‌های پیچیده‌ای بسازیم که مجموعه

---

[1] picker routing problems

داده‌های بزرگی را پردازش کرده و از آن‌ها برای یادگیری خود، بدون برنامه‌نویسی صریح استفاده کند. یادگیری ماشین از الگوریتم‌های پیچیده‌ای استفاده می‌کند که از طریق تجربه یاد می‌گیرند و پیش‌بینی می‌کنند. این الگوریتم‌های هوشمند از طریق داده‌های ورودی‌که داده‌های آموزشی هستند در حال بهبود دائمی می‌باشند. هدف اصلی کاوش، درک داده‌ها و ساخت مدل‌هایی است که روابط بین نقاط داده را یاد می‌گیرند. در مقابل، داده‌کاوی یا فرآیند کشف دانش، عمل کاوش در میان مجموعه داده‌ها را توصیف می‌کند. این رویکرد یک تکنیک محبوب برای کشف الگوها و روندهای ناشناخته است. چنین کشف دانش در پایگاه‌های داده بسیار فراتر از یک تحلیل ساده است. این بدان معناست که داده‌کاوی داده‌ های قابل استفاده را از مجموعه گسترده‌تری از داده‌های خام استخراج می‌کند.

تفاوت اصلی بین یادگیری ماشین و داده‌کاوی در سطح مداخله انسانی است که برای تکمیل یک کار لازم است. یادگیری ماشین، براساس هوش مصنوعی، رایانه‌ای است که برای انجام یک کار تا حدی جایگزین انسان می‌شود. در مقابل، داده‌کاوی به مداخله انسان نیاز دارد تا یک کار را کامل کند. در این سناریو، دانشمندان داده از ابزارهایی برای استخراج و کشف الگوهای مفید در داده‌ها استفاده می‌کنند. در این مورد، فضای زیادی برای خطای انسانی وجود دارد. به صورت مقایسه‌ای، نتایج حاصل از طریق یادگیری ماشین، در مقایسه با داده‌کاوی بسیار دقیق‌تر هستند. یادگیری ماشین از مدل‌های پیشگویانه، الگوریتم‌های آماری و شبکه‌های عصبی برای رسیدن به این کار استفاده می‌کند. داده‌کاوی از انبارهای داده و تکنیک‌های ارزیابی الگو برای یافتن نگرش‌های ارزشمند استفاده می‌کند. می‌توان تفاوت‌های کلیدی بین یادگیری ماشین و داده‌کاوی را در کاربرد، مفاهیم، پیاده‌سازی و قابلیت یادگیری پیدا کرد.

- **کاربرد:** الگوریتم‌های یادگیری ماشین داده‌ها را در قالب داده‌های متعارف (استاندارد) می‌طلبند. برای تجزیه و تحلیل داده‌ها با یادگیری ماشین، باید مجموعه داده را از شکل اصلی خود به یک قالب متعارف تبدیل کنید. این به الگوریتم‌های هوشمند کمک می‌کند تا داده‌ها را به سرعت درک کنند. یادگیری ماشین همچنین به حجم عظیمی از داده‌ها برای ارائه نتایج دقیق نیاز دارد. داده‌کاوی نیز می‌تواند نتایجی را ایجاد کند، اما در حجم کمتری از داده‌ها.

- **مفاهیم:** الگوریتم‌های یادگیری ماشین بر اساس این مفهوم اجرا می‌شوند که ماشین‌ها از داده‌های موجود یاد می‌گیرند. این رویکرد همچنین به بهبود خود کمک می‌کند. یادگیری ماشین، مدل‌هایی را بر اساسِ منطقِ پشت داده‌ها توسعه می‌دهد. این به پیش‌بینی نتایج آینده (با استفاده از روش‌های داده‌کاوی) کمک می‌کند. در مقابل، داده‌کاوی بر استخراج اطلاعات با استفاده از تکنیک‌هایی متمرکز است که به شناسایی الگوها و روندها در داده‌ها کمک می‌کند.

- **پیاده‌سازی:** ما می‌توانیم یادگیری ماشین را با استفاده از الگوریتم‌های هوشمند همانند رگرسیون خطی، درخت تصمیم، شبکه‌های عصبی و غیره پیاده‌سازی کنیم. یادگیری ماشین اساسا از الگوریتم‌های خودکار و شبکه‌های عصبی برای پیش‌بینی نتایج استفاده می‌کند. در مقابل، وقتی صحبت از داده‌کاوی می‌شود، باید مدل‌هایی را با استفاده از پایگاه های داده، موتورهای داده کاوی[1] و تکنیک‌های ارزیابی الگو بسازیم.

- **قابلیت یادگیری:** یادگیری ماشین از تکنیک‌های مشابه داده‌کاوی استفاده می‌کند، اما روش اول خودکار است. این بدان معنی است که یادگیری ماشین به‌طور خودکار یاد می‌گیرد، سازگار می‌شود و تغییر می‌کند. در نتیجه، هنگام پیش‌بینی دقیق‌تر از داده‌کاوی است. در مقابل، داده‌کاوی نیازمند تجزیه و تحلیل انسانی است و آن را به روشی دستی تبدیل می‌کند.

- **عامل انسانی:** در این‌جا یک تفاوت نسبتا قابل توجه وجو دارد. داده‌کاوی بر مداخله انسان متکی است و در نهایت برای استفاده توسط افراد ایجاد می‌شود. در حالی که، دلیل اصلی وجود یادگیری ماشین این است که می‌تواند خودش را آموزش دهد و به تاثیر یا اعمال انسان وابسته نباشد (ارتباط انسان با یادگیری ماشین، تقریبا محدود به تنظیم الگوریتم‌های اولیه است).

# آمار

آمار و یادگیری ماشین ارتباط بسیاری نزدیکی با یکدیگر دارند. با این حال، اهدافی که برای رسیدن به آن‌ها تلاش می کنند بسیار متفاوت است. هدف آمار استنباط در مورد یک جامعه بر اساس نمونه است. در مقابل، یادگیری ماشین برای پیش‌بینی‌های تکرارپذیر با یافتن الگوهای درون داده‌ها استفاده می‌شود. علاوه بر این، یادگیری ماشین به مقادیر زیادی داده نیاز دارد تا بتواند پیش‌بینی‌های دقیقی انجام داد. مدل‌ها با استفاده از داده‌های آموزشی ساخته می‌شوند، با استفاده از مجموعه داده اعتبارسنجی تنظیم دقیق[2] (ریزتنظیم) می‌شوند و با مجموعه داده آزمایشی ارزیابی می‌شوند. همه این مراحل به ماشین کمک می‌کند تا یادگیری بدست آورد. در مقابل، آمار شامل چندین زیرمجموعه داده نمی‌شود، چراکه سعی در پیش‌بینی ندارد. هدف مدل‌سازی در این شرایط، نمایش رابطه بین داده‌ها و متغیر نتیجه است.

---

[1] data mining engines

موتور داده کاوی قلب واقعی معماری داده کاوی است. این شامل ابزارها و نرم‌افزارهایی است که برای بدست آوردن بینش و دانش از داده‌های بدست آمده از منابع داده و ذخیره شده در انبارهای داده استفاده می‌شود.

[2] fine tuned

## ابزار و محیط کاری[1] یادگیری ماشین

محیط کاری یادگیری ماشین یک رابط، کتابخانه یا ابزاری است که به توسعه‌دهندگان و دانشمندان اجازه می‌دهد مدل‌های یادگیری ماشین را آسان‌تر و سریع‌تر بسازند و بکار گیرند؛ بدون اینکه وارد الگوریتم‌های زیربنایی شوند. آن‌ها روشی واضح و مختصر برای تعریف مدل‌های یادگیری ماشین با استفاده از مجموعه‌ای از اجزای از پیش ساخته شده و بهینه‌سازی‌شده ارائه می‌کند. برخی از ویژگی‌های کلیدی یک محیط کاری خوب یادگیری ماشین عبارتند از:

- برای عملکرد بهینه شده است.
- توسعه‌دهنده پسند است.
- به راحتی قابل درک و کدنویسی است.
- ارائه موازی‌سازی برای توزیع فرآیند محاسبات

به طور کلی، یک محیط کاری کارآمد یادگیری ماشین، پیچیدگی یادگیری ماشین را کاهش می‌دهد و آن را برای توسعه‌دهندگان بیشتر در دسترس قرار می دهد.

ما در چند سال گذشته، شاهد انفجاری در ابزارها و پلتفرم‌های توسعه دهنده مرتبط با یادگیری ماشین و هوش مصنوعی بوده‌ایم. از کتابخانه‌ها گرفته تا محیط‌های کاری و مدل‌های از پیش آموزش‌دیده. از همین‌رو، توسعه‌دهندگان انتخاب‌های زیادی برای القاء هوش مصنوعی به برنامه‌های خود دارند. هدف از این بخش معرفی محیط‌های کاری مختلف یادگیری ماشین به توسعه‌دهندگان با تاکید بر ویژگی‌های منحصربه‌فرد آن‌ها است.

## نحوه انتخاب محیط کاری مناسب یادگیری ماشین

چندین محیط کاری یادگیری ماشین برای ساده‌سازی توسعه و استقرار برنامه‌های کاربردی یادگیری ماشین پدید آمده است. با این حال، توسعه‌دهندگان باید در انتخاب محیط کاری مناسب، انتخاب‌های سختی داشته باشند. چرا که، برخی ممکن است بخواهند هنگام آموزش یک الگوریتم یادگیری ماشین بر سهولت استفاده تمرکز کنند، حال آن که برخی دیگر ممکن است بهینه‌سازی ابرپارامترها و استقرار تولید را در اولویت قرار دهند. از این‌رو، چندین نکته کلیدی وجود دارد که باید هنگام انتخاب محیط کاری یادگیری ماشین برای پروژه خود در نظر بگیرید:

- **ارزیابی نیازهای شما.** هنگامی که جستجوی خود را برای یافتن بهترین محیط‌کاری یادگیری ماشین شروع می کنید، این سه سوال را بپرسید:

---

۱. آیا این محیط کاری برای یادگیری عمیق یا الگوریتم‌های یادگیری ماشین سنتی استفاده می‌شود؟

۲. زبان برنامه‌نویسی ترجیحی برای توسعه مدل‌های یادگیری ماشین چیست؟

۳. چه سخت‌افزار، نرم‌افزار و خدمات ابری برای مقیاس‌بندی استفاده می‌شود؟

پایتون و آر زبان‌هایی هستند که به طور گسترده در یادگیری ماشین استفاده می‌شوند، با این حال، زبان‌های دیگری همانند جولیا، سی، جاوا و اسکالا نیز در دسترس هستند. اکثر برنامه‌های کاربردی یادگیری ماشین امروزه به زبان پایتون نوشته می‌شوند و در حال انتقال از آر هستند. زیرا آر توسط متخصصین آمار طراحی شده است و کار با آن تا حدودی ناخوشایند است. پایتون زبان برنامه‌نویسی مدرن‌تری است، نحو ساده و مختصر ارائه می‌دهد و استفاده از آن آسان‌تر است.

■ **بهینه‌سازی ابرپارامترها.** یکی دیگر از ملاحظات کلیدی هنگام انتخاب محیط کاری یادگیری ماشین، بهینه‌سازی ابرپارامترها است. هر الگوریتم یادگیری ماشین رویکرد متفاوتی برای تجزیه و تحلیل داده‌های آموزشی و استفاده از آنچه می‌آموزد در نمونه‌های جدید دارد. الگوریتم‌ها دارای ابرپارامترهایی هستند که می‌توانند آن‌ها را به‌عنوان داشبوردی باکلیدها و شماره‌گیری‌هایی در نظر بگیرید که نحوه‌ی عملکرد الگوریتم را کنترل می‌کنند. آن‌ها وزن متغیرهایی که باید در نظر گرفته شوند را تنظیم می‌کنند، تعیین می‌کنند که چه مقدار باید در نظر گرفته شود و سایر تنظیمات را در الگوریتم انجام می‌دهند. هنگام انتخاب یک محیط کاری یادگیری ماشین، مهم است که در نظر بگیرید که آیا این تنظیم باید خودکار باشد یا دستی.

■ **مقیاس‌بندی توسعه و آموزش.** در مرحله *آموزش* یک الگوریتم، مقیاس‌پذیری، مقدار داده‌ی قابل تحلیل و سرعت تجزیه و تحلیل است. عملکرد را می‌توان از طریق الگوریتم‌های توزیع‌شده و پردازش از طریق استفاده از واحدهای پردازش گرافیکی بهبود بخشید. در مرحله توسعه یک پروژه یادگیری ماشین، مقیاس‌پذیری به تعداد کاربران یا برنامه‌های همزمانی که می‌توانند به‌طور همزمان به مدل دسترسی داشته باشند، مربوط می‌شود. از آنجا که در مرحله آموزش و توسعه نیازمندی‌های متفاوتی وجود دارد، سازمان‌ها تمایل دارند مدل‌هایی را در یک نوع محیط توسعه دهند (مثلا محیط کاری یادگیری ماشین مبتنی‌بر پایتون که در فضای ابری اجرا می‌شود). از این‌رو، هنگام انتخاب یک محیط کاری، مهم است در نظر بگیرید که آیا از هر دو نوع مقیاس‌پذیری پشتیبانی می‌کند یا خیر.

# محبوب‌ترین ابزارهای یادگیری ماشین

در این بخش نگاهی به برخی از محبوب‌ترین ابزارهای یادگیری ماشین که امروزه استفاده می‌شود، می‌اندازیم.

## SciKit-Learn

Scikit-learn یکی از قدیمی‌ترین محیط‌های کاری یادگیری ماشین است که توسط دیوید کورناپو به عنوان پروژه تابستانی کد گوگل در سال ۲۰۰۷

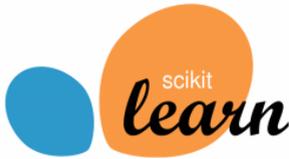

توسعه یافته است. به عنوان یک کتابخانه پایتون در دسترس است و از الگوریتم‌های یادگیری بانظارت و بدون‌نظارت پشتیبانی می‌کند. Scikit-Learn بهترین محیط‌کاری برای توسعه‌دهندگان پایتون برای یادگیری مبانی یادگیری ماشین است. این جعبه‌ابزار، پیاده‌سازی الگوریتم‌های رایج همانند رگرسیون خطی، رگرسیون لجستیک، کا_نزدیکترین همسایه، ماشین بردار پشتیبان، جنگل تصادفی و درخت تصمیم را آسان می‌کند. جدا از یادگیری بانظارت، Scikit-learn را می‌توان برای یادگیری بدون نظارت استفاده کرد و از الگوریتم‌هایی از خوشه‌بندی، تحلیل مؤلفه‌های اصلی و غیره پشتیبانی می‌کند. از آنجایی که Scikit-learn فقط با تکنیک‌های یادگیری ماشین سنتی که از یادگیری عمیق برای آموزش استفاده نمی‌کنند سر و کار دارد، نیازی به GPU ندارد. توسعه دهندگان پایتون می‌توانند با نصب بسته به سرعت با Scikit-learn شروع کنند. حتی آن دسته از توسعه‌دهندگانی که از TensorFlow، Keras و یا PyTorch برای آموزش استفاده می‌کنند، Scikit-learn را برای توابع کمکی همانند پیش‌پردازش داده، رمزگذاری، اعتبارسنجی متقابل و تنظیم ابرپارامترها ترجیح می‌دهند.

## ویژگی‌ها

- مدل‌ها و الگوریتم‌هایی را برای دسته‌بندی، رگرسیون، خوشه‌بندی، کاهش ابعاد، انتخاب مدل و پیش‌پردازش ارائه می‌دهد.
- به داده‌کاوی و تجزیه و تحلیل داده کمک می‌کند.

## PyTorch

PyTorch یک محیط‌کاری یادگیری ماشین مبتنی‌بر Torch است که برای طراحی شبکه عصبی ایده‌آل است. PyTorch توسط آزمایشگاه تحقیقاتی

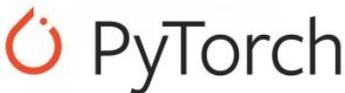

هوش مصنوعی فیس‌بوک توسعه یافته و در ژانویه ۲۰۱۶ به عنوان یک کتابخانه رایگان و منبع‌باز منتشر شد و عمدتا در بینایی رایانه، یادگیری عمیق و برنامه‌های پردازش زبان

طبیعی استفاده می‌شود و از توسعه نرم‌افزار مبتنی‌بر ابر پشتیبانی می‌کند. پیاده‌سازی یک شبکه عصبی در PyTorch نسبت به سایر محیط‌ها ساده‌تر و شهودی است. با پشتیبانی از CPU و GPU، شبکه‌های عصبی عمیق پیچیده را می‌توان با مجموعه داده‌های بزرگ آموزش داد.

## ویژگی‌ها

- الگوریتم‌های بهینه‌سازی متنوعی برای ساخت شبکه‌های عصبی ارائه می‌دهد.
- PyTorch را می‌توان در پلتفرم‌های ابری استفاده کرد.

### TensorFlow

TensorFlow یکی از محبوب‌ترین محیط‌های کاری یادگیری ماشین و یادگیری عمیق است که توسط توسعه‌دهندگان و محققان استفاده می‌شود. TensorFlow در ابتدا در سال ۲۰۰۷

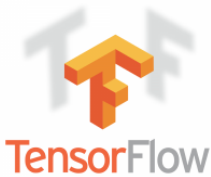

توسط تیم Google Brain راه‌اندازی شد و می‌تواند بر روی CPU و تسریع‌کننده‌های تخصصی هوش مصنوعی، از جمله GPU و TPU اجرا شود. TensorFlow در لینوکس ۶۴ بیتی، macOS، ویندوز و پلتفرم‌های محاسباتی موبایل، از جمله اندروید و iOS در دسترس است. مدل‌های آموزش

دیده در TensorFlow را می‌توان بر روی دسکتاپ، مرورگرها و حتی میکروکنترلرها مستقر[1] کرد. این پشتیبانی گسترده، TensorFlow را منحصر به فرد و آماده تولید می‌کند. چه در حال کار با مسائل بینایی رایانه، پردازش زبان طبیعی یا مدل‌های سری زمانی باشید، TensorFlow یک پلتفرم یادگیری ماشین بالغ و قوی با قابلیت‌های زیاد است.

## ویژگی‌ها

- **استقرار در چندین پلتفرم.** می‌توان TensorFlow را بر روی دسکتاپ، مرورگرها و حتی میکروکنترلرها مستقر کرد.
- **آموزش توزیع‌شده.** TensorFlow پشتیبانی قوی برای آموزش توزیع‌شده در CPU و GPU ارائه می‌دهد.
- **آموزش شبکه عصبی موازی.** TensorFlow خطوط لوله‌ای را ارائه می‌دهد که به شما امکان می‌دهد چندین شبکه عصبی و چندین GPU را به صورت موازی آموزش دهید.

## معایب

- یادگیری آن دشوار است.
- درک برخی از پیام‌های خطا در TensorFlow می‌تواند بسیار دشوار باشد.

---

[1] deployed

## Keras

Keras یک رابط برنامه‌نویسی است که دانشمندان داده را قادر می‌سازد به‌راحتی به پلتفرم یادگیری ماشینی TensorFlow دسترسی داشته باشند و از آن استفاده کنند. این یک رابط

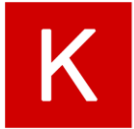

برنامه‌نویسی برنامه کاربردی[1] (API) و محیط کاری یادگیری عمیق منبع‌باز است که در پایتون نوشته شده است که برروی TensorFlow

اجرا می‌شود و اکنون در آن پلتفرم ادغام شده است. Keras قبلا از چندین پشتگاه[2] پشتیبانی می‌کرد اما با شروع نسخه ۲/۴/۰ در ژوئن ۲۰۲۰ به‌طور انحصاری با TensorFlow مرتبط شده است. Keras به‌عنوان یک API سطح بالا، برای انجام آزمایش‌های آسان و سریع طراحی شده است که نسبت به سایر گزینه‌های یادگیری عمیق نیاز به کدنویسی کم‌تری دارد. هدف تسریع اجرای مدل‌های یادگیری ماشین، به ویژه، شبکه‌های عصبی عمیق، از طریق یک فرآیند توسعه با "سرعت تکرار بالا"[3] است. مدل‌های Keras می‌توانند بر روی CPU یا GPU اجرا شوند و در چندین پلتفرم از جمله مرورگرهای وب و دستگاه‌های تلفن همراه Android و iOS مستقر شوند. Keras کندتر از TensorFlow و PyTorch است اما معماری ساده‌ای دارد و خواناتر، مختصرتر، کاربر پسند و قابل‌توسعه است. Keras بیشتر برای مجموعه داده‌های کوچک مناسب است و به دلیل طراحی ساده و قابل درک آن برای مبتدیان توصیه می‌شود.

## ویژگی‌ها

- تمرکز برروی تجربه کاربری
- تولید آسان مدل‌ها
- پشتیبانی از شبکه‌های پیچشی (کانولوشنی)
- پشتیبانی از شبکه‌های بازگشتی
- یک محیط کاری مبتنی‌بر پایتون که اشکال‌زدایی و کاوش را آسان می‌کند.
- توسعه‌یافته با تمرکز بر روی امکان آزمایش سریع
- می‌توان از آن برای نمونه‌سازی[4] آسان و سریع استفاده کرد.
- از ترکیب دو شبکه پشتیبانی می‌کند.
- کاربر پسند و قابل توسعه

---

**Colab**

Colaboratory یا به اختصار Colab یک محصول تحقیقاتی گوگل (سرویس ابری) است که به توسعه‌دهندگان اجازه می‌دهد کدهای پایتون را از طریق مرورگر خود بنویسند و اجرا کنند.

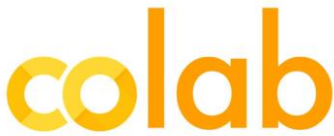

Google Colab یک ابزار عالی برای کارهای یادگیری عمیق است و به توسعه مدل‌ها با استفاده از چندین کتابخانه مانند Keras، Pytorch، OpenCv، Tensorflow و غیره کمک می‌کند.

Colab یک نوت‌بوک مبتنی‌بر Jupyter است که نیازی به نصب ندارد و دارای نسخه رایگان عالی است که دسترسی رایگان به منابع محاسباتی Google مانند GPU و TPU را می‌دهد.

**چرا باید از Colab استفاده کنیم؟**

Colab برای همه چیز ایده‌آل است، از بهبود مهارت‌های کدنویسی پایتون تا کار با کتابخانه‌های یادگیری عمیق، مانند PyTorch، Keras، TensorFlow و OpenCV. می‌توانید نوت‌بوک‌ها را در Colab ایجاد، بارگذاری، ذخیره و به اشتراک بگذارید، Google Drive خود را نصب کنید و از هر چیزی که در آنجا ذخیره کرده‌اید استفاده کنید، نوت‌بوک‌ها را مستقیما از GitHub بارگذاری کنید، فایل‌های Kaggle را بارگذاری کنید، نوت‌بوک‌های خود را باگیری کنید و تقریبا هر کار دیگری را که ممکن است بخواهید انجام دهید را انجام دهید.

از دیگر ویژگی‌های عالی Google Colab، ویژگی همکاری[۱] است. اگر با چند برنامه‌نویس روی یک پروژه کار می‌کنید، استفاده از نوت‌بوک Google Colab عالی است. درست همانند همکاری در یک سند Google Docs، می‌توانید با استفاده از یک نوت‌بوک Colab با چندین برنامه‌نویس کدنویسی کنید. علاوه بر این، شما همچنین می‌توانید کارهای تکمیل شده خود را با توسعه‌دهندگان دیگر به اشتراک بگذارید.

به‌طور خلاصه می‌توان دلایل مختلف استفاده از Colab را به‌صورت زیر فهرست کرد:

- کتابخانه‌های از پیش نصب‌شده
- ذخیره‌شده در ابر
- همکاری
- استفاده از GPU و TPU رایگان

---

[۱] collaboration

با این حال، دو سناریو وجود دارد که شما باید از Jupyter Notebook در ماشین خود استفاده کنید:

۱. اگر به حریم خصوصی اهمیت می‌دهید و می‌خواهید کد خود را مخفی نگه دارید، از Google Colab دوری کنید.

۲. اگر یک ماشین فوق‌العاده قدرتمند با دسترسی به GPU و TPU دارید.

### راه‌اندازی Google Colab

فرآیند راه‌اندازی Colab نسبتا آسان است و می‌تواند با مراحل زیر در هر نوع دستگاهی تکمیل شود:

**۱. از صفحه Google Colab دیدن کنید:**

http://colab.research.google.com

بارگذاری تارنمای فوق شما را به صفحه خوش‌آمدگویی Google Colaboratory هدایت می‌کند.

**۲. روی دکمه ورود به سیستم (Sign in) در بالا سمت راست کلیک کنید:**

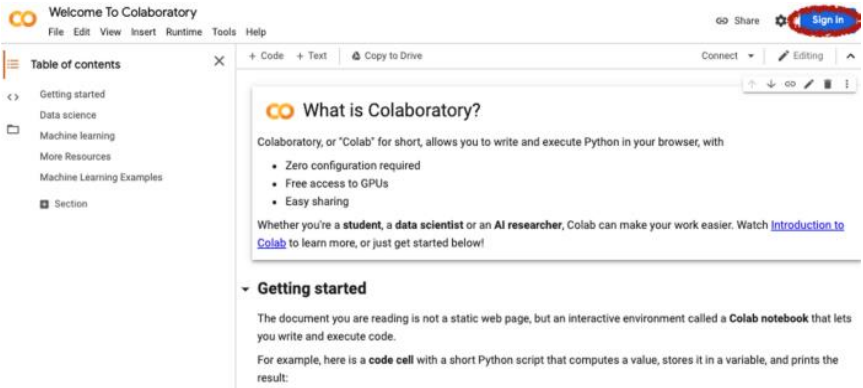

**۳. با حساب GMail خود وارد شوید. اگر حساب GMail ندارید یکی بسازید:**

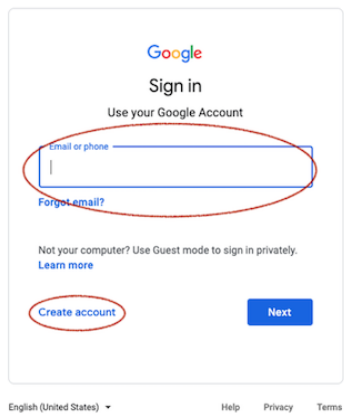

۴. به محض تکمیل فرآیند ورود به سیستم، آماده استفاده از **Google Colab** هستید.

۵. با کلیک بر روی **File< New notebook** به‌راحتی می توانید یک نوت‌بوک **Colab** جدید در این صفحه ایجاد کنید.

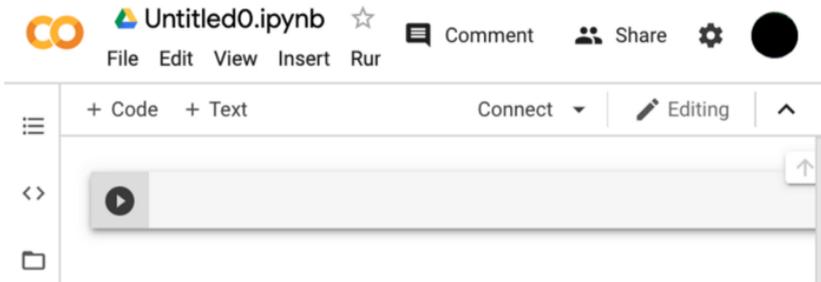

# محدودیت‌های یادگیری ماشین

یادگیری ماشین در دهه گذشته جهان را به شکلی که می‌شناسیم متحول کرده است. انفجار داده‌ها منجر به جمع‌آوری حجم عظیمی از داده‌ها به ویژه توسط شرکت‌های بزرگی مانند فیس‌بوک و گوگل شده است. این حجم از داده‌ها، همراه با توسعه سریع قدرت پردازنده‌ها و موازی‌سازی رایانه‌ها، امکان بدست آوردن و مطالعه مقادیر عظیمی از داده‌ها را با سهولت نسبی فراهم کرده است. به راحتی می‌توان درک کرد که چرا یادگیری ماشین چنین تاثیر عمیقی بر جهان داشته است، چیزی که کم‌تر مشخص است این است که یادگیری ماشین چه محدودیت‌هایی دارد؟ در ادامه این بخش به تشریح این محدودیت‌ها می‌پردازیم.

## اخلاق

دیوید بروکس برای اولین بار در مقاله خود در نیویورک تایمز در سال ۲۰۱۳ با عنوان "فلسفه داده‌ها" به اصطلاح جدیدی به نام "**داده‌باوری**[1]" یا داده‌گرایی اشاره کرد. داده‌باوری یک سیستم اخلاقی است که توسط مورخ مشهور، یووال نوح هراری، به شدت مورد بررسی و محبوبیت قرار گرفته است. به عقیده هراری، داده‌باوری به مرحله جدیدی از تمدن که وارد آن می‌شویم اشاره دارد و در آن ما بیشتر به الگوریتم‌ها و داده‌ها اعتماد داریم تا قضاوت و منطق خودمان. از منظر داده‌باوری، به نظر می‌رسد که ما هم در مدیریت داده‌هایی که جهان تولید می‌کند و هم در درک ذهن و بدن خود دچار مشکل شده‌ایم. در عوض باید تصمیمات خود را به الگوریتم‌هایی بسپاریم که ما را بهتر می‌شناسند. آنها تصمیم می‌گیرند با چه کسی قرار ملاقات بگذاریم، چه بخوریم و کجا برویم. ایده اعتماد به داده‌ها و الگوریتم‌ها بیشتر از قضاوت خودمان، جوانب مثبت و منفی

---

[1] dataism

خود را دارد. بدیهی است که ما از این الگوریتم‌ها سود می‌بریم، در غیر این صورت، در وهله اول از آن‌ها استفاده نمی‌کنیم. این الگوریتم‌ها به ما اجازه می‌دهند تا با قضاوت‌های آگاهانه با استفاده از داده‌های موجود، فرآیندها را خودکار کنیم. با این حال، گاهی اوقات این به معنای جایگزینی شغل شخصی با یک الگوریتم است که با پیامدهای اخلاقی همراه است. علاوه بر این، اگر مشکلی پیش بیاید، چه کسی را مقصر می‌دانیم؟ رایج‌ترین مورد بحث شده در حال حاضر اتومبیل‌های خودران است، اتومبیل در صورت برخورد مرگبار چگونه باید واکنش نشان دهد؟ آیا در آینده هنگام خرید اتومبیل خودران باید انتخاب کنیم که از چارچوب اخلاقی خاصی پیروی کند؟ اگر اتومبیل خودران من کسی را در جاده به قتل برساند، مقصرکیست؟ با این حال، واضح است که یادگیری ماشین نمی‌تواند چیزی در مورد ارزش‌های هنجاری‌که باید بپذیریم، به ما بگوید؛ یعنی در یک موقعیت خاص چگونه باید در جهان عمل کنیم.

## داده

داده. این بارزترین محدودیت است. اگر یک مدل را ضعیف تغذیه کنید، نتایج ضعیفی را هم به شما خواهد داد. این می‌تواند خود را به دو صورت نشان دهد: کمبود داده و نبود داده‌های خوب.

### کمبود داده

بسیاری از الگوریتم‌های یادگیری ماشین قبل از اینکه شروع به ارائه نتایج مفید کنند، به مقادیر زیادی داده نیاز دارند. یک مثال خوب از این شبکه عصبی است. شبکه‌های عصبی ماشین‌های داده‌خواری[1] هستند که به مقدار زیادی از داده‌های آموزشی نیاز دارند. هرچه معماری بزرگ‌تر باشد، داده‌های بیشتری برای تولید نتایج قابل قبول مورد نیاز است. افزایش داده‌ها در بیشتر مواقع راه حل ترجیحی است.

### کمبود داده‌های خوب

با وجود ظاهر، این با مورد بالا یکی نیست. بیایید تصور کنیم که فکر می‌کنید می‌توانید با ایجاد ده هزار نقطه داده جعلی برای قرار دادن در شبکه عصبی تقلب کنید. وقتی آن را به شبکه تغذیه می‌کنید چه اتفاقی می‌افتد؟ با این داده‌ها آموزش می‌بیند و هنگامی که آن را روی یک مجموعه داده دیده نشده (مجموعه آزمایشی) آزمایش می‌کنید، عملکرد خوبی نخواهد داشت. شما داده‌ها را داشتید اماکیفیت داده‌ها در حد بالایی نبود. همان‌طورکه نداشتن ویژگی‌های خوب می‌تواند باعث عملکرد ضعیف الگوریتم شما شود، نداشتن داده‌های خوب نیز می‌تواند قابلیت‌های مدل شما را محدود کند. هیچ شرکتی قرار نیست مدل یادگیری ماشینی را اجراکند که عملکردی بدتر

---

[1] data-eating

از خطای سطح انسانی داشته باشد. به‌طور مشابه، استفاده از مدلی که بر روی مجموعه‌ای از داده‌ها در یک وضعیت خاص آموزش داده شده است، ممکن است لزوما برای وضعیت دوم نیز کاربرد نداشته باشد. بهترین مثالی که می‌توان در این خصوص بیان کرد مربوط به پیش‌بینی سرطان سینه است. پایگاه‌های اطلاعاتی ماموگرافی تصاویر زیادی در خود دارند، اما آن‌ها از یک مشکل رنج می‌برند که در سال‌های اخیر باعث مشکلات مهمی شده است، تقریبا تمام عکس‌های اشعه ایکس از زنان سفیدپوست است. این ممکن است چندان مهم به نظر نرسد، اما در واقع زنان سیاه‌پوست به دلیل طیف گسترده‌ای از عوامل که ممکن است شامل تفاوت در تشخیص و دسترسی به مراقبت‌های بهداشتی باشد، ۴۲ درصد بیشتر در معرض خطر مرگ ناشی از سرطان سینه هستند*. بنابراین، آموزش یک الگوریتم در درجه اول بر روی زنان سفیدپوست تأثیر نامطلوبی بر زنان سیاه‌پوست در این مورد دارد. آنچه در این مورد خاص مورد نیاز است، تعداد بیشتری عکس‌های اشعه ایکس از بیماران سیاه‌پوست در پایگاه داده آموزشی، ویژگی‌های بیشتر مرتبط با علت این احتمال افزایش ۴۲ درصدی است تا الگوریتم طبقه‌بندی عملکرد بهتری داشته باشد.

# قابلیت توضیح

**توضیح‌پذیری**[1] یکی از مشکلات اصلی یادگیری ماشین است. یک شرکت مشاوره هوش مصنوعی که سعی دارد به شرکتی که فقط از روش‌های آماری سنتی استفاده می‌کند پیشنهادهایی را ارائه دهد، اگر مدل قابل توضیح نباشد و نتوانید مشتری خود را متقاعد کنید چگونه متوجه شده‌اید که این الگوریتم به این تصمیم رسیده است، چقدر احتمال دارد به شما و تخصص شما اعتماد کند؟ این مدل‌ها در این صورت می‌توانند چنین بی‌قدرت جلوه کنند، مگر اینکه قابل توضیح باشند. به همین دلیل، توضیح‌پذیری ویژگی بسیار بسیار مهمی است که روش‌های یادگیری ماشین باید به دنبال آن باشند تا در عمل به کار گرفته شوند (تحقیقات زیادی برای نزدیک شدن به توضیح‌پذیری صورت گرفته است).

## توضیح‌پذیری در مقابل تفسیرپذیری

از تشخیص پزشکی گرفته تا سناریوهای تجاری، مدل‌های یادگیری ماشین برای تصمیم‌گیری‌های مهم استفاده می‌شوند. برای اعتماد به سیستم‌هایی که توسط این مدل‌ها کار می‌کنند، باید بدانیم این مدل‌ها چگونه پیش‌بینی می‌کنند. به همین دلیل است که تفاوت بین یک مدل قابل تفسیر و

---

قابل توضیح مهم است. نحوه درک ما از مدل‌ها و درجه‌ای که واقعا می‌توانیم درک کنیم بستگی به این دارد که آیا آن‌ها قابل تفسیر هستند یا قابل توضیح. در زمینه یادگیری ماشین و هوش مصنوعی، **توضیح‌پذیری و تفسیرپذیری**[1] اغلب به جای یکدیگر استفاده می‌شوند. در حالی که آن‌ها بسیار نزدیک بهم مرتبط هستند، بهتر است که تفاوت آن‌ها را درک کنیم. به این دلیل که، ببینیم وقتی شروع به کندوکاو عمیق‌تر در سیستم‌های یادگیری ماشین می‌کنیم چقدر می‌توانند پیچیده شوند.

توضیح‌پذیری در یادگیری ماشین به این معنا است که شما بتوانید آنچه را که در مدل اتفاق می‌افتد از ورودی تا خروجی توضیح دهید که این امر موجب شفافیت در مدل می‌شود. به عبارت دیگر، درک اینکه چه ویژگی‌هایی در پیش‌بینی مدل نقش دارند و چرا این کار را انجام می‌دهند، مفهوم توضیح‌پذیری است.

در مقابل، تفسیرپذیری به درجه‌ای که یک انسان می‌تواند علت یک تصمیم را درک کند یا درجه‌ای که یک انسان می‌تواند به طور مداوم نتایج مدل یادگیری ماشین را پیش‌بینی کند، تعریف می‌شود. به بیان دیگر، درجه‌ای است که شما می‌توانید با توجه به تغییر در پارامترهای ورودی یا الگوریتمی، پیش‌بینی کنید که چه اتفاقی می‌افتد. به عنوان مثال، یک خودرو برای حرکت به سوخت نیاز دارد، یعنی این سوخت است که باعث حرکت موتورها می‌شود: *قابل تفسیر.* درک چگونگی و چرایی مصرف و استفاده موتور از سوخت: *قابل توضیح. به‌طور خلاصه، تفسیرپذیری به معنای توانایی تشخیص مکانیسم است بدون اینکه لزوما دلیل آن را بدانیم. توضیح‌پذیری این است که بتوانیم به‌طور کامل آنچه را که در حال وقوع است توضیح دهیم.*

به طور خلاصه، یک مدل قابل تفسیر می‌تواند بدون هیچ‌گونه کمک یا تکنیک دیگری توسط انسان درک شود. به عبارت دیگر، ما می‌توانیم تنها با نگاه کردن به پارامترهای مدل بفهمیم که این مدل‌ها چگونه پیش‌بینی می‌کنند. ***می‌توان گفت که یک مدل قابل تفسیر، توضیح خود را ارائه می‌دهد. در مقابل، یک مدل قابل توضیح، توضیح خود را ارائه نمی‌دهد و پیچیدگی بیشتری دارد.*** هر چه یک مدل پیچیده‌تر باشد، میزان توضیح‌پذیری آن کمتر و از این‌رو برای انسان کم‌تر قابل درک می‌شود و برای درک چگونگی پیش‌بینی‌ها به تکنیک‌های بیشتری نیاز دارد.

نکته‌ای که باید در نظر داشت، گاها مشاهده می‌شود که تفسیرپذیری و توضیح‌پذیری در کنار یکدیگر قرار گرفته‌اند و براساس میزان پیچیدگی در توضیح‌پذیری مدل‌ها به دو دسته‌یِ مدل‌های جعبه سیاه و جعبه شفاف (جعبه شیشه‌ای یا جعبه سفید)، تقسیم‌بندی شده‌اند. مدلی با قابلیت توضیح‌پذیری مدل جعبه شفاف و در نقطه مقابلش مدل جعبه سیاه قرار دارد. به عبارت دیگر، مدل جعبه سیاه در نقطه‌ی مقابل مدل توضیح‌پذیر قرار می‌گیرد. از این منظر، مدل‌های جعبه

---

[1] interpretability

سیاه اغلب پیچیده هستند و درک عملکرد درونی آن‌ها دشوار است. از طرف دیگر، مدل‌های جعبه شفاف به اندازه‌ای ساده هستند که عملکرد آن‌ها را می‌توان به طور مستقیم توضیح داد.

## یادگیری ماشین تفسیرپذیر

می‌گوییم چیزی قابل تفسیر است که قابل فهم باشد. با در نظر گرفتن این موضوع، می‌گوییم یک مدل در صورتی قابل تفسیر است که به تنهایی بدون کمک تکنیک دیگری، توسط انسان قابل درک باشد. می‌توانیم به پارامترهای مدل یا خلاصه‌ای از مدل نگاه کنیم و دقیقا بفهمیم که چرا یک پیش‌بینی یا به عبارت دیگر یک تصمیم خاص توسط مدل گرفته شده است. نمونه‌هایی از مدل‌های قابل تفسیر شامل درختان تصمیم و رگرسیون خطی هستند. زیرا می‌توانیم مستقیما پارامترهای مدل را بررسی کنیم و استنباط کنیم که چگونه این مدل‌ها و ورودی‌های خود را به خروجی تبدیل می‌کنند. بنابراین، این مدل‌ها خود توضیحی هستند و نیازی به توضیح بیشتر ندارند. به طور خلاصه، تفسیرپذیری به معنای خود توضیحی است.

## یادگیری ماشین توضیح‌پذیر

یک مدل یادگیری ماشین را می‌توان به عنوان یک تابع که ویژگی‌های مدل ورودی‌ها و پیش‌بینی‌ها خروجی هستند. تابعی که توضیح آن برای انسان بسیار پیچیده است به عنوان جعبه سیاه خوانده می‌شود. به عبارت دیگر، ما به یک روش یا تکنیک اضافی نیاز داریم تا بتوانیم به جعبه سیاه نگاه کنیم و نحوه عملکرد مدل را درک کنیم. نمونه‌ای از چنین مدلی یک جنگل تصادفی است. به بیان ساده، یک جنگل تصادفی از درختان تصمیم بسیاری تشکیل شده است که در آن پیش‌بینی‌های همه درختان منفرد در هنگام پیش‌بینی نهایی در نظر گرفته می‌شود. برای درک نحوه عملکرد یک جنگل تصادفی، باید به طور همزمان بفهمیم که همه درختان جداگانه چگونه کار می‌کنند. حتی با تعداد کمی درخت، این امر امکان‌پذیر نخواهد بود.

میزان توضیح‌پذیری، مستقیما با طیف پیچیدگی‌های مدل جعبه سیاه مرتبط است. مدل‌های پیچیده‌تر کم‌تر قابل توضیح هستند (توضیح آن‌ها سخت‌تر است و برای رمزگشایی به کار بیشتری نیاز دارند). به عنوان مثال، وقتی شروع به بررسی الگوریتم‌هایی مانند شبکه‌های عصبی عمیق می‌کنیم، اوضاع خیلی پیچیده‌تر می‌شود. AlexNet، یک شبکه عصبی کانولوشنی است که برای تشخیص تصویر استفاده می‌شود، دارای ۶۲۳۷۸۳٤٤ پارامتر است. تنها با نگاه کردن به وزن پارامترهای این مدل، برای درک اینکه چگونه کار می‌کند، برای انسان ممکن نیست.

## چرا توضیح‌پذیری و تفسیرپذیری در یادگیری ماشین مهم است؟

از آنجایی که حوزه‌هایی مانند مراقبت‌های بهداشتی به دنبال استقرار هوش مصنوعی یا به‌طور دقیق‌تر سیستم‌های یادگیری عمیق هستند، جایی که پرسش‌های مربوط به شفافیت از اهمیت

ویژه‌ای برخوردار است، اگر نتوانیم به‌درستی تفسیرپذیری بهبود یافته و در نهایت توضیح‌پذیری را در الگوریتم‌های خود ارائه دهیم، به‌طور جدی تاثیر بالقوه هوش مصنوعی، کم قدرت جلوه خواهد داد. اما جدای از ملاحظات حرفه‌ای که باید انجام شود، این بحث نیز وجود دارد که بهبود تفسیرپذیری و توضیح‌پذیری حتی در سناریوهای تجاری ساده‌تر نیز مهم است. درک این‌که یک الگوریتم واقعا چگونه کار می‌کند، می‌تواند به همسویی بهتر فعالیت‌های دانشمندان داده و تحلیل‌گران و نیازهای کلیدی سازمان آن‌ها کمک کند.

توضیح‌پذیری می‌تواند درک جنبه‌های مختلف یک مدل را تسهیل کند و منجر به بینش‌هایی شود که می‌تواند توسط ذینفعان مختلف مورد استفاده قرار گیرد تا به نگرانی‌های اساسی زیر هنگام استقرار یک محصول یا تصمیم‌گیری‌هایی که بر اساس پیش‌بینی‌های خودکار ایجاد می‌شوند، کمک کنند:

- **صحت:** آیا مطمئن هستیم که همه و تنها متغیرهای مورد علاقه در تصمیم ما نقش داشته‌اند؟ آیا مطمئن هستیم که الگوها و همبستگی‌های نادرست از نتیجه ما حذف شده‌اند؟

- **نیرومندی:** در حضور داده‌های مفقودی یا نویزدار، آیا مطمئن هستیم که این مدل بد عمل نمی‌کند؟

- **سوگیری:** آیا ما از هرگونه سوگیری خاص داده که به طور ناعادلانه گروهی از افراد را جریمه می‌کند، آگاه هستیم و اگر بله، آیا می‌توانیم آن‌ها را شناسایی و اصلاح کنیم؟

- **بهبود:** از چه طریقی می‌توان مدل پیش‌گویانه را بهبود بخشید؟ داده‌های آموزشی اضافی یا فضای ویژگی‌های پیشرفته چه تاثیری بر مدل خواهند داشت؟ به عبارت دیگر، اگر بدانید چرا و چگونه مدل شما کار می‌کند، دقیقا می‌دانید چه چیزی را باید تنظیم و بهینه کنید.

- **قابلیت انتقال:** به چه طریقی می‌توان مدل پیش‌گویانه یک حوزه کاربردی را در حوزه کاربردی دیگر اعمال کرد؟ چه ویژگی‌هایی از داده‌ها و مدل‌ها باید برای این قابلیت انتقال تطبیق داده شوند؟

- **اعتماد:** در حوزه‌های پرخطر همانند مراقبت‌های بهداشتی یا مالی، اعتماد بسیار مهم است. قبل از اینکه بتوان راه‌حل‌های یادگیری ماشین را مورد استفاده قرار داد و به آن اعتماد کرد، همه ذینفعان باید به طور کامل بفهمند که مدل چه کاری انجام می‌دهد. اگر ادعا می‌کنید که مدل شما تصمیمات بهتری می‌گیرد و متوجه الگوهایی می‌شود که انسان‌ها نمی‌بینند، باید بتوانید با شواهدی از آن حمایت کنید. *متخصصان حوزه به‌طور طبیعی نسبت به هر فناوری که ادعا می‌کند بهتر از آن‌ها مساله را می‌فهمد، بدبین خواهند بود.*

- **انطباق:** قابلیت توضیح مدل برای دانشمندان و تصمیم‌گیرندگان بسیار مهم است تا از انطباق با سیاست‌های شرکت، استانداردهای صنعت و مقررات دولتی اطمینان حاصل شود. طبق ماده ۱۴ قانون حفاظت از داده اروپا (GDPR)، زمانی که یک شرکت از ابزارهای تصمیم‌گیری خودکار استفاده می‌کند، باید اطلاعات معناداری در مورد منطق مربوط و همچنین اهمیت و پیامدهای پیش‌بینی‌شده چنین پردازشی را ارائه دهد. مقررات مشابهی در سرتاسر جهان وضع شده است.

## افزودن پیچیدگی برای مقابله با پیچیدگی

اگرچه ما انسان‌ها در انجام بسیاری از وظایف شناختی از جمله تفکر انتقادی، خلاقیت، همدلی و ذهنیت برتر هستیم، اما در مدیریت پیچیدگی‌ها عالی نیستیم. روانشناسان دریافته‌اند که انسان‌ها تنها می‌توانند حدود ۷±۲ چیز را در حافظه کاری خود پیگیری کنند. اما ماشین‌ها (مثلا یک رایانه) می‌توانند میلیون‌ها و میلیاردها آیتم را (که فقط با اندازه RAM محدود می‌شود) پیگیری کنند. از آنجایی که مساله جعبه سیاه صرفا یک مساله پیچیدگی (بغرنج) است، می‌توانیم از تحلیل به کمک ماشین یا دیگر الگوریتم‌های یادگیری ماشین برای توضیح جعبه‌های سیاه استفاده کنیم.

با این حال، شما یک مرحله اضافی در فرآیند توسعه اضافه می‌کنید. در واقع، احتمالا چندین مرحله را اضافه می‌کنید. از این منظر، به نظر می‌رسد شما در حال تلاش برای مقابله با پیچیدگی با پیچیدگی بیشتر هستید و تا حدی این درست است. معنای این امر در عمل این است که اگر می‌خواهیم واقعا در مورد تفسیرپذیری و توضیح‌پذیری جدی باشیم، باید تغییر گسترده‌تری در شیوه‌ای که علم داده انجام می‌شود و اینکه مردم چگونه معتقدند باید انجام شود، ایجاد شود.

در نهایت، در حالی که تفسیر مدل‌های جعبه سیاه توسط مغز انسان دشوار است، همه آنها با کمک تحلیل‌ها و الگوریتم‌ها قابل توضیح هستند. تعداد فزاینده روش‌ها و چارچوب‌های یادگیری ماشین توسعه‌یافته در این خصوص، به ما این امکان را می‌دهد که به درون جعبه‌های سیاه نگاه کنیم و آن‌ها را به جعبه‌های شیشه‌ای تبدیل کنیم. از این‌رو می‌توان گفت که "مشکل جعبه سیاه" واقعا مشکلی نیست که نتوان آن را حل کرد و به این دلیل قدرت این مدل‌ها را زیر سوال ببریم. رهبران کسب‌وکار که به دلیل ماهیت جعبه سیاه مدل‌ها، استفاده از یادگیری ماشین را قربانی می‌کنند، اساسا از یک روش کارآمد و قابل اعتماد برای بهینه‌سازی تصمیمات تجاری خود برای چیزی که فقط یک مشکل قابل حلی است، چشم‌پوشی می‌کنند.

## تکنیک‌های مورد استفاده برای درک مدل‌های قابل توضیح

می‌توان به دو طریق به توضیح‌پذیری نزدیک شد:

۱. **کلی.** این توضیح کلی رفتار مدل است و تصویری بزرگ از مدل را به ما نشان می‌دهد و اینکه چگونه ویژگی‌های داده به‌طور جمعی بر نتیجه تاثیر می‌گذارد.

۲. **محلی.** این روش به ما می‌گوید که چگونه ویژگی‌ها به صورت جداگانه بر نتیجه تاثیر می‌گذارند.

هنگام نتیجه‌گیری با استفاده از این تکنیک‌ها باید یک سطح احتیاط را رعایت کرد. این به این دلیل است که این تکنیک‌ها فقط می‌توانند تقریبی از نحوه پیش‌بینی واقعی مدل را ارائه دهند. برای تایید هر نتیجه‌گیری، می‌توان از چندین تکنیک به صورت ترکیبی استفاده کرد یا می‌توان آنها را با استفاده از مصورسازی داده‌ها تایید کرد. *دانش دامنه نیز می‌تواند ابزار مهمی باشد، هر نتیجه‌ای که بر خلاف تجربه یا دانش قبلی باشد باید با جزئیات بیشتری تحلیل شود.*

## کدام رویکرد یادگیری ماشین؟

با وجود رویکردهای مختلف یادگیری (یادگیری بانظارت، یادگیری غیرنظارتی و یادگیری تقویتی) و الگوریتم‌های متفاوت، پرسش این‌جاست چگونه تصمیم بگیریم از کدام رویکرد برای یک مساله خاص استفاده کنیم؟

یک راهبرد این است که همه رویکردهایِ ممکن یادگیری ماشین را امتحان کنید و سپس بررسی کنید که کدام رویکرد بهترین نتایج را به همراه دارد. مشکل این روش این است که ممکن است مدت زمان زیادی طول بکشد. ده‌ها الگوریتم یادگیری ماشین وجود دارد و هر کدام زمان اجرای متفاوتی دارند. بسته به مجموعه داده‌ها، تکمیل برخی الگوریتم‌ها ممکن است ساعت‌ها یا حتی روزها طول بکشد. یکی دیگر از ریسک‌های انجام راهبرد "همه رویکردها را امتحان کنید" این است که ممکن است در نهایت از یک الگوریتم یادگیری ماشین برای یک نوع مساله استفاده کنید که واقعا برای آن الگوریتم خاص مناسب نیست. تشبیه آن همانند استفاده از چکش برای سفت کردن یک پیچ است. مسلماً چکش ابزار مفیدی است اما تنها زمانی که برای اهداف مورد نظر خود استفاده می‌شود. اگر می‌خواهید پیچ را سفت کنید، از پیچ‌گوشتی استفاده کنید نه چکش.

وقتی تصمیم گرفتید از چه نوع الگوریتم یادگیری ماشین استفاده کنید، ابتدا باید مساله را به طور کامل درک کنید و سپس تصمیم بگیرید که می‌خواهید به چه چیزی برسید. در اینجا یک چارچوب مفیدی ارائه شده است که می‌تواند برای انتخاب الگوریتم مناسب استفاده شود:

- **آیا می‌خواهید یک مجموعه داده بدون برچسب را به گروه‌هایی تقسیم کنید که هر گروه ویژگی‌های مشابهی داشته باشد (مثلا تقسیم‌بندی مشتری)؟** اگر بله، از یک الگوریتم خوشه‌بندی (یادگیری غیرنظارتی) همانند k-means، خوشه‌بندی سلسله مراتبی یا مدل‌های مخلوط گاوسی استفاده کنید.

▪ **آیا می‌خواهید یک مقدار پیوسته را با توجه به مجموعه‌ای از ویژگی‌ها (مثلا پیش‌بینی قیمت مسکن) پیش‌بینی کنید؟** اگر بله، از الگوریتم رگرسیون (یادگیری بانظارت) مانند رگرسیون خطی استفاده کنید.

▪ **آیا می‌خواهید کلاس‌های مجزا را پیش‌بینی کنیم؟ آیا مجموعه داده‌ای داریم که قبلا با کلاس‌ها برچسب‌گذاری شده است؟** اگر برای هر دو سوال جواب بله است، از یک الگوریتم دسته‌بندی (یادگیری بانظارت) مانند بیزساده، کا‌ـ‌نزدیک‌ترین همسایه، شبکه‌های عصبی یا ماشین‌های بردار پشتیبان استفاده کنید.

▪ **آیا سعی دارید تعداد زیادی از ویژگی‌ها را به تعداد کمتری ویژگی کاهش دهید؟** از الگوریتم‌های کاهش ابعاد استفاده کنید، همانند تحلیل مولفه اصلی.

▪ **آیا به الگوریتمی نیاز دارید که به محیط خود واکنش نشان دهد و به‌طور مداوم از تجربه، روشی که انسان‌ها انجام می‌دهند یاد بگیرد؟** اگر بله، از رویکرد یادگیری تقویتی استفاده کنید.

برای هر یک از سوالات بالا، می‌توانید سوالات بعدی را بپرسید تا الگوریتم مناسب را برای استفاده بررسی کنید. مثلا:

● **آیا به الگوریتمی نیاز داریم که بتوان آن را به سرعت ساخت، آموزش داد و آزمایش کرد؟**

● **آیا به مدلی نیاز داریم که بتواند سریع پیش‌بینی کند؟**

● **مدل چقدر باید دقیق باشد؟**

● **آیا تعداد ویژگی‌ها از تعداد نمونه‌ها بیشتر است؟**

● **آیا ما به مدلی نیاز داریم که تفسیر آن آسان باشد؟**

● **چه معیارهای ارزیابی برای برآوردن نیازهای کسب‌وکار مهم است؟**

● **چه مقدار پیش‌پردازش داده را می‌خواهیم انجام دهیم؟**

## خلاصه فصل

▪ یادگیری ماشین روشی است که با یادگیری تجربی از طریق روش‌های محاسباتی عملکرد سیستم را بهبود می‌بخشد.

▪ در سیستم‌های رایانه‌ای، تجربه در قالب داده‌ها وجود دارند و وظیفه اصلی یادگیری ماشین توسعه الگوریتم‌های یادگیری است که از داده‌ها مدل می‌سازد.

▪ در رویکرد نظارتی، مجموعه‌ای از نمونه‌های آموزشی با پاسخ‌های صحیح به الگوریتم تغذیه می‌شود و الگوریتم سعی می‌کند براساس این داده‌ها و پاسخ‌های صحیح تابعی را بیاموزد تا بتواند مقادیر هدف را برای نمونه‌های جدید به درستی پیش‌بینی کند.

- در یادگیری نظارتی اگر داده‌های مساله جهت یادگیری به‌صورت گسسته باشند، این مساله دسته‌بندی، و اگر مقادیر داده‌ها به‌صورت پیوسته باشند به آن رگرسیون گویند.

- در رویکرد غیرنظارتی، پاسخ‌های درست به الگوریتم ارائه نمی‌شود، اما در عوض الگوریتم سعی می‌کند شباهت‌های بین ورودی‌ها را مشخص کند تا ورودی‌هایی که دارای ویژگی مشترک هستند در کنار هم گروه‌بندی شوند.

- در یادگیری تقویتی، یک عامل سعی می‌کند یک مساله را با آزمایش و خطا از طریق تعامل با محیطی که پویایی آن برای عامل ناشناخته است، حل کند.

- یادگیری انتقالی بر استخراج داده‌ها از یک دامنه مشابه متمرکز است تا توانایی یادگیری را افزایش یا تعداد نمونه‌های برچسب‌دار مورد نیاز در یک دامنه هدف را کاهش دهد.

- یادگیری چندوظیفه‌ای یک الگوی آموزشی است که در راستای به حداکثر رساندن کارآیی مدل، چندین وظیفه مرتبط به‌طور هم‌زمان یاد گرفته می‌شوند و هم‌زمان چندین تابع ضرر بهینه می‌شوند.

- یادگیری بدون‌نمونه یک روش یادگیری نظارتی ولی بدون داده‌های آموزشی از آن کلاس است.

- یادگیری بدون‌نمونه قادر است یک مساله را با وجود عدم دریافت هیچ نمونه آموزشی از آن مساله حل کند.

- یادگیری فعال شاخه‌ای از یادگیری ماشین است که در آن یک الگوریتم یادگیری می‌تواند با کاربر ارتباط برقرار کند تا داده‌ها را با خروجی‌های مورد نظر علامت‌گذاری کند.

- هدف از یادگیری فعال افزایش عملکرد الگوریتم یادگیری ماشین و در عین حال ثابت نگه داشتن تعداد نمونه‌های آموزشی است.

- دو راهبرد اساسی در یادگیری فعال عبارتند از: نمونه‌گیری عدم قطعیت  و نمونه‌گیری فضای نمونه.

- در الگوریتم‌های یادگیری دسته‌ای، داده‌های آموزشی از ابتدا به‌طور کامل موجود و در دسترس عامل یادگیری است.

- در الگوریتم‌های یادگیری افزایشی، داده‌های آموزشی ممکن است از ابتدا معلوم یا کامل نباشند و یا ممکن است در طول زمان اضافه شوند.

- الگوریتم‌های یادگیری ماشین فرآیند یادگیری مدلی را خودکار می‌کنند که رابطه بین ویژگی‌های توصیفی و ویژگی هدف در یک مجموعه داده را نشان می‌دهد.

- هدف یادگیری ماشین یافتن مدل پیش‌گویانه‌ای است که بهترین تعمیم را دارد.

- یادگیری ماشین و داده‌کاوی از تکنیک‌های یکسانی استفاده می‌کنند و هم‌پوشانی دارند.

## مراجع برای مطالعه بیشتر

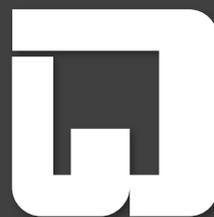

# انتخاب و ارزیابی مدل

## اهداف:

- تفاوت بین مدل و الگوریتم
- آشنایی با مفاهیم بایاس و واریانس
- روش‌های ارزیابی
- روش‌های تنظیم ابرپارمترها
- ارزیابی کارایی

# مدل و الگوریتم

یک سردرگمی رایج برای مبتدیان در یادگیری ماشینی، تفاوت بین "الگوریتم یادگیری ماشین" و "مدل یادگیری ماشین" است. این دو اصطلاح اغلب به جای هم استفاده می‌شوند که موجب سردرگمی می‌شود. به‌طور خلاصه می‌توان گفت آن‌ها یکسان نیستند، یک الگوریتم یادگیری ماشین مانند رویه‌ای است که بر روی داده‌ها اجرا می‌شود تا الگوها و قوانینی را پیدا کند که در آن ذخیره شده و برای ایجاد یک مدل یادگیری ماشین استفاده می‌شود. در ادامه این بخش به شما خواهیم گفت که تفاوت الگوریتم و مدل در یادگیری ماشین چیست؟

## الگوریتم

الگوریتم یادگیری ماشین رویه‌ای است که بر روی داده‌ها برای ایجاد یک "مدل" یادگیری ماشین اجرا می‌شود. به عبارت دیگر، یک الگوریتم در یادگیری ماشین "تشخیص الگو" و "یادگیری" را از داده‌ها انجام می‌دهد. برای ساده‌تر شدن موضوع، می‌توان رابطه بین آن‌ها را به شکل زیر نشان داد:

**مدل یادگیری ماشین ← الگوریتم یادگیری ماشین**

انواع مختلفی از الگوریتم‌ها با عملکردها و اهداف مختلف وجود دارد. سه مورد اصلی عبارتند از:

- **رگرسیون:** برای پیش‌بینی‌هایی که خروجی یک مقدار پیوسته است.
- **دسته‌بندی:** برای پیش‌بینی‌هایی استفاده می‌شود که خروجی آن یک مقدار طبقه‌بندی شده است.
- **خوشه‌بندی:** برای گروه‌بندی چیزهای مشابه یا نقاط داده در خوشه‌ها.

> وقتی یک "الگوریتم" را با داده‌ها آموزش می‌دهید، به یک "مدل" تبدیل می‌شود.

الگوریتم‌های یادگیری ماشین همانند هر الگوریتم دیگری در علوم رایانه هستند. به این ترتیب، الگوریتم‌های یادگیری ماشین دارای ویژگی‌های زیر هستند:

- الگوریتم‌های یادگیری ماشین را می‌توان با استفاده از ریاضیات و شبه‌کد توصیف کرد.
- کارایی الگوریتم‌های یادگیری ماشین را می‌توان تحلیل و توصیف کرد.
- الگوریتم‌های یادگیری ماشین را می‌توان با هر یک از زبان‌های برنامه‌نویسی مدرن پیاده‌سازی کرد.

به عنوان مثال، ممکن است الگوریتم‌های یادگیری ماشین را با شبه کد یا ریاضیات در مقالات و کتاب‌های درسی ببینید. ممکن است کارایی محاسباتی یک الگوریتم یادگیری ماشین خاص را در مقایسه با یک الگوریتم خاص دیگر مشاهده کنید. محققین می‌توانند الگوریتم‌های یادگیری ماشین کاملاً جدیدی ابداع کنند و متخصصان یادگیری ماشین می‌توانند از الگوریتم‌های یادگیری ماشین استاندارد در پروژه‌های خود استفاده کنند. این درست مانند سایر حوزه‌های علوم رایانه است که محققین می‌توانند به عنوان مثال الگوریتم‌های مرتب‌سازی کاملاً جدیدی را ابداع کنند و برنامه‌نویسان می‌توانند از الگوریتم‌های مرتب‌سازی استاندارد در برنامه‌های خود استفاده کنند.

## مدل

یک "مدل" در یادگیری ماشین خروجی یک الگوریتم یادگیری ماشین است که بر روی داده‌ها اجرا می‌شود و نشان‌دهنده آنچه توسط یک الگوریتم یادگیری ماشین آموخته شده است، می‌باشد. برای نشان دادن رابطه بین آن‌ها، می‌توان از رابطه زیر استفاده کنیم:

**مدل یادگیری ماشین = داده‌های مدل + الگوریتم پیش بینی**

در نهایت، مدل "چیزی" است که پس از اجرای الگوریتم یادگیری ماشین روی داده‌های آموزشی ذخیره شده و قوانین، اعداد و سایر ساختار خاص داده‌های الگوریتم مورد نیاز برای پیش‌بینی را نشان می‌دهد.

> "مدل" یادگیری ماشین جایی است که خروجی "الگوریتم" در آن ذخیره می‌شود. مدل را می‌توان برای بعد ذخیره کرد و به عنوان یک برنامه عمل می‌کند و از عملکرد ذخیره شده قبلی الگوریتم برای پیش‌بینی‌های جدید استفاده می‌کند. اگر مدل به‌طور کارآمد و کافی آموزش داده‌شود، می‌توان از آن برای پیش‌بینی‌های بیشتر بر روی داده‌های مشابه با سطح مشخصی از دقت و اطمینان استفاده کرد.

## تفاوت مدل و الگوریتم در یادگیری ماشین

اکنون که می‌دانیم یک الگوریتم و یک مدل چیست، دیدن چگونگی ارتباط آن‌ها آسان‌تر است. همانطور که قبلا ذکر شد، یک الگوریتم بر روی داده‌ها برای ایجاد یک مدل اجرا می‌شود. آن مدل از داده‌ها و روشی برای چگونگی استفاده از داده‌ها برای پیش‌بینی داده‌های جدید تشکیل شده است. این روش تقریبا شبیه یک الگوریتم پیش‌بینی است. اگرچه، همه مدل‌ها یک الگوریتم پیش‌بینی را ذخیره نمی‌کنند. برخی مانند همانند k ـ نزدیکترین همسایه، کل مجموعه داده ذخیره می‌کنند که به عنوان الگوریتم پیش‌بینی عمل می‌کند. ما اساسا یک «مدل» یادگیری ماشین می‌خواهیم و به الگوریتم پشت آن اهمیتی نمی‌دهیم. به عبارت دیگر، الگوریتم فقط مسیری است

که برای بدست آوردن مدل دنبال می‌کنیم. با این حال، مهم است که بدانید کدام الگوریتم را در مدل خود اعمال کنید تا بهترین نتایج را بدست آورید. وقتی این را بدانید، فقط چند خط کد و سطوح کمی از تعامل وجود دارد تا بتوانید یک مدل کاملا کارآمد داشته باشید. بطور خلاصه می‌توان تفاوت مدل و الگوریتم را در یادگیری ماشین اینگونه بیان کرد:

- الگوریتم‌های یادگیری ماشین رویه‌هایی هستند که بر روی داده‌ها برای یافتن الگوها و یادگیری اجرا می‌شوند.

- مدل‌های یادگیری ماشین خروجی الگوریتم‌ها هستند و از داده‌ها و یک الگوریتم پیش‌بینی تشکیل شده‌اند.

- الگوریتم‌های یادگیری ماشین نوعی برنامه‌نویسی خودکار را ارائه می‌کنند که در آن مدل‌های یادگیری ماشین خود برنامه را نشان می‌دهند.

مدل یادگیری ماشین برنامه‌ای است که به‌طور خودکار توسط الگوریتم یادگیری ماشین نوشته یا ایجاد یا یاد گرفته می‌شود تا مساله ما را حل کند. به عنوان یک توسعه‌دهنده، ما کم‌تر به "یادگیری" انجام شده توسط الگوریتم‌های یادگیری ماشین در مفهوم هوش مصنوعی علاقه‌مندیم. ما به شبیه‌سازی فرآیندهای یادگیری اهمیتی نمی‌دهیم. ممکن است برخی از افراد این‌طور باشند و جالب باشد، اما این دلیلی نیست که ما از الگوریتم‌های یادگیری ماشین استفاده می‌کنیم. در عوض، بیشتر به قابلیت برنامه‌ریزی خودکار ارائه‌شده توسط الگوریتم‌های یادگیری ماشین علاقه‌مند هستیم. ما می‌خواهیم یک مدل موثر ایجاد شود تا بتوانیم آن را در پروژه نرم‌افزاری خود بگنجانیم. الگوریتم‌های یادگیری ماشین برنامه‌نویسی خودکار را انجام می‌دهند و مدل‌های یادگیری ماشین برنامه‌هایی هستند که برای ما ایجاد می‌شوند.

## چارچوب کلی برای الگوریتم‌های یادگیری ماشین

چارچوب کلی الگوریتم‌های یادگیری ماشین به این صورت است که با استفاده از مجموعه‌ای از داده‌ها، مدلی برای تولید خروجی برای ورودی‌های جدید که هنوز مشاهده نشده‌اند ساخته می‌شود. یک چارچوب برای یادگیری، که در شکل ۱ ـ ۱ نشان داده شده است، می‌تواند شامل: داده ها (آموزشی و آزمایش)، الگوریتم مورد استفاده برای ساخت مدل و در نهایت ارزیابی مدل برای تولید خروجی باشد که در مورد هر یک از آن‌ها در بخش‌های بعدی توضیح داده خواهد شد.

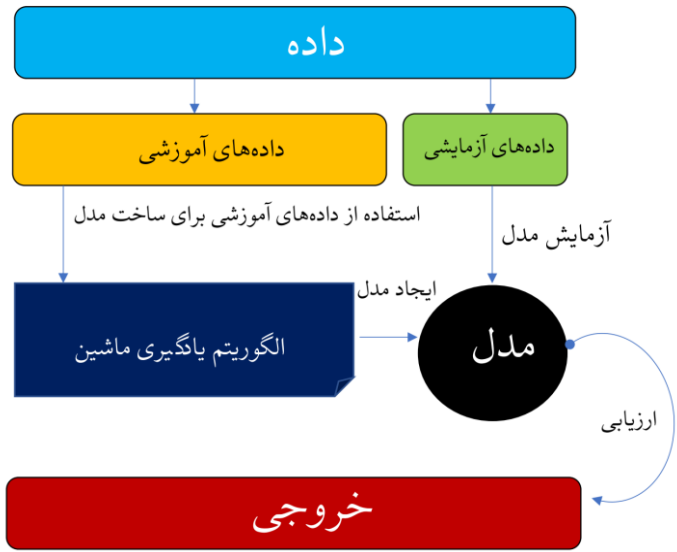

شکل ۱ ـ ۱ چارچوب کلی الگوریتم‌های یادگیری ماشین

## توسعه مدل

## الگوریتم یادگیری و تابع زیان

الگوریتم یادگیری $A$ نگاشتی از یک مجموعه داده محدود $S$ به تابع کاندید $\hat{f}$ است که در آن $\hat{f}$ قابل اندازه‌گیری است. فرض می‌کنیم که داده‌های ما $(x_i, y_i)$ به‌طور مستقل و به‌طور یکسان از فضای احتمال $X \times Y$ با اندازه $\rho$ توزیع شده‌اند. ما معنای "**خوب انجام دادن**" در یک کار را با معرفی یک **تابع زیان**[۱]، یک تابع قابل اندازه‌گیری $L: X \times Y \times F \rightarrow [0, \infty)$ تعریف می‌کنیم. این زیان تقریبا همیشه به شکل $L(x, y, f) = \hat{L}(y, f(x))$ برای برخی از توابع $\hat{L}$ است. از این‌رو، در ادامه از این روش نوشتن استفاده می‌کنیم. طبیعتا، ما باید $L(y, \hat{f}(x))$ را به عنوان اندازه‌گیری هزینه پیش‌بینی $\hat{f}(x)$ که $y$ برچسب واقعی برای $x$ است، در نظر بگیریم. اگر ما $y = f(x)$ را پیش‌بینی کنیم، گوییم پیش‌بینی ما در $x$ عالی (بدون عیب) است و انتظار داریم که هیچ زیانی در $x$ متحمل نشویم (به عنوان مثال $L\left(y, \hat{f}(x)\right) = 0$). انتخاب یک تابع زیان مناسب بخش مهمی از استفاده از یادگیری ماشین در عمل است. در ادامه، نمونه‌هایی از وظایف با فضاهای داده مختلف $X, Y$ و توابع زیان مختلف $L$ را ارائه می‌دهیم.

---

[۱] loss function

## مثال ۱: دسته‌بندی تصاویر

دسته‌بندی تصویر، مساله دسته‌بندی کردن تصویر $x$ به یکی از دسته‌های ممکن $C$ است. در اینجا $x \epsilon R^{H \times W \times ٣}$ است که در آن $H$ و $W$ ارتفاع و عرض تصویر و ٣ مربوط به کانال رنگ (قرمز، سبز و آبی) است. فضای برچسب محدود به یک مجموعه محدود $Y = C$ است. خروجی مدل دسته‌بند یک توزیع گسسته $f(x_i) = (p_١, ..., p_c)$ برروی کلاس‌ها تولید می‌کند که در آن $p_c$ مربوط به احتمال اینکه تصویر ورودی $x$ دارای کلاس $c$ باشد، است. به عنوان تابع زیان برای این مساله از **زیان آنتروپی متقاطع**[1] استفاده می‌کنیم:

$$L\big(y, f(x)\big) = -\frac{١}{N} \sum_{c=١}^{C} ١\{y_i = c\} \log(p_c), \quad p = f(x_i)$$

## مثال ۲: پیش‌بینی خواص بلور

یک کار رایج در علم مواد، پیش‌بینی خواص یک بلور (مثلاً انرژی شکل‌گیری) از ساختار اتمی آن (یک گراف بدون جهت) است. به عنوان یک مساله یادگیری، این یک مساله رگرسیونی با $X$ به عنوان مجموعه گراف‌های بدون جهت و $Y = \mathbb{R}$ است. برای تابع زیان، استفاده از میانگین خطای مطلق (MAE) به دلیل پایداری آن نسبت به نقاط دورافتاده معمول است:

$$L\big(y, f(x)\big) = |y - f(x)|$$

# مساله‌ی یادگیری

یادگیری ماشین به فرآیند استفاده از ابزارهای آماری برای یادگیری و درک داده‌ها اشاره دارد. یادگیری در مورد یافتن تابع $\hat{f}$ است که از داده‌های محدود S به فضای نامتناهی $X \times Y$ تعمیم می‌یابد. این ایده ممکن است به صورت به حداقل رساندن زیان مورد انتظار $\mathcal{E}$ بیان شود که ریسک نیز نامیده می‌شود:

$$\mathcal{E}(f) = \mathbb{E}[L\big(y, f(x)\big)] = \int_{X \times Y} L\big(y, f(x)\big) d_\rho(x, y)$$

هدف ما در یادگیری به حداقل رساندن ریسک است:

$$f^* = argmin_{f \in F} \mathbb{E}\big[L\big(y, f(x)\big)\big]$$

---

[1] cross-entropy loss

$$= argmin_{f \in F} \int_{X \times Y} L\big(y, f(x)\big) d_\rho(x, y)$$

از آنجایی که ما داده‌های محدودی داریم، محاسبه ریسک نیز غیرممکن است. در عوض، ما آن را با استفاده از داده‌های خود تقریب می‌زنیم و **ریسک تجربی**[1] را تولید می‌کنیم:

$$\widehat{\mathcal{E}}(f) = \frac{1}{n} \sum_{i=1}^{N} L(y_i, f(x_i)) \approx \int_{X \times Y} L\big(y, f(x)\big) d_\rho(x, y) \qquad \text{۲–۱}$$

این مفهوم که **به حداقل رساندن ریسک تجربی** نامیده می‌شود، اساس یادگیری ماشین مدرن است.

می‌توان امیدوار بود که با به حداقل رساندن ریسک تجربی بر روی همه توابع قابل اندازه‌گیری، بتوانیم عبارت در سمت راست ۲–۱ را تقریب بزنیم و تابع $\hat{f} = argmin_{f \in F}\widehat{\mathcal{E}}(f)$ مشابه تابع مورد نظر $f^*$ را پیدا کنیم. با این حال، بدون مفروضات یا مقدمات اضافی، این امر امکان‌پذیر نیست. در این تنظیمات بدون محدودیت، هیچ مدلی نمی‌تواند به خطای کم در تمام توزیع‌های داده دست یابد، نتیجه‌ای که به عنوان **قضیه ناهار مجانی نیست**[2] شناخته می‌شود.

تفاوت بین عملکرد تابع یادگیری تجربی ما $\hat{f}$ و بهترین عملکرد ممکن، شکاف تعمیم یا خطای تعمیم[3] نامیده می‌شود. هدف ما به حداقل رساندن احتمال این خطا است $\epsilon$:

$$\mathbb{P}\big(\widehat{\mathcal{E}}(f) - inf_{f \in F}\varepsilon(f) > \epsilon\big)$$

اگر این شکاف در حد بی‌نهایت داده به صفر کاهش یابد، مطلوب است:

$$\lim_{n \to \infty} \mathbb{P}\big(\widehat{\mathcal{E}}(f) - inf_{f \in F}\varepsilon(f) > \epsilon\big) = \cdot \ \forall_\epsilon > \cdot$$

خطای تجربی گاهی اوقات خطای تعمیم نیز نامیده می‌شود. دلیل آن این است که در واقع، در اکثر مسائل، ما به کل دامنه $X$ ورودی‌ها دسترسی نداریم، بلکه فقط به زیر مجموعه آموزشی S خود دسترسی داریم و می‌خواهیم بر اساس S تعمیم‌دهی انجام دهیم که یادگیری استقرایی[4] نیز نامیده می‌شود.

---

[1] empirical risk

[2] No Free Lunch Theorem

[3] generalization error

[4] inductive learning

## قضیه ناهار مجانی نیست!

قضیه ناهار مجانی نیست (NFL) در یادگیری ماشین بانظارت، قضیه‌ای است که اساسا به این معناست که هیچ الگوریتم یادگیری ماشین به طور کلی بهترین الگوریتم برای همه مسائل نیست. شاید کمی عجیب به نظر برسد اما، ایده‌ای که ممکن است الهام بخش قضیه NFL باشد، اولین بار توسط فیلسوفی از دهه ۱۷۰۰ ارائه شد. بله آن را درست خواندید! نه یک ریاضیدان یا یک متخصص آمار، بلکه یک فیلسوف. در اواسط دهه ۱۷۰۰، یک فیلسوف اسکاتلندی به نام دیوید هیوم چیزی را مطرح کرد که او آن را مسئله استقرا[۱] نامید. این مسئله یک سؤال فلسفی است که می‌پرسد آیا استدلال استقرایی واقعا ما را به دانش واقعی می‌رساند؟ استدلال استقرایی شکلی از استدلال است که در آن ما بر اساس مشاهدات گذشته در مورد جهان نتیجه می‌گیریم. در کمال تعجب، این دقیقا همان کاری است که الگوریتم‌های یادگیری ماشین انجام می‌دهند. اگر یک شبکه عصبی ۱۰۰ تصویر از قوهای سفید را ببیند، احتمالا به این نتیجه خواهد رسید که همه قوها سفید هستند. اما اگر شبکه عصبی یک قو سیاه ببیند چه اتفاقی می‌افتد؟ اکنون الگوی آموخته‌شده توسط الگوریتم به‌طور ناگهانی تنها با یک مثال متقابل رد می‌شود. این ایده اغلب به عنوان **پارادوکس قو سیاه**[۲] شناخته می‌شود.

هیوم از این منطق برای برجسته کردن محدودیت استدلال استقرایی استفاده کرد که ما نمی‌توانیم نتیجه‌گیری درباره مجموعه‌ای از مشاهدات را به مجموعه‌ای کلی‌تر از مشاهدات اعمال کنیم. دیوید هیوم در *رساله‌ای درباره طبیعت آدمی* می‌گوید: «هیچ دلیلی برای اثبات این امر وجود ندارد، آن مواردی که ما تجربه‌ای از آن‌ها نداشته‌ایم، مشابه مواردی هستند که ما تجربه آن‌ها را داشته‌ایم». همین ایده بیش از ۲۰۰ سال بعد الهام‌بخش قضیه NFL برای یادگیری ماشین شد.

ولپرت در مقاله[۳] خود در سال ۱۹۹۶، قضیه نهار مجانی نیست را برای یادگیری ماشین بانظارت معرفی کرد و در واقع از نقل قول دیوید هیوم در ابتدای مقاله خود استفاده کرد. این قضیه بیان می‌کند که با توجه به مجموعه داده‌های بدون نویز، برای هر دو الگوریتم یادگیری ماشین A و B، میانگین عملکرد A و B در تمام نمونه‌های مساله‌ی ممکن از توزیع احتمال یکنواخت گرفته شده‌اند، یکسان خواهد بود.

چرا این درست است؟ این به مفهوم استدلال استقرایی برمی‌گردد. هر الگوریتم یادگیری ماشین فرضیات قبلی را در مورد رابطه بین ویژگی‌ها و متغیرهای هدف برای یک مساله یادگیری

---

ماشین ایجاد می‌کند. این مفروضات را اغلب مفروضات پیشین[1] می‌نامند. عملکرد یک الگوریتم یادگیری ماشین در هر مساله‌ای بستگی به این دارد که مفروضات الگوریتم تا چه حد با واقعیت مسئله مطابقت دارند. یک الگوریتم ممکن است برای یک مساله بسیار خوب عمل کند، اما هیچ دلیلی وجود ندارد که باور کنیم آن هم به همان خوبی روی یک مساله متفاوت که ممکن است در آن فرضیات کارساز نباشد، عمل می‌کند. این مفهوم اساسا پارادوکس قو سیاه در زمینه یادگیری ماشین است.

مفروضات محدود کننده‌ای که هنگام انتخاب هر الگوریتمی انجام می‌دهید مانند قیمتی است که برای ناهار می‌پردازید. این مفروضات الگوریتم شما را به طور طبیعی در برخی مسائل بهتر می‌کند در حالی که به طور هم‌زمان آن را به‌طور طبیعی در مسائل دیگر بدتر می‌کند.

همه این تئوری‌ها عالی هستند، اما NFL برای شما به عنوان یک دانشمند داده، یک مهندس یادگیری ماشین یا کسی که فقط می‌خواهد یادگیری ماشین را شروع کند چه معنایی دارد؟ یعنی همه الگوریتم ها برابرند؟ صد البته که نه!! در عمل، همه الگوریتم‌ها یکسان ایجاد نمی‌شوند. این به این دلیل است که کل مجموعه مسائل یادگیری ماشین یک مفهوم نظری در قضیه NFL است و بسیار بزرگ‌تر از مجموعه مسائل یادگیری ماشین عملی است که ما در واقع سعی خواهیم کرد آن‌ها را حل کنیم. برخی از الگوریتم‌ها معمولاً ممکن است در انواع خاصی از مسائل بهتر از بقیه عمل کنند، اما هر الگوریتم به دلیل مفروضات قبلی که با آن الگوریتم ارائه می‌شود، دارای معایب و مزایایی است. الگوریتمی مانند XGBoost ممکن است در صدها مسابقه Kaggle برنده شود، اما به دلیل مفروضات محدود کننده موجود در مدل‌های مبتنی بر درخت، در پیش‌بینی وظایف به‌شدت شکست بخورد. شبکه‌های عصبی ممکن است در انجام وظایف پیچیده‌ای مانند دسته‌بندی تصویر و تشخیص گفتار بسیار خوب عمل کنند، اما اگر به‌درستی آموزش نبینند، به دلیل پیچیدگی‌هایشان، از بیش‌برازش رنج ببرند.

در عمل، معنای NFL این‌گونه است:

- هیچ الگوریتم واحدی تمام مسائل یادگیری ماشین شما را بهتر از هر الگوریتم دیگری حل نمی‌کند.

- قبل از انتخاب یک الگوریتم برای استفاده، مطمئن شوید که مساله یادگیری ماشین و داده‌های مربوط به آن را کاملا درک کرده‌اید.

- مدل‌های ساده‌تر مانند رگرسیون لجستیک دارای بایاس بیشتری هستند و تمایل به کم‌برازش دارند، در حالی که مدل‌های پیچیده‌تر مانند شبکه‌های عصبی واریانس بیشتری دارند و تمایل به بیش‌برازش دارند.

---

[1] priori assumptions

- بهترین مدل‌ها برای یک مسئله معین، در جایی در میانه دو لبه بایاس ــ واریانس وجود دارد.

- برای پیدا کردن یک مدل خوب برای یک مساله، ممکن است مجبور شوید مدل‌های مختلف را امتحان کنید و با استفاده از یک استراتژی اعتبارسنجی متقابل قوی مقایسه کنید.

## پیش‌بینی در مقابل استنباط

به طور کلی، هدف هر فرآیند یادگیری بانظارت، پیش‌بینی یک متغیر کمی یا کیفی $y$ بر اساس مجموعه‌ای از $p$ پیش‌بینی‌کننده $x_1, x_2, \ldots, x_p$ است. علاوه بر این، ما فرض می‌کنیم که یک رابطه بین $y$ و $x_p$ وجود دارد. در ساده‌ترین حالت، این رابطه را می‌توان به صورت بیان کرد:

$$y = f(x_p) + \epsilon$$

جایی که $f$ یک تابع ناشناخته اما با ثابت‌های $x_1, x_2, \ldots, x_p$ و $\epsilon$ عبارت خطایی است که تمام متغیرهایی را که با $y$ مرتبط هستند، اما در مدل گنجانده نشده‌اند را نشان می‌دهد. در آمار، فرض می‌کنیم که عبارت خطا مستقل از $x_p$ است و میانگین آن صفر است. به عبارت دیگر، $f$ یک تابع (ناشناخته) است که رابطه بین متغیر پاسخ و پیش‌بینی کننده را ترسیم می‌کند. از آنجایی که تابع ناشناخته است، باید این تابع را بر اساس نقاط داده مشاهده شده تخمین بزنیم. در تجزیه و تحلیل آماری، دو حوزه اصلی مورد علاقه برای تخمین تابع $f$ وجود دارد، یعنی **استنباط** و **پیش‌بینی**.

یادگیری ماشین عمدتا به پیش‌بینی علاقه دارد. به طور خاص، به شناسایی مجموعه‌ای از پیش‌بینی‌کننده‌هایی که دقیق‌ترین پیش‌بینی‌ها را برای خروجی $y$ ارائه می‌دهند، علاقه‌مند است و کمتر نگران ماهیت رابطه است. به عبارت دیگر، تا زمانی که قدرت پیش‌بینی‌کننده بالا و سازگار باشد، غیرضروری است که آیا ارتباط علی بین $x_p$ پیش‌بینی‌کننده و متغیر پاسخ $y$ وجود دارد یا خیر. در نتیجه، لازم نیست که در مورد داده‌ها و شکل دقیق $f$ فرضیاتی داشته باشیم. از آنجایی که عبارت خطا به‌طور متوسط صفر است، می توانیم $y$ را بر اساس مجموعه پیش‌بینی کننده‌ها بدین گونه پیش‌بینی کنیم

$$\hat{y} = \hat{f}(x)$$

جایی که $\hat{f}$ تابع تخمین $f$ و $\hat{y}$ مقادیر پیش‌بینی شده برای $y$ است. در یادگیری ماشین، هدف معمولا تخمین تابع $\hat{f}$ است که خطای پیش‌بینی را به حداقل می‌رساند.

## تعریف   پیش‌بینی

### استفاده از مدل برای پیش‌بینی نقاط داده جدید

پیش‌بینی زمانی که به‌عنوان اسم استفاده می‌شود، به گزاره‌ای قاطع درباره رویداد یا یک حادثه در آینده اشاره دارد. ارزیابی آن ممکن است بر اساس داده‌ها، حقایق و شواهد باشد یا نباشد. همیشه مشخص نیست که آیا یک پیش‌بینی درست خواهد بود یا خیر. این به این دلیل است که پیش‌بینی که در مورد آینده انجام می‌شود ناشناخته است. هنگامی که به عنوان یک فعل استفاده می‌شود، این اصطلاح به عنوان "پیش‌بینی کردن" معنا می‌گیرد. یک مثال این است که هواشناس «پیش‌بینی» می‌کرد که آیا باران خواهد بارید یا نه. یا فالگیر «پیش‌بینی» کرد که خانه به زودی فروخته خواهد شد. چند نمونه از اصطلاح «پیش‌بینی» در یک جمله عبارتند از: «پیش‌بینی» او درباره آینده اشتباه بود. یا علی "پیش بینی" کرد که تیم قرمز در آن روز برنده مسابقات خواهد شد.

## تعریف   استنباط

### استفاده از مدل برای یادگیری در مورد فرآیند تولید داده

استنباط، زمانی که به عنوان اسم استفاده می‌شود، به عمل رسیدن به نتیجه‌ای اشاره دارد که بر اساس داده‌ها، حقایق و شواهد موجود ارزیابی شده است. این شامل ساخت مدل است که رابطه بین متغیرها و نتیجه یک رویداد یا رخداد را با استفاده از داده‌های آماری توصیف می‌کند. از آنجایی که ارزیابی انجام شده واقعی است، تا حد زیادی اطمینان وجود دارد. علاوه بر این، نتیجه‌گیری ممکن است لزوما حول محور آینده‌ای که تمایل به ناشناخته بودن دارد، نباشد. معمولا زمانی از اصطلاح "استنباط" استفاده می‌شود که نتیجه‌گیری در مورد حال باشد. هنگامی که به عنوان یک فعل استفاده می‌شود، این اصطلاح به عنوان "استنباط کردن" شناخته می‌شود. این به معنای رسیدن به یک نتیجه‌گیری است. به عنوان مثال، اگر کودکان با خوردن یک غذا چهره بدی نشان دهند، مادران آن‌ها «استنباط» می‌کنند که آن‌ها این غذا را دوست ندارند. یا اگر مردم برای رستورانی نظر منفی بدهند، «استنباط» می‌شود که غذای آن‌ها بد است. برخی از نمونه‌های اصطلاح «استنباط» در یک جمله عبارتند از: علی درباره آنچه زیر میز بود «استنباط» کرد، یا کارآگاه از دستیارش خواست که بر اساس سرنخ‌های موجود «استنباط» کند.

در پیش‌بینی، ما علاقه‌مندیم که $f$ را تا حد امکان دقیق تخمین بزنیم تا بتوانیم به‌طور بالقوه برای متغیر هدف $Y$ بر اساس متغیرهای مستقل $X$ پیش‌بینی کنیم. از طرف

دیگر، در استنباط ما همچنان علاقه‌مند به تخمین $f$ هستیم، اما این بار نه برای انجام پیش‌بینی، بلکه برای درک رابطه بین $X$ و $Y$.

## تفسیرپذیری یک ضرورت برای استنباط

در اساس، تفاوت بین مدل‌هایی که برای استنباط مناسب هستند و مدل‌هایی که مناسب نیستند، به قابلیت تفسیر مدل خلاصه می‌شود. منظور از تفسیرپذیری مدل چیست؟ می‌توان مدلی را قابل تفسیر دانست که یک انسان بتواند نحوه تولید تخمین‌هایش را دوباره بررسی کند. برای پیش‌بینی روش‌های زیر را در نظر بگیرید:

- **قابل تفسیر:** مدل‌های خطی تعمیم‌یافته (مانند رگرسیون خطی، رگرسیون لجستیک)، تحلیل تفکیک خطی[1]، ماشین‌های بردار پشتیبان خطی و درخت‌های تصمیم
- **غیرقابل تفسیر (کمتر قابل تفسیر):** شبکه‌های عصبی، ماشین‌های بردار پشتیبان غیرخطی و جنگل‌های تصادفی

*تنها یک زیرمجموعه از روش‌های قابل تفسیر برای استنباط مفید هستند.* به عنوان مثال، ماشین‌های بردار پشتیبان خطی قابل تفسیر هستند، چون آن‌ها یک ضریب برای هر ویژگی فراهم می‌کنند به‌طوری که بتوان تاثیر ویژگی‌های فردی بر پیش‌بینی را توضیح داد. با این حال، ماشین‌های بردار پشتیبان امکان تخمین عدم قطعیت مرتبط با ضرایب مدل (به عنوان مثال واریانس) را نمی‌دهند و نمی‌توان یک معیار ضمنی از اطمینان مدل بدست آورد. توجه داشته باشید که ماشین‌های بردار پشتیبان قادر به ایجاد خروجی احتمالات هستند، اما این احتمالات فقط تبدیلی از مقادیر تصمیم هستند و بر اساس اطمینان مرتبط با تخمین پارامترها نیستند. به همین دلیل است که حتی روش‌های قابل تفسیر مانند ماشین‌های بردار پشتیبان خطی و درخت‌های تصمیم برای استنباط نامناسب هستند. در مقابل، رگرسیون خطی را در نظر بگیرید که فرض می‌کند داده‌ها از توزیع گاوسی پیروی می‌کنند. این مدل‌ها، خطای استاندارد برآورد ضرایب و فواصل اطمینان خروجی را تعیین می‌کنند. از آنجایی که رگرسیون خطی به ما امکان می‌دهد ماهیت احتمالی فرآیند تولید داده را درک کنیم، روش مناسبی برای استنباط است. روش‌های بیزی برای استنباط بسیار محبوب هستند، زیرا این مدل‌ها را می‌توان برای ترکیب مفروضات مختلف در مورد فرآیند تولید داده تنظیم کرد.

*صرف استفاده از مدلی که برای استنباط مناسب است به این معنا نیست که شما واقعا استنباط انجام می‌دهید.* مهم این است که چگونه از مدل استفاده می‌کنید. به عنوان مثال، اگرچه مدل‌های

---

[1] linear discriminant analysis

خطی تعمیم‌یافته برای استنباط مناسب هستند، می‌توان از آن‌ها صرفا برای اهداف پیش‌بینی استفاده کرد. مثال‌های زیر را در نظر بگیرید که تمایز بین پیش‌بینی و استنباط را واضح‌تر می‌کند:

- **پیش‌بینی:** می‌خواهید سطح ازون آینده را با استفاده از داده‌های گذشته پیش‌بینی کنید. از آنجایی که معتقدید یک رابطه خطی بین سطح ازون و اندازه‌گیری دما، تابش خورشیدی و باد وجود دارد، چندین مدل خطی را روی داده‌های آموزشی قرار می‌دهید و مدلی را انتخاب می‌کنید که خطا را در مجموعه آزمایشی به‌حداقل می‌رساند. در نهایت از مدل انتخاب شده برای پیش‌بینی سطح ازون استفاده می‌کنید. توجه داشته باشید، تا زمانی که مدل خطای آزمایش را به حداقل برساند، اصلا به فرض گاوسی مدل یا اطلاعات اضافی که در تخمین مدل موجود است اهمیتی نمی‌دهید.

- **استنباط:** شما می‌خواهید بفهمید که چگونه سطح ازون تحت تاثیر دما، تابش خورشیدی و باد قرار می‌گیرد. از آنجایی که فرض می‌کنید داده‌ها به‌طور عادی (نرمال) توزیع شده‌اند، از مدل رگرسیون خطی استفاده می‌کنید. به منظور بدست آوردن اطلاعات و از آنجایی که دقت پیش‌بینی برای شما مهم نیست، برای مدل‌سازی از کل مجموعه داده استفاده می‌شود. بر اساس مدل برازش شده، نقش ویژگی‌ها را در سطح ازون اندازه‌گیری‌شده تفسیر می‌کنید. به عنوان مثال، با در نظر گرفتن باندهای اطمینان (فواصل اطمینان) برآوردها.

## تقسیم‌بندی داده‌ها

هنگام پیاده‌سازی مدل‌های یادگیری ماشین برای اهداف پیش‌بینی، تقسیم‌بندی مناسب داده‌ها برای ارزیابی عینی عملکرد مدل‌ها و جلوگیری از کمبود داده‌ها بسیار مهم است. در موقعیت‌هایی که داده‌های کافی وجود دارد، تقسیم داده‌ها را می‌توان به‌طور تصادفی به سه گروه دسته‌بندی کرد: یک مجموعه آموزشی، یک مجموعه اعتبارسنجی و یک مجموعه آزمون (آزمایشی). مجموعه آموزشی همان‌طور که از نامش پیداست برای آموزش و برازش مدل‌ها استفاده می‌شود. مجموعه اعتبارسنجی برای بدست آوردن مقادیر بهینه ابرپارامترهای بهینه (بهینه‌سازی ابرپارامترها) و کمک به انتخاب مدل استفاده می‌شود و مجموعه آزمون برای ارزیابی عملکرد مدل نهایی در نمونه‌های دیده شده در فرآیند یادگیری استفاده می‌شود. هیچ قانون واضحی در مورد اندازه مربوط به‌گروه‌های مختلف مجموعه داده‌ها وجود ندارد، زیرا این امر تا حد زیادی به در دسترس بودن داده‌ها بستگی دارد. با این حال، مجموعه آموزشی معمولا بزرگ‌ترین بخش از داده‌ها را تشکیل می‌دهد، چراکه مدل‌های یادگیری ماشین باید بر روی مقادیر زیادی از داده‌ها آموزش داده شوند تا موثر باشند.

# انتخاب و ارزیابی مدل

محور یادگیری ماشینی حول مفهوم الگوریتم‌ها یا مدل‌هایی است که در واقع تخمین‌های آماری انجام می‌دهند. با این حال، هر مدلی، بسته به توزیع داده محدودیت‌های متعددی دارد. هیچ یک از آن‌ها نمی‌توانند کاملا دقیق باشند، چرا که آن‌ها فقط تخمین هستند. این محدودیت‌ها عموما با نام بایاس و واریانس شناخته می‌شوند. یک مدل با بایاس بالا با توجه نکردن به نقاط آموزشی بیش از حد ساده می‌شود (به عنوان مثال، در رگرسیون خطی، صرف نظر از توزیع داده‌ها، مدل همیشه یک رابطه خطی را در نظر می‌گیرد). هم‌چنین، یک مدل با واریانس بالا با تعمیم ندادن نقاط آزمایشی که قبلا ندیده است، خود را به داده‌های آموزشی محدود می‌کند.

یادگیرندگان خوبی که ما به دنبال آن‌ها هستیم، آن‌هایی هستند که در نمونه‌های جدید عملکرد خوبی دارند. از این رو، یادگیرندگان خوب باید قواعد کلی را از نمونه‌های آموزشی بیاموزند به گونه‌ای که قوانین آموخته‌شده برای همه نمونه‌های بالقوه (دیده‌نشده) اعمال شود. با این حال، زمانی که یادگیرنده مثال‌های آموزشی را «خیلی خوب» یاد می‌گیرد، این احتمال وجود دارد که برخی از ویژگی‌های مثال‌های آموزشی به عنوان ویژگی‌های کلی در نظر گرفته شوند که همه نمونه‌های بالقوه خواهند داشت و در نتیجه عملکرد تعمیم کاهش می‌یابد. در یادگیری ماشین، این پدیده به نام بیش‌برازش و برعکس آن به عنوان کم‌برازش شناخته می‌شود، یعنی یادگیرنده در یادگیری ویژگی‌های کلی مثال‌های آموزشی شکست می‌خورد.

## تعریف بیش‌برازش

بیش‌برازش به وضعیتی اشاره دارد که در آن یک مدل آماری به جای دنبال کردن سیگنال در داده‌ها، نویز یا خطاها را خیلی دقیق دنبال می‌کند.

اگر یک مدل یادگیری ماشینی عملکرد بسیار خوبی را در داده‌های آموزشی نشان دهد (خطای آموزش کم) اما هنگام آزمایش روی داده‌های جدید ضعیف عمل کند (خطای آزمون بالا)، این معمولا نشانه‌ای است که مدل دچار بیش‌برازش شده است. این یک وضعیت بسیار بدی است، زیرا به این معنی است که مدل بیش از حد با داده‌های آموزشی مطابقت دارد و قادر به تعمیم روابط ویژگی‌ها به داده‌های جدید نیست. این به این دلیل است که مدل داده‌هایی را که دیده است به خاطر می‌سپارد و نمی‌تواند به نمونه‌های دیده نشده تعمیم پیدا کند.

در یادگیری با نظارت، بیش‌برازش زمانی اتفاق می‌افتد که مدل ما نویز را همراه با الگوی اساسی در داده‌ها ضبط کند. در مقابل بیش‌برازش، در یادگیری بانظارت، کم‌برازش زمانی اتفاق می‌افتد که مدل نتواند الگوی زیربنایی داده‌ها را بدست آورد.

در میان بسیاری از دلایل احتمالی، توانایی یادگیری بیش از حد قوی، یک دلیل رایج برای بیش‌برازش است، چراکه چنین یادگیرندگانی می‌توانند ویژگی‌های غیرکلی مثال‌های آموزشی را بیاموزند. در مقابل، کم‌برازش معمولا به دلیل توانایی یادگیری ضعیف است. در عمل، غلبه بر کم‌برازش نسبتا آسان است. به عنوان مثال، ما می‌توانیم در یادگیری درخت تصمیم انشعاب بیشتری انجام دهیم یا دوره‌های آموزشی بیشتری در یادگیری شبکه‌های عصبی اضافه کنیم. با این حال، همانطور که بعدا خواهیم دید، بیش‌برازش یک مشکل اساسی در یادگیری ماشین است، از این‌رو، متدهای متنوعی برای کاهش آن پیاده‌سازی شده است. با این وجود، باید بدانیم که بیش‌برازش اجتناب‌ناپذیر است و تنها کاری که می‌توانیم انجام دهیم کاهش آن است، نه حذف آن به‌طور کامل. این استدلال را می توان به‌طور خلاصه به شرح زیر توجیه کرد:

*مسائل یادگیری ماشین اغلب NP-hard یا حتی سخت‌تر هستند، اما الگوریتم‌های یادگیری عملی باید یادگیری را در زمان چندجمله‌ای به پایان برسانند. بنابراین، اگر بیش‌برازش قابل اجتناب باشد، به حداقل رساندن خطای تجربی منجر به راه‌حل بهینه می‌شود و بنابراین ما یک اثبات سازنده برای P=NP داریم. به عبارت دیگر، تا زمانی که P≠NP اعتقاد داشته باشیم، بیش‌برازش اجتناب‌ناپذیر است.*

در عمل، اغلب چندین الگوریتم یادگیری نامزد وجود دارد و حتی یک الگوریتم یادگیری ممکن است مدل‌های متفاوتی را تحت تنظیم پارامترهای مختلف تولید کند. از این‌رو، کدام الگوریتم یادگیری را انتخاب کنیم و از کدام تنظیمات پارامتر استفاده کنیم؟ این مساله به عنوان **انتخاب مدل** نامیده می‌شود. راه‌حل ایده‌آل این است که همه مدل‌های نامزد را ارزیابی کنید و مدلی را انتخاب کنید که کم‌ترین خطای تعمیم را دارد. با این حال، ما نمی‌توانیم خطای تعمیم را مستقیما بدست آوریم، در حالی که خطای تجربی از بیش‌برازش رنج می‌برد. بنابراین، چگونه می‌توانیم مدل‌ها را در عمل ارزیابی و انتخاب کنیم؟ در ادامه پس از مروری کوتاه بر موازنه بایاس و واریانس، روش‌ها و معیارهای انتخاب مدل را شرح خواهیم داد.

## موازنه بایاس-واریانس[1]

هدف اصلی مدل یادگیری ماشین، یادگیری از داده‌های تغذیه شده و ایجاد پیش‌بینی بر اساس الگوی مشاهده شده در طول فرآیند یادگیری است. با این حال، وظیفه ما به همین جا ختم نمی‌شود. ما باید به طور مستمر در مدل‌ها، بر اساس نوع نتایجی که ایجاد می‌کند، بهبودهایی ایجاد کنیم. ما عملکرد مدل را با استفاده از معیارهایی همانند دقت، میانگین مربعات خطا، امتیاز F۱ و غیره تعیین می‌کنیم و سعی می‌کنیم این معیارها را بهبود بخشیم. این مساله اغلب زمانی مشکل‌ساز می‌شود که ما باید انعطاف‌پذیری مدل را حفظ کنیم. چراکه عملکرد یک مدل

---

[1] Bias-Variance Trade-Off

یادگیری ماشین بر اساس میزان دقت پیش‌بینی آن و میزان تعمیم آن بر روی مجموعه داده مستقل دیگری که در فرآیند یادگیری آن را ندیده است ارزیابی می‌شود.

یک مدل یادگیری ماشینی بانظارت قصد دارد خود را بر روی متغیرهای ورودی ($X$) به گونه‌ای آموزش دهد که مقادیر پیش‌بینی‌شده ($Y$) تا حد امکان به مقادیر واقعی نزدیک شوند. این تفاوت بین مقادیر واقعی و مقادیر پیش‌بینی شده **خطا** است و برای ارزیابی مدل استفاده می‌شود. به‌طور کلی خطای هر الگوریتم یادگیری ماشین بانظارت به دو دسته قابل تقسیم است:

۱.   خطاهای کاهش‌پذیر
۲.   خطاهای کاهش‌ناپذیر

خطاهای کاهش‌ناپذیر خطاهایی هستند که حتی با استفاده از هر مدل یادگیری ماشین دیگر نمی‌توان آن‌ها را کاهش داد. به عنوان مثال، نویز خطای کاهش‌ناپذیری است که نمی‌توانیم آن را حذف کنیم. از سوی دیگر بایاس و واریانس خطاهای کاهش‌پذیری هستند که می‌توانیم سعی کنیم تا حد امکان آن‌ها را به حداقل برسانیم. به دلیل همین بایاس_واریانس، سبب می‌شود که مدل یادگیری ماشین با داده‌های داده‌شده منجر به بیش‌برازش یا کم‌برازش شود. کاهش خطاها مستلزم انتخاب مدل‌هایی است که دارای پیچیدگی و انعطاف‌پذیری مناسب باشند. دانشمندان داده باید به‌طور کامل تفاوت بین بایاس و واریانس را برای کاهش خطا و ساخت مدل‌های دقیق درک کنند.

بایاس ناتوانی یک مدل یادگیری ماشین برای بدست آوردن رابطه واقعی بین متغیرهای داده است. این امر ناشی از فرضیات اشتباهی است که درون الگوریتم یادگیری است. به عنوان مثال، در رگرسیون خطی، رابطه بین $X$ و متغیر $Y$ خطی فرض می‌شود، در حالی که در واقعیت ممکن است این رابطه کاملا خطی نباشد. در مقابل، برخلاف بایاس، واریانس زمانی است که مدل نوسانات داده‌ها را در نظر می‌گیرد، به عبارت دیگر نویز را نیز در نظر می‌گیرد. بنابراین، وقتی مدل ما واریانس بالایی دارد، یعنی مدل بیش از حد از داده‌های آموزشی یاد می‌گیرد (واریانس چیزی نیست جز مفهوم بیش‌برازش مدل در یک مجموعه داده خاص)، به‌طوری که در مواجهه با داده‌های جدید (آزمایش)، قادر به پیش‌بینی دقیق نیست.

**تعریف** بایاس

بایاس (سوگیری) تفاوت بین میانگین پیش‌بینی مدل ما و مقدار درستی است که در تلاش برای پیش‌بینی آن هستیم.

مدل با بایاس بالا توجه بسیار کمی به داده‌های آموزشی دارد و مدل را بیش از حد ساده می‌کند و همیشه منجر به خطای بالایی در آموزش و داده‌های آزمون می‌شود.

## تعریف | واریانس

**واریانس تغییرپذیری پیش‌بینی مدل برای یک نقطه داده معین، یایک مقدار است که به ما پراکندگی داده‌ها را نشان می‌دهد.**

مدل با واریانس بالا توجه زیادی به داده‌های آموزشی می‌کند و به داده‌هایی که قبلا ندیده تعمیم نمی‌یابد. در نتیجه، چنین مدل‌هایی روی داده‌های آموزشی بسیار خوب عمل می‌کنند، اما نرخ خطای بالایی در داده‌های آزمایشی دارند.

واریانس نشان می‌دهد که در صورت استفاده از داده‌های آموزشی متفاوت، برآورد تابع هدف چقدر تغییر می‌کند. به عبارت دیگر، واریانس بیان می‌کند که یک متغیر تصادفی چقدر با مقدار مورد انتظارش تفاوت دارد. واریانس می‌تواند منجر به بیش‌برازش شود که در آن نوسانات کوچک در مجموعه آموزشی بزرگ می‌شود. یک مدل با واریانس بالا ممکن است به جای تابع هدف، نویز تصادفی را در مجموعه داده‌های آموزشی منعکس کند. یک مدل با واریانس بالا منجر به تغییرات قابل توجهی در پیش‌بینی‌های تابع هدف می‌شود.

هنگام ساختن یک الگوریتم یادگیری ماشین بانظارت، هدف دستیابی به بایاس و واریانس کم برای دقیق‌ترین پیش‌بینی‌ها است. یک مدل با واریانس بالا ممکن است مجموعه داده‌ها را به دقت نشان دهد، اما می‌تواند منجر به بیش‌برازش شود. در مقابل، یک مدل با بایاس بالا ممکن است با داده‌های آموزشی مناسب نباشد. چالش تعادل به نوعِ مدلِ مورد نظر بستگی دارد. یک الگوریتم یادگیری ماشین خطی بایاس بالا اما واریانس کم را نشان می‌دهد. از سوی دیگر، یک الگوریتم غیرخطی بایاس کم اما واریانس بالایی را نشان خواهد داد. استفاده از یک مدل خطی با مجموعه داده‌ای که غیر خطی است، بایاس را به مدل وارد می‌کند. این مدل در مقایسه با مجموعه داده‌های آموزشی، با توابع هدف مناسب نیست. عکس این موضوع نیز صادق است، اگر از یک مدل غیرخطی برروی یک مجموعه داده خطی استفاده کنید، مدل غیرخطی بیش از حد با تابع هدف مطابقت خواهد داشت.

برای مقابله با این چالش‌های موازنه، یک دانشمند داده باید الگوریتم یادگیری را بسازد که به اندازه کافی منعطف باشد تا بدرستی با داده‌ها تطبیق پیدا کند. با این حال، همیشه تنش بین بایاس و واریانس وجود دارد. در واقع، ایجاد مدلی که هم بایاس و هم واریانس کم داشته باشد، دشوار است (گریز از رابطه بین بایاس و واریانس در یادگیری ماشین وجود ندارد):

**افزایش بایاس باعث کاهش واریانس می‌شود.**

**افزایش واریانس باعث کاهش بایاس می‌شود.**

موازنه بایاس ــ واریانس یک مشکل جدی در یادگیری ماشین است. این وضعیتی است که شما نمی‌توانید هم بایاس کم و هم واریانس کم داشته باشید. در واقع، ما نمی‌توانیم بایاس واقعی و واریانس را محاسبه کنیم، زیرا تابع هدف اصلی واقعی را نمی‌دانیم. با این وجود، به عنوان یک چارچوب، بایاس و واریانس ابزارهایی را برای درک رفتار الگوریتم‌های یادگیری ماشین در پیگیری عملکرد پیشگویانه، فراهم می‌کنند و شما باید با آموزش مدلی که نظم‌های موجود را در داده‌ها به اندازه کافی دقیق و قابل تعمیم تعمیم به مجموعه متفاوتی از نقاط از یک منبع بدست می‌آورد، با داشتن بایاس و واریانس بهینه، یک تعادل داشته باشید.

> بایاس و واریانس دو خطا در خطای کل در الگوریتم یادگیری هستند، اگر سعی کنید یک خطا را کاهش دهید، خطای دیگر ممکن است بالا رود.

تصویر زیر بایاس ــ واریانس را بهتر نشان می‌دهد. مرکز نتیجه مدلی است که ما می‌خواهیم به آن برسیم که تمام مقادیر را به درستی پیش بینی می‌کند. همان‌طور که از مرکز دور می‌شویم، مدل ما شروع به پیش‌بینی‌های اشتباه بیشتر و بیشتری می‌کند. مدلی با بایاس پایین و واریانس بالا نقاطی را پیش‌بینی می‌کند که به‌طور کلی در اطراف مرکز هستند، اما بسیار دور از یکدیگر هستند. مدلی با بایاس بالا و واریانس پایین بسیار دور از مرکز است، اما از آنجایی که واریانس پایین است، نقاط پیش‌بینی‌شده به یکدیگر نزدیک‌تر هستند. به عبارت دیگر، اگر واریانس افزایش یابد، داده‌ها بیشتر پخش می‌شوند که منجر به دقت کمتر می‌شود (همان‌طور که در دایره بالا سمت راست در تصویر مشاهده می‌شود). در مقابل اگر بایاس افزایش یابد، خطای محاسبه شده افزایش می‌یابد (همان‌طور که در دایره پایین سمت چپ در تصویر مشاهده می‌شود).

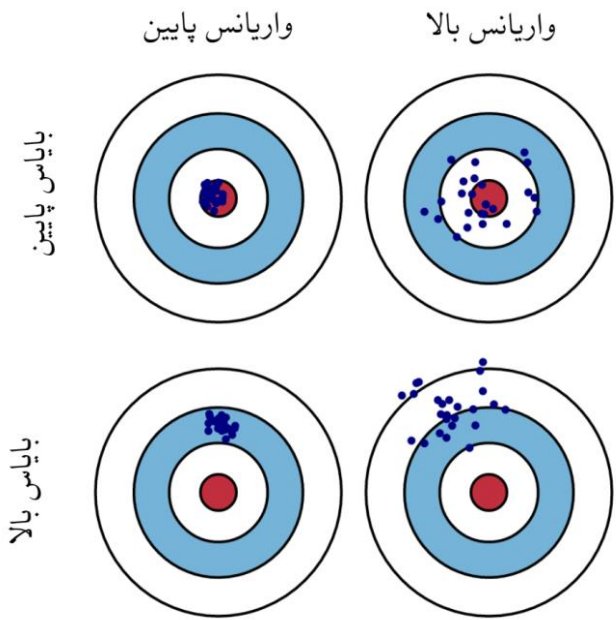

زمانی که الگوریتم یادگیری دارای مشکل بایاس بالا باشد، کار بر روی کاهش بایاس باعث بالا رفتن واریانس می‌شود و مشکل بیش‌برازش را ایجاد می‌کند و هنگامی که الگوریتم یادگیری از مشکل واریانس بالا رنج می‌برد، کار بر روی کاهش واریانس باعث بالا رفتن بایاس می‌شود و مشکل کم‌برازش را ایجاد می‌کند. اینجاست که اصطلاح "موازنه" بوجود می‌آید، چرا که صرفا کاهش بایاس، مدل را بهبود نمی‌بخشد و بالعکس. "نقطه مطلوب" قرار دادن نقاط داده در مکانی است که در آن بایاس بهینه و واریانس بهینه وجود دارد. به‌طور اساسی، با استفاده نکردن از هیچ یک از موارد طرفین که دقت را تغییر می‌دهد، یک الگو پیدا کنید. اغلب اوقات، برنامه‌ریزی و انتخاب این نقاط بزرگ‌ترین چالشی است که دانشمندان و تحلیل‌گران داده با آن روبرو هستند. با این حال، روش‌هایی برای آزمایش برازش مدل وجود دارد. برخی از راه حل‌های ارائه شده برای مقابله با این پدیده عبارتند از:

- **ساخت مدل پیچیده‌تر:** اولین و ساده‌ترین راه‌حل برای یک مشکل کم‌برازش، آموزش یک مدل پیچیده‌تر برای رفع مشکل است و برای مدل بیش‌برازش‌شده، داده‌های بیشتری را وارد کنید.

- **نویز گرادیان[1]:** این روش شامل اضافه کردن نویز گرادیان در طول آموزش است، روشی که ثابت کرد دقت یک مدل را افزایش داده است[*].

- **منظم‌سازی:** روش محبوبی برای کاهش پدیده بیش‌برازش است. این تکنیک که برای حل مشکل واریانس بالا استفاده می‌شود، شامل جریمه کردن ضرایب و وزن‌ها می‌شود تا دقت بالاتری هم برای داده‌های آموزشی و هم برای داده‌های آزمون بدست آید.

- **هنجارسازی:** مسائل مربوط به بیش‌برازش، کم‌برازش را حل می‌کند.

به‌طور خلاصه:

- بایاس فرضیات ساده‌کننده‌ای است که توسط مدل ایجاد می‌شود (به دلیل اینکه داده‌ها را نادیده می‌گیرد) تا تابع هدف را آسان‌تر تقریب بزند.
- مدلی با خطای واریانس بالا، سبب بیش‌برازش می‌شود و از آن چیزهای زیادی یاد می‌گیرد.
- اگر از یک مدل یادگیری ماشین ساده استفاده کنید، مدل دارای بایاس بالا و واریانس کم (کم‌برازش داده‌ها) خواهد بود.
- اگر یک مدل یادگیری ماشین پیچیده استفاده کنید، آنگاه دارای واریانس بالا و بایاس کم (بیش‌برازش داده‌ها) خواهد بود.

---

[1] Gradient Noise

[*] به این مقاله مراجعه کنید:
*Adding Gradient Noise Improves Learning for Very Deep Networks*: https://arxiv.org/abs/1511.06807

▪ شما باید تعادل خوبی بین بایاس و واریانس در مدلی که از آن استفاده کرده‌اید، پیدا کنید. این تعادل همان چیزی است که به عنوان موازنه بایاس و واریانس شناخته می‌شود.

## روش‌های ارزیابی

به طور کلی، ما می‌توانیم خطای تعمیم را از طریق آزمایش‌ها بر روی داده‌های آزمون ارزیابی کنیم. برای انجام این کار، از یک مجموعه آزمایشی برای تخمین توانایی یادگیرنده برای طبقه‌بندی نمونه‌های جدید استفاده می‌کنیم و از خطای آزمون به عنوان تقریبی برای خطای تعمیم استفاده می‌کنیم. در اینجا، ما فقط خطای تعمیم را در نظر می‌گیریم، اما در برنامه‌های کاربردی دنیای واقعی، اغلب عوامل بیشتری مانند هزینه محاسباتی، هزینه حافظه و غیره نیز را در نظر می‌گیریم. معمولا، ما فرض می‌کنیم که نمونه‌های آزمایشی مستقل هستند. باید توجه داشت که مجموعه آزمون و مجموعه آموزشی باید تا حد امکان اختصاصی باشند، یعنی نمونه‌های آزمون باید در مجموعه آموزشی ظاهر نشده و یا به هر نحوی در فرآیند آموزش استفاده نشوند.

چرا نمونه‌های آزمایشی باید در مجموعه آموزشی ظاهر نشوند؟ برای درک این موضوع، اجازه دهید سناریوی زیر را در نظر بگیریم. فرض کنید از مجموعه ده سوالی هم برای تمرین و هم برای امتحان استفاده می‌کنیم، آیا امتحان نتایج یادگیری دانش‌آموزان را منعکس می‌کند؟ پاسخ «خیر» است. چراکه برخی از دانش‌آموزان می‌توانند نمرات خوبی کسب کنند حتی اگر تنها قادر باشند آن ده سوال را حل کنند. به‌طور مشابه، توانایی تعمیمی که ما آرزو داریم مدل داشته باشد همان است که ما می‌خواهیم دانش‌آموزان مطالعه کنند و بر دانش مسلط شوند. بر این اساس، نمونه‌های آموزشی با تمرین‌ها و نمونه‌های آزمایشی مطابق با آزمون است. از این رو، اگر نمونه‌های آزمایشی قبلا در فرآیند آموزش دیده شده باشند، تخمین می‌تواند بسیار خوش‌بینانه باشد.

با این حال، با توجه به تنها یک مجموعه داده $D = \{(x_1, y_1), (x_2, y_2), ... (x_m, y_m)\}$ از نمونه‌ها، چگونه می‌توانیم هم آموزش و هم آزمون انجام دهیم؟ پاسخ این است که هم یک مجموعه آموزشی S و هم یک مجموعه آزمایشی $T$ از مجموعه داده‌های $D$ تولید کنیم. ما چند روش رایج مورد استفاده در این خصوص را در ادامه تشریح خواهیم کرد.

### روش نگهدارنده[1]

روش نگهدارنده، مجموعه داده $D$ را به دو زیرمجموعه مجزا تقسیم می‌کند: یکی به عنوان مجموعه آموزشی $S$ و دیگری به عنوان مجموعه آزمایشی $T$ که در آن:

---

[1] Hold-out Method

$$S \cap T = \emptyset \text{ و } D = S \cup T$$

ما یک مدل را روی مجموعه آموزشی $S$ آموزش می‌دهیم و سپس خطای آزمون را روی مجموعه آزمایشی $T$ به عنوان تخمین خطای تعمیم محاسبه می‌کنیم. به عنوان مثال با در نظر گرفتن یک مساله دسته‌بندی دودویی، فرض کنید $D$ یک مجموعه داده با ۱۰۰۰ نمونه باشد که آن را به یک مجموعه آموزشی $S$ با ۷۰۰ نمونه و یک مجموعه آزمایشی $T$ با ۳۰۰ نمونه تقسیم می‌کنیم. پس از آموزش روی $S$ فرض کنید مدل ۹۰ نمونه را در $T$ به اشتباه دسته‌بندی کرده است، از این‌رو نرخ خطا ۳۰٪=۱۰۰٪ × (۳۰۰/۹۰) را داریم و بر این اساس دقت برابر ۷۰٪=۳۰٪−۱ می‌باشد.

شایان ذکر است که تقسیم باید توزیع اصلی داده را حفظ کند تا از ایجاد بایاس اضافی جلوگیری شود. به عنوان مثال با در نظر گرفتن یک مساله دسته‌بندی، باید سعی کنیم نسبت کلاس را در زیر مجموعه‌های مختلف حفظ کنیم. روش‌های نمونه‌گیری که نسبت کلاس را حفظ می‌کنند، **نمونه‌گیری طبقه‌بندی‌شده**[1] نامیده می‌شود. برای مثال، فرض کنید یک مجموعه داده $D$ داریم که شامل ۵۰۰ نمونه مثبت و ۵۰۰ نمونه منفی است و می‌خواهیم آن را به مجموعه آموزشی $S$ با ۷۰ درصد از نمونه‌ها و مجموعه آزمایشی $T$ را با ۳۰ درصد از نمونه‌ها تقسیم کنیم. از این‌رو، یک روش نمونه‌گیری طبقه‌بندی‌شده تضمین می‌کند که $S$ شامل ۳۵۰ نمونه مثبت و ۳۵۰ نمونه منفی و $T$ شامل ۱۵۰ نمونه مثبت و ۱۵۰ نمونه منفی است. بدون نمونه‌گیری طبقه‌بندی‌شده، نسبت‌های کلاس مختلف در $S$ و $T$ می‌تواند منجر به تخمین خطای بایاس[2] شود زیرا توزیع‌های داده را تغییر می‌کند. با این حال، روش‌های مختلف تقسیم‌بندی منجر به مجموعه‌های آموزشی و آزمایشی متفاوت و بر این اساس، نتایج ارزیابی متفاوتی را نیز در پی خواهند داشت. از این‌رو، یک آزمایش منفرد معمولا به تخمین خطای غیرقابل اعتماد منجر می‌شود.

در عمل، ما اغلب چندین بار آزمون نگه‌دارنده را انجام می‌دهیم، که در آن در هر آزمایش داده‌ها به‌طور تصادفی تقسیم می‌شوند و از میانگین خطا به عنوان تخمین نهایی استفاده می‌کنیم. به عنوان مثال، می‌توانیم مجموعه داده‌ها را به‌طور تصادفی ۱۰۰ بار تقسیم کرده تا ۱۰۰ نتیجه ارزیابی تولید کنیم و سپس میانگین را به عنوان تخمین خطای نگه‌دارنده در نظر بگیریم.

روش نگه‌دارنده $D$ را به یک مجموعه آموزشی و یک مجموعه آزمایشی تقسیم می‌کند، اما مدلی که ما می‌خواهیم ارزیابی کنیم، مدلی است که روی $D$ آموزش داده شده است. بنابراین، ما با یک دوراهی روبرو هستیم. اگر بیشتر نمونه‌ها را در مجموعه آموزشی $S$ قرار دهیم، مدل

---

[1] Stratified sampling

[2] biased error estimation

آموزش‌دیده، تخمینی عالی برای مدل آموزش‌دیده شده روی $D$ است. با این حال، ارزیابی به‌دلیل اندازه کوچک $T$ کم‌تر قابل اعتماد است. از سوی دیگر، اگر تعداد بیشتری از نمونه‌ها را در مجموعه آزمون $T$ قرار دهیم، تفاوت بین مدل آموزش‌داده شده در $S$ و مدل آموزش‌داده شده در $D$ قابل توجه می‌شود، یعنی درستی ارزیابی کم‌تر می‌شود. هیچ راه‌ حل کاملی برای این معضل وجود ندارد و ما باید یک موازنه انجام دهیم. یک روال معمول این است که از $\frac{2}{3}$ تا $\frac{4}{5}$ نمونه‌ها برای آموزش و از بقیه آن‌ها برای آزمون استفاده شود.

## روش اعتبارسنجی متقابل[1]

اعتبارسنجی متقابل مجموعه داده $D$ را به $k$ زیرمجموعه‌ی مجزا با اندازه‌های مشابه تقسیم می‌کند، یعنی:

$$D = D_1 \cup D_2 \cup \cdots \cup D_K \,, D_i \cap D_j = \emptyset \; (i \neq j)$$

معمولا، هر زیرمجموعه $D_i$ سعی می‌کند توزیع اصلی داده را از طریق نمونه‌گیری طبقه‌بندی‌شده حفظ کند. چندین روش برای تقسیم داده‌ها برای اعتبار سنجی متقابل وجود دارد. در روش اعتبارسنجی متقابل چندبخشی، $K-1$ زیرمجموعه به عنوان مجموعه آموزشی برای آموزش یک مدل و از زیر مجموعه باقی‌مانده به عنوان مجموعه آزمایشی برای ارزیابی مدل استفاده می‌شود. این فرآیند را $K$ بار تکرار می‌کنیم و از هر زیر مجموعه به عنوان مجموعه آزمون دقیقا یک بار استفاده می‌کنیم. در نهایت، برای بدست آوردن نتیجه ارزیابی، از $K$ آزمایش میانگین می‌گیریم. متداول‌ترین مقدار استفاده شده از برای $K$ مقدار ۱۰ است. سایر مقادیر رایج $K$ شامل ۵ و ۲۰ است.

> در تکنیک اعتبارسنجی متقابل چند-بخش هر یک از داده‌ها دقیقا یک بار در یک مجموعه آزمایش و یک بار در یک مجموعه آموزش قرار می‌گیرد. این امر به‌طور قابل توجهی بایاس و واریانس را کاهش می‌دهد، چراکه تضمین می‌کند هر نمونه‌ای از مجموعه داده اصلی این شانس را دارد که در مجموعه آموزشی و آزمایشی ظاهر شود. اگر داده‌های ورودی محدودی داشته باشیم، اعتبارسنجی متقابل چند-بخش از بهترین روش‌ها برای ارزیابی کارآیی یک مدل است.

## بوت‌استرپینگ

چیزی که ما می‌خواهیم ارزیابی کنیم مدلی است که با $D$ آموزش داده می‌شود. با این حال، حتی اگر از تکنیک نگهدارنده یا اعتبارسنجی متقابل استفاده کنیم، بازهم مجموعه آموزشی همیشه

---

[1] Cross-Validation

کوچکتر از D خواهد بود. از این‌رو، تخمین بایاس[1] به دلیل اختلاف اندازه بین مجموعه آموزشی و D اجتناب‌ناپذیر است. یک راه حل بوت‌استرپینگ است که از تکنیک نمونه‌گیری بوت‌استرپ استفاده می‌کند. با توجه به یک مجموعه داده $D$ حاوی $m$ نمونه، بوت‌استرپ یک مجموعه داده $Ď$ با انتخاب تصادفی یک نمونه از $D$ و رونوشت کردن آن در $Ď$ و سپس قرار دادن آن در $D$ به‌طوری که هنوز فرصتی برای انتخاب شدن در مرتبه بعدی را داشته باشد، نمونه‌برداری می‌کند. تکرار این فرآیند با $m$ مرتبه منجر به مجموعه داده‌های نمونه‌برداری بوت‌استرپ $Ď$ شامل $m$ نمونه می‌شود. به دلیل جایگزینی، برخی از نمونه‌ها در $D$ ممکن است در $Ď$ ظاهر نشوند، در حالی که برخی دیگر ممکن است بیش از یک بار ظاهر شوند. اگر یک تخمین سریع انجام دهیم: شانس انتخاب نشدن $m$ نوبت برابر با $(۱ - \frac{۱}{m})^m$ است و از این‌رو با گرفتن حد داریم:

$$\lim_{m \to \infty} \left(۱ - \frac{۱}{m}\right)^m = \frac{۱}{e} \approx ۰.۳۶۸$$

به این معنی که تقریباً ۳۶٫۸٪ از نمونه‌های اصلی در مجموعه داده $Ď$ ظاهر نمی‌شوند. از این‌رو، می‌توانیم از $Ď$ به عنوان مجموعه آموزشی و $D \backslash Ď$ به عنوان مجموعه آزمایشی استفاده کنیم، به طوری که هم مدل ارزیابی‌شده و هم مدل واقعی‌ای که می‌خواهیم روی $D$ ارزیابی کنیم، از نمونه‌های آموزشی $m$ استفاده می‌کنند. علاوه بر این، ما هنوز یک مجموعه آزمون جداگانه داریم که شامل حدود $\frac{۱}{e}$ از نمونه‌های اصلی است که برای آموزش استفاده نمی‌شود. نتیجه ارزیابی بدست آمده از این روش **تخمین خارج از کیسه**[2] نامیده می‌شود.

<div style="background:#1ba0e2;color:#fff;padding:1em;">
بوت‌استرپینگ زمانی مفید است که مجموعه داده کوچک باشد یا زمانی که هیچ روش مؤثری برای تقسیم مجموعه‌های آموزش و آزمایش وجود ندارد. علاوه بر این، بوت‌استرپینگ می‌تواند مجموعه داده‌های متعددی را ایجاد کند که می‌تواند برای روش‌هایی مانند یادگیری گروهی مفید باشد. با این وجود، از آنجایی که توزیع داده‌های اصلی با بوت‌استرپ تغییر کرده است، تخمین نیز دارای بایاس است. بنابراین، زمانی که داده‌های فراوانی داریم، اغلب به جای آن از نگه‌دارنده و اعتبارسنجی متقابل استفاده می‌شود.
</div>

## تنظیم ابرپارامترها و مدل نهایی

اکثر الگوریتم‌های یادگیری دارای ابرپارامترهایی برای تنظیم هستند و تنظیم ابرپارامترهای مختلف اغلب به مدل‌هایی با عملکرد متفاوت منجر می‌شوند. از این‌رو، ارزیابی و انتخاب مدل

---

[1] estimation bias

[2] out-of-bag estimate

فقط در مورد انتخاب الگوریتم‌های یادگیری نیست، بلکه در مورد تنظیم ابرپارامترها نیز می‌باشد. فرآیند یافتن ابرپارامترهای مناسب را **تنظیم ابرپارامترها**[1] می‌گویند. ابرپارامترها آرگومان‌های مدل هستند که مقدار آن‌ها قبل از شروع فرآیند یادگیری تنظیم می‌شود. *کلید الگوریتم‌های یادگیری ماشین، تنظیم ابرپارامترها است.* به عبارت دیگر، تنظیم ابرپارامترها فرآیند تعیین ترکیب مناسبی از ابرپارامترها است که به مدل اجازه می‌دهد تا عملکرد مدل را به حداکثر برساند. تنظیم ترکیب صحیح ابرپارامترها تنها راه استخراج حداکثری عملکرد از مدل‌ها است.

خوانندگان ممکن است فکر کنند هیچ تفاوت اساسی بین تنظیم ابرپارامتر و انتخاب الگوریتم وجود ندارد: هر تنظیم ابرپارامتر به یک مدل منتهی می‌شود و ما مدلی را که بهترین نتایج را ایجاد می‌کند به عنوان مدل نهایی انتخاب می‌کنیم. این ایده اساسا صحیح است. با این حال، یک مشکل وجود دارد، از آنجایی که پارامترها اغلب مقدار حقیقی[2] دارند، آزمایش تمام تنظیمات ابرپارامترها غیرممکن است. از این‌رو، در عمل ما معمولا یک محدوده و یک اندازه گام را برای هر ابرپارامتر تعیین می‌کنیم. به عنوان مثال، محدوده [۰٫۲ ، ۰] و اندازه گام ۰٫۰۵ ، که منجر به تنها پنج تنظیم ابرپارامتر کاندید می‌شود. چنین هم‌سنجی[3] بین هزینه محاسباتی و کیفیت تخمین، یادگیری را امکان‌پذیر می‌کند، اگرچه تنظیم ابرپارامتر انتخاب شده معمولا بهینه نیست. در واقعیت، حتی پس از انجام چنین هم‌سنجی، تنظیم ابرپارامتر هنوز می‌تواند بسیار چالش‌برانگیز باشد. می‌توانیم یک برآورد ساده‌ای انجام دهیم. فرض کنید که الگوریتم دارای سه ابرپارامتر است و هر کدام از آن‌ها تنها پنج مقدار کاندید را در نظر می‌گیرند، از این‌رو باید برای هر جفت مجموعه آموزشی و آزمایشی $۵^۳ = ۱۲۵$ مدل را ارزیابی کنیم. الگوریتم‌های یادگیری قدرتمند اغلب دارای ابرپارامترهای بسیار زیادی برای تنظیم هستند که منجر به حجم کاری سنگینی از تنظیم پارامترها می‌شود.

کیفیت تنظیم ابرپارامترها اغلب در برنامه‌های کاربردی دنیای واقعی حیاتی است. با این حال، انتخاب ترکیب مناسب از ابرپارمترها کار آسانی نیست. به‌طورکلی دو راه عمده برای تنظیم آن‌ها وجود دارد:

▪ **تنظیم دستی ابرپارامتر:** در این روش، ترکیب‌های مختلف ابرپارامترها به صورت دستی تنظیم می‌شوند و آزمایش می‌شوند. این یک یک فرآیند خسته کننده است و در مواردی که ابرپارامترهای زیادی برای امتحان وجود دارد، نمی‌تواند عملی باشد. به عبارت دیگر، هر بار که ما ابرپارامترهای مختلف را امتحان می‌کنیم، باید یک مدل را روی داده‌های آموزشی، آموزش دهیم، روی داده‌های اعتبارسنجی پیش‌بینی کنیم و

---

[1] Hyperparameter Tuning

[2] real-valued

[3] trade off

سپس معیار اعتبارسنجی را محاسبه کنیم. این کار با تعداد زیادی ابرپارامتر در مدل‌های پیچیده‌ای همانند یادگیری گروهی یا شبکه‌های عصبی عمیق که ممکن است فرآیند یادگیری چند روز طول بکشد، فرآیند دستی را غیرقابل حل می‌کند!

■ **تنظیم خودکار ابرپارامتر:** در این روش با استفاده از الگوریتمی که فرآیند را خودکار و بهینه می‌کند، ابرپارامترهای بهینه پیدا می‌شوند. جستجوی تصادفی، جستجوی شبکه و بهینه‌سازی بیزی نمونه‌ای تنظیم خودکار ابرپارامتر هستند.

## جستجوی شبکه‌ای

در روش جستجوی شبکه‌ای، شبکه‌ای از مقادیر ممکن برای ابرپارامترها ایجاد می‌کنیم. هر تکرار ترکیبی از ابرپارامترها را به ترتیب خاصی امتحان می‌کند. این مدل را بر روی هر ترکیبی از ابرپارامترهای ممکن برازش می‌دهد و عملکرد مدل را ثبت می‌کند. در نهایت، بهترین مدل را با بهترین ابرپارامترها برمی‌گرداند. از آنجایی که این روش تمام ترکیبات ابرپارامترها را امتحان می‌کند، از این‌رو پیچیدگی زمانی محاسبات را افزایش می‌دهد.

## جستجوی تصادفی

یک روش ساده برای جایگزینی جستجوی شبکه‌ای، نمونه‌گیری از فضای ابرپارامترها به‌صورت تصادفی است. به عبارت دیگر، به جای آزمایش‌های مرتب در تمام مجموعه مقادیر در فضای مسئله، بهتر است مقادیر تصادفی از کل فضای نمونه انتخاب و آزمایش شوند. در روش جستجوی تصادفی، شبکه‌ای از مقادیر ممکن برای ابرپارامترها ایجاد می‌کنیم. هر تکرار ترکیبی تصادفی از ابرپارامترها را از این شبکه امتحان می‌کند، عملکرد را ثبت می‌کند و در نهایت ترکیبی از ابرپارامترها را که بهترین عملکرد را ارائه می‌دهند، برمی‌گرداند.

## بهینه‌سازی بیزی

تنظیم و یافتن ابرپارامترهای مناسب برای مدل، یک مساله بهینه‌سازی است. به عبارت دیگر، ما می‌خواهیم با تغییر ابرپارامترهای مدل، تابع ضرر مدل خود را به حداقل برسانیم. بهینه‌سازی بیزی به ما کمک می‌کند تا نقطه کمینه را در حداقل تعداد گام‌ها پیدا کنیم. بهینه‌سازی بیزی همچنین از یک تابع فراگیری[1] استفاده می‌کند که نمونه‌گیری را به مناطقی هدایت می‌کند که در آن بهبود نسبت به بهترین مشاهدات فعلی محتمل است. به‌طور کلی، مفهوم اصلی بهینه‌سازی بیزی این است که: *"اگر به‌طور تصادفی برخی از نقاط را جستجو کردیم و بدانیم که برخی از این نقاط امیدوارکننده‌تر از سایرین هستند، چرا نگاهی به آن‌ها نیاندازیم؟"* .

---

بهینه‌سازی بیزی به‌طور معمول در رسیدن به مجموعه مطلوب مقادیر ابرپارامترها به تکرار کم‌تری نیاز دارد، چرا که مناطقی از فضای پارامترها که به عقیده وی هیچ کمکی نمی‌کند را نادیده می‌گیرد.

روش‌های جستجوی شبکه‌ای و جستجوی تصادفی نیز نسبتا ناکارآمد هستند. چراکه ابرپارامترهای بعدی را برای ارزیابی بر اساس نتایج قبلی انتخاب نمی‌کنند. جستجوی شبکه‌ای و تصادفی کاملا از ارزیابی‌های گذشته بی‌اطلاع است و در نتیجه، اغلب زمان قابل توجهی را صرف ارزیابی ابرپارامترهای «بد» می‌کنند.

# ارزیابی کارآیی

هر زمان که یک مدل یادگیری ماشین ایجاد می‌کنید، همه مخاطبان از جمله سهام‌داران کسب‌وکار تنها یک سوال دارند، کارآیی مدل چگونه است؟ معیارهای ارزیابی مدل کدامند؟ دقت یک مدل چیست؟ به عبارت دیگر، برای ارزیابی قابلیت تعمیم مدل‌ها، نه تنها به روش‌های تخمین موثر و کارآمد نیاز داریم، بلکه به برخی معیارهای کارآیی نیز نیاز داریم که بتوانند توانایی تعمیم را تعیین کنند. معیارهای کارآیی مختلف منعکس کننده خواسته‌های متنوع مسائل هستند و نتایج ارزیابی متفاوتی را تولید می‌کند. به عبارت دیگر، کیفیت یک مدل یک مفهوم نسبی است که به الگوریتم، داده‌ها و همچنین نیاز کار بستگی دارد.

ارزیابی مدل توسعه‌یافته به شما کمک می‌کند مدل را اصلاح کنید. از این‌رو، شما به توسعه و ارزیابی مدل خود ادامه می‌دهید تا زمانی که به سطح کارآیی مدل بهینه برسید (کارآیی بهینه مدل به معنای دقت ۱۰۰ درصدی نیست!!). بسیاری از تحلیل‌گران داده را می‌توان دید که به کارآیی مدل یا معیارهای ارزیابی مدل اهمیتی نمی‌دهند. شما می‌توانید $n$ تعداد مدل را برای یک مجموعه داده خاص ایجاد کنید، اما اینکه کدام مدل باید انتخاب شود، سوال اصلی است؟ و *معیارهای ارزیابی مدل‌ها* جواب این سوال هستند.

در مسائل پیش‌بینی، یک مجموعه داده $D = \{(x_1, y_1), (x_2, y_2), \dots (x_m, y_m)\}$ که در آن $y_m$ برچسب واقعی برای نمونه $x_m$ است را داریم. برای ارزیابی کارآیی یک یادگیرنده $f$ پیش‌بینی $f(x)$ آن را با برچسب حقیقی $y$ آن مقایسه می‌کنیم.

با توجه به هدف و دامنه کسب و کار خود، می‌توانید معیارهای ارزیابی مدل را انتخاب کنید. وقتی در مورد مدل‌های پیش‌گویانه صحبت می‌کنیم، ابتدا باید انواع مختلف مدل‌های پیش‌گویانه را درک کنیم. به‌طور کلی دو نوع مدل بر اساس متغیرهای وابسته داریم. اگر متغیر وابسته پیوسته باشد، یک مدل رگرسیونی ایجاد می‌کنیم و زمانی که متغیر وابسته گسسته است، یک مدل

دسته‌بندی داریم. اگر مساله از نوع مسائل رگرسیون باشد، رایج‌ترین معیار کارآیی مورد استفاده، میانگین مربعات خطا[1] (MSE) است:

$$E(f; D) = \frac{1}{m} \sum_{i=1}^{m} (f(x_i) - y_i)^2$$

در ادامه این بخش به معرفی و تشریح برخی از معیارهای کارآیی در مسائل دسته‌بندی می‌پردازیم.

## نرخ خطا و دقت

متداول‌ترین معیارهای کارآیی در مسائل دسته‌بندی، از جمله دسته‌بندی دودویی و دسته‌بندی چندگانه (چندکلاسه)، نرخ خطا[2] و دقت[3] هستند. نرخ خطا نسبت نمونه‌های دسته‌بندی اشتباه به همه نمونه‌ها است. در حالی‌که دقت، نسبت نمونه‌های دسته‌بندی درست نسبت به همه نمونه‌ها است. به عنوان مثال، اگر ۱۰۰ نمونه از $x_1$ تا $x_{100}$ وجود داشته باشد و مدل (دسته‌بند) ۲۰ نمونه را به‌درستی دسته‌بندی کند، آن‌گاه نرخ خطا ۰/۲ = ۲۰/۱۰۰ خواهد بود. با توجه به مجموعه داده $D$، نرخ خطا را به‌صورت زیر:

$$E(f; D) = \frac{1}{m} \sum_{i=1}^{m} I(f(x_i) \neq y_i)$$

و دقت را به‌صورت زیر تعریف می‌کنیم:

$$Accuracy(f; D) = \frac{1}{m} \sum_{i=1}^{m} I(f(x_i) = y_i)$$

$$= 1 - E(f; D)$$

که در آن $I(.)$ تابع نشانگر[4] است که ۱ را برای درست و ۰ در غیر این صورت بر می‌گرداند.

## صحت، فراخوانی و F1

زمانی که باید یک مدل را ارزیابی کنیم، در اغلب اوقات از نرخ خطا و دقت استفاده می‌کنیم، اما چیزی که عمدتا روی آن تمرکز می کنیم این است که مدل ما چقدر قابل اعتماد است، چگونه

---

[1] Mean Squared Error

[2] Error Rate

[3] Accuracy

[4] indicator function

روی یک مجموعه داده متفاوت عمل می‌کند (قابلیت تعمیم) و چقدر انعطاف‌پذیری دارد. بدون شک دقت معیار بسیار مهمی است که باید در نظر گرفته شود، اما همیشه تصویر کاملی را از کارآیی مدل ارائه نمی‌دهد.

وقتی می‌گوییم مدل قابل اعتماد است، منظور ما این است که مدل از داده‌ها به‌درستی و مطابق خواسته یادگیری بدست آورده است. بنابراین، پیش‌بینی‌های انجام شده توسط آن به مقادیر واقعی نزدیک است. در برخی موارد، مدل ممکن است منجر به دقت بهتری شود، اما نمی‌تواند داده‌ها را به‌درستی درک کند و بنابراین زمانی که داده‌ها متفاوت هستند، عملکرد ضعیفی دارد. این بدان معنی است که مدل به اندازه کافی قابل اعتماد و قوی نیست و از این رو استفاده از آن را محدود می‌کند.

به عنوان مثال، ما ۹۸۰ سیب و ۲۰ پرتقال داریم و یک مدل داریم که هر میوه را به عنوان یک سیب دسته‌بندی می‌کند. از این‌رو دقت مدل ۹۸٪=۹۸۰/۱۰۰۰ است و بر اساس معیار دقت، ما یک مدل بسیار دقیق داریم. با این حال، اگر ما از این مدل برای پیش‌بینی میوه‌ها در آینده استفاده کنیم، با شکست مواجه خواهیم شد. چراکه این مدل تنها می‌تواند یک کلاس را پیش‌بینی کند.

دریافت تصویری کامل از مدل، به عنوان مثال اینکه چگونه داده‌ها را درک می‌کند و چگونه می‌تواند پیش‌بینی کند، به درک عمیق ما از مدل کمک می‌کند و به بهبود آن کمک می‌کند. بنابراین، فرض کنید مدلی را ایجاد کرده‌اید که دقت ۹۰٪ را بدست می‌آورد، از این‌رو چگونه می‌خواهید آن را بهبود ببخشید؟ برای تصحیح یک اشتباه، ابتدا باید متوجه آن اشتباه شویم. به‌طور مشابه، برای بهبود مدل ما باید به نحوهٔ عملکرد مدل در سطح عمیق‌تری نگاه کنیم. با این حال، این کار تنها با نگاه کردن به معیار دقت بدست نمی‌آید و از این‌رو معیارهای دیگری نیز در نظر گرفته می‌شود. معیارهایی همانند، صحت[1]، فراخوانی[2] و F1 نمونه‌هایی از این معیارها هستند.

برای محاسبه معیارهای ارزیابی کارآیی یک مدل دسته‌بندی نیاز به چهار ترکیب از کلاس واقعی و کلاس پیش‌بینی با عناوین، مثبت راستین، منفی کاذب، مثبت کاذب، منفی راستین و منفی کاذب داریم که می‌توان آن‌ها را در یک ماتریس درهم‌ریختگی[3] نشان داد (جدول ۵ ـ ۱). جایی که:

• **مثبت راستین($TP$):** به عنوان مثال، وقتی مقدار واقعی کلاس "بله" بود، مدل هم "بله" را پیش‌بینی کرد (یعنی پیش‌بینی درست)

---

[1] precision

[2] recall

[3] Confusion Matrix

- **مثبت کاذب($FP$):** به عنوان مثال، یعنی زمانی که مقدار واقعی کلاس "نه" بود، اما مدل "بله" را پیش‌بینی کرد (یعنی پیش‌بینی اشتباه)
- **منفی کاذب($FN$):** به عنوان مثال، زمانی که مقدار واقعی کلاس "**بله**" بود، اما مدل "نه" را پیش‌بینی کرد (یعنی پیش‌بینی اشتباه)
- **منفی راستین($TN$):** به عنوان مثال، یعنی زمانی که مقدار واقعی کلاس "**نه**" بود و مدل هم "**نه**" را پیش‌بینی کرد (یعنی پیش‌بینی درست).

**جدول ۵ ــ ۱ ماتریس درهم‌ریختگی**

|  |  | کلاس پیش‌بینی‌شده | |
|---|---|---|---|
|  |  | **مثبت** | **منفی** |
| **مثبت** |  | مثبت راستین (TP) | منفی کاذب (FN) |
| **منفی** | کلاس واقعی | منفی کاذب (FP) | منفی راستین (TN) |

حال، بر این اساس می‌توانیم معیارهای صحت، فراخوانی و F1 را تعریف کنیم:

- **فراخوانی.** به توانایی یک مدل در پیش‌بینی موارد مثبت از کل موارد مثبت راستین اشاره می‌کند:

$$\text{فراخوانی} = \frac{TP}{FN + TP}$$

- **صحت.** کسری از موارد مثبت راستین را در بین نمونه‌هایی که به عنوان مثبت پیش‌بینی شده‌اند، نشان می‌دهد:

$$\text{صحت} = \frac{TP}{FP + TP}$$

● **F1**. اگر نیاز به تعادل بین صحت و فراخوانی دارید F1 معیار بهتری برای استفاده است. به زبان ساده، F1 صحت و فراخوانی را با محاسبه میانگین هارمونیک بین این دو، در یک معیار ترکیب می‌کند:

$$F1 = ۲ \times \frac{\text{فراخوانی} * \text{صحت}}{\text{فراخوانی} + \text{صحت}}$$

## حساسیت[1] و ویژه‌مندی[2]: بر پیش‌بینی‌های راستین تمرکز کنید

ماهیت حساسیت و ویژه‌مندی این است که هر دو بر نسبت پیش‌بینی‌های درست تمرکز می‌کنند. بنابراین، صورت همیشه معیاری برای پیش‌بینی‌های درست و مخرج همیشه کل پیش‌بینی‌های متناظر آن کلاس است. در حالی‌که حساسیت نسبت مثبت‌های پیش‌بینی‌شده راستین را از تمام مقادیر مثبت واقعی اندازه‌گیری می‌کند، ویژه‌مندی، نسبت منفی‌های پیش‌بینی‌شده راستین را از تمام مقادیر منفی واقعی اندازه‌گیری می‌کند. ویژه‌مندی را همچنین **نرخ منفی راستین** و حساسیت را **نرخ مثبت راستین** می‌نامند.

حساسیت را با منفی کاذب مقابله می‌کند. حساسیت بالا به معنای *نرخ پایین منفی کاذب است*. به عبارت دیگر، حساسیت نسبت موارد مثبت راستین است که به درستی پیش‌بینی شده است:

$$\text{حساسیت} = \text{نرخ منفی کاذب} - ۱ = \frac{TP}{FN + TP}$$

توجه داشته باشید که معادله‌های فراخوانی و حساسیت از نظر ریاضی یکسان هستند.

---

[1] Sensitivity

[2] Specificity

ویژه‌مندی با مثبت کاذب مقابله می‌کند. ویژه‌مندی بالا به معنای نرخ پایین مثبت کاذب است. به عبارت دیگر، ویژه‌مندی نسبت منفی‌های راستین است که به درستی پیش‌بینی شده‌اند:

$$\text{ویژه‌مندی} = 1 - \text{نرخ مثبت کاذب} = \frac{TN}{TN + FP}$$

**ویژه‌مندی در کجا استفاده می‌شود؟**
زمانی که دسته‌بندی منفی‌ها اولویت بالایی دارند.
به عنوان مثال: تشخیص وضعیت سلامتی قبل از معالجه.

# رویکرد برش در مدل دسته‌بندی

هر زمان که مدلی را توسعه می‌دهید، آن مدل به شما احتمال وقوع یا عدم وقوع آن رویداد را می‌دهد. شما می‌توانید یک برش در احتمال بدست آمده ایجاد کنید. به عنوان مثال، اگر احتمال بیش از ۰٫۵۰ باشد، آنگاه پیش‌بینی ۱ و در غیر این صورت پیش‌بینی ۰ است. معمولا همه الگوریتم‌های موجود در هر نرم‌افزار و کتابخانه دارای احتمال پیش فرض ۰٫۵۰ می‌باشد. با این حال، شما می‌توانید این نقطه برش را با توجه به هدف تجاری خود تغییر دهید. *اگر می‌خواهید ریسک بیشتری داشته باشید، می‌توانید برش زیر ۰٫۵۰ را انتخاب کنید و اگر قصد دارید در پیش‌بینی خود از ریسک اجتناب کنید، می‌توانید برش بالای ۰٫۵۰ را انتخاب کنید.*

# خلاصه فصل

- الگوریتم یادگیری ماشین رویه‌ای است که بر روی داده‌ها برای ایجاد یک "مدل" یادگیری ماشین اجرا می‌شود.
- یک "مدل" در یادگیری ماشین خروجی یک الگوریتم یادگیری ماشین است که بر روی داده‌ها اجرا می‌شود.
- هیچ الگوریتم واحدی تمام مسائل یادگیری ماشین شما را بهتر از هر الگوریتم دیگری حل نمی‌کند
- اگر یک مدل یادگیری ماشینی عملکرد بسیار خوبی را در داده‌های آموزشی نشان دهد (خطای آموزش کم) اما هنگام آزمایش روی داده‌های جدید ضعیف عمل کند (خطای آزمون بالا)، این معمولا نشانه‌ای است که مدل دچار بیش‌برازش شده است.
- توانایی یادگیری بیش از حد قوی، یک دلیل رایج برای بیش‌برازش است.
- ابرپارامترها آرگومان‌های مدل هستند که مقدار آن‌ها قبل از شروع فرآیند یادگیری تنظیم می‌شود.
- تنظیم ابرپارامترها فرآیند تعیین ترکیب مناسبی از ابرپارامترها است که به مدل اجازه می‌دهد تا عملکرد مدل را به حداکثر برساند.

# مراجع برای مطالعه بیشتر

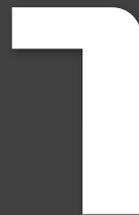

# یادگیری بانظارت

# دسته‌بندی

در یادگیری ماشین، دسته‌بندی به یک مساله مدل‌سازی پیش‌بینی‌کننده اشاره دارد که در آن هر نمونه به یک (یا در برخی مسائل همانند دسته‌بندی چندبرچسبی بیشتر از یک) کلاس اختصاص داده می‌شود. این کلاس‌ها از قبل مشخص هستند و اغلب با عنوان هدف، برچسب یا دسته نامیده می‌شوند. از آنجایی که در فرآیند دسته‌بندی برچسب‌ها وجود دارند، این رویکرد در دسته‌ی یادگیری نظارتی قرار می‌گیرد. به مسائل دسته‌بندی می‌توان از دو دیدگاه متفاوت نگاه کرد، از دیدگاه تعداد برچسب که می‌توان آن را به دو مساله **دسته‌بندی تک‌برچسبی**[1] و **دسته‌بندی چندبرچسبی**[2] تقسیم کرد و از دیدگاه تعداد دسته‌ها به دو مساله **دسته‌بندی دودویی**[3] و **دسته‌بندی چندگانه (چندکلاسی)**[4] تقسیم می‌شود.

دسته‌بندی دودویی که در آن هر نمونه تنها به یکی از دو دسته‌های از پیش تعریف‌شده اختصاص داده می‌شود، ساده‌ترین نوع دسته‌بندی است. دسته‌بندی دودویی با تعریف دسته‌های بیشتر به دسته‌بندی چندگانه گسترش می‌یابد. دسته‌بندی چندبرچسبی حالت تعمیم یافته‌ای از دسته‌بندی تک‌برچسبی است، چراکه در آن هر نمونه می‌تواند به جای یک برچسب با مجموعه‌ای از برچسب‌ها در ارتباط باشد.

**تعریف** دسته‌بندی

دسته‌بندی فرآیند اختصاص متغیرهای ورودی جدید $X$ (براساس مدل دسته‌بندی که از داده‌های آموزشی برچسب‌گذاری شده‌ی قبلی ساخته شده است) به کلاسی است که به احتمال زیاد به آن تعلق دارند

دسته‌بندی سعی در ایجاد ارتباط بین نمونه‌های آموزشی و دسته‌های از پیش تعریف شده برای مساله مورد نظر دارد. داده‌های دارای برچسب برای آموزش دسته‌بند استفاده می‌شود تا بتواند بر روی داده‌های ورودی جدید به‌خوبی عمل کند و بتواند کلاس درست آن نمونه را پیش‌بینی کند. به عبارت دیگر، هدف این است که یک تقریب خوب برای $f(x)$ پیدا شود تا بتواند برای داده‌های دیده‌نشده در فرآیند آموزش پیش‌بینی انجام دهد و بگوید که نمونه جدید به کدام یکی از کلاس‌ها تعلق دارد.

---

[1] single-label classification

[2] multi-label classification

[3] Binary Classification

[4] Multi-Class Classification

## دسته‌بندی تک‌برچسبی

دسته‌بندی تک‌برچسبی (یا سنتی) به‌طور خودکار یک برچسب کلاس را به هر نمونه ورودی اختصاص می‌دهد که در آن یک دسته‌بند یاد می‌گیرد به هر نمونه نادیده‌ای، محتمل‌ترین کلاس یا همان دسته‌اش را مرتبط کند. به طور کلی، مسائل دسته‌بندی تک‌برچسبی را می‌توان به دو گروه اصلی تقسیم کرد: مسائل دودویی و چندگانه.

مساله دسته‌بندی دودویی ساده‌ترین حالت از مسائل دسته‌بندی است که در آن مجموعه کلاس‌ها تنها به دو مورد محدود می‌شود. در این زمینه، ما بین کلاس مثبت و کلاس منفی تمایز قائل می‌شویم. یک مثال ساده از مسائل دسته‌بندی دودویی زمانی است که یک زن به پزشک مراجعه می‌کند تا از باردار بودن خود مطلع شود. نتیجه آزمایش ممکن است مثبت یا منفی باشد.

زمانی که تعداد کلاس‌ها بیشتر از دو باشد، مساله یادگیری را دسته‌بندی چندگانه می‌نامند. فرض بر این است که کلاس‌های هدف مجزا و منحصر به فرد هستند. به عبارت دیگر، هر نمونه دقیقا به یک کلاس تعلق دارد. به عنوان مثال، یک فرد دارای یک گروه خونی در میان چهار نوع A، B، AB یا O است.

### دسته‌بندی دودویی

فرآیندی است که در آن داده‌های ورودی به دو گروه دسته‌بندی می‌شوند. به‌طور اساسی دسته‌بندی دودویی نوعی پیش‌بینی است که به این موضوع می‌پردازد که یک نمونه به کدام یک از دو گروه کلاس تعلق دارد. فرض کنید دو ایمیل برای شما ارسال می‌شود، یکی توسط یک شرکت بیمه‌ای که تبلیغات خود را ارسال می‌کند، ارسال گردیده است و ایمیل دیگر از طرف بانک در مورد صورت‌حساب کارت اعتباری شما. ارائه دهنده خدمات ایمیل دو ایمیل را دسته‌بندی می‌کند، اولین ایمیل به پوشه هرزنامه ارسال می‌شود و ایمیل دوم در ایمیل اصلی ذخیره می‌شود. این فرآیند به‌عنوان دسته‌بندی دودویی شناخته می‌شود، زیرا دو کلاس مجزا وجود دارد یکی اسپم و دیگری اصلی. بنابراین، این یک مساله دسته‌بندی دودویی است. شکل ۱ یک دسته‌بند دودویی را نشان می‌دهد.

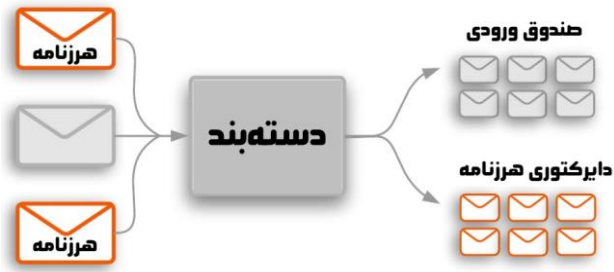

**شکل ۱** دسته‌بندی دودویی (تشخیص هرزنامه)

## دسته‌بندی چندگانه (چندکلاسه)

دسته‌بندی چندگانه یا چندکلاسه، دسته‌بندی عناصر به کلاس‌های مختلف است. برخلاف دسته‌بندی دودویی که محدود به تنها دو کلاس است، محدودیتی در تعداد کلاس‌ها ندارد و می‌تواند دسته‌بندی بیش از دو کلاس را انجام دهد. به عنوان مثال، دسته‌بندی اخبار در دسته‌های مختلف، دسته‌بندی کتاب‌ها براساس موضوع و طبقه‌بندی حیوانات مختلف در یک تصویر نمونه‌هایی از دسته‌بندی چندگانه هستند (شکل ۲ نمونه‌ای از دسته‌بندی چندگانه). با این حال، در حالی‌که دسته‌بند دودویی تنها به یک مدل برای دسته‌بندی نیاز دارد، تعداد مدل مورد استفاده در دسته‌بندی چندگانه بستگی به تکنیک دسته‌بندی دارد. در ادامه دو تکنیک الگوریتم دسته‌بندی چندگانه را شرح خواهیم داد.

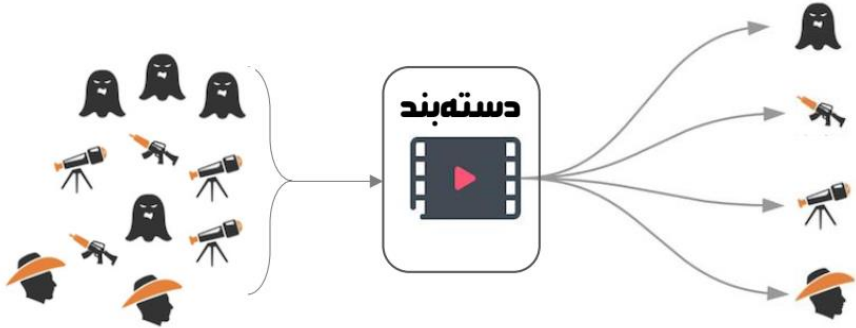

**شکل ۲** دسته‌بندی چندگانه (دسته‌بندی فیلم براساس موضوع)

### یک در مقابل همه (یک در مقابل بقیه)[1]

روش یک در مقابل همه، یک روش ابتکاری استفاده شده از الگوریتم دسته‌بندی دودویی برای دسته‌بندی‌های چندگانه است. این تکنیک شامل تقسیم یک مجموعه داده چند کلاسی به چندین مجموعه از مسائل دودویی است. در دسته‌بندی یک در مقابل همه، برای مجموعه داده با $N$ کلاس، باید $N$ دسته‌بند دودویی را ایجاد کنیم. سپس، هر از یک دسته‌بندهای دودویی آموزش می‌بیند تا یک را پیش‌بینی انجام دهد. به‌عنوان مثال، با وجود داشتن یک مساله دسته‌بندی چندکلاسی با مجموعه داده‌های قرمز، سبز و آبی، دسته‌بندی دودویی برای حل این مساله را می‌توان به‌صورت زیر انجام داد:

- مساله اول: **قرمز در مقابل سبز/آبی**
- مساله دوم: **آبی در مقابل سبز/قرمز**

---

- مساله سوم: **سبز در مقابل آبی/قرمز**

## یک در مقابل یک[1]

همانند رویکرد یک در مقابل همه، رویکرد یک در مقابل یک، یکی دیگر از روش‌های است که از الگوریتم دسته‌بندی دودویی برای دسته‌بندی مجموعه داده‌ای چند کلاسی استفاده می‌کند. در این روش نیز، مجموعه داده‌های چند کلاسی به مجموعه‌ای از چند دسته‌بندی دودویی تقسیم می‌شود. در دسته‌بندی یک در مقابل یک، برای مجموعه داده‌ای با N کلاس، تعداد:

$$\frac{N * (N - ۱)}{۲}$$

دسته‌بند ایجاد می‌شود. با استفاده از این رویکرد دسته‌بندی، مجموعه داده اصلی به یک مجموعه داده تبدیل می‌شود که هر کلاس در مقابل کلاس دیگر قرار می‌گیرد. به‌عنوان مثال، با در نظر گرفتن یک مجموعه داده چندکلاسی با چهار کلاس آبی، قرمز، سبز و زرد، رویکرد یک در مقابل یک آن را به ۶ مجموعه داده دودویی تقسیم می‌کند:

- مساله اول: **قرمز در مقابل سبز**
- مساله دوم: **قرمز در مقابل آبی**
- مساله سوم: **قرمز در مقابل زرد**
- مساله چهارم: **سبز در مقابل زرد**
- مساله پنجم: **آبی در مقابل سبز**
- مساله ششم: **آبی در مقابل زرد**

### دسته‌بندی تک‌کلاسی

در مسائل دسته‌بندی باینری و چندگانه، تابع تصمیم دسته‌بند، با حضور نمونه‌هایی از هر کلاس پشتیبانی می‌شود و الگوریتم‌های مرتبط برای طبقه‌بندی هر نمونه جدید، در یکی از چندین کلاس از پیش تعریف‌شده طراحی شده‌اند. در کاربردهایی همانند سیستم‌های صنعتی، تنها داده‌های موجود حالت‌های عادی عملکرد فرآیند فیزیکی مورد مطالعه را مشخص می‌کنند، حال آن‌که داده‌های مربوط به حالت‌های ناکارآمد و وضعیت‌های بحرانی به سختی بدست می‌آیند. وقتی نوبت به فرآیندهای صنعتی و تشخیص خطاهای ماشین ونفوذ می‌شود، به طور کلی، تعداد حالت‌های خرابی و افزایش تعداد حملات ایجاد شده جدید ممکن است محدود نباشد. به همین دلیل است که محققان در چند سال گذشته الگوریتم‌هایی را برای حل مسائل تک‌کلاسی توسعه داده‌اند که در آن مجموعه داده‌های موجود فقط به یک کلاس منفرد اشاره دارد.

---

طبقه‌بندهای تک‌کلاسی، حالت‌های رفتاری عادی سیستم مورد مطالعه را یاد می‌گیرند. آن‌ها توابع تصمیم‌گیری را به منظور آزمایش نمونه‌های جدیدی که در مجموعه داده آموزشی موجود نیستند را توسعه می‌دهند، به نحوی که تا حد امکان نمونه‌های عادی را بپذیرند و نقاط دورافتاده را شناسایی کنند (هر نمونه‌ای که به توزیع مشابه مجموعه داده آموزشی تعلق ندارد).

## دسته‌بندی تک‌کلاسی در مقابل دسته‌بندی دودویی و چندگانه

دسته‌بندی تک‌کلاسی با دسته‌بندی دودویی و چندگانه تفاوت دارد، چراکه مجموعه آموزشی فقط شامل اشیاکلاس هدف است و هیچ اطلاعاتی در مورد کلاس‌های دیگر در دسترس نیست. وظیفه آن تعیین مرزی است که اشیا هدف را محصور کند و شانس پذیرش اشیا دور از هم را به حداقل برساند.

به عبارت دیگر، دسته‌بندی تک‌کلاسی از طبقه‌بندی دودویی (یا چندگانه) به این دلیل متفاوت است که در مورد دوم، مجموعه آموزشی دادها را از همه طبقات از پیش تعریف‌شده مشخص کرده‌است. حال آن‌که این مجموعه داده جامع در دسته‌بندی تک‌کلاسی وجود ندارد. یعنی، هیچ نمونه‌ای از کلاس دوم (یا چند کلاس دیگر) در طول آموزش موجود نیستند. بنابراین، نمونه‌های کلاس مورد انتظار تنها با استفاده از نمونه‌هایی از یک کلاس دسته‌بندی می‌شوند. علاوه بر این، از آنجایی که مسائل دسته‌بندی کلاسیک، توابع تشخیص‌دهنده[1] را براساس نمونه‌ها از همه دسته‌ها ایجاد می‌کنند، به طور طبیعی تشخیص‌دهنده هستند. این ویژگی به یک مجموعه داده متوازن نیاز دارد تا یک مرز تصمیم‌گیری موثر بسازد. زمانی که نمونه‌هایی از یک کلاس نسبت به دیگر کلاس فراوانی بیشتری دارند، روش‌های تشخیص‌دهنده ممکن است عملکرد خوبی نداشته باشند و نتوانند بکار گرفته شوند. در نتیجه، دسته‌بندی تک‌کلاسی وارد عمل می‌شود. به طور خلاصه، اگر مجموعه داده با تعداد زیادی از نمونه‌ها از همه‌ی کلاس‌ها تشکیل شده باشد، دسته‌بندی دودویی (یا چندکلاسی) توصیه می‌شود. بر عکس، در موارد نامتوازن زمانی که فراوانی نمونه‌هایی از یک کلاس خاص مشاهده می‌شود، راه حل استفاده از دسته‌بندی تک‌کلاسی است.

گاهی اوقات وظیفه دسته‌بندی صرفا تخصیص یک نمونه‌ی آزمایشی به دسته‌های از پیش تعریف‌شده نیست، بلکه تصمیم‌گیری در مورد تعلق آن به یک کلاس خاص است. با این حال، هدف دسته‌بندی چند کلاسه سنتی دسته‌بندی یک نمونه داده ناشناخته به یکی از چندین دسته از پیش تعریف‌شده است. هنگامی که نمونه داده ناشناخته به هیچ یک از این دسته‌ها تعلق نداشته باشد، مشکل ایجاد می‌شود. فرض کنیم ما یک مجموعه داده آموزشی داریم که شامل نمونه‌هایی از میوه‌ها و سبزیجات است. اگر یک نمونه آزمایشی ناشناخته (در حوزه میوه‌ها و سبزیجات، به عنوان مثال سیب یا سیب‌زمینی) برای دسته‌بندی داده شود، می‌توان از دسته‌بندی دودویی برای این مساله استفاده کرد. حال اگر نمونه داده‌های آزمایشی از یک دامنه کاملا متفاوت باشد

---

[1] discriminatory functions

(مثلاً گربه از دسته حیوانات)، طبقه‌بند همیشه گربه را به عنوان میوه یا سبزی دسته‌بندی می‌کند که در هر دو مورد نتیجه اشتباهی است.

## کاربرد دسته‌بندی تک‌کلاسی

دسته‌بندی تک‌کلاسی به طور گسترده در زمینه‌های زیر کاربرد دارند:

- **در تشخیص خطاهای ماشین.** به عنوان مثال، در هنگام *نظارت بر گیربکس هلیکوپتر،* در *کنترل وضعیت عملیاتی یک نیروگاه هسته‌ای* یا در حین *تشخیص نشت نفت.* در اینجا، دسته منفی همه رفتارهای غیرعادی ممکن را شامل می‌شود، اما در درجه اول بسیار نادر هستند و بنابراین، ممکن است برای مردم خطراتی ایجاد کنند و همچنین منجر به هزینه‌های بالا شوند. انتظار برای وقوع خطاها استراتژی خوبی نیست. درعوض، ساختن یک دسته‌بندی تک‌کلاس طبق مشاهدات عادی ماشین، یک راه‌حل است.

- **تشخیص خودکار یک بیماری.** داده‌های مثبت با بیماری‌های "شایع" نشان داده می‌شوند که به راحتی با هم گروه‌بندی می‌شوند. در حالی که دسته منفی توسط بیماری‌های "نادر" تشکیل می‌شود. پر کردن دسته دورافتاده دشوار است، چراکه آزمایش‌های مربوط به بیماری‌های نادر بسیار پرهزینه است و بیماران نادر هستند که باعث می‌شوند طبقه منفی از نمونه‌ها کم باشد.

- **در احراز هویت کُنشیِ تلفن همراه.** ما تنها تصاویر را از کاربر فعلی بدست می‌آوریم. چراکه نمونه‌های دسته منفی (سایر کاربران)، به دلیل مسائل مربوط به حریم خصوصی، به سختی می‌توان جمع‌آوری شوند.

- **برنامه‌های تشخیص ناهنجاری شبکه.**

- **تشخیص نفوذ در سیستم‌های صنعتی.**

- **تشخیص تازگی سری‌های زمانی[1].**

- **تجزیه و تحلیل تشنج[2] از سیگنال‌های الکتروانسفالوگرافی داخل جمجمه‌ای.**

- **تشخیص اشیا بصری در زمینه تعاملات انسان−ربات[3] (HRI).**

می‌توان این موراد را در سه کاربرد کلی دسته‌بندی کرد: **تشخیص تازگی[4]**، **تشخیص ناهنجاری** و **احراز کنشی تلفن همراه.** در تشخیص تازگی، *هدف یافتن موارد جدید با توجه به نمونه‌های مشاهده شده است.* بنابراین طبیعی است که داده‌های کلاس جدید شناخته نشده باشند.

---

هدف از تشخیص ناهنجاری *شناسایی داده‌های غیرعادی است*. از آن‌جایی که آموزش با استفاده از مثال‌های گرداننده‌ی معمولی انجام می‌شود، دسته‌بند ما باید مفهوم هنجار بودن را بیاموزد. در احراز هویت کنشی تلفن همراه، هویت یک کاربر به طور مداوم تایید می‌شود. تنها نمونه‌های او برای تشخیص نمونه‌های منفی در دسترس هستند.

> زمانی از الگوریتم‌های طبقه‌بندی تک‌کلاسه استفاده می‌شود که کلاس منفی وجود نداشته باشد، نمونه‌برداری از آن ضعیف باشد یا به خوبی تعریف نشده باشد.

## دسته‌بندی چندبرچسبی

یادگیری بانظارت علاقمند به استنباط روابط بین نمونه‌های ورودی و برچسب‌های کلاس است. در مسائل دسته‌بندی سنتی، هر نمونه با یک برچسب کلاس مرتبط است. با این حال، در بسیاری از سناریوهای دنیای واقعی، یک نمونه ممکن است با چندین برچسب مرتبط باشد. به عنوان مثال، در دسته‌بندی اخبار، بخشی از اخبار مربوط به عرضه آیفون جدید توسط اپل، هم با برچسب تجارت و هم با برچسب فناوری مرتبط است. به عبارت دیگر، هر نمونه به جای تنها یک برچسب، با مجموعه‌ای از برچسب‌ها مرتبط است. **یادگیری چندبرچسبی** یک زمینه یادگیری ماشین است که به یادگیری از داده‌های چندبرچسبی اشاره دارد که در آن هر نمونه با چندین برچسب بالقوه مرتبط است.

یک تفاوت عمده بین یادگیری چند برچسبی و یادگیری دودویی یا چندگانه سنتی این است که برچسب‌ها در یادگیری چندبرچسبی متقابلا مجزا نیستند. به عبارت دیگر، هر نمونه ممکن است به چندین برچسب مرتبط باشد. بنابراین، یکی از چالش‌های کلیدی یادگیری چندبرچسبی، چگونگی بهره‌برداری مؤثر از همبستگی‌ها بین برچسب‌های مختلف است. علاوه بر این، برخلاف دسته‌بندی تک‌برچسبی، مساله چندبرچسبی تحت تاثیر همبستگی‌های پنهان ذاتی بین برچسب‌ها قرار می‌گیرد. این بدان معناست که عضویت یک نمونه در یک کلاس می‌تواند برای پیش‌بینی مجموعه برچسب‌های آن مفید باشد. به عنوان مثال، بیمار مبتلا به فشار خون بالا بیشتر از افراد دیگر به بیماری قلبی مبتلا می‌شود، اما احتمال ابتلای این فرد به دیستروفی ماهیچه‌ای کمتر است.

## یادگیری چندبرچسبی

یادگیری چندبرچسبی مرتبط با پیش‌بینی برچسب‌های نمونه‌های دیده‌نشده با ساخت یک دسته‌بند بر اساس داده‌های آموزشی است. فرض کنید $\mathcal{X}$ و $\mathcal{Y}$ به ترتیب فضای نمونه ورودی و فضای برچسب خروجی را نشان دهند. در یادگیری چندبرچسبی، فضای برچسب $\mathcal{Y}$ به صورت $\mathcal{Y} = \{0,1\}^k$ تعریف می‌شود، که در آن $k$ تعداد برچسب‌ها است. یعنی، اگر نمونه‌ای مربوط به برچسب $j$ام باشد، جز $j$ام بردار برچسب ۱ است و در غیر این صورت مقدار آن صفر است.

مشابه دسته‌بندی سنتی، با توجه به مجموعه داده‌های آموزشی، هدف یادگیری چندبرچسبی یادگیری $f(\mathcal{X}) \rightarrow \mathcal{Y}$ توسط یک دسته‌بند است که برچسب‌های هر نمونه $x \in \mathcal{X}$ را پیش‌بینی می‌کند. به طور خاص، خروجی دسته‌بند $f$ برای یک نمونه مشخص $x \in \mathcal{X}$ برابر با:

$$f(x) = [f_1(x), f_2(x), \dots, f_k(x)]^T$$

است، که در آن $f_j(x)(j = \cdot, \dots, k)$ یک یا صفر است که نشان‌دهنده‌ی ارتباط $x$ با برچسب $j$ام است.

می‌توان یادگیری چندبرچسبی را اینگونه نیز تعریف کرد. فرض کنید $\mathcal{X}$ فضای نمونه ورودی را نشان دهد و $\mathcal{Y} = \{w_1, \dots, w_Q\}$ مجموعه محدودی از برچسب‌ها باشد. همچنین، اگر $\mathcal{D} = \{(x_1, Y_1), \dots (x_i, Y_i)\}$ مجموعه داده‌ای متشکل از $n$ نمونه چندبرچسبی $(x_i, Y_i), x_i \in \mathcal{X}, Y_i \subseteq \mathcal{Y}$ را نشان دهد، بر این اساس، هدف از یادگیری چندبرچسبی ساختن یک دسته‌بند چندبرچسبی $\mathcal{H}$ است که یک نمونه $x$ را به مجموعه برچسب‌های $Y$ مرتبطِ با آن نگاشت کرده و برخی از معیارهای ارزیابی را بهینه می‌کند.

هنگامی که صحبت از یادگیری از داده‌های چندبرچسبی می‌شود، دو رویکرد اصلی برای حل آن وجود دارد: **تبدیل مساله**[1] و **انطباق الگوریتم**[2]. در رویکرد تبدیل مساله، ابتدا مساله یادگیری چندبرچسبی را با یک سری مسائل تک‌برچسبی تبدیل می‌کند و سپس با استفاده از روش‌های یادگیری تک‌برچسبی موجود به حل آن می‌پردازد. هدف رویکرد دوم، انطباق الگوریتم‌های دسته‌بندی موجود است، بنابراین آن‌ها می‌توانند با داده‌های چند برچسبی سروکار داشته باشند و به جای تنها یک خروجی، چندین خروجی تولید کنند. به عبارت دیگر، انطباق الگوریتم، الگوریتم‌های تک‌برچسبی را برای مقابله مستقیم با داده‌های چندبرچسبی تعمیم می‌دهد.

## رویکرد تبدیل مساله

ساده‌ترین راه‌حل در یادگیری چندبرچسبی، رویکرد تبدیل مساله است که می‌تواند با هر الگوریتم یادگیری استفاده شود. در این روش، مساله دسته‌بندی چند برچسبی به یک یا چند مساله دسته‌بندی تک‌برچسبی تبدیل می‌شود. سپس راه‌حل‌های این مسائل برای حل مساله اصلی یادگیری چندبرچسبی ترکیب می‌شوند. روش تبدیل مساله شامل سه رویکرد اصلی است: **ارتباط دودویی**[3]، **مجموعه توانی برچسب**[4] و **رتبه‌بندی برچسب**[5].

---

**ارتباط دودویی**

رویکرد ارتباط دودویی (BR) که به عنوان استراتژی **یک ــ در برابر ــ همه** نیز شناخته میشود، مساله یادگیری چندبرچسبی با کلاسهای ممکن $Q$ را به مسائل دستهبندی تکبرچسبی $Q$ تقسیم میکند که با آموزش دستهبندهای دودویی $(h_1, ..., h_Q) = Q$ قابل حل است. هر $q$ دستهبند $(q\epsilon\{1, ..., Q\})$ بر روی مجموعه داده اصلی آموزش داده میشود و هدف آن تعیین ارتباط برچسب خاص خود برای یک نمونه معین است. هنگام دستهبندی یک نمونه جدید $x$، BR اجتماع برچسبهایی را که بهطور مثبت توسط دستهبندهای دودویی پیشبینی میشوند، خروجی میدهد. سپس دستهبندهای چندبرچسبی توسط:

$$\mathcal{H} = \{w_Q\epsilon \,\mathcal{Y}|h_Q(x) = 1\}$$

تعیین میشود.

روش BR برای پیادهسازی ساده است و پیچیدگی آن با تعداد برچسبهای ممکن خطی است. با این حال، BR ارتباط بین برچسبها را با هر برچسب به طور مستقل، نادیده میگیرد. برای مقابله با جنبههای منفی BR، **زنجیره دستهبند**[1] (CC) معرفی شده که شامل دستهبندهای باینری $Q$ است که در امتداد یک زنجیره بهم متصل شدهاند. رویکرد مجموعه *توانی برچسب* ارائه شده در بخش بعدی نیز یکی از گزینههای جایگزین برای مقابله با این جنبههای منفی رویکرد BR است.

**مجموعه توانی برچسب**

با توجه به مجموعه داده آموزشی $D$ با $n$ نمونه، رویکرد مجموعه توانی برچسب (LP) هر مجموعه منحصربهفردی از برچسبها را در $D$ به عنوان یک برچسب در نظر میگیرد و سپس یک دستهبند تکبرچسبی را آموزش میدهد. تعداد کلاسها با $\min (2^Q, n)$ محدود میشود. پیچیدگی LP به پیچیدگی دستهبندهای تکبرچسبی با توجه به تعداد کلاسها بستگی دارد. برای یک نمونه جدید، رویکرد LP محتملترین کلاسی را خروجی میدهد که مجموعهای از برچسبها در نمایش چندبرچسبی اصلی است. LP مزیت در نظر گرفتن همبستگیهای برچسب را دارد. با این حال، یک جنبه منفی این رویکرد این است که ممکن است به مجموعه دادههای نامتعادل با تعداد زیادی کلاس همراه با نمونههای کمی منجر شود.

**رویکرد انطباق الگوریتم**

روشهای انطباق با مساله، الگوریتمهای یادگیری ماشین سنتی را سفارشی میکنند تا بهطور مستقیم مفاهیم چندبرچسبی را مدیریت کنند. این روشها مزیت تمرکز بر یک الگوریتم خاص

---

[1] Classifier Chain

را دارند. مزیت دیگر این است که این روش‌ها از تمام مجموعه داده آموزشی به‌طور همزمان برای آموزش یک دسته‌بند چندبرچسبی استفاده می‌کنند. به طور کلی، عملکرد این الگوریتم‌ها در مسائل دشوار دنیای واقعی بهتر از روش‌های تبدیل مساله است و این به قیمت پیچیدگی بیشتر است.

## کاربردهای دسته‌بندی چندبرچسبی

هنگامی که مفاهیم اصلی مرتبط با دسته‌بندی چند برچسبی معرفی شدند، سوال بعدی که احتمالا مطرح می‌شود این است که آن‌ها در کجا کاربرد دارند. همان‌طور که در پیش‌تر بیان شد، هدف یک دسته‌بند چندبرچسبی، پیش‌بینی مجموعه‌ای از برچسب‌های مرتبط برای یک نمونه داده جدید است. در این بخش، چندین حوزه کاربردی که می‌توانند از این قابلیت بهره‌مند شوند، به تصویر کشیده شده است.

### دسته‌بندی صحنه[1]

در دسته‌بندی صحنه، وظیفه تعیین برچسب‌های معنایی مرتبط مانند کوه، دریاچه یا غیره برای تصاویر داده شده است. دسته‌بندی صحنه در بسیاری از زمینه‌ها، از جمله نمایه‌سازی تصویر مبتنی‌بر محتوا و بهبود تصویر حساس به محتوا، کاربرد دارد. برای مثال، بسیاری از سیستم‌های کتابخانه دیجیتال فعلی از بازیابی تصویر مبتنی‌بر محتوا پشتیبانی می‌کنند، که به کاربر اجازه می‌دهد تصاویری را که شبیه به یک تصویر پرس‌وجو هستند، بازیابی کند. در این مورد، آگاهی از برچسب‌های معنایی تصویر پرس‌وجو می‌تواند فضای جستجو را کاهش دهد و دقت بازیابی را بهبود بخشد. از آنجایی که یک صحنه طبیعی ممکن است شامل چندین شی باشد، هر تصویر می‌تواند با چندین برچسب مرتبط شود. از این رو، دسته‌بندی صحنه به‌طور طبیعی یک مساله یادگیری چندبرچسبی است.

### دسته‌بندی متن

دسته‌بندی متن، وظیفه دسته‌بندی اسناد متنی به یک یا چند مجموعه از دسته‌های از پیش تعریف‌شده است. مساله دسته‌بندی متن به اوایل دهه ۱۹۶۰ باز می‌گردد. با این حال، اثربخشی دسته‌بندی متن به‌طور قابل توجهی در دهه‌های گذشته به دلیل پیشرفت روش‌های یادگیری ماشین بهبود یافته است. اسناد متنی را می‌توان در هر جایی یافت. از شرکت‌های بزرگ که انواع توافق‌نامه‌ها و گزارش‌های را ذخیره می‌کنند تا افرادی که فاکتورها و پیام‌های پست الکترونیکی خود را ثبت می‌کنند. تمام کتاب‌ها و مجلات منتشر شده، سوابق پزشکی تاریخی ما و همچنین مقالات در رسانه‌های الکترونیکی، پست‌های وبلاگ و غیره نیز اسناد متنی هستند.

---

[1] Scene Classification

دسته‌بندی متن در زمینه‌های مختلفی از جمله دسته‌بندی صفحات‌وب، تشخیص موضوع متن، فیلتر کردن محتوا و غیره استفاده شده است. معمولاً، برچسب‌ها (یا دسته‌های) از پیش تعریف‌شده در دسته‌بندی متن، متقابلاً منحصربه‌فرد فرض نمی‌شوند. بنابراین دسته‌بندی متن را می‌توان به‌طور طبیعی به عنوان یک مساله یادگیری چندبرچسبی مدل کرد. به عنوان مثال، برچسب‌های تجاری، فناوری، سرگرمی و سیاست را در دسته‌بندی اخبار در نظر بگیرید. یک مقاله خبری در مورد عرضه یک آیفون جدید توسط اپل ممکن است دارای برچسب تجاری و فناوری برچسب باشد.

## تجزیه و تحلیل ژنومیک عملکردی[۱]

ژنومیکس عملکردی یک زمینه مهم در بیوانفورماتیک است که عملکرد ژن و پروتئین را با انجام تجزیه و تحلیل در مقیاس بزرگ بر روی حجم وسیعی از داده‌های جمع‌آوری شده توسط پروژه‌های ژنوم مطالعه می‌کند. به عنوان مثال، ریزآرایه‌های DNA به محققان اجازه می‌دهند تا سطوح بیان هزاران ژن مختلف را به‌طور هم‌زمان اندازه‌گیری کنند.

در تجزیه و تحلیل بیان ژن خودکار، وظیفه پیش‌بینی عملکرد ژن‌ها است و به طور کلی، بر این فرض استوار است که ژن‌هایی با عملکردهای مشابه، پروفایل‌های بیان مشابهی در سلول‌ها دارند. توجه داشته باشید که هر ژن ممکن است با چندین عملکرد در ژنومیک عملکردی همراه باشد. هنگامی که عملکردها به عنوان برچسب در نظر گرفته می‌شوند، مساله پیش‌بینی عملکرد در ژنومیک عملکردی می‌تواند به عنوان یک مساله یادگیری چندبرچسبی مدل شود.

## چالش‌های دسته‌بندی چندبرچسبی

در مقایسه با دسته‌بندی سنتی دودویی و چندگانه، حل مساله دسته‌بندی چندبرچسبی با چالش بیشتری مواجه است. در ادامه، برخی از چالش‌های اساسی در کاربرد موفقیت‌آمیز یادگیری چندبرچسبی در مسائل دنیای واقعی را شرح می‌دهیم.

اولین چالش در چگونگی بهره‌برداری موثر از ساختار برچسب برای بهبود عملکرد دسته‌بندی نهفته است. در یادگیری چندبرچسبی، برچسب‌ها اغلب با هم مرتبط هستند. چراکه آن‌ها متقابلاً منحصربه‌فرد نیستند. از این‌رو، نحوه اندازه‌گیری و ثبت همبستگی‌ها در فضای برچسب برای پیش‌بینیِ بهبودیافته بسیار مهم است.

چالش دوم مربوط به اثربخشی و کارایی یادگیری چند برچسبی برای مسائل در مقیاس بزرگ است؛ به ویژه زمانی که هم ابعاد داده و هم تعداد برچسب‌ها زیاد باشد. یادگیری چند برچسبی نیز از **مشقت بعدچندی**[۲] رنج می‌برد و بسیاری از روش‌های یادگیری چندبرچسبی موجود برای

---

داده‌های با ابعاد بالا کم‌تر موثر هستند، زیرا نقاط داده در فضای با ابعاد بالا پراکنده و از یکدیگر دور می‌شوند. برای مثال، روش‌های BR و LP که پیش‌تر مورد بحث قرار گرفتند، به مسائلی با اندازه برچسب نسبتا کوچک محدود می‌شوند. اخیرا روش‌هایی برای مقابله با برچسب‌ها با تعداد زیاد پیشنهاد شده است. به عنوان مثال، ابعاد فضای برچسب با استفاده از یک *نگاشت تصادفی*[1] کاهش می‌یابد. علاوه بر این، زمانی که تعداد برچسب‌ها زیاد باشد، حفظ تعداد زیادی از مدل‌های پیش‌بینی در حافظه دشوار می‌شود.

# الگوریتم‌های دسته‌بندی

در مساله دسته‌بندی یادگیری ماشین، ما با یک مجموعه داده (جایی که نقاط از فضای نمونه می‌آیند)، همراه با یک برچسب (یا کلاس) برای هر نقطه (جایی که تعداد محدودی از برچسب‌های ممکن وجود دارد) شروع می‌کنیم. فرض می‌کنیم که نقاط در مجموعه داده به‌طور مستقل و یکسان توزیع شده‌اند و یک نقطه داده جدید از توزیع مشابه مجموعه داده داریم که **پرسمان**[2] **(جستار یا پرس‌وجو)** نامیده می‌شود و همچنین فرض می‌شود که مستقل از نقاط مجموعه داده است. با این حال، ما برچسبی برای پرسمان نداریم. از این‌رو، ما می‌خواهیم برچسب را برای پرسمان بر اساس مجموعه داده پیش‌بینی کنیم.

به عنوان مثال، فرض کنید می‌خواهیم پیش‌بینی کنیم که آیا فردی بر اساس ژنوم، مستعد ابتلا به بیماری قلبی است یا خیر. ما مجموعه داده‌ای از ژنوم افراد با توالی ژنومی و اینکه آیا آنها بیماری قلبی دارند یا نه، داریم. ما اکنون یک بیمار جدید داریم که ژنوم آن را داریم اما نمی‌دانیم که آیا بیماری قلبی دارد یا خیر. از این‌رو، می‌خواهیم بر اساس توالی ژنومی، پیش‌بینی کنیم که آیا فرد بیماری قلبی دارد یا خیر؛ با تنها اطلاعاتی که در دسترس ما است، یعنی مجموعه داده‌ها و توالی ژنومی فرد.

اگر $X$ مجموعه داده و $Y$ مجموعه‌ای از کلاس‌ها باشد، یک دسته‌بند $f: X \rightarrow Y$ تابعی است که سعی می‌کند کلاس $y$ را برای نقطه داده $x$ پیش‌بینی کند. دقت دسته‌بند $f$ احتمال این که برچسب درست را برای پرسمان پیش‌بینی کنیم است و خطای دسته‌بند $f$ احتمال این است که برچسب نادرست را پیش‌بینی کنیم. ما می‌خواهیم یک دسته‌بند $f$ پیدا کنیم که دقت آن تا حد امکان بالا باشد (یا به طور معادل، خطای آن تا حد امکان کوچک باشد). فرآیند ساخت دسته‌بند $f$ را **یادگیری** گویند. *یک قانون یادگیری خانواده‌ای از توابع است که مجموعه‌ای از نقاط داده برچسب‌گذاری شده را می‌گیرد و یک دسته‌بند را در خروجی می‌دهد که می‌توانیم از آن برای دسته‌بندی نقاط پرسمان استفاده کنیم. هنگام اعمال یک قانون یادگیری و سپس استفاده از آن*

---

برای دسته‌بندی نقاط، اغلب به ترکیب قانون یادگیری و دسته‌بند با هم به عنوان یک دسته‌بند اشاره می‌کنیم.

برای آزمایش دقت هر دسته‌بند، مجموعه داده را می‌گیریم و آن را به دو زیرمجموعه مجزا تقسیم می‌کنیم، مجموعه آموزشی و مجموعه آزمایشی. مجموعه آموزشی در ساخت $f$ استفاده می‌شود و از این طریق برچسب‌های نقاط مجموعه داده آزمایشی را پیش‌بینی می‌کنیم. سپس برچسب‌های پیش‌بینی‌شده را با برچسب‌های درست در مجموعه آزمون مقایسه می‌کنیم و دقت پیش‌بینی خود را محاسبه می‌کنیم.

## یادگیرنده‌های پارامتری و ناپارامتری

الگوریتم‌های یادگیری ماشین را می‌توان به دو دسته پارامتری یا ناپارامتری طبقه‌بندی کرد. یک پارامتر را می‌توان به عنوان یک متغیر پیکربندی که ذاتی مدل است توصیف کرد. مقدار پارامتر می‌تواند از داده‌های آموزشی در نظر گرفته شود. پس از آموزش، از پارامترها برای تعیین عملکرد مدل در داده‌های آزمون استفاده می‌شود. به عبارت دیگر، مدل از آن‌ها برای پیش‌بینی استفاده می‌کند. یک مدل یادگیری ماشین با تعدادی از پارامترها یک مدل پارامتری است. به طور خلاصه، مدل‌های پارامتری در یادگیری ماشین معمولا رویکردی مبتنی‌بر مدل دارند که در آن فرضی را با توجه به شکل تابعی که باید تخمین زده شود می‌کنیم و سپس مدل مناسبی را بر اساس این فرض به منظور تخمین مجموعه‌ی پارامترها انتخاب می‌کنیم. بزرگ‌ترین نقطه ضعف روش‌های پارامتری این است که فرضیاتی که می‌کنیم ممکن است همیشه درست نباشند. به عنوان مثال، ممکن است فرض کنید که شکل تابع خطی است، در حالی که اینطور نیست. با این حال، روش‌های پارامتری بسیار سریع هستند و همچنین به داده‌های بسیار کمتری در مقایسه با روش‌های ناپارامتری نیاز دارند. یک مثال رایج از الگوریتم پارامتری رگرسیون خطی است.

در مقابل، الگوریتم‌هایی که هیچ فرض خاصی در مورد نوع تابع نگاشت ندارند به عنوان الگوریتم‌های ناپارامتری شناخته می‌شوند. از آنجایی که در این روش‌ها هیچ فرضی وجود ندارد، می‌توانند تابع مجهول $f$ را که می‌تواند به هر شکلی باشد، تخمین بزنند. روش‌های ناپارامتری معمولا دقیق‌تر هستند، زیرا به دنبال بهترین برازش با نقاط داده هستند و می‌توانند جنبه‌های ظریف‌تری از داده‌ها را به تصویر بکشند. با این حال، این به قیمت نیاز به تعداد بسیار زیادی از مشاهدات تمام می‌شود که برای تخمین دقیق تابع مجهول $f$ مورد نیاز است. علاوه بر این، از آنجایی که این الگوریتم‌ها انعطاف‌پذیرتر هستند، ممکن است گاهی اوقات خطاها و نویزها را به گونه‌ای یاد بگیرند که نتوانند به خوبی به نقاط داده جدید و دیده نشده تعمیم داده شوند. به‌طور خلاصه، مبادله بین یادگیرنده‌های پارامتری و ناپارامتری در هزینه و دقت محاسباتی است. یک مثال رایج از یک الگوریتم ناپارامتری کـ نزدیکترین همسایه است.

## یادگیرنده‌های مبتنی‌بر نمونه

یادگیرنده‌های مبتنی‌بر نمونه، الگوریتم‌های دسته‌بندی ناپارامتری هستند که یک نمونه‌ی بدونِ برچسبِ جدید را با توجه به برچسب‌های نمونه‌های مشابه در مجموعه آموزشی دسته‌بندی می‌کنند. در هسته این الگوریتم‌ها، یک روش جستجوی ساده وجود دارد. این تکنیک‌ها می‌توانند دسته‌بندی‌های پیچیده را از تعداد نسبتا کمی نمونه استنتاج کنند و طبیعتا برای حوزه‌های عددی مناسب هستند. با این حال، آن‌ها می‌توانند به ویژگی‌های نامربوط حساسیت بسیار زیادی داشته باشند و قادر به انتخاب ویژگی‌های مختلف در مناطق مختلف فضای نمونه نیستند. علاوه بر این، اگرچه پیچیدگی زمانی برای آموزش این مدل ها‌کم است، دسته‌بندی یک نمونه جدید نسبتا زمان‌بر است.

ابتدایی‌ترین و ساده‌ترین الگوریتم مبتنی‌بر نمونه، یادگیرنده **نزدیک‌ترین همسایه** است که قانون آن برای دسته‌بندی یک الگوی ناشناخته، بر این اساس است: کلاس نزدیک‌ترین نمونه در مجموعه آموزشی را‌که با یک معیار فاصله مشخص اندازه‌گیری و انتخاب کنید. با وجود سادگی، دسته‌بند نزدیک‌ترین همسایه نسبت به روش‌های دیگر مزایای زیادی دارد. به عنوان مثال، می‌تواند از یک مجموعه آموزشی نسبتا کوچک تعمیم یابد. یعنی، در مقایسه با روش‌های دیگر، همانند درخت‌های تصمیم یا شبکه عصبی، دسته‌بند نزدیک‌ترین همسایه به نمونه‌های آموزشی کوچک‌تری برای دستیابی به عملکرد دسته‌بندی یکسان نیاز دارد. دسته‌بند نزدیک‌ترین همسایه می‌تواند به عملکردی دست یابد که با روش‌های مدرن و پیچیده‌تر همانند درخت‌های تصمیم یا شبکه‌های عصبی قابل رقابت باشد.

## تفاوت یادگیرنده‌های مبتنی‌بر نمونه و مبتنی‌بر مدل

تفاوت اصلی یادگیرنده‌های مبتنی‌بر نمونه و مبتنی‌بر مدل در نحوه تعمیم اطلاعات آن‌ها خلاصه می‌شود. یادگیرنده‌های مبتنی‌بر نمونه، تمام داده‌های یک مجموعه آموزشی را به خاطر می‌سپارد و سپس یک نقطه داده جدید را به مقدار خروجی یکسان و یا متوسط نقاط داده مشابه که به خاطر سپرده شده‌است، تعیین می‌کند. در طرف دیگر، یادگیرنده مبتنی‌بر مدل، یک خط پیش‌بینی یا بخش پیش‌بینی را بر اساس ویژگی‌های مختلف داده‌هایی که روی آن آموزش داده است، ایجاد می‌کند. در نهایت، یک نقطه داده جدید، بر اساس ویژگی‌هایی که دارد در امتداد این خط یا در بخش‌های خاصی قرار می‌گیرد.

برای درک بهتر این دو یادگیرنده، مثالی (داستان) که در ادامه آمده است تفاوت آن را بهتر نشان می‌دهد. در وسط یک شهر کوچک، یک فروشگاه پوشاک معروف وجود داشت که توسط یک مادر و دخترش اداره می‌شد. مادر باید بداند که یک مشتری در فروشگاهش چقدر پول خرج کند، زیرا او کسی بود که از غافلگیری متنفر بود. دخترش که دانش‌آموخته‌ی رشته‌ی علوم رایانه است، تصمیم گرفت سیستمی بسازد تا مادرش نیازی به مقابله با استرس عادات

ناشناخته خرج کردن مشتری نداشته باشد. این سیستم به ویژگی‌های یک مشتری در حین ورود به فروشگاه نگاه می‌کند. برخی از ویژگی‌ها شامل نوع ماشینی بود که مشتری‌ها سوار می‌شدند و قیمت لباس‌هایی بود که به تن داشتند. در این شهر معمول بود که همیشه عادات خرج‌کردن خود را از طریق ماشین و لباس خود منعکس کنید. مشتری دائمی فروشگاه مرد جوانی به نام شروین بود. شروین یک تاجر موفق و یکی از ثروتمندترین اعضای شهر بود. او یک BMW می‌راند و مدام لباس‌های گران قیمتش را به رخ می‌کشید. مدلی که دخترش ساخته بود مشتریان جدیدی که اتومبیل‌های مدل بالا و لباس‌های گران‌قیمت داشتند را انتخاب می‌کرد و پیش‌بینی می‌کرد که آن‌ها همان مقداری که شروین در فروشگاه خرج می‌کند، خرج خواهند کرد. در بیشتر موارد، این مدل کاملاً خوب کار می‌کرد، زیرا هر فرد ثروتمند در این شهر کوچک تقریباً به همان میزان پول داشت. سپس یک روز یک فوتبالیست بزرگ به نام علی وارد شهر شد. علی حاضر نشد با چیزی کم‌تر از جدیدترین لامبورگینی و بهترین لباس‌های سفارشی که از ایتالیا خریده بود، دیده شود. وقتی علی به فروشگاه نزدیک شد، مدل پیش‌بینی کرد که علی همان مبلغی را که شروین خرج می‌کند، خرج خواهد کرد. هرچند، ماشین و لباس‌های علی به‌طور قابل توجهی گران‌تر از شروین بودند، این نزدیک‌ترین نقطه داده‌ای بود که سیستم باید به آن ارجاع می‌داد. علی در نهایت بسیار بیشتر از شروین در فروشگاه خرج کرد. مادر از این اتفاق مضطرب بود و به دخترش اجازه داد تا در مدل خود تجدید نظر کند. از این‌رو، دختر به این فکر افتاد که از سیستمی استفاده کند که از یادگیرنده مبتنی‌بر مدل به جای سیستمی که در اختیار داشت (یعنی از یادگیرنده مبتنی‌بر نمونه) استفاده کند. به این ترتیب، اگر یک مشتری جدید وارد شود، بدون ویژگی‌هایی که دقیقاً منعکس‌کننده داده‌هایی است که مدل قبلاً به خاطر سپرده است، پیش‌بینی‌ها احتمالاً منعکس‌کننده مبلغی است که آن‌ها در فروشگاه خرج می‌کنند. دفعه بعد که یک ماشین ناآشنا با یک مشتری با لباس‌های ناآشنا روبه‌رو شد، مدل پیش‌بینی دقیقی در مورد میزان خرج کردن آن‌ها در فروشگاه انجام داد. پس از آن مادر توانست اطمینان خاطر داشته باشد که دیگر همانند اتفاق علی نخواهد افتاد.

در این داستان، ما توانستیم وضعیتی را ببینیم که داشتن یک مدل یادگیری مبتنی‌بر نمونه، پیش‌بینی دقیقی ارائه نمی‌دهد. دلیل این که نقطه داده جدید (علی) در مقایسه با داده‌هایی که مدل روی آن آموزش داده شده بود، به عنوان یک نقطه دورافتاده عمل کرد. مدل‌های یادگیری مبتنی‌بر نمونه می‌توانند عملکرد بسیار خوبی داشته باشند اگر داده‌هایی که با آن‌ها آموزش داده می‌شوند شبیه داده‌های جدیدی باشد که سعی در پیش‌بینی برای آن‌ها دارد. با این حال، در شرایطی که ممکن است موارد دورافتاده وجود داشته باشد، یک مدل مبتنی‌بر نمونه ممکن است در مقدار پیش‌بینی‌شده خطا داشته باشد. با این حال، نوع مدلی که در نهایت برای مساله یادگیری ماشین خود استفاده می‌کنید به وضعیت و شرایط مساله بستگی دارد.

## تفاوت یادگیرنده‌های/الگوریتم‌های پارامتری با ناپارامتری

### یادگیرنده‌های پارامتری

روش‌های پارامتری مفروضات بزرگی را در مورد نگاشت متغیرهای ورودی به متغیر خروجی ایجاد می‌کنند و به نوبه خود سریع‌تر آموزش داده می‌شوند، به داده‌های کمتری نیاز دارند اما ممکن است آن‌قدر قدرتمند نباشند.

### نمونه‌هایی از این الگوریتم‌ها

- **رگرسیون لجستیک**
- **پرسپترون**
- **بیز ساده**

### فواید

- **ساده‌تر و قابل فهم‌تر؛ تفسیر نتایج آسان‌تر است.**
- **یادگیری سریع‌تر از داده‌ها.**
- **داده‌های آموزشی کمتری برای یادگیری تابع نگاشت مورد نیاز است.**

### محدودیت‌ها

- **محدودیت‌های فرم. روش‌های پارامتری یک الگوریتم را به یک فرم تابع مشخص محدود می‌کنند.**
- **تناسب ضعیف در عمل، بعید است که با تابع نگاشت زیربنایی مطابقت داشته باشد. به عبارت دیگر، این روش‌ها بهترین برازش را برای داده‌ها ارائه نمی‌کنند. آن‌ها به احتمال زیاد با تابع نگاشت همخوانی ندارند.**
- **پیچیدگی. الگوریتم‌های پارامتری دارای پیچیدگی محدودی هستند. این به این معنی است که آن‌ها برای مشکلات کمتر پیچیده مناسب‌تر هستند.**

### یادگیرنده‌های ناپارامتری

روش‌های ناپارامتری، فرضیات کمی در مورد تابع هدف ایجاد می‌کنند یا هیچ فرضیاتی در مورد تابع هدف ندارند و به نوبه خود به داده‌های بسیار بیشتری نیاز دارند، آموزش آن‌ها کندتر است و پیچیدگی مدل بالاتری دارند، اما می‌توانند مدل‌های قدرتمندتری ایجاد کنند.

### نمونه‌هایی از این الگوریتم‌ها

- **کا ـ نزدیک‌ترین همسایه**
- **ماشین بردار پشتیبان**

- **درختان تصمیم مانند CART و C4.5**

**فواید**

- قدرت بالا از طریق ایجاد فرضیات ضعیف یا بدون فرضیه در مورد تابع زیربنایی.
- انعطاف‌پذیری بالا، به این معنا که آن‌ها می‌توانند تعداد زیادی از فرم‌های تابع را در خود جای دهند.
- عملکرد بالا در مدل‌های پیش‌گویانه تولید شده.

**محدودیت‌ها**

- داده‌های آموزشی. داده‌های آموزشی بیشتری برای تخمین تابع نگاشت مورد نیاز است.
- سرعت. آموزش کندتر است.
- بیش‌برازش. به همان اندازه که این الگوریتم‌ها تمایل دارند داده‌ها را بهتر از الگوریتم‌های پارامتری برازش دهند، بیشتر مستعد بیش‌برازش هستند.

## یادگیرنده‌های تنبل و مشتاق

هنگامی که یک الگوریتم یادگیری ماشین بلافاصله پس از دریافت مجموعه داده‌های آموزشی، مدلی را می‌سازد، به آن یادگیرنده مشتاق می‌گویند. این روش مشتاق نامیده می‌شود، چراکه وقتی مجموعه داده را دریافت می‌کند، اولین کاری که انجام می‌دهد ساخت مدل است. سپس داده‌های آموزشی را فراموش می‌کند و بعدها وقتی یک داده ورودی می‌آید، از این مدل برای ارزیابی آن استفاده می‌کند. اکثر الگوریتم‌های یادگیری ماشین، یادگیرندگانی مشتاق هستند.

در مقابل، زمانی که یک الگوریتم یادگیری ماشین بلافاصله پس از دریافت داده‌های آموزشی مدلی نمی‌سازد، بلکه منتظر می‌ماند تا داده‌های ورودی برای ارزیابی ارائه شود، به آن یادگیرنده تنبل می‌گویند. به این روش تنبل می‌گویند، چراکه ساختن یک مدل را، تا زمانی که کاملا ضروری باشد، به تاخیر می‌اندازد. به عبارت دیگر، وقتی داده‌های آموزشی را دریافت می‌کند، فقط آن‌ها را ذخیره می‌کند. بعدها، وقتی داده‌های ورودی می‌آیند، تنها در این صورت از این داده‌های ذخیره شده برای ارزیابی نتیجه استفاده می‌کند. یادگیرنده تنبل یک تابع تفکیک‌پذیر را از داده‌های آموزشی یاد نمی‌گیرد، بلکه مجموعه داده‌های آموزشی را به خاطر می‌سپارد. برعکس، یادگیرنده مشتاق وزن مدل (پارامترهای) خود را در طول زمان آموزش می‌آموزد. یک مثال رایج از یک یادگیرنده تنبل کا ـ نزدیک‌ترین همسایه است.

# کا-نزدیک‌ترین همسایه

دسته‌بند **کا-نزدیک‌ترین همسایه (KNN)** یکی از قدیمی‌ترین، ساده‌ترین و در عین حال موثر، از الگوریتم‌های یادگیری بانظارت برای دسته‌بندی مجموعه داده‌ها است. الگوریتم KNN بر اساس این فرض طراحی شده است که چیزهای مشابه در نزدیکی یکدیگر وجود دارند. در مقایسه با سایر الگوریتم‌های دسته‌بندی، کا-نزدیک‌ترین همسایه از رویکرد یادگیری تنبل استفاده می‌کند. به عبارت دیگر، به سادگی نمونه‌ها را در مرحله آموزش ذخیره می‌کند و تا زمانی که نمونه‌های آزمایشی دریافت نشود کاری انجام نمی‌دهد.

شکل ۶_۳ شمایی از دسته‌بند کا-نزیک‌ترین همسایه را ارائه می‌دهد. همچنان که مشاهده می‌شود، پارامتر k نقش مهمی ایفا را دسته‌بندی نمونه‌ی جدید ایفا می‌کند. چراکه مقادیر k مختلف ممکن است به نتایج دسته‌بندی بسیار متفاوتی منجر شود. علاوه بر این، محاسبات مختلف فاصله نیز ممکن است به همسایگی متفاوت، و در نتیجه نتایج دسته‌بندی متفاوت منجر شود. از این‌رو، مقدار K داده شده صحت پیش‌بینی‌ها و تعداد خطاها را تعیین می‌کند، بنابراین انتخاب K مناسب، اهمیت اساسی در این الگوریتم دارد. انتخاب K ایده‌آل به داده‌ها بستگی دارد، اما مقادیر بزرگ K تاثیر نویز را بر دسته‌بندی کاهش می‌دهد، در حالی که مرزها و گروه‌بندی‌ها را کم‌تر متمایز می‌کند.

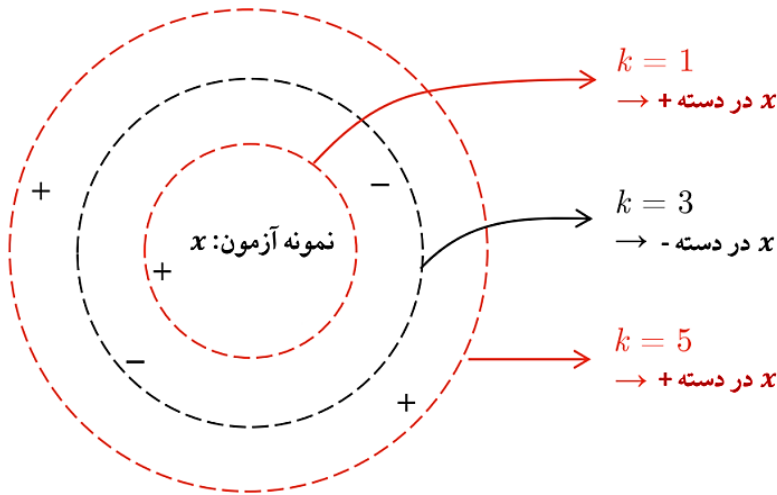

**شکل ۶_۳.** دسته‌بند کا-نزدیکترین همسایه.

دسته‌بند نزدیک‌ترین همسایه بر اساس یادگیری از طریق **تشابه**[1] است. نمونه‌های آموزشی با $p$ ویژگی توصیف می‌شوند. هر نمونه نشان‌دهنده یک نقطه در فضای $p$ بعدی است. به این

---

ترتیب تمام نمونه‌های آموزشی در یک فضای الگوی $p$ بعدی ذخیره می‌شوند. از این‌رو، هنگامی که یک نمونه ناشناخته به آن داده می‌شود، دسته‌بند کا‌ـنزدیک‌ترین همسایه فضای الگو را برای k نمونه آموزشی که نزدیک‌ترین به نمونه ناشناخته هستند، جستجو می‌کند. این k نمونه آموزشی کا‌ـنزدیک‌ترین همسایگان نمونه ناشناخته هستند. برای تعیین کا‌ـنزدیک‌ترین همسایه نقطه داده، باید از معیار تشابه یا عدم تشابه بین نقاط داده استفاده کنیم. معیارهای زیادی برای تشابه یا عدم تشابه وجود دارد، از جمله فاصله اقلیدسی، فاصله مینکوفسکی، فاصله همینگ، ضریب همبستگی پیرسون و شباهت کسینوس که در این بخش فاصله اقلیدسی توضیح داده شده است.

فاصله اقلیدسی به این صورت تعریف می‌شود:

$$d(x_i, x_j) = \sqrt{\sum_{l=1}^{p} (x_{i,l} - x_{j,l})^2}, \, i \neq j.$$

به عبارت دیگر، *برای هر ویژگی عددی*، تفاوت بین مقادیر متناظر آن ویژگی را در نمونه $x_i$ و در نمونه $x_j$ می‌گیریم، این اختلاف را مربع می‌کنیم و در انتها مجذور از مجموع تعداد فاصله جمع‌شده گرفته می‌شود. به طور معمول، قبل از استفاده از معادله، مقادیر هر ویژگی را هنجار (نرمال) می‌کنیم. این کمک می‌کند تا ویژگی‌هایی با دامنه‌های اولیه بزرگ، ویژگی‌هایی با دامنه‌های اولیه کوچک‌تر را بی‌تاثیر نکنند.

> فاصله اقلیدسی معیاری برای **عدم تشابه** بین دو نقطه داده $x_j$ و $x_j$ است. هرچه فاصله اقلیدسی بزرگتر باشد، دو نقطه داده با هم متفاوت‌تر هستند.

الگوریتم KNN هم برای دسته‌بندی و هم رگرسیون استفاده می‌شود. KNN سعی می‌کند با محاسبه فاصله بین داده‌های آزمایشی و تمام نقاط آموزشی، کلاس درست را برای داده‌های آزمون پیش‌بینی کند. معمولا برای مسائل دسته‌بندی، رای‌گیری می‌تواند برای پیش‌بینی نمونه آزمایشی به عنوان متداول‌ترین برچسب کلاس در همسایگان k استفاده شود. برای مسائل رگرسیون، از میانگین‌گیری می‌توان برای پیش‌بینی نمونه آزمایشی به عنوان میانگین k خروجی با مقدار‌ـواقعی استفاده کرد.

> الگوریتم کا‌ـنزدیک‌ترین همسایه، هیچ فرضی در مورد نحوه توزیع داده‌ها ندارد. از این‌رو، عدم نگرانی در مورد توزیع یک مزیت بزرگ است. این بدان معنا است که KNN را می‌توان برروی مجموعه داده‌های مختلف اعمال کرد.

## نحوه کار الگوریتم کا‌-نزدیک‌ترین همسایه

الگوریتم زیر نحوه کار KNN را نشان می‌دهد:

**الگوریتم کا‌ـ نزدیک‌ترین همسایه:**

**مرحله ۱:** عدد K همسایه را انتخاب کنید.

**مرحله ۲:** فاصله اقلیدسی (یا معیارهای فاصله‌ای دیگر) K تعداد همسایه را محاسبه کنید.

**مرحله ۳:** فاصله را مرتب کنید و نزدیکترین همسایگان K را بر اساس حداقل فاصله اقلیدسی محاسبه شده تعیین کنید.

**مرحله ٤:** از میان این K همسایه، تعداد نقاط داده در هر دسته را بشمارید.

**مرحله ۵:** نقاط داده جدید را به دسته‌ای که تعداد همسایه‌ها حداکثر است، اختصاص دهید.

الگوریتم KNN براساس نوع یادگیری به‌صورت زیر است:

- **یادگیری مبتنی‌بر نمونه:** در این روش ما وزن‌ها را از داده‌های آموزشی برای پیش‌بینی خروجی یاد نمی‌گیریم (همانند الگوریتم‌های مبتنی‌بر مدل) بلکه از کل نمونه‌های آموزشی برای پیش‌بینی خروجی برای داده‌های دیده نشده استفاده می‌کنیم.

- **یادگیری تنبل:** مدل با استفاده از داده‌های آموزشی قبلی یاد گرفته نمی‌شود و فرآیند یادگیری به زمانی موکول می‌شود که پیش‌بینی در نمونه جدید درخواست می‌شود.

- **ناپارامتری:** در KNN، هیچ شکل از پیش تعریف‌شده‌ای از تابع نگاشت وجود ندارد.

## مزایا

- اجرای آن ساده است.
- زمان آموزش صفر (یا خیلی کم)
- هیچ فرضی در مورد نحوه توزیع داده‌ها ندارد.
- درک الگوریتم KNN برای مبتدیان در یادگیری ماشین بسیار آسان است.

## معایب

- همیشه نیاز به تعیین مقدار K دارد که ممکن است برخی اوقات پیچیده باشد.
- هزینه محاسبات به دلیل محاسبه فاصله بین نقاط داده برای همه نمونه‌های آموزشی بالا است.
- برروی داده‌های نامتوازن عملکرد خوبی ندارد. بنابراین، داده‌هایی که فراوانی کم‌تری دارند ممکن است به اشتباه گروه‌بندی شوند.

## کا-نزدیک‌ترین همسایه در پایتون

در این بخش، خواهیم دید که چگونه می‌توان از کتابخانه Scikit-Learn پایتون برای پیاده‌سازی الگوریتم KNN استفاده کرد.

### مجموعه داده

ما از مجموعه داده iris[1] معروف برای مثال KNN خود استفاده می‌کنیم. این مجموعه داده شامل چهار ویژگی است: عرض کاسبرگ، طول کاسبرگ، عرض گلبرگ و طول گلبرگ. این‌ها ویژگی‌های انواع خاصی از گیاه زنبق است. وظیفه، پیش‌بینی دسته‌ای است که این گیاهان به آن تعلق دارند. سه کلاس در مجموعه داده وجود دارد: Iris-setosa، Iris-versicolor و Iris-virginica.

### وارد کردن کتاب‌خانه‌ها

```
In [1]:   import numpy as np
          import matplotlib.pyplot as plt
          import pandas as pd
```

### وارد کردن مجموعه داده

برای وارد کردن مجموعه داده و بارگذاری آن در قالب داده pandas، کد زیر را اجرا کنید:

```
In [2]:   url = "https://archive.ics.uci.edu/ml/machine-learning-
          databases/iris/iris.data"

          # Assign colum names to the dataset
          names = ['sepal-length', 'sepal-width', 'petal-length',
          'petal-width', 'Class']

          # Read dataset to pandas dataframe
          dataset = pd.read_csv(url, names=names)
```

برای اینکه ببینید مجموعه داده واقعا چه شکلی دارد، دستور زیر را اجرا کنید:

```
In [3]:   dataset.head(8)
```

اجرای کد بالا ۸ ردیف اول مجموعه داده را مطابق شکل صفحه بعد نمایش می‌دهد:

---

[1] https://archive.ics.uci.edu/ml/datasets/Iris

| | sepal-length | sepal-width | petal-length | petal-width | Class |
|---|---|---|---|---|---|
| 0 | 5.1 | 3.5 | 1.4 | 0.2 | Iris-setosa |
| 1 | 4.9 | 3.0 | 1.4 | 0.2 | Iris-setosa |
| 2 | 4.7 | 3.2 | 1.3 | 0.2 | Iris-setosa |
| 3 | 4.6 | 3.1 | 1.5 | 0.2 | Iris-setosa |
| 4 | 5.0 | 3.6 | 1.4 | 0.2 | Iris-setosa |
| 5 | 5.4 | 3.9 | 1.7 | 0.4 | Iris-setosa |
| 6 | 4.6 | 3.4 | 1.4 | 0.3 | Iris-setosa |
| 7 | 5.0 | 3.4 | 1.5 | 0.2 | Iris-setosa |

**پیش‌پردازش**

گام بعدی این است که مجموعه داده خود را به ویژگی‌ها و برچسب‌های آن تقسیم کنیم. برای این کار از کد زیر استفاده کنید:

```
In [4]:  X = dataset.iloc[:, :-1].values
         y = dataset.iloc[:, 4].values
```

متغیر X شامل چهار ستون اول مجموعه داده (یعنی ویژگی‌ها) است در حالی که y حاوی برچسب‌ها است.

**تقسیم مجموعه داده**

در گام بعدی مجموعه داده خود را به دو بخش آموزشی و آزمایشی تقسیم می‌کنیم که به ما ایده بهتری درباره نحوه عملکرد الگوریتم در مرحله آزمایش می‌دهد. به این ترتیب الگوریتم ما بر روی داده‌های دیده نشده آزمایش می‌شود.

برای تقسیم‌سازی داده‌ها به دو قسمت آموزشی و آزمایشی، کد زیر را اجرا کنید:

```
In [5]:  from sklearn.model_selection import train_test_split

         X_train, X_test, y_train, y_test = train_test_split(X, y,
         test_size =0.25, random_state=42)
```

کد فوق مجموعه داده را به ۷۵ درصد داده‌های آموزشی و ۲۵ درصد داده‌های آزمایشی تقسیم می‌کند. این بدان معناست که از مجموع ۱۵۰ رکورد، مجموعه آموزشی شامل ۱۱۲ رکورد و مجموعه آزمون شامل ۳۸ رکورد خواهد بود:

```
In [6]:  X_train.shape
Out [6]: (112, 4)
In [7]:  X_test.shape
Out [7]: (38, 4)
```

در کد پیشین عدد ٤ نمایانگر تعداد ویژگی‌ها است.

## مقیاس‌بندی ویژگی‌ها

قبل از انجام هر گونه پیش‌بینی واقعی، همیشه بهتر است که ویژگی‌ها را مقیاس‌بندی کنید. کد زیر مقیاس‌بندی ویژگی‌ها را انجام می‌دهد:

```
In [1]:   from sklearn.preprocessing import StandardScaler
          scaler = StandardScaler()
          scaler.fit(X_train)

          X_train = scaler.transform(X_train)
          X_test = scaler.transform(X_test)
```

## آموزش و پیش‌بینی

آموزش الگوریتم KNN و پیش‌بینی با آن، هنگام استفاده از Scikit-Learn، بسیار ساده است:

```
In [1]:   from sklearn.neighbors import KNeighborsClassifier
          classifier = KNeighborsClassifier(n_neighbors=5)
          classifier.fit(X_train, y_train)
```

اولین قدم وارد کردن کلاس KNeighborsClassifier از کتابخانه sklearn.neighbors است. در خط دوم، این کلاس با یک پارامتر، یعنی n_neigbours مقداردهی اولیه می‌شود که همان مقدار K است. پیش‌تر بیان شد هیچ مقدار ایده‌آلی برای K وجود ندارد و پس از آزمایش و ارزیابی انتخاب می‌شود. با این حال، برای شروع، ٥ رایج‌ترین مقدار مورد استفاده برای الگوریتم KNN باشد.

مرحله آخر این است که مدل ساخته‌شده را برروی داده‌های آزمایشی خود پیش‌بینی کنیم. برای انجام این کار، کد زیر را اجرا کنید:

```
In [1]:   y_pred = classifier.predict(X_test)
```

## ارزیابی الگوریتم

برای ارزیابی یک الگوریتم، همچنان‌که پیش‌تر بیان شد، ماتریس درهم‌ریختگی، دقت، فراخوانی و امتیاز F1 رایج‌ترین معیارهای مورد استفاده هستند. برای محاسبه این معیارها می‌توان از confusion_matrix و classification_report استفاده کرد. به کد زیر نگاه کنید:

```
In [3]:   from sklearn.metrics import classification_report,
          confusion_matrix
          print(confusion_matrix(y_test, y_pred))
          print(classification_report(y_test, y_pred))
```

Out [3]:
```
[[15  0  0]
 [ 0 11  0]
 [ 0  0 12]]
```

|                 | precision | recall | f1-score | support |
|-----------------|-----------|--------|----------|---------|
| Iris-setosa     | 1.00      | 1.00   | 1.00     | 15      |
| Iris-versicolor | 1.00      | 1.00   | 1.00     | 11      |
| Iris-virginica  | 1.00      | 1.00   | 1.00     | 12      |
|                 |           |        |          |         |
| accuracy        |           |        | 1.00     | 38      |
| macro avg       | 1.00      | 1.00   | 1.00     | 38      |
| weighted avg    | 1.00      | 1.00   | 1.00     | 38      |

نتایج نشان می‌دهد که مدل KNN ما قادر است تمام ۳۸ رکورد موجود در مجموعه آزمون را با دقت ۱۰۰٪ دسته‌بندی کند که بسیار عالی است. اگرچه الگوریتم با این مجموعه داده بسیار خوب عمل کرد، انتظار یکسان بودن چنین نتایجی را برای همه مجموعه داده‌ها نداشته باشید!!

**مقایسه میزان خطا با مقدار K**

در بخش آموزش و پیش‌بینی گفتیم که هیچ راهی وجود ندارد که از قبل بدانیم کدام مقدار K که در اولین اقدام بهترین نتیجه را دارد. ما به طور تصادفی ۵ را به عنوان مقدار K انتخاب کردیم و اتفاقا به دقت ۱۰۰٪ منجر شد. یکی از راه‌هایی که به شما کمک می‌کند بهترین مقدار K را پیدا کنید، رسم نمودار مقدار K و نرخ خطای مربوط، برای مجموعه داده است.

از این‌رو، در این بخش، میانگین خطای مقادیر پیش‌بینی‌شده مجموعه آزمون را برای همه مقادیر K بین ۱ تا ۳۰ رسم می‌کنیم. در این راستا، ابتدا میانگین خطا را برای همه مقادیر پیش‌بینی‌شده که در آن K از ۱ تا ۳۰ متغیر است محاسبه کنیم. کد زیر را اجراکنید:

In [1]:
```
error = []

# Calculating error for K values between 1 and 40
for i in range(1, 40):
    knn = KNeighborsClassifier(n_neighbors=i)
    knn.fit(X_train, y_train)
    pred_i = knn.predict(X_test)
    error.append(np.mean(pred_i != y_test))
```

کد بالا یک حلقه از ۱ تا ۳۰ را اجرا می‌کند و در هر تکرار میانگین خطای مقادیر پیش‌بینی شده مجموعه آزمون محاسبه می‌شود و نتیجه به لیست خطا اضافه می‌شود.

مرحله بعدی مصورسازی مقادیر خطا در برابر مقادیر K است. کد زیر را برای ایجاد نمودار اجرا کنید:

```
In [1]:    plt.figure(figsize=(12, 6))
           plt.plot(range(1, 30), error, color='red',
           linestyle='dashed', marker='o',
                marker facecolor='blue', markersize=10)
           plt.title('Error Rate K Value')
           plt.xlabel('K Value')
           plt.ylabel('Mean Error')
```

نمودار خروجی به‌صورت زیر است:

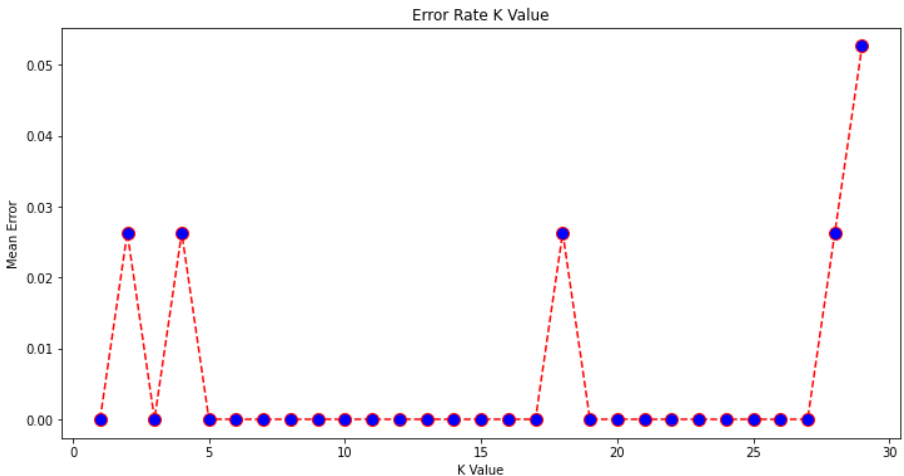

از طریق این نمودار خروجی می‌توانیم بهترین و بدترین مقادیر K انتخابی را مشاهده کنیم. توصیه می‌شود حتما مدل را با مقدار مختلف K آزمایش کنید تا ببینید چگونه بر دقت پیش‌بینی‌ها تاثیر می‌گذارد.

## ماشین بردار پشتیبان

ماشین‌های بردار پشتیبان یا به اختصار SVM در زیرمجموعه روش‌های یادگیری بانظارت قرار می‌گیرند و برای مسائل دسته‌بندی، رگرسیون و تشخیص نقاط دورافتاده استفاده می‌شوند. ماشین‌های بردار پشتیبان با دیگر الگوریتم‌های طبقه‌بندی متفاوت هستند، زیرا آن‌ها مرز تصمیم را انتخاب می‌کنند که فاصله را از نزدیک‌ترین نقاط داده همه کلاس‌ها به حداکثر می‌رساند. مرز تصمیم ایجاد شده توسط ماشین‌های بردار پشتیبان، **حاشیه‌ی حداکثری** یا **ابَرصفحه حاشیه‌ای حداکثری** نامیده می‌شود. یک دسته‌بند SVM خطی ساده با ایجاد یک خط مستقیم (ابرصفحه جداساز) بین دو کلاس کار می‌کند. این بدان معناست که تمام نقاط داده در یک طرف خط، یک

دسته را نشان می‌دهند و نقاط داده در طرف دیگر خط، دسته متفاوتی را نشان می‌دهند. به‌طور شهودی مشخص است که می‌توان تعداد بی‌نهایتی از خطوط را انتخاب کرد. چیزی که الگوریتم SVM خطی را بهتر از برخی از الگوریتم‌های دیگر، همانند کا ـ نزدیک‌ترین همسایه می‌کند، این است که بهترین خط را برای دسته‌بندی نقاط داده شما انتخاب می‌کند.

یک مثال دو بعدی به درک این مورد بهتر کمک می‌کند. فرض کنید شما چند نقطه داده دارید. شما سعی می‌کنید این نقاط داده را بر اساس دسته‌ای که باید در آن قرار گیرند جدا کنید، اما نمی‌خواهید هیچ داده‌ای را در دسته‌ی اشتباه داشته باشید. این بدان معناست که شما در تلاش برای یافتن خطی بین دو نزدیکترین نقطه هستید که سایر نقاط داده را از هم جدا نگه می‌دارد. بنابراین دو نزدیکترین نقطه داده، بردارهای پشتیبانی را به شما می‌دهند که برای یافتن آن خط از آن‌ها استفاده خواهید کرد.

## فرم اولیه و فرم دوگانه (ثانویه)

مسائل بهینه‌سازی را می‌توان به دو فرم مختلف تعریف کرد: مساله اولیه و مساله دوگانه. مزیت این امر این است که گاهی اوقات حل مساله دوگانه آسان‌تر از مساله اولیه است. با این حال، راه حل‌های مساله اولیه و مساله دوگانه ممکن است متفاوت باشد، اما در شرایط خاص، راه حل‌ها برابر هستند. ماشین بردار پشتیبان نیز از آنجایی که یک مساله بهینه‌سازی است، به دو صورت دوگانه و اولیه قابل تعریف است. هر دو نتیجه بهینه‌سازی یکسانی را دریافت می‌کنند، اما نحوه دریافت آن بسیار متفاوت است. قبل از اینکه عمیقا به ریاضیات آن بپردازیم، اجازه دهید به شما بگویم که کدام یک در چه زمانی استفاده می‌شود. فرم اولیه زمانی ترجیح داده می‌شود که نیازی به اعمال **ترفند هسته**[1] برای داده‌ها نداشته باشیم و مجموعه داده بزرگ است، اما ابعاد هر نقطه داده کوچک است. در مقابل، هنگامی که داده‌ها ابعاد بزرگی دارند و ما نیاز به استفاده از ترفند هسته داریم، فرم دوگانه ترجیح داده می‌شود.

## ماشین بردار پشتیبان با حاشیه سخت

با توجه به مجموعه آموزشی $D = \{(x_1, y_1), (x_2, y_2), \ldots, (x_m, y_m)\}$، جایی که $y_m \in \{+1, -1\}$ است، ایده اصلی ماشین بردار پشتیبان این است که از مجموعه آموزشی $D$ استفاده کند تا یک اَبَرصفحه جداساز در فضای نمونه پیدا شود که می‌تواند نمونه‌های کلاس‌های مختلف را جدا کند. با این حال، ممکن است چندین اَبَرصفحه جداساز واجد شرایط، وجود داشته باشد، همان‌طور که شکل ۶ ـ ۴ نشان داده شده است. از این‌رو، کدام یک باید انتخاب شود؟

---

[1] kernel trick

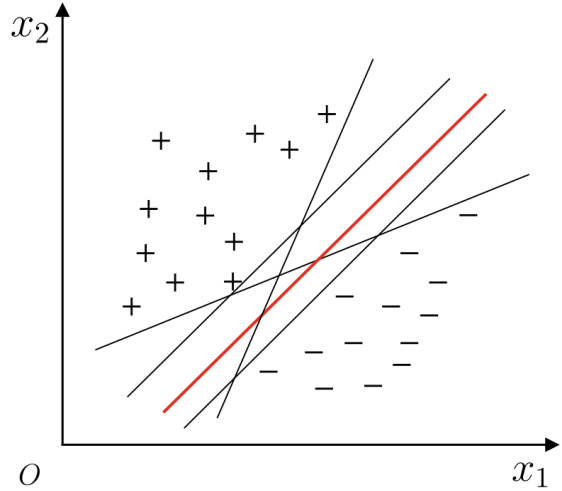

**شکل ۶ ـ ٤.** بیش از یک ابرصفحه می‌تواند نمونه‌های آموزشی را جدا کند.

ما باید از بین این ابرصفحه‌ها موردی را انتخاب کنیم که در وسط دو کلاس قرار دارد، یعنی قرمز (شکل ۶ ـ ٤). به دلیل اینکه، این ابرصفحه جداساز بهترین "**حد مجاز خطا**"[1] را نسبت به انحراف[2] داده‌های محلی دارد. به عنوان مثال، نمونه‌هایی که در مجموعه آموزشی نیستند می‌توانند به دلیل نویز یا محدودیت‌های مجموعه آموزشی به مرز تصمیم‌گیری نزدیک شوند. در نتیجه، بسیاری از ابرصفحه‌های جداسازی که به خوبی در مجموعه آموزشی عمل می‌کنند، اشتباهاتی را مرتکب خواهند شد، در حالی که ابرصفحه جداساز قرمز کمتر احتمال دارد تحت تاثیر قرار گیرد. به عبارت دیگر، این ابرصفحه جداساز دارای قوی‌ترین توانایی تعمیم در دسته‌بندی است.

یک ابرصفحه جداساز در فضای نمونه را می‌توان به صورت تابع خطی زیر نشان داد:

$$w^T x + b = ٠$$

جایی‌که $w = \{w_۱; w_۲; \dots ; w_d\}$ یک بردار طبیعی است که جهت ابرصفحه را کنترل می‌کند و $b$ بایاس است که فاصله بین ابرصفحه و مبدا را کنترل می‌کند. بردار نرمال $w$ و بایاس $b$، ابرصفحه جداسازی را تعیین می‌کند که با $(w, b)$ نشان داده می‌شود. فاصله از هر نقطه $x$ در فضای نمونه به ابرصفحه $(w, b)$ را می‌توان به صورت زیر نوشت:

$$r = \frac{|w^T x + b|}{\|w\|}$$

---

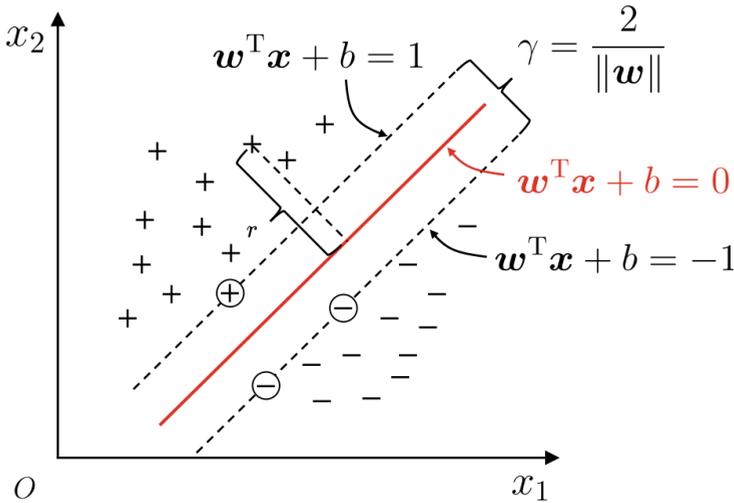

**شکل ٦ ـ ٥.** بردارهای پشتیبان و حاشیه

فرض کنید ابرصفحه $(w, b)$ می‌تواند نمونه‌های آموزشی را به درستی دسته‌بندی کند، یعنی، برای $(x_i, y_i) \in D$، $w^T x_i + b > 0$ وجود دارد که $y_i = +1$ و $w^T x_i + b < 0$ زمانی که $y_i = -1$ است. اگر داشته باشیم:

$$w^T x_i + b \geq +1, y_i = +1,$$
$$w^T x_i + b \leq -1, y_i = -1. \tag{٦ـ١}$$

همان‌طور که شکل ٦ ـ ٥ نشان داده شده است، معادله در (٦ـ١) برای نقاط نمونه نزدیک به ابرصفحه برقرار است. این نقاط نمونه **بردار پشتیبان** نامیده می‌شوند. مجموع دو بردار پشتیبان از کلاس‌های مختلف تا ابرصفحه برابر

$$\gamma = \frac{2}{\|w\|}$$

است که به آن **حاشیه** می‌گویند.

یافتن ابرصفحه جداساز با حداکثر حاشیه، معادل یافتن پارامترهای $w$ و $b$ است که $\gamma$ را با توجه به محدودیت‌های (٦ـ١) به حداکثر می‌رسانند، یعنی:

$$max_{w,b} = \frac{2}{\|w\|} \tag{٦ـ٢}$$

به نحوی که $y_i(w^T x_i + b) \geq 1$، $i = 1, 2, ..., m$.

حاشیه را می‌توان با بیشینه کردن $\|w\|^{-1}$ که معادل با کمینه کردن $\|w\|^2$ است، بهینه کرد. از این‌رو، می‌توانیم معادله (٦ـ٢) را بازنویسی کنیم:

$$min_{w,b} \frac{\backslash}{\gamma} \|w\|^{\gamma} \qquad\qquad (\gamma - \gamma)$$

به نحوی که $۱ \leq y_i(w^T x_i + b)$ و $i = ۱, ۲, ..., m$.

این **فرم اولیه ماشین بردار پشتیبان** نامیده می‌شود. می‌توانیم با معرفی ضرایب لاگرانژ $\alpha_i$ و تبدیل آن به مساله دوگانه این مساله را حل کنیم:

$$L(w, b, \alpha) = \frac{\backslash}{\gamma} w^t w - \sum_{i=\backslash}^{n} \alpha_i (\backslash - y_i(w^T x_i + b))$$

این تابع لاگرانژی ماشین بردار پشتیبان نامیده می‌شود که در آن $b$ و $w$ پارمترهای مدل و $\alpha = (\alpha_\backslash; \alpha_m; ...; \alpha_m)$ و با $w$ و $b$ مشتق‌پذیر است:

$$\nabla_w L(w, b, \alpha) = \cdot \Rightarrow w = \sum_{i=\backslash}^{m} \alpha_i y_i x_i$$

$$\nabla_b L(w, b, \alpha) = \cdot \Rightarrow \sum_{i=\backslash}^{m} \alpha_i y_i = \cdot$$

با جایگزین کردن آن‌ها در $L(w, b, \alpha)$، $w$ را از آن حذف می‌کند و مساله دوگانه ماشین بردار پشتیبان را دریافت خواهیم کرد:

$$max_\alpha \sum_{i=\backslash}^{m} \alpha_i - \frac{\backslash}{\gamma} \sum_{i=\backslash}^{m} \sum_{j=\backslash}^{m} \alpha_i \alpha_j \, y_i y_j x_i^T x_j$$

$$s.t. \quad \sum_{i=\backslash}^{m} \alpha_i y_i = \cdot,$$

$$\alpha_i \geq \cdot, i = ۱, ۲, ..., m.$$

با حل این مساله بهینه‌سازی، $\alpha$ و متعاقبا $w$ و $b$ بدست می‌آید. از این‌رو می‌توانیم دسته‌بندی نهایی یا تابع تصمیم‌گیری را به صورت زیر بنویسیم:

$$f(x) = sign[\sum_{i=1}^{m} \alpha_i \, y_i x_i^T x + b]$$

که در آن $sign(x)$ **تابع علامت** است.

*حل مساله دوگانه آسان‌تر است، چراکه فقط ضریب لاگرانژ دارد.*

## ماشین بردار پشتیبان برای داده‌های غیرخطی قابل تفکیک

تا اینجا فرض بر این بود که نمونه‌های آموزشی به صورت خطی قابل تفکیک هستند، یعنی ابرصفحه‌هایی وجود دارند که می‌توانند همه نمونه‌های آموزشی را به درستی دسته‌بندی کنند. با این حال، این فرض اغلب در عمل صادق نیست. در واقع، بیشتر مسائل غیرخطی هستند و نمی‌توان از SVM خطی قبلی برای حل آن‌ها استفاده کرد. در این صورت برای حل آن باید چه اقدامی کرد؟ در وضع مطلوب، ما باید یک تبدیل غیرخطی $\varphi$ پیدا کنیم به طوری‌که داده‌ها را بتوان در یک فضای ویژگی با ابعاد بالا ترسیم کرد که در آن امکان دسته‌بندی خطی وجود دارد.

به شکل ۶ ـ ۶ نگاه کنید. کلاس‌ها با استفاده از دو متغیر پیش‌بینی‌کننده به صورت خطی قابل تفکیک نیستند. الگوریتم ماشین بردار پشتیبان یک بعد اضافی به داده‌ها اضافه می‌کند، به طوری که یک ابرصفحه خطی می‌تواند کلاس‌ها را در این فضای جدید و ابعاد بالاتر جدا کند. می‌توانیم این را به‌عنوان نوعی تغییر شکل یا کشش فضای ویژگی تصور کنیم. این بعد اضافی **هسته** نامیده می‌شود. هسته‌ها راهی برای حل مسائل غیرخطی با کمک دسته‌بندهای خطی هستند. به این ایده **تدبیر هسته (kernel trick)** گویند.

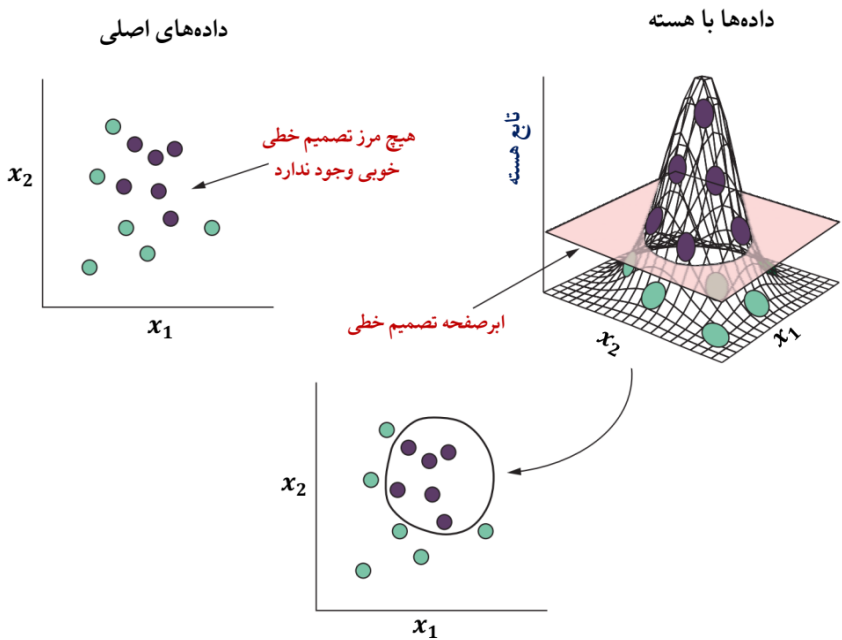

**شکل ۶ ـ ۶ .** الگوریتم **SVM** یک بعد اضافی برای جداسازی خطی داده‌ها اضافه می‌کند. کلاس‌های موجود در داده‌های اصلی به صورت خطی قابل تفکیک نیستند. الگوریتم SVM یک بعد اضافی اضافه می‌کند که در یک فضای ویژگی دو بعدی، می‌تواند به عنوان "کشش" داده‌ها به بعد سوم نشان داده شود. این بعد اضافی اجازه می‌دهد تا داده‌ها به صورت خطی از هم قابل تفکیک شوند.

**حال، پرسش اینجاست که الگوریتم چگونه این هسته جدید را پیدا می‌کند؟** پاسخ، از یک تبدیل ریاضی به داده‌ها به نام **تابع هسته** استفاده می‌کند. توابع هسته زیادی برای انتخاب وجود دارد که هر کدام تبدیل متفاوتی را به داده‌ها اعمال می‌کنند و برای یافتن مرزهای تصمیم‌گیری خطی برای موقعیت‌های مختلف مناسب هستند. شکل ۶-۷ نمونه‌هایی از موقعیت‌هایی را نشان می‌دهد که برخی از توابع رایج هسته می‌توانند داده‌های غیرخطی قابل تفکیک را از هم جدا کنند.

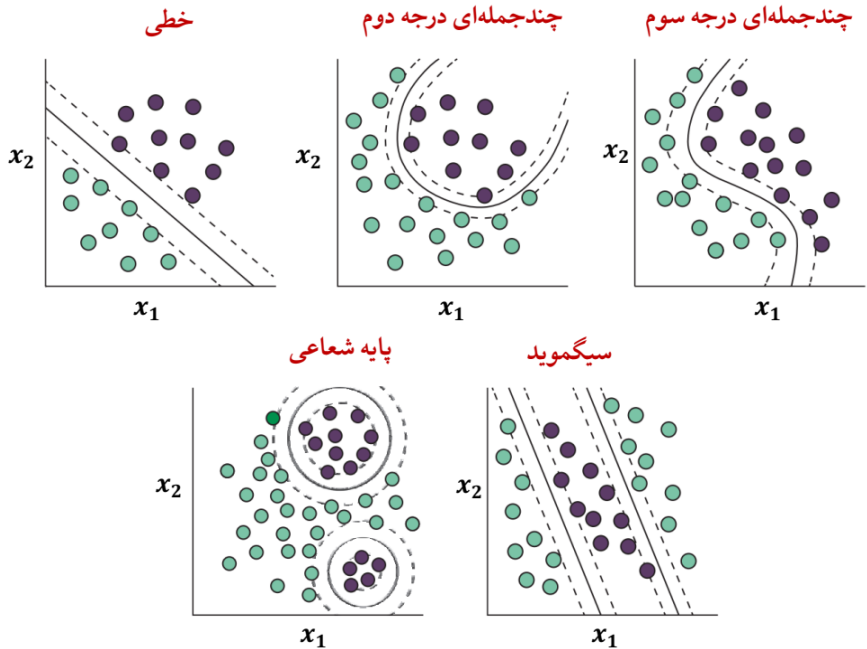

**شکل ۶-۷.** نمونه‌هایی از توابع هسته

## تابع هسته (کرنل)

اگر $\varphi(x)$ بردار ویژگیِ نگاشت شدهٔ $x$ را نشان دهد، از این‌رو مدل ابرصفحه جداساز در فضای ویژگی را می‌توان به صورت بیان کرد:

$$f(x) = w^T \varphi(x) + b$$

که در آن $w$ و $b$ پارمترهای مدل هستند. مشابه معادله (۶-۳) داریم:

$$min_{w,b} \frac{1}{2} \|w\|^2$$

به نحوی (قسمی) که $\geq 1$ $y_i(w^T \varphi(x_i) + b)$ و $i = 1, 2, \ldots, m$.

همچنین مساله دوگانه آن برابر است با:

$$max_\alpha \sum_{i=1}^{m} \alpha_i - \frac{1}{2} \sum_{i=1}^{m} \sum_{j=1}^{m} \alpha_i \alpha_j \, y_i y_j \varphi(x_i^T)\varphi(x_j) \qquad (6-4)$$

$$s.t. \quad \sum_{i=1}^{m} \alpha_i y_i = 0,$$

$$\alpha_i \geq 0, i = 1, 2, \dots, m.$$

حل معادله (۴ـ۶) شامل محاسبه $\varphi(x_i^T)\,\varphi(x_j)$ است که حاصل ضرب داخلی بردارهای ویژگی نگاشت شده $x_i$ و $x_j$ است. از آنجایی که فضای ویژگی نگاشت‌شده می‌تواند ابعاد بسیار بالا یا حتی بی‌نهایت داشته باشد، محاسبه $\varphi(x_i^T)\,\varphi(x_j)$ به‌طور مستقیم اغلب دشوار است. برای جلوگیری از این مشکل، فرض می‌کنیم تابعی به شکل زیر وجود دارد:

$$k(x_i, x_j) = \langle \varphi(x_i), \varphi(x_j) \rangle = \varphi(x_i^T)\varphi(x_j))$$

که می‌گوید ضرب داخلی $x_i$ و $x_j$ در فضای ویژگی را می‌توان در فضای نمونه با استفاده از تابع $k(0,0)$ محاسبه کرد. با چنین تابعی، دیگر نیازی به محاسبه ضرب داخلی در فضای ویژگی نداریم. از این‌رو می‌توانیم معادله (۴ـ۶) را به صورت زیر بازنویسی کنیم:

$$max_\alpha \sum_{i=1}^{m} \alpha_i - \frac{1}{2} \sum_{i=1}^{m} \sum_{j=1}^{m} \alpha_i \alpha_j \, y_i y_j k(x_i, x_j)$$

$$s.t. \quad \sum_{i=1}^{m} \alpha_i y_i = 0,$$

$$\alpha_i \geq 0, i = 1, 2, \dots, m.$$

از حل این معادله خواهیم داشت:

$$f(x) = w^T \varphi(x) + b$$

$$= sign[\sum_{i=1}^{m} \alpha_i \, y_i \varphi(x_i^T)\varphi(x) + b]$$

$$= sign[\sum_{i=1}^{m} \alpha_i \, y_i k(x_i, x_j) + b]$$

که در آن تابع $k(0,0)$ هسته است.

از آنجایی که ما می‌خواهیم نمونه‌ها به صورت خطی در فضای ویژگی قابل تفکیک باشند، کیفیت فضای ویژگی برای عملکرد ماشین‌های بردار پشتیبان حیاتی است. با این حال، ما نمی‌دانیم کدام توابع هسته خوب هستند، چرا که ما نگاشت ویژگی را نمی‌دانیم. بنابراین، انتخاب هسته بزرگ‌ترین عدم قطعیت ماشین‌های بردار پشتیبان است. یک هسته ضعیف نمونه‌ها را به یک فضای ویژگی ضعیف نگاشت می‌کند و در نتیجه عملکرد ضعیفی دارد.

*به عبارت دیگر، نوع تابع هسته برای یک مساله معین از داده‌ها یاد گرفته نمی‌شود و باید آن را مشخص کنیم. از این رو، انتخاب تابع هسته یک ابرپارامتر است. در نتیجه، بهترین رویکرد برای انتخابِ تابع هسته با بهترین عملکرد، تنظیم ابرپارامتر است.*

## انواع توابع هسته

در ادامه برخی از توابع هسته‌ای که در SVM استفاده می‌شود، فهرست شده است:

- **هسته خطی.** فرض کنید دو بردار به نام‌های $x_i$ و $x_j$ داریم، هسته خطی با ضرب داخلی این دو بردار تعریف می‌شود:

$$k(x_i, x_j) = x_i . x_j$$

- **هسته چندجمله‌ای.** هسته چندجمله ای با معادله زیر تعریف می‌شود:

$$k(x_i, x_j) = (x_i . x_j)^d$$

که در آن $d$ درجه چندجمله‌ای است.

- **هسته گاوسی.** معادله هسته گوسی به‌صورت زیر است:

$$k(x_i, x_j) = \exp\left(-\frac{\|x_i - x_j\|^2}{2\sigma^2}\right)$$

سیگمای داده شده نقش بسیار مهمی در عملکرد هسته گاوسی ایفا می‌کند و باید با دقت و با توجه به مساله تنظیم شود.

- **هسته لاپلاسین.** معادله هسته لاپلاسین به‌صورت زیر است:

$$k(x_i, x_j) = \exp\left(-\frac{\|x_i - x_j\|}{\sigma}\right)$$

- **هسته هذلولی‌گون یا سیگموید.** این هسته بیشتر در شبکه‌های عصبی استفاده می‌شود و معادله آن به‌صورت زیر است:

$$k(x_i, x_j) = \tanh(\alpha x^T y + c)$$

## چرا تدبیر هسته مهم است؟

همان‌طور که پیش‌تر بیان شد و در شکل ۵-۶ مشاهده شد، اگر راهی برای نگاشت داده‌ها از فضای دو بعدی به فضای سه بعدی پیدا کنیم، می‌توانیم یک مرز تصمیم‌گیری را پیدا کنیم که

می‌تواند کلاس‌های مختلف را دسته‌بندی کند. اولین تفکر برای حل این مساله در مورد فرآیند تبدیل داده‌ها این است که تمام نقاط داده را به یک بعد بالاتر (در این مورد، ۳ بعد) نگاشت کنم، مرز را پیدا کرده و دسته‌بندی را انجام دهم. ایده کاملا درست به نظر می‌رسد. با این حال، هنگامی که ابعاد داده‌ها بیشتر شود، محاسبات در آن فضا بیشتر بسیار بیشتر می‌شود. این‌جاست که تدبیر هسته کارآمد می‌شود. چراکه به ما این امکان را می‌دهد تا در فضای ویژگی اصلی بدون محاسبه مختصات داده‌ها در فضای ابعاد بالاتر عمل کنیم. برای درک بهتر این موضوع بیایید یک مثال ببینیم:

$$x = (x_1, x_2, x_3)^T$$
$$y = (y_1, y_2, y_3)^T$$

در اینجا $x$ و $y$ دو نقطه داده در فضای ۳ بعدی هستند. فرض کنیم که باید $x$ و $y$ را به فضای ۹ بعدی نگاشت شوند. از این‌رو، باید محاسبات زیر را انجام دهیم:

$$\varphi(x) = (x_1^2, x_1 x_2, x_1 x_3, x_2 x_1, x_2^2, x_2 x_3, x_3 x_1, x_3 x_2, x_3^2)^T$$
$$\varphi(y) = (y_1^2, y_1 y_2, y_1 y_3, y_2 y_1, y_2^2, y_2 y_3, y_3 y_1, y_3 y_2, y_3^2)^T$$

$$\varphi(x)^T \varphi(y) = \sum_{i,j=1}^{3} x_i x_j y_i y_j$$

تا به نتیجه نهایی برسیم. پیچیدگی محاسباتی، در این مورد، $O(n^2)$ است. حال اگر از تابع هسته ($K(x, y)$) به جای انجام محاسبات پیچیده در فضای ۹ بعدی استفاده کنیم، با محاسبه ضرب داخلی $x$ ـ ترانهاده و $y$ به‌همان نتیجه در فضای ۳ بعدی می‌رسیم که پیچیدگی محاسباتی، در این مورد، $O(n)$ است:

$$k(x, y) = (x^T y)^2$$
$$= (x_1 y_1 + x_2 y_2 + x_3 y_3)^2$$
$$= \sum_{i,j=1}^{3} x_i x_j y_i y_j$$

برای درک بهتر این موضوع مثال عددی زیر را در نظر بگیرید:

اگر دو نقطه در فضای ۳بعدی به‌صورت زیر داشته باشیم:

$$x = (2,3,4)$$
$$y = (3,4,5)$$

ابتدا $\varphi(x)$ و $\varphi(y)$ را محاسبه می‌کنیم:

$$\varphi(\Upsilon, \Upsilon, \xi) = (\xi, \Upsilon, \Lambda, \Upsilon, \Upsilon, \Upsilon, \Lambda, \Upsilon, \Upsilon, \Upsilon)$$

$$\varphi(\Upsilon, \xi, \circ) = (\Upsilon, \Upsilon, \Upsilon\circ, \Upsilon, \Upsilon, \Upsilon\circ, \Upsilon\circ, \Upsilon\circ, \Upsilon\circ)$$

$$\varphi(x).\varphi(y) = \varphi(\Upsilon, \Upsilon, \xi).\varphi(\Upsilon, \xi, \circ)$$

$$(\Upsilon\Upsilon + \Upsilon\Upsilon + \Upsilon\Upsilon\circ + \Upsilon\Upsilon + \Upsilon\xi\xi + \Upsilon\xi\circ + \Upsilon\Upsilon\circ + \Upsilon\xi\circ + \xi\circ\circ) = \Upsilon\xi\xi\xi$$

و محاسبه $k(x, y)$ آن برابر است با:

$$k(x, y) = (\Upsilon * \Upsilon + \Upsilon * \xi + \xi * \circ)^{\Upsilon}$$

$$= (\Upsilon + \Upsilon\Upsilon + \Upsilon\circ)^{\Upsilon}$$

$$= \Upsilon\Lambda * \Upsilon\Lambda$$

$$= \Upsilon\xi\xi\xi$$

همان‌طورکه متوجه شدیم، هر دو نتیجه یکسانی را به ما می‌دهند، اما روش استفاده از هسته نیاز به محاسبات کم‌تری دارد.

## ماشین بردار پشتیبان با حاشیه نرم

تاکنون فرض می‌کردیم که نمونه‌ها به‌صورت خطی در **فضای نمونه** یا **فضای ویژگی** قابل تفکیک هستند. با این حال، اغلب یافتن یک تابع هسته مناسب برای جداسازی خطی نمونه‌های آموزشی در فضای ویژگی دشوار است. این را می‌توان به این دلیل نسبت داد که معمولاً ویژگی‌هایی را که از داده‌ها بدست می‌آوریم حاوی اطلاعات کافی نیستند تا بتوانیم به وضوح کلاس‌ها را از یکدیگر جدا کنیم (معمولاً در بسیاری از برنامه‌های کاربردی دنیای واقعی چنین وضعیتی وجود دارد). حتی اگر چنین تابع هسته‌ای را پیدا کنیم، به سختی می‌توان تشخیص داد که آیا این نتیجه *بیش‌برازش* شده است یا خیر.

یکی از راه‌های کاهش این وضعیت این است که به SVM اجازه دهید تعداد مشخصی اشتباه را برروی نمونه‌ها مرتکب شود تا سایر نقاط همچنان بدرستی دسته‌بندی شوند. این ایده با مفهوم **حاشیه نرم** اجرا می‌شود، شکل ۸ ـ ۶ این ایده را نشان می‌دهد. به‌طور خلاصه انگیزه استفاده از این روش به دو دلیل است:

۱. همان‌طورکه پیش‌تر بیان شد، بیشتر برنامه‌های کاربردی دنیای واقعی داده‌هایی دارند که به صورت خطی قابل تفکیک نیستند.

۲. همچنین، در موارد نادری که داده‌ها به صورت خطی قابل تفکیک هستند، ممکن است نخواهیم مرز تصمیم‌گیری را انتخاب کنیم که کاملا داده‌ها را از هم جدا کند. به عبارت دیگر، می‌خواهیم از بیش‌برازش جلوگیری شود. به عنوان مثال، شکل ۶ ـ ۹ را در نظر بگیرید. در اینجا مرز تصمیم قرمز کاملا تمام نقاط آموزشی را از هم جدا می‌کند. با این حال، آیا واقعا داشتن یک مرز تصمیم با چنین حاشیه کم ایده خوبی است؟ آیا فکر می‌کنید چنین مرز تصمیمی به خوبی روی داده‌های دیده نشده تعمیم می‌یابد؟ پاسخ خیر است. مرز تصمیم سبز حاشیه وسیع‌تری دارد که به آن اجازه می‌دهد روی داده‌های دیده نشده تعمیم بهتری داشته باشد. از این‌رو، ماشین بردار پشتیبان با حاشیه نرم به جلوگیری از مشکل بیش‌برازش کمک می‌کند.

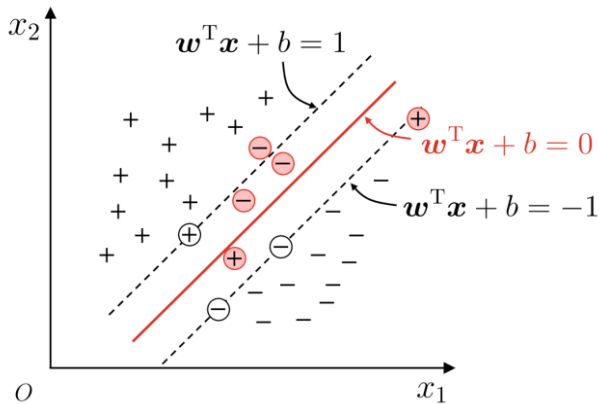

**شکل ۶ ـ ۸.** حاشیه نرم.

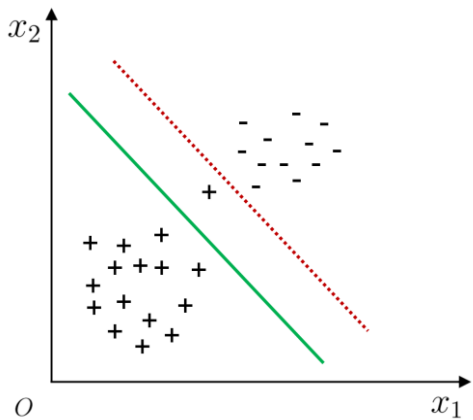

**شکل ۶ ـ ۹.** بهترین مرز تصمیم؟ سبز یا قرمز (خط‌چین)؟

به‌طور دقیق‌تر، SVM معرفی‌شده پیشین (معادله ۶ـ۱) مشمول محدودیت‌هایی است، یعنی **حاشیه سخت** باید همه‌ی نمونه‌ها را بدرستی و بدون خطا دسته‌بندی کند. با این حال، **حاشیه نرم** اجازه نقض این محدودیت را می‌دهد (با مرتکب‌شدن اشتباه برروی چند نمونه). البته، حاشیه نرم باید تعداد نمونه‌هایی که محدودیت را نقض می‌کنند، کمینه (به حداقل رساند) کند و در عین حال حاشیه را بیشینه کند. از این‌رو، هدف بهینه‌سازی را می‌توان به صورت زیر نوشت:

$$min_{w,b} \frac{1}{2} \|w\|^2 + C \sum_{i=1}^{n} \ell_{\frac{0}{1}}(y_i(w^T x_i + b) - 1) \tag{۶ـ۵}$$

که در آن $C > 0$ یک ثابت و $\ell_{\frac{0}{1}}$ تابع زیان ۰/۱ است. در اینجا، $C$ یک ابرپارامتر است که مبادله بین حداکثر کردن حاشیه و به حداقل رساندن اشتباهات را تعیین می‌کند. وقتی $C$ خیلی کوچک است، به اشتباهات دسته‌بندی اهمیت کم‌تری داده می‌شود و تمرکز بیشتر روی به حداکثر رساندن حاشیه است. در حالی که وقتی $C$ بی‌نهایت بزرگ است، تمرکز بیشتر روی اجتناب از دسته‌بندی اشتباه به قیمت کوچک نگه داشتن حاشیه است. به عبارت دیگر، وقتی $C$ بی‌نهایت بزرگ است، همه نمونه‌ها را مجبور می‌کند تا از محدودیت پیروی کنند که معادل با ماشین بردار پشتیبان با حاشیه سخت می‌شود (معادله ۶ـ۳).

حل معادله (۶ـ۵) به‌طور مستقیم دشوار است. زیرا ۰/۱ خواص ریاضی ضعیفی دارد، یعنی غیرمحدب و ناپیوسته است. بنابراین، ما با برخی از توابع زیان دیگری که دارای ویژگی‌های ریاضی خوبی هستند (به عنوان مثال، محدب و پیوسته)، جایگزین می‌کنیم. شکل ۶ـ۱۰ سه تابع زیان متداول را نشان می‌دهد:

- **زیان هینگ:** $\ell_{hinge}(z) = max(0, 1 - z)$
- **زیان نمایی:** $\ell_{exp}(z) = exp(-z)$
- **زیان لجستیک:** $\ell_{log}(z) = log(1 + exp(-z))$

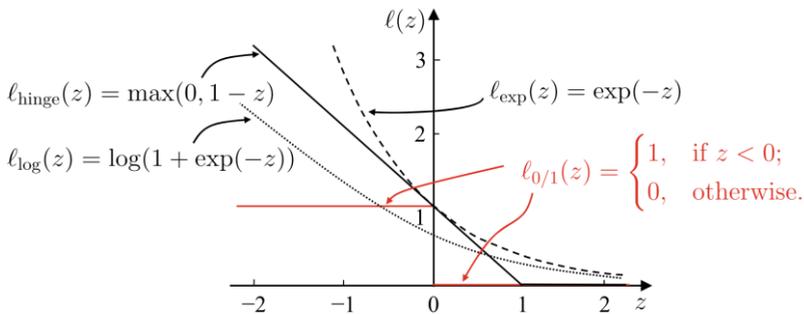

**شکل ۶ـ۱۰.** توابع زیان جایگزین برای $\ell_{0/1}$

هنگامی که از تابع زیان هینگ استفاده می‌شود، معادله ۵ـ۶ به‌صورت زیر تبدیل می‌شود:

$$min_{w,b} \frac{1}{\mathsf{Y}} \|w\|^{\mathsf{Y}} + C \sum_{i=1}^{n} max(\cdot, 1 - y_i(w^T x_i + b)) \tag{۶ـ۶}$$

با معرفی متغیر کمکی $\xi_i \geq \cdot$ معادله ۶ـ۶ به‌صورت زیر بازنویسی می‌شود:

$$min_{w,b} \frac{1}{\mathsf{Y}} \|w\|^{\mathsf{Y}} + C \sum_{i=1}^{m} \xi_i$$

به نحوی که $\xi_i \geq 1 - y_i(w^T \varphi(x_i) + b)$ و $\xi_i \geq \cdot$، $i = 1, \mathsf{Y}, \dots, m$. معمولا از این معادله برای ماشین بردار پشتیبان حاشیه نرم استفاده می‌شود. در این معادله، هر نمونه دارای یک متغیر کمکی متناظر است که میزان نقض محدودیت را نشان می‌دهد.

> عبارت $C$ در تابع هدف، تعادلی را بین حداکثر کردن حاشیه و اطمینان از اینکه حاشیه‌های عملکردی تا حد امکان بالا هستند را کنترل می‌کند. این به ما یک اثر منظم‌سازی می‌دهد که به SVM می‌گوید حتی اگر مقادیر دورافتاده در مجموعه داده وجود داشته باشد، منجر به بیش‌برازش نشود (به‌طور کامل با داده‌ها متناسب نشود).

## حاشیه سخت یا مقابل حاشیه نرم؟

استفاده از حاشیه سخت یا حاشیه نرم در ماشین بردار پشتیبان در تفکیک‌پذیری داده‌ها نهفته است. اگر داده‌های ما به صورت خطی قابل تفکیک باشد، ما به سمت حاشیه سخت می‌رویم. با این حال، در حضور نقاط داده‌ای که یافتن دسته‌بند خطی را غیرممکن می‌کند، باید ملایم‌تر باشیم و اجازه دهیم برخی از نقاط داده به اشتباه دسته‌بندی شوند. به عبارت دیگر، ازحاشیه نرم استفاده می‌کنیم.

گاهی اوقات، داده‌ها به صورت خطی قابل تفکیک هستند، اما حاشیه آنقدر کوچک است که مدل مستعد بیش‌برازش یا حساسیت بیش‌ازحد به موارد دورافتاده دارد. از این‌رو، در این مورد، برای کمک به تعمیم بهتر مدل، می‌توانیم حاشیه بزرگتری را با استفاده از SVM حاشیه نرم انتخاب کنیم.

*باید به این نکته توجه داشت که مسائل دنیای واقعی در اغلب موارد به صورت خطی قابل تفکیک نیستند، از این‌رو نمی‌توانید از حاشیه سخت در این مسائل استفاده کنید.* با این حال، اگر نگاشت مربوط به یک هسته را پیدا کردید که داده‌های تبدیل‌شده را به صورت خطی قابل تفکیک می‌کند، می‌توانید از حاشیه سخت استفاده کنید.

## ابرپارمترهای ماشین بردار پشتیبان

قبل از آموزش مدل‌ها باید ابرپارمترها را تنظیم کنیم. ابرپارمترها در ساخت مدل‌های قوی و دقیق بسیار حیاتی هستند. آن‌ها به ما کمک می‌کنند تا موازنه بین بایاس و واریانس را پیدا کنیم و از این‌رو از بیش‌برازش یا عدم‌کم‌برازش مدل جلوگیری کنیم. هنگام ساخت یک مدل مبتنی‌بر SVM نیز ما نیاز به تنظیم مقادیر زیادی از ابرپارمترها داریم که مهم‌ترین آن‌ها در زیر فهرست شده‌اند:

- **ابرپارمتر هسته** (شکل ٦ ــ ٧).

- **ابرپارمتر درجه**، کنترل می‌کند مرز تصمیم‌گیری برای هسته چندجمله‌ای چقدر انعطاف‌پذیری داشته باشد (شکل ٦ ــ ٧). هرچه درجه چند جمله‌ای بالاتر باشد، مرز تصمیم‌گیری انعطاف‌پذیرتر و پیچیده‌تر می‌شود. با این حال، پتانسیل این را دارد که مدل منجر به بیش‌برازش گردد.

- **ابرپارمتر هزینه یا C**، که میزان "سخت" یا "نرم" بودن حاشیه را کنترل می‌کند (شکل ٦ ــ ١١).

- **پارامتر گاما** میزان تاثیر یک نقطه آموزشی را بر موقعیت مرز تصمیم کنترل می‌کند. این ابرپارمتر توسط تابع هسته پایه شعاعی استفاده می‌شود. مقادیرکم‌گاما، نشان‌دهنده شعاع شباهت زیاد است که منجر به گروه‌بندی نقاط بیشتر می‌شود. برای مقادیر بالای گاما، نقاط باید بسیار نزدیک به یکدیگر باشند تا در یک گروه (یاکلاس) در نظر گرفته شوند. بنابراین، مدل‌هایی با مقادیر گامای بسیار زیاد، تمایل به بیش‌برازش دارند. هرچه گاما کوچکتر باشد، توجه کمتری به هر مورد خواهد داشت و مرز تصمیم‌گیری کم‌تر خواهد بود (به‌طور بالقوه منجر به کم‌برازش می‌شود). اثر گاما برای هسته پایه شعاعی گاوسی در قسمت پایین شکل ٦ ــ ١١ نشان داده شده است.

### پارامتر گاما در مقابل پارامتر C

برای یک هسته خطی، فقط باید پارامتر C را بهینه کنیم. با این حال، اگر بخواهیم از یک هسته پایه شعاعی استفاده کنیم، هر دو پارامتر C وگاما باید به‌طور همزمان بهینه شوند. اگر گاما بزرگ باشد، اثر C ناچیز می‌شود. اگر گاماکوچک باشد، C بر روی مدل تاثیر می‌گذارد؛ دقیقا همانطور که بر مدل خطی تاثیر می‌گذارد. مقادیر معمول برای C وگاما به شرح زیر است. با این حال، مقادیر بهینه خاصی ممکن است به کاربرد ممکن است وجود داشته باشد:

$$۱۰ > گاما > ۰٫۰۰۰۱$$

$$۰٫۰۱ < C < ۱۰۰$$

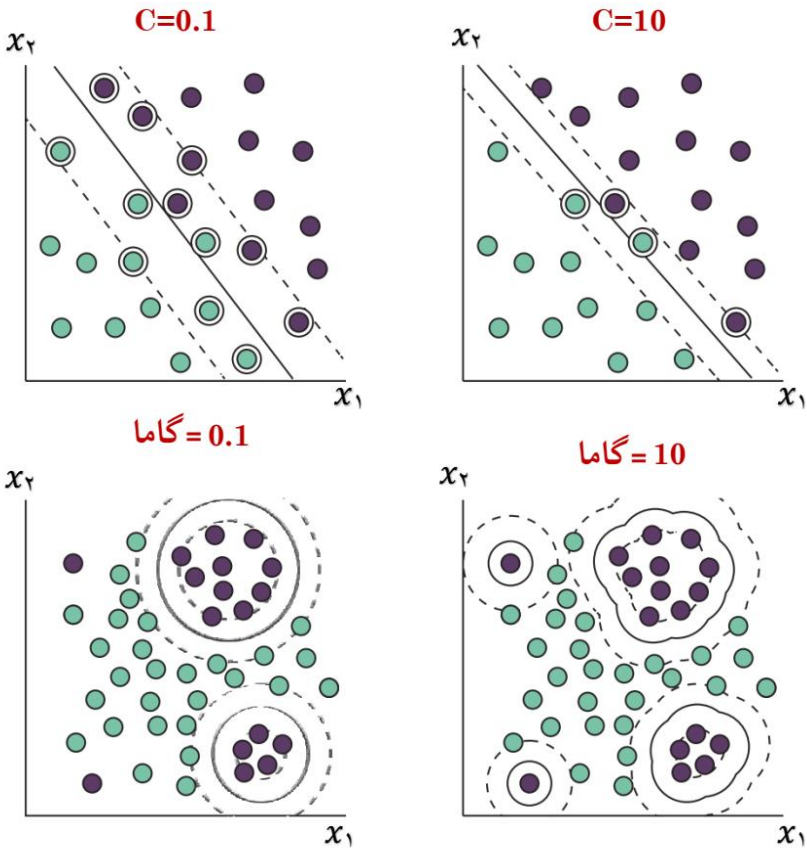

**شکل ۱۱ـ۶.** تاثیر ابرپارمترهای هزینه و گاما.

### مزایا

- حتی با داده‌های با ابعاد بالا بسیار موثر است.
- می‌تواند به‌طور موثر داده‌های غیرخطی را با استفاده از تدبیر هسته مدیریت کند.
- می‌تواند برای حل مسائل دسته‌بندی و رگرسیون استفاده شود.

### معایب

- باید یک هسته بهینه برای SVM انتخاب کنیم که این کار دشوار است.
- در مجموعه داده‌های بزرگ، نسبتا به زمان بیشتری برای آموزش نیاز دارد.
- ماشین بردار پشتیبان یک مدل احتمالی نیست، از این‌رو نمی‌توانیم دسته‌بندی را از منظر احتمال توضیح دهیم.
- درک و تفسیر مدل SVM (در مقایسه با درخت تصمیم) دشوار است.

## ماشین بردار پشتیبان در پایتون

### مجموعه داده

برای مثال SVM از مجموعه داده iris همانند مثال KNN خود استفاده می‌کنیم.

### وارد کردن کتاب‌خانه‌ها

```
In [1]:  import numpy as np
         import matplotlib.pyplot as plt
         import pandas as pd
```

### وارد کردن مجموعه داده

```
In [2]:  url = "https://archive.ics.uci.edu/ml/machine-learning-
         databases/iris/iris.data"

         # Assign colum names to the dataset
         names = ['sepal-length', 'sepal-width', 'petal-length',
         'petal-width', 'Class']

         # Read dataset to pandas dataframe
         dataset = pd.read_csv(url, names=names)
```

برای مشاهده سطرها و ستون‌های مجموعه داده، دستور زیر را اجرا کنید:

```
In  [3]:    dataset.shape

Out [3]:    (150, 5)
```

در خروجی مقدار (۱۵۰،۵) را مشاهده کردید که نشان می‌دهد مجموعه داده دارای ۱۵۰ نمونه با ۵ ستون است.

### پیش‌پردازش

گام بعدی این است که مجموعه داده خود را به ویژگی‌ها و برچسب‌های آن تقسیم کنیم. برای این کار از کد زیر استفاده کنید:

```
In [4]:  X = dataset.iloc[:, :-1].values
         y = dataset.iloc[:, 4].values
```

متغیر X شامل چهار ستون اول مجموعه داده (یعنی ویژگی‌ها) است در حالی که y حاوی برچسب‌ها است.

**تقسیم مجموعه داده**

در گام بعدی مجموعه داده خود را به دو بخش آموزشی و آزمایشی تقسیم می‌کنیم که به ما ایده بهتری درباره نحوه عملکرد الگوریتم در مرحله آزمایش می‌دهد. به این ترتیب الگوریتم ما بر روی داده‌های دیده نشده آزمایش می‌شود.

برای تقسیم‌سازی داده‌ها به دو قسمت آموزشی و آزمایشی، کد زیر را اجرا کنید:

```
In [5]:   from sklearn.model_selection import train_test_split

          X_train, X_test, y_train, y_test = train_test_split(X, y,
          test_size =0.25, random_state=42)
```

کد فوق مجموعه داده را به ۷۵ درصد داده‌های آموزشی و ۲۵ درصد داده‌های آزمایشی تقسیم می‌کند. این بدان معناست که از مجموع ۱۵۰ رکورد، مجموعه آموزشی شامل ۱۱۲ رکورد و مجموعه آزمون شامل ۳۸ رکورد خواهد بود:

```
In  [6]:   X_train.shape

Out [6]:   (112, 4)

In  [7]:   X_test.shape

Out [7]:   (38, 4)
```

در کد پیشین عدد ٤ نمایانگر تعداد ویژگی‌هاست.

**مقیاس‌بندی ویژگی‌ها**

قبل از انجام هر گونه پیش‌بینی واقعی، همیشه بهتر است که ویژگی‌ها را مقیاس بندی کنید. کد زیر مقیاس‌بندی ویژگی‌ها را انجام می‌دهد:

```
In [1]:   from sklearn.preprocessing import StandardScaler
          scaler = StandardScaler()
          scaler.fit(X_train)

          X_train = scaler.transform(X_train)
          X_test = scaler.transform(X_test)
```

**آموزش و پیش‌بینی**

ابتدا داده‌ها را به مجموعه های آموزشی و آزمایشی تقسیم کرده‌ایم و سپس مقیاس‌بندی ویژگی‌ها را برروی داده‌ها انجام دادیم. اکنون زمان آموزش SVM بر روی داده‌های آموزشی است. Scikit-Learn شامل کتابخانه svm است که شامل کلاس‌های داخلی برای الگوریتم‌های مختلف SVM است. از آنجایی که قرار است یک کار دسته‌بندی را انجام دهیم، از کلاس

دسته‌بند بردار پشتیبانی استفاده می‌کنیم که به صورت SVC در کتابخانه Scikit-Learn نوشته شده است. این کلاس یک پارامتر دارد که نوع هسته است. در ادامه ما از سه نوع هسته خطی، گاوسی و سیگموید استفاده خواهیم کرد. در متد fit کلاس SVC برای آموزش الگوریتم بر روی داده‌های آموزشی فراخوانی می‌شود. برای آموزش الگوریتم با هسته خطی کد زیر را اجرا کنید:

```
In [1]:   from sklearn.svm import SVC
          classifier = SVC(kernel='linear')
          classifier.fit(X_train, y_train)
```

مرحله آخر این است که مدل ساخته‌شده را برروی داده‌های آزمایشی خود پیش‌بینی کنیم. برای انجام این کار، کد زیر را اجرا کنید:

```
In [1]:   y_pred = classifier.predict(X_test)
```

### ارزیابی الگوریتم

```
In [3]:   from sklearn.metrics import classification_report,
          confusion_matrix
          print(confusion_matrix(y_test, y_pred))
          print(classification_report(y_test, y_pred))

Out [7]:  [[15  0  0]
           [ 0 10  1]
           [ 0  0 12]]
                        precision   recall f1-score   support

            Iris-setosa      1.00     1.00     1.00        15
         Iris-versicolor     1.00     0.91     0.95        11
          Iris-virginica     0.92     1.00     0.96        12

               accuracy                        0.97        38
              macro avg      0.97     0.97     0.97        38
           weighted avg      0.98     0.97     0.97        38
```

### آموزش با هسته گاوسی و پیش‌بینی

```
In [1]:   from sklearn.svm import SVC
          classifier = SVC(kernel='rbf')
          classifier.fit(X_train, y_train)
In [2]:   y_pred = classifier.predict(X_test)
```

### ارزیابی الگوریتم

```
In [3]:   from sklearn.metrics import classification_report,
          confusion_matrix
```

```
print(confusion_matrix(y_test, y_pred))
print(classification_report(y_test, y_pred))
```

Out [7]:
```
[[15  0  0]
 [ 0 11  0]
 [ 0  0 12]]
              precision   recall f1-score  support

  Iris-setosa      1.00     1.00     1.00       15
Iris-versicolor    1.00     1.00     1.00       11
 Iris-virginica    1.00     1.00     1.00       12

     accuracy                        1.00       38
    macro avg      1.00     1.00     1.00       38
 weighted avg      1.00     1.00     1.00       38
```

### آموزش با هسته سیگموید و پیش‌بینی

In [1]:
```
from sklearn.svm import SVC
classifier = SVC(kernel='sigmoid')
classifier.fit(X_train, y_train)
```

In [1]:
```
y_pred = classifier.predict(X_test)
```

### ارزیابی الگوریتم

In [3]:
```
from sklearn.metrics import classification_report,
confusion_matrix
print(confusion_matrix(y_test, y_pred))
print(classification_report(y_test, y_pred))
```

Out [7]:
```
[[15  0  0]
 [ 0  7  4]
 [ 0  1 11]]
              precision   recall f1-score  support

  Iris-setosa      1.00     1.00     1.00       15
Iris-versicolor    0.88     0.64     0.74       11
 Iris-virginica    0.73     0.92     0.81       12

     accuracy                        0.87       38
    macro avg      0.87     0.85     0.85       38
 weighted avg      0.88     0.87     0.87       38
```

### مقایسه عملکرد هسته‌ها

اگر عملکرد انواع مختلف هسته‌ها را با هم مقایسه کنیم، مشاهده می‌شود که هسته سیگموید در مقایسه با دو هسته دیگر عملکرد بدتری داشته است. دلیل این امر این است که تابع سیگموید دو مقدار ۰ و ۱ بر می‌گرداند، بنابراین برای مسائل دسته‌بندی دودویی مناسب‌تر است. در بین هسته گاوسی و هسته خطی، می‌توانیم ببینیم که هسته گاوسی به یک نرخ پیش‌بینی کامل ۱۰۰٪ دست یافته است در حالی که هسته خطی یک نمونه را به اشتباه دسته‌بندی کرده است. بنابراین هسته گاوسی عملکرد بهتری داشته است. با این حال، هیچ قانون کلی و سریعی وجود ندارد که کدام هسته در هر سناریویی بهترین عملکرد را دارد. از این‌رو، تنها با آزمایش هسته‌های مختلف و مشاهده نتایج می‌توان در هر مساله‌ای نوع هسته را مشخص کرد.

## درخت تصمیم

یکی از محبوب‌ترین الگوریتم‌های یادگیری ماشین، درختان تصمیم به دلیل نحوه عملکرد بسیار ساده آنها است. برخلاف ماشین بردار پشتیبان که برای فهمیدن به یک پایه ریاضی بسیار قوی نیاز دارد، درختان تصمیم به معنای واقعی از روشی که ما انسان‌ها به صورت روزانه عمل می‌کنیم تقلید می‌کنند. به عنوان مثال: فرض کنید یک کتری داریم و می‌خواهیم آن را برداریم، اما در عین حال، نمی‌خواهیم دستمان را بسوزانیم یا به دلیل سنگینی آن را رها کنیم.

در تجزیه و تحلیل تصمیم می‌توان از درختان تصمیم برای نمایش عینی و صریح تصمیمات و تصمیم‌گیری استفاده کرد و همان‌طورکه از نامش پیداست، از یک مدل درخت‌مانند در جهت رسیدن به تصمیم نهایی استفاده می‌کند. اگرچه یک ابزار رایج در داده‌کاوی برای استخراج یک استراتژی برای رسیدن به یک هدف خاص است، اما به طور گسترده در یادگیری ماشین استفاده می‌شود.

در یادگیری ماشین، درختان تصمیم نوعی مدل ناپارامتری هستند که می‌توانند هم برای دسته‌بندی و هم برای رگرسیون استفاده شوند. این بدان معناست که درختان تصمیم مدل‌های انعطاف‌پذیری هستند که با افزودن ویژگی‌های بیشتر (اگر آن‌ها را به درستی بسازیم) تعداد پارامترهای خود را افزایش نمی‌دهند و می‌توانند یک پیش‌بینی دسته‌بندی (مثل اینکه یک گیاه از نوع خاصی است یا خیر) و حتی یک پیش‌بینی عددی (مانند قیمت یک خانه) را تولید کنند.

درختان تصمیم اولین بار توسط لئو بریمن، آماردان دانشگاه کالیفرنیا، برکلی پیشنهاد شدند. ایده او این بود که داده‌ها را به صورت درختی نشان دهد که در آن هر گره داخلی نشان‌دهنده آزمایشی بر روی یک ویژگی (در اصل یک شرط)، هر شاخه نشان‌دهنده نتیجه آزمایش و هر گره برگ (گره پایانی)، دارای یک برچسب کلاس باشد. به‌طور خلاصه، هر سوالی که در فرآیند تصمیم‌گیری پرسیده می‌شود، آزمایشی بر روی یک ویژگی است و هر آزمون یا به یک نتیجه‌گیری و یا به یک آزمون اضافی مشروط به پاسخ فعلی منجر می‌شود. اساسا، درخت‌های تصمیم یک

سری قوانین صریح در مورد مقادیر ویژگی را می‌آموزند که منجر به تصمیمی می‌شود که مقدار هدف را پیش‌بینی می‌کند. شکل زیر نمونه‌ای از درخت تصمیم ساده‌ای است که برای دسته‌بندی یک حیوان به عنوان پرنده، سگ یا ماهی بر اساس ویژگی‌های آن‌ها (شنا کردن یا چهارپا بودن) استفاده می‌شود:

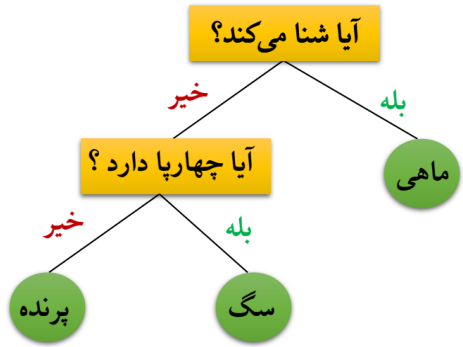

الگوریتم درخت تصمیم برای تجزیه و تحلیل سه کلاس و فرمول‌بندی سوالات لازم برای تمایز بین کلاس‌های مختلف در مثال بالا این سوالات را می‌پرسد:

- آیا شنا می‌کند؟
- آیا چهار پا دارد؟

وقتی به این سوالات بله یا خیر پاسخ داده شد، به‌طور حتم می‌دانیم که داده‌های ما متعلق به کدام کلاس یعنی کدام حیوان است. از این‌رو، درختان تصمیم مفهومی نسبتا ساده برای درک دارند. چراکه بسیار شهودی هستند.

به‌طور معمول، یک درخت تصمیم از یک گره ریشه، چندین گره داخلی و چندین گره برگ تشکیل شده است. گره‌های برگ *با نتایج تصمیم مطابقت دارند* و هر گره دیگر مربوط به یک آزمون ویژگی است. نمونه‌ها در هر گره با توجه به نتایج **انشعاب‌سازی**[1] ویژگی‌ها به گره‌های فرزند تقسیم می‌شوند. هر مسیر از گره ریشه به گره برگ یک دنباله تصمیم‌گیری است. درختان تصمیم سعی می‌کنند مجموعه داده‌ها را به گونه‌ای انشعاب دهند که داده‌های موجود در هر گروه تا حد امکان شبیه به یکدیگر باشد در حالی‌که داده‌های موجود در یک گروه تا حد ممکن با داده‌های موجود در گروه‌های دیگر متفاوت است. هدف تولید درختی است که بتواند نمونه‌های دیده نشده را تعمیم دهد.

ساخت درختان تصمیم از استراتژی **تقسیم و حل**[2] پیروی می‌کند، از این‌رو درخت تصمیم به صورت بازگشتی (بالا به پایین) تولید می‌شود. به عبارت ساده‌تر، الگوریتم درخت تصمیم با

---

[1] splitting

[2] Divide-and-conquer

مجموعه داده‌های آموزشی در گره ریشه شروع می‌شود و به صورت بازگشتی داده‌ها را به گره‌های سطح پایین‌تر بر اساس **معیار انشعاب**[1] تقسیم می‌کند. تنها گره‌هایی که حاوی ترکیبی از کلاس‌های مختلف هستند باید بیشتر انشعاب پیدا کنند. در نهایت الگوریتم درخت تصمیم رشد درخت را بر اساس یک معیار توقف، متوقف می‌کند. ساده‌ترین معیار توقف، معیاری است که در آن همه نمونه‌های آموزشی در برگ متعلق به یک کلاس هستند. یک مشکل این است که ساخت درخت تصمیم تا این سطح ممکن است منجر به بیش‌برازش شود. چنین درختی به خوبی به نمونه‌های آزمایشی دیده‌نشده تعمیم نمی‌یابد. برای جلوگیری از کاهش دقتِ ناشی از بیش‌برازش، دسته‌بند از مکانیزم **هرس**[2] استفاده می‌شود.

پس از اینکه درخت تصمیم ساخته شد، برای دسته‌بندی نمونه‌های آزمون با استفاده از پیمایش بالا به پایین از ریشه تا یک برگ منحصربه‌فرد استفاده می‌شود. شرط انشعاب در هر گره داخلی برای انتخاب شاخه درست درخت تصمیم برای پیمایش بیشتر استفاده و برچسب گره برگی که به آن رسیده است برای نمونه آزمون گزارش می‌شود.

از ویژگی‌های کلیدی درخت تصمیم این است که نیازی به مقیاس‌بندی و هنجارسازی داده‌ها ندارد.

## معیارهای انتخاب ویژگی (انشعاب بهینه) و فرآیند آموزش درخت تصمیم

آموزش درخت تصمیم یک فرآیند از بالا به پایین است که در آن مجموعه داده‌های آموزشی به صورت بازگشتی به زیر مجموعه‌های کوچک‌تر تقسیم می‌شود. این زیرمجموعه‌ها با انتخاب یک محدودیت ویژگی در هر مرحله تعیین می‌شوند که به بهترین وجه قادر به انشعاب مجموعه نمونه‌ها بر اساس معیار خاصی است که به آن معیار انشعاب می‌گویند. هسته الگوریتم یادگیری درخت تصمیم، انتخابِ ویژگیِ انشعاب بهینه است. به طور کلی، همان‌طور که فرآیند انشعاب پیش می‌رود، ما آرزو می‌کنیم که نمونه‌هایی که نمونه‌های بیشتری در هر گره متعلق به یک کلاس واحد باشد. هدف معیار انشعاب، بیشینه کردن جداسازی طبقات مختلف در میان گره‌های فرزند است.

ایده اصلی یک الگوریتم درخت تصمیم، شناسایی ویژگی‌هایی است که حاوی بیشترین اطلاعات در مورد ویژگی هدف هستند و سپس مجموعه داده را در امتداد مقادیر این ویژگی‌ها انشعاب می‌دهد. ویژگی‌ای که عدم قطعیت را از اطلاعات مربوط به ویژگی هدف به بهترین نحو جدا می‌کند، آموزنده‌ترین ویژگی است. روند جستجو برای آموزنده‌ترین ویژگی ادامه می‌یابد تا زمانی که به گره‌های برگ خالص برسد. فرآیند ساخت یک مدل درخت تصمیم شامل پرسیدن

---

[1] split criterion

[2] pruning

یک سوال در هر مورد و سپس ادامه و انشعاب است. حال وقتی چندین ویژگی وجود دارد که مقدار هدف یک نمونه خاص را تعیین می‌کند، سوالات زیر ایجاد می‌شود:

- **کدام ویژگی/صفت باید در گره ریشه باید برای شروع انتخاب شود؟**
- **به کدام ترتیب باید به انتخاب ویژگی‌ها در هر انشعاب بعدی در یک گره ادامه دهیم؟**
- **کدام ویژگی به عنوان گره داخلی یا گره برگ عمل خواهد کرد؟**

برای تصمیم‌گیری در این موارد و نحوه انشعاب درخت، از معیارهای انشعاب استفاده می‌کنیم. فراگیرترین معیارهای انشعاب مورد استفاده برای آموزش درختان تصمیم در ادامه تشریح می‌شوند.

## بهره اطلاعات

این معیار مبتنی بر مفهوم آنتروپی اطلاعات است که بی‌نظمی یا عدم قطعیت را در یک سیستم اندازه‌گیری می‌کند. بهره اطلاعات، اندازه‌گیریِ تغییراتِ آنتروپی پس از تقسیم‌بندی یک مجموعه داده بر اساس یک ویژگی است و محاسبه می‌کند که یک ویژگی چه مقدار اطلاعات در مورد یک کلاس ارائه می‌دهد. با توجه به مقدار بهره اطلاعات، گره را انشعاب داده و درخت تصمیم را می‌سازد. یک الگوریتم درخت تصمیم همیشه سعی می‌کند تا مقدار بهره اطلاعات را بیشینه کرده و گره یا همان ویژگی که بالاترین بهره اطلاعات را دارد ابتدا تقسیم شود.

فرض کنید $p_k$ نسبت کلاس $k$ در مجموعه داده D را نشان می‌دهد و $k = ۱, ۲, ..., |y|$ می‌باشد. بر این اساس آنتروپی به‌صورت زیر تعریف می‌شود:

$$(۷ - ٦)$$
$$Ent(D) - \sum_{k=۱}^{|y|} p_k log_۲ p_k$$

هرچه $Ent(D)$ کم‌تر باشد، خلوص $D$ بالاتر است.

فرض کنید ویژگی گسسته $a$ دارای $V$ مقادیر ممکن $\{a^۱, a^۲, ..., a^V\}$ است. بر این اساس، تقسیم مجموعه داده $D$ بر اساس ویژگی $a$، گره‌های فرزند $V$ را تولید می‌کند، جایی که $V$امین گره فرزند $D^V$ شامل تمام نمونه‌های موجود در $D$ می‌شود که مقدار $a^V$ را برای ویژگی $a$ می‌گیرد. سپس، آنتروپی $D^V$ را می‌توان با استفاده از معادله (٦ - ۷) محاسبه کرد. از آنجایی که تعداد نمونه‌های متفاوتی در گره‌های فرزند وجود دارد، وزن $\frac{|D^v|}{|D|}$ برای نشان دادن اهمیت هر گره اختصاص داده شده است، یعنی هر چه تعداد نمونه‌ها بیشتر باشد، تاثیر گره شاخه بیشتر می‌شود. از این‌رو، بهره اطلاعات حاصل از انشعاب مجموعه داده $D$ با ویژگی $a$ با استفاده از معادله‌ی زیر محاسبه می‌شود:

$$Gain(D, a) = Ent(D) - \sum_{v=1}^{V} \frac{|D^v|}{|D|} Ent(D^v)$$

به طور کلی، هر چه میزان بهره اطلاعات با انشعاب $D$ و ویژگی $a$ بیشتر باشد، خلوص بیشتری را نیز می‌توانیم انتظار داشته باشیم.

> جمعیتی خالص است که همه اعضای آن متعلق به یک دسته واحد باشند.

## شاخص جینی

شاخص جینی معیاری از ناخالصی یا خلوص است که هنگام ایجاد یک مرز تصمیم استفاده می‌شود. یک ویژگی با شاخص جینی پایین باید در مقایسه با شاخص جینی بالا ترجیح داده شود. با استفاده از نمادی مشابه (۷−۶)، مقدار جینی مجموعه داده $D$ به صورت تعریف می‌شود:

$$Gini(D) = \sum_{k=1}^{|y|} \sum_{k' \neq k} p_k \, p_{k'}$$

$$= ۱ - \sum_{k=۱}^{|y|} p_k^{۲}$$

به طور شهودی، $Gini(D)$ احتمال دو نمونه را که به طور تصادفی از مجموعه داده $D$ متعلق به کلاس‌های مختلف انتخاب کردیم را نشان می‌دهد. هرچه جینی $Gini(D)$ کم‌تر باشد، خلوص مجموعه داده $D$ بالاتر است.

با استفاده از نمادی مشابه با بهره اطلاعات، شاخص جینی ویژگی $a$ به‌صورت زیر تعریف می‌شود:

$$Gini\_index(D, a) = \sum_{v=1}^{V} \frac{|D^v|}{|D|} gini(D^v)$$

بر این اساس، با توجه به مجموعه ویژگی‌های نامزد $A$، ویژگی با کم‌ترین شاخص جینی را به عنوان ویژگی انشعاب انتخاب می‌کنیم.

## معیار توقف و هرس

معیار توقف برای رشد درخت تصمیم ارتباط نزدیکی با استراتژی هرس دارد. هنگامی که درخت تصمیم تا انتها رشد می‌کند، یعنی تا زمانی که هر گره برگ فقط نمونه‌هایی متعلق به یک کلاس خاص را شامل شود، درخت تصمیم بدست آمده دقت ۱۰۰ درصدی در نمونه‌های متعلق به داده‌های آموزشی از خود نشان می‌دهد. با این حال، در این حالت، اغلب به نمونه‌های آزمایشی دیده‌نشده تعمیم ضعیفی پیدا می‌کند. چراکه درخت تصمیم در حال حاضر حتی به ویژگی‌های

تصادفی در نمونه‌های آموزشی هم همخوانی دارد. بیشتر این نویز توسط گره‌های سطح پایین‌تر ایجاد می‌شود که حاوی تعداد کم‌تری نقاط داده هستند. از این‌رو، مدل‌های ساده‌تر (درخت‌های تصمیم کم‌عمق) به مدل‌های پیچیده‌تر (درخت تصمیم عمیق) ترجیح داده می‌شوند، *اگر همان خطا را در داده‌های آموزشی ایجاد کنند.*

برای کاهش سطح بیش‌برازش، یک استراتژی این است که رشد درخت را زودتر متوقف کنید. متأسفانه، هیچ راهی برای دانستن نقطه درستی که در آن رشد درخت متوقف شود وجود ندارد. بنابراین، یک استراتژی طبیعی این است که بخش‌های بیش‌برازش درخت تصمیم را هرس کنیم و گره‌های داخلی را به گره‌های برگ تبدیل کنیم. استراتژی‌های کلی هرس شامل **پیشا‌ـ‌هرس**[1] و **پسا‌ـ‌هرس**[2] است.

پیشا‌ـ‌هرس، بهبود توانایی تعمیم دادن هر انشعاب را ارزیابی می‌کند و اگر بهبود کوچک باشد، انشعاب را لغو می‌کند، یعنی گره به عنوان یک گره برگ مشخص می‌شود. در مقابل آن، پسا‌ـ‌هرس، گره‌های غیر برگِ یک درخت تصمیمِ کاملا رشد یافته را دوباره بررسی می‌کند و اگر جایگزینی منجر به بهبود توانایی تعمیم شود، یک گره با یک گره برگ جایگزین می‌شود.

پیشا‌ـ‌هرس این مزیت را دارد که سریع‌تر و کارآمدتر باشد، چراکه از ایجاد زیردرخت‌های بیش از حد پیچیده که با داده‌های آموزشی منطبق (بیش‌برازش) هستند جلوگیری می‌کند. در پسا‌ـ‌هرس، درخت را به طور کامل با استفاده از الگوریتم درخت تصمیم خود رشد می‌دهید و سپس درختان فرعی را به صورت پایین به بالا هرس می‌کنید. شما از گره تصمیم پایین شروع می‌کنید و بر اساس معیارهایی مانند بهره اطلاعات، تصمیم می‌گیرید که آیا این گره تصمیم را حفظ کنید یا آن را با یک گره برگ جایگزین کنید.

**پیشا‌ـ‌هرس:**

- این تکنیک قبل از ساخت درخت تصمیم مورد استفاده قرار می‌گیرد.
- پیشا‌ـ‌هرس را می‌توان با استفاده از تنظیم ابرپارمترها انجام داد.

**پسا‌ـ‌هرس:**

- این تکنیک پس از ساخت درخت تصمیم استفاده می‌شود.
- این تکنیک زمانی استفاده می‌شود که درخت تصمیم عمق بسیار بزرگی داشته باشد و مدل بیش‌برازش را نشان دهد.
- این تکنیک همچنین به عنوان **هرس پس‌رو**[3] شناخته شده است.

---

[1] pre-pruning

[2] post-pruning

[3] backward pruning

## مزایا

- در برابر خطاها مقاوم است و اگر داده‌های آموزشی حاوی خطا باشد، الگوریتم‌های درخت تصمیم برای رسیدگی به چنین مسائلی مناسب‌تر خواهند بود.
- قوانین قابل درکی را تولید می‌کنند و بسیار بصری هستند.
- می‌توانند متغیرهای پیوسته و گسسته را مدیریت کنند.
- نحوه کار آن‌ها بسیار ساده است و می‌توان آن را به راحتی به هرکسی توضیح داد.
- هیچ فرضی در خصوص خطی بودن داده‌ها ندارد و از این‌رو می‌تواند درجایی‌که پارامترها به صورت غیرخطی مرتبط هستند، استفاده شود.
- باعث صرفه‌جویی در زمان آماده‌سازی داده‌ها می‌شود، چراکه آن‌ها به مقادیر مفقودی و مقادیر دورافتاده حساس نیستند.
- نیازی به هنجارسازی و مقیاس‌بندی داده‌ها ندارند.
- مفهوم درخت تصمیم برای برنامه‌نویسان آشناتر است و درک آن نسبت به سایر الگوریتم‌های مشابه آسان‌تر است.

## معایب

- مستعد خطا در مسائل دسته‌بندی باکلاس‌های زیاد و تعداد نسبتاکمی نمونه‌های آموزشی هستند.
- برای درخت تصمیم گاهی اوقات محاسبه می‌تواند بسیار پیچیده‌تر از الگوریتم‌های دیگر باشد.
- اغلب زمان بیشتری برای آموزش مدل نیاز دارد.
- افزودن یک نقطه داده جدید می‌تواند منجر به تولید مجدد درخت شود و همه گره‌ها باید دوباره محاسبه و ایجاد شوند.
- درخت تصمیم منفرد اغلب یادگیرنده ضعیفی است، بنابراین برای پیش‌بینی بهتر به یک دسته درخت تصمیم برای ایجاد جنگل تصادفی نیاز داریم.
- مستعد بیش برازش هستند. به منظور برازش با داده‌ها (حتی داده‌های نویزی)، به تولید گره‌های جدید ادامه می‌دهد و در نهایت درخت برای تفسیر، بیش از حد پیچیده می‌شود. به این ترتیب قابلیت تعمیم خود را از دست می‌دهد.
- به دلیل بیش‌برازش، احتمال واریانس بالایی (به منظور دستیابی به بایاس صفر، منجر به واریانس بالا می‌شود) در خروجی وجود دارد که منجر به خطاهای زیادی در تخمین نهایی می‌شود و دقت پایینی را در نتایج نشان می‌دهد.

## درخت تصمیم در پایتون

### مجموعه داده

در این مثال نیز از مجموعه داده iris استفاده می‌کنیم. در مثال‌های پیشین از طریق تارنمای آن به این مجموعه داده دسترسی پیدا کردیم. این مجموعه داده در کتابخانه Scikit-Learn نیز وجود دارد. در این مثال، از طریق کتابخانه آن را وارد می‌کنیم.

### وارد کردن کتاب‌خانه‌ها

```
In [1]:   import numpy as np
          import matplotlib.pyplot as plt
          import pandas as pd
```

### وارد کردن مجموعه داده

```
In [2]:   from sklearn import datasets
          iris = datasets.load_iris()
```

### آماده‌سازی داده‌ها

گام بعدی این است که مجموعه داده خود را به ویژگی‌ها و برچسب‌های آن تقسیم کنیم. برای این کار از کد زیر استفاده کنید:

```
In [4]:   X = iris.data
          y = iris.target
```

### تقسیم مجموعه داده

```
In [5]:   from sklearn.model_selection import train_test_split

          X_train, X_test, y_train, y_test = train_test_split(X, y,
          test_size =0.25, random_state=42)
```

### آموزش و پیش‌بینی

کتابخانه درخت تصمیم Scikit-Learn شامل متدهایی برای الگوریتم‌های درخت تصمیم مختلف است. از آنجایی که در این مثال قرار است یک کار دسته‌بندی انجام دهیم، برای این مثال از کلاس DecisionTreeClassifier استفاده می‌کنیم.

```
In [1]:   from sklearn.tree import DecisionTreeClassifier
          classifier = DecisionTreeClassifier()
          classifier.fit(X_train, y_train)
```

اکنون که دسته‌بند ما آموزش دیده است، بیایید پیش‌بینی‌هایی را در مورد داده‌های آزمون انجام دهیم. برای انجام این کار، کد زیر را اجرا کنید:

```
In [1]:   y_pred = classifier.predict(X_test)
```

## ارزیابی الگوریتم

در این مرحله ما الگوریتم خود را آموزش داده‌ایم و برخی پیش‌بینی‌ها را انجام داده‌ایم. اکنون می‌خواهیم ببینیم که الگوریتم ما چقدر دقیق است.

```
In [3]:   from sklearn.metrics import classification_report,
          confusion_matrix
          print(confusion_matrix(y_test, y_pred))
          print(classification_report(y_test, y_pred))
```

```
Out [3]:  [[15  0  0]
           [ 0 11  0]
           [ 0  0 12]]
                      precision   recall  f1-score   support

         Iris-setosa      1.00      1.00      1.00        15
      Iris-versicolor     1.00      1.00      1.00        11
       Iris-virginica     1.00      1.00      1.00        12

            accuracy                          1.00        38
           macro avg      1.00      1.00      1.00        38
        weighted avg      1.00      1.00      1.00        38
```

نتایج نشان می‌دهد که مدل درخت تصمیم ما قادر است تمام ۳۸ رکورد موجود در مجموعه آزمون را با دقت ۱۰۰٪ دسته‌بندی کند.

## چاپ بازنمایی به‌صورت متن

خروجی درخت تصمیم به‌صورت بازنمایی متنی می‌تواند هنگام کار برروی برنامه‌های کاربردی بدون رابط کاربری یا زمانی که می‌خواهیم اطلاعات مربوط به مدل را در فایل متنی ذخیره کنیم، مفید باشد. برای این کار کد زیر را اجرا کنید:

```
In [3]:   from sklearn import tree
          text_representation = tree.export_text(classifier)
          print(text_representation)
```

```
Out [3]:  |--- feature_3 <= 0.80
          |   |--- class: 0
```

```
        |--- feature_3 >  0.80
        |   |--- feature_2 <= 4.75
        |   |   |--- feature_3 <= 1.65
        |   |   |   |--- class: 1
        |   |   |--- feature_3 >  1.65
        |   |   |   |--- class: 2
        |   |--- feature_2 >  4.75
        |   |   |--- feature_3 <= 1.75
        |   |   |   |--- feature_2 <= 4.95
        |   |   |   |   |--- class: 1
        |   |   |   |--- feature_2 >  4.95
        |   |   |   |   |--- feature_3 <= 1.55
        |   |   |   |   |   |--- class: 2
        |   |   |   |   |--- feature_3 >  1.55
        |   |   |   |   |   |--- feature_2 <= 5.45
        |   |   |   |   |   |   |--- class: 1
        |   |   |   |   |   |--- feature_2 >  5.45
        |   |   |   |   |   |   |--- class: 2
        |   |   |--- feature_3 >  1.75
        |   |   |   |--- feature_2 <= 4.85
        |   |   |   |   |--- feature_1 <= 3.10
        |   |   |   |   |   |--- class: 2
        |   |   |   |   |--- feature_1 >  3.10
        |   |   |   |   |   |--- class: 1
        |   |   |   |--- feature_2 >  4.85
        |   |   |   |   |--- class: 2
```

اگر می‌خواهید آن را در فایل ذخیره کنید، کد زیر را اجرا کنید:

```
In  [3]:   with open("decistion_tree.log", "w") as fout:
               fout.write(text_representation)
```

## مصورسازی مدل

کتابخانه‌های مختلفی برای مصورسازی درخت تصمیم وجود دارند. با این حال، در این مثال ما از متد plot_tree که در کتابخانه Scikit-Learn وجود دارد، استفاده می‌کنیم. این متد به ما این امکان را می‌دهد تا به راحتی شکل درخت را تولید کنیم. برای این کار کد زیر را اجرا کنید:

```
In  [3]:   fig = plt.figure(figsize=(25,20))
           _ = tree.plot_tree(classifier,
                       feature_names=iris.feature_names,
                       class_names=iris.target_names,
                       filled=True)
```

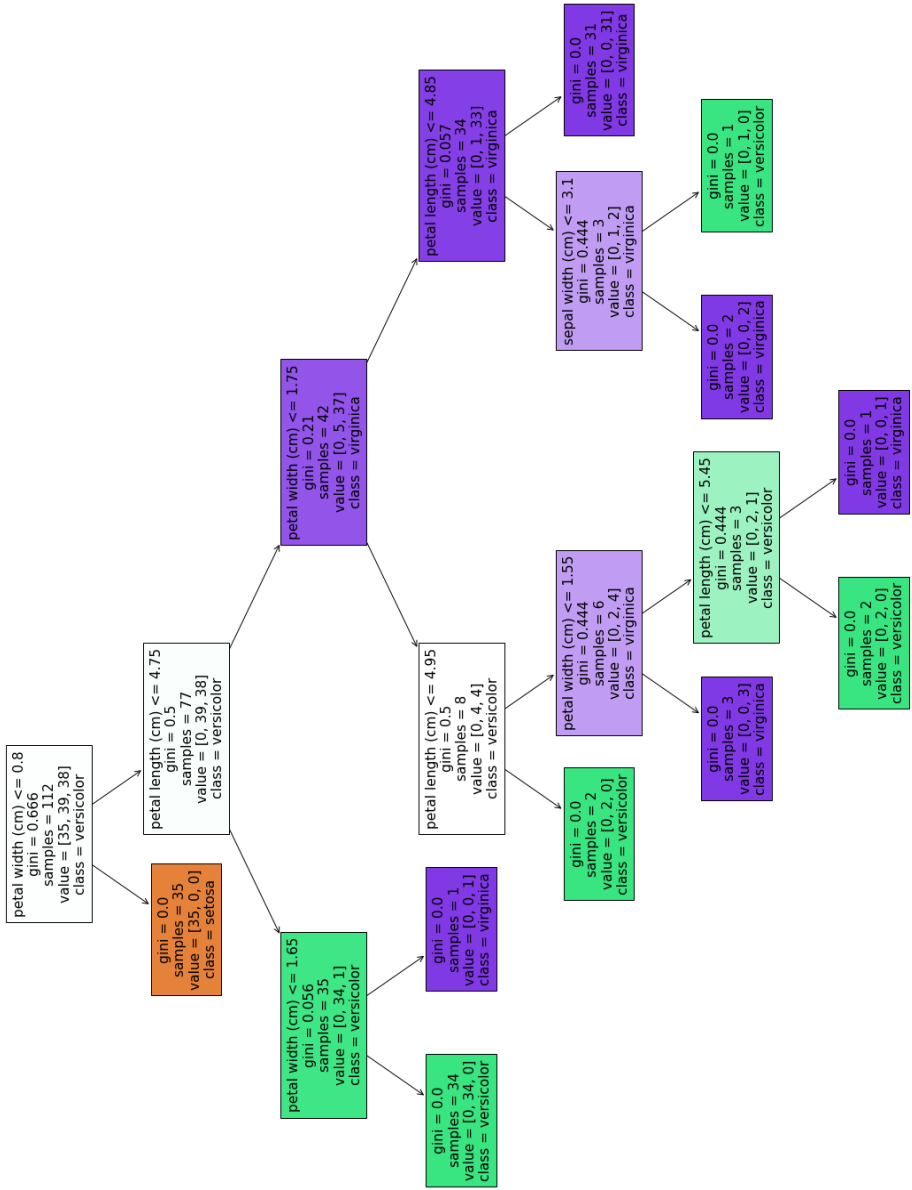

## بیز ساده

بیز ساده یک تکنیک دسته‌بندی بر اساس **قضیه بیز**[1] با این فرض است که تمام ویژگی‌هایی که مقدار هدف را پیش‌بینی می‌کنند مستقل از یکدیگر هستند. این تکنیک، احتمال هر کلاس را

---

[1] Bayes theorem

محاسبه و سپس کلاسی که بیشترین احتمال را دارد انتخاب می‌کند. اما چرا الگوریتم "ساده" نامیده می شود؟ این به این دلیل است که دسته‌بند فرض می‌کند که ویژگی‌های ورودی که وارد مدل می‌شوند مستقل از یکدیگر هستند. از این رو، تغییر یک ویژگی ورودی هیچ یک از سایرین را تحت تاثیر قرار نمی‌دهد. بنابراین ساده است، به این معنا که این فرض ممکن است درست باشد یا نباشد.

پایه اصلی بیز ساده، قضیه بیز یا قاعده بیزی $p(H|S) = \frac{p(S|H)p(S)}{p(H)}$ است که **احتمال پسین**[1] $p(H|S)$ را برای یک فرضیه یا مدل تخمین می‌زند. در اینجا $p(S|H)$ **احتمال درست‌نمایی**[2] نمونه‌های داده است؛ با توجه به اینکه $H$ درست و $p(H)$ **احتمال پیشین**[3] (اولیه) فرضیه $H$ است که به نوعی هرگونه **دانش قبلی**[4] در مورد $H$ را در برمی‌گیرد. اگر دانش قبلی وجود نداشته باشد، می‌توانیم از توزیع‌های یکنواخت به عنوان **پیشین** استفاده کنیم. علاوه بر این، $p(S)$ را می‌توان به عنوان احتمال پیشین  نمونه $S$ در نظر گرفت. بسته به تاکید و فرمول‌بندی، می‌توانیم **بیشینه‌ی پسین**[5] (MAP) فرضیه را جستجو کنیم

$$maximize\ p(H|S) \propto maximize\ P(S|H)P(H)$$

یا می‌توانیم به دنبال **درست‌نمایی بیشینه**[6] (ML) باشیم

$$maximize\ p(S|H)$$

اگر هر دو $p(H)$ و $p(S)$ ثابت باشند، معادل MAP قبلی است.

برای یک مساله دسته‌بندی با تمرکز بر یک ویژگی $F = x$ با $K$ مقادیر ویژگی مختلف $[x_1, x_2, ...., x_k]$ تابع مدل آن $y = f(x)$ مجموعه‌ای از مقادیر گسسته محدود $y_i \in \omega$ را تشکیل می‌دهد. هدف یک دسته‌بند بیزی برآورد احتمال $y$ با داده‌های $x_i$ است تا احتمال کلاس را نسبت دهد

$$\max p(y_i|x_i) = p(y_i|x_1, x_2, ...., x_k)\ , y_i \in \omega$$

که معادل است با:

$$\max \frac{p(x_1, x_2, ...., x_k|y_i)p(y_i)}{p(x_1, x_2, ...., x_k)} \propto \max p(x_1, x_2, ...., x_k|y_i)p(y_i)$$

---

[1] posterior probability

[2] likelihood probability

[3] prior probability

[4] background knowledge

[5] maximum a posterior (MAP)

[6] maximum likelihood (ML)

با این حال، محاسبه احتمال $p(x_1, x_2, \ldots, x_k|y_i)$، نابدیهی است و در بیشتر موارد محاسبه آن غیرممکن است. یک فرض ساده این است که تمام مقادیر نمونه داده از یکدیگر نابستهی مشروط [1] هستند و در نتیجه **احتمال توام**[2] (مشترک)، حاصلضرب احتمال فردی میشود. از اینرو میتوانیم از آن استفاده کنیم:

$$p(x_1, x_2, \ldots, x_k|y_i)p(y_i) = \prod_{i=1}^{k} p(x_i|y_i)$$

که معادل است با:

$$\max p(y_i) \prod_{i=1}^{k} p(x_i|y_i)$$

یک دستهبند احتمالی که از معادله بالا برای تخصیص احتمالات استفاده میکند، به یک دستهبند ساده بیزی تبدیل میشود.

برای آموزش یک دستهبند بیز ساده، احتمال پیشین $p(y_i)$ را از مجموعه آموزشی $D$ محاسبه میکنیم و سپس احتمال شرطی $p(x_i|y_i)$ را برای هر ویژگی محاسبه میکنیم. اگر $D_{y_i}$ زیر مجموعهای از $D$ را نشان دهد که شامل تمام نمونههای کلاس $y_i$ است. با فرض اینکه نمونهها به صورت **مستقل با توزیع یکسان**[3] (i.i.d) باشند، احتمال پیشین را میتوان به راحتی برآورد کرد

$$p(y_i) = \frac{|D_{y_i}|}{|D|}$$

برای ویژگیهای گسسته، اگر $D_{y_i,x_i}$ زیرمجموعهای از $D_{y_i}$ را نشان دهد که شامل تمام نمونههایی است که مقدار $x_i$ را در ویژگی $i$ام دریافت میکنند، احتمال شرطی $p(x_i|y_i)$ را میتوان با

$$p(x_i|y_i) = \frac{|D_{y_i,x_i}|}{|D_{y_i}|}$$

برآورد کرد.

---

[1] conditionally independent

[2] joint probability

[3] Independent and identically distributed

در برخی موارد، به‌ویژه در مجموعه داده‌های پیوسته، می‌توان فرض کرد که نمونه‌ها از توزیع گاوسی گرفته شده‌اند. از این‌رو، فرض می‌کنیم که $p(x_i|y_i) \sim N(\mu_{y_i}, \sigma^{\mathsf{Y}}_{y_i})$ که در آن $\mu_{y_i}$ و $\sigma^{\mathsf{Y}}_{y_i}$ به ترتیب میانگین و واریانس ویژگی $i$اُم نمونه‌های کلاس $y$ هستند. بر این اساس داریم:

$$p(x_i|y_i) = \frac{\mathsf{1}}{\sqrt{\mathsf{Y}\pi}\,\sigma_{y_i}} \exp\left(-\frac{(x_i - \mu_{y_i})^{\mathsf{Y}}}{\mathsf{Y}\sigma^{\mathsf{Y}}_{y_i}}\right).$$

حال، اجازه دهید یک دسته‌بند ساده بیز را با استفاده از مجموعه داده هندوانه در جدول ۶ ـ ۱ را آموزش دهیم و هندوانه T1 زیر را دسته‌بندی کنیم:

| ID | color | root | sound | texture | umbilicus | surface | density | sugar | ripe |
|----|-------|------|-------|---------|-----------|---------|---------|-------|------|
| T1 | green | curly | muffled | clear | hollow | hard | ۰٬۶۹۷ | ۰٬۴۶۰ | ? |

**جدول ۶ ـ ۱** مجموعه داده هندوانه (Zhou, 2021)

| ID | color | root | sound | texture | umbilicus | surface | density | sugar | ripe |
|----|-------|------|-------|---------|-----------|---------|---------|-------|------|
| ۱ | green | curly | muffled | clear | hollow | hard | ۰٬۶۹۷ | ۰٬۴۶۰ | true |
| ۲ | dark | curly | dull | clear | hollow | hard | ۰٬۷۷۴ | ۰٬۳۷۶ | true |
| ۳ | dark | curly | muffled | clear | hollow | hard | ۰٬۶۳۴ | ۰٬۲۶۴ | true |
| ۴ | green | curly | dull | clear | hollow | hard | ۰٬۶۰۸ | ۰٬۳۱۸ | true |
| ۵ | light | curly | muffled | clear | hollow | hard | ۰٬۵۵۶ | ۰٬۲۱۵ | true |
| ۶ | green | slightly curly | muffled | clear | slightly hollow | soft | ۰٬۴۰۳ | ۰٬۲۳۷ | true |
| ۷ | dark | slightly curly | muffled | slightly blurry | slightly hollow | soft | ۰٬۴۸۱ | ۰٬۱۴۹ | true |
| ۸ | dark | slightly curly | muffled | clear | slightly hollow | hard | ۰٬۴۳۷ | ۰٬۲۱۱ | true |
| ۹ | dark | slightly curly | dull | slightly blurry | slightly hollow | hard | ۰٬۶۶۶ | ۰٬۰۹۱ | false |
| ۱۰ | green | straight | dull | clear | flat | soft | ۰٬۲۴۳ | ۰٬۲۶۷ | false |
| ۱۱ | light | straight | dull | blurry | flat | hard | ۰٬۲۴۵ | ۰٬۰۵۷ | false |
| ۱۲ | light | curly | muffled | blurry | flat | soft | ۰٬۳۴۳ | ۰٬۰۹۹ | false |
| ۱۳ | green | slightly curly | muffled | slightly blurry | hollow | hard | ۰٬۶۳۹ | ۰٬۱۶۱ | false |
| ۱۴ | light | slightly curly | dull | slightly blurry | hollow | hard | ۰٬۶۵۷ | ۰٬۱۹۸ | false |
| ۱۵ | dark | slightly curly | muffled | clear | slightly hollow | soft | ۰٬۳۶۰ | ۰٬۳۷۰ | false |
| ۱۶ | light | curly | muffled | blurry | flat | hard | ۰٬۵۹۳ | ۰٬۰۴۲ | false |
| ۱۷ | green | curly | dull | slightly blurry | slightly hollow | hard | ۰٬۷۱۹ | ۰٬۱۰۳ | false |

ابتدا احتمال پیشین را برآورد (تخمین) می‌کنیم:

$$P(ripe = true) = \frac{\mathsf{A}}{\mathsf{1Y}} \approx \mathsf{0.4Y1}$$

$$P(ripe = false) = \frac{\mathsf{9}}{\mathsf{1Y}} \approx \mathsf{0.5Y9}$$

سپس، احتمال شرطی هر ویژگی $p(x_i|y_i)$ را برآورد می‌کنیم:

$$P_{green|true} = P(color = green|ripe = true) = \frac{\mathsf{3}}{\mathsf{A}} = \mathsf{0.3Y5}$$

$$P_{green|false} = P(color = green|ripe = false) = \frac{۳}{۹} \approx ۰.۳۳۳$$

$$P_{curly|true} = P(root = curly|ripe = true) = \frac{۵}{۸} = ۰.۶۲۵$$

$$P_{curly|false} = P(root = curly|ripe = false) = \frac{۳}{۹} \approx ۰.۳۳۳$$

$$P_{muffled|true} = P(sound = muffled|ripe = true) = \frac{۵}{۸} = ۰.۶۲۵$$

$$P_{muffled|false} = P(sound = muffled|ripe = false) = \frac{۴}{۹} \approx ۰.۴۴۴$$

$$P_{clear|true} = P(texture = clear|ripe = true) = \frac{۷}{۸} = ۰.۸۷۵$$

$$P_{clear|false} = P(texture = clear|ripe = false) = \frac{۲}{۹} \approx ۰.۲۲۲$$

$$P_{hollow|true} = P(umbilicus = hollow|ripe = true) = \frac{۵}{۸} = ۰.۶۲۵$$

$$P_{hollow|false} = P(umbilicus = hollow|ripe = false) = \frac{۲}{۹} \approx ۰.۲۲۲$$

$$P_{hard|true} = P(surface = hard|ripe = true) = \frac{۶}{۸} = ۰.۷۵۰$$

$$P_{hard|false} = P(surface = hard|ripe = false) = \frac{۶}{۹} \approx ۰.۶۶۷$$

$$P_{density:۰.۶۹۷|true} = P(density = ۰.۶۹۷|ripe = true)$$

$$= \frac{۱}{\sqrt{۲\pi} * ۰.۱۲۹} \exp\left(-\frac{(۰.۶۹۷ - ۰.۵۷۴)^۲}{۲ * ۰.۱۲۹^۲}\right) \approx ۱.۹۵۹$$

$$P_{density:۰.۶۹۷|false} = P(density = ۰.۶۹۷|ripe = true)$$

$$= \frac{۱}{\sqrt{۲\pi} * ۰.۱۹۵} \exp\left(-\frac{(۰.۶۹۷ - ۰.۴۹۶)^۲}{۲ * ۰.۱۹۵^۲}\right) \approx ۱.۲۰۳$$

$$P_{sugar:۰.۴۶۰|true} = P(sugar = ۰.۴۶۰|ripe = true)$$

$$= \frac{۱}{\sqrt{۲\pi} * ۰.۱۰۱} \exp\left(-\frac{(۰.۴۶۰ - ۰.۲۷۹)^۲}{۲ * ۰.۱۰۱^۲}\right) \approx ۰.۷۸۸$$

$$P_{sugar:۰.۴۶۰|false} = P(sugar = ۰.۴۶۰|ripe = true)$$

$$= \frac{۱}{\sqrt{۲\pi} * ۰.۱۰۸} \exp\left(-\frac{(۰.۴۶۰ - ۰.۱۵۴)^۲}{۲ * ۰.۱۰۸^۲}\right) \approx ۰.۰۶۶$$

از این‌رو داریم:

$$P(ripe = true) \times P_{green|true} \times P_{curly|true} \times P_{muffled|true} \times P_{clear|true}$$
$$\times P_{hollow|true} \times P_{hard|true} \times P_{density:0.697|true}$$
$$\times P_{sugar:۰.٤٦۰|true} \approx ۰.۰۵۲$$

$$P(ripe = false) \times P_{green|false} \times P_{curly|false} \times P_{muffled|false}$$
$$\times P_{clear|false} \times P_{hollow|false} \times P_{hard|false}$$
$$\times P_{density:۰.٦۹۷|false} \times P_{sugar:۰.٤٦۰|false} \approx ٦.۸۰ \times ۱۰^{-۵}$$

از آنجایی که ۱۰$^{-۵}$ × ۶.۸۰ > ۰.۰۵۲، دسته‌بند بیز ساده، کلاس نمونه آزمایشی T1 را به عنوان true انتخاب می‌کند.

### مزایا

- پیاده‌سازی آن آسان است، زیرا فقط احتمال محاسبه می‌شود.
- به داده‌های آموزشی زیادی نیاز ندارد.
- اگر فرض نابسته‌ی مشروط برقرار باشد، می‌تواند بهتر از مدل‌های دیگر عمل کند.
- برای پیش‌بینی کلاس داده‌های آزمایشی سریع‌تر عمل می‌کند.

### معایب

- فرض نابسته‌ی مشروط همیشه صادق نیست.
- در مجموعه داده آزمایشی، اگر یک ویژگی رسته‌ای دارای دسته‌ای باشد که در مجموعه آموزشی مشاهده نشده است، مدل به آن یک احتمال ۰ (صفر) اختصاص می‌دهد و قادر به پیش‌بینی نخواهد بود. این اغلب به عنوان **بسامد صفر**[1] شناخته می‌شود.

### بیز ساده در پایتون

#### مجموعه داده

برای مثال بیز ساده از مجموعه داده iris استفاده می‌کنیم.

#### وارد کردن کتاب‌خانه‌ها

```
In [1]:   import numpy as np
          import matplotlib.pyplot as plt
          import pandas as pd
```

---

۱

### وارد کردن مجموعه داده

```
In [2]:  url = "https://archive.ics.uci.edu/ml/machine-learning-
         databases/iris/iris.data"

         # Assign colum names to the dataset
         names = ['sepal-length', 'sepal-width', 'petal-length',
         'petal-width', 'Class']

         # Read dataset to pandas dataframe
         dataset = pd.read_csv(url, names=names)
```

### پیش‌پردازش

```
In [4]:  X = dataset.iloc[:, :-1].values
         y = dataset.iloc[:, 4].values
```

### تقسیم مجموعه داده

```
In [5]:  from sklearn.model_selection import train_test_split

         X_train, X_test, y_train, y_test = train_test_split(X, y,
         test_size =0.25, random_state=42)
```

### مقیاس‌بندی ویژگی‌ها

```
In [1]:  from sklearn.preprocessing import StandardScaler
         scaler = StandardScaler()
         scaler.fit(X_train)

         X_train = scaler.transform(X_train)
         X_test = scaler.transform(X_test)
```

### آموزش و پیش‌بینی

ابتدا داده‌ها را به مجموعه های آموزشی و آزمایشی تقسیم کرده‌ایم و سپس مقیاس‌بندی ویژگی‌ها را برروی داده‌ها انجام دادیم. اکنون زمان آموزش بیز ساده بر روی داده‌های آموزشی است. در این مرحله از کلاس GaussianNB از کتابخانه sklearn.naive_bayes استفاده می‌کنیم. در اینجا ما از یک مدل گاوسی استفاده کرده‌ایم، چندین مدل دیگر مانند برنولی، چندجمله‌ای و غیره نیز وجود دارد.

```
In [1]:  from sklearn.naive_bayes import GaussianNB
         classifier = GaussianNB()
         classifier.fit(X_train, y_train)
```

مرحله آخر این است که مدل ساخته‌شده را برروی داده‌های آزمایشی خود پیش‌بینی کنیم.

```
In [1]:   y_pred = classifier.predict(X_test)
```

## ارزیابی الگوریتم

```
In [3]:   from sklearn.metrics import classification_report,
          confusion_matrix
          print(confusion_matrix(y_test, y_pred))
          print(classification_report(y_test, y_pred))
```

```
Out [7]:   [[15  0  0]
            [ 0 11  0]
            [ 0  0 12]]
```

|  | precision | recall | f1-score | support |
|---|---|---|---|---|
| Iris-setosa | 1.00 | 1.00 | 1.00 | 15 |
| Iris-versicolor | 1.00 | 1.00 | 1.00 | 11 |
| Iris-virginica | 1.00 | 1.00 | 1.00 | 12 |
|  |  |  |  |  |
| accuracy |  |  | 1.00 | 38 |
| macro avg | 1.00 | 1.00 | 1.00 | 38 |
| weighted avg | 1.00 | 1.00 | 1.00 | 38 |

نتایج نشان می‌دهد که مدل بیز ساده گاوسی ما قادر است تمام ۳۸ رکورد موجود در مجموعه آزمون را با دقت ۱۰۰٪ دسته‌بندی کند.

## رگرسیون

به طور کلی دو راه برای استخراج دانش وجود دارد، اول از طریق کارشناسان حوزه و دوم با استفاده از یادگیری ماشین. برای حجم بسیار زیادی از داده‌ها، کارشناسان بسیار مفید نیستند و بنابراین ما از یادگیری ماشین برای این کار استفاده می‌کنیم. یکی از راه‌هایی که می‌توانیم از یادگیری ماشین استفاده کنیم، تکرار منطق کارشناسان در قالب الگوریتم‌ها است، با این حال، این کار بسیار خسته‌کننده، زمان‌بر و پرهزینه است و علاوه‌بر این شاید نتواند به چیزی که نیاز داریم دست یابد. از این‌رو، راه حل این مشکل حرکت به سمت الگوریتم‌های **یادگیری استقرایی**[1] است که خود استراتژی انجام یک کار را تولید می‌کنند و نیازی به دستورالعمل جداگانه در هر مرحله ندارند. دسته‌بندی و رگرسیون دو الگوریتمی در یادگیری ماشین هستند که در این دسته قرار می‌گیرند. برخلاف فرآیندهای دسته‌بندی، که در آن سعی در پیش‌بینی برچسب‌های کلاس به

---

[1] inductive learning

صورت گسسته است، مدل‌های رگرسیون مقادیر عددی را پیش‌بینی می‌کنند. به عبارت دیگر، رگرسیون یک مساله یادگیری بانظارت است که در آن ورودی $x$ و خروجی $y$ وجود دارد و وظیفه، یادگیری نگاشت از ورودی به خروجی است. فرض کنید می‌خواهیم سیستمی داشته باشیم که بتواند قیمت یک خودروی دست دوم را پیش‌بینی کند. ورودی‌ها، ویژگی‌های خودرو همانند برند، سال، مسافت پیموده شده و اطلاعات دیگری که به اعتقاد ما بر ارزش خودرو تاثیر می‌گذارد و خروجی قیمت خودرو است. یا ناوبری یک ربات متحرک (اتومبیل خودران) را در نظر بگیرید؛ خروجی زاویه‌ای است که در هر بار فرمان باید بچرخد تا بدون برخورد به موانع و انحراف از مسیر پیشروی کند و ورودی‌ها توسط حسگرهای برروی اتومبیل همانند دوربین فیلم‌برداری، GPS و غیره ارائه می‌شوند.

## نحوه کار رگرسیون

در رگرسیون، می‌خواهیم خروجی عددی $y$ را که متغیر وابسته نامیده می‌شود، به عنوان تابعی از ورودی $x$ بنویسیم که متغیر مستقل نامیده می‌شود. فرض می‌کنیم که خروجی، مجموع تابع $f(x)$ ورودی و مقداری خطای تصادفی $\epsilon$ است که به‌صورت زیر نشان داده می‌شود:

$$y = f(x) + \epsilon$$

در اینجا تابع $f(x)$ ناشناخته است و می‌خواهیم آن را با برآوردگر $g(x; \theta)$ که شامل مجموعه‌ای از پارامترهای $\theta$ است، تقریب بزنیم. فرض می‌کنیم که خطای تصادفی از توزیع نرمال با میانگین ۰ پیروی می‌کند. اگر $x_1, \ldots, x_n$ یک نمونه تصادفی از مشاهدات متغیر ورودی $x$ و $y_1, \ldots, y_n$ مقادیر مشاهده شده مربوط به متغیر خروجی $y$ باشد. آنگاه، با استفاده از این فرض که خطا از توزیع نرمال پیروی می‌کند، می‌توانیم از روش برآورد درست‌نمایی بیشینه برای برآورد مقادیر پارامتر $\theta$ استفاده کنیم. می‌توان نشان داد که مقادیر $\theta$ که تابع درست‌نمایی را بیشینه می‌کنند، مقادیر $\theta$ هستند که مجموع مربع‌های زیر کمینه می‌کنند:

$$E(\theta) = (y_1 - g(x_1, \theta))^2 + \cdots + (y_n - g(x_n, \theta))^2$$

روش یافتن مقدار $\theta$ به عنوان مقدار $\theta$ که $E(\theta)$ را کمینه می‌کند، به عنوان روش **کم‌ترین مربعات معمولی** (OLS)[1] شناخته می‌شود. در این بخش، از این روش برای تخمین پارامترها استفاده می‌کنیم.

---

[1] Ordinary Least Squares

## روش کم‌ترین مربعات معمولی

در روش کم‌ترین مربعات معمولی، مقادیر عرض از مبدا ـ $y$ و شیب به گونه‌ای انتخاب می‌شوند که مجموع مجذور خطاها را کمینه کنند؛ یعنی مجموع مجذورات فاصله عمودی بین مقدار $y$ پیش‌بینی شده و مقدار $y$ واقعی (شکل ۶ ـ ۱۲).

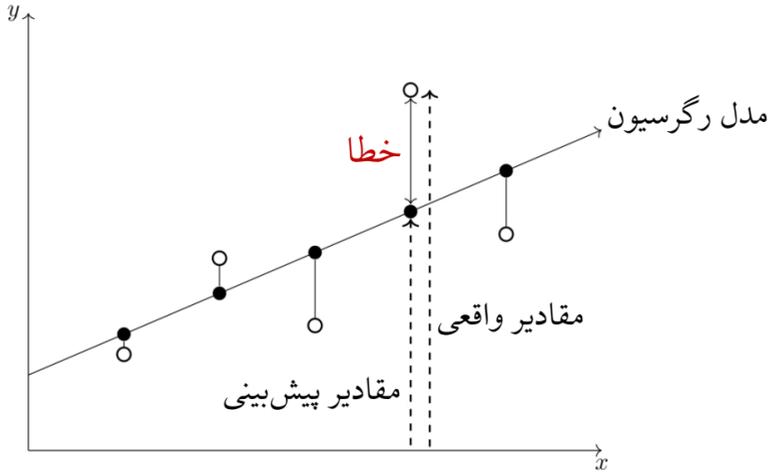

**شکل ۶ ـ ۱۲.** خطا در مقادیر مشاهده شده

اگر $\hat{y}_i$ مقدار پیش‌بینی شده $y_i$ باشد، آنگاه مجموع مربعات خطاها به‌صورت زیر بدست می‌آید:

$$E = \sum_{i=1}^{n} (y_i - \hat{y}_i)^{\mathsf{r}}$$

$$= \sum_{i=1}^{n} (y_i - (\alpha + \beta x_i))^{\mathsf{r}}$$

بنابراین ما باید مقادیر $\alpha$ و $\beta$ را به گونه‌ای پیدا کنیم که $E$ حداقل باشد. می‌توان نشان داد، مقادیر $a$ و $b$ که به ترتیب مقادیر $\alpha$ و $\beta$ هستند که با آن‌ها $E$ حداقل است، می‌تواند با حل معادلات زیر بدست آید:

$$\sum_{i=1}^{n} y_i = na + b \sum_{i=1}^{n} x_i$$

$$\sum_{i=1}^{n} x_i y_i = a \sum_{i=1}^{n} x_i + b \sum_{i=1}^{n} x_i^{\mathsf{r}}$$

**معادله‌های پیدا کردن $a$ و $b$**

برای یادآوری، میانگین $x$ و $y$ به صورت زیر بدست می‌آید:

$$\bar{x} = \frac{1}{n} \sum x_i$$

$$\bar{y} = \frac{1}{n} \sum y_i$$

و همچنین واریانس $x$ توسط

$$Var(x) = \frac{1}{n-1} \sum (x_i - \bar{x})^2$$

حساب می‌شود.

کوواریانس $x$ و $y$ که با $Cov(x, y)$ نشان داده می‌شود به صورت تعریف می‌شود:

$$Cov(x, y) = \frac{1}{n-1} \sum (x_i - \bar{x})(y_i - \bar{y})$$

می‌توان نشان داد که مقادیر $a$ و $b$ را می‌توان با استفاده از معادله‌های زیر محاسبه کرد:

$$b = \frac{Cov(x, y)}{Var(x)}$$

$$a = \bar{y} - b\bar{x}$$

# رگرسیون خطی ساده

اصطلاح "خطی بودن" در جبر به رابطه خطی بین دو یا چند متغیر اشاره دارد که اگر این رابطه را در یک فضای دو بعدی رسم کنیم، یک خط مستقیم به دست می‌آید. بیایید سناریویی را در نظر بگیریم که در آن می‌خواهیم رابطه خطی بین تعداد ساعات مطالعه یک دانش‌آموز و درصد نمره‌هایی را که دانش‌آموز در یک امتحان کسب می‌کند، تعیین کنیم. می‌خواهیم بفهمیم که با توجه به تعداد ساعاتی که یک دانش‌آموز خود را برای امتحان آماده می‌کند، چقدر می‌تواند نمره بالایی کسب کند؟ اگر متغیر مستقل (ساعت) را روی محور $x$ و متغیر وابسته (درصد) را روی محور $y$ رسم کنیم، رگرسیون خطی، خط مستقیمی را به ما می‌دهد که بهترین تناسب را با نقاط داده دارد.

از دوران دبیرستان بیاد داریم که معادله یک خط مستقیم اساسا به صورت زیر است:

$$y = mx + b$$

که در آن $b$ عرض از مبدا و $m$ شیب خط است. بنابراین، اساسا الگوریتم رگرسیون خطی بهینه‌ترین مقدار را برای عرض از مبدا و شیب (در دو بعد) به ما می‌دهد. متغیرهای $y$ و $x$ ثابت

می‌مانند، زیرا آن‌ها ویژگی‌های داده هستند و قابل تغییر نیستند. از این‌رو، مقادیری که ما می‌توانیم کنترل کنیم عبارتند از عرض از مبدا و شیب. بسته به مقادیر عرض از مبدا و شیب می‌تواند چندین خط مستقیم وجود داشته باشد. اساساً کاری که الگوریتم رگرسیون خطی انجام می‌دهد این است که چندین خط را در نقاط داده قرار می‌دهد و خطی را برمی‌گرداند که کم‌ترین خطا را داشته باشد.

فرض کنید $x$ متغیر پیش‌بینی کننده مستقل و $y$ متغیر وابسته باشد و همچنین فرض کنید مجموعه‌ای از مقادیر مشاهده شده $x$ و $y$ داریم. یک مدل رگرسیون خطی ساده، رابطه بین $x$ و $y$ را با استفاده از یک خط، توسط یک معادله به شکل زیر تعریف می‌کند:

$$y = \alpha + \beta x$$

که به منظور تعیین برآوردهای بهینه $\alpha$ و $\beta$، از روش تخمین کم‌ترین مربعات معمولی که پیش‌تر به شرح آن پرداختیم، استفاده می‌شود.

برای درک بهتر اجازه دهید یک مثال را تشریح کنیم. با فرض اینکه $y$ متغیر مستقل است، یک رگرسیون خطی برای داده‌های زیر بدست می‌آوریم:

| $x$ | ۱/۰ | ۲/۰ | ۳/۰ | ٤/۰ | ۵/۰ |
|---|---|---|---|---|---|
| $y$ | ۱/۰۰ | ۲/۰۰ | ۱/۳۰ | ۳/۷۵ | ۲/۲۵ |

داریم:

$$n = ۵$$

$$\bar{x} = \frac{۱}{۵}(۱.۰ + ۲.۰ + ۳.۰ + ٤.۰ + ۵.۰)$$

$$= ۳.۰$$

$$\bar{y} = \frac{۱}{۵}(۱.۰۰ + ۲.۰۰ + ۱.۳۰ + ۳.۷۵ + ۲.۲۵)$$

$$= ۲.۰٦$$

$$Cov(x, y) = \frac{۱}{٤}[(۱.۰ - ۳.۰)(۱.۰۰ - ۲.۰٦) + \cdots + (۵.۰ - ۳.۰)(۲.۲۵ - ۲.۰٦)]$$

$$= ۱.۰٦۲۵$$

$$Var(x) = \frac{۱}{٤}[(۱.۰ - ۳.۰)^۲ + \cdots + (۵.۰ - ۳.۰)^۲]$$

$$= ۲.۵$$

$$b = \frac{۱.۰٦۲۵}{۲.۵}$$

$$= ۰.٤۲۵$$

$$a = ۲.۰٦ - ۰.٤۲۵ \times ۳.۰$$

$$= ۰.۷۸۵$$

بنابراین مدل رگرسیون خطی برای داده‌ها به‌صورت زیر می‌باشد:

$$y = ۰.۷۸۵ + ۰.۴۲۵x$$

همچنین می‌توانید کد زیر را در پایتون اجرا کنید:

```
In [1]:   #data
          X=[[1.0],[2.0],[3.0],[4.0],[5.0]]
          Y=[1.00, 2.00, 1.30, 3.75, 2.25]

          from sklearn.linear_model import LinearRegression
          lr = LinearRegression()
          lr.fit(X, Y)
          #print
          print("b:","%.3f" % round(lr.coef_[0], 3))
          print("a:","%.3f" % round(lr.intercept_, 3))
Out [1]:  b: 0.425
          a: 0.785
```

## رگرسیون چندجمله‌ای

فرض کنید $x$ متغیر پیش‌بینی‌کننده مستقل و $y$ متغیر وابسته باشد، از این‌رو، یک مدل رگرسیون چندجمله‌ای رابطه بین $x$ و $y$ را با یک معادله به شکل زیر تعریف می‌کنند:

$$y = \alpha. + \alpha_۱ x + \alpha_۲ x^۲ + \cdots + \alpha_k x^k$$

برای تعیین مقادیر بهینه پارامترهای $\alpha., \alpha_۱, ..., \alpha_k$ از روش کم‌ترین مربعات معمولی استفاده می‌شود. مقادیر پارامترها، مقادیری هستند که مجموع مربع‌ها را کمینه می‌کنند:

$$E = \sum_{i=۱}^{n} [y_i - (\alpha. + \alpha_۱ x_i + \alpha_۲ x_i^۲ + \cdots + \alpha_k x_i^k)]^۲$$

مقادیر بهینه پارامترها با حل دستگاه معادلات زیر بدست می‌آید:

$$\frac{\partial E}{\partial \alpha_i} = ۰, i = ۰, ۱, ..., k.$$

فرض کنید مقادیر پارامترهایی که $E$ را کمینه می‌کنند به صورت زیر باشند:

$$\alpha_{i=}a_i, i = ۰, ۱, ..., n.$$

براین اساس، می‌توان دید که مقادیر $a_i$ را می‌توان با حل دستگاه معادلات خطی $(k + ۱)$ بدست آورد:

$$\sum y_i = \alpha_0 n + \alpha_1 \left(\sum x_i\right) + \cdots + \alpha_k \left(\sum x_i^k\right)$$
$$\sum y_i x_i = \alpha_0 \left(\sum x_i\right) + \alpha_1 \left(\sum x_i^2\right) + \cdots + \alpha_k \left(\sum x_i^{k+1}\right)$$
$$\sum y_i x_i^2 = \alpha_0 \left(\sum x_i^2\right) + \alpha_1 \left(\sum x_i^3\right) + \cdots + \alpha_k \left(\sum x_i^{k+2}\right)$$
$$\vdots$$
$$\sum y_i x_i^k = \alpha_0 \left(\sum x_i^k\right) + \alpha_1 \left(\sum x_i^{k+1}\right) + \cdots + \alpha_k \left(\sum x_i^{2k}\right)$$

با حل این دستگاه معادلات خطی، مقادیر بهینه پارامترها را بدست می‌آوریم. برای درک بهتر اجازه دهید یک مثال را تشریح کنیم. قصد داریم برای داده‌های زیر، یک مدل رگرسیون درجه دوم پیدا کنیم:

| $x$ | ٣ | ٤ | ٥ | ٦ | ٧ |
|---|---|---|---|---|---|
| $y$ | ٢/٥ | ٣/٢ | ٣/٨ | ٦/٥ | ١١/٥ |

اجازه دهید مدل رگرسیون درجه دوم به صورت زیر باشد:

$$y = \alpha_0 + \alpha_1 x + \alpha_2 x^2.$$

مقادیر $\alpha_0$، $\alpha_1$ و $\alpha_2$ که مجموع مربعات خطا را کمینه می‌کند، $a_0$، $a_1$ و $a_2$ هستند که دستگاه معادلات زیر را برآورده می‌کنند:

$$\sum y_i = na_0 + a_1 \left(\sum x_i\right) + a_2 \left(\sum x_i^2\right)$$
$$\sum y_i x_i = a_0 \left(\sum x_i\right) + a_1 \left(\sum x_i^2\right) + a_2 \left(\sum x_i^3\right)$$
$$\sum y_i x_i^2 = a_0 \left(\sum x_i^2\right) + a_1 \left(\sum x_i^3\right) + a_2 \left(\sum x_i^4\right)$$

با استفاده از داده‌های ارائه شده، داریم:

$$٢٧.٥ = ٥a_0 + ٢٥a_1 + ١٣٥a_2$$

$$١٥٨.٨ = ٢٥a_0 + ١٣٥a_1 + ٧٧٥a_2$$

$$٩٦٦.٢ = ١٣٥a_0 + ٧٧٥a_1 + ٤٦٥٩a_2$$

با حل این دستگاه معادلات بدست می‌آوریم:

$$a_0 = ١٢.٤٢٨٥٧١٤$$

$$a_1 = -٥.٥١٢٨٥٧١$$

$$a_2 = ٠.٧٦٤٢٨٥٧$$

از این‌رو، مدل چند جمله‌ای درجه دوم خواسته شده به‌صورت زیر است:

$$y = ١٢.٤٢٨٥٧١٤ - ٥.٥١٢٨٥٧١x + ٠.٧٦٤٢٨٥٧x^2$$

همچنین می‌توانید کد زیر را در پایتون اجرا کنید:

```
In [1]:  #data
         X=[[3],[4],[5],[6],[7]]
         Y=[2.5, 3.2, 3.8, 6.5, 11.5]

         from sklearn.linear_model import LinearRegression
         from sklearn.preprocessing import PolynomialFeatures
         poly_reg = PolynomialFeatures(degree=2)
         X_poly = poly_reg.fit_transform(X)
         lin_reg2 = LinearRegression()
         lin_reg2.fit(X_poly,Y)
         #print
         print(lin_reg2.coef_)
         print(lin_reg2.intercept_)
Out [1]: array([ 0.        , -5.51285714,  0.76428571])
         12.428571428571345
```

## رگرسیون خطی چندگانه

فرض می‌کنیم $N$ متغیر مستقل $x_1, x_2, \ldots, x_N$ وجود دارد و همچنین متغیر $y$ وابسته است. همچنین فرض کنید که $n$ مقدار مشاهده‌شده از این متغیرها وجود داشته باشد:

| متغیرها (ویژگی‌ها) | مقادیر (نمونه‌ها) | | | |
|---|---|---|---|---|
| | نمونه ۱ | نمونه ۲ | ... | نمونه $n$ |
| $x_1$ | $x_{11}$ | $x_{12}$ | ... | $x_{1n}$ |
| $x_2$ | $x_{21}$ | $x_{22}$ | ... | $x_{2n}$ |
| ... | | | | |
| $x_N$ | $x_{N1}$ | $x_{N1}$ | ... | $x_{Nn}$ |
| $y$ | $y_1$ | $y_2$ | ... | $y_n$ |

مدل رگرسیون خطی چندگانه رابطه بین $N$ متغیر مستقل و متغیر وابسته را با معادله‌ای به شکل زیر تعریف می‌کند:

$$y = \beta_0 + \beta_1 x_1 + \cdots + \beta_N x_N$$

همانند رگرسیون خطی ساده، در اینجا نیز از روش کم‌ترین مربعات معمولی برای بدست آوردن تخمین‌های بهینه $\beta_0, \beta_1, \ldots, \beta_N$ استفاده می‌کنیم. اگر

$$X = \begin{bmatrix} 1 & x_{11} & x_{21} & \ldots & x_{N1} \\ 1 & x_{12} & x_{22} & \ldots & x_{N2} \\ \vdots & & & & \\ 1 & x_{1n} & x_{2n} & \ldots & x_{Nn} \end{bmatrix}, Y = \begin{bmatrix} y_1 \\ y_2 \\ \vdots \\ y_n \end{bmatrix}, B = \begin{bmatrix} \beta_0 \\ \beta_1 \\ \vdots \\ \beta_N \end{bmatrix}$$

آنگاه، می‌توان نشان داد که ضرایب رگرسیون به‌صورت ارائه می‌شود:

$$B = (X^T X)^{-1} X^T Y$$

برای درک بهتر اجازه دهید یک مثال را تشریح کنیم. می‌خواهیم یک مدل رگرسیون خطی چندگانه را برای داده‌های زیر بدست آوریم:

| $x_1$ | ۱ | ۱ | ۲ | ۰ |
|---|---|---|---|---|
| $x_۲$ | ۱ | ۲ | ۲ | ۱ |
| $y$ | ۳/۲۵ | ۶/۵ | ۳/۵ | ۵/۰ |

در این مساله، دو متغیر مستقل و چهار مجموعه از مقادیر متغیرها وجود دارد. از این‌رو، برای نمادهای استفاده شده بالا داریم، ۲ $n =$ و ٤ $= N$. مدل رگرسیون خطی چندگانه برای این مساله به صورت زیر است:

$$y = \beta. + \beta_۱ x_۱ + \beta_۲ x_۲$$

محاسبات در زیر نشان داده شده است:

$$X = \begin{bmatrix} ۱ & ۱ & ۱ \\ ۱ & ۱ & ۲ \\ ۱ & ۲ & ۲ \\ ۱ & ۰ & ۱ \end{bmatrix}, Y = \begin{bmatrix} ۳.۲۵ \\ ۶.۵ \\ ۳.۵ \\ ۵.۰ \end{bmatrix}, B = \begin{bmatrix} \beta. \\ \beta_۱ \\ \beta_۲ \end{bmatrix}$$

$$X^T X = \begin{bmatrix} ٤ & ٤ & ٦ \\ ٤ & ٦ & ٧ \\ ٦ & ٧ & ۱۰ \end{bmatrix}$$

$$(X^T X)^{-۱} = \begin{bmatrix} \frac{۱۱}{٤} & \frac{۱}{۲} & -۲ \\ \frac{۱}{۲} & ۱ & -۱ \\ -۲ & -۱ & ۲ \end{bmatrix}$$

$$B = (X^T X)^{-۱} X^T Y$$

$$= \begin{bmatrix} ۲.۰۶۲۵ \\ -۲.۳۷۵۰ \\ ۳.۲۵۰۰ \end{bmatrix}$$

از این‌رو، مدل خواسته شده به‌صورت زیر است:

$$y = ۲.۰۶۲۵ - ۲.۳۷۵۰ x_۱ + ۳.۲۵۰۰ x_۲$$

همچنین می‌توانید کد زیر را در پایتون اجرا کنید:

```
In [1]:  #data
         X=[[1,1],[1,2],[2,2],[0,1]]
         Y=[3.25, 6.5, 3.5, 5.0]
```

```
from sklearn.linear_model import LinearRegression
lr = LinearRegression()
lr.fit(X, Y)
#print
print(lr.coef_)
print(lr.intercept_)
```
Out [1]:    [-2.375  3.25 ]
            2.0625

صفحه رگرسیون برای مثال داده شده:

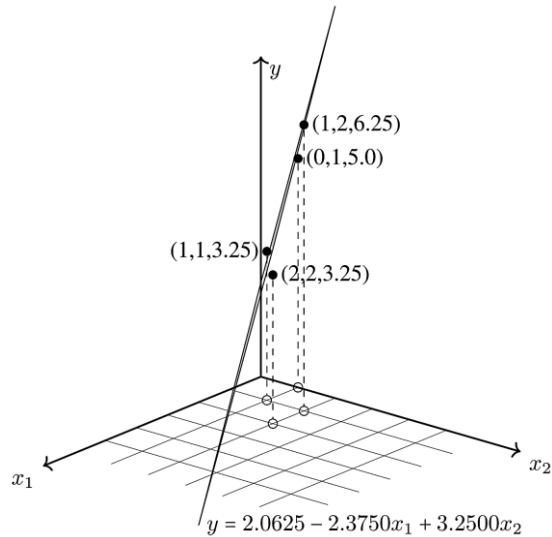

$$y = 2.0625 - 2.3750x_1 + 3.2500x_2$$

# خلاصه فصل

- دسته‌بندی دودویی که در آن هر نمونه تنها به یکی از دو دسته‌ای از پیش تعریف‌شده اختصاص داده می‌شود، ساده‌ترین نوع دسته‌بندی است.

- دسته‌بندی چندبرچسبی حالت تعمیم یافته‌ای از دسته‌بندی تک‌برچسبی است.

- الگوریتم‌های یادگیری ماشین را می‌توان به دو دسته پارامتری یا ناپارامتری طبقه‌بندی کرد.

- بزرگ‌ترین نقطه ضعف روش‌های پارامتری این است که فرضیاتی که می‌کنیم ممکن است همیشه درست نباشند.

- الگوریتم‌هایی که هیچ فرض خاصی در مورد نوع تابع نگاشت ندارند به عنوان الگوریتم‌های ناپارامتری شناخته می‌شوند

- هنگامی که یک الگوریتم یادگیری ماشین بلافاصله پس از دریافت مجموعه داده‌های آموزشی، مدلی را می‌سازد، به آن یادگیرنده مشتاق می‌گویند.

- زمانی که یک الگوریتم یادگیری ماشین بلافاصله پس از دریافت داده‌های آموزشی مدلی را نمی‌سازد، بلکه منتظر می‌ماند تا داده‌های ورودی برای ارزیابی ارائه شود، به آن یادگیرنده تنبل می‌گویند.

- الگوریتم KNN بر اساس این فرض طراحی شده است که چیزهای مشابه در نزدیکی یکدیگر وجود دارند.

- کا‌ـ‌نزدیک‌ترین همسایه فضای الگو را برای k نمونه آموزشی که نزدیک‌ترین به نمونه ناشناخته هستند، جستجو می‌کند.

- الگوریتم KNN هم برای دسته‌بندی و هم رگرسیون استفاده می‌شود.

- ماشین‌های بردار پشتیبان با دیگر الگوریتم‌های طبقه‌بندی متفاوت هستند، زیرا آنها مرز تصمیم را انتخاب می‌کنند که فاصله را نزدیک‌ترین نقاط داده همه کلاس‌ها به حداکثر می‌رساند.

- مسائل دنیای واقعی در اغلب موارد به صورت خطی قابل تفکیک نیستند، از این‌رو نمی‌توانید از ماشین بردار پشتیبان حاشیه سخت در این مسائل استفاده کنید.

- یکی از محبوب‌ترین الگوریتم‌های یادگیری ماشین، درختان تصمیم به دلیل نحوه عملکرد بسیار ساده آنها است.

- درختان تصمیم نوعی مدل ناپارامتری هستند.

- ساخت درختان تصمیم از استراتژی تقسیم و حل پیروی می‌کند.

- آموزش درخت تصمیم یک فرآیند از بالا به پایین است.

- استراتژی‌های کلی هرس در درختان تصمیم شامل پیشا‌ـ‌هرس و پسا‌ـ‌هرس است.

- پیشا‌ـ‌هرس، بهبود توانایی تعمیم دادن هر انشعاب را ارزیابی می‌کند و اگر بهبود کوچک باشد، انشعاب را لغو می‌کند.

- پسا‌ـ‌هرس، گره‌های غیر برگِ یک درخت تصمیمِ کاملا رشد یافته را دوباره بررسی می‌کند و اگر جایگزینی منجر به بهبود توانایی تعمیم شود، یک گره با یک گره برگ جایگزین می‌شود.

- بیز ساده یک تکنیک دسته‌بندی بر اساس قضیه بیز است که تمام ویژگی‌هایی که مقدار هدف را پیش‌بینی می‌کنند مستقل از یکدیگر هستند

- برخلاف فرآیندهای دسته‌بندی، که در آن سعی در پیش‌بینی برچسب‌های کلاس به صورت گسسته است، مدل‌های رگرسیون مقادیر عددی را پیش‌بینی می‌کنند.

- در رگرسیون، می‌خواهیم خروجی عددی y را که متغیر وابسته نامیده می‌شود، به عنوان تابعی از ورودی x بنویسیم که متغیر مستقل نامیده می‌شود.

## مراجع برای مطالعه بیشتر

# یادگیری عمیق

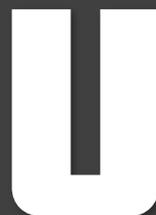

## اهداف:

- یادگیری عمیق چیست؟
- چه تفاوتی با یادگیری ماشین دارد؟
- چه زمانی از یادگیری عمیق استفاده کنیم؟
- معایب و چالش‌ها
- آشنایی با شبکه‌های عصبی عمیق
- بهینه‌سازی در شبکه‌های عمیق
- اهمیت و کابرد علم داده

# یادگیری عمیق چیست؟

یادگیری عمیق زیرمجموعه‌ای از یادگیری ماشین است که به ماشین‌ها می‌آموزد تا کارهایی را انجام دهند که انسان‌ها به طور طبیعی با آن متولد می‌شوند: یادگیری از طریق الگو. اگرچه این فناوری اغلب مجموعه‌ای از الگوریتم‌ها در نظر گرفته می‌شود که "مغز را تقلید می‌کنند"، توصیف مناسب‌تر مجموعه‌ای از الگوریتم‌هایی است که "از طریق لایه‌ها یاد می‌گیرند". به عبارت دیگر، شامل یادگیری از طریق لایه‌هایی است که الگوریتم را قادر می‌سازد تا سلسله‌مراتبی از مفاهیم پیچیده را از مفاهیم ساده‌تر ایجاد کند. این رشته طی سال‌های اخیر توجه زیادی را به خود جلب کرده است و دلایل خوبی برای این امر وجود دارد. چراکه تحولات اخیر به نتایجی منجر شده است که تا پیش از آن تصور نمی‌شد امکان‌پذیر باشد.

یادگیری عمیق الگوریتم‌هایی را توصیف می‌کند که داده‌ها را با ساختاری منطقی، شبیه به نحوهٔ نتیجه‌گیریِ یک انسان تجزیه و تحلیل می‌کند. توجه داشته باشید که این امر می‌تواند هم از طریق یادگیری بانظارت و هم غیرنظارتی اتفاق بیفتد. برای دستیابی به این هدف، برنامه‌های کاربردی یادگیری عمیق از ساختار لایه‌ای (سلسله‌مراتبی) از الگوریتم‌ها به نام شبکه عصبی مصنوعی استفاده می‌کنند. طراحی چنین شبکه عصبی مصنوعی از شبکه عصبی بیولوژیکی مغز انسان الهام گرفته شده است و منجر به فرآیند یادگیری بسیار بهتر از مدل‌های یادگیری ماشین استاندارد می‌شود. از طریق یادگیری عمیق، ماشین‌ها می‌توانند از تصاویر، متن یا فایل‌های صوتی برای شناسایی و انجام هر کاری به روشی شبیه انسان استفاده کنند.

یادگیری عمیق بر روش‌های یادگیری تکراری تمرکز می‌کند که ماشین‌ها را در معرض مجموعه داده‌های عظیم قرار می‌دهد. با انجام این کار، به رایانه‌ها کمک می‌کند تا ویژگی‌ها از داده‌ها پیدا کنند و با تغییرات سازگار شوند. قرار گرفتن مکرر در معرض مجموعه داده‌ها به ماشین‌ها کمک می‌کند تا تفاوت‌ها و منطق داده‌ها را درک کنند و به یک نتیجه‌گیری قابل اعتماد برسند.

## چرا یادگیری عمیق مهم است؟

اهمیت یادگیری عمیق را می‌توان بیشتر به این واقعیت مرتبط دانست که دنیای امروز ما در حال تولید مقادیرِ تصاعدی داده‌ای است. نتیجه آن نیاز به سیستمی است تا بتواند در ساختاردهی این مقیاس بزرگ از داده بپردازد. یادگیری عمیق از این حجم روبه‌رشد داده‌ها استفاده می‌کند. تمام اطلاعات جمع‌آوری شده از این داده‌ها برای دستیابی به نتایج دقیق از طریق مدل‌های یادگیری عمیق استفاده می‌شود. تجزیه و تحلیل مکرر مجموعه داده‌های عظیم، خطاها را کاهش می‌دهد که در نهایت به یک نتیجه قابل اعتماد منجر می‌شود. یادگیری عمیق همچنان در فضاهای تجاری و شخصی تأثیر می‌گذارد و فرصت‌های شغلی زیادی را در زمان آینده ایجاد می‌کند.

احتمالا طی چند سال آینده، استارت‌آپ‌ها و شرکت‌های بزرگ و کوچکِ فناوری از یادگیری عمیق برای ارتقاء مجموعه وسیعی از برنامه‌های کاربردی موجود و ایجاد محصولات و خدمات جدید استفاده خواهند کرد (چه‌بسا که اکنون هم بسیاری از شرکت‌ها از این فناوری بزرگ در حال استفاده هستند). خطوط و بازارهای تجاری کاملا جدیدی پدید خواهند آمد که به نوبه خود باعث نوآوری بیشتر خواهد شد. استفاده از سیستم‌های یادگیری عمیق آسان‌تر و در دسترس‌تر خواهد بود و پیش‌بینی می‌شود که یادگیری عمیق نحوه تعامل مردم با فناوری را بطور اساسی تغییر خواهد داد؛ همان‌طورکه سیستم‌های عامل دسترسی مردم عادی به رایانه را تغییر دادند.

## یادگیری عمیق چگونه کار می‌کند؟

سیستم‌های یادگیری عمیق بر اساس شبکه‌های عصبی در نئوکورتکس مغز انسان، جایی که شناختِ سطح بالاتری رخ می‌دهد، مدل‌سازی می‌شوند. در مغز، یک نورون اطلاعات الکتریکی یا شیمیایی را منتقل می‌کند. هنگامی که با نورون‌های دیگر متصل می‌شود، یک شبکه عصبی را تشکیل می‌دهد. در ماشین‌ها، نورون‌ها به‌صورت مصنوعی ساخته می‌شوند و اگر به اندازه کافی از این نورون‌های مصنوعی را بهم متصل کنید، یک شبکه عصبی مصنوعی بدست می‌آورید.

برای اینکه یک شبکه عصبی مفید باشد، نیاز به آموزش دارد. برای آموزش یک شبکه عصبی، مجموعه‌ای از نورون‌های مصنوعی ایجاد و یک "وزنِ" عددی تصادفی به آن‌ها اختصاص می‌یابد که تعیین می‌کند نورون‌ها چگونه به داده‌های جدید واکنش نشان می‌دهند. مانند هر روش آماری یا یادگیری ماشین، ماشین در ابتدا پاسخ‌های صحیح را نیز می‌بیند. بنابراین اگر شبکه ورودی را به طور دقیق شناسایی نکند، مثلا یک چهره را در یک تصویر نبیند، آنگاه سیستم وزن‌ها را تنظیم می‌کند (میزان توجه هر نورون به داده‌ها، به منظور تولید پاسخ درست). در نهایت، پس از آموزش کافی، شبکه عصبی به طور مداوم الگوهای صحیح در گفتار یا تصاویر را تشخیص خواهد داد.

بطور خلاصه، هسته اصلی یادگیری عمیق به روشی تکراری برای آموزش ماشین‌ها برای تقلید از هوش انسانی متکی است. یک شبکه عصبی مصنوعی این روش تکراری را از طریق چندین سطح سلسله مراتبی انجام می‌دهد. سطوح اولیه به ماشین‌ها کمک می‌کنند تا اطلاعات ساده را بیاموزند. با رفتن به هر سطح جدید، ماشین‌ها اطلاعات بیشتری را جمع‌آوری کرده و آن‌ها را با آنچه در آخرین سطح آموخته بوده‌اند ترکیب می‌کنند. در پایان فرآیند، سیستم یک قطعه اطلاعات نهایی را جمع‌آوری می‌کند که یک ورودی ترکیبی است. این اطلاعات از چندین سلسله مراتب عبور می‌کند و شبیه به تفکرِ منطقیِ پیچیده است. بیایید با کمک یک مثال آن را بیشتر تجزیه کنیم. دستیار صوتی مانند الکسا یا سیری را در نظر بگیرید تا ببینید چگونه از یادگیری عمیق برای تجربیات مکالمه طبیعی استفاده می‌کند. در سطوح اولیه شبکه عصبی، زمانی که دستیار صوتی با داده‌ها تغذیه می‌شود، سعی می‌کند صداها و موارد دیگر را شناسایی کند. در

سطوح بالاتر، اطلاعات مربوط به واژگان را میگیرد و یافتههای سطوح قبلی را به آن اضافه میکند. در سطوح بعدی، اعلانات (فرمانها) را تجزیه و تحلیل میکند و تمام نتایج خود را ترکیب میکند. برای بالاترین سطحِ ساختارِ سلسله مراتبی، دستیار صوتی به اندازه کافی آموخته است که بتواند یک دیالوگ را تجزیه و تحلیل کند و بر اساس آن ورودی، اقدام مربوط را ارائه دهد.

عملکرد سلسلهمراتبی سیستمهای یادگیری عمیق، ماشینها را قادر میسازد تا دادهها را با رویکرد غیرخطی پردازش کنند. شبکههای عصبی مصنوعی مانند مغز انسان ساخته شدهاند و گرههای عصبی مانند یک شبکه بههم متصل هستند. شبکههای عصبی مصنوعی در سطح اول سلسلهمراتب، چیز سادهای را یاد میگیرند و سپس آن را به سطح بعدی میفرستند. در سطح بعدی، این اطلاعات ساده در چیزی که کمی پیچیدهتر است ترکیب میشود و آن را به سطح بعدی و غیره منتقل میکند. هر سطح در سلسلهمراتب، چیز پیچیدهتری از ورودی دریافتی از سطح قبلی میسازد و میتواند بهطور خودکار استخراج ویژگیها را از مجموعه دادهها مانند تصاویر، ویدیو یا متن، بدون برنامهنویسی صریح یا قوانین سنتی، بیاموزد.

در یادگیری عمیق، ما نیازی به برنامهنویسی صریح همهچیز نداریم. آنها میتوانند بهطور خودکار بازنماییهایی را از دادههایی مانند تصاویر، ویدیو یا متن، بدون معرفی قوانین دستی یاد بگیرند. معماریهای بسیار انعطافپذیر آنها میتوانند مستقیما از دادههای خام یاد بگیرند و در صورت ارائه دادههای بیشتر میتوانند عملکرد پیشبینی خود را افزایش دهند.

ایده نورونهای مصنوعی حداقل ۶۰ سال است که وجود داشته است، زمانی که در دهه ۱۹۵۰، فرانک روزنبلت یک "پرسپترون" از آشکارسازهای نور ساخت و با موفقیت آن را آموزش داد تا تفاوت بین اشکال اصلی را تشخیص دهد. اما شبکههای عصبی اولیه از نظر تعداد نورونهایی که میتوانستند شبیهسازی کنند، بسیار محدود بودند. به این معنا که نمیتوانستند الگوهای پیچیده را تشخیص دهند. سه پیشرفت در دهه گذشته یادگیری عمیق را محبوبتر و بادوام ساخت.

ابتدا جفری هینتون و سایر محققان دانشگاه تورنتو روشی ابداع کردند که نورونهای نرمافزاری بتوانند خود را با لایهبندی آموزش دهند. اولین لایه از نورونها یاد میگیرند که چگونه ویژگیهای اساسی، مثلا یک لبه را با میلیونها نقطه داده تشخیص دهند. هنگامی که یک لایه یاد میگیرد که چگونه این چیزها را به طور دقیق تشخیص دهد، به لایه بعدی میرسد که خود را برای شناسایی ویژگیهای پیچیدهتر، مثلا بینی یاگوش، آموزش میدهد. سپس آن لایه به لایه دیگری تغذیه میشود که خود را آموزش میدهد تا سطوح بیشتری از انتزاع را تشخیص دهد و به همین ترتیب، لایهای پس از لایهای دیگر، تا زمانی که سیستم بتواند به طور قابل اعتماد پدیدههای بسیار پیچیده را مانند یک انسان تشخیص دهد.

دومین توسعه‌یِ مسئول پیشرفت‌های اخیر در یادگیری عمیق، حجم انبوه داده‌ای است که اکنون در دسترس است. دیجیتالی شدن سریع منجر به تولید داده‌های در مقیاس بزرگ شده است و این داده‌ها اکسیژنی برای آموزش سیستم‌های یادگیری عمیق هستند.

سرانجام، تیمی در استنفورد به رهبریِ اندرو انگ متوجه شدند تراشه‌های واحد پردازش گرافیکی یا GPU که برای پردازش بصری بازی‌های ویدیویی اختراع شده‌اند، می‌توانند برای یادگیری عمیق تغییر کاربری دهند. تا همین اواخر، تراشه‌های رایانه‌ای معمولی می‌توانستند تنها یک رویداد را در یک زمان پردازش کنند، اما GPUها برای محاسبات موازی طراحی شده‌اند. استفاده از این تراشه‌ها برای اجرای شبکه‌های عصبیِ با میلیون‌ها اتصال بین آن‌ها به طور موازی، آموزش و توانایی‌های سیستم‌های یادگیری عمیق را سرعت زیادی بخشید و این امکان را برای ماشین فراهم کرد که در یک روز چیزی را بیاموزد که قبلا به چندین هفته زمان نیاز داشت.

پیشرفته‌ترین شبکه‌های یادگیری عمیق امروزی از میلیون‌ها نورون شبیه‌سازی شده با میلیاردها اتصال بین آن‌ها تشکیل شده‌اند و می‌توانند از طریق یادگیری غیرنظارتی آموزش ببینند. این موثرترین کاربرد عملی هوش مصنوعی است که تاکنون ابداع شده است. برای برخی از وظایف، بهترین سیستم‌های یادگیری عمیقِ تشخیص‌دهنده‌های الگو، همتراز با افراد هستند.

## تفاوت یادگیری عمیق با یادگیری ماشین؟

الگوریتم‌های یادگیری عمیق، الگوریتم‌های یادگیری ماشین هستند. از این‌رو، شاید برای شما این سوال مطرح شود که چه چیزی یادگیری عمیق را در زمینه یادگیری ماشین خاص می‌کند. پاسخ: **ساختار الگوریتم شبکه‌های عصبی مصنوعی، نیاز به مداخله کمتر انسانی و نیاز به داده‌های بزرگتر.**

اول از همه، در حالی که الگوریتم‌های یادگیری ماشین سنتی ساختار نسبتا ساده‌ای دارند، مانند رگرسیون خطی یا درخت تصمیم، یادگیری عمیق مبتنی‌بر یک شبکه عصبی مصنوعی است. این شبکه عصبی مصنوعی دارای چندین لایه است و همانند مغز انسان، پیچیده و درهم تنیده است. ثانیاً، الگوریتم‌های یادگیری عمیق به مداخله انسانی بسیار کم‌تری نیاز دارند. به عنوان مثال اگر بخواهیم علامت توقف را در یک تصویر تشخیص دهیم، یک الگوریتم یادگیری ماشین سنتی، به یک مهندس نیاز دارد تا به‌صورت دستی ویژگی‌ها و دسته‌بندی را برای مرتب‌سازی تصاویر انتخاب کند و بررسی کند که آیا خروجی مطابق با نیاز است یا خیر و اگر اینطور نیست، الگوریتم را تنظیم می‌کند. با این حال، به عنوان یک الگوریتم یادگیری عمیق، ویژگی‌ها به‌طور خودکار استخراج می‌شوند و الگوریتم از خطاهای خود یاد می‌گیرد. ثالثاً، یادگیری عمیق به داده‌های بسیار بیشتری نسبت به الگوریتم‌های یادگیری ماشین سنتی برای عملکرد صحیح نیاز دارد. یادگیری ماشین با هزاران نقطه داده کار می‌کند، در حالی‌که یادگیری

عمیق اغلب با میلیون‌ها نقطه. با توجه به ساختاری از چندین لایه پیچیده، یک سیستم یادگیری عمیق به مجموعه داده بزرگی برای حذف نوسانات و ایجاد تفسیرهای باکیفیت بالا نیاز دارد.

بر این اساس، اگرچه یادگیری ماشین و یادگیری عمیق، اغلب به‌جای یکدیگر استفاده می‌شوند، یک چیز نیستند. یادگیری ماشین طیفِ وسیع‌تری است که از داده‌ها برای تعریف و ایجاد مدل‌های یادگیری استفاده می‌کند. یادگیری ماشین سعی می‌کند ساختار داده‌ها را با مدل‌های آماری درک کند و با داده‌کاوی شروع می‌شود. جایی که اطلاعات مربوط را از مجموعه داده‌ها به صورت دستی استخراج می‌کند و پس از آن از الگوریتم‌هایی برای هدایت رایانه‌ها برای یادگیری از داده‌ها و انجام پیش‌بینی استفاده می‌کند.

یادگیری ماشین برای مدت طولانی مورد استفاده بوده و در طول زمان تکامل یافته است. یادگیری عمیق یک زمینه نسبتا جدید است (نه به‌طور کامل جدید، چراکه یادگیری عمیق نیز در طول زمان تکامل یافته) است که برای یادگیری و عملکرد فقط بر روی شبکه‌های عصبی تمرکز دارد. در یادگیری عمیق هرچه داده‌هایِ بیشتری به شبکه تغذیه شود، نتایج دقیق‌تر و دقیق‌تر می‌شوند. این ما را به تفاوت دیگری بین یادگیری عمیق و یادگیری ماشین می‌رساند. در حالی که یادگیریِ مدل‌هایِ یادگیری عمیق می‌تواند با حجمِ بیشتری از داده افزایش یابد، یادگیریِ مدل‌هایِ یادگیری ماشین در یک سطح معین محدود می‌شود. به عبارت دیگر، پس از یک سطح معین به سطح بالایی از یادگیری می‌رسند و هر گونه افزودن داده‌هایِ بیشترِ جدید، تفاوتی را ایجاد نمی‌کند. به‌طور خلاصه، تفاوت‌های اصلی بین این دو دامنه به شرح زیر است:

- **اندازه مجموعه داده:** یادگیری عمیق با مجموعه داده‌های کوچک خوب عمل نمی‌کند. با این حال، الگوریتم‌های یادگیری ماشین می‌توانند مجموعه داده‌های کوچک‌تر را پردازش کنند (همچنین داده‌های بزرگ اما نه به اندازهِ مجموعه داده‌های یادگیری عمیق)، بدون به خطر انداختن عملکرد آن.

- **مهندسی ویژگی:** بخش اساسی همه الگوریتم‌های یادگیری ماشین، مهندسی ویژگی‌ها و پیچیدگی آن است. در یادگیری ماشین سنتی، یک متخصص، ویژگی‌هایی را که باید در مدل اعمال شود، تعریف می‌کند. از سوی دیگر، در یادگیری عمیق، مهندسی ویژگی از طریق ساختار سلسله‌مراتبی شبکه‌های عمیق به‌صورت خودکار و بدونِ برنامه نویسی صریح انجام می‌شود.

- **وابستگی‌های سخت‌افزاری:** یادگیری عمیق نیاز به سخت‌افزار پیشرفته برای عملیات سنگین محاسباتی دارد. از سوی دیگر، الگوریتم‌های یادگیری ماشین را می‌توان برروی ماشین‌های رده پایین نیز اجرا کرد. الگوریتم‌های یادگیری عمیق به GPU نیاز دارند تا بتوان محاسبات پیچیده را به طور موثر بهینه کنند.

- **زمان اجرا:** به راحتی می‌توان تصور کرد که یک الگوریتم یادگیری ماشین زمان اجرای کوتاه‌تری در مقایسه با یادگیری عمیق خواهد داشت. چراکه یادگیری عمیق نه تنها

به دلیل مجموعه داده‌های عظیم، بلکه به دلیل پیچیدگی‌های شبکه عصبی، به یک چارچوب زمانی بیشتر برای آموزش نیاز دارد. آموزش یک الگوریتم یادگیری ماشین از چند ثانیه تا چند ساعت طول می‌کشد، اما در مقایسه، الگوریتم یادگیری عمیق می‌توانند تا هفته‌ها طول بکشند.

> مدل‌های یادگیری عمیق به تنهایی قادر به ایجاد ویژگی‌های جدید هستند، در حالی که در رویکرد یادگیری ماشین، ویژگی‌ها باید توسط کاربران به‌طور دقیق شناسایی شوند.

## مزیت‌های یادگیری عمیق

وقتی به فناوری فکر می‌کنیم، نمی‌توان از یادگیری عمیق بحثی به میان نیاورد. نیازی به گفتن نیست که یادگیری عمیق به یکی از حیاتی‌ترین جنبه‌های فناوری تبدیل شده است. امروزه علاوه بر شرکت‌ها و سازمان‌ها، حتی افرادِ به سمتِ جنبه‌های فناوری، به یادگیری عمیق، تمایل دارند. یکی از دلایل متعددی که یادگیری عمیق همه توجه را به خود جلب می‌کند، توانایی آن در دقت پیش‌بینی‌هایی انجام شده است. به طور خلاصه، شرکت‌ها در موقعیتی هستند که به واسطه یادگیری عمیق، از مزایای مالی و عملیاتی مختلفی بهره‌مند می‌شوند.

ممکن است بپرسید چرا تعداد قابل توجهی از غول‌های بزرگ فناوری امروزه در حال استفاده از یادگیری عمیق هستند و همچنان تعداد این شرکت‌ها در استفاده از یادگیری عمیق روز به روز در حال افزایش است. برای درک این دلیل، باید به مزایایی که می‌توان با استفاده از رویکرد یادگیری عمیق بدست آورد، نگاه کرد. در زیر چند مزیت کلیدی که هنگام استفاده از این فناوری وجود دارد را فهرست کرده‌ایم:

- **حداکثر استفاده از داده‌های بدون ساختار:** تحقیقات نشان می‌دهد که درصد زیادی از داده‌های یک سازمان بدون ساختار هستند، زیرا اکثر آن‌ها در قالب‌های مختلفی همانند تصویر، متن و غیره هستند. برای اکثر الگوریتم‌های یادگیری ماشین، تجزیه و تحلیل داده‌های بدون ساختار دشوار است. از این‌رو، اینجاست که یادگیری عمیق مفید می‌شود. چراکه می‌توانید از قالب‌های داده‌ای مختلف برای آموزش الگوریتم‌های یادگیری عمیق استفاده کنید و همچنین بینش‌های مرتبط با هدف آموزش را بدست آورید. برای مثال، می‌توانید از الگوریتم‌های یادگیری عمیق برای کشف روابط موجود بین تجزیه و تحلیل صنعت، گفتگوی رسانه‌های اجتماعی و موارد دیگر برای پیش‌بینی قیمت‌های سهام آینده یک سازمان استفاده کنید.

- **عدم نیاز به مهندسی ویژگی‌ها:** در یادگیری ماشین، مهندسی ویژگی یک کار اساسی و مهم است. چراکه دقت را بهبود می‌بخشد و گاهی اوقات این فرآیند می‌تواند به دانش دامنه در مورد یک مساله خاص نیاز داشته باشد. یکی از بزرگترین مزایای استفاده از رویکرد یادگیری عمیق، توانایی آن در اجرای مهندسی ویژگی به صورت خودکار است.

در این رویکرد، یک الگوریتم داده‌ها را اسکن می‌کند تا ویژگی‌های مرتبط را شناسایی کند و سپس آن‌ها را برای ارتقای سریع‌تر یادگیری، بدون اینکه به طور صریح به او گفته شود، ترکیب می‌کند. این توانایی به دانشمندان داده کمک می‌کند تا مقدار قابل توجهی در زمان صرفه‌جویی کرده و به دنبال آن نتایج بهتری را نیز بدست آورند.

▪ **ارائه نتایج با کیفیت بالا:** انسان‌ها گرسنه یا خسته می‌شوند و گاهی اوقات اشتباه می‌کنند. در مقابل، وقتی صحبت از شبکه‌های عصبی می‌شود، اینطور نیست. هنگامی که یک مدل یادگیری عمیق بدرستی آموزش داده شود، می‌تواند هزاران کار معمولی و تکراری را در مدت زمان نسبتا کوتاه‌تری در مقایسه با آنچه که برای یک انسان لازم است، انجام دهد. علاوه بر این، کیفیت کار هرگز کاهش نمی‌یابد، مگر اینکه داده‌های آموزشی حاوی داده‌های خامی باشد که نشان‌دهنده مساله‌ای نیست که می‌خواهید آن راحل کنید.

با در نظر گرفتن مزایای فوق و استفاده بیشتر از رویکرد یادگیری عمیق، می‌توان گفت که تاثیر قابل توجه یادگیری عمیق در فناوری‌های مختلفِ پیشرفته همانند اینترنت اشیا در آینده بدیهی است. یادگیری عمیق، راهِ درازی را طی کرده است و به سرعت در حال تبدیل شدن به یک فناوری حیاتی است که به‌طور پیوسته توسط مجموعه‌ای از کسب‌وکارها، در صنایع مختلف مورد استفاده قرار می‌گیرد.

## معایب و چالش‌های یادگیری عمیق

اگرچه اهمیت و پیشرفت‌های یادگیری عمیق در حال افزایش است، اما چند جنبه منفی یا چالش وجود دارد که برای توسعه یک مدل یادگیری عمیق باید با آن‌ها مقابله کرد. برای شروع آموزش الگوریتم یادگیری عمیق، مقادیر زیادی داده مورد نیاز است. به عنوان مثال، برای یک برنامه تشخیص گفتار، داده‌هایی با گویش‌های متعدد، جمعیت‌شناسی و داده‌هایی با مقیاس‌های زمانی برای بدست آوردن نتایج مطلوب مورد نیاز است. در حالی که شرکت‌هایی مانند گوگل و مایکروسافت قادر به جمع‌آوری و داشتن داده‌های فراوان هستند، شرکت‌های کوچک با ایده‌های خوب ممکن است نتوانند این کار را انجام دهند. همچنین، ممکن است گاهی اوقات، داده‌های لازم برای آموزش یک مدل از قبل پراکنده یا در دسترس نباشد.

اگرچه مدل‌های یادگیری عمیق بسیار کارآمد هستند و می‌توانند یک راه‌حل مناسب برای یک مشکل خاص پس از آموزش با داده‌ها فرموله کنند، اما برای یک مساله مشابه قادر به انجام این کار نیستند و نیاز به آموزش مجدد دارند. برای نشان دادن این موضوع، یک الگوریتم یادگیری عمیق را در نظر بگیرید که یاد می‌گیرد اتوبوس‌های مدرسه همیشه زرد هستند، اما ناگهان اتوبوس‌های مدرسه آبی می‌شوند. از این‌رو، باید دوباره آموزش داده شود. برعکس، یک کودک پنج ساله مشکلی برای تشخیص وسیله نقلیه به عنوان یک اتوبوس مدرسه آبی ندارد. علاوه بر

این، آن‌ها همچنین در موقعیت‌هایی که ممکن است کمی متفاوت با محیطی باشد که با آن تمرین کرده‌اند، عملکرد مؤثری ندارند. برای مثال DeepMind گوگل سیستمی را برای شکست ۴۹ بازی آتاری آموزش داد. با این حال، هر بار که سیستم یک بازی را شکست می‌داد، باید برای شکست دادن بازی بعدی دوباره آموزش داده می‌شد. این ما را به محدودیت دیگری در یادگیری عمیق می‌رساند، یعنی در حالی که مدل ممکن است در نگاشت ورودی‌ها به خروجی‌ها فوق‌العاده خوب باشد، اما ممکن است در درک زمینه داده‌هایی که آن‌ها مدیریت می‌کنند خوب نباشد.

در نهایت، شناخته‌شده‌ترین نقطه ضعف شبکه‌های عصبی ماهیت "جعبه سیاه" آن‌هاست. به عبارت ساده، شما نمی‌دانید چگونه و یا چرا شبکه عصبی شما خروجی خاصی را بدست آورده است. به عنوان مثال، وقتی تصویری از یک گربه را به یک شبکه عصبی تغذیه می‌کنید و آن را یک ماشین پیش‌بینی می‌کند، درک اینکه چه چیزی باعث شده است که به این پیش‌بینی برسد بسیار سخت است. این سناریو در تصمیمات تجاری مهم خواهند بود. آیا می‌توانید تصور کنید که مدیر عامل یک شرکت بزرگ تصمیمی در مورد میلیون‌ها دلار بگیرد بدون اینکه بفهمد چرا باید این کار را انجام دهد؟ فقط به این دلیل که "رایانه" می‌گوید او باید این کار را انجام دهد؟ در مقایسه، الگوریتم‌هایی مانند درخت تصمیم بسیار قابل تفسیر هستند.

روی هم رفته، به گفته اندرو انگ، یادگیری عمیق راهی عالی برای "ساخت جامعه‌ای مبتنی بر هوش مصنوعی" است و غلبه بر این کاستی‌ها با کمک سایر فن‌آوری‌ها، راه درست برای رسیدن به این هدف است.

## آیا با وجود یادگیری عمیق، از یادگیری ماشین استفاده نکنیم؟

پاسخ منفی است. این به این دلیل است که یادگیری عمیق از نظر محاسباتی می‌تواند بسیار گران باشد. اگر بتوان مساله‌ای را با استفاده از یک الگوریتم یادگیری ماشین ساده‌تر طی حل کرد، یعنی الگوریتمی که نیازی به دست‌وپنجه نرم‌کردن با ترکیب پیچیده‌ای از ویژگی‌های سلسله‌مراتبی در داده‌ها نداشته باشد، آنگاه گزینه‌های محاسباتی کم‌تر، انتخاب بهتری است. از این‌رو اگر به نتایج سریع‌تری نیاز دارید، الگوریتم‌های یادگیری ماشین ممکن است مطلوب‌تر باشند. آموزش آن‌ها سریع‌تر است و به توان محاسباتی کم‌تری نیاز دارند.

یادگیری عمیق همچنین ممکن است بهترین انتخاب برای پیش‌بینی بر اساس داده‌ها نباشد. به عنوان مثال، اگر مجموعه داده کوچک باشد، گاهی اوقات مدل‌های یادگیری ماشینِ خطیِ ساده‌تر ممکن است نتایج دقیق‌تری به همراه داشته باشند. هرچند، برخی از متخصصان یادگیری ماشین استدلال می‌کنند که یک شبکه‌ی عصبیِ عمیقِ آموزش‌دیده‌ی مناسب، همچنان می‌تواند با مقادیرِ کمِ داده، عملکرد خوبی داشته باشد.

بر این اساس، چه زمانی باید از یادگیری ماشین یا یادگیری عمیق استفاده کرد؟ این به نیازهای شما بستگی دارد:

- **آیا به یک مدل بسیار دقیق نیاز دارید؟** از یادگیری عمیق استفاده کنید.
- **آیا نیاز به یک مدل سبک دارید؟** از یادگیری ماشین استفاده کنید.
- **آیا روی یک مساله‌ی بینایی رایانه کار می‌کنید؟** شبکه‌های یادگیری عمیق همانند شبکه‌های پیچشی در حال حاضر در این سناریو استاندارد هستند.
- **در مورد نیازهای خود مطمئن نیستید؟** سعی کنید ابتدا مساله را با یادگیری ماشین حل کنید، اگر نتایج به اندازه کافی رضایت‌بخش نبود، یادگیری عمیق را امتحان کنید.

## شبکه‌های عصبی

### نورون مصنوعی

نورون‌های مصنوعی بلوک‌های اصلی شبکه‌های عصبی مصنوعی هستند. آن‌ها یک مدل ریاضی ساده را توصیف می‌کنند که از نورون‌های مغز الهام گرفته شده است. عملکرد اساسی یک نورون مصنوعی دریافت ورودی‌های متعدد $x_1, \ldots, x_n$ و محاسبه مجموع وزنی $z$ برای این ورودی‌ها با استفاده از وزن‌های $w_1, \ldots, w_n$ است. این مجموع وزنی $z$ یک تبدیل خطی ورودی‌های نورون است. علاوه بر این، بایاس $b$ به مجموع وزنی ورودی‌ها اضافه می‌شود و نتیجه از یک تابع فعال‌سازی $\varphi$ عبور داده می‌شود و در نتیجه خروجی نهایی $\hat{y}$ ایجاد می‌شود:

$$\hat{y} = \varphi(b + \sum_{i=1}^{n} w_i \times x_i)$$

شکل ۷ ــ ۱ یک بلوک دیاگرام سطح بالاست که رابطه بین بردار ورودی و متغیر خروجی را نشان می‌دهد. مقادیر پارامترهای $b$ و $w$ ناشناخته هستند. از این رو، در ابتدای فرآیند یادگیری، این مقادیر به صورت دلخواه تنظیم می‌شوند. سپس سیستم یادگیری ماشین آن‌ها را با استفاده از الگوریتم‌های بهینه‌سازی بهینه می‌کند. از این‌رو، می‌توان نورون مصنوعی را برای یک تقریب یک تابع با توجه به ورودی های $x$ آموزش داد.

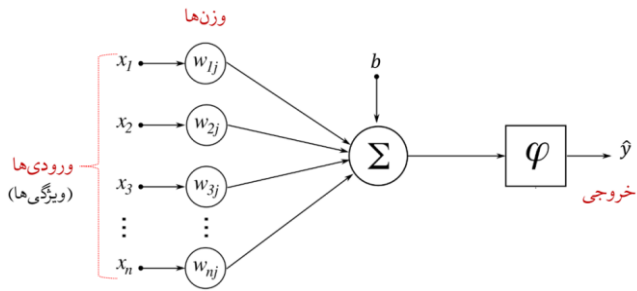

**شکل ۷ ــ ۱.** نورون مصنوعی

اگرچه این یک مدل بسیار ساده است، اما نورون مصنوعی قابلیت استفاده برای مسائل مختلف را دارد. اگر تابع پله‌ای[۱] به عنوان تابع فعال‌سازی انتخاب شود، نورون می‌تواند برای دسته‌بندی دودویی استفاده شود. این نوع نورون‌هایی که برای دسته‌بندی دودویی استفاده می‌شوند **پرسپترون** نامیده می‌شوند. با این حال، الگوریتم پرسپترون تنها زمانی می‌تواند ورودی خود را به درستی دسته‌بندی کند که داده‌ها به صورت خطی بر اساس کلاس قابل تفکیک باشند. برای رفع این مشکل، "پرسپترون چندلایه" پیشنهاد شد، که می‌توان آن را به‌عنوان دنباله‌ای از پرسپترون‌های سازمان‌دهی‌شده در "لایه‌ها" مشاهده کرد که هر کدام ورودی خود را از پرسپترون قبلی می‌گیرند

## شبکه‌ی عصبی پیش‌خور

شبکه عصبی پیش‌خور، مشابه با نورون مصنوعی است که قبلا مورد بحث قرار گرفت. هدف آن نیز تقریب تابع برای ورودی‌های $x$ است. با این حال، به جای محدود شدن به توابع بسیار ساده که فقط از یک مجموع وزن‌دار با یک تابع فعال‌ساز تشکیل شده است، شبکه‌های عصبی پیش‌خور چندین نورون را ترکیب می‌کنند تا یک گراف جهت‌دار بدون دور تشکیل دهند. نمونه‌ای از یک شبکه عصبی پیش‌خور در شکل ۷ـ۲ نشان داده شده است. هر شبکه عصبی پیش‌خور شامل **ورودی‌ها** (که معمولا به عنوان **لایه ورودی** نامیده می‌شود)، تعداد دلخواه لایه‌های میانی نورون‌ها به نام **لایه‌های پنهان**[۲] و لایه‌ای که خروجی‌ها را محاسبه می‌کند به نام **لایه خروجی**[۳] تشکیل می‌شود. این رویکرد مبتنی‌بر لایه، جایی است که نام یادگیری عمیق از آن گرفته شده است، چراکه، عمق یک شبکه عصبی پیش‌خور، تعداد لایه‌هایی را که یک شبکه عصبی پیش‌خور از آن تشکیل شده است را توصیف می‌کند. هنگامی که هر نورون در یک لایه به تمام نورون‌های لایه بعدی متصل می‌شود، به آن **شبکه کاملا متصل**[۴] می‌گویند، به لایه‌هایی که این رفتار را نشان می‌دهند، **لایه‌های کاملا متصل** می‌گویند. شبکه عصبی پیش‌خورِ ساده‌ی نشان داده شده در شکل ۲ـ۷، در حال حاضر بسیار بسیار قدرتمندتر از یک نورون مصنوعی است. می‌توان نشان داد که شبکه‌های عصبیِ پیش‌خور تنها با یک لایه‌ی پنهان، می‌توانند برای تقریب هر تابع پیوسته مورد استفاده قرار گیرند. شبکه‌های عصبی پیش‌خور، پایه و اساس اکثر برنامه‌های کاربردی

---

[۱] step function

[۲] hidden layers

[۳] output layer

[۴] fully-connected network

یادگیری عمیق را نشان می‌دهند. به عنوان مثال، **شبکه‌های عصبی پیچشی**[۱] بسیار محبوب و موفق، صرفا توسعه‌ای برای شبکه‌های عصبیِ پیش‌خورِ استاندارد هستند.

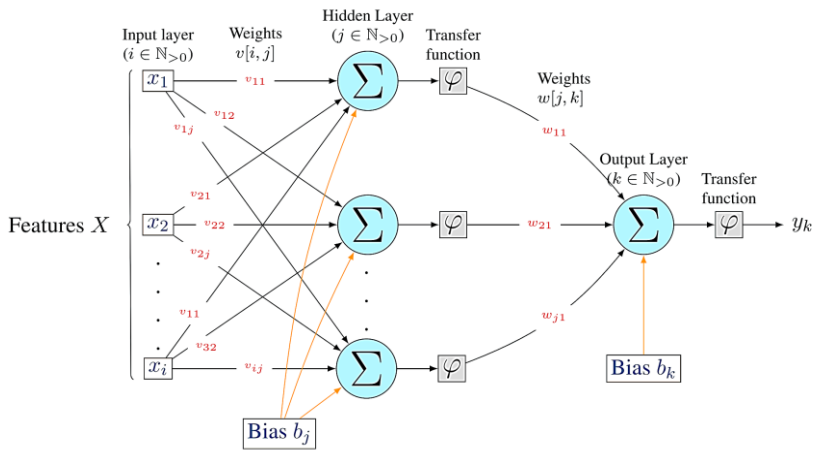

**شکل ۷ـ۲.** معماری ساده‌ی یک شبکه عصبی پیش‌خور

## بهینه‌سازی شبکه‌های عصبی

هدف از بهینه‌سازی شبکه‌های عصبی پیش‌خور، یافتن خودکار وزن‌ها و بایاس‌هایی است که شبکه خروجی هدف مورد نظر $y$ را با ورودی $x$ تقریب می‌زند. به منظور دستیابی به این هدف، لازم است یک معیار برای اینکه شبکه عصبی پیش‌خور چقدر به خروجی تخمین می‌زند، تعریف شود. این معیار معمولا به عنوان **تابع ضرر**[۲] یا **تابع هزینه**[۳] $J(\theta)$ نامیده می‌شود، که در آن $\theta$ پارامترهای شبکه (وزن‌ها و بایاس‌ها) را توصیف می‌کند. با توجه به مجموعه‌ای از $N$ نمونه‌ی آموزشی $J(\theta)$، $y = [y_1, y_2, ..., y_N]$ و برچسب مربوط $x_T = [x_{T_1}, x_{T_2}, ..., x_{T_N}]$ معمولا به عنوان میانگینِ تابعِ ضررِ هر نمونه محاسبه می‌شود و $L(\hat{y}(x_{T_i}; \theta), y_i)$، که در آن $\hat{y}(x_{T_i}; \theta)$ خروجی شبکه‌ی عصبی پیش‌خور، با توجه به نمونه آموزشی $x_{T_i}$ به عنوان ورودی و پارامترهای شبکه $\theta$ است:

$$J(\theta) = \frac{1}{N} \sum_{i=1}^{N} L(\hat{y}(x_{T_i}; \theta), y_i)$$

---

[۱] Convolutional Neural Network

[۲] loss function

[۳] cost function

یک مقدار تابع زیان کوچکتر، معمولا برابر با یک تقریب تابع بهتر شبکه‌ی عصبی پیش‌خور است، بنابراین روش آموزش شبکه‌های عصبی پیش‌خور را می‌توان به عنوان یک مساله بهینه‌سازی فرموله کرد، جایی که هدف کمینه کردن تابع زیان $J(\theta)$ با توجه به پارامترهای شبکه $\theta$ است. این کار معمولا با استفاده از یک نوع الگوریتم با عنوان گرادیان کاهشی انجام می‌شود.

## گرادیان کاهشی[1]

با توجه به یک تابع زیان با مقادیر حقیقی $J(\theta)$، هدف از گرادیان کاهشی یافتن کمینه محلی $J(\theta)$ با توجه به پارامترهای $\theta$ است. در حالی که برای توابع زیان ساده، ممکن است محاسبه‌ی حداقل آن به صورت تحلیلی[2] امکان‌پذیر باشد، برای توابع پیچیده تر با پارامترهای متعدد، مانند توابع زیان شبکه‌های عصبی پیش‌خور با میلیون‌ها پارامتر، این کار غیرممکن می شود. برخلاف محاسبات تحلیلی کمینه‌ها، گرادیان کاهشی یک رویکرد عددی است که با انتخاب پارامترهای تصادفی شروع می‌شود و به‌طور مکرر برروی جهت منفی گرادیان تابع حرکت می‌کند تا کمینه محلی را پیدا کند. برای یک نمونه آموزشی منفرد $x_{T_i}$ با خروجی هدف مربوط $y_i$، گرادیان کاهشی به‌صورت زیر محاسبه می‌شود:

$$-g_\theta = -\nabla_\theta L\big(\hat{y}(x_{T_i}; \theta), y_i\big)$$

سپس گرادیان نهایی برای تابع زیان $J(\theta)$ با محاسبه میانگین همه گرادیان‌ها در کل مجموعه آموزشی $x_T$ بدست می‌آید:

$$-g_\theta = -\frac{1}{N}\sum_{i=1}^{N} -g_{\theta_i}$$

با تعریف یک عامل مثبت کنترل‌کننده‌ی بزرگی گرادیان کاهشی، به نام نرخ یادگیری[3] $\eta$، قانون بروزرسانی گرادیان کاهشی پارامترهای شبکه‌ی عصبی پیش‌خور $\theta$ را می‌توان به صورت تعریف کرد:

$$\theta = \theta + \eta - g_\theta$$

بسته به مقدار اولیه پارامترها، این امکان برای گرادیان کاهشی وجود دارد که کمینه‌ی سراسری $J(\theta)$ را پیدا نکند، *اما این تضمین نمی‌شود، مگر اینکه $J(\theta)$ محدب باشد*. نکته مهم این است که برای همگرا شدن گرادیان کاهشی، تابع زیان باید **هموار**[4] باشد و در همه‌جا گرادیان ارائه کند.

---

[1] gradient descent

[2] analytically

[3] learning rate

[4] smooth

به همین دلیل است که برخی مواقع تابع زیان انتخابی، معمولا با هدف واقعی متفاوت است و به جای آن از یک **تابع زیان جایگزین**[1] استفاده می‌شود. به عنوان مثال، به‌جای بهینه‌سازی تعداد نمونه‌های دسته‌بندی درست در یک مساله دسته‌بندی تصویر، می‌توانیم اطمینان پیش‌بینی‌شده و اطمینان هدف هر کلاس را بر روی خطای میانگین مربعات (MSE) بهینه کنیم. در حالی که تعداد نمونه‌هایی که به درستی دسته‌بندی شده‌اند گسسته و در نتیجه غیرهموار خواهد بود، MSE در همه جا هموار خواهد بود و گرادیان‌های مفیدی برای گرادیان کاهشی فراهم می‌کند. به عنوان مثال دیگر، لگاریتم‌ـ درست‌نمایی منفیِ کلاسِ درست، معمولا به عنوان جایگزینی برای زیان ۰ـ۱ استفاده می‌شود. لگاریتم‌ـ درست‌نمایی منفی، به مدل اجازه می‌دهد تا احتمال شرطی کلاس‌ها را با توجه به ورودی برآورد کند و اگر مدل بتواند این کار را به خوبی انجام دهد، می‌تواند کلاس‌هایی را انتخاب کند که کم‌ترین خطای دسته‌بندی را در امید ریاضی نتیجه دهد.

## محاسبه کارآمد گرادیان با پس‌انتشار

الگوریتم پس انتشار احتمالا اساسی‌ترین بلوک سازنده در یک شبکه عصبی است. پس‌انتشار اساسا تدبیر هوشمندانه‌ای برای محاسبه مؤثر گرادیان‌ها در شبکه‌های عصبی چندلایه است. به عبارت دیگر، پس‌انتشار در مورد محاسبه گرادیان برای توابع تو در تو است که به عنوان یک گراف محاسباتی با استفاده از قاعده‌ی زنجیره‌ای نمایش داده می‌شود. به عبارت ساده، پس از هر گذرِ **پیش‌رو** از یک شبکه، پس‌انتشار یک گذر **پس‌رو** انجام می‌دهد و در عین حال پارامترهای مدل (وزن ها و بایاس‌ها) را تنظیم می‌کند.

قبل از تشریح الگوریتم، به توصیفِ یک نماد مبتنی‌بر ماتریس، برای درک بهتر می‌پردازیم. وزن $w_{jk}^l$ وزنی را توصیف می‌کند که نورون $k$ام را در لایه $(l-1)$ به نورون $j$ام لایه $l$ام متصل می‌کند. از این وزن‌ها می‌توان یک ماتریس $w^l$ تشکیل داد که به آن **ماتریس وزن** برای لایه $l$ام می‌گویند که ورودی ردیف $j$ام و ستون k‌ام برابر با $w_{jk}^l$ است. با توجه به K نورون در لایه $(l-1)$ و J نورون در لایه $l$، این ماتریس به صورت زیر است:

$$w^l = \begin{bmatrix} w_{\backslash\backslash}^l & \cdots & w_{\backslash K}^l \\ \vdots & \ddots & \vdots \\ w_{J1}^l & \cdots & w_{JK}^l \end{bmatrix}$$

---

بهطور مشابه، بایاسهای $b_j^l$، خروجیهای **پیشفعالسازی**[1] $z_j^l$ و خروجیهای $y_j^l$ نورون $j$ام در لایه $l$ را میتوان به ترتیب به صورت $b^l$، $z^l$ و $y^l$ بردارسازی کرد:

$$b^l = \begin{bmatrix} b_1^l \\ \vdots \\ b_j^l \end{bmatrix}, \qquad z^l = \begin{bmatrix} z_1^l = \sum_k w_{1k}^l y_k^{(l-1)} + b_j^l \\ \vdots \\ z_j^l = \sum_k w_{jk}^l y_k^{(l-1)} + b_j^l \end{bmatrix}, \qquad y^l = \begin{bmatrix} y_1^l = \varphi(z_1^l) \\ \vdots \\ y_j^l = \varphi(z_j^l) \end{bmatrix}$$

به عنوان اولین مرحله از الگوریتم پسانتشار، نتایج **انتشار پیشرو** محاسبه میشود. در طی این مرحله، شبکه عصبی پیشخور، یک نمونه آموزشی منفرد $x_{T_i}$ را به عنوان ورودی دریافت میکند و خروجیهای هر لایه $y^l$ و همچنین خروجیهای پیشفعالسازی هر لایه $z^l$ را تا لایه نهایی $D$ محاسبه میکند. علاوه بر این، با توجه به خروجی نهایی شبکهی عصبی پیشخور نهایی $y^D$، و خروجی هدف $y_i$ برای این نمونه آموزشی، زیان هر نمونه $L(y^D, y_i)$ محاسبه میشود. به این امر انتشار پیشرو گفته میشود، زیرا خروجیهای شبکه به صورت تکراری برای هر لایه، از لایه ورودی تا لایه خروجی، محاسبه میشوند. پس از آن، **انتشار پسرو** با محاسبه گرادیان $g_y^D$ در لایه خروجی $D$ با توجه به خروجی شبکهی عصبی پیشخور انجام میشود:

$$g_y^D = \nabla_{y^D} L(y^D, y_i)$$

از آنجایی که ما به گرادیان با توجه به خروجی پیشفعالسازی علاقهمندیم، آن را با انجام یک ضرب هادامارد[2] $\odot$ بین گرادیان **پسفعالسازی**[3] و مشتق تابع فعالسازی محاسبه میکنیم:

$$g_z^D = g_y^D \odot \varphi'(z^D)$$

سپس از این گرادیان پیشفعالسازی $g_z^D$ میتوان برای محاسبه گرادیانهای تابع هزینه $J(\theta)$ با توجه به ماتریس وزن $W^D$ و بردار بایاس $b^D$ استفاده کرد:

$$\nabla_{b^D} L(y^D, y_i) = g_z^D$$

$$\nabla_{w^D} L(y^D, y_i) = g_z^D (y^{(D-1)})^T$$

برای رسیدن به گرادیانهای پسفعالسازیِ لایهیِ پنهانِ سطحِ پایینِ بعدی $(D-1)$، ما به سادگی ماتریس ترانهاده $(W^D)^T$ را در گرادیان پیشفعالسازی لایهیِ فعلی $g_z^D$ ضرب میکنیم:

---

[1] pre-activation

[2] Hadamard product

[3] post-activation

$$g_y^{(D-1)} = \nabla_{z^{D-1}} L(y^D, y_i) = (W^D)^T g_z^D$$

مراحل بالا اکنون می‌توانند به طور مکرر انجام شوند تا زمانی که هر گرادیان منفرد، با توجه به هر وزن و بایاس در لایه اول مشخص شود، بنابراین منجر به گرادیان نهایی برای هر نمونه

$$g_{\theta_i} = \nabla_\theta L(\hat{y}(x_{T_i}; \theta), y_i)$$

از نمونه‌های آموزشی $x_{T_i}$ و خروجی هدف $y_i$ می‌شود. از این‌رو، با استفاده از الگوریتم پس‌انتشار، ما اکنون روشی برای محاسبه گرادیان برای یک نمونه آموزشی منفرد در معادله

$$g_{\theta_i} = \nabla_\theta L(\hat{y}(x_{T_i}; \theta), y_i)$$

داریم، که برای بروزرسانی پارامترهای شبکه‌ی عصبی پیش‌خور به دلیل گرادیان کاهشی لازم است.

با این حال، یک اشکال بزرگ در بهینه‌سازی شبکه‌های عصبی از طریق پس‌انتشار و گرادیان کاهشی، **مشکل محو گرادیان**[1] است. همان‌طور که در بالا توضیح داده شد، گرادیان برای یک وزن خاص با پس‌انتشار از خروجی به سمت نورون مربوط محاسبه می‌شود. در این مسیرِ پس‌انتشار، با استفاده از قاعده‌ی زنجیره‌ای حساب دیفرانسیل و انتگرال، گرادیان‌ها به طور مکرر از پشت به جلو ضرب می‌شوند. بنابراین، هنگامی که گرادیان‌ها کوچک هستند (که می‌تواند به عنوان مثال هنگام استفاده از توابع فعال‌سازی اشباع[2] مانند سیگموید رخ دهد)، می‌تواند منجر به گرادیان‌های کوچکی برای نورون‌ها در لایه‌های جلویی شبکه شود که منجر به یادگیری کند برای این نورون‌ها می‌شود.

در حالی که این یک مشکل در پس‌انتشار و گرادیان کاهشی است، روش‌های مدرن در یادگیری عمیق بیشتر این مشکل را برطرف می‌کنند. فعال‌ساز ReLU، بزرگی گرادیان را در ناحیه ورودی مثبت اشباع نمی‌کند. در نتیجه، محوشدگی گرادیان‌ها را در آن ناحیه تجربه نمی‌کند. علاوه‌بر این، نوع جدیدی از معماری شبکه به نام ResNet، نورون‌ها یا لایه‌هایی با گرادیان محوشونده را با به‌کارگیری مسیرهای میان‌بر دور می‌زند و به گرادیان‌ها اجازه می‌دهد بدون دورزدن در شبکه‌های بسیار عمیق‌تر، جریان پیدا کنند. همچنین می‌توان با یکسان‌سازی لایه‌های میانی با استفاده از **یکسان‌سازی دسته‌ای**[3]، توزیع ورودی‌ها را پایدارتر کرد، در نتیجه احتمالِ گیرکردن در حالت‌های اشباع کم‌تر می‌شود.

---

[1] vanishing gradient problem

[2] saturating

[3] batch normalization

## گرادیان کاهشی تصادفی

در حالی که گرادیان کاهشی استاندارد با استفاده از پس‌انتشار یک روش مفید برای آموزش خودکار شبکه‌های عصبی پیش‌خور است، زمانی که مجموعه آموزشی بسیار بزرگ است، گرادیان کاهشی به منابع محاسباتی زیادی نیاز دارد. چراکه تنها یک مرحله بروزرسانی، نیازمند محاسبهٔ همهٔ گرادیان‌ها برای همهٔ نمونه‌های آموزشی است.

هدف از گرادیان کاهشی تصادفی[1] (SGD) سرعت بخشیدن به فرآیند یادگیری با اندکی تغییر در رویه استاندارد گرادیان کاهشی است. تفاوت اصلی بین SGD و گرادیان کاهشی استاندارد در این است که SGD گرادیان تابع هزینه‌ی $\nabla_\theta j(\theta)$ را با محاسبه‌ی گرادیان‌های زیان هر نمونه فقط برای زیرمجموعه‌ای کوچک از $m$ نمونه‌ی آموزشی انتخابی تصادفی از مجموعه آموزشی $y_M = [y_{M_1}, \ldots, y_{M_m}] \subset y_T$ مربوط هدف خروجی با ،$x_M = [x_{M_1}, \ldots, x_{M_m}] \subset x_T$ برآورد می‌کند. این زیرمجموعه از نمونه‌های آموزشی، **ریزدسته**[2] نامیده می‌شود که $m$ تعداد نمونه‌های این ریزدسته است که به آن **اندازه ریزدسته** می‌گویند. برآورد گرادیان با استفاده از ریزدسته‌ها را می‌توان به صورت خلاصه کرد:

$$\nabla_\theta j(\theta) = \nabla_\theta \frac{1}{N} \sum_{i=1}^{N} L\big(\hat{y}(x_{T_i}; \theta), y_i\big) \approx \nabla_\theta \frac{1}{m} \sum_{i=1}^{m} L\big(\hat{y}(x_{M_i}; \theta), y_{M_i}\big)$$

## گرادیان کاهشی با تکانه[3]

همان‌طور که پیش‌تر بیان شد، از گرادیان کاهشی می‌توان برای یافتن کمینه‌ی محلی یک تابع استفاده کرد. با این حال، بسته به شکل تابع، رویکرد تکراری گرادیان کاهشی اغلب، منجر به تعداد زیادی مرحله می‌شود؛ به خصوص برای توابعی که شامل بسیاری از مناطق تقریبا مسطح با شیب‌های کوچک هستند، گرادیان کاهشی کند است. برای حل این موضوع، **تکانه** به الگوریتم گرادیان کاهشی اضافه می‌شود. ایده اصلی تکانه، اضافه کردن حافظه کوتاه مدت به گرادیان کاهشی است که گاهی اوقات **تندی**[4] نیز نامیده می‌شود. بر این اساس، مرحله بروزرسانی وزن به‌صورت زیر تغییر پیدا می‌کند:

$$g_\beta = \beta g_\beta + \frac{1}{N} \sum_{i=1}^{N} g_{\theta_i}$$

---

[1] Stochastic Gradient Descent

[2] minibatch

[3] Momentum

[4] acceleration

$$\theta = \theta - \eta g_\beta$$

جایی که $g_\beta$ اولیه صفر است، $\eta$ نرخ یادگیری و $\beta$ به عنوان ضابطه‌ی تکانه شناخته می‌شود.

## بهینه‌سازی نرخ یادگیری تطبیقی

هرچند SGD یک روش بهینه‌سازی بسیار قدرتمند برای آموزش شبکه‌های عصبی پیش‌خور است، با این حال، انتخابِ بهترین نرخ یادگیری $\eta$ برای هر مساله‌ای اهمیت زیادی دارد. اگر $\eta$ خیلی بزرگ انتخاب شود، آموزش ممکن است نوسان کند، همگرا نشود یا از کمینه‌های محلی مربوط بگذرد. در مقابل، اگر نرخ یادگیری خیلی کوچک انتخاب شود، به طور قابل توجهی فرآیند همگرایی را به تاخیر می‌اندازد. از این‌رو، یک تکنیک رایج برای دورزدن این مساله استفاده از **نرخ واپاشی**[1] یادگیری است. به عنوان مثال، با استفاده از واپاشی گامی، می‌توان نرخ یادگیری را هر چند دوره به میزانی کاهش داد. این امر این امکان را می‌دهد تا میزان یادگیری زیادی در ابتدای آموزش و نرخ یادگیری کمتری در پایان آموزش وجود داشته باشد. با این حال، این روش واپاشی نیز، به خودی خود یک ابرپارامتر است و بسته به کاربرد باید با دقت طراحی شود.

هدف بهینه‌سازهای نرخ یادگیری تطبیقی، حل مشکل یافتن نرخ یادگیری درست است. در این روش‌ها، نرخ یادگیری $\eta$ یک متغیر سراسری نیست، اما در عوض هر پارامترِ قابل آموزش، نرخِ یادگیری جداگانه‌ای برای خود دارد. در حالی که این روش‌ها اغلب هنوز نیاز به تنظیم ابرپارامتر دارند، بحث اصلی این است که آن‌ها برای طیف وسیع‌تری از پیکربندی‌ها به خوبی کار می‌کنند؛ اغلب زمانی که تنها از ابرپارامترهای پیش‌فرضِ پیشنهادی استفاده می‌کنند.

## تدابیر

همراه با تکنیک گرادیان کاهشی، پس‌انتشار روشی موثر را برای بهینه‌سازی پارامترهای آموزشی شبکه عصبی برای کمینه کردن یک تابع زیان داده‌شده ارائه می‌دهد. با این حال، آموزش شبکه‌های عصبی در عمل توسط عوامل مختلفی دشوار می‌شود. اول، ترکیب بسیاری از لایه‌های پردازش غیرخطی در یک شبکه عصبی، تابعِ زیان حاصل را بسیار غیرمحدب می‌کند. غیرمحدب به این معنی است که هنگام کمینه کردن تابع زیانِ شبکه با روش‌های گرادیان کاهشی، هیچ تضمینی وجود ندارد که یک نقطه‌ی ثابتِ معین، یک کمینه سراسری باشد. همچنین، روشی که پارامترها قبل از بهینه‌سازی مقداردهی اولیه می‌شوند و هم تکنیک گرادیان کاهشی مورد استفاده، ممکن است تاثیر بسیار قوی بر بهترین راه‌حلِ پیدا شده‌ی در طول بهینه‌سازی داشته باشد. همچنین،

---

[1] rate decay

شبکه‌های عصبی پیش‌خور با لایه‌های زیاد نیز مستعد مشکل محو و انفجار گرادیان‌ها هستند. از این‌رو، نیاز به تدابیری است تا از این مشکلات جلوگیری شود.

## وزن‌دهی اولیه

در شبکه‌های عصبی، مقداردهی اولیهٔ وزن‌ها باید با دقت بسیاری انتخاب شود. برای مثال، اگر چندین نورون در یک لایهٔ پنهان، وزن‌های مشابهی داشته باشند، گرادیان‌های یکسانی را دریافت خواهند کرد. از این‌رو نتایج یکسانی را محاسبه می‌کنند که منجر به هدر رفتن ظرفیت مدل می‌شود. به طور معمول، وزن‌های شبکه‌های عصبی با استفاده از یک توزیع گاوسی با میانگین صفر و یک انحراف معیار کوچک مقداردهی اولیه می‌شوند. با این حال، مشکلی که وجود دارد این است که توزیع خروجی‌های یک نورونِ به‌طور تصادفی مقداردهی اولیه‌شده، دارای واریانسی است که با تعداد ورودی‌ها افزایش می‌یابد. برای نرمال کردن واریانس خروجی هر نورون به ۱، کافی است از یک توزیع نرمال استاندارد استفاده کنید و وزن را بر اساس جذر گنجایش ورودی[1] $n_{in}$، که تعداد ورودی‌های آن است، مقیاس کنید:

$$w_0 \sim \frac{\mathcal{N}(0,1)}{\sqrt{n_{in}}}$$

به طور مشابه، گلورت و بنجیو تجزیه و تحلیلی را برروی گرادیان‌های پس‌انتشار انجام دادند و یک مقداردهی اولیه (معروف به مقداردهی اولیه خاویر یا گلورت) را توصیه کردند:

$$w_0 \sim \sqrt{\frac{2}{n_{in} + n_{out}}} \mathcal{N}(0,1)$$

جایی که $n_{out}$ تعداد واحدهای خروجی را توصیف می‌کند. به طور خاص برای نورون‌های با فعال‌سازی ReLU، مقداردهی هی ارائه گردید:

$$w_0 \sim \sqrt{\frac{2}{n_{in}}} \mathcal{N}(0,1)$$

## منظم‌سازی

تا اینجا ما فقط به آموزش یک شبکه‌ی عصبی پیش‌خور با گرادیان کاهشی و پس‌انتشار با استفاده از مجموعه آموزشی $x_T$ با برچسب‌های مربوط $y$ اشاره کردیم. در حالی که از این طریق می‌توانیم شبکه‌ی عصبی خود را برای پیش‌بینی خروجی‌های مجموعه آموزشی، آموزش دهیم، لزوما به این معنی نیست که قادر به پیش‌بینی درست خروجی برای داده‌های دیده‌نشده هم باشد. بنابراین، همان‌طور که در فصل ۵ به آن اشاره شد، دو مجموعه اضافی از داده‌ها برای بهینه‌سازی معرفی

---

[1] fan-in

می‌شوند، مجموعه اعتبارسنجی و مجموعه آزمایشی. هر سه مجموعه داده از هم مستقل هستند، به طوری‌که هیچ نمونه‌ای در بین آن‌ها مشترک نیست.

مجموعه اعتبارسنجی در شبکه‌های عصبی، معمولا برای تنظیم دقیق ابرپارامترهای مدل مانند معماری شبکه یا نرخ یادگیری استفاده می‌شود. مجموعه آزمون فقط برای ارزیابی نهایی در راستای بررسی عملکرد شبکه در داده‌های دیده‌نشده استفاده می‌شود. اگر یک شبکه‌ی عصبی به خوبی تعمیم نیابد، یعنی زیانِ آموزشِ کمتری نسبت به زیان آزمون داشته باشد، همچنان که پیش‌تر اشاره شده، به این حالت بیش‌برازش گفته می‌شود. در حالی‌که در سناریوی معکوس، زمانی که زیان آزمون نسبت به زیان آموزش بسیار کمتر باشد، کم‌برازش نامیده می‌شود (شکل ۷ـ۳). به‌طور معمول، بیش‌برازش و کم‌برازش در شبکه‌های عصبی عمیق، مستقیما با ظرفیت مدل مرتبط است. به زبان ساده، ظرفیت مدلِ یک شبکه‌ی عصبیِ عمیق، به‌طور مستقیم با تعداد پارامترهای داخل شبکه در ارتباط است. ظرفیت مدل تعیین می‌کند که یک شبکه عمیق تا چه حد قادر به برازش با طیف گسترده‌ای از توابع است. اگر ظرفیت خیلی کم باشد، شبکه ممکن است نتواند مجموعه آموزشی را تطبیق دهد (کم‌برازش)، در حالی‌که ظرفیت مدل خیلی بزرگ ممکن است منجر به حفظ نمونه‌های آموزشی (بیش‌برازش) شود. *کم‌برازش معمولاً برای شبکه‌های عصبیِ عمیق، مشکل چندانی ندارد. چراکه این مشکل را می‌توان با استفاده از معماری شبکه‌ی قوی‌تر یا عمیق‌تر با پارامترهای بیشتر برطرف کرد.* با این حال، برای اینکه بتوان از شبکه‌های عمیق برای داده‌های جدید و دیده‌نشده استفاده کرد، باید بیش‌برازش را کنترل کرد. فرآیند کاهش اثر بیش‌برازش یا جلوگیری از آن (نه به‌طور کامل. چرا؟[۱]) را **منظم‌سازی**[۲] می‌گویند. در این بخش، به‌طور خلاصه محبوب‌ترین تکنیک‌های منظم‌سازی برای شبکه‌های عمیق را شرح می‌دهیم.

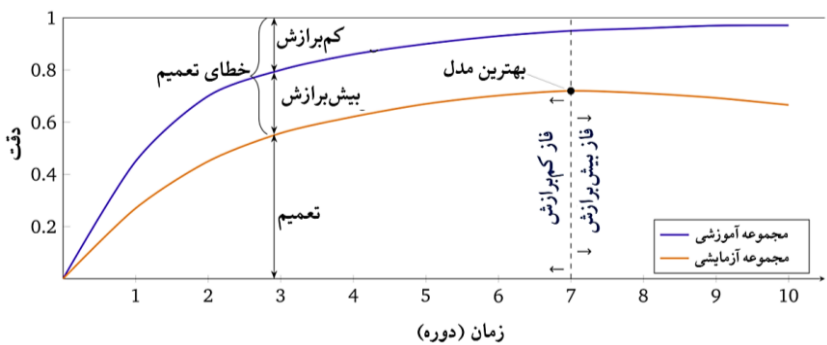

**شکل ۷ـ۳.** رفتار تعمیم‌دهی در منحنی یادگیری با توجه به معیار دقت در داده‌های آموزشی و آزمون

---

**توقف زودهنگام**

زمانی که ظرفیت مدل یک شبکه‌ی عمیق به اندازه‌ی کافی بزرگ باشد که قادر به بیش‌برازش باشد، معمولا مشاهده می‌شود که زیان آموزشی تا زمان همگرایی به‌طور پیوسته کاهش می‌یابد، در حالی که زیان اعتبارسنجی در شروع کاهش یافته و پس از مدتی دوباره افزایش می‌یابد. هدف توقف زودهنگام، منظم‌سازی شبکه‌ی عمیق با یافتن پارامترهای شبکه در نقطه‌ای با کم‌ترین زیان اعتبارسنجی است. با استفاده از پارامترهای شبکه با کم‌ترین زیان اعتبارسنجی، شبکه به‌طور بالقوه بهتر به داده‌های دیده‌نشده تعمیم می‌یابد. چراکه مدل در این مرحله واریانس پایینی دارد و به خوبی داده‌ها را تعمیم می‌دهد. آموزش بیشتر مدل، واریانس مدل را افزایش می‌دهد و منجر به بیش‌برازش می‌شود.

**حذف تصادفی[1]**

"حذف تصادفی" در شبکه‌های عصبی، به فرآیندِ نادیده‌گرفتنِ تصادفیِ گره‌های خاص در یک لایه در طول آموزشِ شبکه اشاره دارد. به عبارت دیگر، نورون‌هایِ مختلَف به‌طور موقت از شبکه

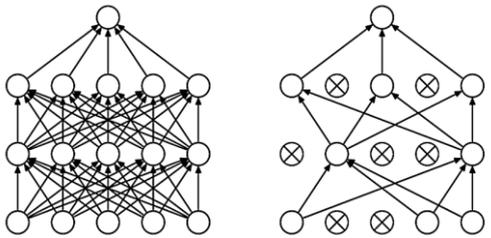

حذف می‌شوند. در طول آموزش، حذف تصادفی، ایده‌یِ یادگیری تمام وزن‌های شبکه را به یادگیری تنها کسری از وزن‌های شبکه تغییر می‌دهد. از شکل مقابل می‌توان دریافت که در مرحله‌ی آموزش استاندارد، همه‌یِ نورون‌ها درگیر هستند و با اعمال

(آ) شبکه‌ی عصبی استاندارد          (ب) شبکه‌ی عصبی با اعمال حذف تصادفی

حذف تصادفی، تنها چند نورون منتخب درگیر آموزش هستند و بقیه "خاموش" هستند. بنابراین پس از هر تکرار، مجموعه‌های مختلفی از نورون‌ها فعال می‌شوند تا از تسلط برخی نورون‌ها بر برخی ویژگی‌ها جلوگیری شود. این رویکرد در عین سادگی به ما کمک می‌کند تا بیش‌برازش را کاهش دهیم و امکان ایجاد معماری‌های شبکه عمیق‌تر و بزرگ‌تری را فراهم کنیم که می‌توانند پیش‌بینی‌های خوبی بر روی داده‌هایی انجام دهند که شبکه قبلا آن‌ها را ندیده است.

**یکسان‌سازی دسته‌ای**

یکی از مشکلاتی که در آموزش شبکه‌های عصبی علاوه بر محو گرادیان وجود دارد، مشکل تغییر متغیرهای داخلی شبکه است. این مشکل از آنجا ناشی می‌شود که پارامترها در طول فرآیند آموزش مدام تغییر می‌کند، این تغییرات به نوبه خود مقادیر توابع فعال‌سازی را تغییر می‌دهد. تغییر مقادیر ورودی از لایه‌های اولیه به لایه‌های بعدی سبب همگرایی کندتر در طول فرآیند آموزش می‌شود، چرا که داده‌های آموزشی لایه‌های بعدی پایدار نیستند. به عبارت دیگر،

---

[1] Dropout

شبکه‌های عمیق ترکیبی از چندین لایه با توابع مختلف بوده و هر لایه فقط یادگیری بازنمایی کلی را از ابتدای آموزش را فرا نمی‌گیرد، بلکه باید با تغییر مداوم در توزیع‌های ورودی با توجه به لایه‌های قبلی تسلط پیدا کند. حال آن‌که بهینه‌ساز بر این فرض بروزرسانی پارامترها را انجام می‌دهد که در لایه‌های دیگر تغییر نکنند و تمام لایه‌ها را هم‌زمان بروز می‌کند، این عمل سبب نتایج ناخواسته‌ای هنگام ترکیب توابع مختلف خواهد شد.

**یکسان‌سازی دسته‌ای** در جهت غلبه بر این مشکل برای کاهش ناپایداری و بهبود شبکه ارائه شده است. در این روش، یکسان‌سازی برروی داده‌های ورودی یک لایه را به گونه‌ای انجام می‌دهد، که دارای میانگین صفر و انحراف معیار یک شوند. با قرار دادن یکسان‌سازی دسته‌ای بین لایه‌های پنهان و با ایجاد ویژگی واریانس مشترک، سبب کاهش تغییرات داخلی لایه‌های شبکه می‌شویم.

# شبکه‌های عصبی بازگشتی

شبکه‌های عصبی بازگشتی یا مکرر، نوعی شبکه‌ی عصبی مصنوعی هستند که برای تشخیص الگوها در توالی داده‌ها، مانند متن، ژنوم، دست‌خط، کلمات گفتاری، داده‌های سری زمان، بازارهای سهام و غیره طراحی شده‌اند. ایده‌ی پشت این شبکه‌های عصبی این است که به سلول‌ها اجازه می‌دهند تا از سلول‌های قبلی متصل به خود یاد بگیرند. می‌توان گفت که به نوعی، این سلول‌ها دارای "حافظه" هستند. از این‌رو، دانش پیچیده‌تری را از داده‌های ورودی می‌سازند.

شبکه‌های عصبی بازگشتی، نواقص شبکه‌های عصبی پیش‌خور را برطرف می‌کنند. چرا که شبکه‌های پیش‌خور تنها می‌توانند ورودی‌های با اندازه ثابت را بپذیرند و تنها خروجی‌هایی با اندازه ثابت تولید کنند و قادر به در نظر گرفتن ورودی‌های قبلی با همان ترتیب نیستند. با در نظر گرفتن ورودی‌های گذشته در توالی‌ها، شبکه‌های عصبی بازگشتی قادر به گرفتن وابستگی‌های زمانی هستند که شبکه‌ی عصبی پیش‌خور قادر به آن نیست.

شبکه‌های عصبی بازگشتی، دنباله‌ای را به عنوان ورودی می‌گیرند و برای هر مرحله زمانی شبکه عصبی را ارزیابی می‌کنند. این شبکه‌ها را می‌توان به عنوان یک شبکه عصبی در نظر گرفت که دارای یک حلقه است که به آن اجازه می‌دهد حالت را حفظ کند. هنگامی که ارزیابی می‌شود،

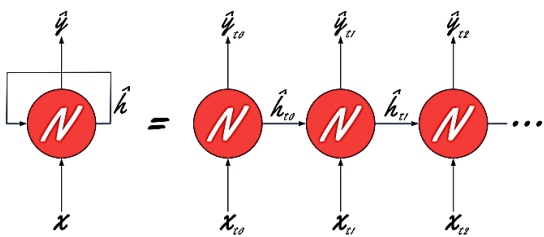

حلقه از طریق مراحل زمانی یک دنباله مانند شکل مقابل باز می‌شود. این حلقه‌ها یا پیوندهای مکرر دلیلی هستند که این شبکه‌ها را شبکه‌های بازگشتی می‌نامند. اینکه یک شبکه بازگشتی شامل حلقه است به این معنی است که خروجی یک نورون در یک نقطه زمانی ممکن است در نقطه زمانی دیگر به همان

نورون بازگردانده شود. نتیجه این امر این است که شبکه نسبت به فعال‌سازی‌های گذشته (و بنابراین ورودی‌های گذشته‌ای که در این فعال‌سازی نقش داشته‌اند) حافظه دارد.

ورودی برای هر مرحله زمانی شامل ویژگی‌های مرحله زمانی و حالت پنهان شبکه عصبی بازگشتی است. حالت پنهان $h$ در هر مرحله زمانی بروز می‌شود و به مرحله زمانی بعدی منتشر می‌شود. به این ترتیب، شبکه عصبی بازگشتی قادر است ورودی‌های قبلی را از طریق حالت پنهان در نظر بگیرد.

شبکه‌های عصبی بازگشتی با استفاده از روشی به نام پس‌انتشارِ در طول زمان آموزش داده می‌شوند. در این الگوریتم، گرادیان‌ها باید در هر مرحله زمانی با استفاده از قاعده‌ی زنجیره‌ای محاسبه شوند. هنگامی که دنباله‌ها طولانی هستند، پس‌انتشار به تعداد زیادی ضرب نیاز دارد. این امر می‌تواند منجر به مشکلی به نام محو یا انفجار گرادیان‌ها شود. هنگامی که گرادیان‌ها کوچک می‌شوند یا به عبارت دیگر با پدیده‌ی محو گرادیان مواجه می‌شوند، بروزرسانی‌های یادگیری به حداقل می‌رسد و باعث توقف یادگیری می‌شود. در مقابل، گرادیان‌های بزرگ باعث می‌شود که گام‌های یادگیری جهش داشته باشند و از نزدیک شدن به یک راه‌حل خوب جلوگیری شود. مشکل محو گردیان و انفجار گرادیان، توانایی شبکه‌های عصبی بازگشتی را برای یادگیریِ وابستگی‌هایِ زمانیِ طولانی محدود می‌کنند. از این‌رو، شبکه‌های عصبی بازگشتی دیگری برای مقابله با این مشکل، توسعه یافته‌اند.

## پس‌انتشار در طول زمان

شبکه‌های عصبی بازگشتی با نوعی خاصی از الگوریتم پس‌انتشار، به نام پس‌انتشار در طول زمان آموزش داده می‌شوند. همانند الگوریتم پس‌انتشار در شبکه عصبی پیش‌خور این الگوریتم نیز برای محاسبه گرادیان از قاعده‌ی زنجیره‌ای استفاده می‌کند. پس‌انتشار در شبکه‌های عصبی بازگشتی به دلیل خاصیت بازگشتی وزن‌ها و از بین رفتن آن‌ها با گذشت زمان، کمی چالش برانگیزتر است. چراکه نیاز است گراف محاسباتی یک RNN را یکبار گسترش داده تا وابستگی‌ها را بین متغیرها و پارامترهای مدل بدست آوریم. سپس، از پس‌انتشار و با استفاده از قاعده‌ی زنجیره‌ای محاسبه گرادیان‌ها و ذخیره گرادیان‌ها انجام گیرد. از آنجایی که توالی‌ها می‌توانند طولانی باشند، بنابراین وابستگی می‌تواند طولانی باشد. به عنوان مثال، برای دنباله‌ای از ۱۰۰۰ نویسه، اولین نویسه به طور بالقوه می‌تواند تأثیر قابل توجهی در نویسه در مکان نهایی داشته باشد. این امر از نظر محاسباتی واقعا امکان‌پذیر نیست. چراکه  بیش از حد به زمان و حافظه نیاز دارد.

در ادامه به تشریح پس‌انتشار در طول زمان به‌صورت ریاضی می‌پردازیم. برای درک اینکه این رویکرد چگونه عمل می‌کند نمایی از نحوه‌ی جریان اطلاعات را در شکل زیر می‌بینید:

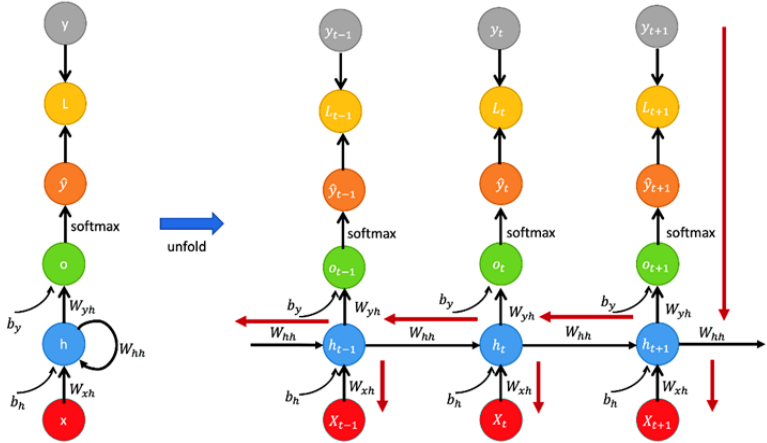

برای اینکه بتوان از پس‌انتشار در طول زمان در فرآیند آموزش شبکه عصبی بازگشتی استفاده شود، ابتدا باید تابع زیان را محاسبه کرد:

$$L(\hat{y}, y) = \sum_{t=1}^{T} L_t(\hat{y}_t, y_t)$$

$$= -\sum_{t}^{T} y_t \log \hat{y}_t$$

$$= -\sum_{t=1}^{T} y_t \log [softmax(o_t)]$$

از آنجایی‌که وزن $w_{yh}$ در تمام توالی زمان تقسیم می‌شود. از همین‌رو، می‌توانیم در هر مرحله از آن مشتق‌گرفته و همه را با هم جمع کرد:

$$\frac{\partial L}{\partial w_{yh}} = \sum_{t}^{T} \frac{\partial L_t}{\partial w_{yh}}$$

$$= \sum_{t}^{T} \frac{\partial L_t}{\partial \hat{y}_t} \frac{\partial \hat{y}_t}{\partial o_t} \frac{\partial o_t}{\partial w_{yh}}$$

$$= \sum_{t}^{T} (\hat{y} - y_t) \otimes h_t$$

که در این معادله $\frac{\partial o_t}{\partial w_{yh}} = h_t$ و $\otimes$ ضرب خارجی دو بردار می‌باشد.

به همین ترتیب، می‌توان گرادیان بایاس $b_y$ را بدست آوریم:

$$\frac{\partial L}{\partial b_y} = \sum_{t}^{T} \frac{\partial L_t}{\partial \hat{y}_t} \frac{\partial \hat{y}_t}{\partial o_t} \frac{\partial o_t}{\partial b_y}$$

$$= \sum_{t}^{T} (\hat{y} - y_t)$$

به‌علاوه بیاید از $L_{t+1}$ برای نشان دادن خروجی مرحله زمانی $t + 1$ استفاده کنیم:

$$L_{t+1} = -y_{t+1} log \hat{y}_{t+1}$$

حال، جزئیات مربوط به گرادیان $w_{hh}$ را با توجه به زمان $t + 1$ را مرور می‌کنیم:

$$\frac{\partial L_{t+1}}{\partial w_{hh}} = \frac{\partial L_{t+1}}{\partial \hat{y}_{t+1}} \frac{\partial \hat{y}_{t+1}}{\partial h_{t+1}} \frac{\partial h_{t+1}}{\partial w_{hh}}$$

از آنجایی که حالت پنهان $h_{t+1}$ با توجه به معادله بازگشتی $h_t$:

$$h_t = \tanh\left(w_{xh}^T . x_t + w_{hh}^T . h_{t-1} + b_h\right)$$

نیز بستگی دارد. بنابراین، در مرحله زمانی $t - 1 \rightarrow t$ می‌توان مشتق جزئی را با توجه به $w_{hh}$ به‌صورت زیر بدست آورد:

$$\frac{\partial L_{t+1}}{\partial w_{hh}} = \frac{\partial L_{t+1}}{\partial \hat{y}_{t+1}} \frac{\partial \hat{y}_{t+1}}{\partial h_{t+1}} \frac{\partial h_{t+1}}{\partial h_t} \frac{\partial h_t}{\partial w_{hh}}$$

بنابراین، در مرحله زمانی $t + 1$، می‌توانیم گرادیان را محاسبه کرده و از طریق پس‌انتشار در طول زمان از $t + 1$ به $t$ استفاده می‌کنیم تا محاسبه گرادیان کلی را با توجه به $w_{hh}$ بدست آوریم:

$$\frac{\partial L_{t+1}}{\partial w_{hh}} = \sum_{k=1}^{t+1} \frac{\partial L_{t+1}}{\partial \hat{y}_{t+1}} \frac{\partial \hat{y}_{t+1}}{\partial h_{t+1}} \frac{\partial h_{t+1}}{\partial h_k} \frac{\partial h_k}{\partial w_{hh}}$$

توجه داشته باشید که $\frac{\partial h_{t+1}}{\partial h_k}$ خود یک قانون زنجیره‌ای است. به عنوان مثال:

$$\frac{\partial h_3}{\partial h_1} = \frac{\partial h_3}{\partial h_2} \frac{\partial h_2}{\partial h_1}$$

همچنین، توجه داشته باشید که چون مشتق یک تابع یک بردار در نظر می‌گیریم، نتیجه یک ماتریس است (ماتریس ژاکوبین*) که همه عناصر آن مشتقات جزئی هستند. می‌توانیم گرادیان فوق را دوباره بازنویسی کنیم:

$$\frac{\partial L_{t+1}}{\partial w_{hh}} = \sum_{k=1}^{t+1} \frac{\partial L_{t+1}}{\partial \hat{y}_{t+1}} \frac{\partial \hat{y}_{t+1}}{\partial h_{t+1}} \left( \prod_{j=k}^{t} \frac{\partial h_{j+1}}{\partial h_j} \right) \frac{\partial h_k}{\partial w_{hh}}$$

جایی‌که:

$$\prod_{j=k}^{t} \frac{\partial h_{j+1}}{\partial h_j} = \frac{\partial h_{t+1}}{\partial h_k} = \frac{\partial h_{t+1}}{\partial h_t} \frac{\partial h_t}{\partial h_{t-1}} \cdots \frac{\partial h_{k+1}}{\partial h_k}$$

گرادیان‌های را با توجه به $w_{hh}$ در کل مراحل پس‌انتشار جمع می‌شود و در نهایت می‌توانیم گرادیان زیر را با توجه به $w_{hh}$ بدست آورد:

$$\frac{\partial L}{\partial w_{hh}} = \sum_{t}^{T} \sum_{k=1}^{t+1} \frac{\partial L_{t+1}}{\partial \hat{y}_{t+1}} \frac{\partial \hat{y}_{t+1}}{\partial h_{t+1}} \frac{\partial h_{t+1}}{\partial h_k} \frac{\partial h_k}{\partial w_{hh}}$$

اکنون بیایید تا گرادیان را با توجه به $w_{xh}$ استخراج کنیم. به‌طور مشابه، مرحله زمانی $t+1$ را در نظر می‌گیریم و گرادیان را با توجه $w_{xh}$ به‌صورت زیر بدست می‌آوریم:

$$\frac{\partial L_{t+1}}{\partial w_{xh}} = \frac{\partial L_{t+1}}{\partial \hat{y}_{t+1}} \frac{\partial \hat{y}_{t+1}}{\partial h_{t+1}} \frac{\partial h_{t+1}}{\partial w_{xh}}$$

از آنجاکه $h_t$ و $x_{t+1}$ هر دو در $h_{t+1}$ مشارکت دارند، بنابراین برای پس‌انتشار به $h_t$ نیاز داریم. اگر این مشارکت را در نظر بگیریم، خواهیم داشت:

$$\frac{\partial L_{t+1}}{\partial w_{xh}} = \frac{\partial L_{t+1}}{\partial \hat{y}_{t+1}} \frac{\partial \hat{y}_{t+1}}{\partial h_{t+1}} \frac{\partial h_{t+1}}{\partial w_{xh}} + \frac{\partial L_{t+1}}{\partial \hat{y}_{t+1}} \frac{\partial \hat{y}_{t+1}}{\partial h_t} \frac{\partial h_t}{\partial w_{xh}}$$

---

* با توجه به تابعی از نگاشت $n-$بعدی بردار $x$ به یک بردار خروجی$m-$بعدی، $f: \mathbb{R}^n \rightarrow \mathbb{R}^m$ ، ماتریس تمام مشتقات جزئی درجه یک این تابع را ماتریس ژاکوبین ($J$) گویند:

$$J = \begin{bmatrix} \frac{\partial f_1}{\partial x_1} & \cdots & \frac{\partial f_1}{\partial x_n} \\ \vdots & \ddots & \vdots \\ \frac{\partial f_m}{\partial x_1} & \cdots & \frac{\partial f_m}{\partial x_n} \end{bmatrix}$$

بنابراین، با جمع همه مشارکت‌ها از ۱ + $t$ به $t$ از طریق پس‌انتشار، می‌توان گرادیان را در مرحه زمانی ۱ + $t$ بدست آوریم:

$$\frac{\partial L_{t+1}}{\partial w_{xh}} = \sum_{k=1}^{t+1} \frac{\partial L_{t+1}}{\partial \hat{y}_{t+1}} \frac{\partial \hat{y}_{t+1}}{\partial h_{t+1}} \frac{\partial h_{t+1}}{\partial h_k} \frac{\partial h_k}{\partial w_{xh}}$$

علاوه بر این، می‌توانیم مشتق را با توجه به $w_{xh}$ در کل دنباله در نظر بگیریم:

$$\frac{\partial L}{\partial w_{xh}} = \sum_{t}^{T} \sum_{k=1}^{t+1} \frac{\partial L_{t+1}}{\partial \hat{y}_{t+1}} \frac{\partial \hat{y}_{t+1}}{\partial h_{t+1}} \frac{\partial h_{t+1}}{\partial h_k} \frac{\partial h_k}{\partial w_{xh}}$$

همچنین فراموش نشود که $\frac{\partial h_{t+1}}{\partial h_k}$ خود یک قانون زنجیره‌ای است.

همان‌طور که بیان شد مشکلات محو و انفجار گرادیان در شبکه عصبی بازگشتی معمولی وجود دارد. به‌طور کلی دو عامل وجود دارد که بر میزان گرادیان‌ها تاثیر گذار می‌باشد: **وزن‌ها و توابع فعال‌سازی یا به‌طور دقیق‌تر، مشتقات آن‌ها** که گرادیان از آن‌ها عبور می‌کند. در شبکه عصبی بازگشتی معمولی، محو و انفجار گرادیان از اتصالات بازگشتی (مکرر) ناشی می‌شود. واضح‌تر، این دو مشکل به دلیل مشتق بازگشتی $\frac{\partial h_{t+1}}{\partial h_k}$ است که در معادله $w_{xh}$ اتفاق می‌افتد و باید محاسبه شود:

$$\prod_{j=k}^{t} \frac{\partial h_{j+1}}{\partial h_j} = \frac{\partial h_{t+1}}{\partial h_k} = \frac{\partial h_{t+1}}{\partial h_t} \frac{\partial h_t}{\partial h_{t-1}} \cdots \frac{\partial h_{k+1}}{\partial h_k}$$

و نمایانگر ضرب ماتریس روی دنباله می‌باشد.

از آنجایی که شبکه عصبی بازگشتی معمولی نیاز دارد تا گرادیان پس‌انتشار را در یک توالی طولانی (با مقادیر کوچک در ضرب ماتریس) بدست آورد، از همین رو مقدار گرادیان لایه به لایه کاهش می‌یابد و در نهایت پس از طی چند مرحله از بین می‌رود. بنابراین، حالاتی که از مرحله زمانی فعلی فاصله دارند، به محاسبه پارامترهای گرادیان که همان پارامترهای یادگیری در شبکه عصبی بازگشتی هستند، هیچ کمکی نخواهند کرد.

محو گرادیان منحصر به شبکه عصبی بازگشتی معمولی نیست. همان‌طور که پیش‌تر بیان شد، آن‌ها در شبکه‌های عصبی پیش‌خور نیز اتفاق می‌افتند. نکته فقط اینجاست که شبکه‌های عصبی بازگشتی به دلیل اینکه عمق زیادی دارد، این مشکلات برای آن رایج‌تر است. این دو مشکل در نهایت نشان می‌دهند که اگر گرادیان از بین برود، به این معنا خواهد بود که حالت‌های پنهان قبلی هیچ تاثیر واقعی در حالت‌های پنهان بعدی ندارند. به عبارت دیگر، هیچ وابستگی طولانی مدتی آموخته نمی‌شود. خوشبختانه، چندین روش برای رفع مشکل محو گرادیان وجود دارد. مقداردهی اولیه مناسب ماتریس‌های وزنی می‌تواند اثر گرادیان‌های محوشده را کاهش دهد. منظم‌سازی نیز می‌تواند کمک‌کننده باشد. راه حل دیگری که از دو مورد قبل بیشتر ترجیح داده

میشود، استفاده از تابع فعالسازی ReLU بهجای توابع فعالسازی تانژانت هذلولویگون یا سیگموید میباشد. مشتق ReLU یک ثابت ۰ یا ۱ است، بنابراین به احتمال زیاد مشکل محو گرادیان را ندارد. راه حل محبوبتر که امروزه بیشتر مورد استفاده میشود، استفاده از شبکههای حافظه طولانیِ کوتاهـ مدت است.

# حافظه طولانی کوتاه-مدت[۱]

شبکههای حافظه طولانیِ کوتاهـ مدت (LSTM) برای غلبه بر مشکل محو یا انفجار گرادیان را ساخته شدند. LSTM دارای یک سلول حالت (وضعیت) است که علاوهبر حالت پنهان، جریان اطلاعات را بین مراحل زمانی کنترل میکند. LSTM همچنین دارای دروازههایی است که برای تغییر حالت و تشکیل یک خروجی استفاده میشوند. یک نمای کلی از یک سلول LSTM در شکل ۷- ۴ ارائه شده است.

دروازهی فراموشی برای فراموش کردن اطلاعات نامربوط از وضعیت سلول استفاده میشود. دروازهی فراموشی ماتریس وزن خود را دارد. ورودیهای دروازه فراموشی، حالت پنهان قبلی $h_{t-1}$ و ورودی فعلی $x_t$ هستند. از یک تابع سیگموید برای ایجاد خروجی با مقدار بین صفر و یک برای هر یک از عناصر در سلول حالت استفاده میشود. یک ضرب عنصری بین خروجیِ دروازهی فراموشی و سلول حالت انجام میشود. مقدار یک در خروجی دروازهی فراموشی به معنای حفظ کاملِ اطلاعاتِ عنصر در سلول حالت است، در مقابل، صفر به معنای فراموش کردن کامل اطلاعات در عنصر سلول حالت است. معادلهی دروازهی فراموشی بهصورت زیر است:

$$f_t = \sigma(W_f.[h_{t-1}, x_t] + b_f)$$

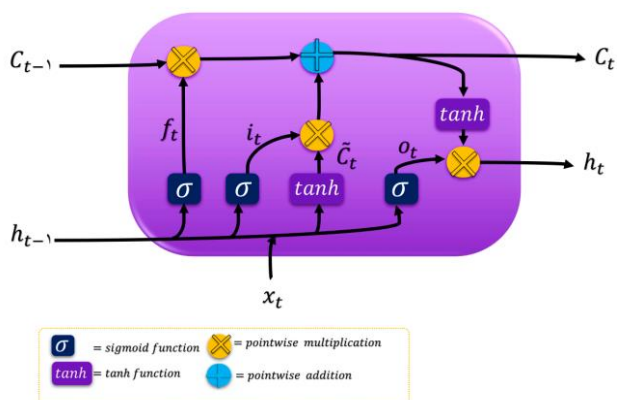

**شکل ۷ ـ ۴.** ساختار یک سلول LSTM

---

[۱] Long Short-Term Memory Networks

دومین عملیات در سلول LSTM، دروازه‌ی ورودی است. دروازه‌ی ورودی اطلاعات جدیدی را از مرحله زمانی فعلی و حالت پنهان قبلی شناسایی می‌کند که باید در حالت سلول گنجانده شود. این کار در دو بخش انجام می‌شود: تصمیم بگیرد که کدام مقادیر را بروزرسانی کند و سپس ایجاد مقادیر برای بروزرسانی است. ابتدا از بردار $i_t$ برای انتخاب مقادیری از نامزدهای جدید بالقوه برای گنجاندن در حالت سلول استفاده می‌شود. بردار نامزد $C_t$ نیز ماتریس وزن مخصوص به خود را دارد و از حالت پنهان قبلی و ورودی‌ها برای تشکیل برداری با ابعاد مشابه سلول حالت استفاده می‌کند. برای ایجاد این بردار نامزد از یک تابع tanh به عنوان یک تابع غیرخطی استفاده می‌شود. سپس یک ضرب عنصری بین بردار ورودی $i_t$ و نامزدهای $C_t$ انجام می‌شود تا انتخاب شود کدام اطلاعات جدید در سلول حالت جدید گنجانده شوند. در نهایت، نتیجه‌ی حاصل‌ضرب به سلول حالت اضافه می‌شود. این فرآیند در معادلات زیر نشان داده شده است:

$$i_t = \sigma(W_i.\,[h_{t-1}, x_t] + b_i)$$

$$\tilde{c}_t = tanh(W_c.\,[h_{c-1}, x_t] + b_c)$$

دروازه‌ی فراموشی و ورودی نحوه‌ی بروزرسانی سلول حالت را در هر مرحله زمانی مشخص می‌کنند. بروزرسانی سلول حالت در یک مرحله زمانی از طریق معادله زیر انجام می‌شود:

$$c_t = f_t * c_{t-1} + i_t * \tilde{c}_t$$

سرانجام باید تصمیم گرفته شود که چه چیزی در خروجی باشد. خروجی نهایی یک سلول LSTM، حالت پنهان $h_t$ است. ابتدا، از یک تابع سیگموید برای محاسبه بردار با مقادیر بین صفر و یک استفاده می‌شود تا انتخاب مقادیرِ سلولِ وضعیت در مرحله زمانی را انجام دهد. سپس مقدار سلول حالت را به یک لایه تانژانت هذلولوی می‌دهیم تا در نهایت مقدار آن را در خروجیِ لایه قبلی سیگموید ضرب کرده، تا قسمت‌های مورد نظر در خروجی به اشتراک گذاشته شوند. معادلات زیر این روند را نشان می‌دهند:

$$o_t = \sigma(W_o.\,[h_{t-1}, x_t] + b_o)$$

$$h_t = o_t * tanh\,(c_t)$$

روشی که LSTM مشکل محو گرادیان یا انفجار را کاهش می‌دهد با در معادله

$$c_t = f_t * c_{t-1} + i_t * \tilde{c}_t$$

LSTMها ساختار داخلی پیچیده‌ای دارند که چندین لایه از نورون‌ها را شامل می‌شود و می‌توان آن‌ها را به تنهایی شبکه در نظر گرفت. با این حال، آن‌ها همچنین می‌توانند به عنوان بلوک ساختمانی یک شبکه عصبی بازگشتی استفاده شوند. این با جایگزینی لایه پنهان در یک شبکه عصبی بازگشتی با یک واحد LSTM بدست می‌آید. LSTMها در پردازش زبان بسیار

موفق بوده‌اند. به عنوان مثال، آن‌ها در حال حاضر شبکه‌ی استانداردی هستند که برای تشخیصِ گفتار در تلفن‌های همراه استفاده می‌شود.

## شبکه‌ی عصبی پیچشی

یک شبکه‌ی عصبی پیش‌خورِ کاملا متصل را در نظر بگیرید که یک تصویر RGB ساده به اندازه [۲۵۶ × ۲۵۶ × ۳] را به عنوان ورودی خود می‌گیرد. بر این اساس، هر نورون به تنهایی دارای ۱۹۶۶۰۸=[۳×۲۵۶×۲۵۶] وزن است و این وزن فقط برای یک نورون است!! حال آن‌که، معماری‌های عمیق به تعداد زیادی نورون و لایه‌های پنهان نیاز دارند تا به اندازه‌ی کافی ساختارهای پیچیده‌ی موجود در داده‌های ورودی را نشان دهند. این بدان معناست که ماهیت کاملا متصل این شبکه‌ها، حافظه زیادی خصوصا برای تصاویر یا ویدیوهای بزرگ مصرف می‌کند. علاوه براین، تعداد زیاد پارامترها میل شبکه را به بیش‌برازش افزایش می‌دهد. از این‌رو، به منظور پرداختن به این مسائل، شبکه‌های عصبی پیچشی یا کانولوشنی (CNN) به عنوان توسعه‌ی بسیار محبوب شبکه‌های عصبی استاندارد معرفی شدند. شبکه‌ی عصبی پیچشی دسته‌ای از شبکه‌های عصبی پیش‌خور هستند که از لایه‌های پیچش برای تجزیه و تحلیل ورودی‌هایی با توپولوژی‌های مشبکی، همانند تصاویر و ویدیوها استفاده می‌کنند. نام این شبکه‌ها بر اساس تابع ریاضی به نام پیچش است که در ساختار خود به کار می‌برند. به‌طور خلاصه، شبکه‌های پیچشی، شبکه‌های عصبی هستند که از کانولوشن به جای ضرب ماتریس، حداقل در یکی از لایه‌های خود استفاده می‌کنند.

## ساختار شبکه‌ی پیچشی

در معماری یک شبکه پیچشی دو بخش اصلی وجود دارد:

- **استخراج ویژگی:** در این بخش با استفاده از **پیچش** و **ادغام** ویژگی‌های مختلف تصویر را شناسایی می‌کند.

- **دسته‌بندی:** در بخش با اسفاده از یک لایه متصل کامل، از خروجی فرآیند استخراج ویژگی استفاده می‌کند و کلاس تصویر را بر اساس ویژگی‌های استخراج شده در مراحل قبلی پیش‌بینی می‌کند. معماری کلی آن در شکل ۷ــ۵ قابل مشاهده است.

### لایه پیچش

برای یک تصویر دو بعدی $I$، پیچش گسسته به صورت زیر تعریف می‌شود:

$$S(i,j) = (I * K)(i,j) \sum_m \sum_n I(i-m, j-n)K(m,n)$$

که در آن $K(m,n)$ یک هسته دو بعدی است و خروجی $S(i,j)$ به عنوان نقشه ویژگی نامیده می‌شود. به‌طور شهودی، این عملیات، هسته را در طول تصویر $I$ "می‌لغزاند[1]" و مجموع وزنی هسته را در هر موقعیت $i$ و $j$ تصویر محاسبه می‌کند. نمونه‌ای از چنین پیچش دو بعدی گسسته

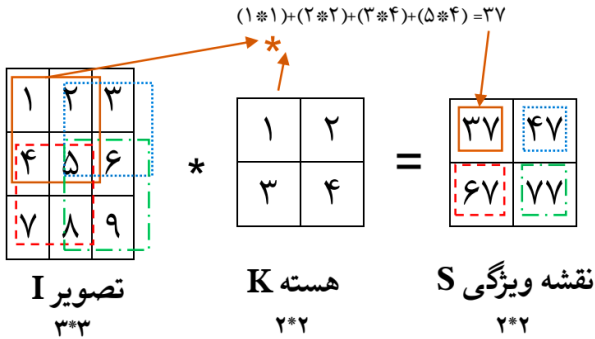

$(1*1)+(2*2)+(3*4)+(5*4) = ۳۷$

**تصویر I**
**۳*۳**

**هسته K**
**۲*۲**

**نقشه ویژگی S**
**۲*۲**

در شکل مقابل نشان داده شده است. در شبکه عصبی پچشی، هسته K وزن‌های قابل یادگیری لایه پیچش را توصیف می‌کند و هر لایه پیچش می‌تواند حاوی تعداد دلخواه هسته باشد که هر کدام منجر به نقشه ویژگی خروجی خود می‌شود.

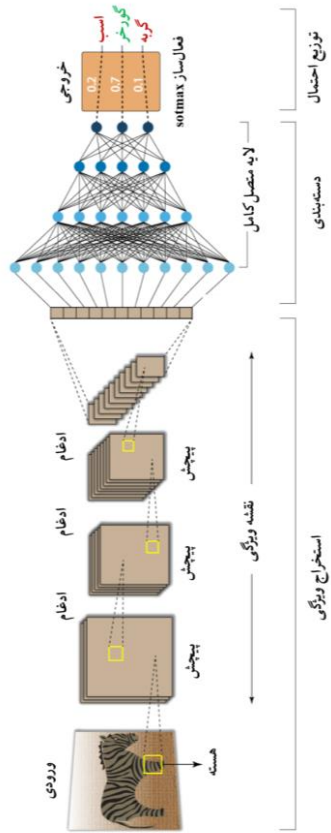

**شکل ۷ ـ ۵. شمایی کلی از ساختار یک شبکه پیچشی**

---

هر لایه پیچش دارای مجموعه خاصی از ابرپارمترها است که هر یک از آن‌ها تعداد ارتباطات و اندازهی خروجیِ نقشه‌های ویژگی را تعیین می‌کند:

- **اندازه هسته:** اندازه‌ی هسته‌ی K (گاهی اوقات اندازه فیلتر نیز نامیده می‌شود) **میدان پذیرنده**[1] را توصیف می‌کند که برای همه مکان‌های ورودی اعمال می‌شود. افزایش این پارامتر به لایه پیچش اجازه می‌دهد تا اطلاعات فضایی بیشتری را دریافت کند، در حالی که به‌طور هم‌زمان تعداد وزن‌های شبکه را افزایش می‌دهد.

- **تعداد هسته:** تعداد هسته‌ها مستقیما با تعداد پارامترهای قابل یادگیری و عمق D حجم خروجی یک لایه پیچش مطابقت دارد. همان‌طور که هر هسته یک نقشه ویژگی خروجی جداگانه تولید می‌کند، هسته‌های D یک نقشه‌ی ویژگی خروجی با عمق D را تولید می‌کنند.

- **گام:** همان‌طور که قبلا توضیح داده شد، پیچش را می‌توان به عنوان جمع‌وزنی با "لغزاندن" یک هسته بر روی یک حجم ورودی درک کرد. با این حال، نیازی نیست که "لغزش" با یک فاصله یک پیکسل در یک زمان اتفاق بیفتد، چیزی که گام توصیف می‌کند. گام $S$ تعداد پیکسل‌هایی را که هسته بین هر محاسبه‌ی ویژگیِ خروجی جابجا می‌شود را مشخص می‌کند. گام‌های بزرگ‌تر، نقشه‌های ویژگیِ خروجیِ کوچک‌تری تولید می‌کنند، زیرا محاسبات کم‌تری انجام می‌شود. این مفهوم در شکل زیر نشان‌داده شده است:

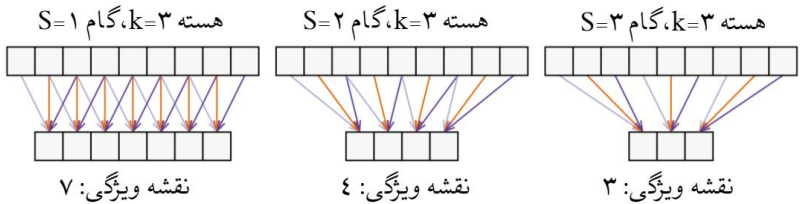

| هسته ۳،k=گام S=۱ | هسته ۳،k=گام S=۲ | هسته ۳،k=گام S=۳ |
|---|---|---|
| نقشه ویژگی: ۷ | نقشه ویژگی: ٤ | نقشه ویژگی: ۳ |

- **لایه‌گذاری ـ صفر:** به دلیل نحوه عملکرد عملیات پیچش، از لایه‌گذاری ـ صفر برای کنترل کاهش ابعاد پس از اعمال فیلترهای بزرگ‌تر از ۱*۱ و جلوگیری از گم شدن اطلاعات در حاشیه استفاده می‌شود. به عبارت دیگر، از لایه‌گذاری ـ صفر اغلب استفاده می‌شود تا ابعاد فضایی لایه‌های ورودی و خروجی را یکسان نگه داشت. با اضافه کردن ورودیِ صفر در اطراف حاشیه، می‌توان کوچک‌شدن ابعاد فضایی هنگام انجام پیچش را دور زد. مقدار صفرهای اضافه شده در هر طرف برای هر بعد فضایی یک ابرپارمتر اضافی $P$ است. نمونه‌ای از لایه‌گذاری صفر در شکل زیر نشان داده شده است:

---

[1] receptive field

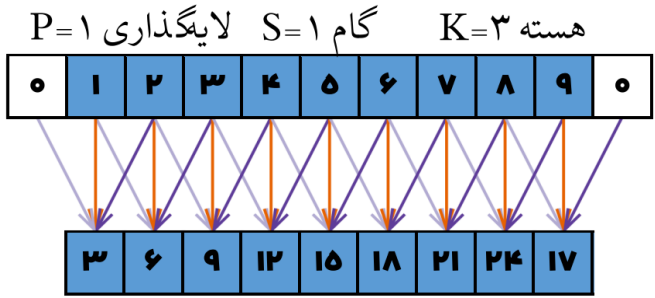

- **فراخش[1] (اتساع)**: فراخش $d$ که اخیرا معرفی شده است ابرپارمتر دیگری است که به لایه پیچش اجازه می‌دهد تا میدان پذیرنده‌ی موثرتری نسبت به ورودی داشته باشد، در حالی که اندازه هسته ثابت نگه می‌دارد. این امر با معرفی $d$ فاصله بین هر سلول از هسته بدست می‌آید. پیچش استاندارد، به سادگی از فراخش ۰ استفاده می‌کند. از این‌رو دارای یک هسته پیوسته است. با افزایش فراخش این امکان برای یک لایه‌ی پیچش وجود دارد که وسعت فضایی بیشتری از ورودی را بگیرد و در عین حال مصرف حافظه را ثابت نگه دارد. مفهوم پیچش‌های فراخش که گاهی اوقات **پیچش‌های آتروس[2]** نیز نامیده می‌شود، با فراخش‌های مختلف در شکل ۷–۶ نشان داده شده است.

با توجه به اندازه‌ی حجم ورودی $W$، اندازه‌ی هسته‌ی $K$، گام $S$، فراخش $d$ و $P$ لایه‌گذاری، حجم خروجی حاصل به صورت زیر محاسبه می‌شود:

$$W_o = \left\lfloor \frac{W + 2P - K - (K-1)(d-1)}{S} \right\rfloor + 1.$$

استفاده از پیچش دارای سه مزیت مهم است. اولا، شبکه‌های عصبی پیچشی معمولا دارای **ارتباط‌های خلوت[3]** هستند. شبکه‌های عصبی پیش‌خور از ماتریسی از پارمترها استفاده می‌کنند که ارتباط بین واحد ورودی و خروجی را توصیف می‌کند. این بدان معناست که هر واحد خروجی با هر واحد ورودی ارتباط دارد. با این حال، شبکه‌های عصبی پیچشی دارای ارتباط خلوت هستند که با کوچک‌تر کردن هسته از ورودی بدست می‌آید. به عنوان مثال، یک تصویر می‌تواند میلیون‌ها یا هزاران پیکسل داشته باشد، اما در حین پردازش آن با استفاده از هسته، می‌توانیم اطلاعات معنی‌داری که ده‌ها یا صدها پیکسل هستند را شناسایی کنیم. این بدان معنی است که ما باید پارمترهای کم‌تری را ذخیره کنیم که نه تنها نیاز به حافظه را کاهش می‌دهد، بلکه کارایی

---

[1] Dilation

[2] atrous convolutions

[3] Sparse interactions

آماری مدل را نیز بهبود می‌بخشد. ثانیاً، شبکه‌های عصبی پیچشی از **اشتراک‌گذاری پارامتر**[1] استفاده می‌کنند. به این معنا که آن‌ها از پارامترهای مشابه برای چندین تابع دوباره استفاده می‌کنند. اشتراک‌گذاری پارامترها هم‌چنین باعث آخرین مزیت اصلی یعنی **هم‌وردایی**[2]می‌شود. هم‌وردایی به این معنی است که اگر ورودی جابجا شود، خروجی نیز به همان صورت جابجا می‌شود. این ویژگی برای پردازش داده‌های دوبعدی ضروری است، چراکه اگر یک تصویر یا بخشی از یک تصویر به جای دیگری در تصویر منتقل شود، نمایش یکسانی خواهد داشت.

تصویر ۹*۹=I، هسته ۳*۳=K، فراخش ۱=d    تصویر ۹*۹=I، هسته ۳*۳=K، فراخش ۰=d

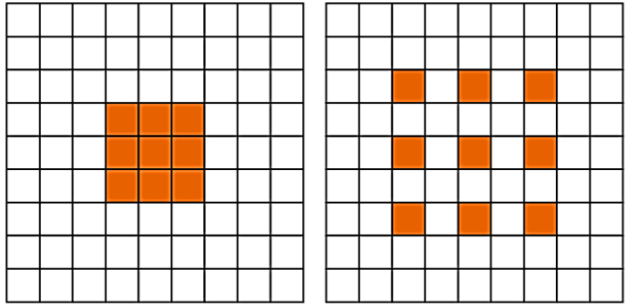

تصویر ۹*۹=I، هسته ۳*۳=K، فراخش ۳=d    تصویر ۹*۹=I، هسته ۳*۳=K، فراخش ۲=d

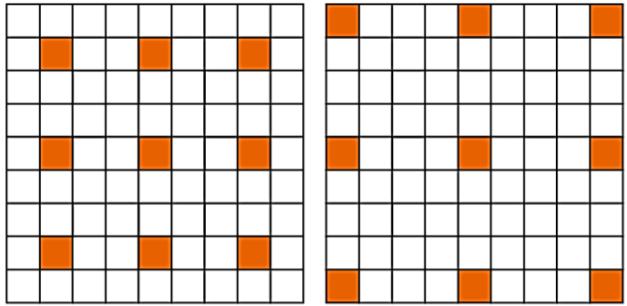

**شکل ۷ـ۶.** فراخش روی ورودی دو بعدی با اندازه‌های مختلف.

## لایه ادغام

به‌منظور پایین نگه داشتن مقدار پارامترها و افزایش بیشتر میدان پذیرنده‌ی مؤثر خروجی‌ها با توجه به ورودی، استفاده از شکل خاصی از نمونه‌برداری فضایی، به نام **ادغام**، پس از چند لایه پیچش در شبکه می‌تواند سودمند باشد. ادغام، مشابه با پیچش، به‌طور شهودی می‌تواند به عنوان یک مکانیسم هسته‌ی لغزنده، با پارامترهای مشابه، مانند گام و اندازه‌ی هسته درک شود. تفاوت

---

اصلی این است که ادغام یک تابع ثابت را روی ورودی‌های خود محاسبه می‌کند که معمولا عملیات **حداکثری** است. رایج‌ترین شکل ادغام شامل یک هسته [۲×۲] باگام ۲ می‌باشد. وقتی این هسته با استفاده از تابع حداکثری برروی حجم ورودی اعمال می‌شود، به‌طور موثری تکه‌های [۲×۲] غیرهم‌پوشانی از حجم ورودی پردازش می‌شود و تنها بیشترین مقدار را در نقشه‌ی ویژگی خروجی نگه می‌دارد و ۷۵ درصد از داده‌های ورودی را دور می‌اندازد. پس‌انتشار برای ادغام حداکثری به سادگی می‌تواند تنها با مسیریابی گرادیان از طریق ورودی که بیشترین مقدار را در گذر پس‌رو دارد، انجام شود.

از آنجایی که این تابع ثابت است، به هیچ *پارامتر قابل آموزش نیازی ندارد* و بنابراین مصرف حافظه و ظرفیت مدل معماری CNN را در مقایسه با **پیچش‌های گامی**[1] افزایش نمی‌دهد. با این حال، به نظر می‌رسد معماری‌های اخیر CNN از استفاده از ادغام برای نمونه‌کاهی[2] دوری می‌کنند و در عوض پیشنهاد می‌کنند همیشه از پیچش‌های گامی برای کاهش ابعاد فضایی استفاده شود. به نظر می‌رسد که این امر به ویژه در هنگام آموزش مدل‌های مولد مانند **شبکه‌های متخاصم مولد**[3] اهمیت دارد.

## لایه‌های غیرخطی

همچنان که پیش‌تر بیان شد، می‌توانیم از شبکه‌های عصبی عمیق برای تشخیص انواع تصویر استفاده کنیم. با این حال، اگر فقط از لایه‌های خطی استفاده کنیم، مانند پیچش، آنگاه می‌توان یک تبدیل خطی تک لایه برای جایگزینی شبکه‌ی عصبی عمیق پیدا کنیم. به عبارت دیگر، مهم نیست که شبکه از چند عملیات خطی تشکیل شده است، کل سیستم قدرتمندتر از یک رگرسیون خطی ساده نیست. به این ترتیب، هیچ راهی برای بهره‌مندی از شبکه‌های عصبی عمیق وجود ندارد. بنابراین، از آنجایی که پیچش یک عملیات خطی است و تصاویر غیرخطی هستند، لایه‌های غیرخطی اغلب مستقیما بعد از لایه پیچش قرار می‌گیرند تا رابطه‌ی غیرخطی بین ورودی و خروجی را برقرار کنند.

در شبکه‌های عصبی، غیرخطی بودن با استفاده از مفهوم تابع فعال‌سازی معرفی می‌شود. انواع مختلفی از توابع غیرخطی وجود دارد که محبوب‌ترین آنها عبارتند از:

- **سیگموید:** تابع فعال‌سازیِ سیگموید به‌صورت زیر تعریف می‌شود:

$$\sigma(x) = \frac{1}{1 + e^{-x}}$$

---

که ورودی با مقدار حقیقی $x$ را در محدوده بین ۰ و ۱ نگاشت می‌کند. تابع

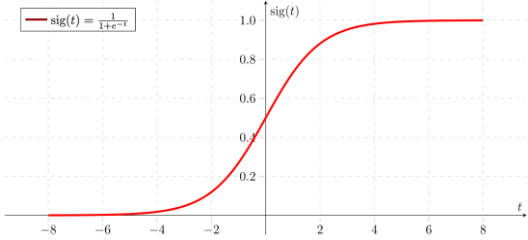

فعال‌سازی سیگموید در شکل مقابل نشان داده شده است. در شبکه‌های عصبی اولیه، سیگموید‌ها انتخابی محبوب بودند. با این حال،

تابع فعال‌سازی سیگموید دارای اشکالات قابل توجهی است. اشکال اصلی سیگموید این است که اشباع می‌شود و بنابراین فقط گرادیان‌های بسیار نزدیک به صفر را در این مناطق ارائه می‌دهند، که به طور موثر از ارائه گرادیان از طریق این نورون به همه ورودی‌ها توسط الگوریتم پس‌انتشار جلوگیری می‌کند. علاوه‌براین، خروجی تابع فعال‌سازی سیگموید **صفر ـ محور**[1] نیست که می‌تواند منجر به پویایی نامطلوب در طول گرادیان کاهشی شود. از این‌رو، استفاده از تابع فعال‌سازی سیگموید برای نورون‌های پنهان همیشه مجاز نیست. با این حال، برای نورون‌های خروجی، محدوده بین ۰ و ۱ می‌تواند مفید باشد. به عنوان مثال، برای تفسیر پیش‌بینی‌ها به عنوان احتمالات.

- **تانژانت هذلولی‌گون (tanh):** تابع فعال‌سازی تانژانت هذلولویی‌گون، ارتباط نزدیکی با تابع فعال‌سازی سیگموید دارد و شکل ریاضی آن به صورت زیر است:

$$\tanh(x) = \frac{\sinh(x)}{\cosh(x)} = \frac{e^x - e^{-x}}{e^x + e^{-x}} = ۲\sigma(۲x) - ۱.$$

همان طورکه در معادله بالا مشاهده می شود، tanh به سادگی یک نسخه مقیاس‌شده

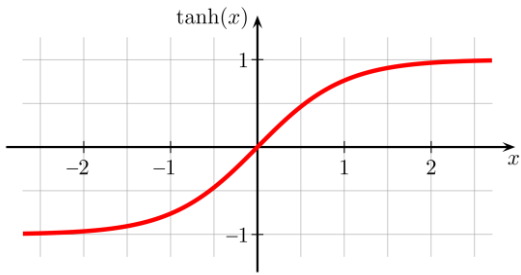

از فعال‌ساز سیگموید است. با این حال، صفر ـ محور است. بنابراین برخی از مسائلی را که فعال‌ساز سیگموید دارد را از خود نشان نمی‌دهد. فعال‌ساز tanh در شکل مقابل نشان داده شده است.

- **واحد یکسوشده‌ی خطی (ReLU):** تابع فعال‌سازی ReLU به صورت زیر تعریف می‌شود:

$$ReLU(x) = \max{(۰, x)}$$

---

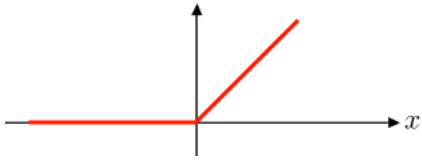

که در شکل مقابل قابل مشاهده است. در مقایسه با سیگموید و تانژانت هذلولی‌گون، ReLU کارآمدتر است و همگرایی را تسریع می‌کند.

## خلاصه فصل

- یادگیری عمیق الگوریتم‌هایی را توصیف می‌کند که داده‌ها را با ساختاری منطقی، شبیه به نحوه‌ی نتیجه‌گیریِ یک انسان تجزیه و تحلیل می‌کند.

- هسته اصلی یادگیری عمیق به روشی تکراری برای آموزش ماشین‌ها برای تقلید از هوش انسانی متکی است.

- در یادگیری عمیق، ما نیازی به برنامه‌نویسیِ صریحِ همه‌چیز نداریم.

- نورون‌های مصنوعی بلوک‌های اصلی شبکه‌های عصبی مصنوعی هستند.

- شبکه‌های عصبی پیش‌خور چندین نورون را ترکیب می‌کنند تا یک گراف جهت‌دار بدون دور تشکیل دهند.

- هدف از بهینه‌سازی شبکه‌های عصبی پیش‌خور، یافتن خودکار وزن‌ها و بایاس‌هایی است که شبکه خروجی هدف مورد نظر y را با ورودی x تقریب می‌زند.

- الگوریتم پس‌انتشار احتمالا اساسی‌ترین بلوک سازنده در یک شبکه عصبی است.

- پس‌انتشار اساسا تدبیر هوشمندانه‌ای برای محاسبه مؤثر گرادیان در شبکه‌های عصبی چندلایه است.

- یک اشکال بزرگ در بهینه‌سازی شبکه‌های عصبی از طریق پس‌انتشار و گرادیان‌کاهشی، مشکل محو گرادیان است.

- هدف از گرادیان‌ کاهشی تصادفی سرعت بخشیدن به فرآیند یادگیری با اندکی تغییر در رویه استاندارد گرادیان کاهشی است.

- هدف بهینه‌سازهای نرخ یادگیری تطبیقی، حل مشکل یافتن نرخ یادگیری درست است.

- در شبکه‌های عصبی، مقداردهی اولیه‌ی وزن‌ها باید با دقت بسیاری انتخاب شود.

- مجموعه اعتبارسنجی در شبکه‌های عصبی، معمولا برای تنظیم دقیق ابرپارمترهای مدل مانند معماری شبکه یا نرخ یادگیری استفاده می‌شود.

- "حذف تصادفی" در شبکه‌های عصبی، به فرآیند نادیده‌گرفتنِ تصادفیِ گره‌هایِ خاص در یک لایه در طول آموزش شبکه اشاره دارد.

- شبکه‌های عصبی بازگشتی یا مکرر، نوعی شبکه‌ی عصبی مصنوعی هستند که برای تشخیص الگوها در توالی داده‌ها طراحی شده‌اند.

- شبکه‌های حافظه طولانی کوتاه‌مدت برای غلبه بر مشکل محو یا انفجار گرادیان را ساخته شدند.
- شبکه‌ی عصبی پیچشی دسته‌ای از شبکه‌های عصبی پیش‌خور هستند که از لایه‌های پیچش برای تجزیه و تحلیل ورودی‌هایی با توپولوژی‌های ِمشبکی، همانند تصاویر و ویدیوها استفاده می‌کنند.

## مراجع برای مطالعه بیشتر

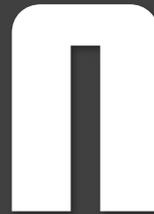

# یادگیری غیرنظارتی

**اهداف:**

- یادگیری غیرنظارتی چیست و چه مزایایی دارد؟
- آشنایی با خوشه‌بندی و انواع مختلف الگوریتم‌ها؟
- تفاوت بین انتخاب ویژگی و استخراج ویژگی
- کاهش ابعاد خطی و غیرخطی
- آشنایی با خودرمزنگار
- تفاوت مدل مولد و تفکیک‌کننده

# یادگیری غیرنظارتی و ضعف یادگیری بانظارت

یادگیری بانظارت در بهینه‌سازی عملکردِ وظایفی با مجموعه داده‌هایی با برچسب‌های فراوان، کارآیی بسیار خوبی از خود نشان می‌دهد. به عنوان مثال، مجموعه دادهی بسیار بزرگی از تصاویری از اشیا را در نظر بگیرید که هر تصویر برچسب‌گذاری شده است. اگر مجموعه داده به اندازه کافی بزرگ باشد، اگر آن را به اندازه کافی با استفاده از الگوریتم‌های یادگیری ماشین مناسب (شبکه‌های عصبی پیچشی) و با رایانه‌ای قدرتمند آموزش دهیم، می‌توانیم یک مدل دسته‌بندی تصویر مبتنی‌بر یادگیری بانظارتِ بسیار خوب بسازیم. همان‌طور که الگوریتم بانظارت بر روی داده‌ها آموزش می‌بیند، می‌تواند عملکرد خود را (از طریق تابع هزینه) با مقایسه برچسب تصویر پیش‌بینی‌شده خود با برچسب تصویر واقعی که در مجموعه داده داریم، اندازه‌گیری کند. الگوریتم، به صورت صریح سعی می‌کند این تابع هزینه را به حداقل برساند؛ به‌طوری که خطای آن در تصاویری که قبلا دیده نشده است (مجموعه آزمون) تا حد امکان کم باشد. به همین دلیل است که برچسب‌ها بسیار قدرتمند هستند، آن‌ها با ارائه یک معیار خطا به هدایت الگوریتم کمک می‌کنند. الگوریتم از معیار خطا برای بهبود عملکرد خود در طول زمان استفاده می‌کند. بدون چنین برچسب‌هایی، الگوریتم نمی‌داند که چقدر در دسته‌بندی درست تصاویر موفق است یا نیست. با این حال، گاهی اوقات هزینه‌ی برچسب‌گذاری دستی یک مجموعه داده بسیار بالا است.

علاوه بر این، به همان اندازه‌ای که مدل‌های یادگیری بانظارت قدرتمند هستند، در تعمیم‌دهیِ دانش فراتر از موارد برچسب‌گذاری شده‌ای که روی آن‌ها آموزش دیده‌اند نیز، محدود هستند. از آنجایی که اکثر داده‌های جهان بدون‌برچسب هستند، با استفاده از یادگیری بانظارت، توانایی هوش مصنوعی برای گسترش عملکرد خود به نمونه‌هایی که قبلا دیده نشده‌اند محدود است. به عبارت دیگر، یادگیری بانظارت در حل مسائل **هوش مصنوعی محدود**[1] **(ضعیف)** عالی است، اما در حل مسائل از نوع **هوش مصنوعی قوی**، چندان خوب نیست.

به عبارت دیگر، برای مسائلی که الگوها ناشناخته هستند یا به‌طور دائم در حال تغییر هستند یا مجموعه داده‌های برچسب‌گذاری‌شده کافی برای آن‌ها نداریم، یادگیری غیرنظارتی واقعا می‌درخشد. یادگیری غیرنظارتی، به جای هدایت شدن توسط برچسب‌ها، با یادگیریِ ساختارِ زیربنایی داده‌هایی که روی آن‌ها آموزش دیده است، کار می‌کند. یادگیری غیرنظارتی این کار را با تلاش برای بازنمایی از داده‌هایی که روی آن آموزش می‌بیند با مجموعه‌ای از پارامترها انجام می‌دهد. با انجام این **یادگیری بازنمایی**[2]، یادگیری غیرنظارتی می‌تواند الگوهای متمایزی را در

---

مجموعه داده شناسایی کند. در مثال مجموعه داده تصویر (این بار بدون‌برچسب)، یادگیری غیرنظارتی ممکن است بتواند تصاویر را بر اساس شباهت آن‌ها به یکدیگر و تفاوت آن‌ها با بقیه شناسایی و گروه‌بندی کند. به عنوان مثال، تمام تصاویری که شبیه صندلی هستند باهم و همه تصاویری که شبیه به گربه هستند با هم گروه‌بندی می‌شوند. البته، خود یادگیری غیرنظارتی نمی‌تواند این گروه‌ها را به عنوان "صندلی" یا "گربه" برچسب‌گذاری کند. با این حال، اکنون که تصاویر مشابه با هم گروه‌بندی شده‌اند، انسان وظیفه برچسب‌گذاری بسیار ساده‌تری دارد. به‌جای برچسب‌گذاری میلیون‌ها تصویر با دست، انسان‌ها می‌توانند به صورت دستی همه گروه‌های مجزا را برچسب‌گذاری کنند و این برچسب‌ها برای همه اعضای هر گروه اعمال شوند.

از این‌رو، یادگیری غیرنظارتی، مسائل حل نشدنی قبلی را قابل حل‌تر می‌کند و در یافتنِ الگوهای پنهان، هم در داده‌های گذشته‌ی در دسترس برای آموزش و هم در داده‌های آینده، بسیار چابک‌تر عمل می‌کند. حتی اگر یادگیری غیرنظارتی در حل مسائل خاص (مسائل محدود هوش مصنوعی) مهارت کم‌تری نسبت به یادگیری بانظارت دارد، اما در مقابله با مشکلات بازتر از نوع هوش مصنوعی قوی و تعمیم این دانش بهتر است. *مهم‌تر از آن، یادگیری غیرنظارتی می‌تواند بسیاری از مشکلات رایجی را که دانشمندان داده هنگام ساخت راه‌حل‌های یادگیری ماشین با آن مواجه می‌شوند، برطرف کند.*

## هوش مصنوعی ضعیف و قوی

امروزه، هوش مصنوعی بر لبان همه است و حتی یک روز هم نمی‌گذرد که در مورد هوش مصنوعی نشنیده باشیم. با این حال، گفتگو در مورد هوش مصنوعی اغلب منجر به سوء تفاهم می‌شود. این امر از آنجایی ناشی می‌شود که هیچ تعریف مشخصی از هوش مصنوعی وجود ندارد. دستیارهای شخصی همانند سیری، الکسای آمازون یا گوگل هوم ممکن است برخی از کاربران را به این فکر فرو ببرد که آن‌ها در حال گفتگو با آن‌ها هستند یا توسط آن‌ها درک می‌شوند.

به طور کلی، هوش مصنوعی شامل الگوریتم‌های پیشرفته‌ای است که از یک تابع ریاضی پیروی می‌کنند که می‌تواند فرآیندهای پیچیده‌ای شبیه به انسان را انجام دهد. به عنوان مثال می‌توان به درک بصری، تشخیص گفتار، تصمیم‌گیری و ترجمه بین زبان‌ها اشاره کرد. به‌طور کلی، دو موضوع فکری در هوش مصنوعی وجود دارد: هوش مصنوعی ضعیف و هوش مصنوعی قوی. به‌اصطلاح دستیارهای شخصی که امروزه به آن‌ها علاقه زیادی داریم، سیری، الکسای آمازون یا گوگل هوم، برنامه‌های ضعیف هوش مصنوعی به حساب می‌آیند، زیرا در مجموعه‌ای از عملکردهای از پیش تعریف‌شده‌ای محدود عمل می‌کنند. حتی برنامه‌های پیشرفته شطرنج نیز هوش مصنوعی ضعیف در نظر گرفته می‌شوند. به نظر می‌رسد این دسته‌بندی ریشه در تفاوت بین برنامه‌نویسی بانظارت و غیرنظارتی دارد. دستیارهای شخصی و شطرنج اغلب یک پاسخ برنامه‌ریزی شده دارند. آن‌ها طبقه‌بندی را براساس چیزهایی مشابه آنچه را که می‌دانند

(از طریق یادگیری از داده‌ها)، انجام می‌دهند. این تجربه‌ای شبیه به انسان را ارائه می‌کند، با این حال، آن تنها یک شبیه‌سازی است. اگر از الکسا بخواهید تلویزیون را روشن کند، برنامه‌نویسی کلمات کلیدی مانند روشن و تلویزیون را می‌فهمد. الگوریتم با روشن کردن تلویزیون پاسخ می‌دهد، اما فقط به برنامه‌های آن پاسخ می‌دهد. به عبارت دیگر، هیچ یک از معنای آنچه شما گفته‌اید را درک نمی‌کند.

در طرف دیگر، ماشین‌هایی با ذهن خودشان هستند که می‌توانند بدون دخالت انسان تصمیمات مستقل بگیرند. این برنامه‌ها را می‌توان هوش مصنوعی قوی در نظر گرفت. به عبارت دیگر، هوش مصنوعی قوی به ماشین‌ها یا برنامه‌هایی اطلاق می‌شود که ذهن خودشان را دارند و می‌توانند به تنهایی و بدون دخالت انسان، وظایف پیچیده‌ای را انجام دهند. هوش مصنوعی قوی دارای الگوریتم‌های پیچیده‌ای است که به سیستم‌ها کمک می‌کند در موقعیت‌های مختلف، عمل مناسبی انجام دهند و ماشین‌های مجهز به هوش مصنوعی قوی می‌توانند بدون تعامل انسانی تصمیم‌گیری مستقل کنند. ماشین‌های قوی با هوش مصنوعی می‌توانند وظایف پیچیده‌ای را به تنهایی انجام دهند، درست مانند انسان‌ها.

هوش مصنوعی قوی که در بسیاری از فیلم‌ها دیده می‌شود، بیشتر شبیه مغز عمل می‌کند. طبقه‌بندی نمی‌کند، اما از خوشه‌بندی و ارتباط برای پردازش داده‌ها استفاده می‌کند. به طور خلاصه، به این معنی است که پاسخ مشخصی برای کلمات کلیدی شما وجود ندارد. تابع، نتیجه را تقلید می‌کند، اما در این مورد، ما از نتیجه مطمئن نیستیم. مانند صحبت کردن با یک انسان، می‌توانید فرض کنید که یک نفر با چه چیزی به یک سوال پاسخ می‌دهد، اما به‌طور یقین از آن اطلاع ندارید. برای مثال، ممکن است دستگاهی "صبح بخیر" را بشنود و شروع به ارتباط آن با روشن شدن قهوه‌ساز کند. اگر رایانه این توانایی را داشته باشد، از نظر تئوری می‌تواند "صبح بخیر" را بشنود و تصمیم بگیرد قهوه ساز را روشن کند.

## تفاوت بین هوش مصنوعی قوی و ضعیف

### معنی

هوش مصنوعی قوی شکلی نظری از هوش مصنوعی است و بر این فرض استوار است که ماشین‌ها واقعا می‌توانند هوش و آگاهی انسان را به همان روشی که یک انسان آن را توسعه می‌دهند، توسعه دهند. هوش مصنوعی قوی به ماشینی فرضی اشاره دارد که توانایی‌های شناختی انسان را نشان می‌دهد. از سوی دیگر، هوش مصنوعی ضعیف، نوعی از هوش مصنوعی است که به استفاده از الگوریتم‌های پیشرفته برای انجام وظایف حل مساله یا استدلال خاص که طیف کاملی از توانایی‌های شناختی انسان را در بر نمی‌گیرد، اشاره دارد.

## عملکرد

عملکرد در هوش مصنوعی ضعیف، در مقایسه با هوش مصنوعی قوی محدود است. هوش مصنوعی ضعیف به خودآگاهی دست نمی‌یابد و یا طیف وسیعی از توانایی‌های شناختی انسان را که ممکن است یک انسان داشته باشد نشان نمی‌دهد. هوش مصنوعی ضعیف به سیستم‌هایی اطلاق می‌شود که برای انجام طیف وسیعی از مشکلات برنامه‌ریزی شده‌اند، اما در محدوده عملکردهای از پیش تعیین‌شده یا از پیش تعریف‌شده عمل می‌کنند. از سوی دیگر، هوش مصنوعی قوی به ماشین هایی اطلاق می شود که هوش انسانی را نشان می‌دهند. ایده این است که هوش مصنوعی را تا جایی توسعه دهیم که انسان با ماشین‌هایی که آگاه، هوشمند و با احساسات و خودآگاهی هدایت می‌شوند، تعامل داشته باشد.

## هدف

هدف هوش مصنوعی ضعیف ایجاد فناوری است که به ماشین‌ها و رایانه‌ها اجازه می‌دهد تا وظایف حل مساله یا استدلال خاص را با سرعتی بسیار سریع‌تر از یک انسان انجام دهند. هدف از هوش مصنوعی قوی توسعه هوش مصنوعی تا جایی است که بتوان آن را هوش واقعی انسانی در نظر گرفت. *هوش مصنوعی قوی نوعی است که هنوز به شکل واقعی خود وجود ندارد.*

### خلاصه ای از هوش مصنوعی قوی در مقابل ضعیف

به طور خلاصه، هوش مصنوعی قوی اساسا نوعی از هوش مصنوعی است که آنقدر پیشرفته است که بتوان آن را هوش واقعی دانست. هوش مصنوعی قوی بر این فرض استوار است که یک ماشین محاسباتی با سازماندهی عملکردی مناسب، دارای ذهنی است که مانند ذهن انسان درک می‌کند، فکر می‌کند و هدف دارد. از سوی دیگر، هوش مصنوعی ضعیف به خودآگاهی دست نمی‌یابد یا طیف وسیعی از توانایی‌های شناختی انسان را نشان نمی‌دهد. از این‌رو، برنامه‌های ضعیف هوش مصنوعی را نمی‌توان در نظر گرفت، چراکه آنها واقعا نمی‌توانند مانند انسان‌ها به تنهایی فکر کنند و تصمیم بگیرند.

# یادگیری غیرنظارتی و بهبود راه‌حل‌های یادگیری ماشین

موفقیت‌های اخیر در یادگیری ماشین به دلیل در دسترس بودن داده‌های زیاد، پیشرفت در سخت‌افزار رایانه و پیشرفت‌هایی در الگوریتم‌های یادگیری ماشین بوده است. اما این موفقیت‌ها در مسائل محدود هوش مصنوعی همانند دسته‌بندی تصویر، بینایی رایانه، تشخیص گفتار، پردازش زبان طبیعی و ترجمه ماشین بوده است.

برای حل مسائل هوش مصنوعی بلندپروازانه، باید ارزش یادگیری غیرنظارتی را نشان دهیم. بیایید رایج‌ترین چالش‌هایی را که دانشمندان داده هنگام ساخت راه‌حل‌ها با آن مواجه هستند و اینکه چگونه یادگیری غیرنظارتی می‌تواند به آن‌ها کمک کند را بررسی کنیم.

## داده‌های برچسب‌دار ناکافی

> *"فکر می‌کنم هوش مصنوعی شبیه ساختن یک فضاپیما است. شما به یک موتور بزرگ و سوخت زیادی نیاز دارید. اگر موتور بزرگ و مقدار کمی سوخت داشته باشید، نمی‌توانید به مدار برسید. اگر یک موتور کوچک و یک تن سوخت دارید، حتی نمی‌توانید آن را بلند کنید. برای ساخت یک موشک به یک موتور بزرگ و سوخت زیادی نیاز دارید."*
> *-اندرو انگ[1]*

اگر یادگیری ماشین یک فضاپیما باشد، داده‌ها سوخت آن خواهند بود. بدون داده‌های زیاد، فضاپیما نمی‌تواند پرواز کند. از این‌رو، برای استفاده از الگوریتم‌های یادگیری بانظارت، به داده‌های برچسب‌گذاری‌شده زیادی نیاز داریم که تولید آن سخت و پرهزینه است.

با یادگیری غیرنظارتی می‌توانیم به‌طور خودکار نمونه‌های بدون‌برچسب را برچسب‌گذاری کنیم. نحوه‌ کار به این صورت است که، همه نمونه‌ها را **خوشه‌بندی**[2] می‌کنیم و سپس برچسب‌ها را از نمونه‌های برچسب‌دار به نمونه‌های بدون برچسب در همان خوشه اعمال می‌کنیم. نمونه‌های بدون‌برچسب، برچسبِ نمونه‌های برچسب‌داری را دریافت می‌کنند که بیشترین شباهت را با آن‌ها دارند.

## مشقت بعدچندی (نفرین ابعاد)

در فضایی با ابعاد بسیار بالا، الگوریتم‌های یادگیری بانظارت، به یادگیریِ نحوه‌ی جداسازیِ نقاط، در راستای ایجاد یک تقریب تابع برای تصمیم‌گیری‌های خوب، نیاز دارند. با این حال، وقتی ویژگی‌ها بسیار زیاد باشد، این جستجو هم از منظر زمانی و هم از منظر محاسباتی بسیار هزینه‌بر می‌شود. در برخی موارد، یافتن یک راه حل خوب با سرعت کافی غیرممکن است. این مشکل به عنوان مشقت بعدچندی (نفرین ابعاد) شناخته می‌شود که یادگیری غیرنظارتی برای کمک به مدیریت آن مناسب است. باکاهش ابعاد، می‌توانیم برجسته‌ترین ویژگی‌ها را در مجموعه ویژگی‌های اصلی پیدا کنیم، تعداد ابعاد را به تعداد قابل مدیریت‌تری کاهش دهیم در حالی که اطلاعات مهم بسیار کمی را در این فرآیند از دست می‌دهیم و سپس الگوریتم‌های بانظارت را برای اجرای کارا برای تقریب عملکرد خوب بکار می‌بریم.

## مهندسی ویژگی[3]

مهندسی ویژگی یکی از کلیدی‌ترین وظایفی است که دانشمندان داده انجام می‌دهند. بدون ویژگی‌های مناسب، الگوریتم یادگیری ماشین قادر به جداکردن به اندازه‌یِ کافیِ نقاط در فضا،

---

برای تصمیم‌گیری خوب در مورد نمونه‌های دیده‌نشده، نخواهد بود. با این حال، مهندسی ویژگی معمولا بسیار کارطلب است. چراکه نیاز به انسان دارد تا به‌طور خلاقانه انواع مناسب انواع ویژگی‌ها را مهندسی کند. درعوض، می‌توانیم از یادگیری بازنمایی از الگوریتم‌های یادگیری غیرنظارتی استفاده کنیم تا به طور خودکار انواع مناسب بازنمایی ویژگی‌ها را یاد بگیرند تا به حلِ مساله در دست، کمک کنند.

برای تولید بازنمایی ویژگی‌های جدید، می‌توانیم از یک شبکه عصبی پیش‌خور و غیر بازگشتی برای انجام یادگیری بازنمایی استفاده کنیم؛ جایی که تعداد نرون‌ها در لایه خروجی با تعداد نرون‌های لایه ورودی مطابقت دارد. این شبکه عصبی به عنوان **خودرمزنگار**[1] شناخته می‌شود و به‌طور موثر ویژگی‌های اصلی را بازتولید می‌کند و با استفاده از لایه‌های پنهان بین آن، بازنمایی جدیدی را یاد می‌گیرد. هر لایه پنهان خودرمزنگار بازنمایی از ویژگی‌های اصلی را می‌آموزد و لایه‌های بعدی بر روی بازنمایی که توسط لایه‌های قبلی آموخته شده است، ساخته می‌شوند. لایه به لایه، خودرمزنگار بازنمایی‌های پیچیده‌تری را از نمونه‌های ساده‌تر می‌آموزد. لایه خروجی آخرین بازنمایی تازه آموخته‌شده از ویژگی‌های اصلی است. در نهایت، این بازنمایی آموخته‌شده می‌تواند به عنوان ورودی در یک مدل یادگیری بانظارت با هدفِ بهبودِ خطای تعمیم، استفاده شود.

## یادگیری بازنمایی

عملکرد هر مدل یادگیری ماشین به شدت وابسته به بازنمایی‌هایی است که می‌آموزد تا خروجی را تولید کند. این بازنمایی یادگرفته شده به نوبه خود به‌طور مستقیم به مدل و آنچه که به عنوان ورودی تغذیه می‌شود بستگی دارد. تصور کنید یک مهندس در حال طراحی یک مدل یادگیری ماشین برای پیش‌بینی سلول‌های بدخیم بر اساس اسکن مغز است. برای طراحی مدل، مهندس باید به شدت به داده‌های بیمار تکیه کند، چراکه تمام پاسخ‌ها اینجاست. هر مشاهده یا ویژگی در آن داده، خصیصه‌های بیمار را توصیف می‌کند. مدل یادگیری ماشین که نتیجه را پیش‌بینی می‌کند باید بیاموزد که چگونه هر ویژگی با نتایج مختلف مرتبط است: خوش‌خیم یا بدخیم. بنابراین در صورت وجود هرگونه نویز یا اختلاف در داده‌ها، نتیجه می‌تواند کاملا متفاوت باشد که به مشکل اکثر الگوریتم‌های یادگیری ماشین است. *اکثر الگوریتم‌های یادگیری ماشین درک سطحی از داده‌ها دارند.* از این‌رو راه حل چیست؟ *پاسخ این است، به ماشین‌ها بازنمایی انتزاعی‌تری از داده‌ها ارائه دهید.* با این حال، برای بسیاری از وظایف، غیرممکن است که بدانید چه ویژگی‌هایی باید استخراج شوند. اینجاست که ایده *یادگیری بازنمایی* مطرح می‌شود.

---

[1] autoencoder

یادگیری بازنمایی زیرمجموعهای از یادگیری ماشین است که هدف آن بدست آوردن ویژگیهای خوب و مفید از دادهها بهطور خودکار، بدون آنکه یک مهندس ویژگی درگیر با مساله باشد. در این رویکرد، ماشین دادههای خام را به عنوان ورودی میگیرد و بهطور خودکار بازنماییهای مورد نیاز برای شناسایی ویژگی را کشف میکند و سپس بهطور خودکار ویژگیهای جدید را یاد میگیرد و آن را اعمال میکند. به عبارت دیگر، هدف یادگیری بازنمایی یافتن تبدیلی است که دادههای خام را به بازنمایی که برای یک وظیفه یادگیری ماشین مناسبتر است (به عنوان مثال دستهبندی) نگاشت میکند. از آنجا که این روش میتواند به عنوان یادگیری ویژگیهای معنادار تفسیر شود، به آن یادگیری ویژگی نیز گفته میشود.

اساسا، یادگیری بازنمایی چیزی نیست جز مجموعهای از ویژگیها که مفاهیم را به صورت جداگانه توصیف میکند. به عنوان مثال، میتوانیم اشیا را با استفاده از رنگها، شکل، اندازه و ویژگیهایشان بازنمایی کنیم. بازنمایی چیزی است که به ما کمک میکند بین مفاهیم مختلف تمایز قائل شویم و به نوبه خود به ماکمک میکند تا شباهتهای بین آنها را پیدا کنیم.

یادگیری بازنمایی دادههای با ابعاد بالا را به دادههای کمبعد کاهش میدهد، یافتن الگوها و ناهنجاریها را آسانتر میکند و همچنین به ما درک بهتری از رفتار دادهها میدهد. با کاهش پیچیدگی دادهها، ناهنجاریها و نویز کاهش مییابد. این کاهش نویز میتواند برای الگوریتمهای یادگیری بانظارت بسیار مفید باشد.

## نقاط دورافتاده

همچنانکه بیشتر بیان شد، کیفیت دادهها بسیار مهم است. اگر الگوریتمهای یادگیری ماشین بر روی نقاط دورافتادهٔ نادر و تحریفکننده آموزش ببینند، تعمیمدهی آنها کمتر از زمانی است که آنها را بهطور جداگانه نادیده میگیرد. با یادگیری غیرنظارتی، میتوانیم تشخیص نقاط دورافتاده را با استفاده از کاهش ابعاد انجام دهیم و یک راهحل برای دادههای هنجار برای ایجاد کنیم.

## رانش داده[۱]

فرض اساسی در توسعه هر مدل یادگیری ماشین این است که دادههایی که برای آموزش مدل استفاده میشوند، از دادههای دنیای واقعی تقلید میکنند. اما چگونه میتوان این فرض را پس از استقرار مدل در تولید، تایید کرد؟ وقتی مدلی را با رهیافت یادگیری بانظارت آموزش میدهید، دادههای آموزشی دارای برچسب هستند، زمانی که مدل را در تولید (عملآوری) مستقر میکنید، هیچ برچسب واقعی وجود ندارد، مهم نیست که مدل شما چقدر دقیق باشد، پیشبینیها تنها در

---

[۱] Data drift

صورتی درست هستند که داده‌های ارائه‌شده به مدل در تولید، داده‌های مورد استفاده در آموزش را تقلید کنند (یا به لحاظ آماری معادل باشند). اگر این کار را نکند چه؟ در این صورت آن را رانش داده می‌نامیم.

مدل‌های یادگیری ماشین باید از رانش داده‌ها آگاه باشند. اگر داده‌هایی که مدل روی آن‌ها پیش‌بینی می‌کند از نظر آماری با داده‌هایی که مدل آموزش داده شده متفاوت است، ممکن است مدل نیاز به آموزش مجدد بر روی داده‌هایی داشته باشد که بیشتر مُعرف داده‌های فعلی هستند. اگر مدل دوباره آموزش پیدا نکند یا رانش را تشخیص ندهد، کیفیت پیش‌بینی مدل بر روی داده‌های فعلی متضرر می‌شود. با ساختن توزیع‌های احتمال با استفاده از یادگیری غیرنظارتی، می‌توانیم ارزیابی کنیم که داده‌های فعلی چقدر با داده‌های مجموعه آموزشی متفاوت هستند. اگر این دو به تفاوت بسیار زیادی با یکدیگر داشته باشند، می‌توانیم به طور خودکار یک **بازآموزی**[1] را آغاز کنیم.

## خوشه‌بندی

خوشه‌بندی یا تحلیل خوشه‌ای وظیفه گروه‌بندی مجموعه‌ای از اشیاء است به گونه‌ای که اشیاء در یک گروه (به نام خوشه) نسبت به سایر گروه‌ها (خوشه‌ها) شباهت بیشتری به یکدیگر داشته باشند. به عبارت دقیق‌تر، هدف تحلیل خوشه‌ای (در حالت ایده‌آل) یافتن خوشه‌هایی است که نمونه‌های درون هر خوشه کاملا شبیه یکدیگر باشند، در حالی که هر خوشه‌ای با یکدیگر کاملا متفاوت باشد. خوشه‌ها چیزی نیستند جز گروه‌بندی نقاط داده به گونه‌ای که فاصله بین نقاط داده درون خوشه‌ای حداقل باشد. به عبارت دیگر، خوشه‌ها مناطقی هستند که تراکم نقاط داده مشابه در آن‌ها زیاد است. به‌طور کلی، خوشه‌بندی برای تجزیه و تحلیل مجموعه داده‌ها، در جهت یافتن **داده‌های بینش‌افزا**[2] و استنتاج از آن استفاده می‌شود. همچنین، استنتاج‌هایی که باید از مجموعه داده‌ها گرفته شود به کاربر بستگی دارد، *چراکه هیچ معیاری تعریف‌شده عمومی برای خوشه‌بندی خوب وجود ندارد.*

برای انجام این تجزیه و تحلیل، الگوریتم‌های خوشه‌بندی، داده‌ها را دریافت می‌کنند و با استفاده از برخی از معیارهای تشابه، این گروه‌ها (خوشه‌ها) را تشکیل می‌دهند. برای رسیدن به یک گروه‌بندی موفق، الگوریتم خوشه‌بندی باید به دو هدف اصلی دست یابد: (۱)، شباهت بین یک نقطه داده با نقطه دیگر و (۲)، تمایز این نقاط داده مشابه با سایر نقاط که قطعا از نظر اکتشافی با آن‌ها تفاوت دارند. از این‌رو، در فرآیند خوشه‌بندی، معیار تشابه مبتنی‌بر فاصله، نقش مهمی را در تصمیم‌گیری خوشه‌بندی بازی می‌کند.

---

[1] retraining

[2] insightful data

با استفاده از دو هدف بالا، شما باید یک الگوریتم خوشهبندی را انتخاب کنید که به احتمال زیاد بهترین نتیجه را در مساله خاص خود بدست میآورد. برخی از الگوریتمهای خوشهبندی تعاریف متفاوتی در مورد این که خوشه چیست، دارند. به عنوان مثال، چندین الگوریتم ممکن است خوشه را به عنوان گروهی از دادهها با شکلهای هندسی خاص تعریف کنند. به عنوان نمونه، k-Means که خوشه را به عنوان گروهی از اشیا با یک شکل کروی تعریف میکند. در عین حال، گروهی دیگر از الگوریتمها ممکن است فرض کنند که خوشهها بهطور متراکم در فضای داده قرار گرفتهاند، به عنوان مثال dbscan. علاوه بر اینها، تعاریف خوشه بیشتری نیز وجود دارد. بر این اساس، کیفیت نتیجه، بستگی به مناسب بودن بین تعریف خوشهی فرض شده الگوریتمها و ساختار خوشهای ذاتی دادهها دارد. از اینرو، هنگام تلاش برای انجام این تحلیل، بهتر است که دانش قبلی درباره مناسبترین تعریف خوشه برای مساله خود داشته باشید. در عین حال، هنگامی که فرایند خوشهبندی را شروع میکنید، این دانش ممکن است در دسترس نباشد. این معمولا تجزیه و تحلیل را پیچیده میکند.

## خوشهبندی و انواع آن

بهطور رسمی، فرض کنید یک مجموعه داده $D = \{x_1, x_2, ..., x_m\}$ حاوی $m$ نمونه بدون برچسب داشته باشیم، که در آن هر نمونه $x_i = (x_{i1}; x_{i2}; ..., x_{in})$ یک بردار $n$ بعدی است. بر این اساس، یک الگوریتم خوشهبندی، مجموعه دادههای $D$ را به $k$ خوشهی مجزا
$$\{C_l | l = 1, 2, ..., k\}$$
تقسیم میکند، که در آن، $C_i \cap_{i \neq l} C_l = \emptyset$ و $D = \cup_{l=1}^k C_l$. از اینرو، $\lambda_j \epsilon \{1, 2, ..., k\}$ را به عنوان برچسب خوشه نمونه $x_j$ نشان میدهیم. در نهایت، نتیجه خوشهبندی را میتوان بهعنوان بردار برچسب خوشه $\lambda = (\lambda_1; \lambda_2; ... ; \lambda_m)$ با $m$ عناصر نشان داد.

از نظر مفهومی، خوشهبندی را میتوان مشابه دستهبندی در نظر گرفت، به این معنا که سعی میکند یک مقدار گسسته را به هر نمونه اختصاص دهد. تنها تفاوت در این است، در حالی که دستهبندی از نمونههای برچسبگذاریشده برای یادگیری الگوهایی در دادهها استفاده میکند تا کلاسها را از هم جدا کند، خوشهبندی هیچ دانش قبلی در مورد عضویت کلاسها و یا اینکه

آیا کلاس‌های مجزایی در داده‌ها وجود دارد یا خیر، ندارد. از این‌رو، خوشه‌بندی مجموعه‌ای از الگوریتم‌ها را توصیف می‌کند که سعی می‌کنند ساختار گروه‌بندی را در یک مجموعه داده شناسایی کنند. به‌طور خلاصه، **هدف از خوشه‌بندی توصیفی و هدف از دسته‌بندی پیش‌بینی است.** بسته به استراتژی یادگیری مورد استفاده، الگوریتم‌های خوشه‌بندی را می‌توان به چندین دسته تقسیم کرد:

- خوشه‌بندی مبتنی‌بر نمونه اولیه
- خوشه‌بندی مبتنی‌بر چگالی
- خوشه‌بندی سلسله‌مراتبی

که در ادامه این بخش به تشریح آن‌ها پرداخته‌ایم. با این حال، قبل از آن، اجازه دهید ابتدا دو مشکل اساسی مربوط به خوشه‌بندی را مورد بحث قرار دهیم: **ارزیابی عملکرد** و **محاسبه فاصله.**

## ارزیابی عملکرد (شاخص‌های اعتبار[1])

معیارهای ارزیابی برای خوشه‌بندی، شاخص‌های اعتبار نیز نامیده می‌شوند. از آنجایی که نتیجه دسته‌بندی توسط معیارهای کارآیی در یادگیری بانظارت ارزیابی می‌شود، نتیجه خوشه‌بندی نیز باید از طریق برخی شاخص‌های اعتبار ارزیابی شود. قبل از آنکه بیشتر به این موضوع بپردازیم باید گفت، ارزیابی اینکه آیا یک خوشه‌بندی خاص خوب است یا خیر، موضوعی مشکل‌ساز و بحث‌برانگیز است. در واقع بونر (۱۹۶۴) اولین کسی بود که استدلال کرد هیچ تعریف عمومی برای خوشه‌بندی خوب وجود ندارد. ارزیابی بیشتر در چشم بیننده است. با این وجود، چندین معیار ارزیابی توسعه داده شده است. این معیارها معمولا به دو دسته **شاخص اعتبار داخلی و خارجی** تقسیم می‌شوند. شاخص خارجی نتیجه خوشه‌بندی را با یک مدل مرجع مقایسه می‌کند، در حالی‌که، شاخص داخلی نتیجه خوشه‌بندی را بدون استفاده از هیچ مدل مرجع ارزیابی می‌کند.

با توجه به مجموعه داده $D = \{x_1, x_2, \ldots, x_m\}$، فرض کنید یک الگوریتم خوشه‌بندی، خوشه‌های $c = \{C, C_2, \ldots, C_k\}$ را تولید می‌کند و یک مدل مرجع خوشه‌های $c^* = \{C^*, C^*_2, \ldots, C^*_k\}$ را ارائه می‌دهد. بر این اساس، اجازه دهید $\lambda$ و $\lambda^*$ به ترتیب بردارهای برچسب خوشه‌بندی $c$ و $c^*$ را نشان دهند. از این‌رو، برای هر جفت نمونه می‌توانیم چهار عبارت زیر را تعریف کنیم:

$$a = |SS| , SS = \{(x_i, x_j) \mid \lambda_i = \lambda_j, \lambda^*_i = \lambda^*_j, i < j\}$$

$$b = |SD| , SD = \{(x_i, x_j) \mid \lambda_i = \lambda_j, \lambda^*_i \neq \lambda^*_j, i < j\}$$

---

[1] validity indices

$$c = |DS|, DS = \{(x_i, x_j) | \lambda_i \neq \lambda_j, \lambda_i^* = \lambda_j^*, i < j\}$$

$$d = |DD|, DD = \{(x_i, x_j) | \lambda_i \neq \lambda_j, \lambda_i^* \neq \lambda_j^*, i < j\}$$

که در آن مجموعه $SS$ شامل جفت نمونه‌هایی است که هر دو نمونه به یک خوشه در $c$ و همچنین به یک خوشه در $c^*$ تعلق دارند. مجموعه $SD$ شامل جفت نمونه‌هایی است که هر دو نمونه به یک خوشه در $c$ تعلق دارند اما در $c^*$ نیستند. مجموعه‌های $DS$ و $DD$ را می‌توان به‌طور مشابه تفسیر کرد. از آنجایی که هر جفت نمونه $(x_i, x_j)$ $(i < j)$ فقط می‌تواند در یک مجموعه ظاهر شود، داریم: $a + b + c + d = \frac{m(m-1)}{2}$.

با توجه به چهار عبارت تعریف شده بالا، برخی از شاخص‌های خارجی رایج را می‌توان به صورت زیر تعریف کرد:

- **ضریب جاکارد[1] (JS):**

$$JS = \frac{a}{a+b+c}$$

- **شاخص فاولکس و مالوز[2] (FMI):**

$$FMI = \sqrt{\frac{a}{a+b} \cdot \frac{a}{a+c}}$$

- **شاخص رند[3] (RI):**

$$RI = \frac{2(a+d)}{m(m-1)}$$

شاخص‌های اعتبار خارجی فوق مقادیری را در بازه [۰، ۱] می‌گیرند. مقدار شاخص بزرگتر نشان‌دهنده کیفیت خوشه‌بندی بهتر است.

شاخص‌های اعتبار داخلی کیفیت خوشه‌بندی را بدون استفاده از مدل مرجع ارزیابی می‌کنند. با توجه به خوشه‌های تولید شده $c = \{C, C_2, \ldots, C_k\}$، می‌توانیم چهار اصطلاح زیر را تعریف کنیم:

$$avg(C) = \frac{2}{|C|(|C|-1)} \sum_{1 \leq i < j \leq |C|} dist(x_i, x_j)$$

$$diam(C) = max_{1 \leq i < j \leq |C|} dist(x_i, x_j)$$

$$d_{min}(C_i, C_j) = min_{x_i \in C_i, x_j \in C_j} dist(x_i, x_j)$$

---

[1] Jaccard Coefficient

[2] Fowlkes and Mallows Index

[3] Rand Index

$$d_{cen}(C_i, C_j) = dist(\mu_i, \mu_j)$$

جایی که $\mu = \frac{1}{|C|}\sum_{1 \le i \le |C|} x_i$ مرکز خوشه $C$ را نشان می‌دهد و $dist(.,.)$ فاصله بین دو نمونه را اندازه می‌گیرد. در اینجا، $avg(C)$ میانگین فاصله بین نمونه‌ها در خوشه $C$ است. $diam(C)$ بزرگترین فاصله بین نمونه‌ها در خوشه $C$ است. $d_{min}(C_i, C_j)$ فاصله بین دو نزدیکترین نمونه در خوشه‌های $C_i$ و $C_j$ و $d_{cen}(C_i, C_j)$ فاصله بین مرکز خوشه‌های $C_i$ و $C_j$ است.

با توجه به اصطلاحات تعریف شده بالا، برخی از شاخص‌های اعتبار داخلی رایج را می‌توان به صورت زیر تعریف کرد:

• **شاخص دیویس‌ـ‌بولدین[1] (DBI):**

$$DBI = \frac{1}{k}\sum_{i=1}^{k} max_{i \neq j}\left(\frac{avg(C_i) + avg(C_j)}{d_{cen}(C_i, C_j)}\right).$$

• **شاخص دان[2] (DI):**

$$DI = min_{1 \le i \le k}\left\{min_{i \neq j}\left(\frac{d_{min}(C_i, C_j)}{max_{1 \le l \le k} diam(C_l)}\right)\right\}$$

مقدار کمتر $DBI$ نشان‌دهنده کیفیت خوشه‌بندی بهتر، در مقابل، مقدار بزرگتر $DI$ نشان‌دهنده کیفیت خوشه‌بندی بهتر است.

## محاسبه فاصله و محاسبه تشابه

از آنجایی که خوشه‌بندی گروه‌بندی نمونه‌ها/اشیاء مشابه است، به نوعی معیاری نیاز است که بتواند تعیین کند، آیا دو شی مشابه یا غیرمشابه هستند. دو نوع معیار اصلی برای تخمین این رابطه مورد استفاده قرار می‌گیرد: **معیار فاصله** و **معیار تشابه.**

بسیاری از روش‌های خوشه‌بندی از معیارهای فاصله برای تعیین شباهت یا عدم شباهت بین هر جفت شی استفاده می‌کنند. یک معیار فاصله $dist(.,.)$ در صورتی که ویژگی‌های زیر را برآورده کند، معیار فاصله متریک نامیده می‌شود:

۱. **غیرمنفی[3]:** $dist(x_i, x_j) \geq \cdot$

۲. **انعکاسی[4]:** $dist(x_i, x_j) = \cdot \Leftrightarrow x_i = x_j$

---

۳. **تقارن**[1] (جابجایی): $dist(x_i, x_j) = dist(x_j, x_i)$

۴. **نابرابری مثلثی**[2]: $dist(x_i, x_j) \leq dist(x_i, x_k) + dist(x_k, x_j)$

ما اغلب معیارهای تشابه را از طریق برخی از انواع فاصله‌ها تعریف می‌کنیم و هر چه فاصله بیشتر باشد، شباهت کم‌تر است.

### مینکوفسکی: معیار فاصله برای ویژگی‌های عددی

با توجه به دو نمونه $n$-بعدی $x_i = (x_{i1}; x_{i2}; \dots, x_{in})$ و $x_j = (x_{j1}; x_{j2}; \dots, x_{jn})$، فاصله بین دو نمونه با استفاده از متریک مینکوفسکی به‌صورت زیر بدست آورد:

$$dist(x_i, x_j) = \left(\left|x_{i1} - x_{j1}\right|^g + \left|x_{i2} - x_{j2}\right|^g + \dots + \left|x_{ip} - x_{jp}\right|^g\right)^{\frac{1}{g}}$$

توجه داشته باشید که اگر $g = 2$، ۱ و ∞ باشد، آن‌گاه به ترتیب فاصله اقلیدسی، فاصله منهتن و حداکثر فاصله را بدست می‌آوریم.

واحدِ سنجشِ استفاده شده می‌تواند بر تحلیل خوشه‌بندی تاثیر بگذارد. برای اجتناب از وابستگیِ به انتخابِ واحد سنجش، داده‌ها باید همسان‌سازی (برسنجیده) شوند. معیار همسان‌سازی‌شده تلاش می‌کند تا به همه متغیرها وزن یکسانی بدهد. با این حال، اگر به هر متغیر با توجه به اهمیتش وزنی نسبت داده شود، فاصله وزنی را می توان به صورت زیر محاسبه کرد:

$$dist(x_i, x_j) = \left(w_1\left|x_{i1} - x_{j1}\right|^g + w_2\left|x_{i2} - x_{j2}\right|^g + \dots + w_i\left|x_{ip} - x_{jp}\right|^g\right)^{\frac{1}{g}}$$

جایی که $w_i \epsilon [0, \infty)$ است.

### معیار فاصله برای ویژگی‌های دودویی

در مورد ویژگی‌های دودویی، فاصله بین اشیاء ممکن است بر اساس **جدول پیشایندی**[3] محاسبه شود. یک ویژگی دودویی در صورتی متقارن است که هر دو حالت آن، یک ارزش داشته باشد. در این صورت، با استفاده از **ضریب تطبیق ساده**[4] می‌توان عدم تشابه بین دو شی را ارزیابی کرد:

$$dist(x_i, x_j) = \frac{r + s}{q + r + s + t}$$

---

که در آن $q$ تعداد ویژگی‌هایی است که برای هر دو شیء برابر با ۱، ۱ $t$ تعداد ویژگی‌هایی است که برای هر دو شیء برابر ۰ و $s$ و $r$ تعداد ویژگی‌هایی است که برای هر دو شی نابرابر هستند.

**معیار فاصله برای ویژگی‌های اسمی**

هنگامی که ویژگی‌ها اسمی هستند، دو رویکرد اصلی ممکن است استفاده شود:

۱. تطبیق ساده:

$$dist(x_i, x_j) = \frac{p - m}{p}$$

که در آن $p$ تعداد کل ویژگی‌ها و $m$ تعداد تطابق‌ها است.

۲. ایجاد یک ویژگی باینری برای هر حالت از هر ویژگی اسمی و محاسبه عدم تشابه آن‌ها.

# خوشه‌بندی مبتنی‌بر نمونه اولیه[۱]

**خوشه‌بندی نمونه اولیه**، که به عنوان **خوشه‌بندی مبتنی‌بر نمونه‌های اولیه** نیز شناخته می‌شود، خانواده‌ای از الگوریتم‌های خوشه‌بندی است که فرض می‌کند ساختار خوشه‌بندی را می‌توان با مجموعه‌ای از نمونه‌های اولیه نشان داد. بطور معمول، چنین الگوریتم‌هایی با برخی از نمونه‌های اولیه شروع می‌شوند و سپس بطور مکرر نمونه‌های اولیه را بروز و بهینه می‌کنند. الگوریتم‌های زیادی با استفاده از رویکرد نمونه اولیه و روش‌های بهینه‌سازی مختلف توسعه داده شده‌اند. با این حال، در ادامه این بخش، تنها دو مورد از الگوریتم‌های خوشه‌بندی مبتنی‌بر نمونه اولیه را مورد بحث قرار خواهیم داد، یعنی **خوشه‌بندی k-Means و خوشه‌بندی مخلوط گاوسی**[۲].

## خوشه‌بندی k-Means

تکنیک خوشه‌بندی K-means ساده است و ما با توضیح الگوریتم اصلی شروع می‌کنیم. ابتدا K مرکز اولیه را انتخاب می‌کنیم، جایی که K یک پارامتر مشخص‌شده توسط کاربر، یعنی تعداد خوشه‌های مورد نظر است. سپس هر نقطه به نزدیک‌ترین مرکز تخصیص داده می‌شود و هر مجموعه‌ای از نقاط اختصاص داده شده به مرکز، یک خوشه است. سپس مرکز هر خوشه بر اساس نقاط اختصاص داده شده به خوشه بروز می‌شود. ما مراحل تخصیص و بروزرسانی را تکرار می‌کنیم تا زمانی که هیچ نقطه‌ای در خوشه‌ها تغییر نکند، به عبارت دیگر، تا زمانی که مرکزها ثابت بمانند.

---

[۱] Prototype clustering

[۲] Mixture-of-Gaussian

داده‌هایی را در نظر بگیرید که معیار تشابه آن‌ها فاصله اقلیدسی است. برای تابع هدف که کیفیت یک خوشه‌بندی را می‌سنجد، از **کمینه‌سازی خطای مربع**[1] استفاده می‌کنیم. به عبارت دیگر، ابتدا محاسبه خطای هر نقطه داده (یعنی فاصله اقلیدسی آن تا نزدیکترین مرکز) و سپس مجموع مربع خطاها را محاسبه می‌کنیم. با توجه به دو مجموعه متفاوت از خوشه‌ها که توسط دو اجرای مختلف K-means تولید می‌شوند، ما یکی را ترجیح می‌دهیم که کمترین مربع خطا را دارد. زیرا بدان معناست که نمونه‌های اولیه‌یِ (مرکز) این خوشه‌بندی، نمایش بهتری از نقاط در خوشه خود هستند.

به طور رسمی، با توجه به مجموعه داده $D = \{x_1, x_2, \dots, x_m\}$، الگوریتم k-means خطای مربع خوشه‌های $c = \{C, C_2, \dots, C_k\}$ را کمینه می‌کند:

$$E = \sum_{i=1}^{k} \sum_{x \in C_i} \|x - \mu_i\|_2^2$$

جایی که $\mu_i = \frac{1}{|C_i|} \sum_{x \in C_i} x$ بردار میانگین خوشه $C_i$ است. به طور شهودی، معادله فوق، **نزدیکیِ**[2] (**همسایگی**) بین میانگین بردار یک خوشه و نمونه‌های درون آن خوشه را نشان می‌دهد، که در آن E کوچکتر، نشان‌دهنده شباهت درون خوشه‌ای بالاتر است. با این وجود، کمینه‌سازی E آسان نیست، چراکه نیاز به ارزیابی تمام افرازهای ممکن مجموعه داده D دارد، که در واقع یک **مساله NP-hard** است. از این رو، الگوریتم k-means یک استراتژی حریصانه اتخاذ می‌کند و یک روش بهینه‌سازی تکراری را برای یافتن راه حل تقریبی اتخاذ می‌کند. در این الگوریتم، ابتدا، بردارهای میانگین را مقداردهی اولیه می‌کند و به ترتیب خوشه‌ها و بردارهای میانگین را به طور مکرر بروز می‌کند. هنگامی که خوشه‌ها پس از یک بار تکرار تغییر نمی‌کنند، خوشه‌های فعلی برگردانده می‌شوند.

در ادامه برای درک بهتر یک مثال را تشرح می‌کنیم. با این حال، ما الگوریتم را در حالتی نشان می‌دهیم که فقط دو متغیر وجود داشته باشد تا نقاط داده و مراکز خوشه‌ای را بتوان به‌صورت هندسی با نقاطی در یک صفحه مختصات نشان داد. فاصله بین نقاط $(x_1, x_2)$ و $(y_1, y_2)$ را با استفاده از فرمول فاصله اقلیدسی محاسبه می‌کنیم:

$$\|\vec{x} - \vec{y}\| = \sqrt{(x_1 - y_1)^2 + (x_2 - y_2)^2}$$

**مثال.** می‌خواهیم با استفاده از الگوریتم نقاط داده جدول زیر را به ۲ خوشه افراز کنیم:

| $x_1$ | ۱ | ۲ | ۲ | ۳ | ۴ | ۵ |
|-------|---|---|---|---|---|---|
| $x_2$ | ۱ | ۱ | ۳ | ۲ | ۳ | ۵ |

---

[1] minimizes the squared error

[2] closeness

نمودار پراکندگی داده‌ها در شکل زیر قابل مشاهده است:

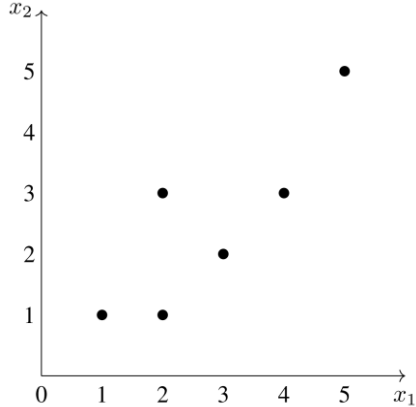

حال به صورت گام به گام به تشریح الگوریتم می‌پردازیم:

۱.   در مساله، تعداد خوشه‌های مورد نیاز ۲ است. از این‌رو، $k = 2$ را قرار می‌دهیم.

۲.   دو نقطه را به صورت دلخواه به عنوان مراکز خوشه اولیه انتخاب می‌کنیم. اجازه دهید به طور دلخواه $\vec{v}_1 = (2,1)$ و $\vec{v}_2 = (2,3)$ را انتخاب کنیم.

۳.   فاصله نقاط دادهیِ داده شده را از مراکز خوشه محاسبه می‌کنیم:

| $\vec{x}_i$ | نقاط داده | فاصله از $\vec{v}_1 = (2,1)$ | فاصله از $\vec{v}_2 = (2,3)$ | کم‌ترین فاصله | مرکز اختصاص داده شده |
|---|---|---|---|---|---|
| $\vec{x}_1$ | (۱,۱) | ۱ | ۲.۲۴ | ۱ | $\vec{v}_1$ |
| $\vec{x}_2$ | (۲,۱) | ۰ | ۲ | ۰ | $\vec{v}_1$ |
| $\vec{x}_3$ | (۲,۳) | ۲ | ۰ | ۰ | $\vec{v}_2$ |
| $\vec{x}_4$ | (۳,۲) | ۱.۴۱ | ۱.۴۱ | ۰ | $\vec{v}_1$ |
| $\vec{x}_5$ | (۴,۳) | ۲.۸۲ | ۳ | ۲ | $\vec{v}_2$ |
| $\vec{x}_6$ | (۵,۵) | ۵ | ۳.۶۱ | ۳.۶۱ | $\vec{v}_2$ |

(فاصله $\vec{x}_4$ از $\vec{v}_1$ و $\vec{v}_2$ برابر است. ما به صورت دلخواه $\vec{v}_1$ را به $\vec{x}_4$ اختصاص داده‌ایم) بر این اساس، داده‌ها به دو خوشه به شکل زیر تشکیل می‌شوند:

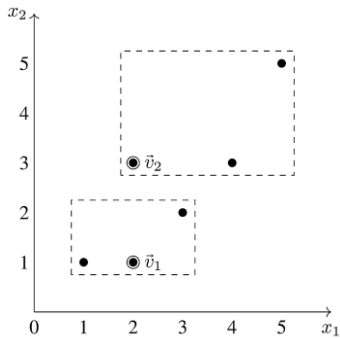

خوشه ۱: $\{\vec{x}_\text{۱}, \vec{x}_\text{۲}, \vec{x}_\text{٤}\}$ به $\vec{v}_\text{۱}$ اختصاص داده شده است.

تعداد نقاط داده در خوشه ۱: $C_\text{۱} = \text{۳}$

خوشه ۲: $\{\vec{x}_\text{۳}, \vec{x}_\text{٥}, \vec{x}_\text{٦}\}$ به $\vec{v}_\text{۲}$ اختصاص داده شده است.

تعداد نقاط داده در خوشه ۲: $C_\text{۲} = \text{۳}$

۴. مراکز خوشه به شرح زیر دوباره محاسبه می‌شوند:

$$\vec{v}_\text{۱} = \frac{\text{۱}}{C_\text{۱}}(\vec{x}_\text{۱} + \vec{x}_\text{۲} + \vec{x}_\text{٤})$$

$$= \frac{\text{۱}}{\text{۳}}(\vec{x}_\text{۱} + \vec{x}_\text{۲} + \vec{x}_\text{٤})$$

$$= (\text{۲٫۰۰}, \text{۱٫۳۳})$$

$$\vec{v}_\text{۲} = \frac{\text{۱}}{C_\text{۲}}(\vec{x}_\text{۳} + \vec{x}_\text{٥} + \vec{x}_\text{٦})$$

$$= \frac{\text{۱}}{\text{۳}}(\vec{x}_\text{۳} + \vec{x}_\text{٥} + \vec{x}_\text{٦})$$

$$= (\text{۳٫٦۷}, \text{۳٫٦۷})$$

۵. فاصله نقاط دادهٍ داده شده را از مراکز خوشه جدید محاسبه می‌کنیم:

| $\vec{x}_i$ | نقاط داده | فاصله از $\vec{v}_\text{۱} = (2, 1.33)$ | فاصله از $\vec{v}_\text{۲} =$ (3.67, 3.67) | کمترین فاصله | مرکز اختصاص داده شده |
|---|---|---|---|---|---|
| $\vec{x}_\text{۱}$ | (۱٬۱) | ۱٫۰٥ | ۳٫۷۷ | ۱٫۰٥ | $\vec{v}_\text{۱}$ |
| $\vec{x}_\text{۲}$ | (۲٬۱) | ۰٫۳۳ | ۳٫۱٤ | ۰٫۳۳ | $\vec{v}_\text{۱}$ |
| $\vec{x}_\text{۳}$ | (۲٬۳) | ۱٫٦۷ | ۱٫۸۰ | ۱٫٦۷ | $\vec{v}_\text{۱}$ |
| $\vec{x}_\text{٤}$ | (۳٬۲) | ۱٫۲۰ | ۱٫۸۰ | ۱٫۲۰ | $\vec{v}_\text{۱}$ |
| $\vec{x}_\text{٥}$ | (٤٬۳) | ۲٫٦۰ | ۰٫۷٥ | ۰٫۷٥ | $\vec{v}_\text{۲}$ |
| $\vec{x}_\text{٦}$ | (٥٬٥) | ٤٫۷٤ | ۱٫۸۹ | ۱٫۸۹ | $\vec{v}_\text{۲}$ |

بر این اساس، داده‌ها به دو خوشه به شکل زیر تشکیل می‌شوند:

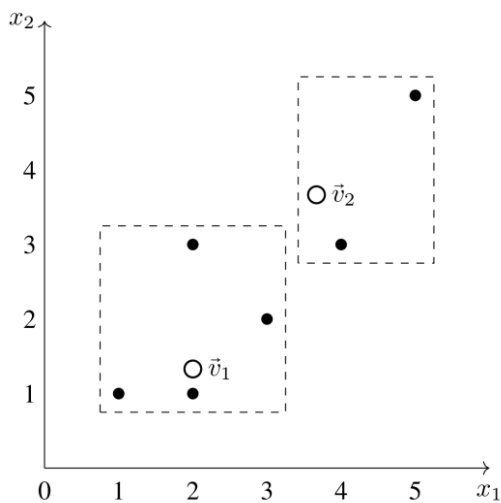

خوشه ۱: $\{\vec{x}_\text{۱}, \vec{x}_\text{۲}, \vec{x}_\text{۳}, \vec{x}_\text{٤}\}$ به $\vec{v}_\text{۱}$ اختصاص داده شده است.

تعداد نقاط داده در خوشه ۱: $C_\text{۱} = ٤$

خوشه ۲: $\{\vec{x}_\text{٥}, \vec{x}_\text{٦}\}$ به $\vec{v}_\text{۲}$ اختصاص داده شده است.

تعداد نقاط داده در خوشه ۲: $C_\text{۲} = ۲$

۶.   مراکز خوشه به شرح زیر دوباره محاسبه می‌شوند:

$$\vec{v}_\text{۱} = \frac{۱}{C_\text{۱}}(\vec{x}_\text{۱} + \vec{x}_\text{۲} + \vec{x}_\text{۳} + \vec{x}_\text{٤})$$

$$= \frac{۱}{٤}(\vec{x}_\text{۱} + \vec{x}_\text{۲} + \vec{x}_\text{۳} + \vec{x}_\text{٤})$$

$$= (۲.۰۰, ۱.۷۵)$$

$$\vec{v}_\text{۲} = \frac{۱}{C_\text{۲}}(\vec{x}_\text{٥} + \vec{x}_\text{٦})$$

$$= \frac{۱}{۲}(\vec{x}_\text{٥} + \vec{x}_\text{٦})$$

$$= (٤.٥, ٤)$$

۷.   فاصله نقاط داده‌ی داده شده را از مراکزِ خوشه جدید محاسبه می‌کنیم:

| $\vec{x}_i$ | نقاط داده | فاصله از $\vec{v}_1$ | فاصله از $\vec{v}_2$ | کم‌ترین فاصله | مرکز اختصاص داده شده |
|---|---|---|---|---|---|
| $\vec{x}_\text{۱}$ | (۱,۱) | ۱.۲۵ | ۴.۶۱ | ۱.۲۵ | $\vec{v}_\text{۱}$ |
| $\vec{x}_\text{۲}$ | (۲,۱) | ۰.۷۵ | ۳.۹۱ | ۰.۷۵ | $\vec{v}_\text{۱}$ |
| $\vec{x}_\text{۳}$ | (۲,۳) | ۱.۲۵ | ۲.۶۹ | ۱.۲۵ | $\vec{v}_\text{۱}$ |
| $\vec{x}_\text{٤}$ | (۳,۲) | ۱.۰۳ | ۲.۵۰ | ۱.۰۳ | $\vec{v}_\text{۱}$ |
| $\vec{x}_\text{٥}$ | (٤,۳) | ۲.۳۶ | ۱.۱۲ | ۱.۱۲ | $\vec{v}_\text{۲}$ |
| $\vec{x}_\text{٦}$ | (۵,۵) | ۴.۴۲ | ۱.۱۲ | ۱.۱۲ | $\vec{v}_\text{۲}$ |

بر این اساس، داده‌ها به دو خوشه به شکل زیر تشکیل می‌شوند:

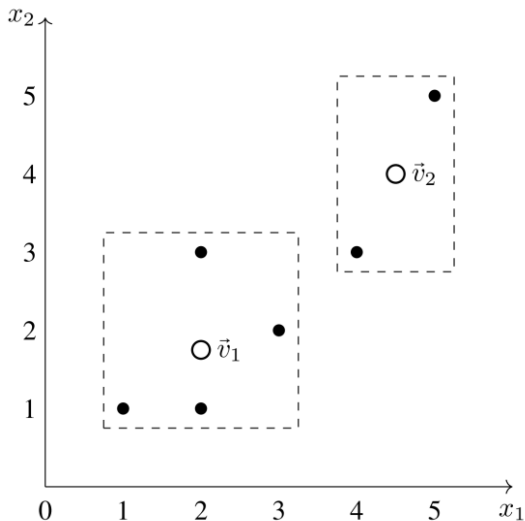

خوشه ۱: $\{\vec{x}_1, \vec{x}_2, \vec{x}_3, \vec{x}_4\}$ به $\vec{v}_1$ اختصاص داده شده است.

تعداد نقاط داده در خوشه ۱: $C_1 = 4$

خوشه ۲: $\{\vec{x}_5, \vec{x}_6\}$ به $\vec{v}_2$ اختصاص داده شده است.

تعداد نقاط داده در خوشه ۲: $C_2 = 2$

۸. مراکز خوشه به شرح زیر دوباره محاسبه می‌شوند:

$$\vec{v}_1 = \frac{1}{C_1}(\vec{x}_1 + \vec{x}_2 + \vec{x}_3 + \vec{x}_4)$$

$$= \frac{1}{4}(\vec{x}_1 + \vec{x}_2 + \vec{x}_3 + \vec{x}_4)$$

$$= (2.00, 1.75)$$

$$\vec{v}_2 = \frac{1}{C_2}(\vec{x}_5 + \vec{x}_6)$$

$$= \frac{1}{2}(\vec{x}_5 + \vec{x}_6)$$

$$= (4.5, 4)$$

۹. از آنجایی که اینها با مراکز خوشه محاسبه شده پیشین یکسان هستند، بنابراین هیچ گونه تخصیص مجدد نقاط داده به خوشه‌های دیگر وجود نخواهد داشت و بنابراین محاسبات در اینجا متوقف می‌شوند.

۱۰. در نتیجه، خوشه‌ها با مراکز زیر به‌صورت زیر بدست می‌آیند:

خوشه ۱: $\{\vec{x}_1, \vec{x}_2, \vec{x}_3, \vec{x}_4\}$ به $\vec{v}_1 = (2, 1.75)$ اختصاص داده شده است.

خوشه ۲: $\{\vec{x}_5, \vec{x}_6\}$ به $\vec{v}_2 = (4.5, 4)$ اختصاص داده شده است.

**مزایا خوشه‌بندی k-means**

▪ دارای پیچیدگی زمانی خطی است. از این‌رو، سریع و کارآمد است.

▪ خوشه‌هایی را برمی‌گرداند که به‌راحتی قابل تفسیر و حتی تجسم هستند. این سادگی باعث می‌شود در برخی مواردی که نیاز به یک مرور کلی سریع از بخش‌های داده دارید، بسیار مفید باشد.

▪ پیاده‌سازی آسان.

▪ هنگام برخورد با خوشه‌های کروی عالی عمل می‌کند.

**معایب خوشه‌بندی k-means**

▪ الگوریتم یادگیری نیاز به تعیین پیشینِ تعداد مراکز خوشه دارد.

▪ مراکز خوشه نهایی، بستگی به مراکز انتخابی اولیه دارد. به عبارت دیگر، افراز اولیه متفاوت، منجر به خوشه‌های نهایی متفاوت می‌شوند.

▪ حساس به نقاط دورافتاده.

**پیچیدگی فضا و زمان**

فضای مورد نیاز برای K-means زیاد نیست، چراکه فقط نقاط داده و مرکزها ذخیره می‌شوند. به طور کلی، ذخیره‌سازی مورد نیاز $O((m + K)n)$ است، که در آن $m$ تعداد نقاط و $n$ تعداد ویژگی‌ها است. زمان مورد نیاز برای K-means از نظر تعداد نقاط داده خطی است. به طورکلی، زمان مورد نیاز $O(I \times K \times m \times n)$ است، جایی‌که $I$ تعداد تکرارهای مورد نیاز برای همگرایی است. $I$ اغلب کوچک است و معمولا می‌توان آن را با خیال راحت محدود کرد، زیرا اکثر تغییرات معمولا در چند تکرار اول رخ می‌دهند. بنابراین، K-means بر حسب $m$ (تعداد نقاط)، خطی، کارآمد و ساده است؛ مشروط بر اینکه $K$ (تعداد خوشه‌ها)، به‌طور قابل توجهی کم‌تر از $m$ باشد.

**خوشه‌بندی مخلوط گاوسی**

در قسمت قبل، خوشه‌بندی k-means را معرفی کردیم که یکی از پرکاربردترین روش‌های خوشه‌بندی است. این روش ناپارامتری برای خوشه‌بندی داده‌هایی با برخی ویژگی‌ها، عالی عمل می‌کند. در این بخش، یک روشِ پارامتری را بررسی خواهیم کرد که از توزیع گاوسی استفاده می‌کند که به **مدل مخلوط گاوسی** معروف است. یکی از نقاط قوت خوشه‌بندی مدل مخلوط گاوسی این است که یک روش خوشه‌بندی نرم است. به عبارت ساده‌تر، مجموعه‌ای از مدل‌های احتمالی را با داده‌ها برازش می‌دهد و به هر مورد احتمال تعلق اختصاص می‌دهد. این امر به ما این امکان را می‌دهد تا احتمال تعلقِ هر مورد به هر خوشه را بررسی کنیم. بنابراین خوشه‌بندی مدل مخلوط مجموعه‌ای از مدل‌های احتمالی را با داده‌ها برازش می‌دهد. این مدل‌ها می‌توانند انواع توزیع‌های احتمالی باشند، اما معمولا توزیع‌های گاوسی هستند. این رویکرد خوشه‌بندی، مدل‌سازی مخلوط نامیده می‌شود، زیرا توزیع‌های احتمالی متعدد (مخلوطی از) را به داده‌ها برازش می‌کنیم. بنابراین، یک مدل مخلوط گاوسی به سادگی مدلی است که چندین توزیع گاوسی را به مجموعه‌ای از داده‌ها برازش می‌دهد. هر گاوسی در مدل مخلوط، نشان‌دهنده یک خوشه بالقوه است. هنگامی که مخلوط گاوسی ما تا حد امکان با داده‌ها مطابقت داشت، می‌توانیم احتمال تعلق هر مورد به هر خوشه را محاسبه کنیم و موارد را به محتمل‌ترین خوشه اختصاص دهیم. سوالی که اینجا بوجود می‌آید این است که ما اطلاعی از توزیعی که هر نمونه آموزشی از آن حاصل شده است داریم و نه پارامترهای مدل مخلوط، با این حال، ما چگونه می‌توانیم ترکیبی از گاوسی‌ها را پیدا کنیم که به خوبی با داده‌های زیربنایی مطابقت داشته باشد؟ می‌توانیم روش مورد استفاده برای الگوریتم خوشه‌بندی k-means را اتخاذ کرده و تکرار کنیم. به این صورت که، با حدس‌های اولیه برای پارامترها شروع کرده و از آن‌ها برای محاسبه احتمالات خوشه برای هر نمونه استفاده می‌کنیم. سپس، از این احتمالات برای تخمین مجدد پارامترها استفاده کرده و

این چرخه را تا همگرایی تکرار می‌کنیم. این کار توسط الگوریتمی به نام **بیشینه‌سازی انتظار**[1] انجام می‌شود.

تحت شرایط خاص، k-means و مدل مخلوط گاوسی را می‌توان برحسب یکدیگر توضیح داد. در k-means، نقاطی که به یک مرکز خوشه نزدیک‌ترین باشند، به‌طور مستقیم به آن مرکز خوشه اختصاص پیدا می‌کنند، با این فرض که خوشه‌ها به‌طور مشابه مقیاس‌بندی شده و کوواریانس ویژگی آن‌ها متفاوت نیست. به همین دلیل است که اغلب منطقی است که داده‌های خود را قبل از استفاده از k-means هنجار کنید. با این حال، مخلوط گاوسی از چنین محدودیتی رنج نمی‌برد، به این دلیل که آن‌ها به دنبال مدل‌سازی کوواریانس ویژگی برای هر خوشه هستند. به عبارت دیگر، برخلاف k-means، خوشه‌بندی مخلوط گاوسی از بردارهای نمونه اولیه استفاده نمی‌کند، بلکه از مدل‌های احتمالی برای نمایش ساختارهای خوشه‌بندی استفاده می‌کند. مدل‌های مخلوط گوسی فرض می‌کنند که هر مشاهده در یک مجموعه داده از یک توزیع گاوسی با میانگین و واریانس متفاوت می‌آید. با برازش داده‌ها به مدل مخلوط گاوسی، هدف ما تخمین پارامترهای توزیع گاوسی با استفاده از داده‌ها است. اگر اینها کمی گیج‌کننده به نظر می‌رسد، نگران نباشید! ما این مفاهیم را با جزئیات بیشتر مرور خواهیم کرد. با این حال، قبل از این که در مورد جزئیات بیشتر این الگوریتم بحث کنیم، اجازه دهید تعریف توزیع گاوسی (چند متغیری) را مرور کنیم.

احتمالا نام توزیع گاوسی را شنیده باشید که گاهی اوقات نیز به عنوان توزیع نرمال شناخته می‌شود، اما توزیع گوسی دقیقا چیست؟ به زودی تعریف ریاضی را به شما ارائه خواهیم داد، اما از نظر کیفی می‌توان آن را توزیعی دانست که به‌طور طبیعی و بسیار مکرر اتفاق می‌افتد.

برای یک بردار تصادفی $x$ در یک فضای نمونه $n$ بعدی $\chi$، اگر $x$ از توزیع گاوسی پیروی کند، تابع چگالی احتمال آن برابر است با:

$$p(x) = \frac{1}{(2\pi)^{\frac{n}{2}} |\Sigma|^{\frac{1}{2}}} e^{\{-\frac{1}{2}(x-\mu)^T \Sigma^{-1}(x-\mu)\}}$$

که در آن $\mu$ یک بردارِ میانگین $n$ـ بعدی و $\Sigma$ یک ماتریس کوواریانس $n \times n$ است. از معادله فوق، می‌توانیم دید که توزیع گاوسی به طور کامل توسط بردار میانگین $\mu$ و ماتریس کوواریانس $\Sigma$ آن تعیین می‌شود. برای واضح‌تر نشان دادن این وابستگی، تابع چگالی احتمال را به صورت $p(x \mid \mu, \Sigma)$ می‌نویسیم.

توزیع مخلوط گاوسی به صورت زیر تعریف می‌شود:

$$p_{\mathcal{M}}(x) = \sum_{i=1}^{k} \alpha_i . p(x \mid \mu, \Sigma)$$

---

که از k مولفه مخلوط تشکیل شده که هرکدام مربوط به توزیع گاوسی هستند. $\mu_i$ و $\Sigma_i$ پارامترهای مولفه‌های مخلوط $i$ام هستند و $\alpha_i > 0$ ضرایب مخلوط مربوطه هستند که $\sum_{i=1}^{k} \alpha_i = 1$ است. فرض کنید که نمونه‌ها از یک توزیع مخلوط گاوسی با فرآیند زیر تولید می‌شوند:

انتخاب مولفهٔ مخلوط گاوسی با استفاده از توزیع پیشین تعریف‌شده توسط $\alpha_1, \alpha_1, \ldots, \alpha_k$ جایی که $\alpha_i$ احتمال انتخاب مولفه مخلوط $i$ام است. سپس با نمونه‌برداری از توابع چگالی احتمال مولفه مخلوط انتخابی، نمونه‌ها را تولید می‌کند.

فرض کنید $D = \{x_1, x_2, \ldots, x_m\}$، یک مجموعه آموزشی باشد که از فرآیند بالا تولید شده است و $z_j \epsilon \{1, 2, \ldots, k\}$ متغیر تصادفی مولفه مخلوط گاوسی که نمونه $x_j$ را تولید می‌کند، جایی که مقادیر $z_j$ ناشناخته است. از آنجایی که احتمال پیشین $P(z_j = i)$ برای $z_j$ با $\alpha_i \epsilon \{i = 1, 2, \ldots, k\}$ مطابقت دارد، توزیع پسین $z_j$، طبق قضیه بیز، برابر است با:

$$p_{\mathcal{M}}(z_j = i | x_j) = \frac{\alpha_i . p_{\mathcal{M}}(x_j | z_j = i)}{p_{\mathcal{M}}(x_j)}$$

$$= \frac{P(z_j = i) . P(x_j | \mu_i, \Sigma_i)}{\sum_{l=1}^{k} \alpha_l . P(x_j | \mu_l, \Sigma_l)}$$

به عبارت دیگر، $p_{\mathcal{M}}(z_j = i | x_j)$ احتمال پسینی را می‌دهد که $x_j$ توسط مولفه مخلوط گاوسی $i$ تولید می‌شود. برای سهولت، آن را با $\gamma_{ji}$ نشان می‌دهیم، جایی که $k, \ldots, 1, 2 = i$.

وقتی توزیع مخلوط گاوسی شناخته شده است، مجموعه داده $D$ را می‌توان به $k$ خوشه تقسیم کرد و تخصیص خوشه $\lambda_j$ برای هر نمونه $x_j$ توسط $\lambda_j = argmax_{i \epsilon \{1, 2, \ldots, k\}} \gamma_{ji}$ داده می‌شود. **از این رو، از منظر خوشه‌بندی نمونه اولیه، خوشه‌بندی مخلوط گاوسی از مدل‌های احتمالی (با توزیع گاوسی) برای نشان‌دادن نمونه‌های اولیه استفاده می‌کند و تخصیص خوشه‌ها توسط احتمالات پسین نمونه‌های اولیه انجام می‌شود.**

حال پرسش اینجاست چگونه پارامترهای مدل را بهینه کنیم؟ یک روش اعمال برآورد درست‌نمایی بیشینه در مجموعه داده $D$ است، یعنی بیشینه‌کردن ($log$) درست‌نمایی:

$$LL(D) = \ln \left( \prod_{j=1}^{m} p_{\mathcal{M}}(x_j) \right)$$

$$= \sum_{j=1}^{m} \ln \left( \sum_{i=1}^{k} \alpha_i . P(x_j | \mu_i, \Sigma_i) \right)$$

که معمولا توسط الگوریتم بیشینه‌سازی انتظار (EM) حل می‌شود.

الگوریتم EM (همان‌طور که از نامش پیداست) دو مرحله دارد: **انتظار (توقع)** و **بیشینه‌سازی**. مرحله انتظار جایی است که احتمالات پسین برای هر مورد، برای هر گاوس محاسبه می‌شود (شکل (ب) ۷ـ۱ برای گاوسی یک بعدی و شکل ۷ـ۲ (ب) برای گاوسی بیشتر از یک بعد). در این مرحله، الگوریتم از قضیه بیز برای محاسبه احتمالات پسین استفاده می‌کند. مرحله بعدی بیشینه‌سازی است. کار مرحله بیشینه‌سازی بروزرسانی پارامترهای مدل مخلوط، برای بیشینه کردن درست‌نمایی داده‌ها است. بر اساس قضیه بیز و $\gamma_{ji} = p_{\mathcal{M}}(z_j = i | x_j)$ بروزرسانی میانگین به‌صورت زیر انجام می‌شود:

$$\mu_i = \frac{\sum_{j=1}^m \gamma_{ji} x_j}{\sum_{j=1}^m \gamma_{ji}}$$

به عبارت دیگر، میانگین (مرکز خوشه را مشخص می‌کند) هر گاوسی را می‌توان به عنوان میانگین وزنی نمونه‌ها محاسبه کرد که در آن هر نمونه با احتمال پسین تعلق این نمونه به گاوسی داده شده وزن‌دهی می‌شود. کوواریانس (عرض را مشخص می‌کند) هر گاوسی به روشی مشابه بروز می‌شود:

$$\Sigma_i = \frac{\sum_{j=1}^m \gamma_{ji} (x_j - \mu_i)(x_j - \mu_i)^T}{\sum_{j=1}^m \gamma_{ji}}$$

آخرین چیزی که باید بروز شود، احتمالات پیشین برای هر گاوسی است. پیشین‌های جدید (ضریب مخلوط) با جمع احتمالات پسین برای یک گاوسی خاص و تقسیم بر تعداد نمونه‌ها محاسبه می‌شوند:

$$\alpha_i = \frac{\sum_{j=1}^m \gamma_{ji}}{m}$$

**جدول ۸ـ۱** مجموعه داده هندوانه (Zhou, 2021)

| ID | density | sugar | ID | density | sugar | ID | density | sugar |
|----|---------|-------|----|---------|-------|----|---------|-------|
| 1 | 0.697 | 0.460 | 11 | 0.245 | 0.057 | 21 | 0.748 | 0.232 |
| 2 | 0.774 | 0.376 | 12 | 0.343 | 0.099 | 22 | 0.714 | 0.346 |
| 3 | 0.634 | 0.264 | 13 | 0.639 | 0.161 | 23 | 0.483 | 0.312 |
| 4 | 0.608 | 0.318 | 14 | 0.657 | 0.198 | 24 | 0.478 | 0.437 |
| 5 | 0.556 | 0.215 | 15 | 0.360 | 0.370 | 25 | 0.525 | 0.369 |
| 6 | 0.403 | 0.237 | 16 | 0.593 | 0.042 | 26 | 0.751 | 0.489 |
| 7 | 0.481 | 0.149 | 17 | 0.719 | 0.103 | 27 | 0.532 | 0.472 |
| 8 | 0.437 | 0.211 | 18 | 0.359 | 0.188 | 28 | 0.473 | 0.376 |
| 9 | 0.666 | 0.091 | 19 | 0.339 | 0.241 | 29 | 0.725 | 0.445 |
| 10 | 0.243 | 0.267 | 20 | 0.282 | 0.257 | 30 | 0.446 | 0.459 |

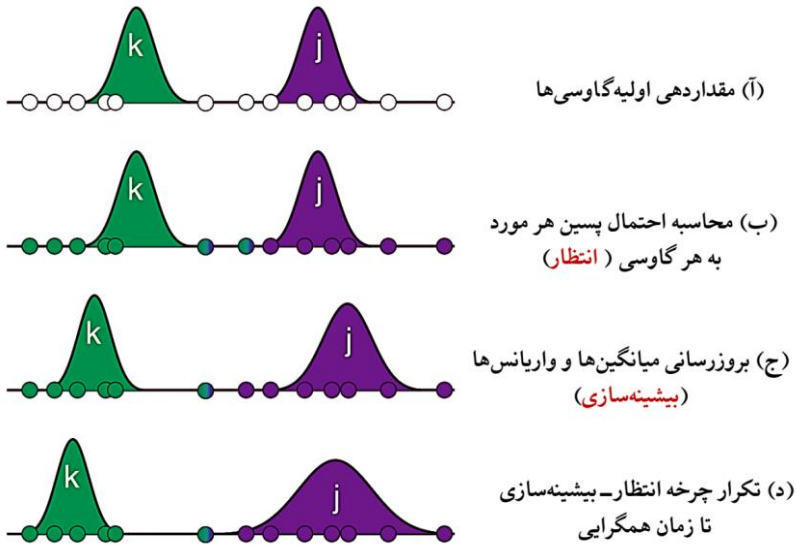

**شکل ۸ ـ ۱. الگوریتم بیشینه‌سازی انتظار برای دو گاوسی یک بعدی.** نقطه‌ها مواردی را در امتداد یک خط عددی نشان می‌دهند. دو گاوسی به طور تصادفی در طول خط مقداردهی اولیه می‌شوند. در مرحله انتظار، احتمال پسین هر مورد برای هر گاوس محاسبه شده و در مرحله بیشینه‌سازی، میانگین‌ها، واریانس‌ها (در این مثال به دلیل اینکه در یک بعد هستند از واریانس به جای کواریانس استفاده می‌شود) و احتمالات پیشین برای هر گاوسی بر اساس پسین‌های محاسبه‌شده بروز می‌شوند. این روند تا زمانی ادامه می‌یابد که درست‌نمایی همگرا شود.

**(آ) مقداردهی اولیه‌گاوسی‌ها به‌صورت تصادفی**

**(ب) مرحله انتظار**

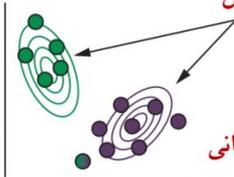

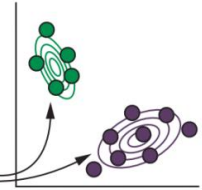

۱. محاسبه پسین‌ها برای هر مورد

**(ج) مرحله بیشینه‌سازی**

**(د) تکرار تا همگرایی**

۲. بروزرسانی ماتریس‌های کوواریانس و پیشین‌ها برای هر گاوسی

۳. تکرار چرخه انتظار ـ بیشینه‌سازی تا زمانی که بهبود در درست‌نمایی ناچیز باشد

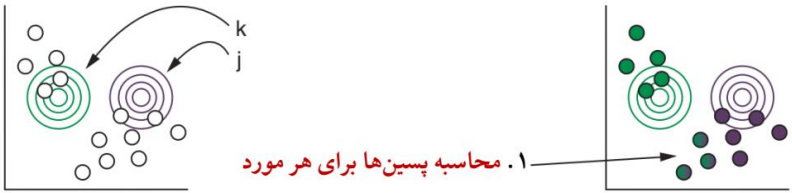

**شکل ۸ ـ ۲. الگوریتم بیشینه‌سازی انتظار برای دو گاوسی دو بعدی.** دو گاوسی به طور تصادفی در فضای ویژگی مقداردهی اولیه می‌شوند. در مرحله انتظار، احتمالات پسین برای هر مورد برای هر گاوس محاسبه

می‌شود. در مرحله بیشینه‌سازی، میانگین‌ها، ماتریس‌های کوواریانس و پیشین‌ها برای هر گاوسی بر اساس پسین‌ها بروز می‌شوند. این روند تا زمانی ادامه می‌یابد که درست‌نمایی همگرا شود.

هنگامی که مرحله بیشینه‌سازی کامل شد، تکرار دیگری از مرحله انتظار را انجام می‌دهیم، این بار احتمالات پسین را برای هر مورد تحت گاوسی‌های جدید محاسبه می‌کنیم. هنگامی که این کار انجام شد، مرحله بیشینه‌سازی را مجددا اجرا می‌کنیم و دوباره میانگین‌ها، کوواریانس‌ها و ضریب مخلوط (پیشین‌ها) را برای هر گاوسی بر اساس موارد پسین بروزرسانی می‌کنیم. این چرخه انتظار-بیشینه‌سازی به صورت تکراری ادامه می‌یابد تا زمانی که به تعداد مشخصی از تکرارها برسد یا درست‌نمایی کلی داده‌های تحت مدل، کمتر از مقدار مشخصی تغییر کند (همگرایی).

مجموعه داده هندوانه در جدول ۷ـ۱ را به عنوان مثال در نظر بگیرید تا نمایش دقیق‌تری از خوشه‌بندی مدل مخلوط گوسی ارائه دهیم. فرض کنید که تعداد اجزای مخلوط گاوسی ۳ $= k$ است و الگوریتم با مقداردهی اولیه پارامتر زیر شروع می‌شود:

$$\alpha_١ = \alpha_٢ = \alpha_٣ = \frac{١}{٣}; \; \mu_١ = x_٦; \mu_٢ = x_{٢٢}; \mu_٣ = x_{٢٧};$$

$$\Sigma_١ = \Sigma_٢ = \Sigma_٣ = \begin{pmatrix} ٠٫١ & ٠٫٠ \\ ٠٫٠ & ٠٫١ \end{pmatrix}$$

در اولین تکرار، الگوریتم احتمالات پسین نمونه‌ها را با توجه به اینکه آن‌ها توسط هر مولفه مخلوط تولید می‌شوند محاسبه می‌کند. به عنوان مثال با در نظر گرفتن $x_١$، احتمالات پسین محاسبه شده توسط $\gamma_{ji}$ به‌صورت زیر است:

$$\gamma_{١١} = ٠٫٢١٩; \; \gamma_{١٢} = ٠٫٤٠٤; \; \gamma_{١٣} = ٠٫٣٧٧$$

پس از محاسبه احتمالات پسین همه نمونه‌ها با توجه به تمام مولفه‌های مخلوط، پارامترهای مدل بروز شده زیر را بدست می‌آوریم:

$$\acute{\alpha}_١ = ٠٫٣٦١; \acute{\alpha}_٢ = ٠٫٣٢٣; \; \acute{\alpha}_٣ = ٠٫٣١٦$$

$$\acute{\mu}_١ = (٠٫٤٩١, ٠٫٢٥١); \acute{\mu}_٢ = (٠٫٥٧١, ٠٫٢٨١); \acute{\mu}_٣ = (٠٫٥٣٤, ٠٫٢٩٥)$$

$$\acute{\Sigma}_١ = \begin{pmatrix} ٠٫٠٢٥ & ٠٫٠٠٤ \\ ٠٫٠٠٤ & ٠٫٠١٦ \end{pmatrix}$$

$$\acute{\Sigma}_٢ = \begin{pmatrix} ٠٫٠٢٣ & ٠٫٠٠٤ \\ ٠٫٠٠٤ & ٠٫٠١٧ \end{pmatrix}$$

$$\acute{\Sigma}_٣ = \begin{pmatrix} ٠٫٠٢٤ & ٠٫٠٠٥ \\ ٠٫٠٠٥ & ٠٫٠١٦ \end{pmatrix}$$

روند بروز رسانی بالا تا زمان همگرایی تکرار می‌شود. . شکل ۷ـ۳ نتایج خوشه‌بندی را پس از تکرارهای مختلف نشان می دهد.

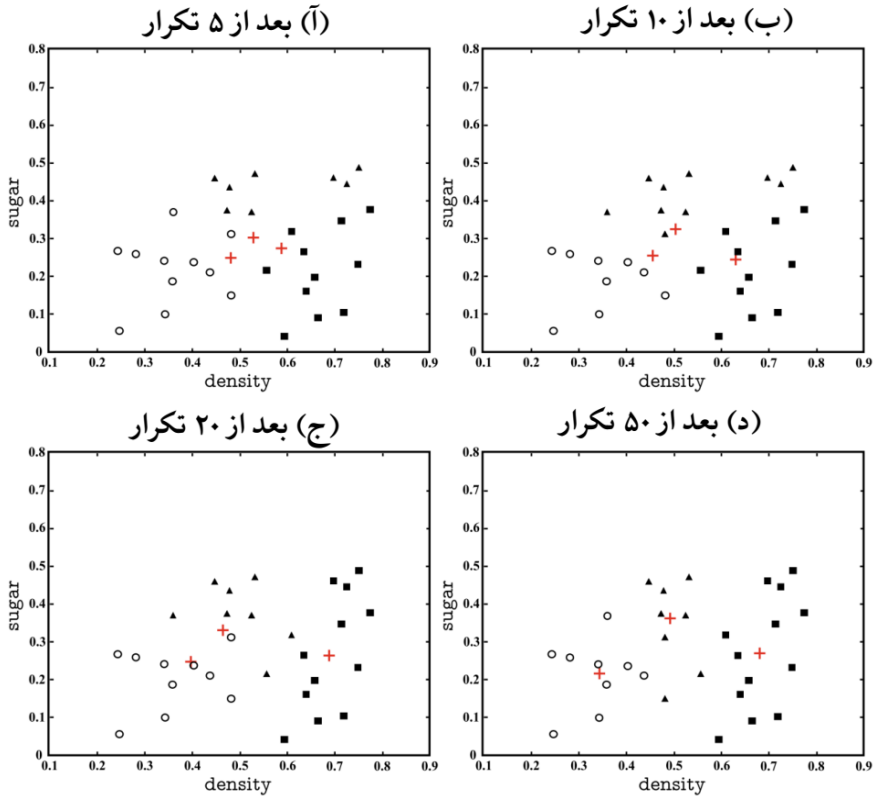

**شکل ۸ـ۳. نتایج الگوریتم خوشه‌بندی مدل مخلوط گوسی پس از تکرارهای مختلف در مجموعه داده هندوانه (جدول ۸ـ۱) با ۳ = $k$. بردارهای میانگین مولفه مخلوط گاوسی با "+" نشان داده شده‌اند.**

**سناریوهای مورد استفاده از مدل مخلوط گاوسی**

- در مورد تجزیه و تحلیل سری‌های زمانی، مدل مخلوط گاوسی را می‌توان برای کشف چگونگی ارتباط نوسان با روند و نویز استفاده کرد که می‌تواند به پیش‌بینی قیمت سهام در آینده کمک کند. یک خوشه می‌تواند از یک روند در سری زمانی تشکیل شود، در حالی که خوشه دیگر می‌تواند نویز و نوسانات ناشی از عوامل دیگر مانند رویدادهای فصلی یا رویدادهای خارجی که بر قیمت سهام تاثیر می‌گذارد را داشته باشد. برای جداکردن این خوشه‌ها، می‌توان از مدل مخلوط گاوسی استفاده کرد. چراکه به جای تقسیم ساده داده‌ها به دو بخش مانند K-means، احتمالی را برای هر دسته ارائه می‌دهند.

- زمانی است که گروه‌های مختلفی در یک مجموعه داده وجود دارد و به سختی می‌توان آن‌ها را به عنوان متعلق به یک گروه یا گروه دیگر برچسب‌گذاری کرد. مدل مخلوط گاوسی را می‌توان در این مورد استفاده کرد. چراکه آن‌ها مدل‌های گاوسی را پیدا می‌کنند که به

بهترین شکل هر گروه را توصیف می‌کند و احتمالی را برای هر خوشه ارائه می‌دهد که هنگام برچسب‌زدن خوشه‌ها مفید است.

■ مثال دیگری که در آن مدل مخلوط گاوسی می‌تواند مفید باشد، زمانی است که می‌خواهیم گروه‌های زیربنایی از دسته‌ها مانند انواع سرطان یا عوامل خطر مرتبط با انواع مختلف سرطان را کشف کنیم.

**کاربرد مدل مخلوط گاوسی**

بسیاری از مسائل مختلف در دنیای واقعی وجود دارد که می‌توان آن‌ها را با مدل‌های مخلوط گاوسی حل کرد. مدل‌های مخلوط گاوسی زمانی بسیار مفید هستند که مجموعه داده‌های بزرگی وجود داشته باشد و یافتن خوشه‌ها دشوار باشد. این جایی است که مدل‌های مخلوط گاوسی می‌توانند خوشه‌های گاوسی را کارآمدتر از سایر الگوریتم‌های خوشه‌بندی مانند k-means بیابند. برخی از مسائلی است که با استفاده از مدل‌های مخلوط گاوسی قابل حل هستند، در زیر فهرست شده‌اند:

■ **یافتن الگو در مجموعه داده‌های پزشکی:** مدل‌های مخلوط گاوسی می‌توانند برای بخش‌بندی تصاویر به دسته‌های چندگانه براساس محتوای آن‌ها یا یافتن الگوهای خاص در مجموعه داده‌های پزشکی استفاده شود.

■ **مدل‌سازی پدیده‌های طبیعی:** از مدل‌های مخلوط گاوسی می‌توان برای مدل‌سازی پدیده‌های طبیعی استفاده کرد که در آن مشخص شده است که نویز از توزیع‌های گاوسی تبعیت می‌کند.

■ **تجزیه و تحلیل رفتار مشتری:** مدل‌های مخلوط گاوسی را می‌توان برای انجام تجزیه و تحلیل رفتار مشتری در بازاریابی برای پیش‌بینی خریدهای آینده بر اساس داده‌های گذشته استفاده کرد.

■ **پیش‌بینی قیمت سهام:** حوزه دیگری که از مدل‌های مخلوط گاوسی استفاده می‌شود، در امور مالی است که می‌توان آن‌ها را در سری‌های زمانی قیمت سهام اعمال کرد. مدل‌های مخلوط گاوسی می‌توانند برای شناسایی نقاط تغییر در داده‌های سری زمانی استفاده شوند و به یافتن نقاط عطف قیمت سهام یا سایر حرکات بازار کمک کنند که به دلیل نوسانات و نویز تشخیص آن‌ها دشوار است.

■ **تجزیه و تحلیل داده‌های بیان ژن:** مدل‌های مخلوط گاوسی را می‌توان برای تحلیل داده‌های بیان ژنی استفاده کرد. به طور خاص، مدل‌های مخلوط گاوسی می‌توانند برای شناسایی ژن‌های بیان‌شده بین دو حالت و تشخیص اینکه کدام ژن‌ها ممکن است در فنوتیپ یا وضعیت بیماری خاصی نقش داشته باشند، استفاده کرد.

**مزایا خوشه‌بندی مدل مخلوط گاوسی**

- از یک رویکرد احتمالی استفاده می‌کند و برای هر نقطه داده‌ای که به خوشه‌ها تعلق دارد، احتمال ارائه می‌کند.
- می‌تواند خوشه‌های غیر کروی با قطرهای مختلف را شناسایی کند.
- نسبت به متغیرها در مقیاس‌های مختلف حساس نیست.

> اگر به دنبال راهی کارآمد برای یافتن الگوها در مجموعه داده‌های پیچیده هستید یا برای مدل‌سازی پدیده‌های طبیعی مانند بلایای طبیعی یا تجزیه و تحلیل رفتار مشتری در بازاریابی خود به کمک نیاز دارید، مدل‌های مخلوط گاوسی می‌توانند انتخاب مناسبی باشند.

**معایب خوشه‌بندی مدل مخلوط گاوسی**

- به مجموعه داده‌های بزرگی نیاز دارد و تخمین تعداد خوشه‌ها دشوار است.
- به دلیل تصادفی بودن گاوسی‌های اولیه، پتانسیل همگرایی به یک مدل بهینه محلی را دارد.
- به موارد دورافتاده حساس است.

# خوشه‌بندی سلسله‌مراتبی

در بخش قبل دیدیم که چگونه خوشه‌بندی مبتنی‌بر نمونه اولیه، k مرکز را در فضای ویژگی پیدا می‌کند و به‌طور مکرر آن‌ها را برای یافتن مجموعه‌ای از خوشه‌ها بروز می‌کند. خوشه‌بندی سلسله مراتبی رویکرد متفاوتی دارد و همان‌طور که از نامش پیداست، سلسله‌مراتب خوشه‌ها را به شکل درخت توسعه می‌دهد. این ساختار درختی شکل، به **درخت‌واره‌نگار**[1] معروف است. به جای دریافت یک خروجی "هموار" از خوشه‌ها، خوشه‌بندی سلسله‌مراتبی درختی از خوشه‌ها را به ما می‌دهد. در نتیجه، خوشه‌بندی سلسله‌مراتبی نسبت به روش‌های خوشه‌بندی مسطح همانند k-means، بینشِ بیشتری نسبت به ساختارهای گروه‌بندی پیچیده فراهم می‌کند.

از این‌رو، مزیت اصلی خوشه‌بندی سلسله مراتبی نسبت به رویکرد مبتنی‌بر نمونه اولیه، این است که ما درک دقیق‌تری از ساختار داده‌های خود بدست می‌آوریم و این رویکرد اغلب قادر به بازسازی سلسله‌مراتب واقعی در طبیعت است. به عنوان مثال، تصور کنید که ما ژنوم همه‌یِ نژادهای گربه را توالی‌یابی کنیم. می‌توانیم با خیال راحت فرض کنیم که ژنوم یک نژاد بیشتر شبیه ژنوم نژاد(هایی) است که از آن مشتق شده است تا به ژنوم نژادهایی که از آن مشتق نشده است. اگر خوشه‌بندی سلسله‌مراتبی را برای این داده‌ها اعمال کنیم، سلسله‌مراتبی را که می‌توان به

---

[1] dendrogram

صورت درخت‌گواره‌نگار تجسم کرد، می‌تواند به این صورت تفسیر شود که کدام نژادها از نژادهای دیگر مشتق شده‌اند.

هنگام تلاش برای یادگیری سلسله‌مراتبی از داده‌ها، دو رویکرد وجود دارد:

- **تجمیعی**[1]: یک رویکرد از پایین به بالا است که در آن الگوریتم با گرفتن تمام نقاط داده به عنوان خوشه‌های منفرد و ادغام آنها تا زمانی که یک خوشه باقی بماند انجام می‌شود.

- **تقسیمی**[2]: این رویکرد، برعکس رویکرد تجمیعی عمل می‌کند و به‌صورت بالا به پایین است. یعنی، با همه موارد در یک خوشه شروع می‌شود و به صورت بازگشتی آن‌ها را به خوشه‌هایی تقسیم می‌کند تا زمانی که هر مورد در خوشه خودش قرار گیرد.

**درخت‌واره‌نگار**

خوشه‌بندی سلسله‌مراتبی را می‌توان با یک درخت دودویی ریشه‌دار نشان داد. گره‌های درختان گروه‌ها یا خوشه‌ها را نشان می‌دهند و گره ریشه، کل مجموعه داده را نشان می‌دهد. همچنین، گره‌های پایانی هر کدام یکی از مشاهدات منفرد (خوشه‌های منفرد) را نشان می‌دهند. هر گره غیر پایانی دو گره دختر دارد.

درخت‌واره‌نگار، یک نمودار درختی است که برای نشان‌دادن آرایش خوشه‌های تولید شده توسط خوشه‌بندی سلسله‌مراتبی استفاده می‌شود. درخت‌واره‌نگار ممکن است با گره ریشه در بالا و رشد شاخه‌ها به صورت عمودی به سمت پایین ترسیم شود (شکل (آ) ۸ ـ ۴). همچنین ممکن است با گره ریشه در سمت چپ و رشد شاخه‌های افقی به سمت راست کشیده شود (شکل (ب) ۸ ـ ۴).

**شکل ۸ ـ ۴.** روش‌های مختلف نمایش درخت‌واره‌نگار

---

[1] Agglomerative

[2] Divisive

در شکل ۸ـ۵ یک درخت‌واره‌نگار برای مجموعه داده {$a, b, c, d, e$} نشان داده شده است. توجه داشته باشید که گره ریشه، کلِ مجموعه داده را نشان می‌دهد و گره‌های پایانی مشاهدات فردی را نشان می‌دهند. با این حال، درخت‌واره‌نگارها در قالبی ساده‌تر ارائه می‌شوند که در آن فقط گره‌های پایانی (یعنی گره‌هایی که خوشه‌های تک را نشان می‌دهند) به صراحت نمایش داده می‌شوند. شکل ۸ـ۴ نمایشِ ساده شدهِ درخت‌واره‌نگار شکل ۸ـ۵ را نشان می‌دهد.

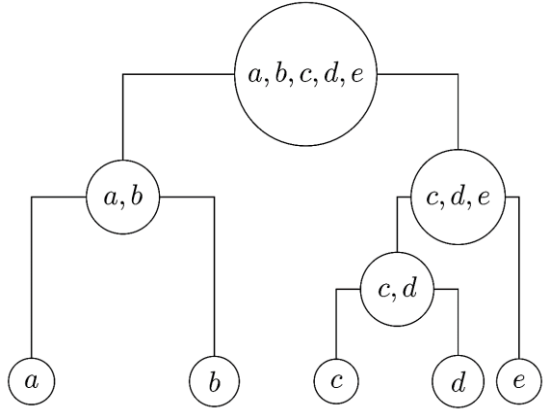

**شکل ۸ـ۵.** درخت‌واره‌نگار برای مجموعه داده {$a, b, c, d, e$}

## خوشه‌بندی سلسله‌مراتبی تجمیعی

این رویکرد با در نظر گرفتن هر نمونه در مجموعه داده به عنوان یک خوشه اولیه شروع می‌شود. سپس در هر دور، دو نزدیک‌ترین خوشه به عنوان یک خوشه جدید ادغام می‌شوند و این روند تا زمانی تکرار می‌شودکه تعداد خوشه‌ها به مقدار از پیش تعیین‌شده برسد. از این‌رو می‌توان مراحل این الگوریتم را به صورت زیر خلاصه کرد:

۱.  **تعریف یک معیار فاصله (تعریف‌شده توسط کاربر) بین هر خوشه**
۲.  **ادغام شبیه‌ترین خوشه‌ها در یک خوشه**
۳.  **مراحل ۱ و ۲ را تکرار کنید تا همه نمونه‌ها در یک خوشه قرار گیرند.**

نحوه کار این الگوریتم در شکل ۸ـ۶ نشان داده شده است. از آنجایی که در این شکل ۹ نمونه وجود دارد، با ۹ خوشه شروع می‌کنیم. این الگوریتم یک معیار فاصله بین هر یک از خوشه‌ها را محاسبه کرده و خوشه‌هایی که بیشترین شباهت به یکدیگر را دارند باهم ادغام می‌کند. این عمل تا زمانی ادامه می‌یابد که تمام نمونه‌ها، در ابرخوشه نهایی قرار گیرند.

در اینجا، نکته کلیدی نحوه اندازه‌گیری فاصله بین خوشه‌ها است. از آنجایی که هر خوشه مجموعه‌ای از نقاط داده است، باید یک اندازه‌گیری فاصله در مورد مجموعه‌ها تعریف کنیم. از این‌رو، تصمیم در مورد ادغام یا عدم ادغام دو خوشه بر اساس اندازه‌گیری عدم‌تشابه بین خوشه‌ها

گرفته می‌شود. به عنوان مثال، با توجه به خوشه های $C_i$ و $C_j$، می‌توانیم فواصل زیر را تعریف کنیم:

■ کم‌ترین فاصله:

$$d_{min}(C_i, C_j) = min_{x \epsilon C_i, z \epsilon C_j} dist(x, z)$$

■ بیشترین فاصله:

$$d_{max}(C_i, C_j) = max_{x \epsilon C_i, z \epsilon C_j} dist(x, z)$$

■ میانگین فاصله:

$$d_{avg}(C_i, C_j) = \frac{1}{|C_i||C_j|} \sum_{x \epsilon C_i} \sum_{z \epsilon C_j} dist(x, z)$$

کم‌ترین فاصله بین دو خوشه توسط دو نزدیک‌ترین نمونه آن‌ها تعیین می‌شود. حداکثر فاصله توسط دو نمونه دورتر از خوشه‌ها تعیین می‌شود. میانگین فاصله توسط همه نمونه‌ها در هر دو خوشه تعیین می‌شود. هنگامی که فواصل خوشه با $d_{min}$، $d_{max}$ یا $d_{avg}$ اندازه‌گیری می‌شوند، الگوریتم مربوطه را به ترتیب **پیوند–تکی**[۱]، **پیوند–کامل**[۲] یا **پیوند–میانگین**[۳] می‌گویند. هر یک از این پیوندها در شکل ۷ــ۸ نشان داده شده است.

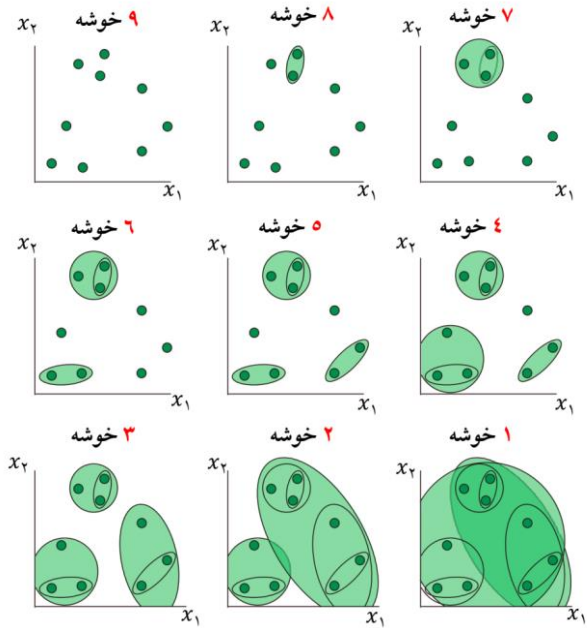

**شکل ۸ ــ ٦.** الگوریتم خوشه‌بندی سلسله‌مراتبی تجمیعی

---

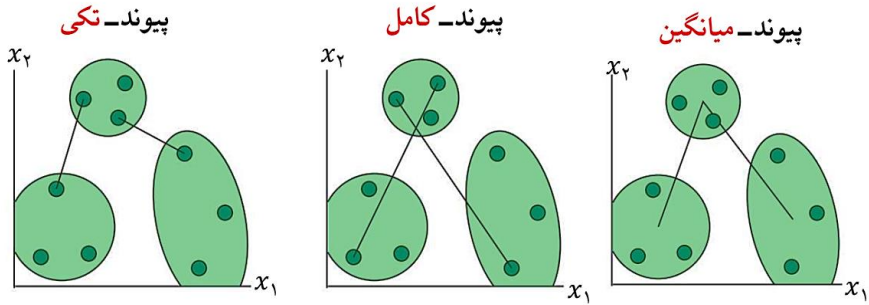

**شکل ۷ـ۸. روش‌های مختلف پیوند.** پیوندتکی فاصله بین نزدیکترین موارد دو خوشه را به عنوان فاصله بین آن خوشه‌ها می‌گیرد. پیوندـ کامل فاصله بین دورترین حالت دو خوشه را به عنوان فاصله بین آن خوشه‌ها می‌گیرد. پیوند میانگین، میانگین فاصله بین تمام موارد دو خوشه را به عنوان فاصله بین آن خوشه‌ها می‌گیرد.

**مثال.** با توجه به مجموعه داده $\{a, b, c, d, e\}$ و ماتریس فاصله زیر، می‌خواهیم یک درخت‌واره‌نگار با خوشه‌بندی سلسله‌مراتبی پیوندـ کامل بسازیم:

|   | **a** | **b** | **c** | **d** | **e** |
|---|---|---|---|---|---|
| **a** | ۰ | ۹ | ۳ | ۶ | ۱۱ |
| **b** | ۹ | ۰ | ۷ | ۵ | ۱۰ |
| **c** | ۳ | ۷ | ۰ | ۹ | ۲ |
| **d** | ۶ | ۵ | ۹ | ۰ | ۸ |
| **e** | ۱۱ | ۱۰ | ۲ | ۸ | ۰ |

خوشه‌بندی پیوندـ کامل از "معادله **بیشترین فاصله**"، یعنی معادله زیر برای محاسبه فاصله بین دو خوشه $C_j$ و $C_i$ استفاده می‌کند:

$$d_{max}(C_i, C_j) = max_{x \in C_i, z \in C_j} dist(x, z)$$

۱.   با توجه به مجموعه داده $\{a, b, c, d, e\}$ خوشه‌های اولیه برابر است با:

$C_1: \{a\}, \{b\}, \{c\}, \{d\}, \{e\}$

۲.   جدول زیر فواصل بین خوشه‌های مختلف در $C_1$ را نشان می‌دهد:

|   | **a** | **b** | **c** | **d** | **e** |
|---|---|---|---|---|---|
| **a** | ۰ | ۹ | ۳ | ۶ | ۱۱ |
| **b** | ۹ | ۰ | ۷ | ۵ | ۱۰ |
| **c** | ۳ | ۷ | ۰ | ۹ | ۲ |
| **d** | ۶ | ۵ | ۹ | ۰ | ۸ |
| **e** | ۱۱ | ۱۰ | ۲ | ۸ | ۰ |

در جدول فوق، حداقل فاصله، فاصله بین خوشه‌های $\{c\}$ و $\{e\}$ است. از این‌رو، $\{c\}$ و $\{e\}$ را ادغام می‌کنیم.

بر این اساس، مجموعه جدید خوشه‌ها برابر است با:

$$C_2: \{a\}, \{b\}, \{d\}, \{c, e\}$$

۳. فاصله $\{c, e\}$ را از خوشه‌های دیگر محاسبه می‌کنیم:

$$dist(\{c, e\}, \{a\}) = \max\{dist(c, a), dist(e, a)\} = \max\{3, 11\} = 11$$
$$dist(\{c, e\}, \{b\}) = \max\{dist(c, b), dist(e, b)\} = \max\{7, 10\} = 10$$
$$dist(\{c, e\}, \{d\}) = \max\{dist(c, d), dist(e, d)\} = \max\{9, 8\} = 9$$

بر این اساس، جدول زیر فواصل بین خوشه‌های مختلف در $C_2$ را نشان می‌دهد.

|  | **a** | **b** | **d** | **c, e** |
|---|---|---|---|---|
| a | ۰ | ۹ | ۶ | ۱۱ |
| b | ۹ | ۰ | ۵ | ۱۰ |
| d | ۶ | ۵ | ۰ | ۹ |
| c, e | ۱۱ | ۱۰ | ۹ | ۰ |

در جدول فوق، حداقل فاصله، فاصله بین خوشه‌های $\{b\}$ و $\{d\}$ است. از این‌رو، $\{b\}$ و $\{d\}$ را ادغام می‌کنیم.

بر این اساس، مجموعه جدید خوشه‌ها برابر است با:

$$C_3: \{a\}, \{b, d\}, \{c, e\}$$

۴. فاصله $\{b, d\}$ را از خوشه‌های دیگر محاسبه می‌کنیم:

$$dist(\{b, d\}, \{a\}) = \max\{dist(b, a), dist(d, a)\} = \max\{9, 6\} = 9$$
$$dist(\{b, d\}, \{c, e\}) = \max\{dist(b, c), dist(b, e), dist(d, c), dist(d, e)\}$$
$$= \max\{7, 10, 9, 8\} = 10$$

بر این اساس، جدول زیر فواصل بین خوشه‌های مختلف در $C_3$ را نشان می‌دهد.

|  | **a** | **b, d** | **c, e** |
|---|---|---|---|
| a | ۰ | ۹ | ۱۱ |
| b, d | ۹ | ۰ | ۱۰ |
| c, e | ۱۱ | ۱۰ | ۰ |

در جدول فوق، حداقل فاصله، فاصله بین خوشه‌های $\{a\}$ و $\{b, d\}$ است. از این‌رو، $\{a\}$ و $\{b, d\}$ را ادغام می‌کنیم.

بر این اساس، مجموعه جدید خوشه‌ها برابر است با:

$$C_4: \{a, b, d\}, \{c, e\}$$

۵. فقط دو خوشه باقی مانده است. از این‌رو، آن‌ها را با هم ادغام می‌کنیم و یک خوشه منفرد حاوی تمام نقاط داده را تشکیل می‌دهیم. داریم:

$$dist(\{a, b, d\}, \{c, e\})$$
$$= \max\{dist(a, c), dist(a, e), dist(b, c), dist(b, e), dist(d, c), dist(d, e)\}$$
$$= \max\{3, 11, 7, 10, 9, 8\} = 11$$

۶. شکل ۸-۸ درخت‌واره‌نگار خوشه‌بندی سلسله‌مراتبی را نشان می‌دهد.

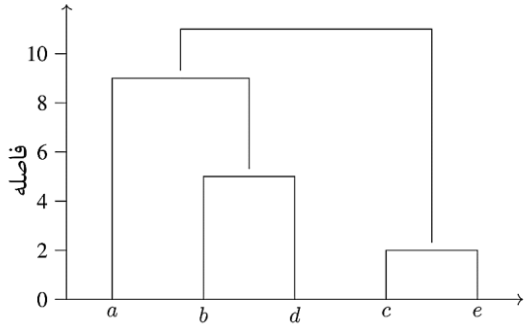

**شکل ۸ ـ ۸.** درخت‌واره‌نگار خوشه‌بندی سلسله‌مراتبی مثال ارائه‌شده

## خوشه‌بندی سلسله‌مراتبی تقسیمی

برخلاف خوشه‌بندی تجمیعی، خوشه‌بندی تقسیمی با همهٔ موارد در یک خوشه شروع می‌شود و به صورت بازگشتی آن را به خوشه‌های کوچک‌تر و کوچک‌تر تقسیم می‌کند، تا زمانی که هر مورد در خوشه خودش قرار گیرد. یافتن تقسیم بهینه، در هر مرحله از خوشه‌بندی کار دشواری است. از این‌رو، خوشه‌بندی تقسیمی، از یک **رویکرد ابتکاری**[1] **(اکتشافی)** استفاده می‌کند.

بر این اساس، در هر مرحله از خوشه‌بندی، خوشه‌ای با بیشترین قطر انتخاب می‌شود. قطر یک خوشه بزرگ‌ترین فاصله (عدم تشابه) بین هر دو نمونه آن است. از این‌رو، الگوریتم نمونه‌ای را در این خوشه پیدا می‌کند که بیشترین فاصله متوسط را با سایر نمونه‌ها در خوشه دارد. این غیرمشابه‌ترین مورد، گروه تقسیمی خود را ایجاد می‌کند. این روند تا زمانی تکرار می‌شود که همه موارد در خوشه خودشان قرار بگیرند. اساسا، خوشه‌بندی تقسیمی، خوشه‌بندی k-means (با k=۲) را در هر سطح از سلسله‌مراتب به‌منظور تقسیم هر خوشه اعمال می‌کند.

تنها یک پیاده‌سازی از خوشه‌بندی تقسیمی با عنوان الگوریتم DIANA وجود دارد. خوشه‌بندی تجمعی بیشتر مورد استفاده قرار می‌گیرد و از نظر محاسباتی هزینه کم‌تری نسبت به الگوریتم DIANA دارد. با این حال، اشتباهاتی که در ابتدای خوشه‌بندی سلسله مراتبی انجام می‌شوند را نمی‌توان در پایین‌تر برطرف کرد. بنابراین، در حالی که خوشه‌بندی تجمیعی ممکن است در یافتن خوشه‌های کوچک بهتر عمل کند، DIANA ممکن است در یافتن خوشه‌های بزرگ بهتر عمل کند.

## الگوریتم DIANA

طرح کلی الگوریتم DIANA در زیر آمده است:

**مرحله ۱.** فرض کنید که خوشه $C_l$ قرار است به خوشه های $C_i$ و $C_j$ تقسیم شود.

---

[1] heuristic approach

**مرحله ۲.** اگر داشته باشیم: $C_i = C_l$ و $C_j = \emptyset$.

**مرحله ۳.** برای هر شی $x \in C_i$:

(آ): برای اولین تکرار، میانگین فاصله $x$ را تا تمام اشیاء دیگر محاسبه کنید.

(ب): برای تکرارهای باقی‌مانده، محاسبه زیر را انجام دهید:

$$D_x = avg\{dist(x,y): y \epsilon C_i\} - avg\{dist(x,y): y \epsilon C_j\}$$

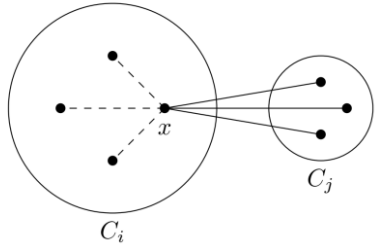

**شکل ۸ـ۹.** (میانگین خط ممتد)ـ (میانگین خط‌چین)= $D_x$

**مرحله ٤.**

(آ): برای اولین تکرار، شی با حداکثر فاصله متوسط را به $C_j$ منتقل کنید.

(ب): برای تکرارهای باقی‌مانده، یک شی $x$ را در $C_i$ پیدا کنید که $D_x$ برای آن بزرگ‌ترین است. اگر $D_x > 0$ باشد، $x$ را به $C_j$ منتقل کنید.

**مرحله ٥.** مراحل(ب) ۳ و (ب) ٤ را تکرار کنید تا زمانی‌که تمام تفاوت‌های $D_x$ منفی شوند. سپس $C_l$ به $C_i$ و $C_j$ تقسیم می‌شود.

**مرحله ٦.** خوشه کوچک‌تر با بیشترین قطر را انتخاب کنید. سپس این خوشه را طبق مراحل ۱ تا ٥ تقسیم کنید.

**مرحله ۷.** مرحله ٦ را تکرار کنید تا زمانی که همه خوشه‌ها فقط یک شی واحد داشته باشند.

**مثال.** با توجه به مجموعه داده $\{a, b, c, d, e\}$ و ماتریس فاصله زیر، می‌خواهیم یک درخت‌واره‌نگار با خوشه‌بندی سلسله‌مراتبی با استفاده از الگوریتم DIANA بسازیم:

|   | a | b | c | d | e |
|---|---|---|---|---|---|
| a | ۰ | ۹ | ۳ | ٦ | ۱۱ |
| b | ۹ | ۰ | ۷ | ٥ | ۱۰ |
| c | ۳ | ۷ | ۰ | ۹ | ۲ |
| d | ٦ | ٥ | ۹ | ۰ | ۸ |
| e | ۱۱ | ۱۰ | ۲ | ۸ | ۰ |

۱. در ابتدا داریم: $C_l = \{a, b, c, d, e\}$. همچنین:

$$C_j = \emptyset \text{ و } C_i = C_l$$

۲. تقسیم‌سازی:

**(آ): تکرار اولیه.**

میانگین عدم تشابه اشیاء در $C_i$ را با سایر اشیاء در $C_i$ محاسبه کنیم.

میانگین عدم تشابه $a$:

$$= \frac{1}{\xi}\big(dist(a,b) + dist(a,c) + dist(a,d) + dist(a,e)\big)$$

$$= \frac{1}{\xi}(9 + ۳ + ٦ + ۱۱) = ۷.۲۵$$

به همین ترتیب داریم:

میانگین عدم تشابه $b = ۷.۷۵$

میانگین عدم تشابه $c = ٥.۲۵$

میانگین عدم تشابه $d = ۷.۰۰$

میانگین عدم تشابه $e = ۷.۷۵$

بالاترین میانگین فاصله ۷.۷۵ است و دو شی متناظر برای آن وجود دارد. یکی از آن‌ها، به‌طور دلخواه انتخاب می‌کنیم. ما $b$ را انتخاب کرده و به $C_j$ منتقل می‌کنیم. اکنون داریم:

$$C_i = \{a,c,d,e\}, C_j = \emptyset \cup \{b\} = \{b\}$$

**(ب): تکرارهای باقی‌مانده.**

- دومین تکرار.

$$D_a = \frac{1}{۳}\big(d(a,c) + d(a,d) + d(a,e)\big) - \frac{1}{1}\big(d(a,b)\big)$$

$$= \frac{۲۰}{۳} - 9 = -۲.۳۳$$

$$D_c = \frac{1}{۳}\big(d(c,a) + d(c,d) + d(c,e)\big) - \frac{1}{1}\big(d(c,b)\big)$$

$$= \frac{۱٤}{۳} - ۷ = -۲.۳۳$$

$$D_d = \frac{1}{۳}\big(d(d,a) + d(d,c) + d(d,e)\big) - \frac{1}{1}\big(d(d,b)\big)$$

$$= \frac{۲۳}{۳} - ۷ = ۰.٦۷$$

$$D_e = \frac{1}{۳}\big(d(e,c) + d(e,c) + d(e,d)\big) - \frac{1}{1}\big(d(e,b)\big)$$

$$= \frac{۲۱}{۳} - ۷ = ۰$$

$D_d$ بزرگترین است و $D_d > ۰$. بنابراین، $d$ را به $C_j$ منتقل می‌کنیم. اکنون داریم:

$$C_i = \{a,c,e\}, C_j = \{b\} \cup \{d\} = \{b,d\}$$

● سومین تکرار.

$$D_a = \frac{1}{2}\big(d(a,c) + d(a,e)\big) - \frac{1}{2}\big(d(a,b) + d(a,d)\big)$$

$$= \frac{14}{2} - \frac{15}{2} = -0.5$$

$$D_c = \frac{1}{2}\big(d(c,a) + d(c,e)\big) - \frac{1}{2}\big(d(c,b) + d(c,d)\big)$$

$$= \frac{5}{2} - \frac{16}{2} = -13.5$$

$$D_e = \frac{1}{2}\big(d(e,c) + d(e,c)\big) - \frac{1}{2}\big(d(e,b) + d(e,d)\big)$$

$$= \frac{13}{2} - \frac{18}{2} = -2.5$$

همه منفی هستند. بنابراین توقف کرده و خوشه‌های $C_i$ و $C_j$ را تشکیل می‌دهیم.

۳. برای تقسیم $C_i$ و $C_j$، قطر آن‌ها را محاسبه می‌کنیم:

$$diameter(C_i) = \max\{d(a,c), d(a,e), d(c,e)\}$$

$$= \max\{3, 11, 2\}$$

$$= 11$$

$$diameter(C_j) = \max\{d(b,d)\}$$

$$= 5$$

خوشه‌ای با بیشترین قطر $C_i$ است. بنابراین ما اکنون $C_i$ را تقسیم می‌کنیم. با گرفتن $C_l = \{a, c, e\}$ محاسبات باقیمانده به عنوان تمرین به خواننده سپرده می‌شود.

## پیچیدگی زمانی خوشه‌بندی‌های سلسله‌مراتبی

تکنیک‌های تجمیعی و تقسیمی در نیازهای محاسباتی خود بسیار متفاوت هستند. فرض کنید $N$ نقطه داده داریم، از این رو، تعداد ادغام‌های احتمالی که یک الگوریتم تجمیعی باید در مرحله اول در نظر بگیرد $\frac{N(N-1)}{2}$ است (هر دو نقطه داده را می‌توان با هم ادغام کرد تا یک خوشه به اندازه دو تشکیل شود). در مجموع $N-1$ ادغام باید انجام شود تا سلسله‌مراتبِ کاملِ خوشه ساخته شود و در مجموع $O(N^3)$ ادغام احتمالی باید در نظر گرفته شود. به‌طور کلی، رویکرد تجمیعی به زمانی که نیاز دارد $O(N^3)$ یا $O(N^2 log N)$ است که بستگی به این دارد که آیا بعد از هر ادغام باید تمام فواصل بین همه خوشه‌ها محاسبه شود یا خیر. در مقابل تعداد تقسیم‌های ممکنی که در یک الگوریتم تقسیمی باید به تنهایی در مرحله اول در نظر گرفته شود، $O(2^N)$

است. بنابراین، الگوریتم‌های تقسیمی که هر تقسیم ممکن را برای یافتن تقسیم بهینه در نظر می‌گیرند، برای مجموعه‌های داده با اندازه متوسط **رام‌نشدنی**[1] هستند.

## مزایا و معایب خوشه‌بندی سلسله‌مراتبی

**مزایا**

- نیازی به مشخص کردن تعداد خوشه‌ها نیست.
- درخت‌واره‌نگار، می‌تواند اطلاعات مفیدی به شما بدهد.
- درک و پیاده‌سازی آن آسان است.

**معایب**

- به داده‌ها در مقیاس‌های مختلف حساس است.
- از منظرِ محاسباتی در مجموعه داده‌های بزرگ هزینه‌ی زیادی دارد.
- به موارد دورافتاده حساس است.
- به ندرت بهترین راه‌حل را می‌دهد.
- با مجموعه داده‌های بزرگ، تعیین تعداد خوشه‌های مناسب از درخت‌واره‌نگار دشوار است.

# خوشه‌بندی مبتنی‌بر چگالی

رویکرد خوشه‌بندی مبتنی بر چگالی، روشی است که قادر به یافتن خوشه‌هایی با شکل دلخواه است و همان‌طور که از نام آن پیداست، از چگالی نمونه‌ها برای اختصاص عضویت در خوشه استفاده می‌کند. این الگوریتم‌ها فرض می‌کنند ساختار خوشه‌بندی را می‌توان با چگالی توزیع‌های نمونه تعیین کرد. به طور معمول، الگوریتم‌های خوشه‌بندی مبتنی‌بر چگالی، ارتباط بین نمونه‌ها را از منظر چگالی ارزیابی کرده و با افزودن نمونه‌های قابل ارتباط، خوشه‌ها را گسترش می‌دهند.

روش‌های مختلفی برای اندازه‌گیری چگالی وجود دارد، اما می‌توانیم آن را به عنوان تعداد نمونه‌هایِ در واحدِ حجمِ فضایِ ویژگی تعریف کنیم. می‌توان گفت، مناطقی از فضای ویژگی که حاوی نمونه‌های زیادی هستند (نزدیک به هم قرار گرفته‌اند) دارای چگالی بالا هستند، در حالی که مناطقی از فضای ویژگی که حاوی موارد کمی هستند یا هیچ موردی ندارند، دارای چگالی کم هستند. شهود ما در اینجا بیان می‌کند که خوشه‌های متمایز در یک مجموعه داده با مناطقی با چگالی بالا نشان داده می‌شوند و با مناطق با چگالی کم از هم جدا می‌شوند. الگوریتم‌های خوشه‌بندی مبتنی بر چگالی تلاش می‌کنند تا این مناطق متمایز با چگالی بالا را یاد بگیرند و آن‌ها

---

[1] intractable

را به خوشه‌ها افراز کنند. الگوریتم‌های خوشه‌بندی مبتنی بر چگالی چندین ویژگی خوب دارند. یک الگوریتم مبتنی‌بر چگالی فقط به یک پویش از مجموعه داده‌های اصلی نیاز دارد و می‌تواند نویز را مدیریت کند. علاوه بر این‌ها، تعداد خوشه‌ها در این روش مورد نیاز نیست، چراکه الگوریتم‌های خوشه‌بندی مبتنی‌بر چگالی می‌توانند به‌طور خودکار تعداد خوشه‌ها را شناسایی کنند.

برخلاف بسیاری دیگر از الگوریتم‌های خوشه‌بندی سنتی، الگوریتم‌های خوشه‌بندی مبتنی‌بر چگالی توانایی مقابله با موارد دورافتاده را دارند. در خوشه‌بندی مبتنی‌بر چگالی، نقاط دورافتاده به عنوان نمونه‌هایی در نظر گرفته می‌شوند که به مناطق پراکنده (خلوت) تعلق دارند و در نتیجه باعث ایجاد این شهود می‌شوند که در مقایسه با سایر نمونه‌ها از مکانیسم‌های متفاوتی ایجاد می‌شوند.

محبوب‌ترین و رایج‌ترین روش خوشه‌بندی مبتنی‌بر چگالی DBSCAN است که در ادامه این بخش به تشریح آن می‌پردازیم

## الگوریتم DBSCAN

DBSCAN یک الگوریتم خوشه‌بندی مبتنی‌بر چگالی است که چگالی توزیع‌های نمونه را با یک جفت پارامتر "همسایگی" ($\varepsilon, \mu$) مشخص می‌کند. برای درک الگوریتم DBSCAN، ابتدا باید این دو ابرپارامتر را درک کنید. الگوریتم با انتخاب یک نمونه از داده‌ها و جستجوی موارد دیگر در یک شعاع جستجو شروع می‌شود. این ابرپارامتر شعاع اپسیلون ($\varepsilon$) است. ابرپارامتر $\mu$ حداقل تعداد نقاط (موارد) را که یک خوشه باید داشته باشد تا خوشه ایجاد شود را مشخص می‌کند. بنابراین ابرپارامتر $\mu$ یک عدد صحیح است. اگر یک مورد خاص دارای حداقل موارد $\mu$ در داخل شعاع اپسیلون خود باشد، آن مورد یک نقطه مرکزی در نظر گرفته می‌شود.
با توجه به مجموعه داده $D = \{x_1, x_2, \ldots, x_m\}$ مفاهیم زیر را تعریف می‌کنیم:

- **$\varepsilon$ـ همسایگی:** $\varepsilon$ـ همسایگی $p \epsilon D$ که با $N_\varepsilon(p)$ نمایش داده می‌شود، به صورت زیر تعریف می‌شود:

$$N_\varepsilon(p) = \{q \epsilon D | dist(p, q) \leq \varepsilon\}$$

- **خصوصیات نقاط هسته:** هر نمونه در $D$ بسته به همسایگی‌اش به عنوان نقطه مرکزی، نقطه مرزی یا نقطه نویز طبقه‌بندی می‌شود. یک نمونه $p$ یک نقطه مرکزی است اگر بیش از $\mu$ نمونه در $\varepsilon$ـ همسایگی خود داشته باشد. اگر $p$ کمتر از $\mu$ نمونه در داخل $\varepsilon$همسایگی خود داشته باشد و هیچ یک از همسایگان آن نمونه‌های مرکزی نباشند، آنگاه $p$ به عنوان نمونه نویز یا دورافتاده طبقه‌بندی می‌شود. در غیر این صورت، $p$ یک نمونه مرزی نامیده می‌شود.

- **مستقیما قابل‌دستیابی به چگالی:** یک نمونه $q \epsilon D$ مستقیما از نمونه $p \epsilon D$ قابل دستیابی است و با $q \triangleright p$ نشان داده می‌شود، اگر و تنها اگر، $|N_\varepsilon(p)| \geq \mu$ و $q \epsilon N_\varepsilon(p)$.

- **متصل به چگالی:** دو نمونه $p$ و $q$ متصل به چگالی هستند و با $p \bowtie q$ نشان داده می‌شود، اگر دنباله‌ای ($x_1, x_2, ..., x_m$) از نمونه‌ها وجود داشته باشد به طوری که:
$$\forall_{x_i}: |N_\varepsilon(p)| \geq \mu \quad q \triangleright x_m \triangleright \cdots \triangleright x_1 \triangleright p.$$

- **خوشه:** یک خوشه به عنوان مجموعه حداکثری از نمونه‌های متصل به چگالی تعریف می‌شود و از نمونه‌های مرکزی و نمونه‌های مرزی تشکیل شده است. در DBSCAN، یک نمونه مرزی به ترتیب نمونه‌ها می‌تواند به چندین خوشه تعلق داشته باشد. یک نمونه نویز به هیچ خوشه‌ای تعلق ندارد و به آن دورافتاده می‌گویند. یک زیرمجموعه $C \subseteq D$ اگر دو شرط زیر را داشته باشد، خوشه نامیده می‌شود:

  ۱. **بیشینگی:** $\forall_p \epsilon C: p, \forall_q \epsilon C: p \backslash C: \neg p \bowtie q$

  ۲. **همبندی:** $\forall_{p,q} \epsilon C: p \bowtie q$

DBSCAN از ساختار داده‌ای به نام لیست بذر S استفاده می‌کند که شامل مجموعه‌ای از نمونه دانه برای گسترش خوشه است. برای ساخت یک خوشه، DBSCAN به طور تصادفی یک نمونه پردازش نشده را انتخاب می‌کند و آن را به عنوان مقداردهی اولیه در لیست خالی S قرار می‌دهد. سپس، به طور مداوم یک نمونه $p$ را از $S$ استخراج می‌کند و پرس‌وجوی $\varepsilon$ـ محدوده را در $p$ انجام می‌دهد تا نمونه‌هایی را پیدا کند که مستقیما از $p$ قابل دستیابی هستند و اگر تاکنون پردازش نشده باشند، آن‌ها را در $S$ قرار می‌دهد. هنگامی که لیست دانه $S$ خالی است، ساخت خوشه کامل شده و ساخت یک خوشه جدید آغاز می‌شود. کل فرآیند گسترش تا زمانی که همه نمونه‌ها برچسب‌گذاری شوند تکرار می‌شود.

مجموعه داده هندوانه در جدول ۸ـ۱ را به عنوان مثال در نظر می‌گیریم تا نمایش دقیق‌تری ارائه دهیم. فرض کنید که ابرپارامترهای همسایگی ($\mu = 5, \varepsilon = 0.11$) باشد. ما با پیدا کردن $\varepsilon$ـ همسایگی برای هر نمونه شروع می‌کنیم تا بتوانیم مجموعه نمونه‌های هسته (مرکزی) را شناسایی کنیم:

$$\Omega = \{x_3, x_5, x_6, x_8, x_9, x_{13}, x_{14}, x_{18}, x_{19}, x_{24}, x_{25}, x_{28}, x_{29}\}$$

سپس، به‌طور تصادفی یک نمونه هسته را به‌عنوان بذر انتخاب می‌کنیم و آن را بسط می‌دهیم تا همه‌ی نمونه‌های قابل دستیابی با چگالی را شامل شود. این نمونه‌ها یک خوشه را تشکیل می‌دهند. بدون از دست دادن کلیت، فرض کنید نمونه مرکزی $x_8$ به عنوان بذر انتخاب شده است، از این‌رو، اولین خوشه به‌صورت زیر تولید شده است:

$$C_1 = \{x_6, x_7, x_8, x_{10}, x_{12}, x_{18}, x_{19}, x_{20}, x_{23}\}$$

پس از آن، DBSCAN تمام نمونه‌های اصلی در $C_1$ را از $\Omega$ حذف می‌کند:

$$\Omega = \Omega \setminus C_\mathtt{1} = \{x_\mathtt{3}, x_\mathtt{5}, x_\mathtt{9}, x_\mathtt{13}, x_\mathtt{14}, x_\mathtt{24}, x_\mathtt{25}, x_\mathtt{28}\}$$

سپس، خوشه بعدی با انتخاب تصادفی یک نمونه هسته دیگر از $\Omega$ بروزرسانده به عنوان بذر، تولید می‌شود. این روند تا زمانی تکرار می‌شود که عنصر دیگری در $\Omega$ وجود نداشته باشد. شکل ۸ـ۱۱ خوشه‌های تولید شده در دورهای مختلف را نشان می‌دهد. علاوه بر $C_\mathtt{1}$، سه خوشه دیگر نیز تولید شده‌اند:

$$C_\mathtt{2} = \{x_\mathtt{3}, x_\mathtt{4}, x_\mathtt{5}, x_\mathtt{9}, x_\mathtt{13}, x_\mathtt{14}, x_\mathtt{16}, x_\mathtt{17}, x_\mathtt{21}\}$$

$$C_\mathtt{3} = \{x_\mathtt{1}, x_\mathtt{2}, x_\mathtt{22}, x_\mathtt{26}, x_\mathtt{29}\}$$

$$C_\mathtt{4} = \{x_\mathtt{24}, x_\mathtt{25}, x_\mathtt{27}, x_\mathtt{28}, x_\mathtt{30}\}.$$

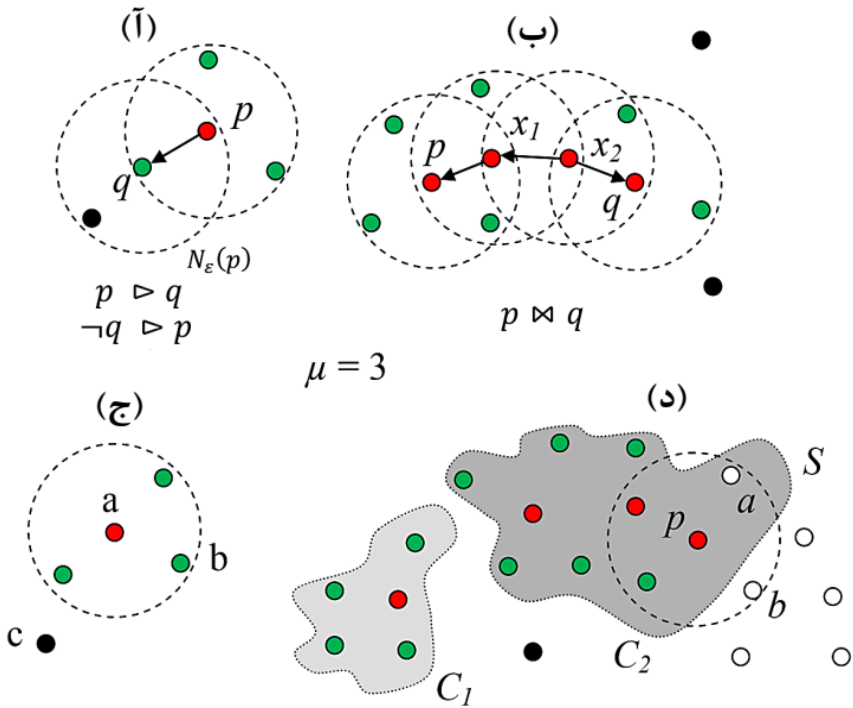

**شکل ۸ـ۱۰. مفاهیم DBSCAN.** (آ) $q$ به طور مستقیم از $p$ قابل دستیابی است. (ب) $p$ و $q$ به هم متصل به چگالی هستند. (ج) نمونه $a$ (قرمز) یک نمونه مرکزی (هسته) است، $b$ (سبز) نمونه مرزی است، $c$ (سیاه) نمونه نویز است. (د) لیست دانه $S$ برای‌گسترش خوشه. DBSCAN در حال حاضر در حال ساخت خوشه $C_\mathtt{2}$ است. نمونه $p$ از $S$ استخراج شده و مورد بررسی قرار می‌گیرد. نمونه‌های $a$ و $b$ که در همسایگی $p$ قرار دارند پردازش نمی‌شوند و بنابراین در $S$ قرار می‌گیرند.

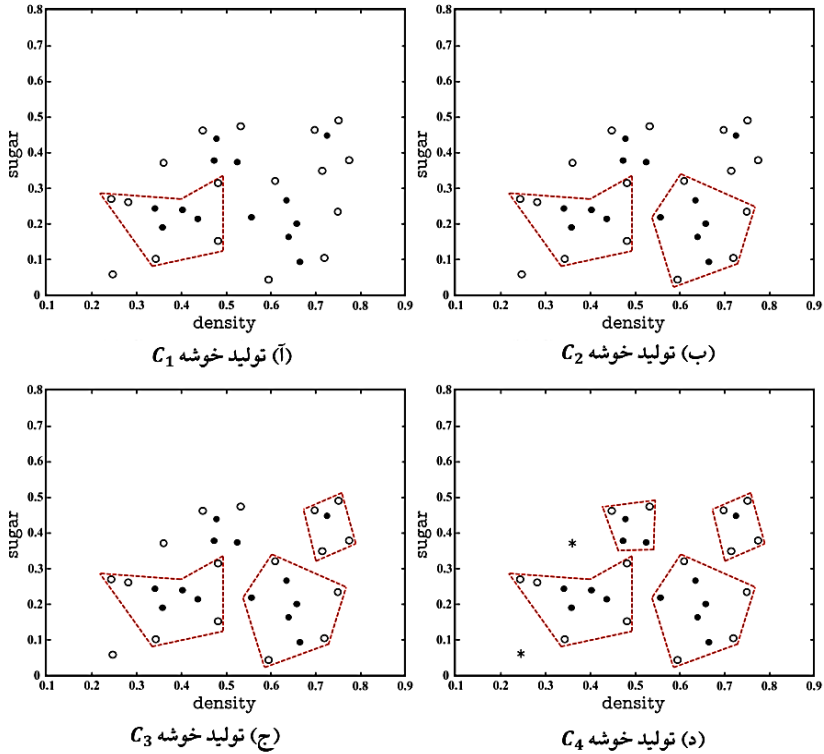

**شکل ۸ـ۱۱. نتایج الگوریتم DBSCAN با $\mu = 5$ و $\varepsilon = 0.11$. نمادهای "●"، "○"، "\*" به ترتیب
نشان‌دهنده نمونه‌های مرکزی (هسته)، نمونه‌های غیر مرکزی و نمونه‌های نویزدار هستند . خطوط چین‌دار،
خوشه‌ها را نشان می‌دهد.**

## مزایای DBSCAN

- مفهومی به نام نویز دارد، از این‌رو نسبت به موارد دورافتاده قوی است.
- نیازی به تعیین تعداد خوشه‌ها از قبل ندارد.

## معایب DBSCAN

- نمی‌تواند مجموعه داده‌ها با تفاوت‌های زیاد در چگالی را به خوبی خوشه‌بندی کند.
- کاملا قطعی نیست. بنابراین نقاط مرزی که از بیش از یک خوشه قابل دستیابی هستند،
  می‌توانند بخشی از هر خوشه باشند.

## پیچیدگی زمان و فضا در الگوریتم DBSCAN

پیچیدگی زمانی الگوریتم DBSCAN برابر ($m$ × زمان برای یافتن نقاط در $\varepsilon$ ـ همسایگی)$O$
است که $m$ تعداد نقاط است. در بدترین حالت، این پیچیدگی $O(m^2)$ است. با این حال، در
فضاهای کم‌بعد (مخصوصا فضای دوبعدی)، داده‌ساختارهایی مانند kd ـ درختان امکان بازیابی

کارآمد همه نقاط در یک فاصله معین از یک نقطه مشخص را فراهم می‌کنند و پیچیدگی زمانی می‌تواند تا $O(mlogm)$ در حالت متوسط کم شود. فضای مورد نیاز DBSCAN، حتی برای داده‌های با ابعاد بالا، $O(m)$ است، زیرا لازم است فقط مقدار کمی از داده‌ها برای هر نقطه، یعنی برچسب خوشه و شناسایی هر نقطه به عنوان مرکز، مرز یا نقطه نویز نگهداری شود. توجه به این نکته ضروری است که پیچیدگی زمانی معیارهای تشابه بین اشیا در اینجا در نظر گرفته نشده است. با فرض اینکه معیار تشابه بین اشیا دارای پیچیدگی زمانی $\Psi$ باشد، پیچیدگی نهایی DBSCAN برابر $O(\Psi m^2)$ یا $O(\Psi mlogm)$ است.

# خوشه‌بندی با پایتون

## خوشه‌بندی k-means

### وارد کردن کتاب‌خانه‌ها

```
In [1]:    import matplotlib.pyplot as plt
           import numpy as np
           from sklearn.cluster import KMeans
```

### آماده‌سازی داده‌ها

مرحله بعدی آماده‌سازی داده‌هایی است که می‌خواهیم خوشه‌بندی کنیم. بیایید یک آرایه numpy از ۲۰ ردیف و ۲ ستون ایجاد کنیم.

```
In [1]:    X = np.array([[1,3],
              [11,16],
              [16,10],
              [20,10],
              [35,23],
              [75,78],
              [69,82],
              [63,75],
              [65,70],
              [83,98],
              [71,96],
              [25,18],
              [18,5],
              [92,98],
              [67,67],
              [5,3],
              [13,17],
              [19,16],
              [24,10],
              [30,45]])
```

**مصورسازی داده‌ها**

بیایید این نقاط را ترسیم کنیم و بررسی کنیم که آیا می‌توانیم خوشه‌ای را مشاهده کنیم. برای این کار کد زیر را اجرا کنید:

In [2]: ```python
plt.scatter(X[:,0],X[:,1], label='True Position',marker = 's')
```

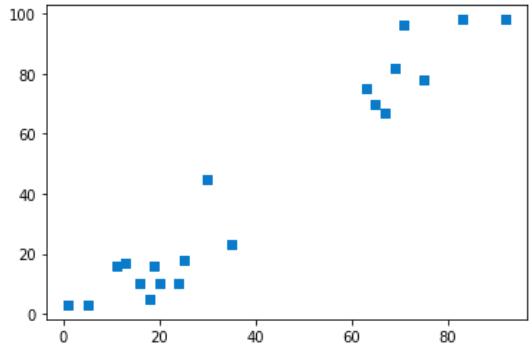

همان‌طور که در شکل فوق مشاهده می‌شود، اگر بخواهیم دو خوشه از نقاط داده را تشکیل دهیم، احتمالاً یک خوشه از هشت نقطه در بالا سمت راست و یک خوشه از دوازده نقطه در پایین سمت چپ ایجاد خواهیم کرد. بیایید ببینیم آیا الگوریتم خوشه‌بندی K-means ما همین کار را می‌کند یا خیر.

**ایجاد خوشه‌ها**

برای اجرای الگوریتم خوشه‌بندی K-means با دو خوشه، به سادگی کد زیر را اجرا کنید:

In [4]: ```python
kmeans = KMeans(n_clusters=2)
kmeans.fit(X)
```

در خط اول، یک شی KMeans ایجاد می‌کنید و عدد ۲ را به عنوان مقدار پارامتر تعداد خوشه‌ها n_clusters ارسال می‌کنید. بعد، شما به سادگی باید متد fit را روی kmeans فراخوانی کنید و داده‌هایی که می‌خواهید خوشه‌بندی کنید را به آن ارسال کنید. در این مثال داده‌ها در آرایه X هستند که قبلاً آن‌ها ایجاد کردیم.

حالا بیایید ببینیم که الگوریتم ایجاد شده برای خوشه‌های نهایی چه مقادیری از مرکز را دارند.

In [3]: ```python
print(kmeans.cluster_centers_)
```

Out [7]: ```
[[18.08333333 14.66666667]
 [73.125      83.        ]]
```

خروجی یک آرایه دو بعدی به شکل ۲×۲ است. در اینجا ردیف اول حاوی مقادیر مختصات مرکز اول یعنی (۱۴.۱۶۶۶۶۶۷، ۱۸.۰۸۳۳۳۳۳) و ردیف دوم حاوی مقادیر مختصات مرکز دیگر یعنی (۸۳.۰، ۷۳.۱۲۵) است.

برای دیدن برچسب‌های نقاط داده، کد زیر را اجرا کنید:

```
In [3]:   print(kmeans.labels_)

Out [7]:  [0 0 0 0 0 1 1 1 1 1 1 0 0 1 1 0 0 0 0 0]
```

خروجی یک آرایه یک بعدی از ۲۰ عنصر مربوط به خوشه‌های اختصاص داده شده به ۲۰ نقطه داده ما است. *در اینجا ۰ و ۱ صرفا برای نشان دادن شناسه‌های خوشه‌ها استفاده می‌شوند و هیچ اهمیت ریاضی ندارند. اگر سه خوشه وجود داشت، خوشه سوم با رقم ۲ نشان داده می‌شد.*

**مصورسازی داده‌ها**

بیایید دوباره نقاط داده را روی نمودار رسم کنیم و نحوه خوشه‌بندی داده‌ها را مصورسازی کنیم. این بار داده‌ها را به همراه برچسب اختصاص داده‌شده ترسیم می‌کنیم تا بتوانیم بین خوشه‌ها تمایز قائل شویم. کد زیر را اجرا کنید:

```
In [2]:   plt.scatter(X[:,0],X[:,1], c=kmeans.labels_, cmap='rainbow',marker = 's')
```

در اینجا ما ستون اول آرایه X را در مقابل ستون دوم ترسیم می‌کنیم، همچنین در این مورد kmeans.labels_ را نیز به عنوان مقدار پارامتر c که مربوط به برچسب‌ها است، ارسال می‌کنیم. پارامتر 'cmap='rainbow' برای انتخاب نوع رنگ برای نقاط داده مختلف و برای نشانگذار نقاط، پارامتر 's' = marker ارسال شده است. خروجی به‌صورت زیر است:

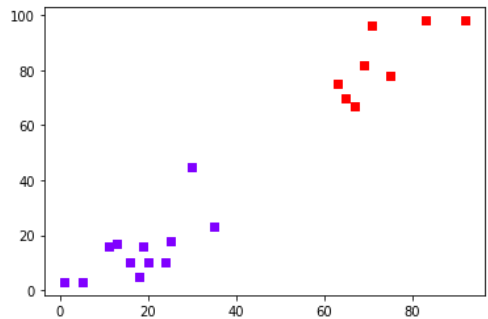

همانطور که انتظار می‌رفت و در شکل فوق مشاهده می‌شود، دوازده نقطه پایین سمت چپ باهم گروه‌بندی شده و نقاط باقی‌مانده در سمت راست بالا باهم در یک خوشه قرار گرفته‌اند.

حالا بیایید نقاط را به همراه مختصات مرکز هر خوشه ترسیم کنیم تا ببینیم موقعیت مرکز چگونه بر خوشه‌بندی تاثیر می‌گذارد. کد زیر را برای رسم نمودار اجرا کنید:

```
In [2]:   plt.scatter(X[:,0], X[:,1], c=kmeans.labels_, cmap='rainbow',marker = 's')
          plt.scatter(kmeans.cluster_centers_[:,0] ,kmeans.cluster_centers_[:,1],
          color='black',marker = '+')
```

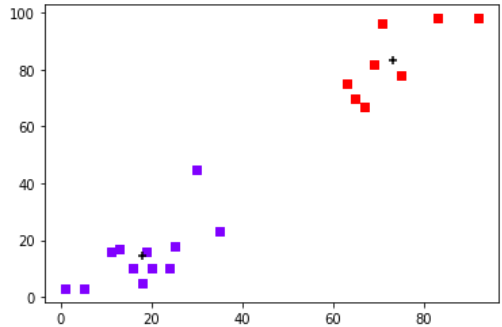

بار دیگر، الگوریتم K-means را با ۳ خوشه اجرا می‌کنیم. نمودار خروجی زیر را بدست می‌آید:

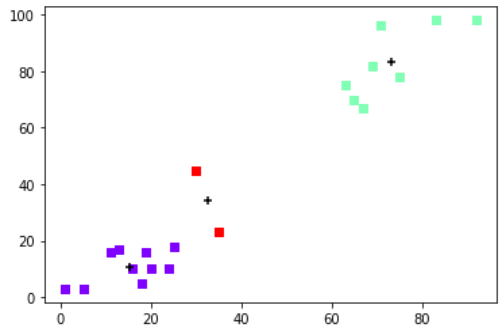

می‌بینید که دوباره، نقاطی که بهم نزدیک هستند در کنار هم قرار گرفته‌اند.

به عنوان تمرین می‌توان این داده‌ها را با الگوریتم‌های دیگر با استفاده از کتابخانه scikit-learn خوشه‌بندی کرد. می‌توان از قطعه کدهای زیر استفاده کنید. همچنین می‌توانید پارامترهای مختلف هر الگوریتم خوشه‌بندی را تغییر داده و نتایج را مشاهده کنید.

**خوشه‌بندی سلسله‌مراتبی**

```
In [4]:   from sklearn.cluster import AgglomerativeClustering
          cluster = AgglomerativeClustering(n_clusters=2,
          affinity='euclidean', linkage='single')
          cluster.fit_predict(X)
```

**خوشه‌بندی DBSCAN**

```
In [4]:   from sklearn.cluster import DBSCAN
          clustering = DBSCAN(eps=3, min_samples=2).fit(X)
          clustering.labels_
```

**خوشه‌بندی مدل مخلوط گاوسی**

```
In [4]:   from sklearn import mixture
          model = mixture.GaussianMixture(n_components=2,
          covariance_type='full').fit(X)
          labels = model.predict(X)
```

# چه زمانی از خوشه‌بندی استفاده کنیم؟

مهم نیست که چه نوع تحقیقی انجام می‌دهید یا مدل یادگیری ماشین شما چه وظیفه‌ای را بر عهده دارد، در یک نقطه از خط، به تکنیک‌های خوشه‌بندی نیاز خواهید داشت. چرا که در بسیاری از مواقع (حداقل در ابتدا)، با مجموعه داده‌هایی که عمدتا بدون‌ساختار هستند و دسته‌بندی نشده‌اند، کار می‌کنید و خوشه‌بندی و آماده‌سازی داده‌ها در پیوند با یکدیگر و لازمه این کارند. مهم‌تر از آن، خوشه‌بندی روشی آسان برای انجام بسیاری از تحلیل‌های سطحی است که می‌تواند به شما یک برد سریع در زمینه‌های مختلف بدهد. به عنوان مثال، بیمه‌گران می‌توانند به سرعت عوامل خطر و مکان‌ها را بررسی کنند و یک پروفایل ریسک اولیه برای متقاضیان ایجاد کنند.

در علم داده، می‌توانیم از تحلیل خوشه‌ای برای بدست آوردن بینش‌های ارزشمند از داده‌های خود با دیدن اینکه نقاط داده در هنگام اعمال یک الگوریتم خوشه‌بندی در چه گروه‌هایی قرار می‌گیرند، استفاده کنیم. به عبارت دیگر، خوشه‌بندی و تجزیه و تحلیل داده‌ها می‌تواند فرآیند مدیریت داده شما را متحول کند. از آنجایی که خوشه‌بندی توسط الگوریتم انجام می‌شود، این احتمال وجود دارد که بتوانید همبستگی‌های ناشناخته قبلی را در داده‌ها کشف کنید که می‌تواند به شما در برخورد با یک چالش تجاری از دیدگاه جدید کمک کند.

وقتی نوبت به داده‌کاوی یا استخراج داده می‌شود، می‌توانید از خوشه‌بندی داده‌ها به‌عنوان یک ابزار مستقل برای بدست آوردن بینش در مورد توزیع داده‌ها یا برای تقویت خوشه‌های خاصی که می‌خواهید تجزیه و تحلیل بیشتری روی آنها انجام دهید، استفاده کنید. شما همچنین می‌توانید از آن در هوش تجاری برای سازماندهی جدید مشتریان، سازماندهی پروژه‌های معلق و برنامه‌های متعدد دیگر استفاده کنید. خوشه‌بندی با به حداقل رساندن تعداد پویش‌های مورد نیاز برای جستجوی داده‌ها، به کارآمدتر شدن داده‌کاوی کمک می‌کند.

با وجود تمام کارهای بزرگی که تحلیل خوشه‌ای می تواند برای پروژه و سازمان شما انجام دهد، به همان اندازه چیزهایی وجود دارد که وقتی به دنبال بینش عمیق هستید، آن را نه چندان مطلوب می‌کند. به عبارت دیگر، خوشه‌بندی به خودی خود چالش‌های مهمی را ایجاد می‌کند و آن را برای کارهای پیچیده‌تر یادگیری ماشین و تجزیه و تحلیل ایده‌آل نمی‌کند. بزرگترین مشکلی که در اکثر روش‌های خوشه‌بندی مطرح می‌شود این است که اگرچه در ابتدا داده‌های شما را به زیرمجموعه‌ها تفکیک می‌کنند، استراتژی‌های مورد استفاده گاهی اوقات لزوما به خود داده‌ها مربوط نمی‌شوند، بلکه به موقعیت آن در رابطه با سایر نقاط مربوط می‌شوند. برای مثال، خوشه‌بندی K-means، بسته به تعداد گروه‌هایی که تنظیم می‌کنید، می‌تواند نتایج بسیار متفاوتی داشته باشد و معمولا وقتی با خوشه‌های غیرکروی استفاده می‌شود، کارکرد خوبی ندارد. علاوه بر این، این مساله که مرکز جرم‌ها به صورت تصادفی تنظیم می‌شوند نیز بر نتایج تاثیر می‌گذارد

و می‌تواند منجر به مشکلاتی شود. الگوریتم‌های دیگر می‌توانند این مشکل را حل کنند، اما نه بدون هزینه.

خوشه‌بندی سلسله مراتبی تمایل به تولید نتایج دقیق‌تری دارد، اما به قدرت محاسباتی قابل توجهی نیاز دارد و زمانی که با مجموعه داده‌های بزرگ‌تر کار می‌کنید ایده‌آل نیست. این روش به مقادیر دورافتاده نیز حساس است و در نتیجه می‌تواند خوشه‌های نادرست ایجاد کند. با این حال، این‌ها به این معنی نیست که هرگز نباید از خوشه‌بندی استفاده کنید، بلکه باید آن را در جایی و زمانی که بیشترین تاثیر و بینش را به شما می‌دهد، به کار ببرید. همچنین، موقعیت‌های زیادی وجود دارد که در آن خوشه‌بندی نه تنها می‌تواند نقطه شروع خوبی به شما بدهد، بلکه ویژگی‌های مهم داده‌های شما را که می‌توان با تجزیه و تحلیل عمیق‌تر بهبود بخشید، روشن‌تر می‌کند. این‌ها فقط برخی از مواقعی هستند که باید از خوشه‌بندی استفاده کنید:

- **زمانی که با یک مجموعه داده بزرگ و بدون ساختار شروع می‌کنید:** همانند سایر ابزارهای یادگیری غیرنظارتی، خوشه‌بندی می‌تواند مجموعه داده‌های بزرگی را بگیرد و بدون آموزش، آن‌ها را به سرعت به موارد قابل استفاده‌تر سازماندهی کند. مزیت این امر این است که اگر به دنبال انجام یک تجزیه و تحلیل گسترده نیستید، خوشه‌بندی می‌تواند به شما پاسخ‌های سریعی را در مورد داده‌های شما بدهد.

- **زمانی که نمی‌دانید داده‌های شما به چند یا کدام دسته تقسیم می‌شود:** حتی اگر با مجموعه داده‌های ساختاریافته‌تر شروع می‌کنید، ممکن است هنوز دسته‌بندی مورد نظر شما را نداشته باشد. خوشه‌بندی اولین قدم عالی در آماده‌سازی داده‌های شما است، چراکه شروع به پاسخ دادن به سوالات کلیدی در مورد مجموعه داده شما می‌کند. برای مثال، ممکن است متوجه شوید که آنچه فکر می‌کردید دو زیرمجموعه اصلی می‌باشد، در واقع بیشتر است.

- **زمانی که تقسیم دستی و حاشیه‌نویسی داده‌های شما بیش از حد نیاز به هزینه و زمان دارد:** برای مجموعه داده های کوچک‌تر، حاشیه‌نویسی و سازماندهی دستی، اگر ایده آل نباشد، امکان‌پذیر است. با این حال، با شروع افزایشی داده‌های شما، حاشیه‌نویسی آن‌ها به طور تصاعدی سخت‌تر می‌شود. خوشه‌بندی (بسته به الگوریتمی که استفاده می‌کنید)، می‌تواند زمان حاشیه‌نویسی شما را کاهش دهد. چراکه کمتر به نتایج خاص علاقه دارد و بیشتر به خود طبقه‌بندی توجه دارد.

- **وقتی به دنبال ناهنجاری در داده‌های خود هستید:** نکته جالب اینجاست که یکی از با ارزش‌ترین کاربردهای خوشه‌بندی این است که به دلیل حساسیت زیاد الگوریتم‌ها به نقاط دورافتاده، آن‌ها می‌توانند به عنوان شناسه‌ای برای ناهنجاری‌های داده عمل کنند. درک داده‌های نابهنجار می‌تواند به شما کمک کند داده‌های موجود خود را سازگارتر کنید و به نتایج دقیق‌تری برسید.

زمانی که خوشه‌بندی را نه به‌عنوان یک مدل مستقل، بلکه به‌عنوان بخشی از استراتژی کشف داده‌های گسترده‌تر به کار می‌برید، بیشترین بهره را از خوشه‌بندی خواهید برد.

# کاهش ابعاد

در یادگیری ماشین، "بُعد" به تعدادِ ویژگی‌ها (متغیرهای ورودی) در مجموعه داده اشاره دارد. وقتی تعداد ویژگی‌ها نسبت به تعداد نمونه‌ها در مجموعه داده شما زیادتر است، نیاز به افزایش تعداد نمونه‌ها است تا مدل به بهترین کارآیی دست یابد. به عبارت دیگر، با افزایش تعداد ویژگی‌ها، تعداد نمونه‌ها نیز به نسبت افزایش می‌یابد. زیرا، هرچه ویژگی‌های بیشتری داشته باشیم، تعداد نمونه‌های بیشتری نیاز خواهیم داشت تا همه‌یِ ترکیب‌های مقادیر ویژگی به خوبی در مجموعه داده‌ها نشان داده شوند. علاوه بر این، هرچه تعداد ویژگی‌ها بیشتر باشد، امکان بیش‌برازش بیشتر می‌شود. یک مدل یادگیری ماشین که بر روی تعداد زیادی ویژگی آموزش دیده است، به‌طور فزاینده‌ای به داده‌هایی که روی آن‌ها آموزش داده شده است وابسته می‌شود و به نوبه منجر به بیش‌برازش می‌شود، که در نتیجه عملکرد ضعیفی برروی داده‌های دیده‌نشده خواهد داشت. اجتناب از بیش‌برازش، انگیزه اصلی برای انجام کاهش ابعاد است. هر چه داده‌های آموزشی ما ویژگی‌های کم‌تری داشته باشد، مفروضات مدل ماکم‌تر و ساده‌تر خواهد بود. اما این همه چیز نیست و کاهش ابعاد مزایای بسیار بیشتری برای ارائه دارد. هرچند، وقتی ابعاد یک مجموعه داده را کاهش می‌دهیم، درصدی از انعطاف‌پذیریِ داده‌های اصلی را از دست می‌دهیم. با این حال، نگران از دست دادن این درصد از انعطاف‌پذیری در داده‌های اصلی نباشید، چراکه کاهش ابعاد مزایای زیادی دارد.

داده‌ها پایه و اساس هر الگوریتم یادگیری ماشین را تشکیل می‌دهند، بدون آن، علم داده نمی‌تواند اتفاق بیفتد. گاهی اوقات، ممکن است این مجموعه داده‌ها دارای تعداد زیادی ویژگی باشد که برخی از آن‌ها حتی مورد نیاز نیستند. چنین اطلاعات اضافی، مدل‌سازی را پیچیده می‌کند. علاوه بر این، تفسیر و درک داده‌ها از طریق مصورسازی به‌دلیل ابعاد بالا دشوار می‌شود. *اینجاست که کاهش ابعاد مطرح می‌شود.*

تعداد ابعاد کم‌تر در داده‌ها به معنای زمان آموزش کم‌تر و منابع محاسباتی کم‌تر است. مسائلِ یادگیری ماشین که شامل ویژگی‌های زیادی هستند، آموزش را بسیار کند می‌کند. در یک مجموعه داده با ابعاد بالا، بیشتر نقاط داده احتمالاً از یکدیگر دور هستند. بنابراین، الگوریتم‌ها نمی‌توانند به طور موثر و کارآمدی روی داده‌های با ابعاد بالا آموزش ببینند (مشکل مشقت بعدچندی).

کاهش ابعاد از مشکل بیش‌برازش جلوگیری می‌کند. وقتی ویژگی‌های زیادی در داده‌ها وجود داشته باشد، مدل‌ها پیچیده‌تر می‌شوند و تمایل دارند که روی داده‌های آموزشی بیش‌برازش کنند.

کاهش ابعاد برای مصورسازی داده‌ها بسیار مفید است. وقتی ابعاد داده‌هایی با ابعاد بالا را به دو یا سه جز کاهش دهیم، آنگاه داده‌ها را می‌توان براحتی در یک نمودار دو بعدی یا سه بعدی ترسیم کرد.

کاهش ابعاد نویز در داده‌ها را حذف می‌کند. با حفظ مهم‌ترین ویژگی‌ها و حذف ویژگی‌های اضافی، کاهش ابعاد نویز در داده‌ها را حذف می‌کند. در نتیجه، دقت مدل را بهبود می‌بخشد.

## تعریف ‖ کاهش ابعاد

کاهش ابعاد به فرآیند کاهش تعداد ویژگی‌ها در مجموعه داده‌ها اشاره دارد، در حالی که تا آنجا که ممکن است تغییرات در مجموعه داده اصلی حفظ می‌شود. فرآیند کاهش ابعاد اساسا داده‌ها را از فضای ویژگی‌های با ابعاد بالا به فضای ویژگی‌های با بعد کم‌تر تبدیل می‌کند. به طور همزمان، مهم است که ویژگی‌های معنی‌دار موجود در داده‌ها در طول تبدیل از بین نروند.

کاهش ابعاد یک مرحله پیش‌پردازش داده است. به این معنا که قبل از آموزش مدل، کاهش ابعاد را انجام می‌دهیم.

به‌طور کلی، دو رویکرد برای کاهش ابعاد وجود دارد: انتخاب ویژگی و استخراج (تبدیل) ویژگی. رویکرد انتخاب ویژگی سعی می‌کند یک زیرمجموعه از ویژگی‌های مهم را انتخاب و ویژگی‌های نه‌چندان مهم را به‌منظور کاهش پیچیدگی مدل، افزایش کارایی محاسباتی مدل و کاهش خطای تعمیم به دلیل ایجاد نویز، حذف کند. در مقابل، انتخاب ویژگی که همچنین به عنوان تبدیل ویژگی شناخته می‌شود، سعی می‌کند یک زیرفضای ویژگی جدید ایجاد کند. ایده اصلی پشت استخراج ویژگی فشرده‌سازی داده‌ها با هدف حفظ بیشتر اطلاعات مربوط است.

## تعریف ‖ انتخاب ویژگی

انتخاب ویژگی، فرآیند انتخاب خودکار یا دستی زیرمجموعه‌ای از مناسب‌ترین و مرتبط‌ترین ویژگی‌ها برای استفاده در ساخت مدل است

انتخاب ویژگی با گنجاندن ویژگی‌های مهم یا حذف ویژگی‌های نامربوط در مجموعه داده بدون تغییر آن‌ها انجام می‌شود.

## تعریف ‖ استخراج ویژگی

استخراج ویژگی، فرآیند کاهش تعداد ویژگی‌های یک مجموعه داده با ایجاد ویژگی‌های جدید از ویژگی‌های اصلی است.

اهداف اصلی کاهش ابعاد، عبارتند از: بهبود دقت عملکرد مدل پیش‌گویانه، کاهش زمان محاسبه و بهبود تفسیرپذیری مدل است.

# انتخاب ویژگی در مقابل استخراج ویژگی

هر دو روش تعداد ابعاد را کاهش می‌دهند اما به روش‌های مختلف. تمایز بین این دو نوع روش بسیار مهم است. هدف استخراج ویژگی کاهش تعداد ویژگی‌های یک مجموعه داده با ایجاد ویژگی‌های جدید از ویژگی‌های موجود (و سپس کنار گذاشتن ویژگی‌های اصلی) است. این مجموعه جدید کاهش یافته از ویژگی‌ها باید بتواند بیشتر اطلاعات موجود در مجموعه اصلی ویژگی‌ها را خلاصه کنند. به این ترتیب، یک نسخه خلاصه شده از ویژگی‌های اصلی می‌تواند از ترکیب مجموعه اصلی ویژگی‌ها ایجاد شود. در مقابل، هدف انتخاب ویژگی، مهم‌ترین ویژگی‌ها را در مجموعه داده نگه می‌دارد و ویژگی‌های اضافی را حذف می‌کند. تفاوت بین انتخاب ویژگی و استخراج ویژگی این است که هدف انتخاب ویژگی، رتبه‌بندی اهمیت ویژگی‌های موجود در مجموعه داده و کنار گذاشتن ویژگی‌های کم‌تر مهم است. به عبارت دیگر هیچ ویژگی جدیدی از این طریق ایجاد نمی‌شود. در مقابل، از استفاده از استخراج ویژگی، به ایجاد یک مجموعه کاملا جدیدی از ویژگی‌ها منجر می‌شود. رویکرد استخراج ویژگی را می‌توان به روش‌های خطی و روش‌های غیرخطی دسته‌بندی کرد. روش‌های غیرخطی به عنوان یادگیری منیفلد نیز شناخته می‌شوند.

کاهش ابعاد فرآیند کاهش تعداد ابعاد در داده‌ها یا با حذف ویژگی‌های کم‌تر مفید (انتخاب ویژگی) یا تبدیل داده‌ها به ابعاد پایین‌تر (استخراج ویژگی) است.

## تکنیک‌های انتخاب ویژگی

انتخاب ویژگی می‌تواند به صورت دستی یا به صورت استفاده از تکنیک‌ها (خودکار) رایجی که برای این منظور توسعه یافته‌اند، صورت گیرد. برای مثال، در نظر بگیرید که در حال تلاش برای ساخت مدلی هستید که وزن افراد را پیش‌بینی می‌کند و مجموعه بزرگی از داده‌ها را جمع‌آوری کرده‌اید که هر فرد را توصیف می‌کند. اگر ستونی داشتید که رنگ لباس هر فرد را توصیف می‌کند، آیا برای پیش‌بینی وزن آن‌ها کمک زیادی می‌کند؟ فکر می‌کنم با خیال راحت می‌توانیم توافق کنیم که اینطور نخواهد بود. این ویژگی است که ما می‌توانیم بدون هیچ مشکلی کنار بگذاریم. زمانی که ارتباط یا نامرتبط بودن ویژگی‌های خاص مشخص باشد، می‌توانیم این ویژگی‌ها را به‌صورت دستی انتخاب کنیم و ابعاد را کاهش دهیم و زمانی که به وضوح این ویژگی‌ها مشخص نیستند، تکنیک‌ها و ابزارهای زیادی وجود دارد که می‌توانیم برای کمک به انتخاب ویژگی‌ها استفاده کنیم.

تکنیک‌های انتخاب ویژگی می‌توانند غیرنظارتی یا بانظارت (مانند الگوریتم‌های ژنتیک) باشند. همچنین در صورت نیاز می‌توان چندین روش را ترکیب کرد. روند انتخاب ویژگی را می‌توان در دو مرحله توصیف کرد:

- ترکیبی از یک تکنیک جستجو برای پیشنهاد یک زیرمجموعه ویژگی جدید.
- معیار ارزیابی که امتیازی را به زیرمجموعه‌های مختلف ویژگی می‌دهد.

## الگوریتم ژنتیک

از منظر ریاضی، انتخاب ویژگی‌ها به عنوان یک مساله بهینه‌سازی ترکیباتی فرموله می‌شود. تابع هدف، عملکرد تعمیم مدل پیش‌گویانه است که با عبارت خطا در ویژگی‌های انتخابی یک مجموعه داده نشان داده می‌شود. یک انتخاب جامع از ویژگی‌ها، $2^N$ ترکیب مختلف را ارزیابی می‌کند، که در آن $N$ تعداد ویژگی‌ها است. این فرآیند، مستلزم کار محاسباتی زیادی است و اگر تعداد ویژگی‌ها زیاد باشد، انجام آن غیرعملی می‌شود. به عبارت دیگر، انتخاب ویژگی‌ها یک مساله NP-Hard است. از این‌رو، نیاز به روش‌های هوشمندی داریم که امکان انتخاب ویژگی‌ها را در عمل فراهم کنند. یکی از پیشرفته‌ترین الگوریتم‌ها برای انتخاب ویژگی، الگوریتم ژنتیک است.

الگوریتم ژنتیک یک روش تصادفی برای بهینه‌سازی تابع بر اساس مکانیک ژنتیک طبیعی و تکامل بیولوژیکی است. در طبیعت، ژن‌های موجودات در طول نسل‌های متوالی تکامل می‌یابند تا بهتر با محیط سازگار شوند. الگوریتم ژنتیک یک روش بهینه‌سازی اکتشافی است که از رویه‌های تکامل طبیعی الهام گرفته شده است. الگوریتم‌های ژنتیک روی جمعیتی از افراد عمل می‌کنند تا تقریب‌های بهتر و بهتری تولید کنند. این الگوریتم در هر نسل با انتخاب افراد، جمعیت جدیدی ایجاد می‌کند. سپس این افراد با استفاده از عملگرهایی که از ژنتیک طبیعی به عاریت گرفته شده‌اند، باهم ترکیب می‌شوند. فرزندان نیز ممکن است دچار جهش شوند. این فرآیند منجر به تکامل جمعیت‌هایی می‌شود که نسبت به افرادی که آن را به وجود آورده‌اند، مناسب‌تر با محیط خود هستند.

در یادگیری ماشین، الگوریتم ژنتیک دو کاربرد اصلی دارد. اولین مورد برای بهینه‌سازی است، مانند یافتن بهترین وزن برای یک شبکه عصبی. مورد دوم برای انتخاب ویژگی به صورت بانظارت است. در این مورد، "ژن‌ها" ویژگی‌های فردی را نشان می‌دهند و "ارگانیسم" مجموعه‌ای از ویژگی‌ها را نشان می‌دهد. هر ارگانیسم در "جمعیت" بر اساس یک امتیاز برازندگی[1] درجه‌بندی می‌شود. مناسب‌ترین موجودات زنده می‌مانند و تولید مثل می‌کنند و تکرار می‌شوند، تا چند نسلِ بعدِ جمعیت، به یک راه‌حل همگرا شود.

---

[1] fitness

**مزیت:**

- الگوریتم‌های ژنتیک می‌توانند به طور موثر ویژگی‌ها را از مجموعه داده‌های با ابعاد بسیار بالا انتخاب کنند، جایی که جستجوی جامع غیرممکن است. هنگامی که نیاز به پردازش داده‌ها برای الگوریتمی دارید که انتخاب ویژگی داخلی ندارد (مثلا کا ـ نزدیک‌ترین همسایه) و زمانی که باید ویژگی‌های اصلی را حفظ کنید (یعنی استخراج ویژگی مجاز نیست)، احتمالا الگوریتم ژنتیک بهترین گزینه برای شما خواهد بود.

**ضعف:**

- الگوریتم ژنتیک، سطح بالاتری از پیچیدگی را به پیاده‌سازی شما می‌افزاید و در بیشتر موارد ارزش این کار را ندارد. در صورت امکان، استفاده از PCA یا استفاده مستقیم از یک الگوریتم با انتخاب ویژگی داخلی، سریع‌تر و ساده‌تر است.

## روش پوشش‌دهنده[1]

در روش پوشش‌دهنده، انتخاب ویژگی‌ها با در نظر گرفتن آن به عنوان یک مساله جستجو انجام می‌شود که در آن ترکیبات مختلف ساخته، ارزیابی و با سایر ترکیب‌ها مقایسه می‌شوند. این روش، الگوریتم را با استفاده از زیرمجموعه ویژگی‌ها به صورت تکراری آموزش می‌دهد. بر اساس خروجی مدل، ویژگی‌ها اضافه یا کم می‌شوند و با این مجموعه ویژگی‌ها، مدل دوباره آموزش می‌بیند. برخی از تکنیک‌های روش پوشش‌دهنده عبارتند از:

- **انتخاب پیش‌رو:** انتخاب پیش‌رو یک فرآیند تکراری است که با مجموعه‌ای خالی از ویژگی‌ها آغاز می‌شود. پس از هر بار تکرار، یک ویژگی را اضافه می‌کند و عملکرد را ارزیابی می‌کند تا بررسی شود آیا عملکرد را بهبود می‌بخشد یا خیر. این فرآیند تا زمانی ادامه می‌یابد که افزودن یک ویژگی جدید باعث بهبود عملکرد مدل نشود.

- **حذف پس‌رو:** حذف پس‌رو نیز یک رویکرد تکراری است، اما برعکس انتخاب پیش‌رو است. این تکنیک، فرآیند را با در نظر گرفتن تمام ویژگی‌ها آغاز می‌کند و ویژگی با اهمیت کم‌تر را حذف می‌کند. این روند حذف تا زمانی ادامه می‌یابد که حذف ویژگی‌ها باعث بهبود عملکرد مدل نشود.

- **انتخاب گام‌به‌گام (تدریجی):** انتخاب گام‌به‌گام یا حذف دوطرفه، شبیه به انتخاب پیش‌رو است، اما تفاوت در این است که با افزودن یک ویژگی جدید، اهمیت ویژگی‌های قبلا اضافه شده را نیز بررسی می‌کند و اگر هر یک از ویژگی‌های قبلا انتخاب‌شده را ناچیز بیابد، به سادگی آن ویژگی خاص را از طریق حذف پس‌رو حذف می‌کند. از این رو، ترکیبی از انتخاب پیش‌رو و حذف پس‌رو است.

---

[1] Wrapper Methods

**روش فیلتر**

در روش فیلتر، ویژگی‌ها بر اساس معیارهای آماری انتخاب می‌شوند. این روش به الگوریتم یادگیری بستگی ندارد و ویژگی‌ها را به عنوان مرحله پیش‌پردازش انتخاب می‌کند. روش فیلتر ویژگی‌های نامربوط را از مدل با استفاده از معیارهای مختلف از طریق رتبه‌بندی فیلتر می‌کند. مزیت استفاده از روش‌های فیلتر این است که به زمان محاسباتی کمی نیاز دارد و منجر به بیش‌برازش داده‌ها نمی‌شود.

# کاهش خطی: تحلیل مولفه اصلی (PCA)[1]

تحلیل مولفه اصلی (PCA)، یکی از رایج‌ترین تکنیک‌های کاهش ابعاد خطی است که از یک تبدیل متعامد برای تبدیل مجموعه‌ای از مشاهدات متغیرهای احتمالا همبسته به مجموعه‌ای از مقادیر متغیرهای خطی ناهمبسته به نام مولفه‌های اصلی استفاده می‌کند. قبل از معرفی جزئیات، اجازه دهید سوال زیر را در نظر بگیریم:

برای نمونه‌هایی که در یک **فضای ویژگی متعامد**[2] قرار دارند، چگونه می‌توانیم از یک اَبَرصفحه برای نمایش نمونه‌ها استفاده کنیم؟

به طور شهودی، اگر چنین ابرصفحه‌ای وجود داشته باشد، احتمالا باید ویژگی‌های زیر را داشته باشد:

**کم‌ترین خطای بازسازی:** نمونه‌ها باید فاصله کوتاهی با این ابرصفحه داشته باشند.
**بیشترین واریانس:**[3] نگاشت (افکنش) نمونه‌ها روی ابرصفحه باید از یکدیگر دور باشد.

PCA، بازنمایی کم‌بعدی از داده‌ها را پیدا می‌کند و در عین حال تا آنجا که ممکن است تغییرات (یعنی اطلاعات برجسته) را حفظ می‌کند. PCA این کار را با پرداختن به همبستگی بین ویژگی‌ها انجام می‌دهد. اگر همبستگی بین زیرمجموعه‌ای از ویژگی‌ها بسیار زیاد باشد، PCA سعی می‌کند ویژگی‌های بسیار همبسته را ترکیب کند و این داده‌ها را با تعداد کمتری از ویژگی‌های خطی غیرهمبسته نشان دهد. الگوریتم به اجرای این کاهش همبستگی ادامه می‌دهد، جهت‌های حداکثر واریانس را در داده‌های بابعد اصلی پیدا می‌کند و آن‌ها را در فضای ابعادی کوچک‌تر نمایش می‌دهد. این مولفه‌های تازه مشتق شده به عنوان مولفه‌های اصلی شناخته می‌شوند. این تبدیل به گونه‌ای تعریف می‌شود که اولین مولفه اصلی بیشترین واریانس ممکن را داشته باشد (یعنی تا آنجا که ممکن است تغییرپذیری در داده‌ها را به خود اختصاص دهد) و هر مولفه بعدی به نوبه خود بیشترین واریانس ممکن را داشته باشد. با این مولفه‌ها، بازسازی ویژگی‌های اصلی

---

امکان‌پذیر می‌شود (نه به‌طورکامل). الگوریتم PCA به طور فعال تلاش می‌کند تا خطای بازسازی را در طول جستجوی مولفه‌های بهینه، کمینه کند.

در زیر یک طرح کلی از روش انجام تحلیل مولفه اصلی بر روی یک مجموعه داده، ارائه شده است.

**مرحله ۱. مجموعه داده**

فرض کنید مجموعه داده‌ای با $n$ ویژگی یا متغیر داریم که با $X_1, X_2, \ldots, X_n$ نشان داده شده است. اگر تعداد $N$ نمونه وجود داشته باشد، مقادیر $i$امین ویژگی $X_i$ برابر با $X_{i1}, \ldots, X_{iN}$ است (همانند جدول زیر).

| ویژگی‌ها | نمونه ۱ | نمونه ۲ | $\cdots$ | نمونه $N$ |
|---|---|---|---|---|
| $X_1$ | $X_{11}$ | $X_{12}$ | $\cdots$ | $X_{1N}$ |
| $X_2$ | $X_{21}$ | $X_{22}$ | $\cdots$ | $X_{2N}$ |
| $\vdots$ | | | | |
| $X_i$ | $X_{i1}$ | $X_{i2}$ | $\cdots$ | $X_{iN}$ |
| $\vdots$ | | | | |
| $X_n$ | $X_{n1}$ | $X_{n2}$ | $\cdots$ | $X_{nN}$ |

**مرحله ۲. محاسبه میانگین متغیرها**

میانگین $\bar{X}_i$ متغیر $X_i$ را محاسبه می‌کنیم:

$$\bar{X}_i = \frac{1}{N}(x_{i1} + x_{i2} + \cdots + x_{iN})$$

**مرحله ۳. محاسبه ماتریس کوواریانس**

متغیرهای $X_i$ و $X_j$ را در نظر بگیرید. کوواریانس جفت مرتب شده $(X_i , X_j)$ به صورت زیر تعریف می‌شود:

$$Cov(X_j , X_i) = \frac{1}{N-1} \sum_{k=1}^{N} (x_{ik} - \bar{X}_i)(x_{jk} - \bar{X}_j)$$

ما ماتریس $S_{n \times n}$ را محاسبه می‌کنیم که ماتریس کوواریانس نامیده می‌شود:

$$S = \begin{bmatrix} Cov(X_1 , X_1) & Cov(X_1 , X_2) & \cdots & Cov(X_1 , X_n) \\ Cov(X_2 , X_1) & Cov(X_2 , X_2) & \cdots & Cov(X_2 , X_n) \\ \vdots & & & \\ Cov(X_n , X_1) & Cov(X_n , X_n) & \cdots & Cov(X_n , X_n) \end{bmatrix}$$

**مرحله ۴. محاسبه مقادیر ویژه و بردارهای ویژه ماتریس کوواریانس**

فرض کنید $S$ ماتریس کوواریانس باشد و همچنین $I$ ماتریس همانی باشد که ابعادی مشابه با بعد $S$ دارد.

أ. معادله زیر را تشکیل دهید:

$$\det(S - \lambda I) = \cdot$$

این یک معادله چند جمله‌ای درجه $n$ در $\lambda$ است. $n$ ریشه حقیقی دارد (برخی از ریشه‌ها ممکن است تکرار شوند) و این ریشه‌ها مقادیر ویژه $S$ هستند. ما $n$ ریشه $\lambda_1, ..., \lambda_n$ را از معادله فوق پیدا می‌کنیم.

ب. اگر $\acute{\lambda} = \lambda$ یک مقدار ویژه باشد، بردار ویژه مربوط به یک بردار به شکل زیر است:

$$U = \begin{bmatrix} u_1 \\ u_2 \\ \vdots \\ u_\Upsilon \end{bmatrix}$$

به‌طوری‌که:

$$(S - \acute{\lambda}I)U = \cdot$$

سپس مجموعه‌ای از $n$ بردار ویژه متعامد $U_1, ..., U_n$ را پیدا می‌کنیم، به طوری که $U_i$ یک بردار ویژه مربوط به $\lambda_i$ باشد.

ج. اکنون بردارهای ویژه را نرمال می‌کنیم. با توجه به هر بردار $X$، آن را با تقسیم $X$ بر طول آن نرمال می‌کنیم. طول (یا نرم) بردار

$$X = \begin{bmatrix} x_1 \\ x_2 \\ \vdots \\ x_\Upsilon \end{bmatrix}$$

به‌صورت

$$\|X\| = \sqrt{x_1^2 + x_2^2 + \cdots + x_n^2}$$

تعریف می‌شود.

با توجه به هر بردار ویژه $U$، بردار ویژه نرمال‌شده مربوط به صورت

$$\frac{1}{\|U\|} U$$

محاسبه می‌شود.

ما $n$ بردار ویژه نرمال‌شده $e_1, ..., e_n$ را توسط

$$e_i = \frac{1}{\|U_i\|} U_i, i = 1, 2, ..., n.$$

محاسبه می‌کنیم.

**مرحله ۵. استخراج مجموعه داده جدید**

مقادیر ویژه را از بیشترین به کم‌ترین مرتب کنید. بزرگ‌ترین مقدار ویژه واحد، اولین مولفه اصلی است.

أ.  بگذارید مقادیر ویژه به ترتیب نزولی $\lambda_1, \ldots, \lambda_n$ باشد و بردارهای ویژه واحد متناظر $e_1, \ldots, e_n$ باشند.

ب.  یک عدد صحیح $p$ را طوری انتخاب کنید که $n \leq p \leq 1$ باشد.

ج.  بردارهای ویژه مربوط به مقادیر ویژه $\lambda_1, \ldots, \lambda_p$ را انتخاب کنید و ماتریس $n \times p$ زیر را تشکیل دهید:

$$F = \begin{bmatrix} e_1^T \\ e_2^T \\ \vdots \\ e_p^T \end{bmatrix}$$

د.  ماتریس $N \times n$ زیر را تشکیل می‌دهیم:

$$X = \begin{bmatrix} X_{11} - \bar{X}_1 & X_{12} - \bar{X}_1 & \cdots & X_{1N} - \bar{X}_1 \\ X_{21} - \bar{X}_2 & X_{22} - \bar{X}_2 & \cdots & X_{2N} - \bar{X}_1 \\ \vdots & & & \\ X_{31} - \bar{X}_3 & X_{n2} - \bar{X}_n & \cdots & X_{nN} - \bar{X}_n \end{bmatrix}$$

ه.  سپس ماتریس زیر را محاسبه می‌کنیم:

$$X_{\text{جدید}} = FX$$

توجه داشته باشید که این یک ماتریس $N \times p$ است که به ما مجموعه داده‌ای از $N$ نمونه با $p$ ویژگی را می‌دهد.

**مرحله ۶. مجموعه داده جدید**

ماتریس $X_{\text{جدید}}$ مجموعه داده جدید است. هر ردیف از این ماتریس مقادیر یک ویژگی را نشان می‌دهد.

**مرحله ۷. نتیجه**

به این صورت تحلیل مولفه اصلی به ما در کاهش ابعاد مجموعه داده کمک می‌کند. **توجه داشته باشید که امکان بازگرداندن مجموعه داده $n$ بعدی اصلی از مجموعه داده جدید وجود ندارد.**

ما ایدهِ تحلیل مولفه اصلی را با در نظر گرفتن یک مثال نشان می‌دهیم. در این مثال، تمام جزئیاتِ محاسبات آورده شده است. این کار برای این است که به خواننده ایده‌ای از پیچیدگی محاسبات بدهد و همچنین به خواننده کمک کند تا یک را با محاسبات دستی بدون توسل به بسته‌های نرم‌افزاری انجام دهد.

**مثال.** با توجه به داده‌های جدول زیر، از PCA برای کاهش از بعد ۲ به ۱ استفاده کنید.

| ویژگی‌ها | نمونه ۱ | نمونه ۲ | نمونه ۳ | نمونه ٤ |
|---|---|---|---|---|
| $X_1$ | ٤ | ٨ | ۱۳ | ۷ |
| $X_2$ | ۱۱ | ٤ | ٥ | ۱٤ |

## ۱. نمودار نقطه‌ای (پراکندگی) داده‌ها

داریم:

$$\bar{X}_1 = \frac{1}{4}(4 + 8 + 13 + 7) = 8$$

$$\bar{X}_1 = \frac{1}{4}(11 + 4 + 5 + 14) = 8.5$$

شکل زیر نمودار نقطه‌ای داده‌ها را همراه با نقطه $(\bar{X}_1, \bar{X}_2)$ نشان می‌دهد:

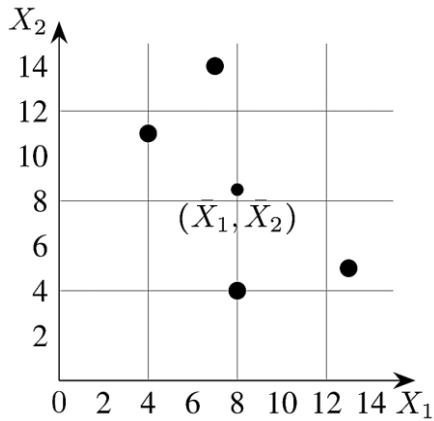

**شکل ۸ ـ ۱۲.** پراکندگی داده‌ها

## ۲. محاسبه ماتریس کوواریانس

کوواریانس‌ها به صورت زیر محاسبه می‌شوند:

$$Cov(X_1, X_1) = \frac{1}{N-1}\sum_{k=1}^{N}(x_{1k} - \bar{X}_1)^2$$

$$= \frac{1}{3}\left((4-8)^2 + (8-8)^2 + (13-8)^2 + (7-8)^2\right)$$

$$= 14$$

$$Cov(X_1, X_2) = \frac{1}{N-1}\sum_{k=1}^{N}(x_{1k} - \bar{X}_1)(x_{2k} - \bar{X}_2)$$

$$= \frac{1}{3}\left((4-8)(11-8.5) + (8-8)(4-8.5) + (13-8)(5-8.5) + (7-8)(14-8.5)\right)$$

$$= -11$$

$$Cov(X_2, X_1) = -11$$

$$Cov(X_2, X_2) = \frac{1}{N-1}\sum_{k=1}^{N}(x_{2k} - \bar{X}_2)^2$$

$$= \frac{1}{3}\left((11-8.5)^2 + (4-8.5)^2 + (5-8.5)^2 + (14-8.5)^2\right)$$

$$= 23$$

بر این اساس، ماتریس کوواریانس برابر است با:

$$S = \begin{bmatrix} Cov(X_1, X_1) & Cov(X_1, X_2) \\ Cov(X_2, X_1) & Cov(X_2, X_2) \end{bmatrix}$$

$$= \begin{bmatrix} 14 & -11 \\ -11 & 23 \end{bmatrix}$$

### ۳. مقادیر ویژه ماتریس کوواریانس

معادله مشخصه ماتریس کوواریانس برابر است با:

$$0 = \det(S - \lambda I)$$

$$= \begin{vmatrix} 14 - \lambda & -11 \\ -11 & 23 - \lambda \end{vmatrix}$$

$$= (14 - \lambda)(23 - \lambda) - (-11) \times (-11)$$

$$= \lambda^2 - 37\lambda + 201$$

از حل معادله مشخصه داریم:

$$\lambda = \frac{1}{2}(37 \pm \sqrt{565})$$

$$= 30.3849, 6.6151$$

$$= \lambda_1, \lambda_2$$

### ۴. محاسبه بردارهای ویژه

برای یافتن اولین مولفه اصلی، فقط باید بردار ویژه مربوط به بزرگترین مقدار ویژه را محاسبه کنیم. در مثال حاضر، بزرگترین مقدار ویژه $\lambda_1$ است. بنابراین ما بردار ویژه مربوط به $\lambda_1$ را محاسبه می‌کنیم.

بردار ویژه مربوط به $\lambda = \lambda_1$ بردار $U = \begin{bmatrix} u_1 \\ u_2 \end{bmatrix}$ است که معادله زیر را برآورده می‌کند:

$$\begin{bmatrix} 0 \\ 0 \end{bmatrix} = (S - \lambda_1 I) X$$

$$= \begin{bmatrix} 14 - \lambda_1 & -11 \\ -11 & 23 - \lambda_1 \end{bmatrix} \begin{bmatrix} u_1 \\ u_2 \end{bmatrix}$$

$$= \begin{bmatrix} (14 - \lambda_1)u_1 - 11u_2 \\ -11u_1 + (23 - \lambda_1)u_2 \end{bmatrix}$$

که معادل با دو معادله زیر است:

$$(14 - \lambda_1)u_1 - 11u_2 = 0$$

$$-11u_1 + (23 - \lambda_1)u_2 = 0$$

با استفاده از نظریه دستگاه معادلات خطی، متوجه می‌شویم که این معادلات مستقل نیستند و جواب‌ها توسط

$$\frac{u_1}{11} = \frac{u_2}{14 - \lambda_1} = t$$

ارائه می‌شوند، به این معنا که

$$u_1 = 11t, u_2 = (14 - \lambda_1)t$$

جایی که $t$ هر عدد حقیقی می‌باشد. با گرفتن $t = 1$، بردار ویژه مربوط به $\lambda_1$ را به صورت

$$U_1 = \begin{bmatrix} 11 \\ 14 - \lambda_1 \end{bmatrix}$$

دریافت می‌کنیم. برای یافتن بردار ویژه واحد، طول $X_1$ را محاسبه می‌کنیم که با

$$\|U_1\| = \sqrt{11^2 + (14 - \lambda_1)^2}$$

$$= \sqrt{11^2 + (14 - 30.3849)^2}$$

$$= 19.7348$$

بدست می‌آید. بنابراین، بردار ویژه واحد مربوط به $\lambda_1$ برابر است با

$$e_1 = \begin{bmatrix} 11 \big/ \|U_1\| \\ 14 - \lambda_1 \big/ \|U_1\| \end{bmatrix}$$

$$= \begin{bmatrix} 11 \big/ 19.7348 \\ 14 - 30.3849 \big/ 19.7348 \end{bmatrix}$$

$$= \begin{bmatrix} 0.5574 \\ -0.8303 \end{bmatrix}$$

با انجام محاسبات مشابه، بردار ویژه $e_2$ مربوط به مقدار ویژه $\lambda = \lambda_1$ را می‌توان بدست آورد:

$$e_2 = \begin{bmatrix} 0.8303 \\ 0.5574 \end{bmatrix}.$$

## ۵. محاسبه اولین مولفه اصلی

فرض کنید k-امین نمونه در جدول داده‌ها باشد. اولین مولفه اصلی این مثال برابر $\begin{bmatrix} X_{1k} \\ X_{2k} \end{bmatrix}$ است با:

$$e_1^T \begin{bmatrix} X_{1k} - \overline{X}_1 \\ X_{2k} - \overline{X}_2 \end{bmatrix} = [0.5574 \quad -0.8303] \begin{bmatrix} X_{1k} - \overline{X}_1 \\ X_{2k} - \overline{X}_2 \end{bmatrix}$$

$$= 0.5574(X_{1k} - \overline{X}_1) - 0.8303(X_{2k} - \overline{X}_2).$$

به عنوان مثال، اولین مولفه اصلی مربوط به مثال اول $\begin{bmatrix} X_{11} \\ X_{21} \end{bmatrix} = \begin{bmatrix} 4 \\ 11 \end{bmatrix}$ به صورت زیر محاسبه می‌شود:

$$[\cdot.۰۰۷۴ \quad -\cdot.۸۳۰۳]\begin{bmatrix} X_{۱۱} - \bar{X}_۱ \\ X_{۲۱} - \bar{X}_۲ \end{bmatrix}$$

$$= \cdot.۰۰۷۴(X_{۱۱} - \bar{X}_۱) - \cdot.۸۳۰۳(X_{۲۱} - \bar{X}_۲)$$

$$= \cdot.۰۰۷۴(۴ - ۸) - \cdot.۸۳۰۳(۱۱ - ۸.۵)$$

$$= -۴.۳۰۵۳۵$$

نتایج محاسبات در جدول زیر خلاصه شده است:

| $X_1$ | ۴ | ۸ | ۱۳ | ۷ |
|---|---|---|---|---|
| $X_۲$ | ۱۱ | ۴ | ۵ | ۱۴ |
| اولین مولفه اصلی | $-۴.۳۰۵۲$ | ۳.۷۳۶۱ | ۵.۶۹۲۸ | $-5.1238$ |

## ٦. معنای هندسی اولین مولفه اصلی

برای شکل ۸ـ۱۲، محورهای مختصات جدیدی را معرفی می‌کنیم. ابتدا مبدا را به "مرکز" $(\bar{X}_۱, \bar{X}_۲)$ تغییر داده و سپس جهت محورهای مختصات را به جهت بردارهای ویژه $e_۱$ و $e_۲$ تغییر می‌دهیم (شکل ۸ـ ۱۳ را ببینید).

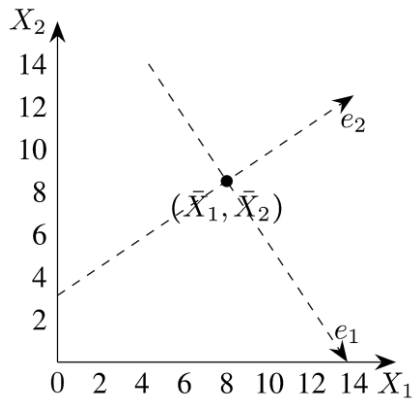

**شکل ۸ـ۱۳.** دستگاه مختصات برای مولفه‌های اصلی

سپس، عمودها را از نقاط داده شده به محور $e_۱$ می‌کشانیم (شکل ۸ـ۱۴ را ببینید). اولین مولفه اصلی مختصات $e_۱$ پاهای عمود هستند. نگاشت نقاط داده در محور $e_۱$ ممکن است به عنوان تقریبی از نقاط داده ارائه شده در نظر گرفته شود، در نتیجه، می‌توان مجموعه داده ارائه شده را با این نقاط جایگزین کنیم.

اکنون، هر یک از این تقریب‌ها را می‌توان به طور واضح با یک عدد مشخص کرد، یعنی مختصات تقریبی $e_۱$. بنابراین مجموعه داده‌های دو بعدی ارائه شده را می توان تقریبا با مجموعه داده‌های یک بعدی زیر نشان داد (شکل ۷ـ۱۵ را ببینید):

| اولین مولفه اصلی | $-4.3052$ | $3.7361$ | $5.6928$ | $-5.1238$ |
|---|---|---|---|---|

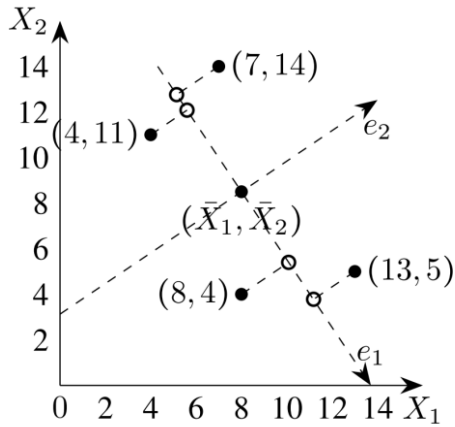

**شکل ۸ ــ ١٤.** نگاشت داده‌ها در محور اولین مؤلفه اصلی

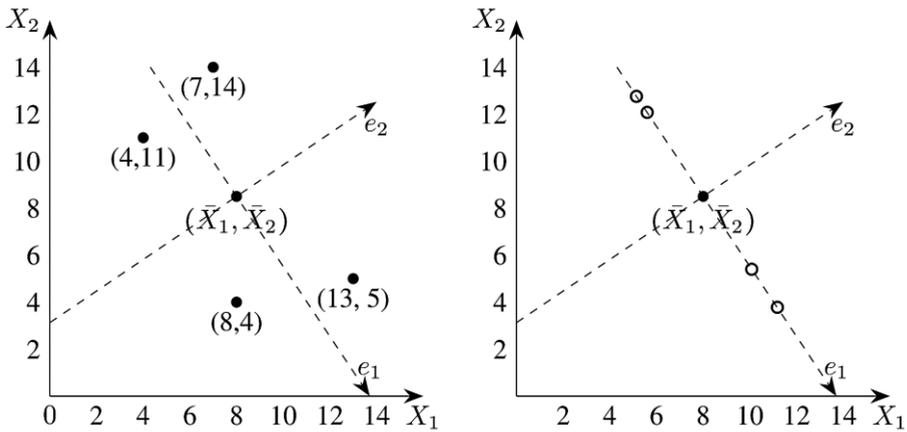

**شکل ۸ ــ ١٤.** نمایش هندسی تقریبی یک بعدی داده‌ها

همچنین می‌توانید کد زیر را در پایتون اجرا کنید:

```
#data
import numpy as np
import matplotlib.pyplot as plt
from sklearn.decomposition import PCA
X = np.array([[4,11],[8,4],[13,5],[7,14]])
X1=np.mean(X[:, 0])
X2=np.mean(X[:, 1])
pca = PCA(n_components=1)
pca.fit(X)
X_pca = pca.transform(X)
```

```
X_new = pca.inverse_transform(X_pca)
plt.scatter(X[:, 0], X[:, 1], alpha=0.2)
plt.scatter(X1, X2, alpha=0.2)
plt.scatter(X_new[:, 0], X_new[:, 1], alpha=0.8)
plt.axis('equal');
print(pca.components_)
```

Out [1]:    [[ 0.55738997 -0.83025082]]

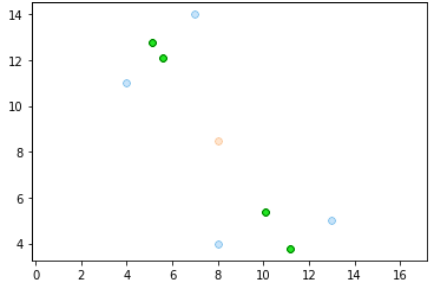

## مزایا و معایب PCA

**مزایا**

- **آسان برای محاسبه:** PCA مبتنی‌بر جبر خطی است که از نظر محاسباتی به راحتی توسط رایانه قابل حل است.
- **سرعت سایر الگوریتم‌های یادگیری ماشین را افزایش می‌دهد:** الگوریتم‌های یادگیری ماشین زمانی که بر روی مولفه اصلی به جای مجموعه داده اصلی آموزش داده می‌شوند، سریع‌تر همگرا می‌شوند.

**معایب**

- **متعارف‌سازی داده:** داده‌ها باید قبل از اجرای PCA متعارف شوند، در غیر این صورت شناسایی اجزای اصلی بهینه دشوار می‌شود.
- **PCA یک رابطه خطی بین ویژگی‌ها را فرض می‌کند:** الگوریتم برای روابط غیرخطی مناسب نیست.

# یادگیری منیفلد (کاهش غیرخطی)

در بخش پیشین دیدیم که چگونه می‌توان از تحلیل مولفه اصلی در کار کاهش ابعاد استفاده کرد. در حالی که PCA منعطف و سریع است، در صورت وجود روابط غیرخطی در داده‌ها، آن‌قدر خوب عمل نمی‌کند. برای رفع این محدودیت، می‌توانیم به دسته‌ای از روش‌ها به نام یادگیری

منیفلد مراجعه کنیم؛ دسته‌ای از برآوردگرهای غیرنظارتی که به دنبال توصیف مجموعه داده‌ها به عنوان منیفلدهای کم‌بعدی تعبیه‌شده در فضاهای با ابعاد بالا هستند.

وقتی به یک منیفلد فکر می‌کنید، پیشنهاد می‌کنم یک ورق کاغذ را تصور کنید. این یک شی دو ـ بعدی است که در دنیای سه ـ بعدی ما وجود دارد و می‌تواند در این دو بعد خم شود یا لوله شود. در اصطلاح یادگیری منیفلد، می‌توان این ورق را یک منیفلد دوبعدی تعبیه‌شده در فضای سه بعدی در نظر گرفت. چرخش، جهت‌دهی مجدد، یا کشش تکه کاغذ در فضای سه ـ بعدی، هندسه صاف کاغذ را تغییر نمی‌دهد. اگر کاغذ را خم کنید، بپیچید یا مچاله کنید، همچنان یک منیفلد دو ـ بعدی است، اما جاسازی در فضای سه ـ بعدی دیگر خطی نیست. الگوریتم‌های یادگیری منیفلد به دنبال یادگیری ماهیت دو بعدی این کاغذ، یا به عبارت دیگر، تعیین ساختار این منیفلد هستند.

اساسا، فرضیه منیفلد بیان می‌کند که داده‌های با ابعاد بالا در دنیای واقعی بر روی منیفلدهای کم‌بعدی قرار گرفته‌اند که در فضای با ابعاد بالا تعبیه شده‌اند. به عبارت ساده‌تر، به این معنی است که داده‌های با ابعاد بالاتر بیشتر اوقات بر روی یک منیفلد بسیار نزدیک‌تر با ابعاد پایین‌تر قرار دارند. فرآیند مدل‌سازیِ منیفلدی که نمونه‌های آموزشی روی آن قرار دارند، یادگیری منیفلد نامیده می‌شود.

نظریه هندسه دیفرانسیل نشان می‌دهد که منیفلد را می‌توان به عنوان یک فضای ریاضی انتزاعی در نظر گرفت که هندسه ذاتی آن را می‌توان به طور کامل توسط متریکِ (سنجه‌ای) محلی و اطلاعات همسایگیِ بسیار کوچک که شبیه فضای اقلیدسی است، تعیین کرد. بنابراین ممکن است فکر کنیم که ساختار همسایگی محلی یک منیفلد می‌تواند برای جاسازی نقاط داده در فضای با ابعاد بالا در فضای ویژگی‌هایِ بعدـ پایین استفاده شود. از این‌رو، اگر یک منیفلد بعدـ پایین در فضای با ابعاد بالا تعبیه شده باشد، نمونه‌هایی که در فضای با ابعاد بالا بسیار پیچیده به نظر می‌رسند، به طور محلی همان ویژگی‌های نمونه‌های موجود در فضای اقلیدسی را دارند. بنابراین، می‌توانیم نگاشت کاهش ابعاد را به صورت محلی ایجاد کنیم و سپس به کل فضا گسترش دهیم. هنگامی که ابعاد به دو یا سه کاهش می‌یابد، به طور طبیعی می‌توانیم داده‌ها را تجسم کنیم و از این رو یادگیری منیفلد برای اهداف مصورسازی نیز مفید است. در ادامه این بخش به بررسی چند الگوریتم محبوب یادگیری منیفلد می‌پردازیم.

**ایزومپ[1]**

ایزومپ، یک الگوریتم یادگیری منیفلد است که مدل داخلی داده‌ها را یاد می‌گیرد. نزدیک‌ترین همسایگان را بهم متصل می‌کند و تشکیل یک گراف می‌دهد. سپس کوتاه‌ترین مسیر را بین تمام گره‌های گراف محاسبه می‌کند. این فاصله ژئودزیکی (فاصله دو نقطه نسبت به سطح) از نقاط

---

[1] Isomap

را برآورد می‌کند. در نهایت، مقیاس‌گذاری چند‌ـ‌بعدی را روی ماتریس فواصل گراف اعمال می‌کند که باعث می‌شود داده‌های اصلی با ابعاد پایین جاسازی شوند.

دو رویکرد کلی برای ساختن گراف همسایگی وجود دارد. اولین رویکرد این است که تعداد همسایگان را مشخص کنید. به عنوان مثال، با استفاده از کا‌ـ‌نزدیکترین همسایه با معیار فاصله اقلیدسی. روش دیگر تعیین آستانه فاصله $\epsilon$ برای در نظر گرفتن همه نقاط با فاصله کوچکتر از $\epsilon$ به عنوان همسایه است. با این حال، هر دو رویکرد محدودیت یکسانی دارند: اگر محدوده همسایگی مشخص شده $k$ یا $\epsilon$، بزرگ باشد، ممکن است "اتصال کوتاه" اتفاق بیفتد، که در آن برخی از نقاط دوردست به اشتباه نزدیک به یکدیگر در نظر گرفته شوند. از سوی دیگر، اگر محدوده همسایگی مشخص شده خیلی کوچک باشد، ممکن است "مدار باز" اتفاق بیفتد که در آن برخی از مناطق از یکدیگر جدا شوند.

یک الگوریتم کارآمد برای محاسبه کوتاه‌ترین مسیر بین هر جفت رئوس در یک گراف، الگوریتم فلوید است که با گراف‌های چگال (گراف‌هایی با یال‌های زیاد) بهتر کار می‌کند. با این حال، الگوریتم دایکسترا زمانی ترجیح داده می‌شود که گراف خلوت باشد. الگوریتم فلوید دارای پیچیدگی $O(n^{3})$ در بدترین حالت است، در حالی که الگوریتم دایکسترا با هرم‌های فیبوناچی دارای پیچیدگی $O(Kn^{2}logn)$ است که در آن $K$، اندازه همسایگی است.

ایزومپ فقط مختصات بعد‌ـ‌پایین نمونه‌های آموزشی را ارائه می‌دهد، اما چگونه می‌توانیم نمونه‌های جدید را به فضای بعد‌ـ‌پایین برسانیم؟ یک رویکرد کلی این است که یک مدل رگرسیونی را با استفاده از مختصاتِ ابعادیِ بالایِ نمونه‌هایِ آموزشی به عنوان ورودی و مختصات بعد‌ـ‌پایین مربوط را به عنوان خروجیِ آموزش دهیم. سپس از مدل رگرسیون آموزش دیده برای پیش‌بینی مختصات بعد‌ـ‌پایین نمونه‌های جدید استفاده کنیم. چنین روشی فاقد عمومیت به نظر می‌رسد، اما در حال حاضر به نظر می‌رسد راه‌حل بهتری وجود ندارد!!

## جاسازی خطی محلی[۱] (LLE)

برخلاف ایزومپ که فاصله‌یِ بینِ نمونه را حفظ می‌کند، جاسازیِ خطی محلی (LLE) با هدف حفظ روابطِ خطیِ بینِ نمونه‌هایِ همسایه است. جمله‌ی معروفی که در خصوص الگوریتم LLE به کرات نقل می‌شود این است که "به‌صورت جهانی فکر کنید، به‌صورت محلی برازش کنید"[۲]: به عبارت دیگر، الگوریتم به تکه‌های کوچک و محلی در اطراف هر نمونه نگاه می‌کند و از این قطعه‌ها برای ساخت منیفلد وسیع‌تر استفاده می‌کند. همان‌طور که در شکل ۱۵‌ـ‌۸ نشان داده

---

شده است، فرض کنید مختصات یک نقطه نمونه $x_i$ را می‌توان از طریق یک ترکیب خطی از مختصات نمونه‌های همسایه $x_k, x_j$ و $x_l$ بازسازی کرد، یعنی:

$$x_i = w_{ij}x_j + w_{ik}x_k + w_{il}x_l$$

هدف LLE حفظ رابطه بالا در فضای بعد‌ـ‌پایین است.

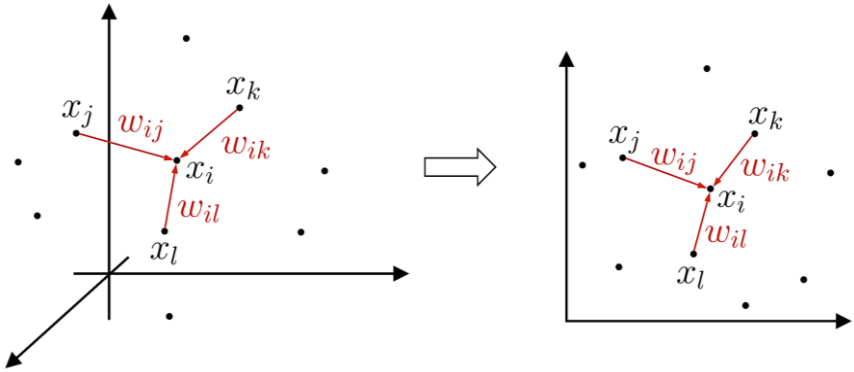

**شکل ۸ـ۱۵.** رابطه بازسازی نمونه‌ها در فضای با ابعاد بالا در فضای بعد‌ـ‌پایین حفظ می‌شود

LLE با شناسایی شاخص‌های همسایگی $Q_i$ برای نمونه $x_i$ شروع می‌کند و سپس وزن‌های بازسازی (نوسازی) خطی $w_i$ نمونه‌ها را در $Q_i$ پیدا می‌کند:

$$min_{w_1,\ldots,w_m} \sum_{i=1}^{m} \left\| x_i - \sum_{j \in Q_i} w_{ij}x_j \right\|_{\Upsilon}^{\Upsilon}$$

$$s.t \sum_{j \in Q_i} w_{ij}x_j = 1$$

که در آن $x_i$ و $x_j$ شناخته‌شده هستند. اگر $C_{jk} = (x_i - x_j)^T(x_i - x_k)$ باشد، آن‌گاه $w_{ij}$ یک راه‌حل فرم بسته[۱] دارد:

$$w_{ij} = \frac{\sum_{j \in Q_i} C_{jk}^{-1}}{\sum_{l,s \in Q_i} C_{ls}^{-1}}$$

از آن‌جایی که LLE، $w_i$ را در فضای بعد‌ـ‌پایین حفظ می‌کند، مختصات بعد‌ـ‌پایین $z_i$ از $x_i$ را می‌توان به صورت زیر بدست آورد:

$$min_{z_1,\ldots,z_m} \sum_{i=1}^{m} \left\| z_i - \sum_{j \in Q_i} w_{ij}z_j \right\|^{\Upsilon}$$

$z_i$ مختصات بعد پایین $x_i$ را بهینه می‌کند.

---

اگر $\mathbf{Z} = (z_1, \ldots, z_m) \epsilon \mathbb{R}^{\hat{d} \times m}$ و $(\mathbf{W})_{ij} = w_{ij}$، $\mathbf{M} = (\mathbf{I} - \mathbf{W})^T(\mathbf{I} - \mathbf{W})$، آن‌گاه معادله پیشین را می‌توان به صورت

$$min_Z \ \ tr(\mathbf{Z}\mathbf{M}\mathbf{Z}^T)$$
$$s.t \ \mathbf{Z}\mathbf{Z}^T = 1$$

بازنویسی کرد. می‌توانیم معادله فوق را با تجزیه مقادیر ویژه حل کنیم: $\mathbf{Z}\mathbf{T}$ ماتریس متشکل از بردارهای ویژه $\hat{d}$ با کوچک‌ترین مقادیر ویژه $\mathbf{M}$ است.

## تی-جاسازی همسایگی تصادفی توزیع‌شده (t-SNE) [1]

تی‌ـ‌جاسازی همسایگی تصادفی توزیع‌شده (t-SNE)، یکی از محبوب‌ترین تکنیک‌های کاهش ابعاد غیرخطی برای مصورسازی داده‌های با ابعاد بالا است. t-SNE این کار را با مدل‌سازی هر نقطه با ابعاد بالا در یک فضای دو یا سه‌ـ‌بعدی انجام می‌دهد، جایی که نقاط مشابه نزدیک به یکدیگر و نقاط غیرمشابه، دورتر مدل‌سازی می‌شوند. برای انجام این کار، t-SNE دو توزیع احتمال می‌سازد، یکی بر روی جفت نقاط در فضای با ابعاد بالا و دیگری بر روی جفت نقاط در فضای بعد پایین، به‌طوری‌که نقاط مشابه احتمال زیاد و نقاط غیرمشابه احتمال کم‌تری دارند. به بیان دقیق‌تر، t-SNE **واگرایی کولبک‌ـ‌لیبلر** [2]، بین دو توزیع احتمال را کمینه می‌کند.

اولین مرحله در الگوریتم t-SNE محاسبه فاصله بین هر نمونه با نمونه دیگر در مجموعه داده است. به‌طور پیش‌فرض، برای معیار فاصله، از فاصله اقلیدسی استفاده می‌شود، که فاصله خط مستقیم بین هر دو نقطه در فضای ویژگی است. سپس این فاصله‌ها به احتمالات تبدیل می‌شوند. می‌توانید در شکل ۸ـ۱۶ آن را مشاهده کنید.

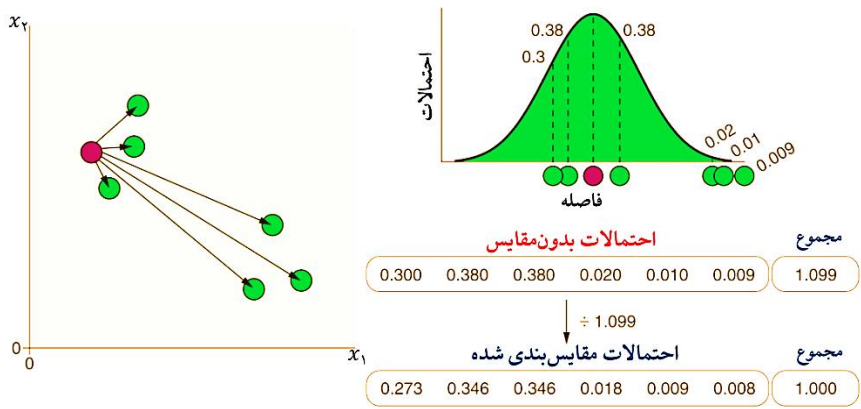

**شکل ۸ـ۱۶.** محاسبه فاصله‌ها و تبدیل آن‌ها به احتمالات.

---

[1] t-distributed stochastic neighbor embedding

[2] Kullback-Leibler

برای یک نمونه خاص در مجموعه داده، فاصله بین این نمونه و سایر نمونه‌ها اندازه‌گیری می‌شود. سپس یک توزیع نرمال بر روی این حالت متمرکز می‌شود و فواصل با نگاشت آن‌ها بر روی چگالی احتمال توزیع نرمال، به احتمالات تبدیل می‌شوند. انحراف معیار این توزیع نرمال، با چگالی نمونه‌ها در اطراف نمونه مورد نظر، رابطه معکوس دارد. به عبارت دیگر، اگر نمونه‌های زیادی در این نزدیکی وجود داشته باشد (چگال‌تر)، انحراف معیار توزیع نرمال کوچک‌تر است. اما اگر موارد کمی در این نزدیکی وجود داشته باشد (چگالی کم‌تر)، آنگاه انحراف معیار بزرگ‌تر است.

پس از تبدیل فواصل به احتمالات، احتمالات برای هر نمونه با تقسیم آن‌ها بر مجموع آن‌ها مقیاس‌بندی می‌شوند. این باعث می‌شود که مجموع احتمالات برای هر نمونه در مجموعه داده به ۱ برسد. استفاده از انحراف معیار مختلف برای چگالی‌های مختلف و سپس نرمال کردن احتمالات به ۱ برای هر نمونه، به این معناست که اگر خوشه‌های چگال و خوشه‌های خلوتی از نمونه‌ها در مجموعه داده وجود داشته باشد، t-SNE خوشه‌های چگال را گسترش داده و خوشه‌های پراکنده را فشرده می‌کند تا بتوانند باهم، راحت‌تر مصورسازی شوند.

هنگامی‌که احتمالات مقیاس‌بندی شده برای هر نمونه در مجموعه داده محاسبه شد، ماتریسی از احتمالات داریم که تشریح می‌کند که هر نمونه چقدر شبیه به هر یک از نمونه‌های دیگر است. این مصورسازی در شکل ۸ ـ ۱۷ به عنوان یک نقشه‌حرارتی[1] نشان داده شده است.

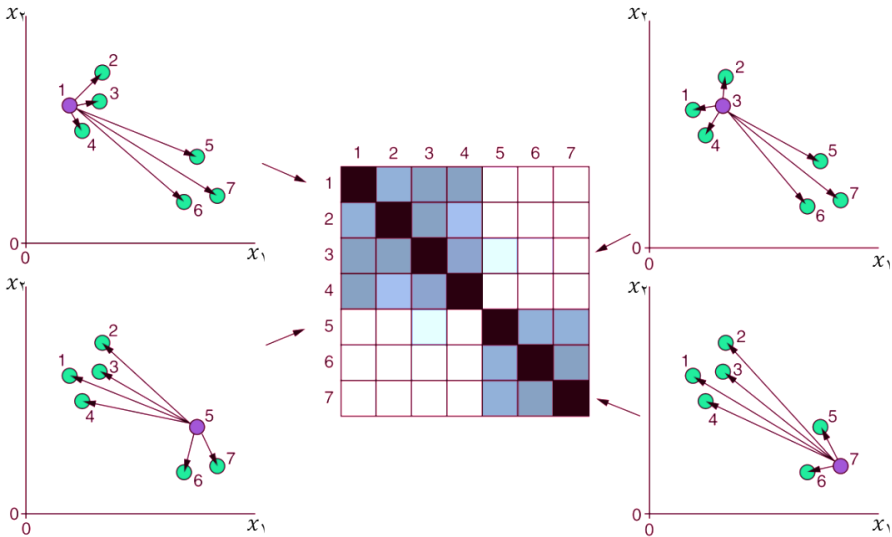

**شکل ۸ ـ ۱۷.** احتمالات مقیاس‌بندی شده برای هر نمونه به عنوان ماتریسی از مقادیر ذخیره می‌شوند. این در اینجا به‌عنوان یک نقشه حرارتی مصورسازی شده‌اند: هر چه دو نمونه نزدیک‌تر باشند، کادر تیره‌تر است که نشان دهنده فاصله آن‌ها در نقشه حرارتی است.

---

[1] heatmap

ماتریس احتمالات ما اکنون برای الگوی مرجع یا الگوی ما برای چگونگی ارتباط مقادیر داده با یکدیگر در فضای اصلی و با ابعاد بالا است. گام بعدی در الگوریتم t-SNE، تصادفی‌کردن موارد در امتداد دو محور (لازم نیست دو محور باشد، اما معمولا اینطور است) جدید است (این همان جایی است که t-SNE نام تصادفی می‌گیرد).

t-SNE فواصل بین نمونه‌ها را در این فضای تصادفیِ و بعد پایینِ جدید، محاسبه می‌کند و آن‌ها را مانند قبل به احتمالات تبدیل می‌کند. تنها تفاوت این است که به جای استفاده از توزیع نرمال، اکنون از توزیع تی‌ـ‌استیودنت استفاده می‌کند. توزیع تی تاحدودی شبیه یک توزیع نرمال است، با این تفاوت که وسط آن چندان بلند نیست و دو طرف آن هموارتر است و بیشتر به سمت بیرون کشیده می‌شود (شکل ۸ـ۱۸).

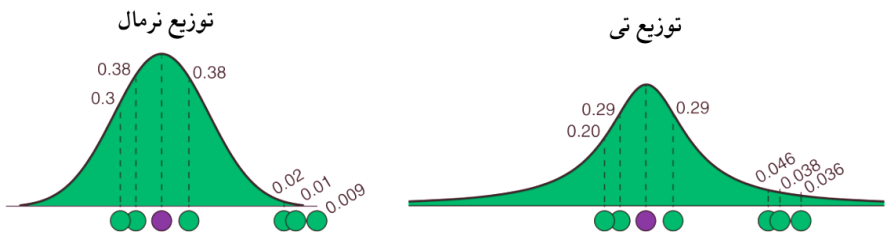

**شکل ۸ـ۱۸.** توزیع نرمال و توزیع تی

اکنون کار t-SNE این است که نقاط داده را در اطراف این محورهای جدید، گام به گام "درهم بزند" تا ماتریس احتمالات در فضای ابعاد پایین‌تر را تا حد امکان به ماتریس احتمالات اصلی در فضای با ابعاد بالا نزدیک کند. شهودی که در اینجا وجود دارد این است که اگر ماتریس‌ها تا حد ممکن مشابه باشند، آنگاه داده‌ها در هر دو فضای، نزدیک به یکدیگر هستند. برای اینکه ماتریس احتمال در فضای بعد پایین، مشابه ماتریس فضای با ابعاد بالا باشد، هر نمونه باید به نمونه‌های نزدیک‌تر شود که در داده‌های اصلی به آن نزدیک بوده است و از نمونه‌هایی که دور بودند، فاصله بگیرد. در نتیجه، نمونه‌هایی که باید در همان نزدیکی باشند، همسایه خود را به سمت خود می‌کشاند، اما مواردی که باید دور باشند را از خود دور می‌کنند. تعادل این نیروهای جذابه و دافعه باعث می‌شود که هر مورد در مجموعه داده به سمتی حرکت کند که دو ماتریس احتمال را کمی شبیه‌تر کند. اکنون، در این موقعیت جدید، ماتریس احتمال بعد پایین دوباره محاسبه می‌شود و نمونه‌ها دوباره حرکت می‌کنند و باعث می‌شود که ماتریس‌های بعد پایین و بالا، دوباره کمی شبیه‌تر به نظر برسند. این روند تا زمانی ادامه می‌یابد که به تعداد تکرارهای از پیش تعیین‌شده برسیم، یا تا زمانی که واگرایی (تفاوت) بین ماتریس‌ها بهبود نیابد. هنگامی که این فرآیند تکراری با واگرایی کولبک‌ـ‌لیبلر کم، همگرا شد، باید بازنمایی با ابعاد پایین از

---

[1] shuffle

داده‌های اصلی خود را داشته باشیم که شباهت های بین نمونه‌های نزدیک را حفظ کند. کل این فرآیند در شکل ۸ــ۱۹ نشان داده شده است.

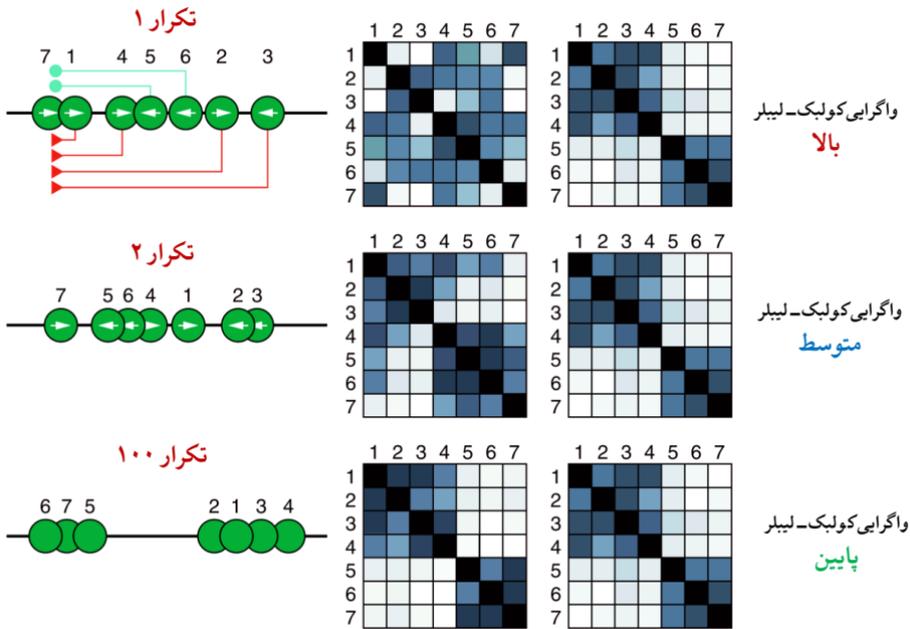

**شکل ۸ــ۱۹. فرآیند الگوریتم t-SNE.** نمونه‌ها به طور تصادفی روی محورهای جدید مقداردهی اولیه می‌شوند (در این مثال یک محور نشان داده شده است). ماتریس احتمال برای این محور محاسبه می‌شود و موارد را درهم می‌ریزند تا این ماتریس شبیه به ماتریس اصلی با ابعاد بالا با کمینه کردن واگرایی کولبک لیبلر شود. در حین جابجایی، نمونه‌ها به سمت نمونه‌هایی که شبیه به آن‌ها هستند (خطوط با دایره) جذب می‌شوند و از موارد غیرمشابه (خطوط با مثلث) دور می‌شوند.

شاید برای شما این پرسش بوجود آید که چرا از توزیع تی برای تبدیل فواصل به احتمالات در فضای بعد پایین استفاده می‌کنیم؟ برای فهم بهتر، دوباره به شکل ۸ــ۱۹ توجه کنید. دوطرف توزیع تی گسترده‌تر از توزیع نرمال است. این بدان معناست که برای بدست آوردن احتمال مشابه از توزیع نرمال، موارد غیرمشابه باید از حالتی که توزیع تی متمرکز است، دورتر شوند. این کمک می‌کند تا خوشه‌هایی از داده‌ها را که ممکن است در داده‌ها وجود داشته باشند پخش کنیم و به ما کمک می‌کند آن‌ها را راحت‌تر شناسایی کنیم. با این حال، نتیجه مهم این است که اغلب گفته می شود t-SNE ساختار محلی را در بازنماییِ بعد پایین حفظ می‌کند، اما معمولا ساختار جهانی (سراسری) را حفظ نمی‌کند. در عمل، این بدان معناست که ما می‌توانیم نمونه‌هایی را که در بازنمایی نهایی به یکدیگر نزدیک هستند، شبیه به یکدیگر تفسیر کنیم، اما نمی‌توانیم به راحتی بگوییم که کدام دسته از نمونه‌ها، شبیه به سایر خوشه‌هایِ نمونه‌ها در داده‌های اصلی هستند.

در زیر طرح کلی الگوریتم t-SNE ارائه شده است:

**مرحله ۱. محاسبه فاصله بین جفت نقاط**

در مرحله اول، t-SNE جفت نقاط داده را در فضای با ابعاد بالا مقایسه می‌کند. احتمال شرطی برای هر جفت نشان می‌دهد که نقاط چقدر نزدیک هستند. متریک شباهت مورد استفاده برای ساخت توزیع احتمال می‌تواند فاصله اقلیدسی باشد. بر اساس آن، نقاط داده‌ی نزدیک، احتمال بالایی دارند در حالی که نقاط داده‌ای که به‌طور توجهی ازهم جدا شده‌اند، احتمال بی‌نهایت کوچکی دارند. از نظر ریاضی، احتمال شرطی بین دو نقطه داده $x_i$ و $x_j$ به شکل زیر است:

$$p_{j|i} = \frac{\exp\left(\frac{-\|x_i - x_j\|^{\Upsilon}}{\Upsilon\sigma_i^{\Upsilon}}\right)}{\sum_{k \neq i} \exp\left(\frac{-\|x_i - x_k\|^{\Upsilon}}{\Upsilon\sigma_i^{\Upsilon}}\right)}$$

که در آن $\sigma_i^{\Upsilon}$ واریانس توزیع گاوسی است. الگوریتم مقدار $\sigma_i$ را با توجه به **سرگشتگی**[1] این توزیع پیدا می‌کند. سرگشتگی قابلیت پیش‌بینی برخی از توزیع‌های احتمال را اندازه‌گیری می‌کند و یک پارامتر مهم در t-SNE است. کاربر می‌تواند این پارامتر را مشخص کند، مقادیر مفید در محدوده ۵ تا ۵۰ است.

چگالی احتمال یک جفت نقطه متناسب با شباهت آن است. برای نقاط داده نزدیک، $p_{j|i}$ نسبتا زیاد خواهد بود و برای نقاطی که به طور قابل توجهی از هم جدا شده‌اند، $p_{j|i}$ کوچک خواهد بود. احتمالات شرطی را در فضای بعد بالا متقارن کنید تا شباهت‌های نهایی را در فضای ابعاد بالا بدست آورید. برای اندازه‌گیری شباهت جفتی بین دو نقطه داده، احتمالات شرطی با در نظر گرفتن $N$ نقطه در مجموع، با میانگین‌گیری دو احتمال، به احتمالات مشترک $p_{ij}$ متقارن می‌شوند:

$$p_{ij} = \frac{p_{j|i} + p_{i|j}}{N}$$

**مرحله ۲. ساخت فضای بعد پایین**

پس از اندازه‌گیری احتمالات زوج نقاط برای فضای داده‌های اصلی، مرحله بعدی ساخت یک فضای بعد پایین است. شباهت‌های $q_{ij}$ بین دو نقطه داده $y_i$ و $y_j$ در فضای بعد پایین به صورت زیر محاسبه می‌شود:

$$p_{j|i} = \frac{(\Upsilon + \left\|y_i - y_j\right\|^{\Upsilon})^{-\Upsilon}}{\sum_{k \neq l}(\Upsilon + \left\|y_i - y_j\right\|^{\Upsilon})^{-\Upsilon}}$$

---

[1] perplexity

$y_i$ و $y_j$ همتایان با بعد پایین نقاط داده با ابعاد بالا با $x_i$ و $x_j$ هستند.

**مرحله ٣. کمینه کردن تفاوت بین توزیع‌های احتمال فضای با بعد بالا و فضای جدید**

گام بعدی کمینه کردنِ تفاوتِ بینِ توزیع احتمال فضای اولیه با ابعاد بالا و فضای جدید ایجاد شده با تعداد ابعاد کاهش یافته است. برای کمینه کردن این تفاوت با گرادیان کاهشی، از تابع زیان استفاده می‌شود. در مورد t-SNE این تابع زیان واگرایی کولبک ـ لیبلر (KL) است:

$$D_{KL}(p_{ij}||p_{ji}) = \sum_{i \neq j} p_{ij} log \frac{p_{ij}}{q_{ij}}$$

t-SNE با استفاده از گرادیان کاهشی و تابع زیان واگرایی کولبک ـ لیبلر نقاط را در فضای ابعاد پایین‌تر بهینه می‌کند. چرا از واگرایی KL استفاده می‌کنیم؟ وقتی واگرایی KL را کمینه می‌کنیم، $q_{ij}$ عملاً با $p_{ij}$ یکسان می‌شود، بنابراین ساختار داده‌ها در فضای ابعادی بالا، شبیه ساختار داده‌ها در فضای بعد پایین خواهد بود. بر اساس معادله واگرایی KL:

• اگر $p_{ij}$ بزرگ باشد، به یک مقدار بزرگ برای $q_{ij}$ نیاز داریم تا نقاط محلی با شباهت بالاتر را نشان دهیم.
• اگر $p_{ij}$ کوچک است، به مقدار کوچک‌تری برای $q_{ij}$ نیاز داریم تا نقاط محلی دور از هم را نشان دهیم.

**مرحله ٤. استفاده از توزیع تی برای محاسبه شباهت بین دو نقطه در فضای بعد پایین**

t-SNE از توزیع تی ـ استیودنت برای محاسبه شباهت بین دو نقطه در فضای بعد پایین به جای توزیع گاوسی استفاده می‌کند. توزیع تی، توزیع احتمالِ نقاط را در فضای ابعاد پایین‌تر ایجاد می‌کند و به کاهش مشکل ازدحام کمک می‌کند.

این روش معایب خود را نیز دارد. کاربرد مستقیم برای مجموعه داده‌های چندبعدی غیرعملی است، چراکه فواصل در ابعاد بزرگ بسیار شبیه هستند. از این‌رو، قبل از استفاده از t-SNE، داده‌های چند بعدی را می‌توان ابتدا در روش کاهش ابعاد دیگری اعمال کرد. پس از آن، در صورت لزوم، می‌توان t-SNE را اعمال کرد. قبل از وارد کردن داده‌ها، هنجارسازی مرحله مهمی است، چراکه معیار فاصله‌یِ استفاده شده فاصله اقلیدسی است. از آنجایی که الگوریتم تصادفی است، می‌توان آن را چندین بار اجرا کرد تا با توجه به واگرایی KL، راهِ حلی با کم‌ترین زیان پیدا شود.

## برآورد منیفلد یکنواخت و افکنش (UMAP)

در سال‌های اخیر، یک روش جدید کاهش ابعاد با عنوان برآورد منیفلد یکنواخت و افکنش (UMAP) ارائه شده است. این تکنیک غیرخطی است و از ایده t-SNE برای ایجاد فضای بعد

پایین برای نمایش داده‌ها پیروی می‌کند. با این حال، تفاوت‌های عمده زیادی نیز دارد. مدل ریاضی آن بر اساس هندسه و توپولوژی ریمانی است. UMAP از نرمال‌سازی در توزیع‌های احتمال استفاده نمی‌کند. در مقابل، یک نسخه هموار از کـ نزدیک‌ترین همسایه استفاده می‌شود. UMAP از نظر عملکرد نسبت به t-SNE چندین مزیت دارد. زمانی که جاسازی در فضای ابعادی جدید بزرگ‌تر از سه بعد باشد، سریع‌تر است. UMAP ساختار جهانی را بهتر حفظ می‌کند و در عین حال اطلاعات محلی را ذخیره می‌کند. طبق الگوریتم، هر نقطه در نمایش با ابعاد بالا به همسایه خود متصل است. در حالی که t-SNE دارای یک ابرپارامتر مهم (سرگشتگی) است، UMAP بسیاری از آن‌ها را دارد. مهم‌ترین ابرپارامتر تعداد همسایگان، حداقل فاصله، تعداد مولفه‌ها و متریک هستند. با این حال، مزیت این است که این ابرپارامترها به‌طور مستقیم قابل درک هستند و ما آن‌ها را با جزئیات بیشتر توضیح می‌دهیم.

ابرپارامتر تعداد نزدیک‌ترین همسایگان یکی از مهم‌ترین ابرپارامترها است. این ابرپارامتر اندازه همسایگان محلی در نظر گرفته شده را تعیین می‌کند که بر ساخت گراف اولیه با ابعاد بالا تاثیر می‌گذارد. هنگامی که مقادیر کم برای این ابرپارامتر مشخص می‌شود، جزئیات محلی حفظ می‌شود. در مقابل، با مقادیر بالا، الگوریتم به ساختار جهانی در داده‌های اولیه توجه می‌کند. این ابرپارامتر ممکن است به عنوان تغییری از ابرپارامتر سرگشتگی در t-SNE در نظر گرفته شود.

دومین ابرپارامتر مهم، حداقل فاصله است که کنترل می‌کند نقاط تا چه حد می‌توانند در یک نمایش با ابعاد پایین با یکدیگر قرار گیرند. مقادیر پایین منجر به نمایش چگال‌تر می‌شود در حالی که مقادیر بالا به تخصیص غیرمتراکم‌تری از نقاط داده منجر می‌شوند. ابرپارامترِ تعداد مولفه‌ها ابعاد فضای بعد کاهش یافته‌ای را که داده‌ها به آن نمایش داده می‌شوند را تعیین می‌کند. ابرپارامتر متریک، نحوهٔ محاسبهٔ فاصله در فضای محیطی داده‌های ورودی را کنترل می‌کند.

تابع زیانی که UMAP استفاده می‌کند، آنتروپیِ متقابلِ بین نمایش‌های توپولوژیکی فضاهای با ابعاد بالا و پایین است. این نیز یک تفاوت مهم در مقایسه با t-SNE است که از واگرایی KL به عنوان تابع زیان استفاده می‌کند. با این حال، مشابه بسیاری از تکنیک‌های کاهش ابعاد، UMAP از گرادیان کاهشی برای کمینه کردن این تابع زیان استفاده می‌کند.

اشکال این تکنیک کاهش ابعاد، این است که انتخاب مجموعه خوبی از ابرپارمترها بی‌اهمیت نیست. فواصل بین خوشه‌ها ممکن است بی‌معنی باشد، چراکه فواصل محلی هنگام ساخت گراف در نظر گرفته می‌شود. الگوریتم UMAP تصادفی است و به کمینه محلی ختم می‌شود، از این‌رو، بهتر است که با ابرپارامترهای یکسان چندین بار بدست آوریم. UMAP از تصادفی بودن هم برای سرعت بخشیدن به مراحل برآورد و هم برای کمک به حل مسائل سخت بهینه‌سازی استفاده می‌کند. این بدان معناست که اجراهای مختلف UMAP می‌توانند نتایج متفاوتی تولید کنند. UMAP نسبتا پایدار است. بنابراین واریانس بین اجراها در حالت ایده‌آل باید نسبتا کوچک

باشد، اما اجراهای مختلف ممکن است تغییراتی داشته باشند. برای اطمینان از اینکه نتایج را می‌توان دقیقا بازتولید کرد، می‌توان یک حالت دانه تصادفی تنظیم کرد.

# خودرمزنگار

خودرمزنگارها (یا رمزنگار خودکار) شبکه‌های عصبی هستند که برای کاهش ابعاد استفاده می‌شود. هدف یک خودرمزنگار یادگیریِ بازنمایی فشرده از داده‌های ورودی، با دستیابی به بازسازی در خروجی است. به عبارت ساده‌تر، خودرمزنگارها، توانایی کشف بازنمایی‌هایِ بعد پایین داده‌های با ابعاد بالا را دارند و قادر به بازتولید ورودی در خروجی هستند که سعی می‌کنند ورودی $x$ را به خودش منتقل کنند. یادگیری رونوشت کردن از ورودی به خروجی ممکن است بی‌اهمیت به نظر برسد، اما با تحمیل برخی محدودیت‌ها بر ساختار شبکه، خودرمزنگار را مجبور می‌کند که مهم‌ترین ویژگی‌های داده‌های آموزشی را بیاموزد، چراکه نمی‌تواند همه چیز را مدل‌سازی کند.

ایده پشت خودرمزنگارها این است که داده‌های ورودی با ابعاد بالا را از طریق یک گلوگاه اطلاعاتی دریافت کند و در طول فرآیند آن را تبدیل به یک بازنمایی با ابعاد پایین کند، سپس بازتولید داده‌های ورودی را از این بازنمایی آموخته‌شده، ایجاد کند. با نگاشت ورودی در فضایی با ابعاد کاهش یافته، شبکه عصبی قادر به یادگیری و استخراج ویژگی‌های مختلف است.

حال، شاید برای خواننده این پرسش بوجود آید که چرا ما به خود زحمت می‌دهیم تا بازنمایی ورودی اصلی را فقط برای بازتولید هرچه بهتر خروجی یاد بگیریم؟ پاسخ این است که وقتی ورودی با ویژگی‌هایِ زیاد داریم، تولید یک بازنمایی فشرده از طریق لایه‌های پنهان شبکه عصبی، می‌تواند به فشرده‌سازی نمونه آموزشی ورودی کمک کند. بنابراین وقتی شبکه عصبی تمام داده‌های آموزشی را مرور می‌کند و وزن تمام گره‌های لایه پنهان را تنظیم می‌کند، آنچه اتفاق می‌افتد این است که وزن‌ها واقعا نوعی ورودی را نشان می‌دهند که ما معمولا می‌بینیم. در نتیجه، اگر بخواهیم نوع دیگری از داده‌ها را وارد کنیم، همانند داشتن داده‌هایی با مقداری نویز، شبکه خودرمزنگار، قادر خواهد بود نویز را در ورودی تشخیص داده و حداقل بخشی از نویز را هنگام تولید خروجی حذف کند.

خودرمزنگارها از دو قطعه شبکه عصبی، **رمزگذار** و **رمزگشا** تشکیل شده‌اند. رمزگذار ابعاد یک مجموعه داده با ابعاد بالا را به یک مجموعه با ابعاد کم کاهش می‌دهد در حالی که رمزگشا اساسا داده‌های ابعاد پایین را به داده‌هایی با ابعاد بالا گسترش می‌دهد. هدف چنین فرآیندی تلاش برای بازتولید ورودی اصلی است. اگر شبکه عصبی به خوبی ساخته شده باشد، شانس خوبی برای بازتولید ورودی اصلی از داده‌های رمزشده وجود دارد. ساختار یک خودرمزنگار خودکار در شکل ۸ ــ ۲۰ نشان داده شده است.

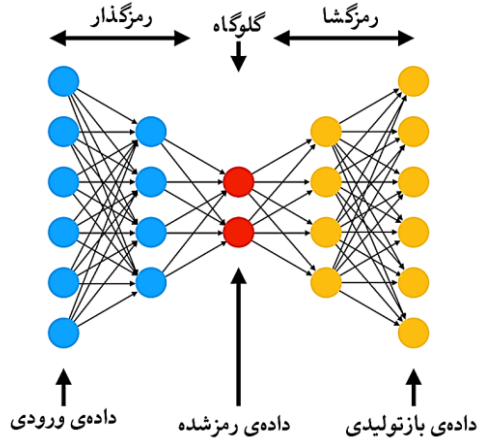

**شکل ۸ـ۲۰.** ساختار یک خودرمزنگار

در عمل، از خودرمزنگار می‌توان برای کاهش ابعاد ویژگی، مشابه با سایر روش‌ها استفاده کرد، اما عملکرد آن قوی‌تر است. چراکه مدل‌های شبکه عصبی می‌توانند ویژگی‌های جدید را کارآمدتر استخراج کنند. از منظرِ ساختاری، خودرمزنگارها بسیار شبیه به شبکه‌های پیش‌خور هستند، در ساده‌ترین شکل، آن‌ها یک لایه ورودی، یک لایه پنهان و یک لایه خروجی دارند. تفاوت ساختاری کلیدی با شبکه‌های پیش‌خور این است که خودرمزنگارها تعداد گره‌های یکسانی در لایه‌های ورودی و خروجی خود دارند.

با این حال، تفاوت عمده‌یِ بین خودرمزنگار و شبکه‌ی پیش‌خور، در فرآیند یادگیری آن‌ها نهفته است؛ خودرمزنگارها از داده‌های بدون‌برچسب به جای داده‌های برچسب‌دار استفاده می‌کنند. انواع مختلفی از خودرمزنگارها وجود دارد. ابتدا نسخه استاندارد آن را شرح می‌دهیم و سپس نسخه‌های دیگر آن را مورد بررسی قرار خواهیم داد.

یکی از کاربردهای رایج خودرمزنگار برای کاهش ابعاد است. ابعاد بازنمایی نهفته معمولا بسیار کوچکتر از ابعاد ورودی اصلی است و اگر رمزگذار به خوبی عمل کند، بازنمایی نهفته به عنوان یک نسخه فشرده خوب از نسخه اصلی، با ابعاد ورودیِ بسیار کم‌تر عمل می‌کند.

### خودرمزنگار استاندارد

خودرمزنگار استاندارد ساختاری متشکل از دو بخش است. بخش اول یک رمزگذار است که داده‌های ورودی را دریافت می‌کند و یاد می‌گیرد که آن‌ها را برای بدست آوردن یک رمزگذاری، فشرده کند (که به عنوان **رمز** یا **بازنمایی نفهته**[1] نیز شناخته می‌شود). این عمل می‌تواند با

---

[1] latent representation

یادگیریِ مهم‌ترین ویژگی‌هایِ داده‌ها صورت گیرد. از منظرِ ریاضی، رمز، برای یک لایه پنهان و ورودیِ $x$ را می‌توان به صورت زیر نشان داد:

$$h = \sigma(Wx + b)$$

که در آن $h$ بازنماییِ نهفته، $\sigma$ تابع فعال‌سازی لایه شبکه، $W$ و $b$ به ترتیب ماتریس‌وزن و بایاس هستند.

بخش دوم رمزگشا است که داده‌یِ رمزشده را دریافت می‌کند. هدف رمزگشا بازسازی رمز به گونه‌ای است که بازتولید حاصل، تا حد امکان به داده‌هایِ ورودیِ رمزگذار نزدیک شود. بازتولید $\acute{x}$ در این مورد را می‌توان با

$$\acute{x} = \acute{\sigma}(\acute{W}h + \acute{b})$$

نشان داد، که $\acute{\sigma}$، $\acute{W}$ و $\acute{b}$ پارامترهای مختلف تابع فعال‌سازی، ماتریس وزن و بایاس هستند.

اگر رمزگذار با تابعی به شکل $h = f(x)$ و رمزگشا را با تابعی به شکل $r = g(h)$ نشان دهیم، آنگاه، کل یک خودرمزنگار را می‌توان با یک تابع $r = g\big(f(x)\big)$ توصیف کرد، که در آن خروجی $r$ مشابه ورودی اصلی $x$ است. هدف از آموزش خودرمزنگار بدست آوردن یک $h$ مفید است که بتواند ورودی $x$ را به خوبی نشان دهد. مدل ایده آل خودرمزنگار، نکات زیر را در نظر می‌گیرد:

- به اندازه کافی به ورودی حساس است تا بتواند بازتولید را با دقت بسازد.
- به اندازه کافی نسبت به ورودی حساس نیست، به طوری که مدل به سادگی داده‌های آموزشی را به خاطر نمی‌سپارد. به عبارت دیگر، بیش‌برازش نمی‌کند.

این موازنه را مدل مجبور می‌کند که فقط تغییرات داده‌ای را که برای بازتولید ورودی لازم است را بدون حفظ افزونگی در ورودی حفظ کند. در بیشتر موارد، این عمل شامل ساخت یک تابع زیان است. تابع زیان، برای ارزیابیِ میزان تفاوت مقدار پیش‌بینی شده‌ی مدل با مقدار واقعی استفاده می‌شود و هرچه زیان کمتر باشد، معمولا عملکرد مدل بهتر است. تابع زیان برای خودرمزنگار به صورت $L(x, g(f(x)))$ تعریف می‌شود، که در آن $L$ تابع زیان برای محاسبه اختلاف $x$ و $g(f(x))$ است. هدف آموزشِ خودرمزنگار، کمینه کردن تابع زیان (زیان بازتولید) است. شبکه را می‌توان با استفاده از تکنیک‌های استاندارد شبکه‌های عمیق همانند پس‌انتشار آموزش داد.

با آموزش یک خودرمزنگار که در بازتولید داده‌های ورودی خوب عمل کده است، امیدواریم که بازنماییِ نفهته $h$ بتواند برخی از ویژگی‌های مفید در داده‌ها را به تصویر بکشد. برای جلوگیری از راه‌حل‌های بی‌اهمیت و یادگیری ویژگی‌های مفید، باید محدودیت هایی را به خودرمزنگار اضافه کنیم.

خودرمزنگار می‌تواند برای استخراج ویژگی‌های مفید با مجبور کردن $h$ به ابعاد کوچکتر از $x$ استفاده شود. خودرمزنگاری که بعد نهفته آن کمتر از بعد ورودی باشد، **خودرمزنگار ناقص**[1] نامیده می‌شود. یادگیری یک بازنمایی ناقص، خودرمزنگار را مجبور می‌کند تا برجسته‌ترین ویژگی‌های داده‌های آموزشی را ثبت کند. به عبارت دیگر، بازنمایی نهفته $h$ یک بازنمایی توزیع شده است که مختصات را در امتداد عوامل اصلی تغییر در داده‌ها بدست می‌آورد. این شبیه به روشی است که نگاشتِ (افکنش) مؤلفه‌های اصلی تغییر در داده‌ها را نشان می‌دهد. در واقع، اگر یک لایه‌ی پنهان خطی وجود داشته باشد و از معیار میانگین مربعات خطا برای آموزش شبکه استفاده شود، آنگاه واحدهای پنهان یاد می‌گیرند که ورودی را در گستره اولین مؤلفه‌های اصلی بازنمایی کنند. اگر لایه پنهان غیرخطی باشد، خودرمزنگار با PCA متفاوت رفتار می‌کند و می‌تواند جنبه‌های چندوجهی توزیع ورودی را به تصویر بکشد.

انتخاب دیگر این است که $h$ را به‌گونه‌ای محدود کنیم که بُعدِ بزرگتر از $x$ داشته باشد. خودرمزنگاری که ابعاد نهفته آن بزرگتر از بعد ورودی است، **خودرمزنگار فراکامل**[2] نامیده می‌شود. با این حال، به دلیل ابعاد بزرگ، رمزگذار و رمزگشا میل به بیش‌برازش پیدا می‌کنند. از این‌رو، در چنین مواردی، حتی یک رمزگذار و رمزگشای خطی هم می‌تواند یاد بگیرد که ورودی را در خروجی رونوشت کند، بدون اینکه هیچ بازنماییِ مفیدی در مورد توزیع داده‌ها یاد بگیرد. خوشبختانه، هنوز هم می‌توانیم ساختار جالبی را با اعمال محدودیت‌های دیگر بر روی شبکه کشف کنیم. یکی از پرکاربردترین محدودیت‌ها، محدودیت **پراکندگی**[3] در $h$ است. خودرمزنگار فراکامل با محدودیت پراکندگی، **خودرمزنگار خلوت**[4] نامیده می‌شود.

## خودرمزنگار حذف نویز

خودرمزنگار حذف نویز با خودرمزنگار استاندارد از یک جهت متفاوت است. سیگنال ورودی در ابتدا تا حدی در خودرمزنگار حذف نویز خراب می‌شود و بعدا به شبکه تغذیه می‌شود. آموزش شبکه به گونه‌ای انجام می‌شود که جریان داده ورودی از داده‌های نسبتا خراب بازیابی می‌شود. این امر به خودرمزنگار اجازه می‌دهد تا ساختار اولیه سیگنال‌های ورودی را برای بازتولید مناسبِ بردار ورودیِ اصلی درک کند، تا بردار ورودی اصلی را به اندازه کافی بازتولید کند. همچنان که پیش‌تر بیان شد، معمولا خودرمزنگارها تابع زیان $L$ را کاهش می‌دهند. خودرمزنگار حذف نویز تابع زیان زیر را کاهش می‌دهد:

$$L(x, g(f(\hat{x})))$$

---

که در آن $\acute{x}$ یک رونوشت از $x$ است که با نویز خراب شده است. مکانیسم خودرمزنگار حذف نویز در شکل ۸ ــ ۲۱ نشان داده شده است.

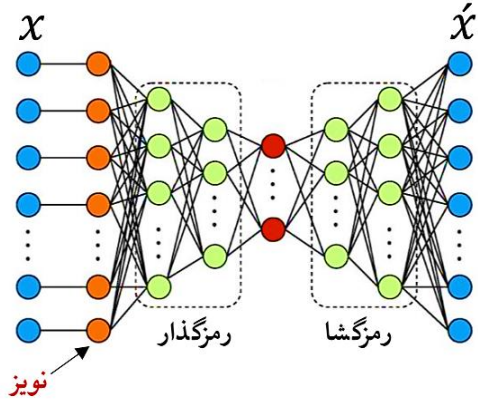

**شکل ۸ ــ ۲۱.** خودرمزنگار حذف نویز

اضافه کردن نویز به ورودی $x$ در طول آموزش برای اطمینان از اینکه خودرمزنگار ویژگی‌های مفید توزیع داده را می‌آموزد، انجام می‌شود.

## خودرمزنگار خلوت

خودرمزنگار خلوت می‌تواند ساختارهای ارزشمندی را در داده‌های ورودی با تحمیل پراکندگی به واحدهای پنهان در طول آموزش بیاموزد. پراکندگی عموما می‌تواند با افزودن عبارت‌های اضافی ($\Omega$) به تابع زیان در طول آموزش یا با صفر کردن دستی همه واحدهای پنهان به جز چند مورد ضروری بدست آید:

$$L\left(x, g\big(f(x)\big)\right) + \Omega(h)$$

با جریمه‌سازی تابع زیان در لایه‌های پنهان، منجر به این می‌شود که با ورود یک نمونه جدید، تنها چند چند فعال شود. شهود پشت این روش این است که، برای مثال، اگر شخصی ادعا کند که در ریاضیات، علوم رایانه، روانشناسی، فیزیک و شیمی متخصص است، ممکن است در حال یادگیری دانش کاملا سطحیِ این موضوعات باشد. با این حال، اگر او فقط ادعا می‌کند که در علوم رایانه تخصص دارد، مایلیم بینش‌های مفیدی را از او بدست آوریم و برای خودرمزنگاری که آموزش می‌دهیم نیز همین‌طور است؛ تعداد گره‌های کمتری که هنوز با فعال‌سازی عملکرد خود را حفظ می‌کنند، تضمین می‌کنند که خودرمزنگار در واقع به‌جای اطلاعات اضافی در داده‌های ورودی، بازنمایی پنهان را یاد می‌گیرد.

یکی از پیشرفت‌هایی که در زمینه خودرمزنگار خلوت وجود دارد، **خودرمزنگار کـ ــ خلوت** است. در این خودرمزنگار ما **k** نورون با بالاترین توابع فعال‌سازی را انتخاب می‌کنیم و سایر

توابع فعال‌سازی را با استفاده از توابع فعال‌سازی ReLU نادیده می‌گیریم و آستانه را برای یافتن بزرگترین نورون‌ها تنظیم می‌کنیم. این مقدار k را تنظیم می‌کند تا سطح پراکندگی به بهترین وجه برای مجموعه داده به دست آید.

### خودرمزنگار پشته‌ای[1]

برخی از مجموعه داده‌ها رابطهٔ پیچیده‌ای در ویژگی‌ها دارند. بنابراین، استفاده از تنها یک خودرمزنگار کافی نیست. چرا که ممکن است یک خودرمزنگارِ منفرد نتواند ابعاد ویژگی‌های ورودی را کاهش دهد. بنابراین برای چنین مواردی، از خودرمزنگار پشته‌ای استفاده می‌کنیم. خودرمزنگار پشته‌ای، همان‌طور که از نام آن پیداست، رمزگذارهای متعددی هستند که روی هم چیده شده‌اند.

# کاهش ابعاد با پایتون

## PCA در پایتون

### مجموعه داده

در حالی که بیشتر مجموعه داده‌های دنیای واقعی مانند تصاویر و داده‌های متنی ابعاد بسیار بالایی دارند، ما از مجموعه داده‌های ارقام دست‌نویس MNIST برای سادگی استفاده می‌کنیم. مجموعه داده MNIST مجموعه‌ای از تصاویر خاکستری از ارقام دست‌نویس بین ۰ تا ۹ است که شامل ۶۰۰۰۰ تصویر در اندازه ۲۸×۲۸ پیکسل است. بنابراین، این مجموعه داده دارای ۶۰۰۰۰ نمونه داده با ابعاد ۷۸۴ است. برای نشان دادن کاهش ابعاد در این مجموعه داده، ما از PCA برای کاهش ابعاد داده‌ها استفاده می‌کنیم و داده‌ها را در فضای ویژگی با ابعاد پایین نمایش می‌دهیم. این مثال داده‌ها را با ۷۸۴ ویژگی به فضای ویژگی‌های دو بعدی ترسیم می‌کند و نتایج را مصورسازی می‌کند.

### وارد کردن کتاب‌خانه‌ها

```
In [1]:   from keras.datasets import mnist, fashion_mnist
          import time
          import numpy as np
          import pandas as pd
          from sklearn.decomposition import PCA
          import matplotlib.pyplot as plt
          from matplotlib import colors as mcolors
          import seaborn as sns
```

---

[1] Stacked Autoencoder

**وارد کردن مجموعه داده**

```
In [2]:    (X_train, y_train) , (X_test, y_test) = mnist.load_data()
```

**آماده‌سازی داده‌ها**

یک آرایه به تعداد تصاویر و تعداد پیکسل‌ها در تصویر ایجاد می‌کنیم و داده‌های X_train را در X رونوشت می‌کنیم.

```
In [4]:    X = np.zeros((X_train.shape[0], 784))
           for i in range(X_train.shape[0]):
               X[i] = X_train[i].flatten()
```

مجموعه داده را درهم می‌زنیم و ۲۵٪ از داده‌های آموزشی MNIST را برمی‌داریم و آن‌ها را در یک قاب داده ذخیره می‌کنیم.

```
In [4]:    X = pd.DataFrame(X)
           Y = pd.DataFrame(y_train)
           X = X.sample(frac=0.25,
           random_state=1400).reset_index(drop=True)
           Y = Y.sample(frac=0.25,
           random_state=1400).reset_index(drop=True)
           df = X
```

**استفاده از PCA در مجموعه داده MNIST**

```
In [5]:    pca = PCA(n_components=2)

           pca_results = pca.fit_transform(df.values)
```

PCA دو بعد ایجاد می‌کند، مولفه اصلی ۱ و مولفه اصلی ۲. دو مولفه PCA را به همراه برچسب آن‌ها در یک قاب داده اضافه می‌کنیم. *برچسب تنها برای مصورسازی مورد نیاز است.*

```
In [1]:    pca_df = pd.DataFrame(data = pca_results
                  , columns = ['pca_1', 'pca_2'])
           pca_df['label'] = Y
```

**مصورسازی**

در این مرحله ما الگوریتم خود را آموزش داده‌ایم و برخی پیش‌بینی‌ها را انجام داده‌ایم. اکنون می‌خواهیم ببینیم که الگوریتم ما چقدر دقیق است.

```
In [3]:    fig = plt.figure(figsize = (8,8))
           ax = fig.add_subplot(1,1,1)
           ax.set_xlabel('Principal Component 1', fontsize = 15)
           ax.set_ylabel('Principal Component 2', fontsize = 15)
           ax.set_title('2 component PCA', fontsize = 20)
```

```
targets = [0,1,2,3,4,5,6,7,8,9]
colors = ['r', 'g', 'b']
colors = dict(mcolors.BASE_COLORS,
**mcolors.CSS4_COLORS)
colors=['yellow', 'black', 'cyan', 'green', 'blue', 'red',
'brown','crimson', 'gold', 'indigo']
for target, color in zip(targets,colors):
    indicesToKeep = pca_df['label'] == target
    ax.scatter(pca_df.loc[indicesToKeep, 'pca_1']
            , pca_df.loc[indicesToKeep, 'pca_2']
            , c = color
            , s = 50)
ax.legend(targets)
ax.grid()
```

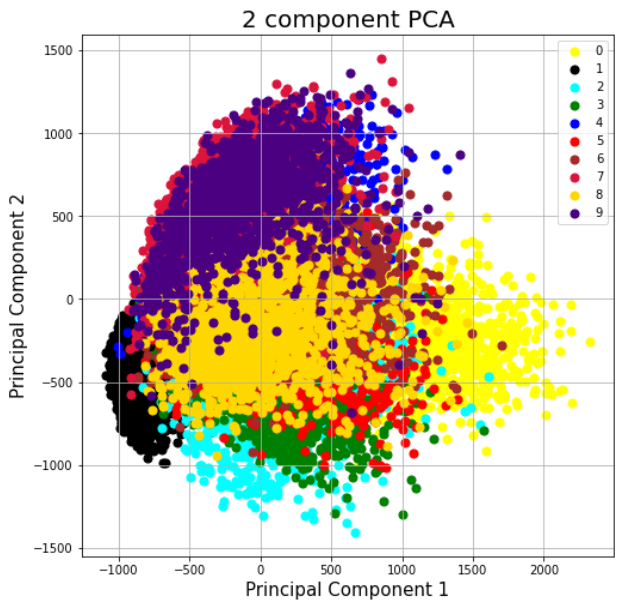

### t-SNE در پایتون

### وارد کردن کتاب‌خانه‌ها

```
In [1]:  from sklearn.manifold import TSNE
         from keras.datasets import mnist
         import matplotlib.pyplot as plt
         from mpl_toolkits.mplot3d import Axes3D
         import seaborn as sns
         import matplotlib.patheffects as PathEffects
         import numpy as np
         import pandas as pd
```

### وارد کردن مجموعه داده

```
In [2]:  (X_train, y_train) , (X_test, y_test) = mnist.load_data()
```

### آماده‌سازی داده‌ها

```
In [4]:  X = np.zeros((X_train.shape[0], 784))
         for i in range(X_train.shape[0]):
             X[i] = X_train[i].flatten()
```

مجموعه داده را درهم می‌زنیم و ۸۰٪ از داده‌های آموزشی MNIST را برمی‌داریم و آن‌ها را در یک قاب داده ذخیره می‌کنیم.

```
In [4]:  X = pd.DataFrame(X)
         Y = pd.DataFrame(y_train)
         X = X.sample(frac=0.80,
         random_state=10).reset_index(drop=True)
         Y = Y.sample(frac=0.80,
         random_state=10).reset_index(drop=True)
         df = X
         df['label'] = Y
```

### استفاده از **t-SNE** در مجموعه داده **MNIST**

```
In [5]:  tsne = TSNE(n_components=2, verbose=1,
         perplexity=40, n_iter=300)
         tsne_results = tsne.fit_transform(df)

out [5]:  [t-SNE] Computing 121 nearest neighbors...
          [t-SNE] Indexed 48000 samples in 0.026s...
          [t-SNE] Computed neighbors for 48000 samples in 139.468s...
          [t-SNE] Computed conditional probabilities for sample 1000 / 48000
          [t-SNE] Computed conditional probabilities for sample 2000 / 48000
          [t-SNE] Computed conditional probabilities for sample 3000 / 48000
          [t-SNE] Computed conditional probabilities for sample 4000 / 48000
          [t-SNE] Computed conditional probabilities for sample 5000 / 48000
          [t-SNE] Computed conditional probabilities for sample 6000 / 48000
          [t-SNE] Computed conditional probabilities for sample 7000 / 48000
          [t-SNE] Computed conditional probabilities for sample 8000 / 48000
          [t-SNE] Computed conditional probabilities for sample 9000 / 48000
          [t-SNE] Computed conditional probabilities for sample 10000 / 48000
          [t-SNE] Computed conditional probabilities for sample 11000 / 48000
          [t-SNE] Computed conditional probabilities for sample 12000 / 48000
          [t-SNE] Computed conditional probabilities for sample 13000 / 48000
          [t-SNE] Computed conditional probabilities for sample 14000 / 48000
          [t-SNE] Computed conditional probabilities for sample 15000 / 48000
          [t-SNE] Computed conditional probabilities for sample 16000 / 48000
          [t-SNE] Computed conditional probabilities for sample 17000 / 48000
          [t-SNE] Computed conditional probabilities for sample 18000 / 48000
          [t-SNE] Computed conditional probabilities for sample 19000 / 48000
          [t-SNE] Computed conditional probabilities for sample 20000 / 48000
          [t-SNE] Computed conditional probabilities for sample 21000 / 48000
          [t-SNE] Computed conditional probabilities for sample 22000 / 48000
          [t-SNE] Computed conditional probabilities for sample 23000 / 48000
          [t-SNE] Computed conditional probabilities for sample 24000 / 48000
          [t-SNE] Computed conditional probabilities for sample 25000 / 48000
          [t-SNE] Computed conditional probabilities for sample 26000 / 48000
```

```
[t-SNE] Computed conditional probabilities for sample 27000 / 48000
[t-SNE] Computed conditional probabilities for sample 28000 / 48000
[t-SNE] Computed conditional probabilities for sample 29000 / 48000
[t-SNE] Computed conditional probabilities for sample 30000 / 48000
[t-SNE] Computed conditional probabilities for sample 31000 / 48000
[t-SNE] Computed conditional probabilities for sample 32000 / 48000
[t-SNE] Computed conditional probabilities for sample 33000 / 48000
[t-SNE] Computed conditional probabilities for sample 34000 / 48000
[t-SNE] Computed conditional probabilities for sample 35000 / 48000
[t-SNE] Computed conditional probabilities for sample 36000 / 48000
[t-SNE] Computed conditional probabilities for sample 37000 / 48000
[t-SNE] Computed conditional probabilities for sample 38000 / 48000
[t-SNE] Computed conditional probabilities for sample 39000 / 48000
[t-SNE] Computed conditional probabilities for sample 40000 / 48000
[t-SNE] Computed conditional probabilities for sample 41000 / 48000
[t-SNE] Computed conditional probabilities for sample 42000 / 48000
[t-SNE] Computed conditional probabilities for sample 43000 / 48000
[t-SNE] Computed conditional probabilities for sample 44000 / 48000
[t-SNE] Computed conditional probabilities for sample 45000 / 48000
[t-SNE] Computed conditional probabilities for sample 46000 / 48000
[t-SNE] Computed conditional probabilities for sample 47000 / 48000
[t-SNE] Computed conditional probabilities for sample 48000 / 48000
[t-SNE] Mean sigma: 451.266121
[t-SNE] KL divergence after 250 iterations with early exaggeration: 95.580673
[t-SNE] KL divergence after 300 iterations: 4.694413
```

**مصورسازی**

اکنون که دو بعد حاصل را داریم، می‌توانیم با ایجاد یک نمودار نقطه‌ای از دو بعد و رنگ‌آمیزی هر نمونه با برچسب مربوط، آن‌ها را مصورسازی کنیم.

```
In [3]:   df['tsne-2d-one'] = tsne_results[:,0]
          df['tsne-2d-two'] = tsne_results[:,1]
          plt.figure(figsize=(16,10))
          sns.scatterplot(
              x="tsne-2d-one", y="tsne-2d-two",
              hue=df['label'],
              palette=sns.color_palette("hls", 10),
              data=df,
              legend="full",
              alpha=0.3)
```

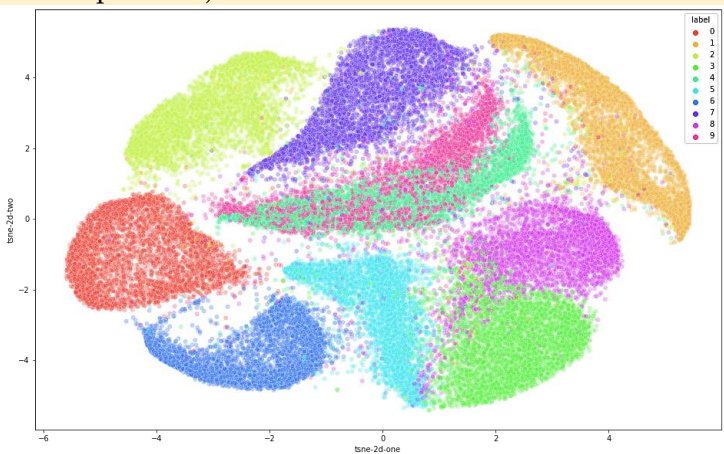

# مدل‌های مولد

مدل‌های مولد (تولیدی) دسته‌ای از مدل‌های یادگیری ماشین هستند که برای توصیف نحوه تولید داده‌ها استفاده می‌شوند. برای آموزش یک مدل مولد، ابتدا حجم زیادی از داده‌ها را در هر حوزه‌ای جمع‌آوری می‌کنیم و بعدا مدلی را برای ایجاد یا تولید داده‌هایی مانند آن آموزش می‌دهیم. به عبارت دیگر، مدل‌های مولد توانایی این را دارند تا یاد بگیرند تا داده‌هایی شبیه به داده‌هایی که ما به آنها تغذیه کرده‌ایم را ایجاد کنند.

اساسا در یادگیری ماشین دو نوع مدل وجود دارد: **مدل‌های مولد**[1] و **مدل‌های تفکیک‌کننده**[2]. مدل تفکیک‌کننده، $P(Y|X)$ را یاد می‌گیرد، که رابطه شرطی بین متغیر هدف Y و ویژگی‌های X است و رویکردی برای مرتب کردن رابطه بین متغیرها است. در مقابل، هدف مدل مولد توصیف احتمالیِ کاملِ مجموعه داده است. به عبارت دیگر، روشی برای یادگیری هر نوع توزیع داده است. مدل مولد، در یادگیری ماشین غیرنظارتی به عنوان وسیله‌ای برای توصیف پدیده‌ها در داده‌ها استفاده می‌شود و رایانه‌ها را قادر می‌سازد دنیای واقعی را درک کنند.

فرض کنید ما یک مساله یادگیری بانظارت داریم، که در آن $x_i$ ویژگی‌های داده شده نقاط داده و $y_i$ برچسب‌های مربوط هستند. یکی از راه‌های پیش‌بینی $y$، یادگیری تابع $f()$ از $(x_i, y_i)$ است؛ به‌طوری که $x$ را دریافت می‌کند و محتمل‌ترین $y$ را به عنوان خروجی برمی‌گرداند. چنین مدل‌هایی در دسته مدل‌های تفکیک‌کننده قرار می‌گیرند، چراکه در حال یادگیری نحوه تفکیک بین $x$ از کلاس‌های مختلف هستید. روش‌هایی مانند شبکه‌های بردار پشتیبان، شبکه‌های عصبی و درخت تصمیم در این دسته قرار می‌گیرند. با این حال، حتی اگر بتوانید داده‌ها را بسیار دقیق دسته‌بندی کنید، هیچ تصوری از نحوه تولید داده‌ها ندارید.

رویکرد دوم این است که چگونه داده‌ها را تولید کرده و یک تابع $f(x, y)$ را یاد بگیریم که به پیکربندی تعیین‌شده توسط $x$ و $y$ با هم امتیاز می‌دهد. سپس می‌توان $y$ را برای $x$ جدید با یافتن $y$ای که امتیاز $f(x, y)$ را بیشینه می‌کند، پیش‌بینی کرد. برای درک بهتر این دو دسته از مدل‌ها، مثالی می‌زنیم. $x$ را به عنوان یک تصویر و $y$ را نوعی حیوان همانند یک گربه در تصویر، تصور کنید. احتمالی که به صورت $p(y|x)$ نوشته می‌شود به ما می‌گوید که مدل چقدر معتقد است که یک گربه وجود دارد، با توجه به یک تصویر ورودی در مقایسه با همه احتمالاتی که درباره آن می‌داند. الگوریتم‌هایی که سعی می‌کنند به‌طور مستقیم این نگاشت احتمال را مدل کنند، مدل‌های تفکیک‌کننده نامیده می‌شوند. از سوی دیگر، مدل‌های مولد سعی می‌کنند تابعی به نام احتمال توأم $p(y, x)$ را یاد بگیرند. می‌توانیم این را اینگونه بخوانیم که چقدر مدل معتقد

---

است که $x$ یک تصویر است و یک گربه $y$ هم‌زمان در آن وجود دارد. این دو احتمال به‌هم مرتبط هستند و می‌توان آن‌ها را به صورت $p(y,x) = p(x)p(y|x)$ نوشت. در این رابطه، $p(x)$ نشان می‌دهد که چقدر احتمال دارد ورودی $x$ یک تصویر باشد. احتمال $p(x)$ معمولاً تابع چگالی نامیده می‌شود.

دلیل اصلی نامیدن این مدل‌ها به مولد به این واقعیت مربوط می‌شود که مدل به احتمال ورودی و خروجی هم‌زمان دسترسی دارد. بر این اساس، می‌توانیم با نمونه‌برداری از انواع حیوانات $y$ و تصاویر جدید $x$ در $p(y,x)$ تصاویری از حیوانات تولید کنیم. از این‌رو، مدل‌های مولد یک مزیت جالب نسبت به مدل‌های تفکیک‌کننده دارند، یعنی آن‌ها پتانسیل درک و توضیح ساختار زیربنایی داده‌های ورودی را دارند؛ حتی زمانی که هیچ برچسبی در دسترس نباشد.

> به طور کلی، یک مدل تفکیک‌کننده مستقیماً مدلی را برای توزیع شرطی می‌آموزد، به عنوان مثال، $p(y,x)$ که در آن $y$ یک متغیر هدف و $x$ ورودی‌ها هستند. در مقابل، یک مدل مولد توزیع توأم $p(y,x)$ را می‌آموزد و به اصطلاح گفته می‌شود که می‌تواند نمونه‌هایی از توزیع داده تولید کند.

اگر مدلی واقعاً قادر به تولید نمونه‌های جدیدی باشد که از پیدایش اشیا دنیای واقعی پیروی می‌کنند، در واقع می‌توان گفت که مفهومی را بدون آموزش یاد گرفته و درک کرده است. از این‌رو، این دسته از مدل‌ها، در رده مدل‌های غیرنظارتی (مدل‌های مولد را می‌توان در رده مدل‌های خودنظارتی نیز قرار داد) قرار می‌گیرد.

این مدل‌ها عموماً بر روی شبکه‌های عصبی اجرا می‌شوند و می‌توانند به طور طبیعی ویژگی‌های متمایز داده‌ها را تشخیص دهند. شبکه‌های عصبی این درک بنیادی از داده‌های دنیای واقعی را دریافت می‌کنند و سپس از آن‌ها برای مدل‌سازی داده‌هایی استفاده می‌کنند که مشابه داده‌های دنیای واقعی هستند.

هدف اصلی انواع مدل‌های مولد یادگیری توزیعِ واقعیِ داده‌هایِ مجموعه آموزشی است تا نقاط داده جدید با تغییراتی تولید شوند. اما این امکان برای مدل وجود ندارد که توزیع دقیق داده‌های ما را بیاموزد و بنابراین توزیعی را مدل‌سازی می‌کنیم که مشابه توزیع داده‌های واقعی است. برای این کار، ما از دانش شبکه‌های عصبی برای یادگیری تابعی استفاده می‌کنیم که می‌تواند توزیع مدل را به توزیع واقعی تقریب بزند.

## انواع مدل مولد

هدف مدل‌های مولد یادگیری تابع چگالی احتمال $p(x)$ است. این چگالی احتمال به‌طور مؤثر رفتار داده‌های آموزشی ما را توصیف می‌کند و ما را قادر می‌سازد تا داده‌های جدیدی را با نمونه‌گیری از توزیع تولید کنیم. در حالت ایده‌آل، می‌خواهیم مدل ما چگالی احتمال $p(x)$ را

یاد بگیرد که با چگالی داده‌های $p_{data}(x)$ یکسان باشد. برای رسیدن به این هدف، راهبردهای مختلفی وجود دارد.

دسته اول مدل‌ها می‌توانند تابع چگالی $p$ را به طور **صریح**[1] محاسبه کنند یا سعی کنند آن را تقریب بزنند. یعنی، بعد از آموزش می‌توانیم یک نقطه داده $x$ را به مدل وارد کنیم و مدل به ما درست‌نمایی نقطه داده را می‌دهد که حاصل $p(x)$ است. از این مدل‌ها به عنوان مدل‌های مولد صریح نام برده می‌شود. دسته دوم که به عنوان مدل‌های مولد **ضمنی**[2] شناخته می‌شوند، $p(x)$ را محاسبه نمی‌کنند. با این حال، پس از آموزش مدل، می‌توانیم از توزیع زیربنایی نمونه‌برداری کنیم.

## مدل‌های متغیر نهفته

مساله اصلی در یادگیری ماشین، یادگیری توزیع احتمال پیچیده $p(x)$ با مجموعه محدودی از نقاط داده با ابعاد بالای $x$ است که از این توزیع گرفته شده‌اند. به عنوان مثال، برای یادگیری توزیع احتمال بر روی تصاویر گربه‌ها، باید توزیعی را تعریف کنیم که بتواند همبستگی‌های پیچیده بین تمام پیکسل‌هایی که هر تصویر را تشکیل می‌دهند، مدل کند. مدل‌سازی مستقیم این توزیع یک کار چالش‌برانگیز یا حتی غیرممکن است.

به‌جای مدل‌سازی مستقیم $p(x)$ می‌توانیم یک متغیر نهفته z را معرفی کنیم و یک توزیع شرطی $p(x|z)$ را برای داده‌ها تعریف کنیم که به آن درست‌نمایی می‌گویند. در روابط احتمالی z را می‌توان به عنوان یک متغیر تصادفی پیوسته تفسیر کرد. برای مثال در تصاویر گربه، z می‌تواند حاویِ بازنمایی نهفته‌ای از نوعِ گربه، رنگ یا شکل آن باشد.

از این‌رو، با داشتن z، می‌توانیم یک توزیع پیشین $p(z)$ را روی متغیرهای نهفته برای محاسبه توزیع توام برروی متغیرهای مشاهده شده و نهفته معرفی کنیم:

$$p(x, z) = p(x|z)p(z)$$

این توزیع توام به ما این امکان را می دهد که توزیع پیچیده $p(x)$ را به روشی ساده‌تر حل کنیم. برای بدست آوردن توزیع داده $p(x)$ باید متغیرهای نهفته را به حاشیه ببریم:

$$p(x) = \int p(x, z)dz = \int p(x, z)p(z)dz \qquad (۸ - ۱)$$

علاوه بر این، با استفاده از قضیه بیز می‌توانیم توزیع پسین $p(z|x)$ را به صورت زیر محاسبه کنیم:

$$p(z|x) = \frac{p(x, z)p(z)}{p(x)} \qquad (۸ - ۲)$$

---

[1] explicitly

[2] implicit

توزیع پسین به ما اجازه می‌دهد تا متغیرهای پنهان را با توجه به مشاهدات استنباط کنیم.
با توجه به این ایده، اکنون اصطلاحات اساسی زیر را داریم:

- **توزیع پیشین** $p(z)$: رفتار متغیرهای نهفته را مدل می‌کند.
- **درست‌نمایی** $p(x|z)$: نحوه نگاشت متغیرهای نهفته به نقاط داده را مشخص می‌کند.
- **توزیع توام** $p(x,z) = p(x|z)p(z)$: ضرب درست‌نمایی و پیشین است و اساساً مدل ما را توصیف می‌کند.
- **توزیع حاشیه‌ای** $p(x)$: توزیع داده‌های اصلی است و هدف نهایی مدل است. توزیع حاشیه‌ای به ما می‌گوید که چقدر امکان تولید یک نقطه داده وجود دارد.
- **توزیع پسین** $p(z|x)$: متغیرهای پنهانی را توصیف می‌کند که می‌تواند توسط یک نقطه داده خاص تولید شود.

*توجه داشته باشید که ما از هیچ شکلی از برچسب استفاده نکردیم!!*

بر این اساس، دو اصطلاح دیگر را می‌توان تعریف کرد:

- **تولید:** به فرآیند محاسبه نقطه داده $x$ از متغیر نهفته $z$ اشاره دارد. اساسا، ما از فضای نهفته به توزیع واقعی داده‌ها حرکت می‌کنیم. از منظر ریاضی این با درست‌نمایی $p(x|z)$ نشان داده می‌شود.
- **استنباط:** فرآیند یافتن متغیر پنهان $z$ از نقطه داده $x$ است و با توزیع پسین $p(z|x)$ فرموله می‌شود. بدیهی است که استنباط معکوس تولید است و بالعکس. به صورت بصری می‌توانیم شکل زیر را در نظر بگیریم:

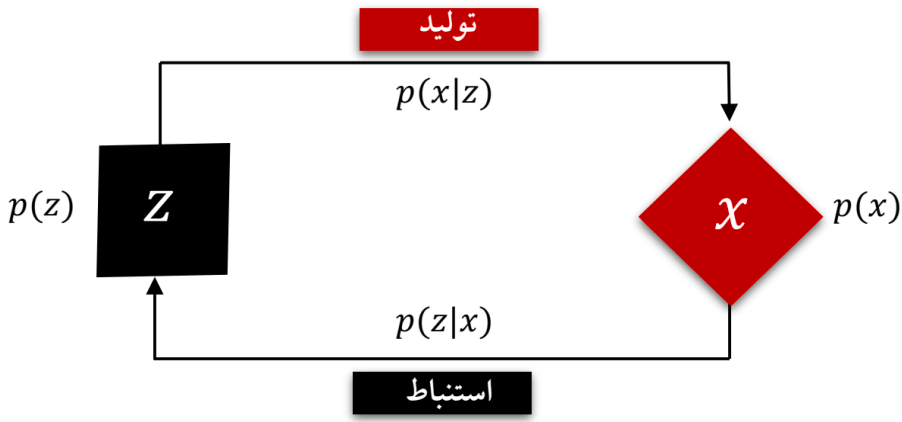

اینجا نقطه‌ای است که همه چیز با هم ترکیب می‌شود. اگر فرض کنیم که درست‌نمایی $p(x|z)$، پسین $p(z|x)$، حاشیه‌ای $p(x)$ و پیشین $p(z)$ را به نحوی می‌دانیم، می‌توانیم موارد زیر را انجام دهیم:

- **تولید:** برای تولید یک نقطه داده، می‌توانیم $z$ را از $p(z)$ و سپس نقطه داده $x$ را از $p(x|z)$ نمونه‌برداری کنیم:

$$z \sim p(z)$$
$$x \sim p(x|z)$$

- **استنباط:** از طرف دیگر، برای استنباط یک متغیر نهفته، $x$ را از $p(x)$ و سپس $z$ را از $p(z|x)$ نمونه‌برداری می‌کنیم:

$$x \sim p(x)$$
$$z \sim p(z|x)$$

مزیت مدل‌های دارای متغیرهای نهفته این است که می‌توانند فرآیندِ تولیدی که داده‌ها از آن ایجاد شده‌اند را بیان کنند. به طور کلی، به این معنی است که اگر می‌خواهیم یک نقطه داده جدید تولید کنیم، ابتدا باید یک نمونه $z \sim p(z)$ دریافت کنیم و سپس از آن برای نمونه‌برداری از یک مشاهده جدید $x$ از توزیع شرطی $p(x|z)$ استفاده کنیم. در حین انجام این کار، همچنین می‌توانیم ارزیابی کنیم که آیا مدل تقریب خوبی برای توزیع داده $p(x)$ ارائه می‌دهد یا خیر. شما می‌توانید متغیرهای نهفته را به عنوان گلوگاهی در نظر بگیرید که تمام اطلاعاتی که برای تولید داده‌ها لازم است باید از آن عبور کنند. ما از فرضیه منیفلد می‌دانیم که داده‌هایی که با ابعاد بالا روی منیفلدهای ابعادِ پایین‌تر تعبیه شده در این فضای با ابعاد بالا قرار دارند. این فضای نهفته، ابعاد پایین‌تر را توجیه می‌کند. به عبارت ساده‌تر، متغیرهای نهفته، تبدیل نقاط داده به یک فضای پیوسته با ابعاد پایین‌تر هستند. به طور شهودی، متغیرهای نهفته داده‌ها را به روشی ساده‌تر توصیف می‌کنند.

توجه داشته باشید که انتگرال در معادله (۸ـ۱) برای اکثر داده‌هایی که با آنها سر و کار داریم **راه‌حل تحلیلی ندارد**[1] و باید از روشی برای استنباط پسین در رابطه (۸ـ۲) استفاده کنیم.

## استنباط پسین

توزیع پسین $p(z|x)$، که یک جزء کلیدی در استدلال احتمالی است، باورهای ما را در مورد متغیرهای نهفته پس از مشاهده یک نقطه دادهٔ جدید بروز می‌کند. با این حال، پسین برای داده‌های دنیای واقعی اغلب غیرقابل حل است، چراکه هیچ راه‌حل تحلیلی برای انتگرال در معادله (۸ـ۱) که در مخرج معادله (۸ـ۲) ظاهر می‌شود، وجود ندارد. دو روش برای تقریب این توزیع وجود دارد. یک روش نمونه‌گیری به نام روش زنجیره مارکوف مونت‌کارلو است. با این حال، این روش‌ها از نظر محاسباتی گران هستند و به مجموعه داده‌های بزرگ مقیاس نمی‌شوند. روش دوم تکنیک‌های تقریب قطعی هستند. از جمله این تکنیک‌ها **استنباط**

---

[1] no analytical solution

**تغییرپذیر**[1] است که در **خودرمزنگار تغییرپذیر** استفاده می‌شود و این کار را با تبدیل آن به یک مساله بهینه‌سازی انجام می‌دهد. با این حال، توجه داشته باشید که اشکال این روش این است که نمی‌تواند نتایج دقیقی را در زمان محاسباتی بی‌نهایت ایجاد کند.

## خودرمزنگار تغییرپذیر

خودرمزنگار تغییرپذیر یک چارچوب کلی برای یادگیری مدل‌های متغیر نهفته با استنباط تغییرپذیر است. به طور کلی، خودرمزنگار تغییرپذیر، از شبکه‌های عصبی برای مدل مولد و همچنین مدل استنباط استفاده می‌کند.

فرض کنید که مشاهدات را به صورت متغیرهای تصادفی مستقل با توزیع یکسان (i.i.d) در نظر می‌گیریم و فرض می‌کنیم که داده‌ها توسط برخی از متغیرهای نهفته تصادفی $z$ تولید می‌شوند. از دیدگاه استنباط متغیر، ما $p(z)$ را رام‌شدنی فرض می‌کنیم. سپس سعی می‌کنیم مدل شواهد $p_\theta(x)$ را با تقریب توزیع نهفته‌ی برآوردشده، بیشینه کنیم:

$$\log p_\theta(x) = D_{KL}(q_\varphi(z|x)||p_\theta(z|x)) + \mathcal{L}(\theta, \varphi; x)$$

که در آن $D_{KL}$ واگرایی کولبک ـ لیبلر، $\varphi$ پارامترهای تغییرپذیر، $\theta$ پارامترهای مولد و $\mathcal{L}$ حد پایین تغییرپذیر یا حد پایین مشاهدات (ELBO) است. طبق قضیه بیز می‌توان آن را بازنویسی کرد:

$$\mathcal{L}(\theta, \varphi; x) = -D_{KL}(q_\varphi(z|x)||p_\theta(z)) + \mathbb{E}_{q_\varphi(z|x)}[log p_\theta(x|z)]$$

هنگامی که وزن شبکه عصبی را به عنوان پارامتر انتخاب می‌کنیم، ارتباط واضحی بین معادله بالا و خودرمزنگار تغییرپذیر وجود دارد. آیتم واگرایی KL مانند یک منظم‌ساز روی رمزگذار عمل می‌کند تا توزیع متغیر نهفته تخمین‌زده شده را به پیشین تخمین بزند، در حالی که آیتم دوم خطای بازتولید را تحت متغیر نهفته $z$ کمینه می‌کند. با این حال، نکته‌ای که وجود دارد این است که پس‌انتشار نمی‌تواند از لایه تصادفی عبور کند. چرا که مشتق‌گرفتن با توجه به پارامترهای متغیر $\varphi$ امکان‌پذیر نیست. به عبارت دیگر، گرادیان‌ها را نمی‌توان از طریق متغیر پنهان $z$ به عقب انتشار داد. این مشکل از آنجا ناشی می‌شود که پس‌انتشار از طریق گره‌های تصادفی قابل جریان نیست و پس‌انتشار برای تعیین این پارامترها، انتظار مقادیر قطعی را دارد. از این‌رو، یک تدبیر پارامترسازی مجدد برای حل این مشکل پیشنهاد شده است. به این معنی که متغیر نهفته $z$ را با یک تبدیل مشتق‌پذیر $g_\varphi(\epsilon, x)$ با یک متغیر نویز اضافی $\epsilon$ مجددا پارامتر کنیم. برای گاوسی، پارامترسازی مجدد را می‌توان به صورت بیان کرد:

$$z = \mu + \sigma\epsilon, \quad \epsilon \sim \mathcal{N}(0, I)$$

---

[1] variational inference

در نهایت، واگرایی KL را می‌توان به صورت تحلیلی محاسبه کرد و MSE اغلب به عنوان زیان بازسازی استفاده می‌شود. بنابراین تابع زیان خودرمزنگار تغییرپذیر (VAE) به‌صورت زیر بدست می‌آید:

$$\mathcal{L}_{VAE} = \frac{1}{\Upsilon} \sum_{i=1}^{M} \sum_{j=1}^{J} (1 + \log(\sigma_{i,j}^{\Upsilon}) - \mu_{i,j}^{\Upsilon} - \sigma_{i,j}^{\Upsilon}) + \sum_{i=1}^{M} \|x_i - \tilde{x}_i\|_{\Upsilon}^{\Upsilon}$$

که در آن $M$ تعداد نمونه داده‌ها، $j$ بعد رمزگذاری‌های نهفته، $x_i$ و $\tilde{x}_i$ به ترتیب $i$امین داده‌های اصلی و بازتولید هستند. شماتیک خودرمزنگار تغییرپذیر در شکل ۸ ـ ۲۲ نشان داده شده است[1].

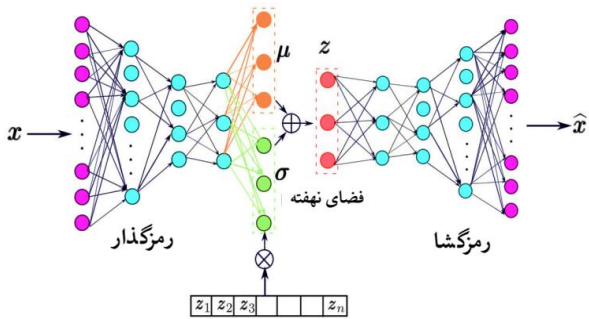

**شکل ۸ ـ ۲۲.** خودرمزنگار تغییرپذیر

## خودرمزنگار تغییرپذیر در پایتون

### وارد کردن کتاب‌خانه‌ها

```
In [1]:    import numpy as np
           import matplotlib.pyplot as plt
           import pandas as pd
           import seaborn as sns
           import warnings
           import tensorflow
           tensorflow.compat.v1.disable_eager_execution()
```

### وارد کردن مجموعه داده

```
In [2]:    from tensorflow.keras.datasets import mnist
           (trainX, trainy), (testX, testy) = mnist.load_data()
```

---

[1] برای اطلاعات بیشتر در خصوص خودرمزنگار تغییرپذیر می‌توانید به مرجع زیر مراجعه کنید:

**میلاد وزان، یادگیری عمیق: اصول، مفاهیم و رویکردها، ویرایش نخست، تهران، میعاد اندیشه، ۱۳۹۹**

```
In [2]:   for j in range(5):
              i = np.random.randint(0, 10000)
              plt.subplot(550 + 1 + j)
              plt.imshow(trainX[i], cmap='gray')
              plt.title(trainy[i])
          plt.show()
```

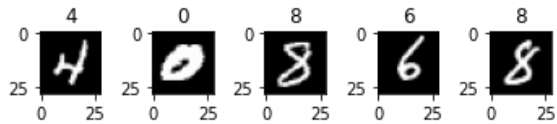

## آماده‌سازی داده‌ها

```
In [4]:   train_data = trainX.astype('float32')/255
          test_data = testX.astype('float32')/255

          train_data = np.reshape(train_data, (60000, 28, 28, 1))
          test_data = np.reshape(test_data, (10000, 28, 28, 1))

          print (train_data.shape, test_data.shape)

out [4]:  (60000, 28, 28, 1) (10000, 28, 28, 1)
```

## ساخت رمزگذار

در این قسمت رمزگذار مدل VAE خود را تعریف می‌کنیم.

```
In [5]:   import tensorflow

          input_data = tensorflow.keras.layers.Input(shape=(28, 28, 1))

          encoder = tensorflow.keras.layers.Conv2D(64, (5,5),
          activation='relu')(input_data)
          encoder = tensorflow.keras.layers.MaxPooling2D((2,2))(encoder)

          encoder = tensorflow.keras.layers.Conv2D(64, (3,3),
          activation='relu')(encoder)
          encoder = tensorflow.keras.layers.MaxPooling2D((2,2))(encoder)

          encoder = tensorflow.keras.layers.Conv2D(32, (3,3),
          activation='relu')(encoder)
          encoder = tensorflow.keras.layers.MaxPooling2D((2,2))(encoder)

          encoder = tensorflow.keras.layers.Flatten()(encoder)
          encoder = tensorflow.keras.layers.Dense(16)(encoder)
```

**توزیع و نمونه‌گیری نهفته**

این بخش وظیفه گرفتن ویژگی‌های پیچشی از آخرین بخش و محاسبه میانگین و واریانس ویژگی‌های نهفته را بر عهده دارد (همان‌طور که فرض کردیم ویژگی‌های نهفته از یک توزیع نرمال استاندارد پیروی می‌کنند و توزیع را می‌توان با میانگین و واریانس نشان داد).

```
In     def sample_latent_features(distribution):
[5]:       distribution_mean, distribution_variance = distribution
           batch_size = tensorflow.shape(distribution_variance)[0]
           random = tensorflow.keras.backend.random_normal(shape=(batch_size,
       tensorflow.shape(distribution_variance)[1]))
           return distribution_mean + tensorflow.exp(0.5 * distribution_variance) *
       random

       distribution_mean = tensorflow.keras.layers.Dense(2, name='mean')(encoder)

       distribution_variance = tensorflow.keras.layers.Dense(2,
       name='log_variance')(encoder)

       latent_encoding =
       tensorflow.keras.layers.Lambda(sample_latent_features)([distribution_mean,
       distribution_variance])
```

این ویژگی‌های پنهان (محاسبه شده از توزیع آموخته‌شده) در واقع قسمت رمزگذار مدل راکامل می‌کند. اکنون مدل رمزگذار را می‌توان به صورت زیر تعریف کرد:

```
In   [5]:    encoder_model = tensorflow.keras.Model(input_data,
             latent_encoding)
             encoder_model.summary()
```

```
Model: "model"

Layer (type)                     Output Shape          Param #    Connected to
==================================================================================
input_1 (InputLayer)             [(None, 28, 28, 1)]   0          []

conv2d (Conv2D)                  (None, 24, 24, 64)    1664       ['input_1[0][0]']

max_pooling2d (MaxPooling2D)     (None, 12, 12, 64)    0          ['conv2d[0][0]']

conv2d_1 (Conv2D)                (None, 10, 10, 64)    36928      ['max_pooling2d[0][0]']

max_pooling2d_1 (MaxPooling2D)   (None, 5, 5, 64)      0          ['conv2d_1[0][0]']

conv2d_2 (Conv2D)                (None, 3, 3, 32)      18464      ['max_pooling2d_1[0][0]']

max_pooling2d_2 (MaxPooling2D)   (None, 1, 1, 32)      0          ['conv2d_2[0][0]']

flatten (Flatten)                (None, 32)            0          ['max_pooling2d_2[0][0]']

dense (Dense)                    (None, 16)            528        ['flatten[0][0]']

mean (Dense)                     (None, 2)             34         ['dense[0][0]']

log_variance (Dense)             (None, 2)             34         ['dense[0][0]']

lambda (Lambda)                  (None, 2)             0          ['mean[0][0]',
                                                                  'log_variance[0][0]']
==================================================================================
Total params: 57,652
Trainable params: 57,652
Non-trainable params: 0
```

رمزگذار بسیار ساده است و فقط حدود ۵۷ هزار پارامتر قابل آموزش دارد.

## توزیع و نمونه‌گیری نهفته

بخش رمزگذار مدل یک تصویر را به عنوان ورودی می‌گیرد و بردار رمزگذاری نهفته را به عنوان خروجی می‌دهد که از توزیع آموخته‌شده مجموعه داده ورودی نمونه‌برداری می‌شود. وظیفه **رمزگشا** این است که این بردار تعبیه شده را به عنوان ورودی گرفته و تصویر اصلی (یا تصویری متعلق به کلاس مشابه تصویر اصلی) را دوباره ایجاد کند. از آنجایی که بردار نهفته یک نمایش کاملاً فشرده از ویژگی‌ها است، بخش رمزگشا از چندین جفت از لایه‌های Deconvolutional و لایه‌های upsampling تشکیل شده است. یک لایه Deconvolutional اساساً کاری را که یک لایه پیچشی انجام می‌دهد معکوس می‌کند. لایه های upsampling برای بازگرداندن وضوح اصلی تصویر استفاده می‌شود. به این ترتیب تصویر را با ابعاد اصلی بازسازی می‌کند.

```
In  [5]:   decoder_input = tensorflow.keras.layers.Input(shape=(2))
           decoder = tensorflow.keras.layers.Dense(64)(decoder_input)
           decoder = tensorflow.keras.layers.Reshape((1, 1, 64))(decoder)
           decoder = tensorflow.keras.layers.Conv2DTranspose(64, (3,3),
           activation='relu')(decoder)

           decoder = tensorflow.keras.layers.Conv2DTranspose(64, (3,3),
           activation='relu')(decoder)
           decoder = tensorflow.keras.layers.UpSampling2D((2,2))(decoder)

           decoder = tensorflow.keras.layers.Conv2DTranspose(64, (3,3),
           activation='relu')(decoder)
           decoder = tensorflow.keras.layers.UpSampling2D((2,2))(decoder)

           decoder_output = tensorflow.keras.layers.Conv2DTranspose(1, (5,5),
           activation='relu')(decoder)
```

مدل رمزگشا را می‌توان به صورت زیر تعریف کرد:

```
In  [5]:   decoder_model = tensorflow.keras.Model(decoder_input,
           decoder_output)
           decoder_model.summary()
```

```
Model: "model_1"

Layer (type)                  Output Shape          Param #
=================================================================
input_2 (InputLayer)          [(None, 2)]           0

dense_1 (Dense)               (None, 64)            192

reshape (Reshape)             (None, 1, 1, 64)      0

conv2d_transpose (Conv2DTra   (None, 3, 3, 64)      36928
nspose)

conv2d_transpose_1 (Conv2DT   (None, 5, 5, 64)      36928
ranspose)

up_sampling2d (UpSampling2D   (None, 10, 10, 64)    0
)

conv2d_transpose_2 (Conv2DT   (None, 12, 12, 64)    36928
ranspose)

up_sampling2d_1 (UpSampling   (None, 24, 24, 64)    0
2D)

conv2d_transpose_3 (Conv2DT   (None, 28, 28, 1)     1601
ranspose)

=================================================================
Total params: 112,577
Trainable params: 112,577
Non-trainable params: 0
```

## ساخت خودرمزنگار تغییرپذیر

در نهایت، خودرمزنگار متغییر را می‌توان با ترکیب بخش‌های رمزگذار و رمزگشا تعریف کرد.

```
In [5]:    encoded = encoder_model(input_data)
           decoded = decoder_model(encoded)

           autoencoder = tensorflow.keras.models.Model(input_data, decoded)

           autoencoder.summary()
```

```
Model: "model_2"

Layer (type)              Output Shape           Param #
=================================================================
input_1 (InputLayer)      [(None, 28, 28, 1)]    0

model (Functional)        (None, 2)              57652

model_1 (Functional)      (None, 28, 28, 1)      112577

=================================================================
Total params: 170,229
Trainable params: 170,229
Non-trainable params: 0
```

## تابع زیان

```
In [5]:    def get_loss(distribution_mean, distribution_variance):

               def get_reconstruction_loss(y_true, y_pred):
                   reconstruction_loss = tensorflow.keras.losses.mse(y_true, y_pred)
                   reconstruction_loss_batch = tensorflow.reduce_mean(reconstruction_loss)
                   return reconstruction_loss_batch*28*28

               def get_kl_loss(distribution_mean, distribution_variance):
                   kl_loss = 1 + distribution_variance -
               tensorflow.square(distribution_mean) -
               tensorflow.exp(distribution_variance)
                   kl_loss_batch = tensorflow.reduce_mean(kl_loss)
                   return kl_loss_batch*(-0.5)

               def total_loss(y_true, y_pred):
                   reconstruction_loss_batch = get_reconstruction_loss(y_true, y_pred)
                   kl_loss_batch = get_kl_loss(distribution_mean, distribution_variance)
                   return reconstruction_loss_batch + kl_loss_batch

               return total_loss
```

در نهایت، مدل برای آموزش آماده است:

```
In [5]:    autoencoder.compile(loss=get_loss(distribution_mean,
           distribution_variance), optimizer='adam')
```

## آموزش مدل

```
In [5]:    autoencoder.fit(train_data, train_data, epochs=20, batch_size=64,
           validation_data=(test_data, test_data))
```

```
Epoch 7/20
60000/60000 [==============================] - 21s 342us/sample - loss: 34.7669 - val_loss: 34.9932
Epoch 8/20
60000/60000 [==============================] - 20s 340us/sample - loss: 34.5290 - val_loss: 34.4761
Epoch 9/20
60000/60000 [==============================] - 21s 342us/sample - loss: 34.3419 - val_loss: 34.1162
Epoch 10/20
60000/60000 [==============================] - 21s 344us/sample - loss: 34.1046 - val_loss: 33.9851
Epoch 11/20
60000/60000 [==============================] - 21s 342us/sample - loss: 33.9861 - val_loss: 33.8626
Epoch 12/20
60000/60000 [==============================] - 20s 341us/sample - loss: 33.8211 - val_loss: 33.9228
Epoch 13/20
60000/60000 [==============================] - 21s 343us/sample - loss: 33.6959 - val_loss: 34.0816
Epoch 14/20
60000/60000 [==============================] - 21s 342us/sample - loss: 33.6299 - val_loss: 33.8098
Epoch 15/20
60000/60000 [==============================] - 20s 341us/sample - loss: 33.4760 - val_loss: 33.9145
Epoch 16/20
60000/60000 [==============================] - 20s 339us/sample - loss: 33.3355 - val_loss: 33.6437
Epoch 17/20
60000/60000 [==============================] - 21s 343us/sample - loss: 33.3159 - val_loss: 33.6046
Epoch 18/20
60000/60000 [==============================] - 21s 343us/sample - loss: 33.1844 - val_loss: 33.4338
Epoch 19/20
60000/60000 [==============================] - 20s 341us/sample - loss: 33.1549 - val_loss: 33.4451
Epoch 20/20
60000/60000 [==============================] - 20s 340us/sample - loss: 33.0717 - val_loss: 33.5015
```

**نتایج**

در این قسمت قابلیت‌های بازتولید مدل خود را بر روی تصاویر آزمون خواهیم دید. کد زیر ۹
تصویر را از مجموعه داده آزمایشی انتخاب می‌کند و ما تصاویر بازتولید مربوط را برای آن‌ها
ترسیم می‌کنیم.

```
In [5]:    offset=400
           print ("Real Test Images")
           # Real Images
           for i in range(9):
               plt.subplot(330 + 1 + i)
               plt.imshow(test_data[i+offset,:,:, -1], cmap='gray')
           plt.show()

           # Reconstructed Images
           print ("Reconstructed Images with Variational Autoencoder")
           for i in range(9):
               plt.subplot(330 + 1 + i)
               output = autoencoder.predict(np.array([test_data[i+offset]]))
               op_image = np.reshape(output[0]*255, (28, 28))
               plt.imshow(op_image, cmap='gray')
           plt.show()
```

Real Test Images

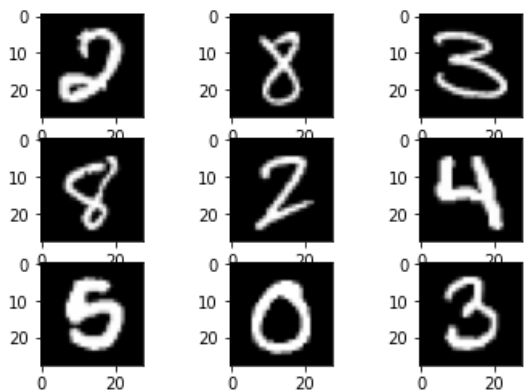

Reconstructed Images with Variational Autoencoder

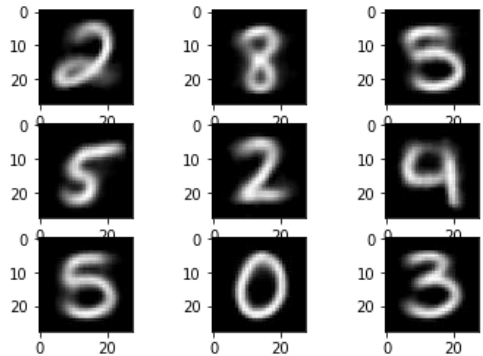

نتایج فوق نشان می‌دهد که مدل قادر به بازسازی تصاویر رقمی با کارایی مناسب است. با این حال، نکته مهمی که در اینجا باید به آن توجه کرد این است که برخی از تصاویر بازسازی شده از نظر ظاهری بسیار متفاوت از تصاویر اصلی هستند. این به این دلیل اتفاق می‌افتد که بازسازی فقط به تصویر ورودی وابسته نیست، بلکه از توزیعی است که آموخته شده است. جالب است، نه!

دومین چیزی که در اینجا باید به آن توجه کرد این است که تصاویر خروجی کمی تار هستند. این یک مورد رایج در خودرمزنگارهای تغییرپذیر است، آن‌ها اغلب خروجی‌های پر نویز (یا کیفیت پایین) تولید می‌کنند، زیرا بردارهای نهفته (گلوگاه) بسیار کوچک هستند. خودرمزنگارهای تغییرپذیر در واقع برای بازسازی تصاویر طراحی نشده‌اند، هدف واقعی یادگیری توزیع است که به آن‌ها قدرت فوق العاده‌ای می‌دهد تا داده‌های جعلی تولید کنند.

## خوشه‌های ویژگی‌های نهفته

همان‌طور که پیش‌تر بیان شد، خودرمزنگارهای تغییرپذیر، توزیع زیربنایی ویژگی‌های نهفته را یاد می‌گیرند، این اساسا به این معنی است که رمزگذاری‌هایِ نهفته‌یِ نمونه‌هایِ متعلق به یک کلاس، نباید در فضای نهفته خیلی از یکدیگر دور باشند.

بیایید جاسازی‌های نهفته را برای همه تصاویر آزمایشی خود ایجاد کنیم و آن‌ها را مصورسازی کنیم:

```
In [5]:   x = []
          y = []
          z = []
          for i in range(10000):
              z.append(testy[i])
              op = encoder_model.predict(np.array([test_data[i]]))
              x.append(op[0][0])
              y.append(op[0][1])
          df = pd.DataFrame()
          df['x'] = x
          df['y'] = y
          df['z'] = ["digit-"+str(k) for k in z]
```

```
plt.figure(figsize=(8, 6))
sns.scatterplot(x='x', y='y', hue='z', data=df)
plt.show()
```

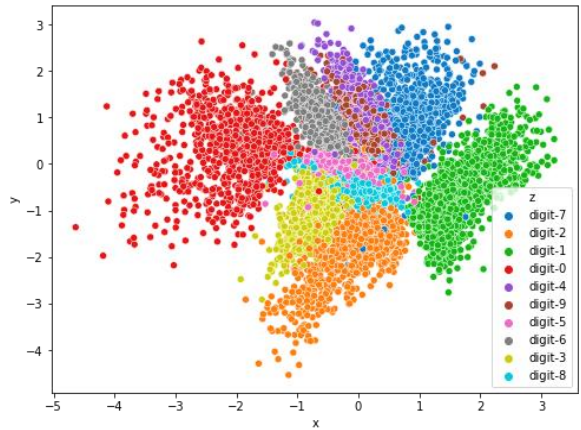

نمودار بالا نشان می‌دهد که جاسازی ارقام یکسان در فضای نهفته نزدیک‌تر است. این تقریبا چیزی است که می‌خواستیم از خودرمزنگارهای تغییرپذیر بدست آوریم. بیایید به قسمت پایانی برویم، جایی که قابلیت‌های تولیدی مدل خود را آزمایش می‌کنیم.

## تولید تصویر جعلی

بخش قبل نشان می‌دهد که رمزگذاری‌هایِ نهفته‌یِ داده‌هایِ ورودی، از توزیع نرمال استاندارد پیروی می‌کنند و مرزهای واضحی برای کلاس‌های مختلف ارقام قابل مشاهده است. از این‌رو، مدل می‌داند که کدام قسمت از فضا به چه کلاسی اختصاص دارد. این بدان معناست که ما در واقع می‌توانیم تصاویر رقمی با ویژگی‌های مشابه مجموعه داده‌های آموزشی را با عبور دادن نقاط تصادفی از فضا (فضای توزیع نهفته) تولید کنیم. در نتیجه، خودرمزنگارهای تغییرپذیر می‌توانند به عنوان مدل‌های مولد برای تولید داده‌های جعلی استفاده شوند.

بیایید دسته‌ای از ارقام با رمزگذاری‌های نهفته تولید کنیم.

```
In  [5]:    generator_model = decoder_model
            x_values = np.linspace(-3, 3, 30)
            y_values = np.linspace(-3, 3, 30)
            figure = np.zeros((28 * 30, 28 * 30))
            for ix, x in enumerate(x_values):
                for iy, y in enumerate(y_values):
                    latent_point = np.array([[x, y]])
                    generated_image = generator_model.predict(latent_point)[0]
                    figure[ix*28:(ix+1)*28, iy*28:(iy+1)*28,] =
            generated_image[:,:,-1]

            plt.figure(figsize=(15, 15))
            plt.imshow(figure, cmap='gray', extent=[3,-3,3,-3])
            plt.show()
```

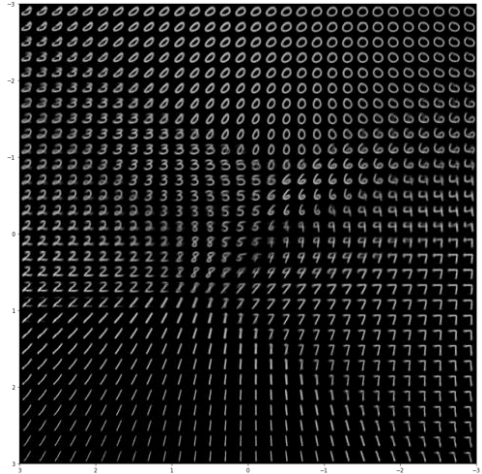

شما می‌توانید تمام ارقام (از ۰ تا ۹) را در ماتریس تصویر بالا بیابید زیرا ما سعی کرده‌ایم از تمام بخش‌های فضای نهفته تصاویر تولید کنیم.

# شبکه‌های متخاصم مولد[1] (GANs)

شبکه‌های متخاصم مولد (GANs) یکی از قدرتمندترین مدل‌های مولد محسوب می‌شوند. در معماری شبکه GAN دو مولفه وجود دارد، فضای بازنمایی G و D که باید در نظر گرفته شوند. این دو مولفه را می‌توان با دو شبکه عصبی مجزا ساخت. یکی تفکیک‌کننده D و دیگری مولد G است. آن‌ها مانند یک بازی دو نفره عمل می‌کنند تا یکدیگر را در طولِ بازیِ رقابتی بهبود بخشند. D سعی می‌کند تصاویر جعلی تولید شده توسط G را از تصاویر واقعی تشخیص دهد، در حالی که G سعی می‌کند تصاویر مشابه بیشتری با تصاویر واقعی تولید کند تا D را به اشتباه اندازد. در نهایت، G آموزش‌دیده، تصاویر واقعی تولید می‌کند.

ورودی مولد مقداری نویز تصادفی نمونه‌برداری شده از فضای نهفته است و خروجی آن یک تصویر $x$ است که قرار است تصاویر موجود در فضای داده اصلی را تخمین بزند. فرض کنید $z$ یک متغیر تصادفی نمونه‌برداری شده از $p_z(z)$ و $\theta_g$ پارامترهای مولد G باشد. بر این اساس، خروجی مولد را می‌توان به صورت $G(z; \theta_g)$ نشان داد. مولد G تمام تلاش خود را می‌کند تا تصویری واقعی تولید کند تا بتواند تفکیک‌کننده D را فریب دهد. در همین حال، تشخیص‌دهنده به عنوان یک دسته‌بند دودویی کار می‌کند که تصاویر اصلی ($x$) و خروجی‌های مولد $G(z; \theta_g)$ را به عنوان ورودی می‌گیرد. سپس، تفکیک‌کننده D با پارامترهای $\theta_d$ تلاش می‌کند تصاویر

---

[1] Generative adversarial networks

اصلی را از جعلی تشخیص دهد. هدف یک شبکه متخاصم مولد از منظر ریاضی به‌صورت زیر است:

$$\min_G max_D L(D, G) = \mathbb{E}_{x \sim p_r(x)} log D(x, \theta_d) + \mathbb{E}_{z \sim p(z)} \log\left(1 - D(G(z; \theta_z))\right)$$

که در معادله فوق $p_r(x)$ توزیع داده‌های واقعی و $p_g(x)$ توزیع داده‌های تولید شده توسط مولد است.

تفکیک‌کننده D می‌تواند تصاویر واقعی و جعلی را به راحتی در مراحل اولیه آموزش تشخیص دهد. از این‌رو، $\log\left(1 - D(G(z; \theta_z))\right)$ اشباع می‌شود. در عمل، می‌توانیم $\log G(z; \theta_z)$ را به جای کمینه کردن $\log\left(1 - D(G(z; \theta_z))\right)$ بیشینه کنیم. چارچوب شبکه‌های متخاصم مولد در شکل ۸ــ۲۳ نشان داده شده است.

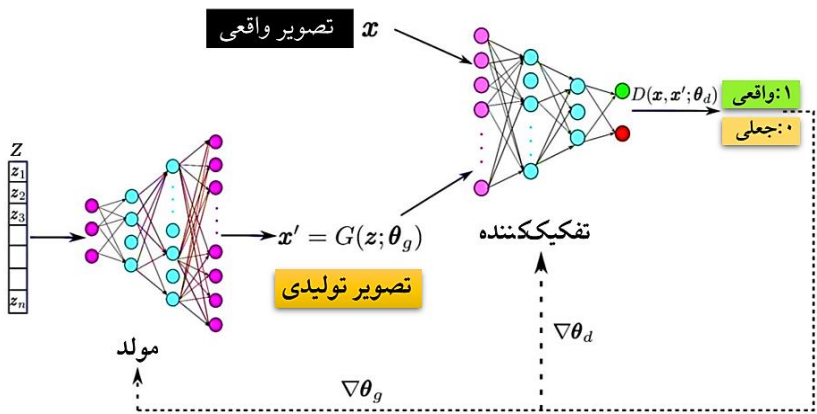

**شکل ۸ــ۲۳.** شبکه‌های متخاصم مولد

# خلاصه فصل

- برای مسائلی که الگوها ناشناخته هستند یا به‌طور دائم در حال تغییر هستند یا مجموعه داده‌های برچسب‌گذاری‌شده کافی برای آن‌ها نداریم، یادگیری غیرنظارتی می‌درخشد.
- یادگیری غیرنظارتی، مسائل حل نشدنی قبلی را قابل حل‌تر می‌کند و در یافتن الگوهای پنهان هم در داده‌های گذشته‌ی در دسترس برای آموزش و هم در داده‌های آینده، بسیار چابک‌تر عمل می‌کند.
- با یادگیری غیرنظارتی می‌توان به‌طور خودکار نمونه‌های بدون‌برچسب را برچسب‌گذاری کرد.
- خوشه‌ها مناطقی هستند که تراکم نقاط داده مشابه در آن‌ها زیاد است.

- خوشه‌بندی یا تحلیل خوشه‌ای وظیفه گروه‌بندی مجموعه‌ای از اشیاء است به گونه‌ای که اشیاء در یک گروه (به نام خوشه) نسبت به سایر گروه‌ها (خوشه‌ها) شباهت بیشتری به یکدیگر داشته باشند.

- خوشه‌بندی می‌تواند به تنهایی برای شناسایی ساختار ذاتی داده‌ها مورد استفاده قرار گیرد، در حالی که می‌تواند به عنوان یک تکنیک پیش‌پردازش برای سایر وظایف یادگیری مانند دسته‌بندی نیز عمل کند.

- هدف از خوشه‌بندی توصیفی و هدف از دسته‌بندی پیش‌بینی است.

- بسیاری از روش‌های خوشه‌بندی از معیارهای فاصله برای تعیین شباهت یا عدم شباهت بین هر جفت شی استفاده می‌کنند.

- خوشه‌بندی نمونه اولیه، خانواده‌ای از الگوریتم‌های خوشه‌بندی است که فرض می‌کند ساختار خوشه‌بندی را می‌توان با مجموعه‌ای از نمونه‌های اولیه نشان داد.

- الگوریتم $k-means$ یک استراتژی حریصانه اتخاذ می‌کند و یک روش بهینه‌سازی تکراری را برای یافتن راه حل تقریبی اتخاذ می‌کند.

- یکی از نقاط قوت خوشه‌بندی مدل مخلوط گاوسی این است که یک روش خوشه‌بندی نرم است.

- یک مدل مخلوط گاوسی به سادگی مدلی است که چندین توزیع گاوسی را به مجموعه‌ای از داده‌ها برازش می‌دهد.

- هر گاوسی در مدل مخلوط، نشان‌دهنده یک خوشه بالقوه است.

- خوشه‌بندی سلسله مراتبی رویکرد متفاوتی دارد و همانطور که از نامش پیداست، سلسله‌مراتب خوشه‌ها را به شکل درخت توسعه می‌دهد.

- رویکرد خوشه‌بندی مبتنی بر چگالی، روشی است که قادر به یافتن خوشه‌هایی با شکل دلخواه است و همانطور که از نام آن پیداست، از چگالی نمونه‌ها برای اختصاص عضویت در خوشه استفاده می‌کند.

- الگوریتم‌های خوشه‌بندی مبتنی‌بر چگالی، ارتباط بین نمونه‌ها را از منظر چگالی ارزیابی کرده و با افزودن نمونه‌های قابل ارتباط، خوشه‌ها را گسترش می‌دهند.

- برخلاف بسیاری دیگر از الگوریتم‌های خوشه‌بندی سنتی، الگوریتم‌های خوشه‌بندی مبتنی‌بر چگالی توانایی مقابله با موارد دورافتاده را دارند.

- $DBSCAN$ یک الگوریتم خوشه‌بندی مبتنی‌بر چگالی است که چگالی توزیع‌های نمونه را با یک جفت پارامتر "همسایگی" $(\varepsilon, \mu)$ مشخص می‌کند.

- زمانی که خوشه‌بندی را نه به‌عنوان یک مدل مستقل، بلکه به‌عنوان بخشی از استراتژی کشف داده‌های گسترده‌تر به کار می‌برید، بیشترین بهره را از خوشه‌بندی خواهید برد.

- اجتناب از بیش‌برازش، انگیزه اصلی برای انجام کاهش ابعاد است.

- تعداد ابعاد کمتر در داده‌ها به معنای زمان آموزش کمتر و منابع محاسباتی کمتر است.
- کاهش ابعاد برای مصورسازی داده‌ها بسیار مفید است.
- کاهش ابعاد نویز در داده‌ها را حذف می‌کند.
- به‌طور کلی، دو رویکرد برای کاهش ابعاد وجود دارد: انتخاب ویژگی و استخراج ویژگی
- رویکرد انتخاب ویژگی سعی می‌کند یک زیرمجموعه از ویژگی‌های مهم را انتخاب و ویژگی‌های نه‌چندان مهم را حذف کند.
- انتخاب ویژگی سعی می‌کند یک زیرفضای ویژگی جدید ایجاد کند.
- ایده اصلی پشت استخراج ویژگی فشرده‌سازی داده‌ها با هدف حفظ بیشتر اطلاعات مربوط است.
- خودرمزنگارها شبکه‌های عصبی هستند که برای کاهش ابعاد استفاده می‌شود.
- خودرمزنگارها از دو قطعه شبکه عصبی، رمزگذار و رمزگشا تشکیل شده‌اند.
- مدل‌های مولد (تولیدی) دسته‌ای از مدل‌های یادگیری ماشین هستند که برای توصیف نحوه تولید داده‌ها استفاده می‌شوند.
- هدف مدل مولد توصیف احتمالی کامل مجموعه داده است.

## مراجع برای مطالعه بیشتر

# ۹ مباحث منتخب

**اهداف:**

- یادگیری گروهی چیست و چه مزایایی دارد؟
- آشنایی با یادگیری همیشگی و ارتباط آن با سایر رویکردها
- یادگیری تقویتی چیست؟
- الگوریتم‌های یادگیری تقویتی
- یادگیری تقلیدی و تفاوت آن با یادگیری تقویتی

# یادگیری گروهی

در بسیاری از مواقعی که نیاز به تصمیم‌گیری مهم داریم، این طبیعت ما است که از نظر یک متخصص برای کمک به تصمیم‌گیری بهره ببریم. در بیشتر موارد، ما یک گام فراتر می‌رویم و به دنبال نظر دوم و سوم نیز هستیم. از این جهت که فکر می‌کنیم هیچ فردی به تنهایی نمی‌تواند در مورد موضوعی که از نظر عینی دشوار است (مثلا تشخیص پزشکی)، دانش کاملی داشته باشد. با در نظر گرفتن تعدادی از عوامل، ما سپس به طور شهودی، به نوعی توصیه‌های دریافت‌شده را برای اطلاع از تصمیم نهایی خود وزن کرده و ترکیب می‌کنیم. ما از همه‌ی این تصمیم‌ها استفاده می‌کنیم، چراکه انتظار داریم این تصمیم بهتر از تصمیمی باشد که خودمان می‌گیریم.

علی‌رغم مهارت ما در حل مساله، انسان‌ها هنوز از مشورت با منابع متعدد سود می‌برند. بنابراین، طبیعی است که حوزه یادگیری ماشین از این عادت انسان الهام گرفته باشد، زمانی که به دنبال راه‌هایی برای بهبود مدل‌های خود می‌گردند. حوزه‌ای از یادگیری ماشین که شامل رویه‌هایی است تا با آموزش مدل‌های متعدد و روش‌هایِ ترکیبِ خروجی‌های آنها به بهبودِ عملکردِ مدل دست یابد، به عنوان **یادگیری گروهی**[1] شناخته می‌شود. به عبارت ساده‌تر، یادگیری گروهی، یعنی هنر استفاده از چندین مدل، برای بدست آوردن عملکردِ پیش‌بینی بهتر.

اصل اساسی پشت مدل‌های گروهی، این است که گروهی از یادگیرنده‌های ضعیف گِرد هم آیند تا به کمک یکدیگر یک یادگیرنده قوی تشکیل دهند. شهودِ پشتِ مدل‌سازی گروهی، مترادف با چیزی است که ما در زندگی روزمره به آن عادت کرده‌ایم، مانند درخواست نظر از چندین متخصص قبل از اتخاذ یک تصمیم خاص به‌منظور به‌حداقل رساندن احتمال تصمیم بد یا نتیجه نامطلوب. یک سناریوی نمونه که در آن یک گروهِ سودمند تشکیل شده است در شکل ۹ ـ ۱ قابل مشاهده است. در این مثال، ما ۵ مدل را بر روی داده‌های آموزشی یکسان آموزش می‌دهیم هر مدل به کلاس تصویر رای می‌دهد و گروه، کلاسی که بیشترین رای را دارد پیش‌بینی می‌کند.

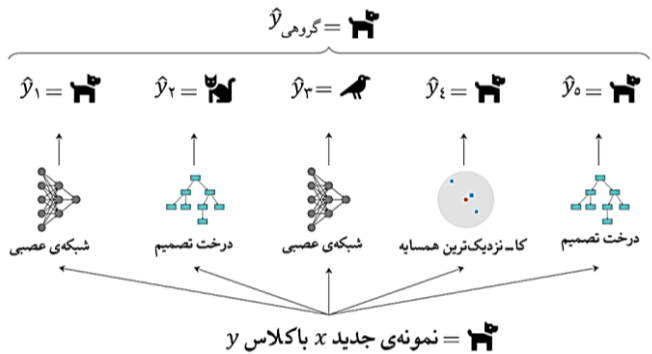

**شکل ۹ ـ ۱.** مثالی از یادگیری گروهی

---

۱

هیچ مدلِ یادگیری ماشین کامل نیست. همهٔ آنها محدودیت‌هایی دارند و اشتباه می‌کنند؛ همان‌طور که یک انسان ممکن است قبل از تصمیم‌گیری دشوار از چندین متخصص، مشاوره بخواهد. هم‌چنین، نشان داده شده که ترکیبِ پیش‌بینی‌های چندین مدل، پیش‌بینی‌هایِ بسیار دقیق‌تری را ارائه می‌دهد. بسیاری از مطالعات نظری و تجربی موقعیت‌هایی را شناسایی کرده‌اند که در آن سیستم‌هایِ گروهی سودمند هستند. به‌طور کلی، سه دلیل وجود دارد که چرا یادگیری گروهی می‌تواند نتایج بهتری نسبت به یک یادگیرندهٔ منفرد ایجاد کند:

۱. **داده‌هایِ آموزشیِ محدود:** برای مسائل پیچیده، داده‌ها ممکن است اطلاعات کافی را برای همهٔ ویژگی‌هایِ احتمالی داده‌هایی که یادگیرنده ممکن است با آنها مواجه شود، ارائه نکند. بنابراین، یک فرضیهٔ منحصر به فرد برای یک مجموعه دادهٔ محدود، بعید است به دلیل اطلاعات ناکافی برای مجموعه دادهٔ دیگری کار کند. در داده‌هایِ آموزشی با ترکیبِ فرضیه‌های یادگیرندگان، شانس بیشتری وجود دارد که یادگیرندهٔ گروهی با موفقیت داده‌هایی با ویژگی‌های ناشناخته را مدیریت کند.

۲. **جبرانِ فرآیندهایِ جستجویِ ناقص:** فرآیندهای جستجویِ الگوریتم‌هایِ یادگیری، ممکن است ناقص باشند. از این‌رو، حتی اگر یک فرضیهٔ بهینهٔ منحصر به فرد وجود داشته باشد، ممکن است هرگز پیدا نشود. یادگیری گروهی می‌تواند به جبران چنین فرآیندهایِ جستجوی ناقصی کمک کند.

۳. **هیچ فرضیهٔ بهینهٔ منحصر به فردی وجود ندارد:** فضای موردِ جستجو برای مجموعه دادهٔ معین، ممکن است حاویِ یک فرضیهٔ بهینهٔ واحد نباشد. از این‌رو یادگیری گروهی می‌تواند تقریب‌های خوبی را با ترکیب فرضیه‌های متعدد ارائه دهد.

## تا یادگیری عمیق هست، چرا یادگیری گروهی؟!

پیشرفت‌های محاسباتی و روش‌شناختی اخیر در آنچه به عنوان یادگیریِ عمیق شناخته می‌شود، سودمندیِ درک‌شده از روش‌های یادگیریِ گروهی را تغییر داده است. با یک معماریِ کافی و تنظیم ابرپارمترها، شبکه‌های عصبیِ عمیق می‌توانند عملکردی نزدیک به کامل در طیف وسیعی از مجموعه داده‌ها داشته باشند. علاوه بر این، بسیاری از تدبیرهای منظم‌سازی برای اطمینان از همگراییِ این مدل‌هایِ پیچیده بدون بیش‌برازش در دسترس هستند.

از این‌رو، با این عملکرد فوق‌العادهٔ یادگیریِ عمیق، آیا روش‌هایِ گروهی هنوز نقشِ مفیدی دارند؟! اول این‌که، آموزش شبکه‌هایِ عصبی عمیق، هزینهٔ محاسباتی زیادی دارد. تنظیم ابرپارمترهایِ جامع برای این مدل‌ها به دلیل کمبود منابع محاسباتی به ندرت امکان‌پذیر است. با این حال، یافتن معماریِ مناسب برای تولید مدل‌هایی با عملکرد پیش‌بینی بالا ضروری است. برخلاف این، یک سیستم گروهی می‌تواند عملکرد پیش‌بینی مشابهی با چنین مدل‌های منفرد

ایجاد کند، اما بدون متحمل شدن هزینه‌های محاسباتی مشابه. ممکن است در نگاه اول غیر شهودی به نظر برسد، چراکه چندین مدل برای تولید یک گروه نیز، نیاز به آموزش دارند! با این حال، مدل‌های فردی می‌توانند پیچیدگی کم‌تری باشند (لایه‌های کمتر) که امکان کاوش بهتر در فضای دسته‌بند را در همان بودجه محاسباتی فراهم می‌کند. علاوه بر این، بسیاری از این تنظیمات را می‌توان در سیستم نهایی گروهی (به جای انتخاب بهترین مدل) در نظر گرفت.

دوم اینکه، روش‌های منظم‌سازی به‌صورت کامل از بیش‌برازش جلوگیری نمی‌کنند. بنابراین، ترکیب مدل‌هایی که همگرا نشده‌اند، یا از مشکل بیش‌برازش داده‌ها رنج می‌برند، می‌توانند مدل‌های بهتری را بدون آموزش مجدد تولید کنند. در نهایت، هنگام کاوش مجموعه داده‌های دنیای واقعی، در بسیاری از مسابقات دیده می‌شود که سیستم‌های گروهی بهترین نتایج را ایجاد می‌کنند.

در حالی که شبکه‌های عصبی عمیق در برخی وظایف به عملکردی، نزدیک به انسان دست یافته‌اند، واقعیت این است که این مدل‌ها بسیار تخصصی هستند. از این‌رو، استفاده از یک شبکه‌ی عصبی که بتواند در هنگام تشخیص بین تصاویر گربه‌ها و سگ‌ها به دقت کامل برسد و آن را در یک مساله متفاوت اما مرتبط (مثلا شناسایی کلاس‌های مشابه بر اساس نمونه‌های ویدیویی) بکار برد، عملکرد بسیار ضعیفی را در پی خواهد داشت. این به یک موضوع بسیار مورد بحث معروف، به نام قضیه ناهار مجانی نیست (که پیش‌تر درباره‌ی آن بحث شد) مربوط می‌شود: **هیچ مدل یادگیری ماشین به تنهایی، در همه‌ی انواع داده‌ها و مجموعه داده‌ها بهترین عملکرد را نخواهد داشت.** این نشان می‌دهد که مجموعه‌ای از مدل‌های بسیار تخصصی، راه‌حلِ بهتری برای این موضوع هستند.

## تکنیک‌های یادگیری گروهی

یادگیری گروهی یک الگوی یادگیری ماشین است که در آن چندین مدل که اغلب "**یادگیرنده‌ی ضعیف**" نامیده می‌شوند برای حل یک مساله آموزش داده می‌شوند و برای بدست آوردن نتایج بهتر ترکیب می‌شوند. فرضیه‌ی اصلی این است که وقتی مدل‌های ضعیف بدرستی ترکیب شوند، می‌توانیم مدل‌های دقیق‌تری/قوی‌تری بدست آوریم.

در نظریه‌ی یادگیری گروهی، ما **یادگیرندگان ضعیف** (یا مدل‌های پایه) را می‌نامیم که می‌توانند با ترکیب چندین مدل از آن‌ها به عنوان بلوک‌هایی برای طراحی مدل‌های پیچیده‌تر استفاده شوند. اغلب اوقات، این مدل‌های پایه به‌خودی خود چندان خوب عمل نمی‌کنند به این دلیل که بایاس یا واریانس بالایی دارند. از این‌رو، ایده روش‌های گروهی این است که سعی کنیم بایاس یا واریانس چنین یادگیرندگان ضعیفی را با ترکیب چند مورد از آن‌ها، کاهش دهیم تا یک **یادگیرنده‌ی قوی** (یا مدل گروهی) ایجاد کنیم که به عملکرد بهتری دست یابد.

برای ایجاد یک مدل یادگیری گروهی، ابتدا باید مدل‌های پایه خود را برای تجمع انتخاب کنیم. بیشتر اوقات از یک الگوریتم یادگیری پایه استفاده می‌شود تا یادگیرندگان ضعیف همگنی داشته باشیم که به روش‌های مختلف آموزش دیده‌اند. به این نوع مدل‌ها "**همگن**" گفته می‌شود. با این حال، روش‌هایی نیز وجود دارد که از انواع مختلف الگوریتم‌های یادگیری پایه استفاده می‌کنند. سپس برخی از یادگیرندگان ضعیف ناهمگن در یک "مدل مجموعه‌های ناهمگن" ترکیب می‌شوند.

یک نکته مهم این است که انتخاب یادگیرندگان ضعیف ما، باید با روش تجمع این مدل‌ها هماهنگ باشد. اگر مدل‌های پایه با بایاس کم اما واریانس بالا را انتخاب کنیم، باید با روش تجمیع‌کننده‌ای[1] باشد که تمایل به کاهش واریانس دارد، در حالی که اگر مدل‌های پایه با واریانس کم اما بایاس بالا را انتخاب کنیم، باید با روش تجمعی باشد که تمایل به کاهش بایاس دارد.

تکنیک‌های گروهی زیادی برای ترکیب چندین یادگیرندهٔ ماشین در راستای ایجاد یک مدل پیش‌گویانه وجود دارد. رایج‌ترین تکنیک‌ها عبارتند از: **تجمیع‌پردازی**[2] برای کاهش واریانس و **توان‌افزایی**[3] برای کاهش بایاس.

## توان‌افزایی (بوستینگ)

توان‌افزایی، دسته‌ای از الگوریتم‌ها است که یادگیرندگان ضعیف را به یادگیرندگان قوی تبدیل می‌کند. الگوریتم توان‌افزایی با آموزش یک یادگیرندهٔ پایه شروع می‌شود و سپس توزیع نمونه‌های آموزشی را با توجه به نتیجهٔ یادگیرندهٔ پایه تنظیم می‌کند تا نمونه‌های دسته‌بندی نادرست مورد توجه بیشتر یادگیرندگان پایهٔ بعدی قرار گیرد. پس از آموزش اولین یادگیرندهٔ پایه، یادگیرندهٔ پایه دوم با نمونه‌های آموزشی تنظیم‌شده آموزش می‌بیند و از نتیجه برای تنظیم مجدد توزیع نمونه آموزشی استفاده می‌شود. چنین فرآیندی تکرار می‌شود تا زمانی که تعداد یادگیرندگان پایه به مقدار از پیش تعریف شده T برسد و در نهایت این یادگیرندگان پایه وزن و ترکیب شوند.

شناخته‌شده‌ترین الگوریتم توان‌افزایی **AdaBoost** است که الگوریتم توان‌افزایی ساده را از طریق یک فرآیند تکراری بهبود می‌بخشد. ایدهٔ اصلی پشت این الگوریتم تمرکز بیشتر برروی الگوهایی است که دسته‌بندی آن‌ها سخت‌تر است. مقدار تمرکز با وزنی که به هر الگوی مجموعه آموزشی اختصاص‌داده می‌شود، تعیین می‌شود. در ابتدا، وزن یکسانی به همهٔ الگوها اختصاص داده می‌شود. در هر تکرار وزن تمام نمونه‌های دسته‌بندی اشتباه افزایش می‌یابد در حالی که وزن

---

نمونه‌های دسته‌بندی صحیح کاهش می‌یابد. در نتیجه، یادگیرنده‌ی ضعیف مجبور می‌شود با انجام تکرارهای اضافی و ایجاد دسته‌بندهای بیشتر، برروی نمونه‌های دشوار مجموعه آموزشی تمرکز کند. علاوه بر این، وزنی به هر دسته‌بند اختصاص داده می‌شود. این وزن دقت کلی دسته‌بند را اندازه‌گیری می‌کند و تابعی از وزن کل الگوهای دسته‌بندی صحیح است. بنابراین، وزن‌های بالاتری به دسته‌بندهای دقیق‌تر داده می‌شود. از این وزن‌ها برای دسته‌بندی الگوهای جدید استفاده می‌شود.

*از منظر تجزیه و تحلیل بایاس ـ واریانس، توان‌افزایی عمدتا بر کاهش بایاس تمرکز دارد.* به همین دلیل است که مجموعه‌ای از یادگیرندگان با توانایی تعمیم ضعیف می‌توانند بسیار قدرتمند باشند.

## تجمیع‌پردازی (بگینگ)

تجمیع‌پردازی یک روش ساده و در عین حال موثر برای ایجاد مجموعه‌ای از دسته‌بندها است. دسته‌بند گروهی که با این روش ایجاد می‌شود، خروجی‌های دسته‌بندهای مختلف را در یک دسته‌بند واحد ادغام می‌کند. این منجر به یک دسته‌بند می‌شود که دقت آن بالاتر از دقت هر دسته‌بند منفردی است.

ایده‌ی پشت تجمیع‌پردازی، ترکیب نتایج چندین مدل برای بدست آوردن یک نتیجه کلی است. در اینجا یک سوال وجود دارد: اگر همه‌ی مدل‌ها را برروی یک مجموعه داده ایجاد کنیم و آن‌ها را ترکیب کنید، مفید خواهد بود؟ احتمال زیادی وجود دارد که این مدل‌ها نتیجه یکسانی داشته باشند، زیرا ورودی یکسانی دارند. پس چگونه می‌توانیم این مشکل را حل کنیم؟ یکی از این راه‌ها ایجاد یادگیرندگان پایه‌ی مختلف، با تقسیم مجموعه‌ی آموزشی اصلی به چندین زیرمجموعه غیرهمپوشان و استفاده از هر زیرمجموعه برای آموزش یک یادگیرنده‌ی پایه است. از آنجایی که زیرمجموعه‌های آموزشی متفاوت هستند، احتمالا یادگیرندگان پایه‌ی آموزش‌دیده نیز متفاوت هستند. با این حال، اگر زیرمجموعه‌ها کاملا متفاوت باشند، به این معنی است که هر زیرمجموعه فقط بخش کوچکی از مجموعه آموزشی اصلی را در برمی‌گیرد که احتمالا منجر به یادگیری ضعیفی می‌شود. از آنجایی که یک گروه خوب مستلزم آن است که هر یادگیرنده‌ی پایه به‌طور معقولی خوب باشد، ما اغلب اجازه می‌دهیم زیرمجموعه‌ها به گونه‌ای همپوشانی داشته باشند که هر یک از آن‌ها شامل نمونه‌های کافی باشد.

تجمیع‌پردازی بر اساس نمونه‌برداری **بوت‌استرپ** کار می‌کند. با توجه به مجموعه داده با $m$ نمونه، به طور تصادفی یک نمونه را انتخاب کرده و در مجموعه نمونه‌برداری رونوشت می‌کند. سپس، آن را در مجموعه داده‌های اصلی نگه می‌داریم به‌طوری که هنوز فرصتی برای برداشتن آن در دفعه بعد وجود داشته باشد. با تکرار این فرآیند $m$ مرتبه‌ی این مجموعه داده حاوی $m$ نمونه بدست می‌آید که در آن برخی از نمونه‌های اصلی ممکن است بیش از یک بار ظاهر شوند در

حالی که برخی ممکن است هرگز ظاهر نشوند. از فصل ۵ می‌دانیم که تقریباً ٪۶۳٫۲ از نمونه‌های اصلی در مجموعه داده‌ها ظاهر می‌شوند.

اعمال فرآیند بالا در $T$ مرتبه، $T$ مجموعه داده را تولید می‌کند که هر کدام حاوی $m$ نمونه است. سپس، یادگیرندگانِ پایه برروی این مجموعه داده‌ها آموزش دیده و ترکیب می‌شوند. چنین رویه‌ای، روند کار اولیه‌ی تجمیع‌پردازی است. هنگام ترکیب پیش‌بینی‌های یادگیرندگان پایه، تجمیع‌پردازی روش رای‌گیری ساده را برای وظایف دسته‌بندی و روش میانگین‌گیری ساده را برای وظایف رگرسیونی اتخاذ می‌کند. هنگامی که چندین کلاس تعداد آرای یکسانی را دریافت می‌کنند، می‌توانیم به طور تصادفی یکی را انتخاب کنیم یا اطمینان آرا را بیشتر بررسی کنیم.

نمونه‌گیری بوت‌استرپ مزیت دیگری را برای تجمیع‌پردازی به ارمغان می‌آورد: از آنجایی که هر یادگیرنده‌ی پایه فقط از ٪۶۳٫۲ از نمونه‌های آموزشی اصلی برای آموزش استفاده می‌کند، ٪۳۶٫۸ نمونه‌های باقی‌مانده (خارج از کیسه) را می‌توان به عنوان مجموعه اعتبارسنجی برای بدست آوردن توانایی تعمیم استفاده کرد. برای به دست آوردن این تخمین، باید نمونه‌های آموزشی مورد استفاده هر یادگیرنده‌ی پایه را ردیابی کنیم. فرض کنید $D_t$ مجموعه‌ای از نمونه‌های استفاده شده توسط یادگیرنده $h_t$ را نشان دهد و $H^{oob}(x)$ نشان‌دهنده‌ی پیش‌بینی برای نمونه $x$ دیده‌نشده باشد، یعنی فقط پیش‌بینی‌های انجام شده توسط یادگیرندگان پایه‌ای را در نظر بگیریم که از نمونه $x$ برای آموزش استفاده نکرده‌اند. از این‌رو، داریم:

$$H^{oob}(x) = arg_{y \in Y} \, max \sum_{t=1}^{T} \mathbb{I}(h_t(x) = y) . \mathbb{I}(x \notin D_t)$$

و برآورد نمونه‌های دیده‌نشده، خطای تعمیم تجمیع‌پردازی است:

$$\epsilon^{oob}(x) = \frac{1}{|D|} \sum_{(x,y) \in D} \mathbb{I}(H^{oob}(x) \neq y)$$

فرض کنید که پیچیدگیِ محاسباتیِ یک یادگیرنده‌ی پایه $O(m)$ است، پس پیچیدگیِ تجمیع‌پردازی تقریبا $T(O(m) + O(s))$ است، که در آن $O(s)$ پیچیدگی رای‌گیری یا میانگین‌گیری است. از آنجایی که پیچیدگی $O(s)$ کم است و $T$ ثابتی است که اغلب خیلی بزرگ نیست، تجمیع‌پردازی دارای همان پیچیدگی یادگیرنده پایه است، یعنی تجمیع‌پردازی یک الگوریتم یادگیری گروهی کارآمد است.

از منظر تجزیه و تحلیل بایاس‌ـ‌واریانس، تجمیع‌پردازی به کاهش واریانس کمک می‌کند و این به ویژه برای درختان تصمیم هرس‌نشده مفید است.

## تفاوت تجمیع‌پردازی با توان‌افزایی

تجمیع‌پردازی، همانند توان‌افزایی، تکنیکی است که دقت دسته‌بند را با تولید یک مدل ترکیبی که چندین دسته‌بند را ترکیب می‌کند، بهبود می‌بخشد. هر دو روش از یک رویکرد رای‌گیری

پیروی می‌کنند که به طور متفاوتی اجرا می‌شود تا خروجی‌های دسته‌بندهای مختلف را ترکیب کند. در توان‌افزایی، برخلاف تجمیع‌پردازی، هر دسته‌بند تحت تاثیر عملکرد دسته‌بندی‌هایی است که قبل از ساخت آن، ساخته شده‌اند. به طور خاص، دسته‌بند جدید به خطاهای دسته‌بندی که توسط دسته‌بندهای قبلی ساخته شده‌اند، توجه بیشتری می‌کند، جایی که میزان توجه با توجه به عملکرد آن‌ها تعیین می‌شود. در تجمیع‌پردازی، هر نمونه با احتمال مساوی انتخاب می‌شود، در حالی که در توان‌افزایی، نمونه‌ها با احتمال متناسب با وزن آن‌ها انتخاب می‌شوند.

## یادگیری همیشگی[1] (دیرپای)

با در دسترس قرار گرفتن مجموعه داده‌های بزرگ‌تر و کاهش هزینه‌های محاسباتی، مدل‌هایی که قادر به حل وظایف بزرگ‌تر هستند نیز در دسترس شدند. با این حال، آموزش یک مدل هر بار که نیاز به یادگیری یک کار جدید است، ممکن است غیر ممکن باشد. چراکه ممکن است داده‌های قدیمی‌تر در دسترس نباشند، داده‌های جدید به دلیل مشکلات حفظ حریم خصوصی نتوانند ذخیره شوند یا دفعاتی که سیستم باید در آن بروزرسانی شود، نمی‌تواند از آموزش یک مدل جدید با تمام داده‌ها به اندازه کافی مکرر پشتیبانی کند. راه حلی برای این مشکلات را می‌توان در یادگیری همیشگی یافت. هدف از یادگیری همیشگی این است که بتوانیم کارهای جدید را بدون نیاز به دسترسی به داده‌های مربوط به کارهایی که قبلا آموخته‌ایم یاد بگیریم. وقتی شبکه‌های عصبی وظایف جدیدی را یاد می‌گیرند، در صورت عدم استفاده از معیارهای خاص، دانش جدید بر دانش قدیمی‌تر اولویت داده می‌شود و معمولا باعث فراموشی دانش دوم می‌شود. این موضوع معمولا به عنوان **فراموشی فاجعه‌آمیز**[2] شناخته می‌شود (شکل ۹ـ۲ را ببینید). فراموشی فاجعه‌آمیز زمانی اتفاق می‌افتد که یک شبکه عصبی آموزش‌دیده، قادر به حفظ توانایی خود برای انجام وظایفی که قبلا آموخته است، زمانی که برای انجام وظایف جدید تطبیق داده می‌شود، نیست.

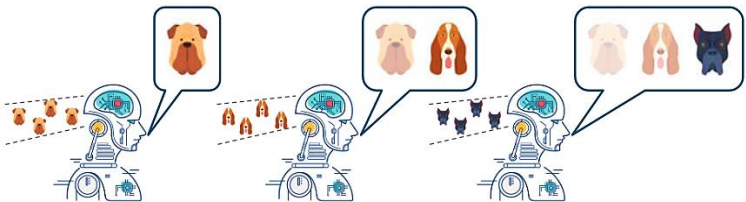

**شکل ۹ـ۲. تصویری از فراموشی فاجعه‌آمیز.** دانشِ آموخته شدهِ قبلی هنگام یادگیری کلاس‌های جدیدی که برای مدتی دیده نشده‌اند، فراموش می‌شوند (به صورت تدریجی محو می‌شوند).

---

[1] lifelong

[2] catastrophic forgetting

یادگیری همیشگی، یک الگوی یادگیری ماشین پیشرفته است که به طور مداوم یاد می‌گیرد، دانش‌آموخته شده در کارهای قبلی را جمع‌آوری می‌کند و از آن برای کمک به یادگیری آینده استفاده می‌کند. در این فرآیند، یادگیرنده، بیشتر و بیشتر در یادگیری آگاه و مؤثر می‌شود. توانایی یادگیری یکی از شاخصه‌های هوش انسان است. با این حال، الگوی یادگیری ماشین غالب فعلی به صورت مجزا یاد می‌گیرد: با توجه به مجموعه داده‌های آموزشی، الگوریتم یادگیری ماشین را روی مجموعه داده اجرا می‌کند تا یک مدل تولید کند. هیچ تلاشی برای حفظ دانش آموخته شده و استفاده از آن در یادگیری آینده انجام نمی‌دهد. اگرچه این الگوی یادگیری مجزا بسیار موفقیت‌آمیز بوده است، اما به تعداد زیادی نمونه آموزشی نیاز دارد و فقط برای کارهایی که به خوبی تعریف شده و محدود هستند مناسب است.

در مقایسه، ما انسان‌ها می‌توانیم با چند مثال، به‌طور مؤثر یاد بگیریم. چراکه در گذشته دانش زیادی جمع‌آوری کرده‌ایم که به ما امکان می‌دهد با داده‌ها یا تلاش اندک یاد بگیریم. هدف یادگیری همیشگی دستیابی به این قابلیت است. برنامه‌هایی مانند دستیاران هوشمند، بات‌های مکالمه و روبات‌های فیزیکی که با انسان‌ها و سیستم‌ها در محیط‌های واقعی تعامل دارند نیز نیازمند چنین قابلیت‌های یادگیری همیشگی هستند. بدون توانایی جمع‌آوری دانش آموخته‌شده و استفاده از آن برای یادگیری تدریجی دانش، احتمالا یک سیستم هرگز واقعا هوشمند نخواهد بود.

## تعریف یادگیری همیشگی

یک یادگیرنده که به‌روی یک دنباله از وظایف، از ۱ تا $N-1$، یادگیری را انجام داده است. هنگامی که با تکلیف $N$ روبرو می‌شود، از دانش بدست آمده در کارهای $1-N$ گذشته برای کمک به یادگیری برای تکلیف $N$ استفاده می‌کند.

برای تأکید بر زمینه همیشگی، وظایف ۱ تا N-۱ را وظایف/دامنه‌های گذشته و N-امین را وظیفه فعلی/دامنه می‌نامیم.

چندین سوال و چالش در طراحی یک سیستم یادگیری ماشین همیشگی [1] (LML) وجود دارد:

- چه اطلاعاتی باید از وظایف یادگیری گذشته حفظ شود؟
- چه اَشکالی از دانش برای کمک به یادگیری آینده استفاده خواهد شد؟
- سیستم چگونه دانش را بدست می‌آورد؟
- چگونه سیستم از دانش برای کمک به یادگیری آینده استفاده می‌کند؟

برای پاسخ به سوالات ذکر شده در بالا، یک سیستم LML به چهار جزء کلی زیر نیاز دارد:

---

[1] Lifelong Machine Learning

- **ذخیرهی اطلاعات گذشته[1] (PIS):** اطلاعات حاصل از یادگیریِ گذشته را ذخیره میکند. این ممکن است شامل ذخیرههای فرعی برای اطلاعاتی مانند (۱) دادههای اصلی مورد استفاده در هر کار گذشته، (۲) نتایج میانی از یادگیری هر کار گذشته و (۳) مدل نهایی یا الگوهای آموختهشده از وظایفِ گذشته باشد.

- **پایگاه دانش[2] (KB):** دانشِ استخراجشده یا ادغامشده از PIS را ذخیره میکند. این نیاز به یک طرح بازنماییِ دانش مناسب برای کاربرد دارد. مقیاسپذیریِ پایگاهِ دانش نیز در مورد کلانداده ضروری است.

- **استخراجکنندهی دانش[3] (KM):** دانش را از PIS استخراج میکند. این کاویدن را میتوان به عنوان یک فرآیند **فرایادگیری** در نظر گرفت. چراکه دانش را از اطلاعات حاصل از یادگیری وظایف گذشته میآموزد. دانش به KB موجود (دانش پایه) اضافه میشود.

- **یادگیرنده مبتنی بر دانش[4] (KBL):** با توجه به دانشِ موجود در KB، این یادگیرنده میتواند از دانش و یا برخی اطلاعات در PIS برای کار جدید استفاده کند.

از دیدگاهِ دیگری عناصر زیر برای یک عامل LML ضروری هستند:

**(۱) حفظِ دانشِ وظیفهی آموخته شده.**
**(۲) انتقالِ انتخابی یا استفاده از دانش قبلی هنگام حل وظایف جدید.**
**(۳) یک رویکرد سیستمی که تعامل موثر و کارآمد عناصر حفظ و انتقال فوق را تضمین میکند.**

در این زمینه، صحبت در مورد حفظِ دانش، با دیدگاهِ بازنماییِ دانش انجام میشود. هر دانش آموختهشده را میتوان به اشکال مختلف نشان داد. سادهترین شکل میتواند به سادگیِ ذخیرهی نمونههایِ آموزشی باشد. ذخیرهسازی دادههایِ آموزشیِ خام از مزیت دقت و خلوصِ دانش (حفظِ دانش) برخوردار است. با این حال، به دلیل حجم زیادی از ذخیرهسازی که نیاز دارد، ناکارآمد است. متناوبا، نمایشی از یک فرضیهی دقیق ایجاد شده از نمونههای آموزشی را میتوان ذخیره کرد. مزایای دانش بازنمایی، اندازهی کوچکِ آن در مقایسه با فضای مورد نیاز برای دادههای آموزشیِ اصلی و تواناییِ آن برای تعمیمِ فراتر از نمونههای آموزشی است.

صحبت در مورد انتقال دانش از دیدگاه یادگیریِ ماشین انجام میشود. انتقالِ بازنمایی، شامل تخصیص یک بازنماییِ وظیفهی شناخته شده، به یک سیستم یادگیری با یک وظیفه هدف جدید است. با انجام این کار، مدل جدید در یک حوزهی خاص از فضای خاص فرضیهیِ سیستم، مقداردهی میشود. انتقالِ بازنمایی، اغلب، زمانِ آموزشِ مدلِ جدید را بدون ضرر قابل توجهی در عملکرد

---

تعمیم فرضیه‌های حاصل کاهش می‌دهد. رویکرد سیستمی بر تعامل ضروری بین حفظ دانش و یادگیری انتقالی تاکید دارد. LML فقط یک الگوریتم نیست. LML می‌تواند از تحقیقات جدید در مورد الگوریتم‌های یادگیری و تکنیک‌های آموزشی بهره‌مند شود، اما شامل حفظ و سازماندهی دانش نیز می‌شود.

# حوزه‌های مرتبط با یادگیری ماشین همیشگی

زمینه‌های مرتبط زیادی با یادگیری ماشین همیشگی وجود دارد، از جمله: **یادگیری انتقالی**، **یادگیری چندوظیفه‌ای**، **یادگیری بی‌پایان**، **یادگیری خودآموخته**، **یادگیری برخط** و **یادگیری جهان باز**. به طور کلی، می‌توانیم آن‌ها را به عنوان انواع مختلف LML در نظر بگیریم در حالی که هر یک از آن‌ها بر مسائل فرعیِ خاصی تمرکز دارند. در بخش‌های زیر به‌طور خلاصه هر یک از آن‌ها شرح می‌دهیم.

## یادگیری انتقالی

یادگیری انتقالی، در سال‌های اخیر به‌طور گسترده‌ای مورد تحقیق قرار گرفته است. به‌طور کلی، یادگیری انتقالی شامل دو حوزه است: یک دامنه منبع و یک دامنه هدف. دامنه منبع دارای مقدار مناسبی از داده‌های آموزشی برچسب‌گذاری‌شده است، در حالی‌که دامنه هدف دارای داده‌های آموزشی برچسب‌گذاری شده کمی است یا اصلا وجود ندارد. هدف، استفاده از اطلاعاتِ بانظارت از دامنه‌ی منبع، برای کمک به پیش‌بینی در حوزه‌ی هدف است. به عبارت ساده‌تر، یادگیری انتقالی، کمک به فرآیندِ یادگیریِ یک کار مشخص با بهره‌برداری از دانش یک حوزه دیگر است. یادگیری انتقالی یک مورد خاص از LML است. چراکه معمولا فقط داده‌های دامنه منبع را حفظ می‌کند. یادگیری انتقالی همچنین معمولا فرض می‌کند که دامنه منبع و دامنه هدف ارتباط نزدیکی دارند.

## یادگیری چندوظیفه‌ای

یادگیری چندوظیفه‌ای، یادگیری چند کار مرتبط به‌طور همزمان، با هدف دستیابی به عملکرد بهتر با استفاده از اطلاعاتِ مربوطِ به اشتراک‌گذاری‌شده توسط وظایف است. یادگیری چندوظیفه‌ای همچنین از بیش‌برازش در وظیفه‌ی منفرد جلوگیری می‌کند و در نتیجه تعمیم بهتری دارد. یادگیری چندوظیفه‌ای معمولا بر کمینه‌کردن خطاها در همه وظایف متمرکز است و بنابراین وقتی یک وظیفه‌ی جدید وارد می‌شود، باید روی همه کارها از جمله تمام کارهای گذشته اجرا شود. از سوی دیگر، LML، دانش را از وظایفِ گذشته استخراج و انباشته می‌کند و تنها بر روی کار جدید با استفاده از دانشِ حفظ شده اجرا می‌شود. مشابه با یادگیری انتقالی، یادگیری چندوظیفه‌ای معمولا فرض می‌کند که وظایف به یکدیگر مرتبط هستند.

## یادگیری بی‌پایان[1]

یادگیری بی‌پایان منطقی مشابه با LML دارد به این معناکه هدف آن دستیابی به عملکرد بهتر و بهتر پس از مشاهده داده‌های بیشتر و بیشتر است. شناخته‌شده‌ترین سیستم یادگیری بی‌پایان **یادگیرنده‌ی زبان بی‌پایان[2]** (NELL) نام دارد که هدف آن بدست آوردن اطلاعات از وب برای ایجاد یک پایگاه دانش ساختاریافته است. در هر روز، هدف یادگیری، دستیابی به عملکرد بهتر از روز قبل است.

## یادگیری خودآموخته[3]

یادگیری خودآموخته، نوع خاصی از یادگیری انتقالی است که در آن حوزه منبع همان حوزه هدف است. بنابراین فقط روی یک دامنه متمرکز می‌شود. دانش از حجم زیادی از داده‌های بدون برچسب (کلان‌داده) بدست می‌آید که بسیار آسان‌تر از داده‌های برچسب‌دار است. داده‌های برچسب‌دار و داده‌های بدون برچسب به ترتیب با $D_L$ و $D_U$ نشان داده می‌شوند. هیچ فرضی در مورد رابطه‌ی بین $D_U$ و $D_L$ وجود ندارد. $D_U$ می‌تواند توزیع مولد متفاوتی از $D_L$ داشته باشد. $D_U$ نیازی به داشتن برچسب‌های $D_L$ ندارد.

مراحل اساسی یادگیری خودآموخته به شرح زیر است:

**۱.  آموختن یک بازنمایی سطح بالا از $D_U$.**

**۲.  نگاشت این ویژگی‌های بازنمایی[4] برای $D_L$.**

**۳.  یک مدل یادگیری بانظارت (به عنوان مثال، SVM) بر روی ویژگی‌های بازنمایی‌شده از مرحله ۲ ایجاد کنید.**

منطق یادگیری یک بازنمایی سطح بالاتر از $D_U$ این است که از طریق مقدار زیادی داده‌ی بدون برچسب، الگوریتم ممکن است بتواند "عنصر اساسی" که یک شی را تشکیل می‌دهد، یاد بگیرد. به عنوان مثال، برای تصاویر، ویژگی اصلی برای $D_L$ می‌تواند مقادیر شدت پیکسل باشد. از طریق یادگیری داده‌های بدون برچسب، الگوریتم ممکن است یاد بگیرد که تصاویر را با استفاده از لبه‌های روی تصاویر به جای مقادیر شدت پیکسل خام نمایش دهد. با اعمال این بازنمایی آموخته‌شده برای $D_L$، بازنمایی سطح بالاتری برای $D_L$ بدست می‌آوریم که انتظار می‌رود قابل تعمیم‌تر باشد. پس از یادگیری بازنمایی بدون‌نظارت، هر نمونه آموزشی اصلی، به فضای بُعد

---

جدید تبدیل می‌شود و یک الگوریتم یادگیری بانظارت، به عنوان مثال SVM، می‌تواند با استفاده از داده‌های آموزشی تبدیل‌شده ساخته شود.

## یادگیری برخط[1]

یادگیری برخط به‌طور گسترده‌ای در جامعه یادگیری ماشین مورد مطالعه قرار گرفته است. وظیفه آن، یادگیری از یک جریان ثابت داده است. پیکربندی یادگیری برخط شبیه LML است. به این معنا که در سناریوی جریان داده کار می‌کند. اما یادگیری برخط معمولا فرض می‌کند که داده‌های جدید توزیع یکسانی را با داده‌های موجود به اشتراک می‌گذارند، در حالی که LML همچنین در نظر می‌گیرد که داده‌های جدید ممکن است از یک وظیفه‌ی جدید ناشی شوند که توزیع یکسانی (یا حتی نامربوط) ندارد.

## یادگیری جهان باز[2]

یادگیری جهان باز با مشکلِ شناساییِ کلاس‌هایِ جدید، در زمان آزمون سروکار دارد، بنابراین از نسبت‌دهی اشتباه به کلاس‌های شناخته‌شده اجتناب می‌کند. وقتی کلاس‌های جدید در مدل ادغام می‌شوند، مساله یادگیری همیشگی را برطرف می‌کند. به این ترتیب، یادگیری جهان باز را می‌توان به عنوان وظیفه‌ی فرعی یادگیری همیشگی در نظر گرفت.

# یادگیری ماشین همیشگی برای مدل‌های تفکیک‌کننده

LML برای مسائل طبقه‌بندی به خوبی مورد بررسی قرار گرفته است. کارهای گذشته را می‌توان به سه دسته طبقه‌بندی کرد: **رویکردهای ترکیبی پودمانی**[3]، **رویکردهای مبتنی‌بر منظم‌سازی**[4] و **رویکردهای مرور**[5].

## رویکرد ترکیبی پودمانی

رویکردهای ترکیبی پودمانی، اجزا یا پارامترهای مدل متفاوتی را برای وظایف مختلف بکار می‌گیرند تا از فراموشی مدل جلوگیری کنند. ساده‌ترین رویکرد ترکیبی پودمانی، آموزش یک مدل جداگانه برای هر کار است. با این حال، این رویکرد در استفاده از مدل‌های آموزش‌دیده قبلی شکست خورده و در نتیجه منجر به اتلاف منابع می‌شود. برای حل این مشکل، **شبکه‌های**

---

[1] Online learning

[2] Open world learning

[3] modular compositional approaches

[4] regularization based approaches

[5] rehearsal approaches

**پیش‌رو**[1] با استفاده از دانش قبلی از مدل‌های آموزش‌دیده قبلی پیشنهاد شد. فرض کنید یک شبکه عصبی حاوی $L$ لایه‌ی $\{h_i^{(1)}\}_{i=1}^{L}$ در کار ۱ آموزش دیده است. هنگامی که وظیفه جدید می‌آید، پارامترهای مدل آموزش داده شده‌ی قبلی ثابت می‌شوند و شبکه‌های فرعیِ جدید معرفی و بروی نقشه‌های ویژگی‌های پیشین به منظور ساختن نقشه‌هایِ ویژگی برای این وظیفه‌ی جدید اعمال می‌شوند.

## رویکرد مبتنی‌بر منظم‌سازی

اگرچه رویکردهای ترکیبی پودمانی می‌توانند مشکل فراموشی فاجعه‌آمیز را برطرف کنند، نیاز به حافظه در آن‌ها بسیار زیاد است؛ چندین پودمان یا حتی یک پودمان در هر وظیفه برای کمک به یادگیری همه وظایف مورد نیاز است. نوع دیگری از رویکرد، **تنظیم ـ دقیق**[2] مدل آموزش‌دیده قبلی برای کارهای جدید و منظم‌سازی شبکه برای حفظ عملکرد مدل در وظایف قبلی است. این نوع رویکرد مبتنی‌بر منظم‌سازی معمولا یک مدل را برای همه وظایف نگه می‌دارد و یا پارامترهای بسیار محدودی برای هر کار جدید اضافه می‌کند. دو راه برای تنظیم پارامترهای شبکه وجود دارد. اولین راه این است که پارامترهای شبکه‌ی مهم وظایف قبلی را از تغییر در هنگام یادگیری وظایف بعدی جریمه کنید. راه دوم این است که پارامترهای شبکه را منظم‌سازی کنید تا خروجی‌های مدل‌های آموزش دیده قبلی هنگام یادگیری کارهای بعدی تغییر نکنند.

## رویکرد مرور

رویکردهای مرور، شامل یک بافر حافظه است که تعداد کمی از نمونه‌ها را برای وظیفه‌های قبلی ذخیره می‌کند. پیشنهاد شده است که زیرمجموعه‌ای از نمونه‌ها در هر دسته نگهداری شود که می‌تواند میانگین هر کلاس را به بهترین نحو برآورد کند. روش مرور مستعد بیش‌برازش برروی نمونه‌های ذخیره شده است. برای پرداختن به این مشکل، اخیرا رویکردهای مبتنی‌بر حافظه‌ی اپیزودیک گرادیان[3] (GEM) ارائه شده است. هدف GEM بروزرسانی پارامترهای مدل آموزش دیده به گونه‌ای است که زیان برای تمام وظیفه‌های قبلی افزایش پیدا نکند.

## تشخیص خارج از توزیع[4]

شبکه‌های عصبی عمیق اغلب با فرضیات جهان بسته آموزش داده می‌شوند، یعنی فرض می‌شود که توزیع داده‌های آزمایشی مشابه توزیع داده‌های آموزشی است. با این حال، هنگامی که در

---

[1] Progressive Networks

[2] fine-tune

[3] Gradient Episodic Memory

[4] Out-of-distribution detection

کارهای دنیای واقعی بکار گرفته می‌شود، این فرض درست نیست و منجر به کاهش قابل توجهی از عملکرد آن‌ها می‌شود. اگرچه این افت عملکرد برای کاربردهای همانند توصیه‌گرهای محصول قابل قبول است، اما استفاده از چنین سیستم‌هایی در حوزه‌های همانند پزشکی و رباتیک خانگی خطرناک است، چراکه می‌توانند باعث بروز حوادث جدی شوند.

هنگامی که شبکه‌های عصبی عمیق دادههایی را پردازش می‌کنند که شبیه توزیع مشاهده شده در زمان آموزش نیستند (به اصطلاح خارج از توزیع نامیده می‌شوند)، اغلب پیش‌بینی‌های اشتباهی انجام می‌دهند و این کار را با اطمینان بیش از حد انجام می‌دهند (شکل ۹ ـ ۳ را ببینید). یک سیستم هوش مصنوعی ایده‌آل باید در صورت امکان به نمونه‌های خارج از توزیع تعمیم پیدا کند. بنابراین، توانایی تشخیصِ خارج از توزیع برای بسیاری از برنامه‌های کاربردی دنیای واقعی بسیار مهم است.

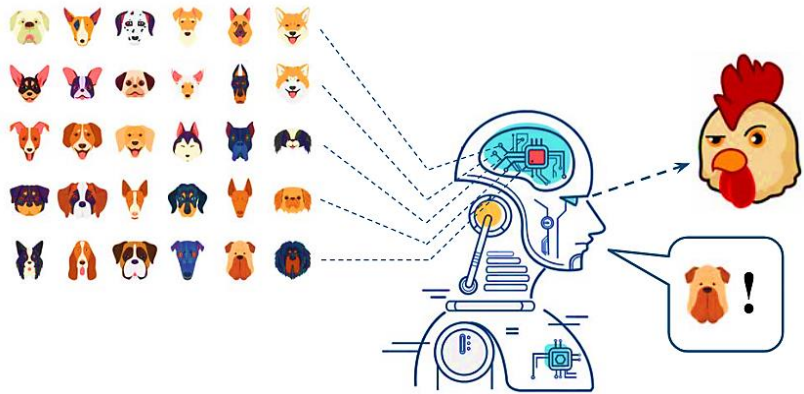

**شکل ۹ ـ ۳.** هنگامی که یک نمونه‌ی جدیدِ خارج از توزیع آموخته شده ارائه شود، شبکه‌های عصبی یک کلاس را از توزیع آموخته‌شده با اطمینان زیادی پیش‌بینی می‌کنند. تشخیص خارج از توزیع الگوریتم‌هایی را برای رفع این مشکل پیشنهاد می‌کند.

تشخیص خارج از توزیع برای تضمین قابلیت اطمینان و ایمنی سیستم‌های یادگیری ماشین ضروری است. به عنوان مثال، در رانندگی خودران، ما می‌خواهیم سیستم رانندگی زمانی که صحنه‌های غیرعادی یا اشیایی راکه قبلا ندیده است و نمی‌تواند تصمیم ایمن بگیرد را تشخیص دهد، هشدار بدهد و کنترل را به انسان واگذار کند.

این مشکل برای اولین بار در سال ۲۰۱۷ ظاهر شد و از آن زمان به بعد توجه روزافزونی از سوی جامعه تحقیقاتی را به خود جلب کرده است. بیشتر کارهای اخیر در مورد تشخیص خارج از توزیع مبتنی بر آموزش بانظارت شبکه‌های عصبی است که از **زیان آنتروپی متقابل**[1] را بهینه

---

[1] cross-entropy loss

می‌کند. در این موارد خروجی شبکه یک تناظر مستقیم با راه‌حل مساله دارد، یعنی احتمال برای هر کلاس. با این حال، جمع نمایشِ بردارِ خروجی مجبور است همیشه به یک برسد. این بدان معناست که وقتی به شبکه یک ورودی نشان داده می‌شود که بخشی از توزیع آموزشی نیست، بازهم احتمال را به نزدیک‌ترین کلاس می‌دهد تا جمع احتمالات به یک برسد. این پدیده منجر به مشکل شناخته شده‌ی شبکه‌های عصبی زیاد مطمئن[1] به محتوایی شده است که هرگز ندیده‌اند.

# یادگیری تقویتی

با وجود اینکه یادگیری عمیق توانایی بازنمایی قدرتمند داده‌ها را دارد و در بسیاری از مسائل دسته‌بندی و پردازش تصویر از بسیاری روش‌های دیگر بهتر عمل می‌کند، اما برای ساخت یک سیستمِ هوشمندِ هوش مصنوعی کافی نیست. این امر از آنجایی ناشی می‌شود که یک سیستم هوش مصنوعی نه تنها باید توانایی آموختن از داده‌ها را داشته باشد، بلکه باید همانند انسان از تعاملات با محیط دنیای واقعی نیز بیاموزد. یادگیری تقویتی یکی از حوزه‌های یادگیری ماشین می‌باشد و تمرکز بر این موضوع دارد تا ماشین را قادر به تعامل با محیط دنیای واقعی کند.

یادگیری تقویتی از طریق عامل در تلاش است با آزمون و خطا مساله را از طریق تعامل با محیطی که برای عامل ناشناخته می‌باشد، حل کند. عامل می تواند ضمن انجام بازخورد فوری از محیط، با اقدامات خود وضعیت محیط را تغییر دهد. بازخورد معمولا به عنوان **پاداش** در یادگیری تقویتی نامیده می‌شود. عامل با دریافت پاداش‌های مثبت بیشتر از محیط، توانایی یادگیری بهتر را کسب می‌کند. به‌طور کلی، هدف عامل پیدا کردن یک زنجیره بهینه از اقدامات است تا مساله را حل کند. یادگیری تقویتی، معمولا به عنوان یک فرآیند تصمیم‌گیری مارکوف مدل‌سازی می‌شود و می‌توان آن را همانند شکل ۹ ـ ٤ توصیف کرد.

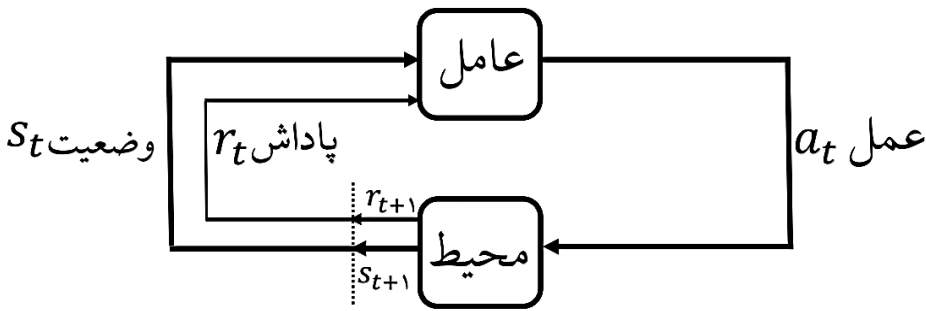

**شکل ۹ ـ ٤.** نمایش یادگیری تقویتی

---

[1] overconfident

همان‌طور که در شکل ۹ـ٤ قابل مشاهده است، واحد یادگیرنده که عامل نامیده می‌شود، با انتخاب عمل $a$ از مجموعه اعمال ممکن، محیط خود را به‌طور فعال تغییر می‌دهد. پس از اجرای عملی، محیط مطابق با آن تغییر می‌کند و وضعیت جدید $s$ را به عامل می‌گوید. علاوه براین، محیط با انتشار یک سیگنال پاداش $r$ به عامل، بازخوردی درباره‌ی عمل انتخاب شده با توجه حالات داده‌شده ارائه می‌دهد. یک عامل همچنان به فعالیت خود ادامه می‌دهد، در نتیجه از یک وضعیت به وضعیت دیگر منتقل، تا زمانی که به یک حالت نهایی برسد.

عناصر تشکیل‌دهنده یک سیستم یادگیری تقویتی را می‌توان به‌صورت زیر فهرست کرد:

- **عامل:** برنامه‌ای که با هدف انجام کاری مشخص، آموزش می‌بیند.
- **محیط:** جهانی واقعی یا مجازی، که در آن عامل اقداماتی را انجام می‌دهد.
- **عمل:** حرکتی که توسط عامل انجام می‌شود و سبب تغییر وضعیت (حالت) در محیط می‌شود.
- **پاداش:** یک تابع پاداش، هدف را در یک مساله یادگیری تقویتی تعریف می‌کند و هر وضعیت مشاهده شده از محیط را به یک عدد واحد نگاشت می‌کند که مطلوبیت ذاتی آن وضعیت را نشان می‌دهد. هدف از پاداش در یادگیری تقویتی، ارزیابی یک عمل که می‌تواند مثبت یا منفی باشد و بازخوردی است که عامل پس از هر عمل از محیط دریافت می‌کند. پاداش واقعی یک عمل درست انجام شده در یک وضعیت خاص، ممکن است بلافاصله محقق نشود.
- **وضعیت:** کلیه اطلاعاتی که عامل در محیط فعلی خود در اختیار دارد. برای مثال در یک بازی شطرنج، وضعیت، موقعیت همه مهره‌ها در صفحه است.
- **مشاهدات:** با توجه به اینکه در برخی مسائل، عامل بر وضعیت کامل محیط دسترسی ندارد، معمولا مشاهده بخشی از وضعیتی است که عامل می‌تواند مشاهده کند. به عبارت دیگر، مشاهدات اطلاعاتی هستند که محیط در اختیار عامل قرار می‌دهد و آنچه در اطراف عامل رخ می دهد را نشان می‌دهد. با این حال، اغلب در ادبیات، وضعیت و مشاهده به‌جای یکدیگر استفاده می‌شوند.
- **خط‌مشی:** مشخص می‌کند که عامل با توجه به وضعیت فعلی چه اعمالی انجام خواهد داد. در زمینه یادگیری عمیق، می‌توانیم یک شبکه عصبی را آموزش دهیم تا این تصمیمات را بگیرد. در طول دوره آموزش، عامل سعی می‌کند خط‌مشی خود را اصلاح کند تا تصمیمات بهتری بگیرد. وظیفه یافتن خط مشی بهینه، بهبود خط‌مشی (کنترل) نامیده می‌شود و یکی از اصلی ترین مسائل در یادگیری تقویتی است.
- **تابع مقدار:** مشخص می‌کند که چه چیزی برای عامل در درازمدت خوب است. به عبارت دیگر، وقتی تابع مقدار را روی یک حالت معین اعمال می‌کنیم، اگر از آن حالت شروع کنیم، کل بازدهی را که می‌توان در آینده انتظار داشت را به ما می‌دهد.

مثال‌هایِ ساده‌یِ زیر کمک می‌کنند تا سازوکار یادگیری تقویتی را بهتر درک کنید:

فرض کنید گربه‌ای دارید و قصد دارید به آن آموزش دهید تا اقدامات خاصی را انجام دهد. از آنجا که گربه فارسی یا هر زبان دیگری را نمی‌فهمد، نمی‌توان به‌طور مستقیم به او گفت که چه کاری را باید انجام دهد. در عوض، می‌توان یک استراتژی متفاوتی را دنبال کرد. ما یک وضعیت را ارائه و گربه سعی می‌کند به طرق مختلف به آن پاسخ دهد. اگر پاسخ گربه روش مطلوبی باشد، به او ماهی می‌دهیم. اکنون هر زمان که گربه در معرض همان وضعیت قرار گیرد، گربه نیز با اشتیاق بیشتر انتظار مشابهی را برای دریافت پاداش (غذا) دارد. چراکه یادگرفته است که اگر یک عمل خاص را انجام دهد، پاداش دریافت می‌کند.

در این مثال:

- گربه عاملی است که در معرض محیط که در این مورد خانه می‌باشد، قرار می‌گیرد.
- وضعیت می‌تواند نشستن گربه باشد و شما از گفتن کلمه‌ای خاص برای راه رفتن گربه استفاده می‌کنید.
- عامل با انجام یک عمل، با انتقال از یک وضعیت به وضعیت دیگر واکنش نشان می‌دهد. به عنوان مثال، گربه از حالت نشستن به راه رفتن می‌رود.

می‌توان مثال دیگری از کودکان ارائه کرد. بچه ها اغلب اشتباهاتی را انجام می‌دهند. بزرگسالان سعی می‌کنند اطمینان حاصل کنند که کودک از این اشتباه درس گرفته و سعی می‌کنند دیگر آن را تکرار نکنند. در این حالت می‌توانیم از مفهوم بازخورد استفاده کنیم. اگر والدین سخت‌گیر باشند، فرزندان را برای هرگونه خطا سرزنش می‌کنند، که بازخوردی منفی است. کودک از این پس با انجام این خطا به یاد می‌آورد که اقدام اشتباهی انجام داده است، چراکه با سرزنش والدین همراه خواهد بود. پس از آن بازخورد مثبتی نیز وجود دارد. جایی که والدین ممکن است آن‌ها را بخاطر انجام کاری درست، تحسین کنند. در اینجا، ما یک عمل صحیح را به طریقی خاص اجرا یا در تلاش برای اجرای آن هستیم.

به‌طور خلاصه می‌توان این‌گونه بیان کرد که، یادگیری تقویتی نوعی روش شناسی یادگیری است که در آن به الگوریتم با پاداش بازخورد می‌دهیم تا از آن یاد گرفته، تا بهبود نتایج را در آینده به‌همراه داشته باشد.

## فرآیندهای تصمیم‌گیری مارکوف

فرآیندهای تصمیم‌گیری مارکوف، یک مدل ریاضی تصادفی برای یک سناریوی تصمیم‌گیری است. در هر مرحله، تصمیم گیرنده یا به عبارت‌دیگر همان عامل، عملی را انتخاب می‌کند. در این مدل، بخشی از نتایج آن تصادفی و بخشی دیگر نیز نتیجه عمل است. فرآیندهای تصمیم‌گیری مارکوف، برای مدل‌سازی انواع مسائل بهینه‌سازی استفاده و از طریق برنامه‌نویسی پویا و یادگیری‌تقویتی قابل حل هستند.

فرآیند تصمیم‌گیری مارکوف، همانند یک نمودار جریانی با دایره‌هایی که وضعیت‌ها را نشان می‌دهند. از هر دایره فلش‌های خارج می‌شود که نمایانگر تمام اقدامات احتمالی قابل انجام از آن وضعیت می‌باشد. به عنوان مثال، یک فرآیند تصمیم‌گیری مارکوف در بازنمود یک بازی شطرنج، دارای وضعیت‌هایی است که نشان‌دهنده‌ی محل قرارگیری مهره‌ها صفحه شطرنج است و اعمالی که نمایانگر حرکات احتمالی براساس مهره‌های در صفحه‌ی شطرنج می‌باشند.

یک ویژگی اصلی یک فرآیند تصمیم‌گیری مارکوف، این است که هر وضعیت باید تمام اطلاعات مورد نیاز عامل را برای تصمیم‌گیری آگاهانه در اختیار داشته باشد، الزامی که "دارایی مارکوف" نامیده می‌شود. اساساً دارایی مارکوف می‌گوید که نمی‌توان انتظار داشت عامل از خود حافظه تاریخی خارج از وضعیت خود داشته باشد. به عنوان مثال، وضعیت فعلی تخته شطرنج، همه مواردی را که برای انجام بهترین حرکت بعدی است را می‌گوید و نیازی به حرکاتی که قبلا انجام شده نیست تا آن‌ها را به‌خاطر بیاورد .

در عمل، برای اینکه یادگیری تقویتی بتواند مساله‌ای را حل کند، الزامی به الگو قرار دادن مساله دنیای واقعی نیست. به عنوان مثال، خاطره‌ی من از چگونگی بازی شطرنج توسط یک حریف خاص ممکن است در روند تصمیم‌گیری من در دنیای واقعی نقش داشته باشد، اما می‌توان با یادگیری تقویتی یک عامل بازی شطرنج را برنده ساخت بدون اینکه نیازی به این اطلاعات داشته باشد.

فرآیندهای تصمیم گیری مارکوف، توسط یک ۵تایی از عناصر $< S, A, P, R, \gamma >$ تعریف می‌شود، جایی که:

- **S**: مجموعه‌ای از وضعیت‌ها که شامل تمام نمایش‌های احتمالی محیط است.
- **A**: در هر وضعیت، محیط مجموعه‌ای از اعمال را در یک فضای عملیاتی در اختیار عامل قرار می‌دهد تا عامل، از این اعمال انتخاب نماید. عامل از طریق اعمال بر محیط تاثیر می‌گذارد.
- **P**: $P = (s, a, \acute{s}) = P_r(s_{t+1} = \acute{s} | s_t = s, a_t = a)$ ماتریس انتقالی است که احتمال می‌دهد عمل $a$ در وضعیت $s$ در زمان $t$ منجر به وضعیت $\acute{s}$ در زمان $t+1$ می‌شود.
- **R**: $R = (s, a, \acute{s})$ پاداش مورد انتظار است که عامل پس از عمل $a$ در وضعیت $s$ و رسیدن به وضعیت $\acute{s}$ دریافت می‌کند.
- **γ**: فاکتور نزول است و نشان‌دهنده اهمیت بین پاداش‌های کوتاه‌مدت و درازمدت می‌باشد.

مساله اساسی در فرآیندهای تصمیم‌گیری مارکوف، یافتن "خط‌مشی" برای تصمیم‌گیرنده است؛ یک تابع $P$ که وضعیت‌ها را به اعمال $a = \pi(s)$ نگاشت می‌کند. این خط‌مشی می‌تواند

قطعی یا تصادفی باشد. هدف این است که یک خطمشی را بیابیم که مجموع پاداش فاکتور نزول از هر وضعیت به وضعیت بعد را بیشینه کند:

$$G_t = \sum_{i=t}^{\infty} \gamma^i . R(s_i, a_i, s_{t+1})$$

از $G_t$ به عنوان بازده نام برده میشود.

# عامل

عامل کسی یا چیزی است که با انجام برخی اعمال، مشاهدات و دریافت پاداشهای نهایی با این محیط ارتباط برقرار میکند. عامل مولفهای است که براساس پاداش و تنبیه تصمیم میگیرد، چه اقدامی باید صورت پذیرد. برای تصمیمگیری، عامل مجاز است از هرگونه مشاهده از محیط و هرگونه قانون داخلی استفاده کند. این قوانین داخلی میتوانند هر چیزی باشند، اما بهطور معمول در یادگیری تقویتی، انتظار براین است که وضعیت فعلی توسط محیط تامین تا وضعیت، خصوصیت تصمیمگیری مارکوف را داشته باشد و پس از آن، با استفاده از یک تابع خطمشی تصمیم میگیرد چه تصمیمی باید صورت گیرد.

در بیشتر سناریوهای عملی یادگیری تقویتی، عامل نرمافزار ماست که قرار است برخی از مسائل و مشکلات را به روشی کموبیش کارآمد حل کند. عامل یکی از مهمترین عناصر تشکیلدهنده یک سیستم مبتنیبر یادگیری تقویتی است. چرا که شامل هوش در جهت تصمیمگیری و توصیه اقدامات (اعمال) بهینه در هر شرایط است. از آنجاکه عامل در یادگیری تقویتی نقش بسیار مهمی را داراست، تحقیقات زیادی در مورد معماری یادگیری و مدلهای مرتبط انجام شده است. در ادامه الگوریتمها را بر اساس عاملها به الگوریتمهای **مبتنیبر مقدار**، **مبتنیبر خطمشی** و **مبتنیبر مدل** تقسیمبندی میکنیم.

## الگوریتمهای مبتنیبر مقدار

در الگوریتمهای مبتنیبر مقدار، توابع مقدار $V^\pi(s)$ به وضعیتها اختصاص داده میشوند و عامل تصمیمات خود را برمبنای مقادیر وضعیتها میگیرد. تابع مقدار، تابعی است که میزان خوب بودن وضعیت را برمبنای پیشبینی پاداش آینده ارزیابی میکند. دو نوع متفاوت از تابع مقدار وجود دارد:

- **تابع وضعیت_مقدار** که معمولا به عنوان تابع مقدار خوانده میشود، بازده $G_t$ مورد انتظار با شروع از یک وضعیت $s$ و پیروی از خطمشی $\pi$ است و توسط معادله بلمن بهصورت زیر تعریف میشود:

$$V^\pi(s) = \mathbb{E}[G_t | S_t = s]$$

$$= \sum_{a \in A} \pi(a|s) \sum_{\acute{s} \in S} \pi(\acute{s}|s,a)[R = (s,a,\acute{s}) + \gamma^{V^{\pi(\acute{s})}}]$$

- **تابع وضعیت ــ عمل** که معمولا با عنوان $Q$ ــ مقدار خوانده می‌شود، بازده $G_t$ مورد انتظار یک جفت وضعیت ــ عمل در زمان $t$ و پیروی از خط‌مشی $\pi$ است و به‌طور مشابه توسط معادله بلمن به‌صورت زیر تعریف می‌شود:

$$Q^{\pi}(s,a) = \mathbb{E}[G_t|S_t = s, A_t = a]$$
$$= \sum_{\acute{s} \in S} P(\acute{s}|s,a)[R = (s,a,\acute{s}) + \gamma \sum_{a \in A} \pi(\acute{a},\acute{s}) + Q^{\pi}(\acute{s},\acute{a})]$$

### معادلات بلمن

معادلات بلمن به مجموعه معادلاتی گفته می‌شود که تابع مقدار را به پاداش فوری به علاوه مقادیر آتی تنزیل‌یافته تجزیه می‌کند.

$$V^{\pi}(s) = \mathbb{E}[G_t|S_t = s]$$
$$= \mathbb{E}[R_{t+1} + \gamma R_{t+2} + \gamma^2 R_{t+3} + \ldots |S_t = s]$$
$$= \mathbb{E}[R_{t+1} + \gamma(R_{t+2} + \gamma R_{t+3} + \ldots)|S_t = s]$$
$$= \mathbb{E}[R_{t+1} + \gamma G_{t+1}|S_t = s]$$
$$= \mathbb{E}[R_{t+1} + \gamma V(S_{t+1})|S_t = s]$$

به‌طور مشابه برای $Q$ ــ مقدار:

$$Q^{\pi}(s,a) = \mathbb{E}[R_{t+1} + \gamma V(S_{t+1})|S_t = s, A_t = a]$$
$$= \mathbb{E}[R_{t+1} + \gamma \mathbb{E}_{a \sim \pi} Q(S_{t+1},a)|S_t = s, A_t = a]$$

معادله زیر بین توابع وضعیت ــ مقدار و وضعیت ــ عمل برقرار است:

$$V^{\pi}(s) = \sum_{a} \pi(a|s) Q^{\pi}(s,a)$$

هدف عامل بیشینه‌کردن کل پاداش تجمعی در بلند مدت است. به عبارتی دیگر، هدف از یادگیری تقویتی، یافتن خط‌مشی بهینه است. خط‌مشی که حداکثر پاداش تجمعی را بیشینه می‌کند، خط‌مشی بهینه نامیده می‌شود و با $\pi^*$ نشان داده می‌شود. خط‌مشی بهینه $\pi^*$ به این صورت است که مقدار هر وضعیت $s$ تحت $\pi^*$ بیشتر یا برابر با مقدار وضعیت $s$ تحت خط‌مشی دیگر $\pi'$ برای همه‌ی $s \in S$ باشد:

$$V^{\pi^*}(s) = V^*(s) \geq V^{\pi'} \ \forall \ s \in S, \pi'$$

اگر تابع وضعیت ــ مقدار بهینه باشد، عامل از خط‌مشی بهینه استفاده کرده و ممکن است چندین خط‌مشی بهینه وجود داشته باشد که منجر به همان تابع وضعیت ــ مقدار بهینه شود. تابع وضعیت ــ مقدار بهینه $V^*$ را می‌توان به شرح زیر تعریف کرد:

$$V^*(s) = max_{\pi} V^{\pi}(s) \ \forall \ s \in S$$

علاوه براین، خط‌مشی بهینه منجر به تابع وضعیت ـ عمل بهینه $Q^*$ نیز می‌شود:

$$Q^*(s,a) = max_\pi Q^\pi(s,a) \ \forall \ s \in S, a \in A$$

$$= \mathbb{E}[R_{t+1} + \gamma V^*(\acute{s})|S_t = s, A_t = a]$$

در پایان، می‌توان معادله بهینه‌سازی بلمن را از معادلات معرفی شده قبلی استخراج کرد:

$$V^*(s) = max_{a \in A(s)} Q^{\pi^*}(s,a)$$

$$= max_a \ \mathbb{E}[R_{t+1} + \gamma V^*(\acute{s})|S_t = s, A_t = a]$$

$$= max_a \sum_{\acute{s} \in S} P(\acute{s}|s,a)[R(s,a,\acute{s}) + \gamma V^*(\acute{s})]$$

$$= max_a \sum_{\acute{s} \in S} P(\acute{s}|s,a)[R(s,a,\acute{s}) + \gamma max_{\acute{a}} Q^{\pi^*}(\acute{s},\acute{a})]$$

## نمودار پشتیبان‌گیری در یادگیری تقویتی

معادلات بلمن با استفاده از درختی از حالات و کنش‌ها (نمودار پشتیبان‌گیری) بهتر قابل درک هستند. نمودار پشتیبان‌گیری، یک نمایش بصری از الگوریتم‌ها و مدل‌های مختلف در یادگیری تقویتی را نشان می‌دهد.

فرآیند پشتیبان‌گیری (عملیات بروزرسانی)، نمایش گرافیکی الگوریتم، با نشان‌دادن وضعیت، عمل، وضعیت انتقال، پاداش و غیره است. در این نمودار وضعیت ـ مقدار با یک دایره توخالی درحالیکه عمل ـ مقدار با یک دایره توپر نمایش داده می‌شود. همچنین، عمل با یک فلش شروع‌شده از وضعیت نمایش داده می‌شود.

حال بیایید نحوه نمایش یک وضعیت ـ مقدار را با استفاده از نمودارهای پشتیبان‌گیری بررسی کنیم (شکل ۹ ـ ۵).

۱. $s$ حالت شروع و گره اصلی است.

۲. از حالت $s$ سه عمل می‌تواند صورت گیرد. همان‌طور که توسط فلش نشان‌داده شده و عامل طبق خط‌مشی $\pi$ عمل می‌کند.

۳. در صورتی که یک محیط تصادفی با احتمال انتقال معین باشد، این عامل می‌تواند در حالت‌های مختلفی به پایان برسد. همان‌طور که در شکل مشاهده می‌شود ۳ حالت احتمالی نشان داده شده که عامل می‌تواند پس از انجام بیشترین عمل درست به آن وارد شود.

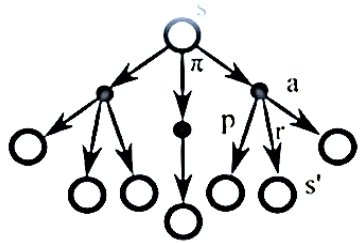

**شکل ۹ ـ ۵.** نمودار پشتیبان‌گیری وضعیت ـ مقدار $V^\pi(s)$

## الگوریتم‌های مبتنی‌بر خط‌مشی

خط‌مشی، استراتژی‌ای است که سازوکار یادگیری برای تعیین بهترین اقدام بعدی را براساس وضعیت فعلی امکان‌پذیر می‌کند. خط‌مشی، هسته اصلی عامل یادگیری تقویتی است؛ به این معنا که به تنهایی برای تعیین رفتار کافی است.

الگوریتم‌های مبتنی‌بر خط‌مشی، در مقابل با الگوریتم‌های مبتنی‌بر مقدار، رویکرد مستقیم‌تری را در پیش می‌گیرند. در این رویکرد، به‌جای یافتن مقدار هر وضعیت ممکن و سپس بدست آوردن خط‌مشی بهینه، به دنبال یافتن مستقیم خط‌مشی هستند تا بازده مورد انتظار بیشینه شود.

به عبارت دیگر، هسته اصلی روش‌های مبتنی‌بر خط‌مشی، بروزرسانی مجدد پارامترهای خط‌مشی $\theta$ است؛ به‌طوری که بازده $G_t$ مورد انتظار افزایش یابد. به عبارت ساده‌تر، در رویکرد مبتنی‌بر خط‌مشی، در ابتدا یک خط‌مشی تصادفی انتخاب و در مرحله ارزیابی تابع مقدار آن خط‌مشی پیدا می‌شود. سپس، خط‌مشی جدید را با استفاده از تابع مقدار محاسبه تا در هر مرحله بهبود پیدا کند. این روند را تا زمانی که خط‌مشی بهینه پیدا شود تکرار می‌کند. روند بهینه‌سازی به‌صورت زیر توصیف می‌شود:

$$\theta_{i+1} = \theta_i + \Delta\theta_i$$

جایی که $\theta_i$ مجموعه‌ای از پارامترهای خط‌مشی است که برروی خط‌مشی $\pi$ پارامترسازی می‌شوند و $\Delta\theta_i$ تغییرات پارامترهای خط‌مشی است.

## استخراج[1] در مقابل اکتشاف[2]

انسان‌ها قبل از انجام هرکاری سعی می‌کند هرچه بیشتر اطلاعات کسب کنند. به عنوان مثال، قبل از اینکه یک رستوران جدید را امتحان کند، سعی می‌کند نظرات را بخواند یا از دوستانی که قبلاً آن را امتحان کرده‌اند پرسش‌وجو کند. از طرف دیگر، در یادگیری تقویتی، انجام این کار امکان پذیر نیست. اما برخی از تکنیک‌ها وجود دارد که به کشف بهترین استراتژی کمک می‌کند.

در یادگیری تقویتی، عامل در زمان تصمیم گیری در محیط، می‌تواند دو استراتژی را اعمال کند:

• **اکتشاف:** یک عمل تصادفی را انتخاب کند. با پیروی از این روش، عامل می‌تواند از وضعیت‌های جدید بازدید کند و خط‌مشی‌های جدید و بهتری را پیدا کند. نتیجه آن، جمع‌آوری اطلاعات بیشتر که ممکن است به تصمیم‌گیری بهتر در آینده منجر شود.

---

[1] exploitation

[2] exploration

- **استخراج:** حریصانه عمل کند. با استفاده از پاداش کلی، بهترین اعمال شناخته شده را با توجه به دانش موجود انتخاب کند. به عبارت دیگر، قصد دارد با توجه به اطلاعات فعلی بهترین تصمیم را بگیرد.

در یادگیری تقویتی به این نوع دو از تصمیم‌گیری: زمانی که ادامه کار قبلی را انجام می‌دهید استخراج و در هنگام امتحان کارهای دیگر اکتشاف گفته می‌شود.

یکی از ویژگی‌های کلیدی یادگیری تقویتی، معضل استخراج در مقابل اکتشاف است. اگر عامل بخواهد اعمال بهتری بیاموزد یا به عبارت دیگر، اعمالی که در نهایت به پاداش انباشته بیشتری منجر شود، باید اعمال جدید را امتحان کند. همچنین اگر عامل از دانش فعلی خود استخراج کرده و اقدامات قبلاً شناخته شده برای بازخورد پاداش‌های خوب را دنبال کرده باشد، تضمین نمی شود که بازدهی بالاتر از پاداشی که عامل می‌توانست دریافت کند، داشته باشد. بنابراین این معضلی است که یک عامل هنگام تصمیم‌گیری برای اعمال بعدی با آن روبرو می‌شود: یا اعمال تصادفی را امتحان کند و حدس بزند که پاداش‌های بیشتری می‌گیرد، اما با خطر بدتر شدن امتیاز مواجه است. یا اینکه با عملکرد مطابق با شرایط فعلی خود به پاداش احتمالاً پایین‌تر اما مطمئن دست پیدا کند. به عبارت دیگر، اگر عامل تنها اکتشاف را انجام دهد، ممکن است در این کار به امتیازات بالاتری نرسد و اعمال خود را بهبود ندهد. در طرف دیگر، اگر فقط از استخراج استفاده شود، ممکن است در خط‌مشی فعلی خود با دیدن تمام خط سیرهای احتمالی گیر کند. از همین رو، عامل احتمالا خط‌مشی بهینه را از دست خواهد داد. بنابراین، باید یک تعادل خوب بین اکتشاف و استخراج وجود داشته باشد.

این معضل از آنجایی ناشی می‌شود که، فرآیند یادگیری در یادگیری تقویتی به‌صورت برخط صورت می‌گیرد. به عبارت دیگر، به یادگیری تقویتی هیچ داده‌ای همانند یادگیری بانظارت داده نمی‌شود. بنابراین، عامل خود به نوعی به‌دنبال جمع‌آوری داده‌ها است و از طریق اعمالی که انجام می‌دهد، بر داده‌های مشاهده‌شده اثر می‌گذارد. از همین رو، گاهی ارزش دارد که اعمال مختلفی را برای بدست آوردن داده‌های جدید انجام دهد.

معضل اکتشاف در مقابل استخراج یک موضوع تکرارشونده در یادگیری تقویتی و به‌طور کلی هوش مصنوعی است. آیا باید از دانش به دست آمده استخراج کنیم، یعنی باید یک مسیر شناخته شده با پاداش بالا را دنبال کنیم؟ یا باید در جستجوی خط‌مشی جدید بهتر، وضعیت‌های ناشناخته را کشف کنیم؟ تعادل بین هر دو استراتژی تصمیم‌گیری، عملکرد یادگیری عامل بسیار بهبود می‌بخشد. یک پاسخ قابل قبول برای حل آن، این است که ابتدا یک عامل لازم است بیشترین وضعیت‌ها را کشف و به دنبال آن با استخراج از دانش جمع‌آوری شده، نتایج بهتری را کسب کند. اما نباید فراموش کرد که در محیط‌های پویا، بدست‌آوردن اینکه آیا اکتشاف کافی صورت گرفته یا نه بسیار پیچیده می‌باشد. با این وجود، چند روش برای انتخاب اعمال ارائه شده است که در ادامه به تشریح آن‌ها می‌پردازیم.

**استخراج:** به‌معنای بیشینه‌کردن عملکرد عامل با استفاده از دانش موجود است که عملکرد آن معمولاً با پاداش مورد انتظار ارزیابی می‌شود. به عنوان مثال، یک جوینده طلا اکنون یک سنگ معدن دارد که روزانه دو گرم طلا برای او فراهم می‌کند و او می‌داند که بزرگ‌ترین سنگ طلا می‌تواند پنج گرم طلا در روز به او بدهد. با این حال، همچنین می‌داند که یافتن یک سنگ معدن جدید نه تنها او را مجبور به متوقف کردن استخراج از سنگ معدن فعلی می‌کند، بلکه برای او هزینه اضافی را نیز در پی دارد چرا که در نهایت خطر پیدا نکردن هیچ چیز را نیز به همراه دارد. با توجه به این موارد، او تصمیم می‌گیرد سنگ معدن فعلی را حفر کند تا حداکثر پاداش (در این حالت طلا) از طریق استخراج بیشتر شود و با توجه به خطرات بزرگ اکتشاف، از اکتشاف صرف نظر می‌کند. خط‌مشی که وی در اینجا اتخاذ کرده خط‌مشی حریصانه است، به این معنی که عامل بطور مداوم عملی را انجام می‌دهد که بالاترین پاداش مورد انتظار را براساس اطلاعات فعلی بدست می‌آورد، نه اینکه آزمایش‌های خطرناکی را انجام دهد که منجر به کاهش پاداش‌های مورد انتظار شود.

**اکتشاف:** به‌معنای افزایش دانش موجود با اقدامات و تعامل با محیط است. به مثال حفار طلا برگردیم، او آرزو دارد مدتی را صرف یافتن سنگ معدن جدید کند و اگر سنگ معدن طلای بزرگ‌تری پیدا کرد، می‌تواند روزانه جایزه بیشتری دریافت کند. برای داشتن بازده بلندمدت بیشتر، ممکن است بازده کوتاه مدت فدا شود. حفار طلا همیشه با معضل استخراج و اکتشاف روبرو است، چراکه باید تصمیم بگیرد که معدن طلا چه میزان بازده برای ماندن دارد و بازده معدن طلا برای ادامه کار در او چقدر است. تعامل بین اکتشاف و استخراج تعادل بین میزان تلاش عامل به ترتیب در اکتشاف و بهره برداری را توصیف می‌کند. تعامل بین اکتشاف و استخراج، موضوع اصلی تحقیقات یادگیری تقویتی و توسعه الگوریتم‌های یادگیری تقویتی است.

### استراتژی انتخاب حریصانه-اپسیلون

یک روش ممکن، ساده و محبوب و در عین حال موثر برای انتخاب عمل در هر مرحله به عنوان استراتژی انتخاب، انتخاب حریصانه‌ـ اپسیلون است. در این روش، یک پارامتر $e$ بین ۰ تا ۱ اقدام عامل مبنی‌بر اینکه، استخراج یا اکتشاف انجام می‌دهد را کنترل می‌کند. با استفاده از این روش در هر زمان عامل بطور احتمالی بین اکتشاف و استخراج یکی را انتخاب می‌کند. با احتمال $e$ و انتخاب تصادفی از بین تمام اعمال موجود و با احتمال $e-۱$ اکتشاف می‌کند.

مقادیر بالای $e$ عامل را وادار به جستجوی بیشتر و در نتیجه، احتمال اعمال بهینه را کاهش می‌دهد. در حالی که، به عامل توانایی واکنش سریع به تغییراتی را می‌دهد که در محیط ایجاد می‌شود. در مقابل، مقادیر کم $e$ عامل را به سوی اعمال بهینه‌تر سوق می‌دهد.

## اکتشاف بولتزمن

روش دیگر برای انتخاب اعمال، خط‌مشی توزیع بولتزمن است. توزیع بولتزمن، یک خط‌مشی یادگیری است که با گذشت زمان تمایل به اکتشاف را کاهش می‌دهد. و بر این فرض است که با پیشرفت یادگیری، مدل فعلی بهبود می‌یابد. توزیع بولتزمن، احتمالی را با استفاده از یک پارامتر به نام دما به هر عمل اختصاص می‌دهد.

توزیع بولتزمن، با استفاده از معادله زیر یک احتمال مثبت را برای هر عمل ممکن $a \epsilon A$ اختصاص می‌دهد:

$$P(a|s) = \frac{e^{\frac{Q(s,a)}{T}}}{\sum \acute{a} \epsilon A e^{\frac{Q(s,\acute{a})}{T}}}$$

جایی که

$$T_{\text{جدید}} = e^{-dj} * T_{\text{بیشینه}} + ١$$

عمل با $Q(s,a)$ بیشتر با احتمال $P$ بیشتری همراه است. $T$ با افزایش تکرار $j$ با گذشت زمان کاهش پیدا می‌کند. بنابراین، با پیشرفت یادگیری گرایش به اکتشاف در عامل کاهش پیدا می‌کند و در نتیجه، خط‌مشی یادگیری توزیع بولتزمن تمایل به استخراج از اعمال با $Q(s,a)$ بالا را دارد. پارامترهای $T_{\text{بیشینه}}$ و نرخ تنزل $dj$ در ابتدا تنظیم می‌شوند.

## مبتنی‌بر مدل در مقابل بدون مدل

همان‌طور که پیش‌تر بیان شد، هدف اصلی عامل در یادگیری تقویتی، جمع‌آوری بیشترین میزان پاداش در "درازمدت" است. در جهت انجام این کار، عامل باید یک خط‌مشی بهینه برای رفتار در محیط پیدا کند. محیط می‌تواند قطعی یا تصادفی (غیرقطعی) باشد. یعنی اگر عامل در یک وضعیت خاص عملی را انجام دهد، وضعیت بعدی حاصل از محیط ممکن است لزوماً همیشه یکسان نباشد. قطعاً این عدم اطمینان‌ها، مساله را برای یافتن خط‌مشی بهینه دشوارتر می‌کند.

> ### پیش‌بینی در مقابل بازبینی
> پیش‌بینی و بازبینی به دو مساله اساسی اشاره دارد که یک عامل یادگیری تقویتی باید آن را حل کند. پیش‌بینی، مستلزم توانایی محاسبه یا برآورد پیامدهای یک عمل است. مسائل پیش‌بینی معمولا به معنای برآورد مقادیر وضعیت‌ها یا مقادیر عمل جفت‌های وضعیت ـ عمل است. در مقابل، بازبینی نیاز به توانایی تصمیم‌گیری دارد. بدون بازبینی، عامل هرگز عملی را انجام نمی‌دهد.

مساله پیش‌بینی: با توجه به فرآیند تصمیم‌گیری مارکوف $< S, A, P, R, \gamma >$ و خط‌مشی $\pi$، باید تابع مقدار $v(\pi)$ پیدا شود. به عبارت دیگر، هدف این است که بفهمیم یک خط‌مشی چقدر خوب است.

مساله بازبینی: با توجه به فرآیند تصمیم‌گیری مارکوف $< S, A, P, R, \gamma >$، باید تابع مقدار بهینه $v(\pi)$ و خط‌مشی بهینه $\pi*$ پیدا شود. به عبارت دیگر، هدف این است که خط‌مشی پیدا شود که بیشترین پاداش را با بهترین عمل برای انتخاب در اختیار قرار دهد.

همان‌طور که می‌دانیم، در یادگیری تقویتی مساله اغلب از منظر ریاضی به عنوان یک فرآیند تصمیم‌گیری مارکوف صورت می‌گیرد. فرآیند تصمیم‌گیری مارکوف، روشی برای نمایش "پویایی" محیط است؛ یعنی، نحوه واکنش محیط به اقدامات احتمالی عامل در یک وضعیت خاص. به عبارت دقیق‌تر، فرآیند تصمیم‌گیری مارکوف به یک تابع انتقال مجهز شده که تابعی است که با توجه به وضعیت فعلی محیط و علمی که ممکن است عامل انجام دهد، احتمال انتقال به هر یک از خروجی‌ها را ایجاد می‌کند. یک تابع پاداش نیز با فرآیند تصمیم‌گیری مارکوف مرتبط است.

تابع پاداش با توجه به وضعیت فعلی محیط و احتمالا عملی که توسط عامل و وضعیت بعدی محیط انجام می‌شود، پاداش دریافت می‌کند. توابع پاداش و انتقال اغلب الگوی محیط نامیده می‌شوند. با این حال، گاهی اوقات توابع پاداش و انتقال را نداریم. از همین رو، نمی‌توانیم خط‌مشی را تخمین بزنیم، چراکه ناشناخته است. در غیاب این توابع، در جهت تخمین خط‌مشی بهینه، نیاز به تعامل با محیط و مشاهده پاسخ‌های آن است که اغلب به عنوان "مشکل یادگیری تقویتی" نامیده می‌شود. چراکه، عامل باید با تقویت باورهای خود در مورد پویایی محیط، یک خط‌مشی را تخمین بزند.

باگذشت زمان، عامل شروع به درک نحوه واکنش محیط به اقدامات خود می‌کند و می‌تواند خط‌مشی بهینه را تخمین بزند. بنابراین، در مسائل یادگیری تقویتی، عامل خط‌مشی بهینه را برای رفتار در یک محیط ناشناخته با تعامل با آن، با استفاده از روش "آزمون و خطا" تخمین می‌زند. بر این اساس، الگوریتم‌های یادگیری تقویتی را می‌توان به الگوریتم‌های مبتنی‌بر مدل یا بدون مدل است تقسیم‌بندی کرد.

در الگوریتم‌های مبتنی‌بر مدل، عامل به یک مدل کامل از محیط دسترسی دارد، یا سعی می‌کند آن را از طریق تعامل بیاموزد و به‌منظور برآورد درست خط‌مشی بهینه از تابع انتقال و پاداش استفاده می‌کند. به عبارت دیگر، عامل سعی می‌کند مدل احتمالاتی را نمونه‌گیری کرده و بیاموزد و از آن برای تعیین بهترین اعمال استفاده کند. عامل ممکن است فقط به تقریبی از توابع انتقال و پاداش دسترسی که توسط عامل آموخته می‌شود، دسترسی داشته باشد. در حالی که، با محیط تعامل دارد یا می‌تواند به عامل مثلا توسط عامل دیگری داده شود.

به‌طور کلی، در یک الگوریتم مبتنی‌بر مدل، عامل می‌تواند به‌طور بالقوه پویایی محیط را در طول یا بعد از مرحله یادگیری پیش‌بینی کند. چراکه، تخمینی از توابع انتقال و تابع پاداش را دارد و اگر احتمال انتقال با موفقیت آموخته شود، عامل می‌داند که با توجه به وضعیت فعلی و عمل، چقدر احتمال دارد به یک وضعیت خاص وارد شود. با این حال، باید توجه داشت که توابع انتقال و پاداشی که عامل برای بهبود تخمین خط‌مشی بهینه خود استفاده می‌کند، ممکن است فقط تقریبی از توابع "واقعی" باشد. از این‌رو خط‌مشی بهینه به دلیل این تقریب‌ها ممکن است هرگز یافت نشود.

در مقابل الگوریتم‌های مبتنی‌بر مدل، الگوریتم‌های بدون مدل هیچ دانش اولیه‌ای در مورد تابع انتقال ندارد و باید ضمن یادگیری در یافتن مسیرهای کارآمد، آن را بیاموزد. به عبارت دیگر، یک الگوریتم بدون مدل، یا "تابع مقدار" یا "خط‌مشی" را مستقیما از تجربه، یعنی با تعامل بین عامل و محیط تخمین می‌زند، بدون اینکه از توابع انتقال و پاداش استفاده کند.

یکی از راه‌های تمیز دادن بین روش‌های مبتنی‌بر مدل و بدون مدل این است که: آیا عامل می‌تواند پیش از شروع هر اقدامی پس از یادگیری، وضعیت و پاداش بعدی را پیش‌بینی کند. به عبارت دیگر، راه تمایز بین الگوریتم‌های مبتنی‌بر مدل یا بدون مدل این است که الگوریتم ها را بررسی کنید و ببینید آیا از توابع انتقال و پاداش استفاده می‌کنند یا نه. اگر استفاده می‌کنند، یک الگوریتم یادگیری تقویتی مبتنی‌بر مدل است.

هر دو روش دارای نقاط قوت و ضعف هستند. روش‌های بدون مدل تاحدودی تضمین می‌کنند که در نهایت خط‌مشی بهینه را پیدا می‌کنند و زمان محاسبه بسیار کمی در هر آزمایش دارند. با این حال، آن‌ها از داده‌ها در طول آزمایشات بسیار ناکارآمد استفاده می‌کنند و بنابراین اغلب برای دستیابی به عملکرد خوب به تجربه زیادی نیاز دارند. در مقابل، الگوریتم‌های مبتنی‌بر مدل می‌توانند بر این مشکل فائق آیند، اما عامل تنها برای یک مدل خاص یاد می‌گیرد و گاهی اوقات برای برخی از مدل‌های دیگر مناسب نیست. همچنین، برای یادگیری یک مدل دیگر نیاز به زمان دارد.

مزیت مهم در داشتن مدل این است که به عامل اجازه می‌دهد تا از قبل برنامه‌ریزی کند و ببیند برای طیف وسیعی از اعمال مختلفی که می‌تواند در وضعیت فعلی خود انجام دهد چه اتفاقی می‌افتد. پس از آن، هنگام تصمیم‌گیری برای عمل به مقایسه این نتایج بپردازد. این امر می‌تواند در مقایسه با الگوریتم‌هایی که از مدل استفاده نمی‌کنند، منجر به بهبود قابل توجهی در کارایی شود. الگوریتم‌های بدون مدل برای بروزرسانی دانش خود، تنها به "آزمون و خطا" اعتماد می‌کنند. از همین‌رو، به ذخیره تمام ترکیبات وضعیت‌ها و اعمال، نیازی ندارند.

رویکرد بدون مدل را می‌توان به‌صورت در ـ خط‌مشی[1] و خارج‌از ـ خط‌مشی[2] طبقه‌بندی کرد. روش‌های در ـ خط‌مشی، از خط‌مشی فعلی برای ایجاد اقدامات استفاده می‌کنند و از آن برای به‌روزرسانی خط‌مشی فعلی استفاده می‌کنند. در حالی‌که، در روش‌های خارج‌از ـ خط‌مشی، از یک خط‌مشی اکتشافی متفاوت برای ایجاد اقدامات در مقایسه با خط‌مشی که بروز می‌شود، استفاده می‌کنند.

# رویکردهای کلاسیك یادگیری تقویتی

اکنون که با مفاهیم پایه یادگیری تقویتی آشنا شده‌اید، در این بخش قصد داریم تا رویکردهای کلاسیک یادگیری تقویتی در حل مسائل را تشریح کنیم. این رویکردها را می‌توان براساس رویکردهای مبتنی‌بر مدل، بدون مدل و ترکیبی از این دو همانند شکل ۹ ـ ۶ تقسیم‌بندی کرد. در ادامه به بررسی این رویکردها می‌پردازیم.

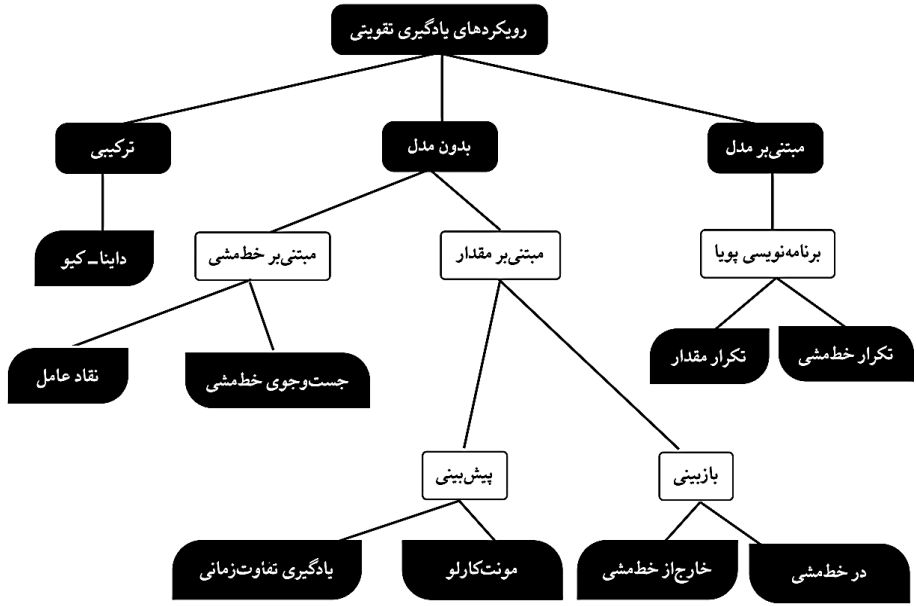

**شکل ۹ ـ ۶.** نحوه تقسیم‌بندی رویکردهای یادگیری تقویتی

## برنامه‌نویسی پویا

اصطلاح برنامه نویسی پویا، به مجموعه‌ای از الگوریتم‌ها گفته می‌شود که می‌تواند برای محاسبه خط‌مشی بهینه با توجه به مدل کاملی از محیط به عنوان فرآیند تصمیم‌گیری مارکوف مورد استفاده

---

[1] on-policy
[2] off-policy

قرار گیرد. ایده اصلی برنامه نویسی پویا و به طور کلی یادگیری تقویتی، استفاده از توابع مقدار برای سازماندهی و ساختار جستجوی خط‌مشی‌های خوب است. الگوریتم های کلاسیک برنامه‌نویسی پویا، به دلیل فرض داشتن یک مدل کامل و همچنین به دلیل هزینه محاساتی زیاد، کاربرد محدودی در یادگیری تقویتی دارند. با این همه، این روش‌ها از لحاظ نظری هنوز هم مهم هستند.

برنامه‌نویسی پویا، شامل دو نسخه متفاوت از چگونگی پیاده‌سازی است: **تکرار خط‌مشی و تکرار مقدار**. در ادامه به‌طور خلاصه این دو رویکرد را شرح خواهیم داد.

## تکرار خط‌مشی

هنگامی که یک خط‌مشی $\pi$ با استفاده از $v_\pi$ بهبود یافته تا یک خط‌مشی بهتر $\pi'$ داشته باشد، می‌توان $v_{\pi'}$ را محاسبه کرده و دوباره آن را بهبود بخشید تا خط‌مشی $\pi''$ حتی بهتر بدست آورد. در نتیجه می‌توان به یک ترتیب به طور یکنواخت خط‌مشی‌ها و توابع مقدار را بدست آورد:

$$\pi_0 \xrightarrow{\mathrm{E}} v_{\pi_0} \xrightarrow{\mathrm{I}} \pi_1 \xrightarrow{\mathrm{E}} v_{\pi_1} \xrightarrow{\mathrm{I}} \pi_2 \xrightarrow{\mathrm{E}} \cdots \xrightarrow{\mathrm{I}} \pi_* \xrightarrow{\mathrm{E}} v_*,$$

تضمین شده است که هر خط‌مشی نسبت به خط‌مشی قبلی اکیدا بهبود می‌یابد؛ مگر اینکه از قبل بهینه باشد. از آنجا که فرآیند تصمیم‌گیری مارکوف محدود، تنها تعداد محدودی از خط‌مشی‌ها را دارد، این فرآیند باید در یک تعداد تکرار محدود، به یک خط‌مشی و تابع مقدار بهینه همگرا باشد.

این روش برای یافتن یک خط‌مشی بهینه، تکرار خط‌مشی نامیده می‌شود. باید توجه داشت که هر ارزیابی خط‌مشی، که خود یک محاسبه تکراری است، با تابع مقدار برای خط‌مشی قبلی شروع می‌شود. این امر، معمولا منجر به افزایش بسیار زیاد سرعت همگرایی ارزیابی سیاست می‌شود؛ احتمالا به این دلیل که تابع مقدار از یک خط‌مشی به خط‌مشی دیگر تغییر کمی می‌کند.

## تکرار مقدار

یکی از اشکالاتی که در روش تکرار خط‌مشی وجود دارد، این است که هر یک از تکرارهای آن شامل ارزیابی خط‌مشی است، که این امر ممکن است خود یک محاسبه تکراری طولانی مدت را به‌همراه داشته باشد که مستلزم جابجایی‌های متعدد در مجموعه وضعیت‌ها است. اگر ارزیابی خط‌مشی به‌صورت تکراری انجام شود، همگرایی دقیقا به $v_\pi$ فقط در محدوده مجاز اتفاق می‌افتد. در نتیجه، سوالی بوجود می‌آید، آیا ما باید منتظر همگرایی دقیق باشیم، یا می‌توانیم از آن دست بکشیم؟

مرحله ارزیابی خط‌مشی، تکرار خط‌مشی می‌تواند به چندین روش بدون از دست دادن تضمین همگرایی تکرار خط‌مشی کوتاه شود. یک مورد خاص مهم هنگامی است که ارزیابی خط‌مشی فقط پس از یک بار حرکت متوقف شود. این الگوریتم تکرار مقدار نامیده شده که می‌تواند به

عنوان یک عملیات پشتیبان‌گیری ساده نوشته شود که ترکیبی از بهبود خط‌مشی و مراحل ارزیابی خط‌مشی است:

$$v_{k+1}(s) = max_a \mathbb{E}[R_{t+1} + \gamma v_k(S_{t+1})|S_t = s, A_t = a]$$

$$= max_a \sum_{\acute{s},r} p(\acute{s}, r|s, a)[r + \gamma v_k(\acute{s})]$$

برای هر $v_{\circ}$ دلخواه، می‌توان نشان داد که توالی $\{v_k\}$ می‌تواند تحت همان شرایطی که وجود $v^*$ را تضمین می‌کند به $v^*$ همگرا می‌شود.

در نهایت، بیایید در نظر بگیریم که تکرار مقدار چگونه به پایان می‌رسد. همانند ارزیابی خط‌مشی، تکرار مقدار رسما به تعداد نامحدودی از تکرارها نیاز دارد تا دقیقا در عمل همگرا شود، ما زمانی متوقف می‌شویم که تابع مقدار تنها با مقدار کمی در یک جابه‌جایی تغییر می‌کند.

### تکرار خط‌مشی تعمیم‌یافته[1]

تکرار خط مشی شامل دو فرآیند همزمان و متقابل است که یکی تابع مقدار را با خط‌مشی فعلی (ارزیابی خط‌مشی) سازگار می‌کند، و دیگری خط‌مشی را با توجه به تابع مقدار فعلی (بهبود سیاست) حریص می‌کند. در تکرار خط مشی، این دو فرآیند متناوب هستند و هر یک قبل از شروع روند دیگر تکمیل می شوند، اما این واقعا لازم نیست. به عنوان مثال، در تکرار مقدار، تنها یک تکرار ارزیابی خط‌مشی بین هر بهبود خط‌مشی انجام می‌شود.

از اصطلاح تکرار خط‌مشی تعمیم‌یافته برای اشاره به ایده کلی اجازه دادن به تعامل فرآیندهای ارزیابی خط‌مشی و بهبود خط‌مشی، مستقل از جزئیات دو فرآیند استفاده می‌شود. طرح کلی برای تکرار خط‌مشی تعمیم‌یافته در شکل ۷ ـ ۹ نشان داده شده است.

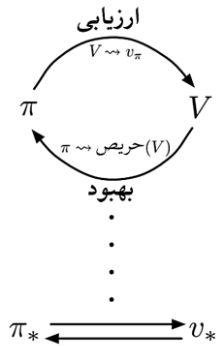

**شکل ۹ ـ ۷.** تکرار خط‌مشی تعمیم‌یافته. توابع مقدار و خط‌مشی تا زمانی که بهینه و در نتیجه با هم سازگار باشند تعامل دارند.

---

[1] generalized policy iteration (GPI)

به راحتی می‌توان دریافت که اگر هم فرآیند ارزیابی و هم فرآیند بهبود پایدار شوند، یعنی دیگر تغییری ایجاد نکنند، آنگاه تابع مقدار و خط‌مشی باید بهینه باشد. تابع مقدار تنها زمانی پایدار می‌شود که با خط‌مشی فعلی سازگار باشد، و خط‌مشی تنها زمانی پایدار می‌شود با توجه به تابع مقدار فعلی، حریص باشد. بنابراین، هر دو فرآیند فقط وقتی ثبات پیدا می‌کنند که خط‌مشی پیدا شود که نسبت به تابع ارزیابی خود حریصانه عمل کند. این نشان می‌دهد که معادله بهینه‌سازی بلمن برقرار است و بنابراین خط‌مشی و تابع مقدار بهینه هستند.

فرایندهای ارزیابی و بهبود در تکرار خط‌مشی تعمیم‌یافته را می‌توان هم به عنوان یک رقابت و هم همکاری در نظر گرفت. در درازمدت، این دو فرآیند برای یافتن یک راه مشترک تعامل دارند: تابع مقدار بهینه و خط‌مشی بهینه.

## مونت کارلو

برخلاف برنامه‌نویسی پویا، روش‌های مونت‌کارلو تنها از تجربه می‌آموزند. از بسیاری جهات می‌توان آن را ساده‌ترین روش یادگیری تقویتی دانست. یک روش مونت‌کارلو مقدار یک وضعیت را با گذراندن چندین مرتبه و میانگین کل پاداش دریافتی پس از تصویب وضعیت، تعیین می‌کند. از آنجا که کل پاداش دریافتی قبل از سپری‌شدن مرحله ناشناخته است، روش‌های مونت‌کارلو فقط برای کارهای مرحله‌ای قابل اجرا هستند. بروزرسانی تنها پس از پایان حرکت در فضای وضعیت که بروزرسانی خارج‌از خط نامیده می‌شود (بروزرسانی در حالی‌که هنوز در فضای وضعیت حرکت می‌کنید، بروزرسانی برخط نامیده می‌شود)، صورت می‌گیرد. علاوه براین، در روش مونت‌کارلو، مقادیر براساس تجربه واقعی است نه بر اساس مقادیر وضعیت‌های جانشین.

روش‌های مونت کارلو بر مبنای ایده تکرار خط‌مشی تعمیم‌یافته کار می‌کنند. همان‌طور که پیش‌تر بیان شد، تکرار خط‌مشی تعمیم‌یافته یک طرح تکراری است و از دو مرحله تشکیل شده است. در اولین گام، سعی می‌شود تقریب تابع مقدار را براساس خط‌مشی فعلی ایجاد کند که به عنوان مرحله ارزیابی خط‌مشی شناخته می‌شود. در گام دوم، خط‌مشی با توجه به تابع مقدار فعلی که به عنوان مرحله بهبود خط‌مشی شناخته می‌شود، بهبود می‌یابد. در روش‌های مونت‌کارلو، برای تخمین تابع مقدار، اجرای برنامه‌ها با اجرای خط‌مشی فعلی برروی سیستم انجام می‌شود. پاداش تجمعی در کل مرحله و توزیع حالاتی که با آن روبرو می‌شوند برای تشکیل تخمینی از تابع مقدار استفاده می‌شود. سپس، خط‌مشی فعلی به‌صورت حریصانه، با توجه به تابع مقدار فعلی تخمین‌زده می‌شود. با استفاده از این دو مرحله به صورت تکراری، می‌توان نشان داد که الگوریتم به تابع مقدار و خط‌مشی بهینه همگرا می‌شود. اگرچه اجرای روش‌های مونت‌کارلو ساده است، اما برای همگرایی آن‌ها به تعداد تکرار زیادی نیاز دارد و از واریانس بالا در تخمین تابع مقدار رنج می‌برد.

## یادگیری تفاوت‌زمانی

یادگیری تفاوت‌زمانی، ایده‌های برنامه‌نویسی پویا و مونت‌کارلو را در خود جای می‌دهد. رویکرد تفاوت‌زمانی با مقایسه تخمین‌ها در دو نقطه از زمان، مقدار جفت وضعیت ـ عمل را تقریب می‌زند، از همین‌رو نام تفاوت‌زمانی را به خود گرفته است. همانند برنامه‌نویسی پویا، الگوریتم‌های یادگیری تفاوت‌زمانی، برآورد مقادیر را براساس سایر تخمین‌ها یاد می‌گیرد (که به آن بوت‌استرپینگ[1] گویند). همچنین، یادگیری تفاوت زمانی همانند روش مونت‌کارلو، می‌تواند بدون اطلاع قبلی از محیط و به‌طور مستقیم از تجربیات بیاموزد. این بدان معناست که یادگیری تفاوت‌زمانی یک رویکرد یادگیری بدون مدل (یا می‌توان آن را معادل با یادگیری بدون‌نظارت دانست) است.

ایده اصلی در رویکرد یادگیری تفاوت‌زمانی، یادگیری براساس تفاوت بین پیش‌بینی‌های پی‌درپی زمانی است و برای بروزرسانی نیازی به صبرکردن تا پایان مسیر نیست. به عبارت دیگر، هدف از یادگیری این است که پیش‌بینی کنونی یادگیرنده برای الگوی فعلی ورودی، بیشتر با پیش‌بینی بعدی در مرحله بعدی مطابقت داشته باشد. بینش این روش بر این است که، پس از مشاهده برخی از پاداش‌هایی که عامل پس از بازدید یک وضعیت و انجام یک عمل معین بدست آورده، می‌توان تخمین بهتری برای مقدار یک جفت وضعیت ـ عمل ارائه کرد.

روش‌های مونت‌کارلو و برنامه‌نویسی پویا دارای نقاط ضعف جدی هستند؛ مونت‌کارلو قادر به حل مسائل پیوسته نیست و برنامه‌نویسی پویا به یک مدل از محیط نیاز دارد. یادگیری تفاوت‌زمانی، از نسخه پشتیبان‌گیری تهیه نمونه در روش مونت‌کارلو و بوت‌استرپینگ از روش برنامه‌نویسی پویا همزمان استفاده می‌کند. از همین‌رو، تنها قسمت‌های خوب روش‌های قبلی را با یک‌دیگر ترکیب می‌کند.

روش مونت‌کارلو، از پاداش کل برای بروزرسانی تابع مقدار استفاده می‌کند. در حالی‌که، روش تفاوت‌زمانی، به‌جای بروزرسانی مقادیر در پایان مرحله (با استفاده از مجموع پاداش)، مقادیر را به‌صورت برخط در هر مرحله بروز می‌کند.

---

[1] bootstrapping

## مونت‌ـ‌کارلو

$$V(S_t) \leftarrow V(S_t) + \alpha\,(G_t - V(S_t))$$

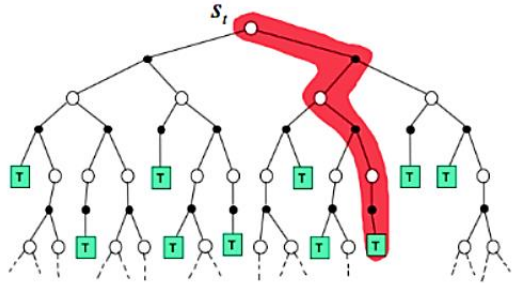

## تفاوت‌ـ‌زمانی

$$V(S_t) \leftarrow V(S_t) + \alpha\,(R_{t+1} + \gamma V(S_{t+1}) - V(S_t))$$

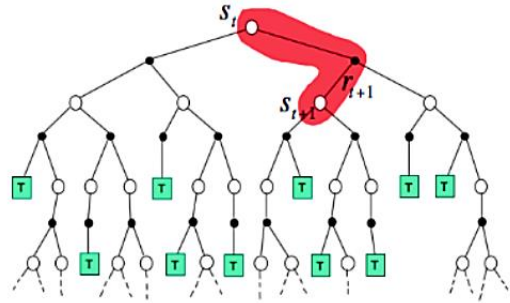

## برنامه‌نویسی پویا

$$V(S_t) \leftarrow \mathbb{E}_\pi\,[R_{t+1} + \gamma V(S_{t+1})]$$

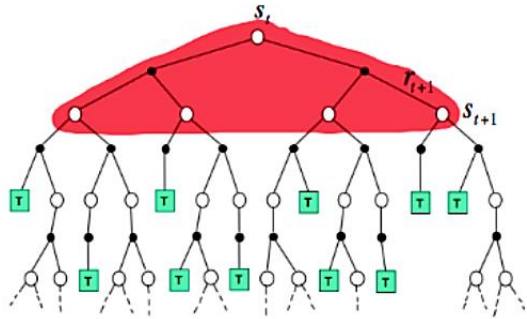

**شکل ۹ـ۷** مقایسه نمودارهای پشتیبان‌گیری مونت‌کارلو ، یادگیری تفاوت‌زمانی و برنامه‌نویسی پویا برای توابع وضعیت مقدار

برای مقایسه سه رویکرد، مونت‌کارلو، برنامه‌نویسی پویا و یادگیری تفاوت‌زمانی می‌توان از نمودار پشتیبان‌گیری استفاده کرد. مقایسه بین این سه رویکرد براساس نمودار پشتیبان‌گیری وضعیت‌ـ مقدار در شکل ۴ ـ ۵ قابل مشاهده است.

یادگیری تفاوت زمانی اغلب به یک مساله پیش‌بینی با یک قانون بروزرسانی برای تابع مقدار داده شده، اشاره دارد:

$$V(s) \leftarrow V(s) + \alpha\big(r + \gamma V(\acute{s}) - V(s)\big)$$

جایی که $\alpha$ نرخ یادگیری و $\gamma$ فاکتور تنزیل است. قسمت داخل پرانتز به عنوان خطای تفاوت‌زمانی شناخته می‌شود:

$$\delta_t = r + \gamma V(\acute{s}) - V(s)$$

روش یادگیری تفاوت‌زمانی برای پیش‌بینی در دو روش مختلف برای انجام بازبینی استفاده می‌شود. تفاوت اصلی این دو روش این است که یکی در خطمشی و دیگری خارج از خطمشی است. این دو الگوریتم که به‌طور گسترده در یادگیری تقویتی استفاده می‌شود، سارسا و کیوـ یادگیری می‌باشد. در ادامه به بررسی این دو نوع از الگوریتم تفاوت‌زمانی می‌پردازیم.

## سارسا (در خطمشی)

سارسا یک روش یادگیری تقویتی تفاوت‌زمانی است که در تلاش برای تخمین یک تابع مقدار عمل را به‌جای یک تابع مقدار یاد بگیرد. این وراثت نام از: ("وضعیت"، "عمل"، "پاداش"، "وضعیت بعدی"، "عمل بعدی") گرفته شده است. این رویکرد، در خطمشی است، چرا که $Q_\pi(s, a)$ را برای $\pi$ برای خطمشی فعلی تخمین می‌زند. قانون بروزرسانی وضعیت‌ـ مقدار به‌صورت زیر می‌باشد:

$$Q(s, a) \leftarrow Q(s, a) + \alpha\big(r + \gamma Q(\acute{s}, \acute{a}) - Q(s, a)\big)$$

می‌توان نشان داد که سارسا هنگامی که همه جفت‌های وضعیت‌ـ عمل به تعداد نامحدودی بازدید شود، به یک خطمشی بهینه همگرا می‌شود.

## کیو-یادگیری (خارج از خطمشی)

یادگیری تفاوت‌زمانی خارج از خطمشی کیوـ یادگیری (Q-learning) نامیده می‌شود که یکی از اساسی‌ترین و محبوب‌ترین روش‌ها برای تخمین توابع مقدار Q به روشی بدون مدل است. قانون بروزرسانی این روش به صورت زیر است:

$$Q(s, a) \leftarrow Q(s, a) + \alpha\big(r + \gamma max_{\acute{a}} Q(\acute{s}, \acute{a}) - Q(s, a)\big)$$

کیوـ یادگیری در تلاش است با توجه به شرایط فعلی، بهترین اقدامات را انجام دهد. این الگوریتم خارج از خطمشی در نظر گرفته می‌شود. چراکه، تابع کیوـ یادگیری از اقداماتی خارج از خطمشی فعلی یاد می‌گیرد. به‌طور کلی می‌توان گفت که، کیوـ یادگیری به‌دنبال یادگیری خطمشی

است تا مجموع پاداش را بیشینه کند. امروزه، بسیاری از الگوریتم‌های یادگیری تقویی عمیق مبتنی‌بر کیو_یادگیری هستند.

کیو_یادگیری بی‌تفاوت به اکتشاف است. این بدان معناست که، بدون توجه به خط‌مشی اکتشافی که دنبال می‌شود، به خط‌مشی مطلوب همگرا می‌شود؛ با این فرض که هر جفت اقدام وضعیت به تعداد نامحدودی بازدید شده و پارامتر یادگیری $\alpha$ به‌طور مناسب کاهش می‌یابد.

## جست‌وجوی خط‌مشی

روش‌های جست‌وجوی خط‌مشی نیازی به حفظ یک مدل تابع مقدار ندارند، بلکه به‌طور مستقیم یک خط‌مشی بهینه $\pi^*$ را جست‌وجو می‌کنند. از بین الگوریتم‌های جست‌وجوی خط‌مشی در یادگیری تقویتی، الگوریتم گرادیان خط‌مشی محبوب‌ترین هستند.

هدف از یادگیری تقویتی، یافتن یک استراتژی رفتار بهینه برای عامل برای به‌دست آوردن پاداش‌های بهینه است. روش‌های گرادیان خط‌مشی به طور مستقیم به مدل‌سازی و بهینه‌سازی خط‌مشی می‌پردازند. این خط مشی با انجام اقدامات ارائه شده از خط‌مشی فعلی و محاسبه پاداش ارزیابی می‌شود. سپس پارامترهای خط‌مشی در جهت افزایش بازده مورد انتظار با استفاده از گرادیان نزولی بروز می‌شوند. قانون بروزرسانی برای پارامترهای خط‌مشی را می توان با توجه به بازده مورد انتظار $J$ به‌صورت زیر نوشت:

$$\theta_{i+1}^\pi = \theta_i^\pi + \alpha \nabla_{\theta^\pi} j \ , j = \mathbb{E}_\pi(\sum_{k=0}^\infty \gamma^k r_k)$$

جست‌وجوی خط‌مشی، همگرایی بهتری دارد و می‌تواند خط‌مشی‌های تصادفی را بیاموزد که با روش‌های مبتنی بر مقدار امکان‌پذیر نیست. اشکال عمده الگوریتم‌های خط‌مشی، مرحله ارزیابی خط‌مشی آن‌ها است که از واریانس زیادی رنج می‌برد و بنابراین می‌تواند در یادگیری خط‌مشی‌های خوب، کند باشد.

## نقاد عامل

روش‌های گرادیان خط‌مشی محض، به دلیل تخمین‌های با واریانس بالا، به آهستگی یاد می‌گیرند و برای اجرای مسائل برخط نامناسب هستند. با این حال، می‌توان از روش‌های تفاوت‌زمانی برای مقابله با این مشکلات استفاده کرد.

رویکردهای تابع مقدار از لحاظ نظری، به کل پوشش فضای وضعیت و مقادیر تقویت‌شده مربوطه از همه اعمال ممکن در هر وضعیت نیاز دارند. بنابراین، هنگام کار با برنامه‌های کاربردی با ابعاد بالا، پیچیدگی محاسباتی زیادی داشته و همچنین، یک تغییر کوچک در مقادیر تقویت‌کننده محلی ممکن است باعث تغییر زیادی در خط‌مشی شود.

در مقابل روش‌های تابع مقدار، روش‌های جست‌وجوی خطمشی، خطمشی فعلی و خطمشی بعدی را به خطمشی فعلی در نظر می‌گیرند و سپس تغییرات پارامترهای خطمشی را محاسبه می‌کنند که در نتیجه، پیچیدگی محاسباتی بسیار کمتری در مقابل روش‌های تابع مقدار است. با این حال، رویکردهای جستجوی خطمشی ممکن است سبب بهینه‌سازی محلی شود و نتواند به بهینه سراسری برسد.

روش‌های نقاد عامل با ترکیب این دو رویکرد به‌صورت همزمان یک خطمشی و یک تابع وضعیت ـ مقدار را یاد می‌گیرند. این رویکرد که از تلفیق تابع مقدار و رویکرد جست‌وجوی خطمشی برای بهره‌وری از مزیت هر دو روش ارائه شده است، با ساختار نقاد عامل شناخته می‌شود. می‌توان گفت، روش نقاد عامل، روش‌های یادگیری تفاوت‌زمانی است که به‌طور صریح خطمشی را ذخیره می‌کند. همان‌طور که در شکل ۹ـ۸ نشان داده شده است، انتخاب اقدام توسط عامل کنترل و از نقاد برای انتقال مقادیر به عامل استفاده می‌شود. بنابراین، تصمیم‌گیری در مورد به روزرسانی خطمشی براساس این نقاد می‌باشد.

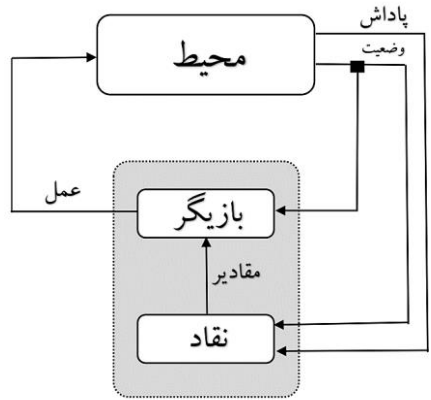

**شکل ۹ـ۸.** ساختار نقاد عامل

## رویکرد ترکیبی (داینا-کیو)

داینا ـ کیو، یکی از معماری‌هایی است که توانایی تلفیق قابلیت یادگیری در روش بدون مدل و قابلیت برنامه‌ریزی در روش‌های مبتنی‌بر را دارد. الگوریتم با دیدن این که کدام وضعیت‌ها بازدید می‌شود و چه پاداشی در یک وضعیت خاص می‌گیرد، یک مدل را یاد گرفته و از اطلاعات برای بروزرسانی احتمالات انتقال و تابع پاداش استفاده می‌کند.

# یادگیری تقلیدی[1]

در حالی که یادگیری تقویتی، طی سال‌های اخیر برخی از متعارف‌ترین پیشرفت‌ها را در هوش مصنوعی انجام داده است، موفقیت آن تا حد زیادی به حوزه‌های شبیه‌سازی‌شده محدود شده است که در آن سیگنال پاداش به خوبی تعریف شده و داده می‌شود. اگرچه ممکن است به‌طور شهودی مشخص باشد که تابع پاداش در حوزه‌های شبیه‌سازی‌شده (مانند امتیاز در بازی‌های آتاری) چیست، در بسیاری از حوزه‌های دنیای واقعی، فرض دسترسی به تابع پاداش بسیار سخت است. وظیفه یادگیری یک خط‌مشی رانندگی خوب را در نظر بگیرید. در حالی که یک راننده انسانی خوب می‌تواند با خیال راحت در جاده‌ها رانندگی کند، ممکن است نتواند از نظر ریاضی تابع پاداشی را فرموله کند که به‌طور دقیق سیاست‌های رانندگی خوب و بد را از هم جدا کند. بدون عملکرد پاداش خوب، یادگیری تقویتی برای حل مساله رانندگی خودران مناسب نیست.

علی‌رغم دشواری تعریف تابع پاداش، یک خط‌مشی خوب را می‌توان با تقلید مستقیم از مسیرهای ارائه شده توسط یک کارشناس آموخت که ممکن است به تابع پاداش واقعی دسترسی داشته باشد یا نداشته باشد؛ با این حال، می‌داند که چگونه در محیط بهینه رفتار کند. در حالت ایده‌آل، خط‌مشی صرفا مسیرهای خبره را به خاطر نمی‌سپارد، بلکه رفتار کارشناس را در حالت‌های غیرقابل مشاهده در مسیرهای کارشناس نیز تعمیم می‌دهد. این رویکرد برای یادگیری یک خط‌مشی از طریق تقلید به عنوان **یادگیری تقلیدی** شناخته می‌شود.

هدف تکنیک‌های یادگیری تقلیدی، تقلید رفتار انسان در یک وظیفهٔ مشخص است. یک عامل برای انجام یک وظیفه، از *نمایش‌ها*[2] با یادگیری نگاشت بین مشاهدات و اعمال آموزش می‌بیند. ایده یادگیری تقلیدی سال‌هاست که وجود داشته است. با این حال، این زمینه اخیرا به دلیل پیشرفت در محاسبات و همچنین افزایش تقاضا برای کاربردهای هوشمند مورد توجه قرار گرفته است. پارادایم یادگیری از طریق تقلید در حال محبوبیت است، چراکه آموزش کارهای پیچیده را با حداقل دانش تخصصی از وظایف تسهیل می‌کند. روش‌های یادگیری تقلیدی به‌طور بالقوه می‌توانند مسالهٔ آموزش یک کار را به مسالهٔ ارائه نمایش، بدون نیاز به برنامه‌نویسی صریح یا طراحی توابع پاداش خاص برای آن کار کاهش دهند.

---

## تفاوت یادگیری تقلیدی با یادگیری تقویتی

در یادگیری تقلیدی، یادگیرنده ابتدا اعمال یک کارشناس (اغلب انسانی) را در مرحله‌ی آموزش مشاهده می‌کند. سپس یادگیرنده از این مجموعه‌ی آموزشی برای یادگیریِ خط‌مشی استفاده می‌کند که سعی می‌کند اقدامات نشان داده‌شده توسط کارشناس را تقلید کند تا به بهترین عملکرد برسد. در طرف مقابل، در یادگیری تقویتی چنین کارشناسی وجود ندارد. در عوض، عامل دارای عملکرد "پاداش" است و از استراتژی‌هایی برای کاوش موثر وضعیت و فضای عمل استفاده می‌کند و به تنهایی (با آزمون و خطا) یک خط‌مشی بهینه ارائه می‌دهد. بیایید عامل خودران را در نظر بگیریم. فضای حالت می‌تواند همان چیزی باشد که عامل در هر لحظه می‌بیند: جاده، علائم، ماشین‌های دیگر و عابران پیاده. اقدامات می‌تواند فرمان اتومبیل، شتاب و توقف باشد.

یک عامل یادگیری تقلیدی یک راننده‌ی انسانی (کارشناس) را مشاهده می‌کند و اقدامات او را در وضعیت‌های مختلف ثبت می‌کند. از این‌رو، خط‌مشی را ایجاد می‌کند که در هر وضعیت، بر اساس آن‌چه کارشناس انجام داده است، چه اقداماتی باید انجام شود. در زمان اجرا، تمام تلاش خود را می‌کند تا اقدام درست را بر اساس خط‌مشی تقریب بزند، زیرا وضعیت‌ها دقیقا شبیه هم نیستند و یک عنصر احتمال نیز وارد خواهد شد.

در مقابل، یک عامل یادگیری تقویتی، یک تابع پاداش دریافت می‌کند. برای مثال، هر ثانیه‌ای که می‌گذرد به عنوان یک امتیاز مثبت محاسبه می‌شود و اگر تصادف کند یا با عابر پیاده یا ماشین دیگری برخورد کند، وظیفه با پاداش صفر به پایان می‌رسد. سپس عامل با اقدامات تصادفی شروع می‌کند و از طریق آزمون و خطا یاد می‌گیرد که کدام اقدامات پاداش را بیشینه می‌کند و کدام اقدامات به صفر می‌رسد. پس از مدتی، زمانی که خط‌مشی به اندازه کافی خوب باشد (بر اساس معیارهای عملکرد)، عامل از خط‌مشی برای رانندگی استفاده می‌کند. بنابراین هر دو روش منجر به یک استراتژی زمان اجرا می‌شوند. فقط نحوه یادگیری خط‌مشی متفاوت است.

## یادگیری تقلیدی مستقیم (شبیه‌سازی رفتار)

فرآیند تصمیم‌گیری چند مرحله‌ای در یادگیری تقویتی با فضای جستجوی عظیمی روبرو است و از این‌رو استفاده از پاداش‌های تجمعی مناسبی برای یادگیریِ تصمیمات از چندین مرحله قبل گرفته شده است، آسان نیست. با این حال، می‌توان چنین مشکلی را با تقلید مستقیم از جفت‌های وضعیت ـ عمل ارائه‌ شده توسط کارشناسان انسانی به خوبی کاهش داد، که **یادگیری تقلیدی مستقیم** (معروف به **شبیه‌سازی رفتار**[1]) نامیده می‌شود.

فرض کنید کارشناسان انسانی مجموعه‌ای از مسیرها را به ما ارائه کرده‌اند $\{\tau_1, \ldots, \tau_m\}$، که در آن هر مسیر شامل دنباله‌ای از وضعیت‌ها و اقدامات است:

---

$$\tau_i = <s_1^i, a_1^i, s_2^i, a_2^i, \ldots, s_{n_i+1}^i>$$

که در آن $n_i$ تعداد انتقال در مسیر $i$ام است. چنین اطلاعاتی به عامل می‌گوید که در هر وضعیت چه کاری انجام دهد. بنابراین، ما می‌توانیم از یادگیری بانظارت برای یادگیری خط‌مشی استفاده کنیم که با داده‌های مسیر بدست آمده از کارشناسانِ انسانی مطابقت دارد.

می‌توانیم جفت‌های وضعیت ـ عمل را از همه‌ی مسیرها استخراج کنیم تا یک مجموعه داده جدید بسازیم:

$$D = \{(s_1, a_1), (s_2, a_2), \ldots, (s_{\sum_{i=1}^{m} n_i}, a_{\sum_{i=1}^{m} n_i})\}$$

که وضعیت‌ها را به عنوان ویژگی‌ها و اقدامات را به عنوان برچسب‌ها حاشیه‌نویسی می‌کند.

سپس، از مجموعه داده‌ی $D$، می‌توانیم یک مدل خط‌مشی را با استفاده از الگوریتم‌های دسته‌بندی برای اقدامات گسسته یا الگوریتم‌های رگرسیون برای اقدامات پیوسته بیاموزیم.

پس از آن، شبیه‌سازی رفتار (BC) با کمینه کردن زیان آموزشی بانظارت، خط‌مشی را حل می‌کند:

$$min_\pi J_{BC}(\pi) = -\frac{1}{N} \sum_{k=1}^{N} \log \pi(s_k, a_k)$$

## یادگیری تقویتی معکوس[۱]

در بسیاری از کاربردها، طراحی تابع پاداش اغلب چالش‌برانگیز است، اما ممکن است بتوانیم تابع پاداش را به‌طور معکوس از مثال‌های ارائه شده توسط کارشناسِن انسانی استخراج کنیم. این رویکرد **یادگیری تقویتی معکوس** نامیده می‌شود.

در یادگیری تقویتی معکوس، فضای وضعیت X و فضای عمل A مشخص است و هم‌چنین مجموعه داده‌ی مسیر $\{\tau_1, \ldots, \tau_m\}$، درست مانند آنچه در یادگیری تقلیدی مستقیم داریم. ایده‌ی اصلی یادگیری تقویتی معکوس به شرح زیر است:

*اجازه دادن به عامل برای انجام اقداماتی که با مثال‌های ارائه‌شده مطابقت دارد، معادل یافتن خط‌مشی بهینه در محیطِ تابع پاداش است، جایی که خط‌مشی بهینه همان مسیرهایی را ایجاد می‌کند که به عنوان مثال‌ها ارائه شده است. به عبارت دیگر، ما به دنبال تابع پاداشی می‌گردیم که نمونه‌های ارائه شده بهینه باشند و سپس از این تابع پاداش برای آموزش خط‌مشی یادگیری تقویتی استفاده می‌کنیم.*

فرض کنید که تابع پاداش را می‌توان به عنوان یک تابع خطی از وضعیت‌ها نشان داد، یعنی $R(x) = w^T x$. سپس، پاداش‌های تجمعی خط‌مشی $\pi$ را می‌توان به صورت زیر نوشت:

---
[۱] Inverse Reinforcement Learning

$$\rho^\pi = \mathbb{E}\left[\sum_{t=0}^{+\infty} \gamma^t R(x_t)\,|\pi\right] = \mathbb{E}\left[\sum_{t=0}^{+\infty} \gamma^t w^T x_t\,|\pi\right]$$

$$= w^T \mathbb{E}\left[\sum_{t=0}^{+\infty} \gamma^t x_t\,|\pi\right].$$

که حاصل ضرب داخلی ضرایب $w$ و امید ریاضی جمع‌وزنی بردارهای وضعیت است.

اجازه دهید $\bar{x}^\pi$ نشانگر امید ریاضی بردارهای وضعیت $\mathbb{E}[\sum_{t=0}^{+\infty} \gamma^t x_t\,|\pi]$ باشد. امید ریاضی $\bar{x}^\pi$ را می‌توان با استفاده از روش مونت کارلو تقریب زد:

مسیرهای مثال را می‌توان به عنوان نمونه‌ای از خط‌مشی بهینه مشاهده کرد، از این رو می‌توانیم مجموع وزنی وضعیت‌ها را در هر مسیر مثال، محاسبه کنیم و سپس میانگین بگیریم که با $\bar{x}^*$ نشان داده شده است. سپس، برای تابع پاداش بهینه $R(x) = w^{*T}x$ امید ریاضی $\bar{x}^\pi$ تولید شده توسط هر خط‌مشی دیگر، داریم:

$$w^{*T}\bar{x}^* - w^{*T}\bar{x}^\pi = w^{*T}(\bar{x}^* - \bar{x}^\pi) \geq \cdot.$$

اگر بتوانیم $(\bar{x}^* - \bar{x}^\pi)$ را برای هر خط‌مشی محاسبه کنیم، آنگاه می‌توانیم

$$w^* = argmax_w\, min_\pi w^T(\bar{x}^* - \bar{x}^\pi)$$
$$s.t. \|w\| \leq \cdot.$$

را حل کنیم.

از آنجایی که بدست آوردن همه‌ی خط‌مشی‌ها دشوار است، یک رویکرد جایگزین، شروع با یک خط‌مشی تصادفی و استفاده از آن برای یافتن تابع پاداش بهتر به صورت تکراری است. سپس از تابع پاداش برای یافتن خط‌مشی بهتر استفاده می‌شود. این روند تا زمانی ادامه می‌یابد که تابع پاداش و خط‌مشی متناسب با مسیرهای مثال (نمونه) را داشته باشیم.

به‌طور خلاصه، در یادگیری تقویتی معکوس، ما با مجموعه‌ای از نمایش‌های کارشناس شروع می‌کنیم (فرض می‌کنیم که این‌ها بهینه هستند)، سپس سعی می‌کنیم تابعِ پاداشِ پارامتری را تخمین بزنیم، که باعث رفتار (خط‌مشی) کارشناس می‌شود.

## خلاصه فصل

- اصل اساسیِ پشت مدل‌های گروهی، این است که گروهی از یادگیرنده‌های ضعیف گرد هم آیند تا به کمک یکدیگر یک یادگیرنده قوی تشکیل دهند.

- هیچ مدل یادگیری ماشین به تنهایی، در همه‌ی انواع داده‌ها و مجموعه داده‌ها بهترین عملکرد را نخواهد داشت.

- در نظریه‌ی یادگیری گروهی، یادگیرندگانِ ضعیف (یا مدل‌های پایه) را می‌نامیم که می‌توانند با ترکیب چندین مدل از آن‌ها به عنوان بلوک‌هایی برای طراحی مدل‌های پیچیده‌تر استفاده شوند.

- تجمیع‌پردازی یک روش ساده و در عین حال موثر برای ایجاد مجموعه‌ای از دسته‌بندها است.

- ایده‌ی پشت تجمیع‌پردازی، ترکیبِ نتایجِ چندین مدل برای بدست آوردن یک نتیجه کلی است.

- تجمیع‌پردازی بر اساس نمونه‌برداری بوت‌استرپ کار می‌کند.

- در توان‌افزایی، هر دسته‌بند تحت تاثیر عملکرد دسته‌بندی‌هایی است که قبل از ساخت آن، ساخته شده‌اند.

- یادگیری همیشگی، یک الگوی یادگیری ماشین پیشرفته است که به طور مداوم یاد می‌گیرد، دانش‌آموخته شده در کارهای قبلی را جمع‌آوری می‌کند و از آن برای کمک به یادگیری آینده استفاده می‌کند.

- شبکه‌های عصبی عمیق اغلب با فرضیات جهان بسته آموزش داده می‌شوند، یعنی فرض می‌شود که توزیع داده‌های آزمایشی مشابه توزیع داده‌های آموزشی است.

- هنگامی‌که شبکه‌های عصبی عمیق داده‌هایی را پردازش می‌کنند که شبیه توزیع مشاهده شده در زمان آموزش نیستند، اغلب پیش‌بینی‌های اشتباهی انجام می‌دهند و این کار را با اطمینان بیش از حد انجام می‌دهند.

- یک سیستم هوش مصنوعی ایده‌آل باید در صورت امکان به نمونه‌های خارج از توزیع تعمیم پیدا کند.

- تشخیص خارج از توزیع برای تضمین قابلیت اطمینان و ایمنی سیستم‌های یادگیری ماشین ضروری است.

- یادگیری تقویتی از طریق عامل در تلاش است با آزمون و خطا مساله را از طریق تعامل با محیطی که برای عامل ناشناخته می‌باشد، حل کند.

- فرآیندهای تصمیم‌گیری مارکوف، یک مدل ریاضی تصادفی برای یک سناریوی تصمیم‌گیری است.

## مراجع برای مطالعه بیشتر

# Machine Learning
## and Data Science

---

Foundations, concepts, algorithms, and tools

---

## Milad Vazan

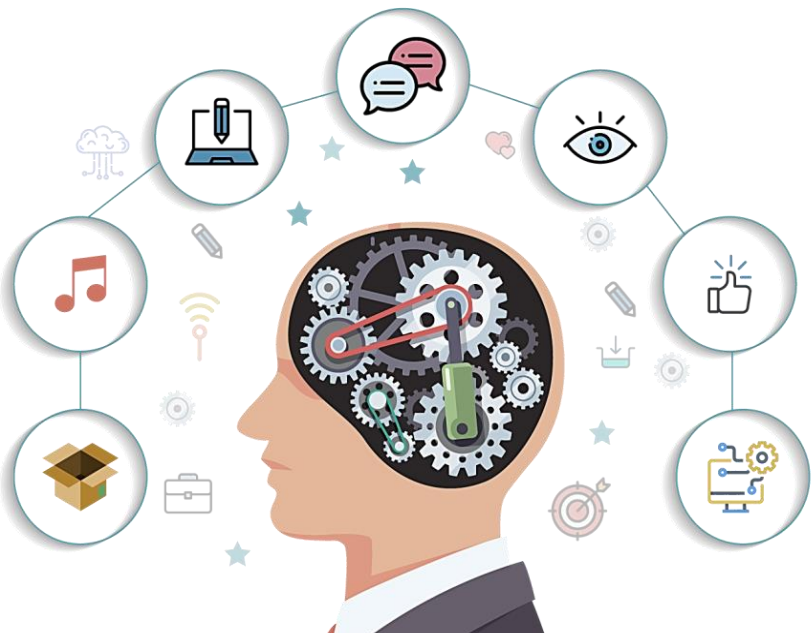

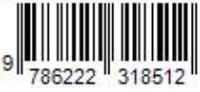

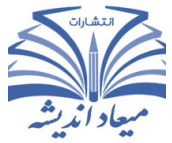